%% file: arxiv.tex
\documentclass{article}

\usepackage[preprint]{neurips_2026}

\usepackage[utf8]{inputenc} 
\usepackage[T1]{fontenc}    
\usepackage{hyperref}       
\usepackage{url}            
\usepackage{booktabs}       
\usepackage{amsfonts}       
\usepackage{nicefrac}       
\usepackage{microtype}      
\usepackage{xcolor}         

\usepackage{microtype}
\usepackage{graphicx}
\usepackage{subcaption}
\usepackage{booktabs} 

\usepackage[x11names]{xcolor}

\usepackage{amsmath}
\usepackage{amssymb}
\usepackage{mathtools}
\usepackage{amsthm}
\usepackage{multirow}

\usepackage[capitalize,noabbrev]{cleveref}

\usepackage{siunitx}
\usepackage{ifthen}

\theoremstyle{plain}

\theoremstyle{definition}

\theoremstyle{remark}

\usepackage[textsize=tiny]{todonotes}

\title{Slicing and Dicing: Configuring Optimal Mixtures of Experts}

\author{
  Margaret Li \\
   Paul G Allen School of Computer Science\\
    University of Washington \\
    Seattle, USA \\
  \texttt{margsli@cs.washington.edu} 
  \And 
  Sneha Kudugunta \\
   Paul G Allen School of Computer Science\\
    University of Washington \\
    Seattle, USA \\
  \And
  Danielle Rothermel \\
  New York University \\
  Courant School of Data Science \\
  New York, USA \\
  \And 
  Luke Zettlemoyer \\
   Paul G Allen School of Computer Science\\
    University of Washington \\
    Seattle, USA \\
}

\begin{document}

\maketitle

\begin{abstract}

Mixture-of-Experts (MoE) architectures have become standard in large language models, yet many of their core design choices — expert count, granularity, shared experts, load balancing, token dropping — have only been studied one or two at a time over narrow configuration ranges. 
It remains an open question whether these choices can be optimized independently, without considering interactions. We present the first systematic study of over 2,000 pretraining runs spanning models up to 6.6B total parameters, in which we exhaustively vary total experts, expert dimension, heterogeneous expert sizing within a single layer, shared expert size and load-balancing mechanisms. 
We find that at every active-parameter scale that we study, performance consistently improves with total MoE parameters even at extreme active expert parameter ratios like 128.
Further, the optimal expert size is nearly invariant to total parameter count and depends only on active parameter count. 
Third, we see that other choices like shared experts, heterogeneous experts and load-balancing settings have small effects relative to expert count and granularity, although dropless routing yields a consistent gain. 
Overall, our results suggest a simpler recipe: focus on expert count and granularity, other choices have minimal effect on final quality.

\end{abstract}

\section{Introduction}

Mixture of Experts (MoE) \citep{shazeer2017outrageously} models decouple computational overhead from model capacity by adding \emph{experts} which are conditionally activated by a learned router. 
Recent MoE LLMs demonstrate improved efficiency when training at massive scales \citep{switch,deepseekv3,du2022glam}.
Yet, there is a lack of analysis to understand the effect of the central design choices controlling this sparse activation; we do not know exactly what makes existing recipes work so well.

Several fundamental MoE design choices introduce substantial complexity beyond dense LLMs \citep{raffel2019exploringtl,brown2020language}. To date, these innovations have been evaluated in isolation or in limited combinations. First, the granularity and topology of experts — including total expert count \citep{clark2022unified}, the use of \emph{shared} FFN components alongside routed ones \citep{rajbhandari2022deepspeed}, and the implementation of fine-grained experts \citep{deepseek} have significant implications for both model quality and throughput. 
Second, the choice of routing algorithm \citep{clark2022unified,zoph2022stmoe,lewis2021base,zhou2022mixture,deepseek} greatly affects model quality. We present the first systematic study of all of these factors. Our main contributions are:

\paragraph{Disentangling expert count from granularity:} Unlike previous studies, we vary both expert granularity and total expert count \emph{separately} to measure the independent impact of expert size and expert count under a fixed compute budget.  We conduct over 2000 experiments ranging from 10 million to 6.6 billion total parameters, and find that performance improves monotonically with total MoE parameters even at an active expert parameter ratio of 128.
\paragraph{Expert flexibility offers little:} We test heterogeneous expert pools (mixed granularities within a layer) and shared ``generalist'' experts across many configurations. Neither improves over a well-configured homogeneous MoE, while generalists consistently hurt.
\paragraph{Simplifying the routing design space:} Across multiple expert configurations and scales, we see that MoE quality is robust to load-balancing hyperparameters within reasonable ranges, while dropless routing yields a small but consistent gain. 

We therefore make the practical recommendation to minimally sweep hyperparameters, but focus MoE performance optimization on expert count and granularity, setting expert size according to active parameter scale, then setting total expert count to match memory allowances. We also release our code and data at \url{https://github.com/hadasah/slicing_and_dicing}.

\section{Mixture of Experts}\label{sec:bg}

\begin{figure}[!ht]
  \centering
    \includegraphics[width=0.48\textwidth]{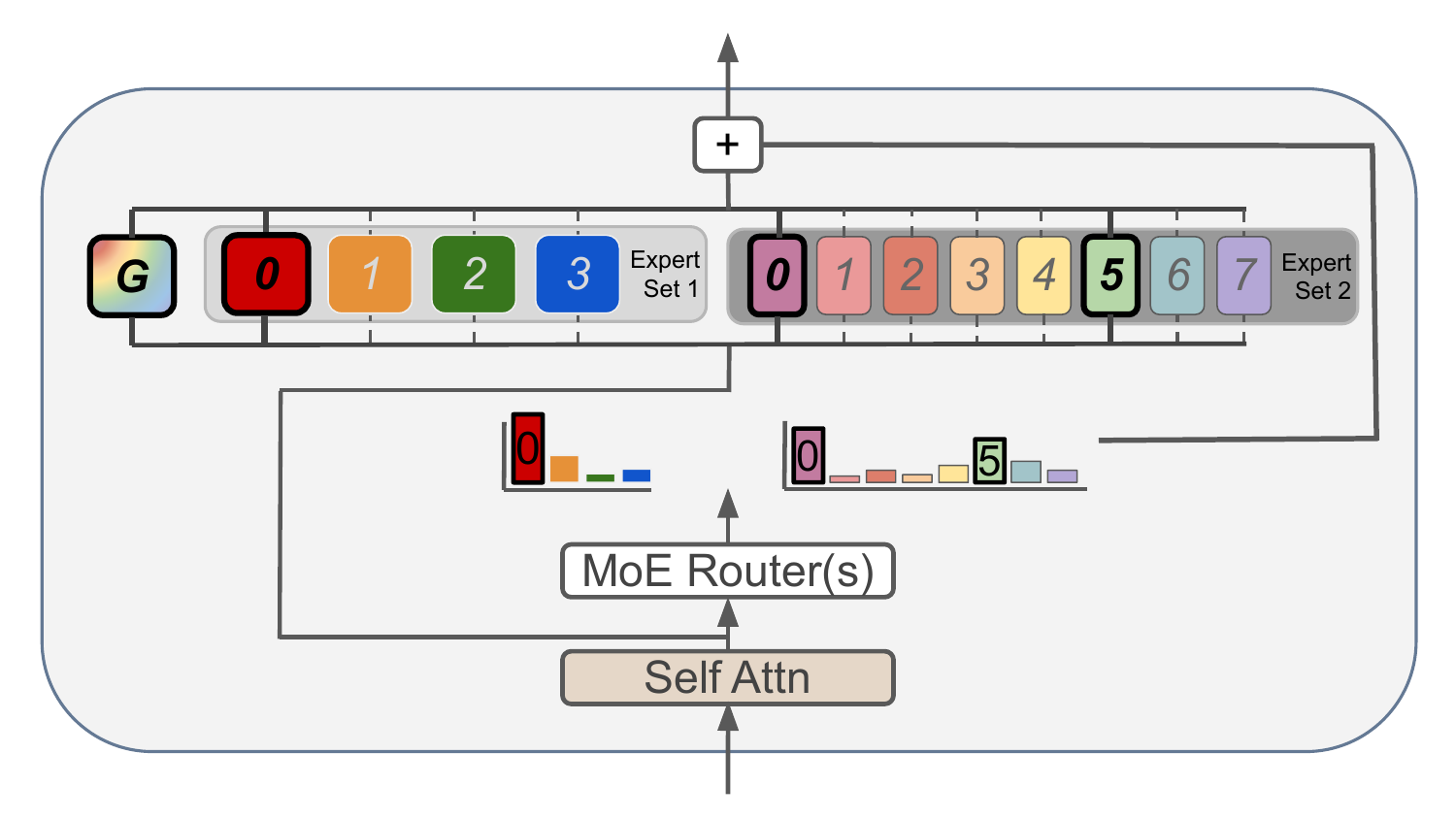}
  \caption{
  \textbf{A Mixture of Experts Layer}. In an MoE Transformer layer, each token first passes through the same self attention mechanism. Then, router(s) activate the highest affinity experts. Finally, the outputs of any activated FFN modules are combined in a weighted sum. A standard \emph{homogeneous} MoE layer includes $n$ experts of identical granularity $g$, of which $k$ are activated.
  We introduce more flexible configurations with \emph{heterogeneous} MoE layers consisting of experts with different granularity, as shown above (\emph{Expert Sets 1 and 2}). 
  An always-active \emph{generalist} 
  can also be added (labeled `G').
   }
  \label{fig:diagram_2}
\end{figure}

Mixture of Experts models contain conditionally activated submodules, selected by a routing mechanism. Recent MoE LLMs, which are based on the Transformer architecture, replace the feed-forward network (FFN) in some or all transformer layers. Instead of a single FFN, a router activates $k$ out of $n$ expert feed-forward networks. The chosen subset of $k$ depends on the input, and is commonly selected per-timestep (i.e., per-token), per-sequence, or per-document. The selected subset of expert FFNs generates outputs which are combined via an average weighted by the per-expert router weights.

\subsection{Expert Configurations} \label{sec:bg_expts}

\paragraph{Expert Size.} In \emph{dense} Transformer models, the intermediate FFN dimension is canonically 4 times the hidden dimension. 
MoEs were previously configured to have FFN experts of identical dimensions to their dense FFN counterparts. However, more recent work has explored fine-grained experts with decreased intermediate FFN dimension \citep{deepseek}.
We define granularity: 
\[g = \frac{\text{intermediate dimension of expert FFN}}{\text{intermediate dimension of dense FFN}}\]
A granularity of 1 indicates expert FFNs of the same size as the dense FFN, whereas a granularity of $\frac{1}{4}$ indicates expert FFNs with an intermediate dimension equivalent to $\frac{1}{4}$ that of the dense FFN.

\paragraph{Active Expert Count.} Increased active expert count incurs a higher FLOPs-per-token compute cost and memory usage, as each additional active expert entails additional computation. 
\citet{gshard} studied the tradeoffs in higher expert activation count with increased compute budget, and chose to activate 2 experts, as more than 2 experts yielded relatively marginal gains, but activating only 1 expert reduced performance. \citet{switch} chose to activate 1 expert, but \citet{stmoe} found that with properly configured load balancing, 2 experts indeed achieved better performance, perhaps as a result of enabling gradients to directly update router rows in relation to each other.

Active expert count is independent of expert size. However, to match FLOPs per step in performance comparisons, active expert count can be constrained to fix activated expert parameters:
\[\text{\# of active experts}\space\cdot\space\text{expert granularity} = \text{1}\]

In other words, to FLOP-match an MoE with expert granularity $g$ to a dense model, we activate exactly $k=\frac{1}{g}$ experts, thus maintaining an equal number of active parameters.

\paragraph{Total Expert Count.}
Total expert count may also be varied independently of expert size and active expert count. A larger total expert count results in only a negligible increase in compute cost.
However, each additional total expert incurs increased memory costs, depending on infrastructure.  

\paragraph{Activation Sparsity.} To enable fair comparison, we define MoE \emph{activation sparsity} $s$, the factor by which total FFN parameters exceed active FFN parameters
$s := n \cdot g$. This is sometimes referred to as \emph{expansion factor}, but this term is overloaded and also used to refer to FFN dimension expansion.

\paragraph{Heterogeneous Experts.} The granularity $g$ is typically shared for all experts; all experts have identical dimensions.
However, this homogeneous sizing of experts is not a requirement of the architecture. For example, instead of a set of 16 experts with granularity $\frac{1}{2}$, an MoE layer might have 8 experts of granularity $\frac{1}{2}$, as well as 16 experts of $\frac{1}{4}$ granularity. Intuitively, this additional flexibility may allow for additional improvement, as not all functions learned by the model require the same dimensions. With the flexibility to use experts with a variety of dimensions, models may be biased to more efficiently group learned functions and use available compute and parameter activation capacity.

\paragraph{Generalists.}
Also referred to as \emph{shared experts} \citep{deepseek}, generalists are experts which are always activated, for all inputs. Their outputs are combined with experts activated by the router. 

Generalists may have any granularity. MoE layers are not inherently limited to one generalist, but no evidence supports the possibility that 2 or more generalists may outperform a single larger generalist. Our experiments (\S\ref{sec:expt_main}) support aggregation of all shared parameters in a single generalist FFN.

\subsection{Router Configurations}

At each timestep, the router for a particular MoE layer selects a subset of experts to activate. The router may take as input any combination of document-, sequence- or token-level information, and in some cases uses additional metadata or pre-trained classifiers. However, the most straightforward token-level router uses only the hidden state of that token-layer to compute an affinity between each (hidden state, expert) pair. There are several possible algorithms for choosing token-expert pairings given these affinity scores. One of the most common is \emph{token choice} \citep{shazeer2017outrageouslylargeneuralnetworks}. For each token, the router activates the experts which correspond to the token's highest affinity values.

Without additional mechanisms, routers do not usually learn to evenly balance token \emph{loads} across experts, and instead route to, and therefore allow gradient flow to, only a small subset of available experts \citep{eigen2014learningfactoredrepresentationsdeep,bengio2016conditionalcomputationneuralnetworks}.

\subsubsection{Load Balancing Loss}
Auxiliary losses are a direct and explicit tool to induce load balancing. Auxiliary losses are added to the primary loss, such as a language modeling cross-entropy loss, weighted by a hyperparameter $\alpha$:
\begin{equation}
\label{eq:loss}
\mathcal{L} = \mathcal{L}_{\textit{CE}} + \alpha \mathcal{L}_{\textit{auxiliary}}
\end{equation}
A simple load balancing auxiliary loss \citep{shazeer2017outrageouslylargeneuralnetworks} penalizes overreliance on expert $E_i$ in each batch by multiplying the fraction of tokens in its load $f_i$ by its total routing probability $P_i$:
\begin{equation}  
\label{eq:lbl}
\mathcal{L}_{\textit{LB}} = N_{E} \cdot \sum_{i=1}^{N_{E}} f_i \cdot P_i
\end{equation}

However, this, like all auxiliary losses, introduces additional complexities. As each loss term changes in magnitude at different rates over the course of training, their weights must be carefully calibrated. If given too low of a weight, that auxiliary loss will have insufficient effect. Conversely, if any auxiliary loss dominates, optimization may bias the model to the detriment of its primary task.

\subsubsection{Loss Free Load Balancing}

A loss-free mechanism avoids the challenge of balancing loss terms, though it does introduce other inductive biases to the LM. \citet{deepseekv3} introduces a loss-free load balancing mechanism, which uses a bias to adjust the load on each expert towards more balanced routing. This per-expert bias is adjusted by an interval of $\gamma$, defined as a hyperparameter, at each train step.
They find that this loss-free mechanism is insufficient without a load balancing auxiliary loss, and results in MoE LMs that underperform those with the auxiliary loss term. By pairing the two mechanisms, with a lowered weight given to the load balancing loss, they are able to achieve improved performance.

\subsubsection{Dropless Routing}
Even with a router that perfectly load balances on average, variations between batches result in locally imbalanced routing. In these situations, one possible solution is to set a \emph{capacity factor}, the factor by which an expert's capacity to accept incoming tokens exceeds a perfectly balanced load. A common capacity factor is 2. However, this is inefficient, as at least half of the available capacity is unused, and it only reduces the incidence of overflow rather than eliminating it entirely. Any imbalance in excess of the capacity factor remains a problem. These remaining overflow tokens may be routed to the next best available expert, or they may be dropped altogether, passed directly to the next layer.
Dropless routing \citep{megablocks} uses block-sparsity to ensure that no tokens are dropped.

\subsubsection{Other Router Mechanisms}
\citet{stmoe} introduce an additional auxiliary loss, the Z-loss, designed to address instability in router training, especially at larger model scales:
\begin{equation}
\mathcal{L}_{\textit{RZ}}(x) = \frac{1}{B} \cdot \sum_{i=1}^B \left(\log \sum_{j=1}^{N_{E}} \exp({x_j^{(i)}}) \right)^2
\end{equation}
This is weighted by some $\alpha_{RZ}$ and summed with the cross-entropy loss and other auxiliary losses. 

\input{fig_tex/4_1_experts}

\section{Experiments} \label{sec:expt}

\paragraph{Model Architecture}
We train MoE Transformer LMs based loosely on \citet{olmoe}.
Our models range from roughly 10 million to 300 million active parameters,
and 10 million to 6.6 billion total parameters. Specifically, we refer to our model architectures 10M, 20M, 50M, 80M, 110M, 200M, 300M according to their active parameter count. 
See Appendix~\ref{app:param_counts} for additional discussion and Table~\ref{tab:param_counts} listing architecture details.

To FLOP-match all configurations on a per-timestep basis, we match active FFN parameters at each model scale.
We also compare architectures by total parameters via \emph{activation sparsity} $s$ (\S\ref{sec:bg_expts}).

\paragraph{Training Data} We use a training data mixture comprised of web text, code, math, and encyclopedic text, introduced and released openly by \citet{olmoe}. Additional details are in \S\ref{app:train_data}. 

\paragraph{Training Hyperparameters}
Full settings with discussion and ablations may be found in \S\ref{app:hparam}.

\paragraph{Evaluation} 

We show the macro-average cross-entropy loss on a diverse set of held-out validation data. Trends are similar across domains. Additional results for all experiments can be found in Appendix~\S\ref{app:expt_experts}-\ref{app:expt_experts_downstream_dataset}, including loss on individual domains and downstream task evaluations. In Appendix \S\ref{app:random_seed}, we experiment with 5 random seeds and find near-zero standard deviation on language modeling cross-entropy loss. Downstream task results are subject to some variance.

Additional details and discussion in \S\ref{app:eval_tasks}.

\subsection{Optimal Expert Configuration} \label{sec:expt_main}

We study experts with granularity $g\in\{1, \frac{1}{2}, \frac{1}{4}, \frac{1}{8}, \frac{1}{16}, \frac{1}{32}, \frac{1}{64}\}$ 
and total expert count $n \in\{2, 4, 8, 16, 32, 64, 128, 256, 512, 1024\}$, resulting in $s \in\{2, 4, 8, 16, 32, 64, 128\}$.
We FLOP-match by matching active FFN parameters, setting number of active experts $k = \frac{1}{g}$, or equivalently, $k \cdot g = 1$. Exact configurations are shown in Appendix \S\ref{app:param_counts} Figure~\ref{fig:hgn_configs}.

Results for selected model scales are shown here, with all other model scales in Appendix~\S\ref{app:results}. 

\paragraph{MoEs outperform dense models (above some compute minimum).}
At 50M and larger active parameter scale, even our most suboptimal MoE configurations often outperform dense baselines. 
At 10M and 20M scales, MoEs consistently fail to outperform dense baselines, across all settings (Figure~\ref{fig:10M20M_experts}). 
We hypothesize that this is a result of the interaction between data and parameter budget, rather than of model size alone. 
To verify this, we train 10M and 20M models with a 20 times and 5 times greater data budget, respectively. 
These results approximate the 50M models, which have a similar compute budget, supporting our hypothesis. See Appendix \S\ref{app:expt_small} for details and full results. 

\input{fig_tex/4_1_gran_4_2_het}

\paragraph{Increased inactive expert parameters improve performance in FLOP-matched settings.} 
At a fixed total number of experts, performance improves with increased expert size (Figure~\ref{fig:moe_count_and_granularity}). 
Similarly, at a fixed expert granularity, additional total (inactive) experts improve performance. There may exist a critical point, at which additional total expert parameters do not improve, or even actively harm performance. However, in this setting, we do not observe this phenomenon. (Additional discussion for other settings in \S\ref{sec:expt_router}.)
In general, the optimal configuration maximizes total expert parameters.

\paragraph{Optimal expert count and granularity vary by total parameter count.} 
We now consider the optimal expert (total count, granularity) configuration, constrained by a fixed total MoE parameter count, equivalent to a memory budget. We hold MoE activation sparsity $s$ fixed for $s\in\{2, 4, 8, 16, 32, 64\}$, and trade off total expert count for expert granularity. 
In Figure~\ref{fig:moe_count_and_granularity}, the optimal granularity of experts varies only slightly by MoE activation sparsity. 
As $s$ increases, optimal expert size increases slowly, suggesting that sparsity should be scaled up primarily by increasing expert count. However, optimal granularity clearly varies by effective model size.  
For 50M active parameters, this optimal expert granularity lies between $[\frac{1}{4}, \frac{1}{2}]$, shifting to $\frac{1}{2}$ at $s \geq 32$; for 110M, 
between $[\frac{1}{8}, \frac{1}{4}]$,
shifting to $\frac{1}{4}$ at larger $s$; for 300M, at $\leq \frac{1}{8}$.

\paragraph{Top-1 routing underperforms in FLOP-matched settings}
Models with only one activated expert ($g = 1$) underperform compared to the rest of the trend.
Fixing $n \in\{2, 4, 8, 16, 32, 64, 128, 256\}$, performance tends to improve as $g$ increases, but regresses when increasing $g$ from $\frac{1}{2}$ to $1$. 
It is possible that this degradation is a result of only activating 1 expert at a time, so that gradients from each token update only 1 row in the router weight matrix, disabling rows calibrating against each other. 
These findings are in line with prior work \citep{stmoe}.

\input{fig_tex/4_2_gen}

\paragraph{Reweighting finegrained FFNs without sparsity underperforms}

We hypothesize that the MoE architecture provides performance improvements which are \emph{not} a result of simply splitting the FFN mechanism into smaller, absent an increase in total parameters. We test this hypothesis by training a number of LMs which split the FFN into $n$ identical smaller FFN components, of granularity $g = \frac{1}{n}$. Thus, both the total and active parameters are unchanged, and the only modification is the division of FFN parameters into multiple, granular FFN mechanisms. We further consider two settings for these always-active components: weighting all components equally, or uniform pseudo-routing, which weights all components according to a learned pseudo-router. That is, though all components are active, they are weighted the same, or with learned affinity weights. 

Thus, both the total and active parameters are unchanged, and the only modification is the division of FFN parameters into multiple, granular modules. The dimension of each individual FFN component is reduced, which might suggest a negative impact on performance. Conversely, the inductive bias introduced by dividing the FFN may allow for more parallel learning and benefits performance.

For our experiments, we set $n \in\{1, 2, 4, 8\}$. Across all active parameter scales and all settings, we find that increasing granularity in a densely activated model degrades performance (Figure~\ref{fig:dense_ablation}). Thus, we find additional support for our conclusion that increasing total model parameters is the primary mechanism for improved MoE performance.

\subsection{Additional Expert Configurations: Generalists and Heterogeneity} \label{sec:expt_hetgen}

\input{fig_tex/4_3_dropless}

The results of \S\ref{sec:expt_main} provide evidence that a significant determinant of MoE performance is the optimal division of \emph{total expert parameters} between \emph{expert size} and \emph{total expert count}. However, the architectures considered were tightly constrained to consist only of one set of conditionally activated, homogeneously sized experts. We expand our analysis to a more flexible definition of Mixture of Experts, including \emph{heterogeneous} experts and \emph{generalists}, both separately and combined.

To FLOP-match in this flexible setting, we again match active FFN parameters: we ensure that all configurations satisfy $g_{gen} + \sum_i k_i \cdot g_i = 1$. That is, we match, to the dense baseline, the active FFN parameters summed across the generalist and all activated experts. We also generalize our definition of sparsity to the case of multiple expert pools and a generalist: $s = g_{gen} + \sum_i n_i \cdot g_i$.

\paragraph{Expert heterogeneity alone does not improve performance.} 
We now consider heterogeneous experts. Given the combinatorial explosion of possibilities, this exploration rapidly becomes prohibitively costly, and we constrain the settings we explore thus: layers consist of 2 pools of $(n_1, n_2)$ experts, between which we equally divide both active and total parameters, so that each incurs half the total compute and memory cost. 
We experiment with granularity $(g_1, g_2) =  \{(\frac{1}{2}, \frac{1}{4}), (\frac{1}{4}, \frac{1}{8}), (\frac{1}{8}, \frac{1}{16})\}$ and total expert count $(n_1, n_2) \in \{(2, 4), (4, 8), (8, 16)\}$. Additional discussion in Appendix~\S\ref{app:het_configs}.

Results in Figure~\ref{fig:heterogeneous} show that, as with homoegeneous experts \S\ref{sec:expt_main}, when holding the granularity of each expert pool fixed, performance improves with increased total expert count. 
Similarly, when holding total number of experts fixed, the settings with largest experts, and equivalently, with the greatest number of total expert parameters, is optimal. 
Holding MoE activation sparsity $\{s_1,s_2\}$ fixed, the optimal numbers of experts $\{n_1, n_2\}$ increases almost linearly with $\{s_1, s_2\}$, while the optimal granularity $\{g_1, g_2\}$ increases slowly, remaining nearly constant for each model scale (Fig~\ref{fig:heterogeneous}).

In our heterogeneous settings, results lie on a plausible interpolated line between the most similar homogeneous settings, suggesting that heterogeneity alone does not yield improved performance.

\paragraph{Generalists do not improve performance.} 
We experiment with the inclusion of a generalist, varying its granularity within $\{\frac{1}{2}, \frac{1}{4}, \frac{1}{8}\}$, across many expert configurations. 
We consider both homogeneous and heterogeneous expert setups, focusing on the best performing setups from \S\ref{sec:expt_main} and \S\ref{sec:expt_hetgen}. In the homogeneous setting, experts are constrained by definition to be of equal or finer granularity than the generalist. For example, with a generalist of granularity $\frac{1}{8}$, the remaining experts must have a granularity sum of $\frac{7}{8}$ under the FLOP-matching constraint, so it is not possible to use only experts of granularity $\frac{1}{2}$.
As shown in Figure~\ref{fig:generalist},
we find that having any generalist of sparsity $\frac{1}{2}$, $\frac{1}{4}$ or $\frac{1}{8}$ consistently results in performance that is comparable to or slightly degraded from the no-generalist setting. See Appendix \S\ref{app:expt_experts} for additional scales and heterogeneous MoEs with generalists. 

\input{fig_tex/4_3_lb}

\subsection{Ablations}
\label{sec:expt_router}

We ablate the importance of load balancing mechanisms and dropless routing to understand their respective impacts on MoE performance across model scales and expert configurations. Other hyperparameter settings, including z-loss, are varied in preliminary experiments (Appendix~\S\ref{app:settings}).

\paragraph{Dropless routing improves performance}

We ablate the benefits of dropless routing across homogeneous and heterogeneous expert configurations, various sized generalists, and load balancing. 

Comparing setups with and without dropless routing, as in Figure~\ref{fig:dropless}, we find that dropless routing results in improved performance. However, at our largest total number of experts $n=1024$, default routing begins to perform comparably to dropless routing. This may be due to the reduced incidence rate of token dropping when average expert load decreases. See additional discussion in Appendix~\S\ref{app:settings}.

\paragraph{Load balancing settings have minimal impact within specific ranges.}

Finally, we vary the weight applied to the load balancing loss between $\{\num{1e-4}, \num{1e-2}\}$, as well as the loss-free load balancing bias $\gamma$ between $\{0, \num1e-3\}$ ($\gamma=0$ indicates no loss-free load balancing). 
We consider a variety of homogeneous and heterogeneous expert configurations, and generalist settings.

Results for 50M homogeneous MoEs are shown in Figure~\ref{fig:lb}, with heterogeneous MoEs, generalists, and additional model scales in Appendix \ref{app:expt_experts}.
At all scales, we find that the worst setting uses the lower load balancing loss weight $\alpha_{LB} = \num{1e-4}$ and $\gamma=0$.
This is likely because there is not enough bias applied to the model for load balancing.
This is supported by higher load imbalance and load balancing loss (Appendix Figure~\ref{fig:lb_loss}-\ref{fig:load_imbalance}). 
No setting clearly dominates at total expert count $n\leq 128$. However, with higher $n$,
loss-free load balancing damages performance. This may be due to its design, which adjusts the bias term on each expert's routing probability by increments of the hyperparameter $\gamma$, set to \num{1e-3} in our experiments, following \citet{deepseekv3}. 
At large $n$, bias $\gamma = \num{1e-3}$ may result in overly coarse-grained adjustments. Thus, the bias $\gamma$ requires tuning, especially at high $n$. Additionally, \citet{deepseekv3} did not consider dropless routing, and may have required a loss-free mechanism to avoid token dropping without over-weighting the auxiliary loss.

\section{Related Work}

Earlier work \citep{clark2022unified,ludziejewski2025jointmoescalinglaws,abnar2025parameters,wang2024scaling} established scaling laws for MoE models in simplified settings \citep{switch,gshard}. 
However, recent MoE architectures define a more complex design space \citep{deepseekv3}. 
\citet{deepseekv3} demonstrate that combining fine-grained specialization \citep{krajewski2024scalinglawsfinegrainedmixture} with shared experts \citep{rajbhandari2022deepspeed} and improved routing algorithms results in a superior recipe. 
Consequently, determining the optimal expert granularity and shared expert allocation have added complexity to determining the best MoE settings at scale. 
In this work, we expand this investigation with a wider range of jointly varied expert counts and granularities, and by considering various combinations of heterogeneous experts \citep{wang2025hmoe} and generalists.

\citet{olmoe} ablate various training factors, but focus on a single scale to train a 1B active (7B total) parameter MoE model. 
They individiually consider one factor while fixing all others: expert size, generalists, expert and token choice routing \citep{zhou2022mixture}, load balancing losses \citep{shazeer2017outrageously,deepseek} and Router Z-loss \citep{stmoe}. 
Consistent with our results, they find that using shared experts degrades performance.

\citet{tian2025greaterleveragescalinglaws} look at activation ratio, expert granularity and shared experts through the lens of computational efficiency across scales. 
However, their methodology studies these factors in isolation and ties expert granularity to expert count. 
This difference likely explains why they recommend much more fine-grained experts, whereas our strictly disentangled grid search recommends significantly coarser experts and finds no benefit from shared experts. 
Finally, \citet{zhao2025comprehensivescalinglawmixtureofexperts} argue that shared experts are essential and identify a fixed optimal active expert count of $\approx 7$. 
Our results contradict this, suggesting that these optima are scale-dependent rather than fixed constants. 
We hypothesize that these may be artifacts of overfitting to a specific regime; as noted by \citep{misfitting}, scaling laws derived from high-degree-of-freedom fits often fail to extrapolate reliably.

\section{Conclusion} \label{sec:conclusion}
We exhaustively search MoE configurations, varying total experts, expert granularity, heterogeneous experts, generalist size, load balancing mechanisms, and dropless routing. Through pretraining over 2,000 LMs with up to 6.6B total parameters, we simplify practical recommendations for optimizing MoE architectures: Maximize total (inactive) expert parameters by increasing total expert count, while keeping coarse expert granularity around $\frac{1}{4}$
in our settings. Auxiliary losses and other mechanisms have marginal impact on improving performance but must be tuned to avoid interference with the language modeling objective.

\bibliography{example_paper}
\bibliographystyle{icml2026}

\newpage
\appendix
\onecolumn
\section{Additional Training Details}

\subsection{Training Data} \label{app:train_data}
Our training data is directly taken from \citet{olmoe} and consists of documents from: 
DCLM-Baseline~\citep{li2024datacomplm}, StarCoder~\citep{li2023starcoder,kocetkov2022stack3tbpermissively}, peS2o~\citep{peS2o,soldaini2024dolma}, arXiv~\citep{together2023redpajama}, OpenWebMath~\citep{paster2023openwebmath}, Algebraic Stack~\citep{azerbayev2023llemma}, English Wikipedia \& Wikibooks~\citep{soldaini2024dolma}. 

Each model is trained with a token-to-active-parameter ratio of about 20, following the recommendations of \citet{chinchilla}. In other words, a model with 1 billion active parameters is trained with about 20 billion tokens.

\subsection{Evaluation Tasks} \label{app:eval_tasks}
Our evaluation suite consists of held-out language modeling tasks and downstream tasks. 

The language modeling tasks are a subset of Paloma
\citep{magnusson2024palomabenchmarkevaluatinglanguage}:
\textsc{C4} (\citet{raffel2019exploringtl} via \citet{dodge-etal-2021-documenting}), \textsc{The Pile} \citep{Gao2020ThePA}, \textsc{WikiText-103} \citep{Merity2016PointerSM}, \textsc{Dolma} \citep{dolma}, \textsc{M2D2 S2ORC} \citep{reid-etal-2022-m2d2}, \textsc{ICE} (\citet{GREENBAUM_1996} via \citet{Liang2022HolisticEO}). \textsc{Dolma} is subdivided into six domains: books, common-crawl, pes2o, reddit\_uniform, stack\_uniform, wiki.

The downstream tasks consist of BoolQ \citep{clark2019boolq}, HellaSwag \citep{zellers2019hellaswag}, and MMLU \citep{son2024kmmlumeasuringmassivemultitask}. MMLU is subdivided into four domains: humanities, STEM, social sciences, and other. 

For language modeling tasks, we compute and report the macro-average cross-entropy loss.

\subsection{Hyperparameters} \label{app:hparam}
\input{tables/hyperparams}

\subsection{Hardware and GPU-hours} \label{app:gpus}
We train on a mix of A40, L40, L40S, and H200 GPUs. Our main experiments represent roughly 32,000 GPU-hours of training and evaluation compute, with an additional 9,500 GPU-hours represented in preliminary investigations of hyperparameters and architectures. Exact GPU-hours per experiment varies greatly by architecture and hardware, but is roughly 2, 4, 8, 12, 24, 56, and 108 GPU-hours for 10M, 20M, 50M, 80M, 110M, 200M, and 300M model training.

\subsection{Active and Total Parameter Counts}
\label{app:param_counts}

In Table~\ref{tab:param_counts}, we list all architecture details for each model configuration. 

In Figure~\ref{fig:hgn_configs}, we show the (number of experts, granularity) configurations represented in \S\ref{sec:expt}.

\input{tables/param_counts}
\input{tables/hgn_configs}

\clearpage
\subsection{Heterogeneous Expert Configurations} \label{app:het_configs}
In \S\ref{sec:expt_hetgen}, we constrain our heterogeneous MoE settings, only considering those expert configurations consisting of exactly two pools of experts, consisting of total expert counts $(n_1, n_2)$, active expert counts $k_1, k_2$, and granularity $(g_1, g_2)$, such that $n_1 = \frac{1}{2} \cdot n_2$, $k_1 = \frac{1}{2} \cdot k_2$, and $g_1 = 2 \cdot g_2$. Thus, $k_1 \cdot g_1 = k_2 \cdot g_2$ and $n_1 \cdot g_1 = n_2 \cdot g_2$, so each pool of expert has identical total and active expert parameters.

Each such setting may be seen as an interpolated point between two homogeneous settings that we study. Firstly, the setting with $2 \cdot n_1$ total and $2 \cdot k_1$ active experts of granularity $g_1$; secondly, the setting with $2 \cdot n_2$ total and $2 \cdot k_2$ active experts of granularity $g_2$.

\input{tables/het_configs}

\clearpage
\section{Discussion of Hyperparameters and Configurations}
\label{app:settings}

\subsection{Preliminary Hyperparameter Investigations}

\paragraph{Learning Rate}

We sweep the learning rate in $\{\num{1e-4}, 4e-4, \num{1e-3}, 4e-3, \num{1e-2}\}$. We find that the optimal learning rate does indeed vary by active parameter count, but also by total parameter count, and with other training settings. As such, it would be necessary to sweep each setting if we were to find an approximate optimal learning rate. Even then, we would only be able to approximate the optimum, because of the computational infeasibility of finding, with high resolution, the exact optimal point for every single setting. 
\citet{misfitting} demonstrates that performing a sweep for optimal learning introduces additional random noise, and that fixing a learning rate allows for a much more stable scaling law fit. This may be a result of the difficulty of finding the true optimal learning rate with sufficient resolution. Therefore, we choose to fix a learning rate which lies near the optimum for the largest scale we consider, at over 100 million active parameters. This trades off a closeness to the exact optimal performance metrics, which are unique to our setup, with consistency and legibility of trends, which may be a greater contribution to the understanding of the research community.

\paragraph{Learning Rate Scheduler}

We test the Warmup-Stable-Decay (WSD) learning rate schedule, which we compare to the Cosine Decay schedule we use in the experiments of \S\ref{sec:expt}. Although initial results showed that WSD slightly outperformed Cosine Decay, we observed some sensitivity to settings as we increased model scale. As WSD was more recently introduced, it has not yet been thoroughly studied with MoEs, and, bounded by computational constraints, we chose to utilize the dominant Cosine Decay Schedule instead.

\paragraph{MoE Z-Loss Weight}

There is a general lack of work investigating the effects of the Z-Loss since its introduction, with many works omitting any discussion of it, or adopting the mechanism and the default weight from \citet{stmoe}. In initial experiments, we remove the Z-Loss to ablate its importance. We find little impact on model performance. However, the Z-Loss is argued to aid in router instability. Router instability disrupts model training stability, but generally at much larger active parameter size, above 1 Billion. We retain the Z-Loss as a default in anticipation of its benefit at larger scales.

\subsection{Additional Router Configuration Discussion} \label{app:router_configs}

\paragraph{Heterogeneous Router Implementation
}
In each Transformer layer, the FFN is replaced by a module consisting of one or more routed expert pools, and of zero or one generalists. In the case of one expert pool, all experts have the same granularity. In the case of two expert pools, each pool consists of experts which are identical in granularity to each other, but experts across pools have different granularity. Though it's theoretically possible to route to all experts of varied granularity from a single router, our implementation includes one router per pool of experts, due to compute infrastructure constraints. This implementation is functionally equivalent to the case with a single router, and does not impact performance. 

\paragraph{Token vs Expert Choice} 

\citet{expertchoice} introduces \emph{Expert Choice} routing, in which the router selects, for each expert, its highest affinity tokens. This is in contrast to the Token Choice routing we use, in which the router selects, for each token, its highest affinity experts. Token Choice routing guarantees that each token may be seen by the same number of experts. Expert Choice routing guarantees load balancing for each expert on even a batch scale, without any need for additional load balancing mechanisms. However, it suffers from 2 primary drawbacks. First, each token is not guaranteed to be processed with the same number of experts, or equivalently, the same compute, and some tokens are effectively dropped altogether. Modifications may be made to ensure each token is seen by at least one expert, but these add complexity. In some cases, adaptive computation cost may be desirable, but such a study lies outside the scope of our investigation. Secondly, the expert choice algorithm leaks information about future tokens. In other words, the routing decision for tokens early in a sequence is informed by the tokens later in that sequence. 

For these reasons, we employ Token Choice routing and use load balancing mechanisms and dropless routing to manage its drawbacks.

\paragraph{Dropless vs Default Routing}
We observe, in some preliminary experiments, that dropless and default routing perform comparably at our highest total expert counts of 256, 512, and 1024. This result can be partially observed in Figure~\ref{fig:lm_avg_dropless_with_lf}. It is likely that with such high sparsity, the dropless mechanism is of negligible benefit. However, such high expert counts have been very rare in the literature, and we focus our hyperparameter and configuration selection on settings more commonly seen in prior work.

\newpage
\section{Additional Results}
\label{app:results}

\subsection{Sensitivity to Random Seed} \label{app:random_seed}
We verify the stability of our results without requiring multiple random seeds for all settings. For one dense and one MoE setting at 50M scale, we run experiments with 5 random seeds. In Table~\ref{tab:random_seeds}, we report the mean and standard deviation of the cross-entropy loss for each dataset. Standard deviations for language modeling domains remain close to 0, indicating stable results even without multiple random seeds. Standard deviations for downstream tasks, which are shown only in Appendix~\ref{app:results}, are higher, with the exception of HellaSwag.  
\input{tables/random_seeds}

\clearpage
\subsection{Train Cross-entropy Loss, Load Balancing Loss, and Load Imbalance}
\label{app:expt_lb}
\input{fig_tex/hist_plots}

\clearpage
\subsection{10M and 20M Scale Results}
\label{app:expt_small}

We present all results for models at our 10M and 20M active parameter scales. As shown in Figure~\ref{fig:10M20M_experts}, all MoEs underperform dense baselines at these scales. This pattern holds as we add hetereogeneous experts (Figure~\ref{fig:10M20M_het}) and a generalist (Figure~\ref{fig:10M20M_gen}), and as we ablate load balancing mechanisms (Figure~\ref{fig:10M20M_lb}).
However, we hypothesize that this performance gap is a result of more than active parameter count alone, and train 10M and 20M LMs with a 5 times and 20 times greater data budget, respectively, such that total train tokens is roughly 100 times and 400 times active parameter count, respectively  (Figure~\ref{fig:10M_more_data}-\ref{fig:20M_more_data}). In this setting, we see that MoEs once again outperform dense baselines. This setting also has a similar compute budget to our 50M scale experiments, and results in similar tradeoffs between performance and expert granularity and count.

\input{fig_tex/10M20M}

\clearpage
\subsection{All Model Scale Results} \label{app:expt_experts}
\input{fig_tex/lm_avg}

\clearpage
\subsection{Hellaswag Accuracy Results} \label{app:expt_experts_hellaswag_acc}

\input{fig_tex/downstream/hellaswag_acc}

\clearpage
\subsection{Language Modeling Results -- by Dataset} \label{app:expt_experts_lm_dataset}
For full reporting, we include results on each Language Modeling validation dataset from \S\ref{app:eval_tasks}, which together comprise the Language Modeling macro-average reported in the main text of our paper. Trends are consistent across datasets.

\subsubsection{\textsc{C4}}
\input{fig_tex/lm/c4}

\clearpage
\subsubsection{\textsc{Dolma} Books}
\input{fig_tex/lm/dolma_books}

\clearpage
\subsubsection{\textsc{Dolma} common-crawl}
\input{fig_tex/lm/dolma_common_crawl}

\clearpage
\subsubsection{\textsc{Dolma} pes2o}
\input{fig_tex/lm/dolma_pes2o}

\clearpage
\subsubsection{\textsc{Dolma} reddit}
\input{fig_tex/lm/dolma_reddit}

\clearpage
\subsubsection{\textsc{Dolma} stack}
\input{fig_tex/lm/dolma_stack}

\clearpage
\subsubsection{\textsc{Dolma} wiki}
\input{fig_tex/lm/dolma_wiki}

\clearpage
\subsubsection{\textsc{ICE}}
\input{fig_tex/lm/ice}

\clearpage
\subsubsection{\textsc{M2D2 S2ORC}}
\input{fig_tex/lm/m2d2_s2orc}

\clearpage
\subsubsection{\textsc{The Pile}}
\input{fig_tex/lm/pile}

\clearpage
\subsubsection{\textsc{WikiText-103}}
\input{fig_tex/lm/wikitext_103}

\clearpage
\subsection{Downstream Task Results -- by Dataset} \label{app:expt_experts_downstream_dataset}
For full reporting, we include results on each Downstream task dataset from \S\ref{app:eval_tasks}. As discussed and seen in Appendix~\S\ref{app:random_seed} Table~\ref{tab:random_seeds}, downstream tasks are subject to random noise which overwhelms any signal. The exception is HellaSwag (\S\ref{app:hellaswag}), which exhibits trends similar to our language modeling datasets.

\subsubsection{BoolQ} \label{app:boolq}
\input{fig_tex/downstream/boolq}

\clearpage
\subsubsection{HellaSwag} \label{app:hellaswag}
\input{fig_tex/downstream/hellaswag}

\clearpage
\subsubsection{MMLU Humanities}
\input{fig_tex/downstream/mmlu_humanities}

\clearpage
\subsubsection{MMLU Other}
\input{fig_tex/downstream/mmlu_other}

\clearpage
\subsubsection{MMLU Social Sciences (5-shot)}
\input{fig_tex/downstream/mmlu_social_sciences}

\clearpage
\subsubsection{MMLU STEM (5-shot)}
\input{fig_tex/downstream/mmlu_stem}

\clearpage

\end{document}

%% file: fig_tex/4_1_experts.tex
\begin{figure*}[!t]
    \centering
    \begin{subfigure}[t]{\textwidth}
        \begin{subfigure}[t]{0.33\textwidth}
            \centering
            \caption*{\scriptsize Fixed total experts (n)}
            \includegraphics[width=\linewidth]{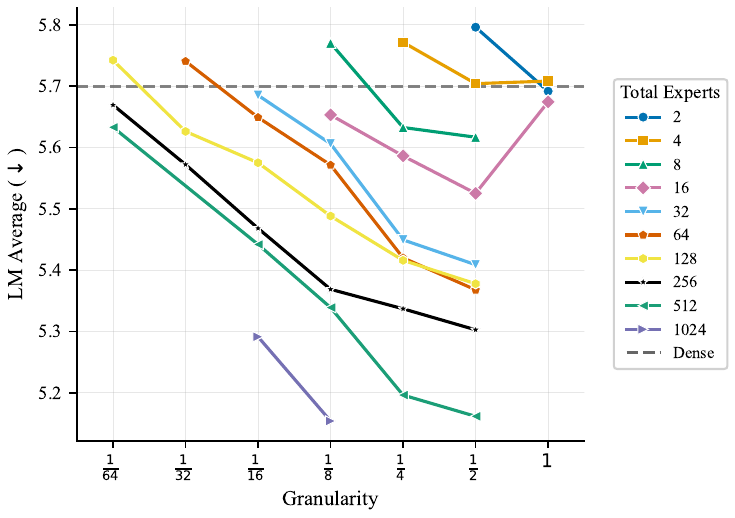}
        \end{subfigure}
        \begin{subfigure}[t]{0.33\textwidth}
            \centering
            \caption*{\scriptsize Fixed granularity (g)}
            \includegraphics[width=\linewidth]{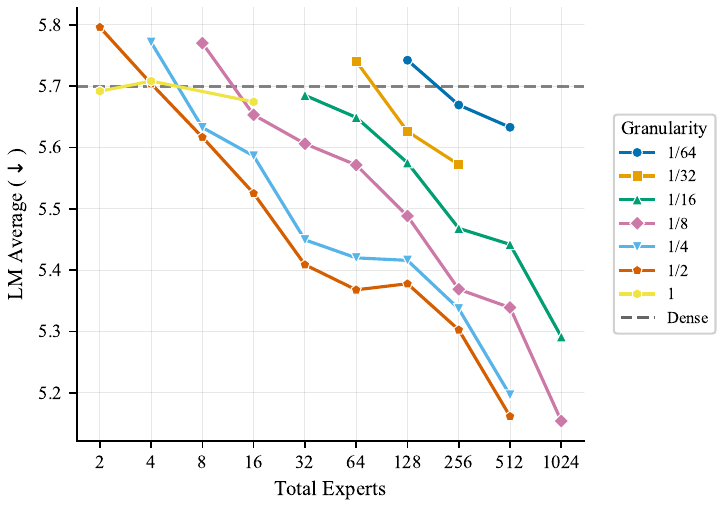}
        \end{subfigure}
        \begin{subfigure}[t]{0.33\textwidth}
            \centering
            \caption*{\scriptsize Fixed activation sparsity (s)}
            \includegraphics[width=\linewidth]{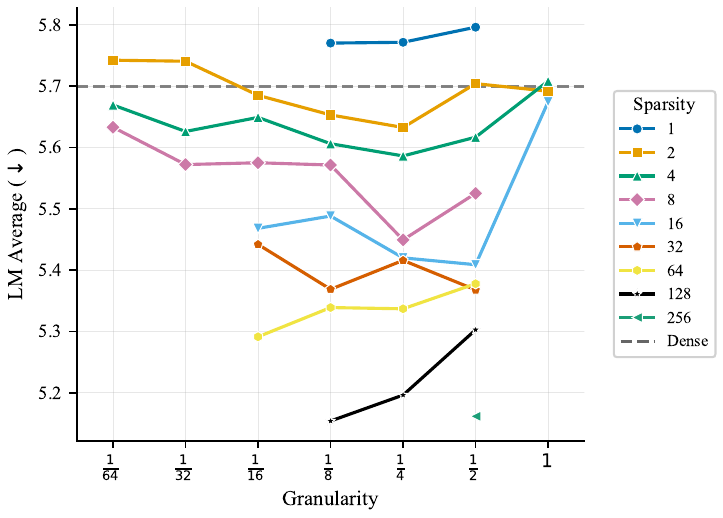}
        \end{subfigure}
        \caption{50M active, 50M - 930M total parameters}
    \end{subfigure}
    \begin{subfigure}[t]{\textwidth}
        \begin{subfigure}[t]{0.33\textwidth}
            \centering
            \includegraphics[width=\linewidth]{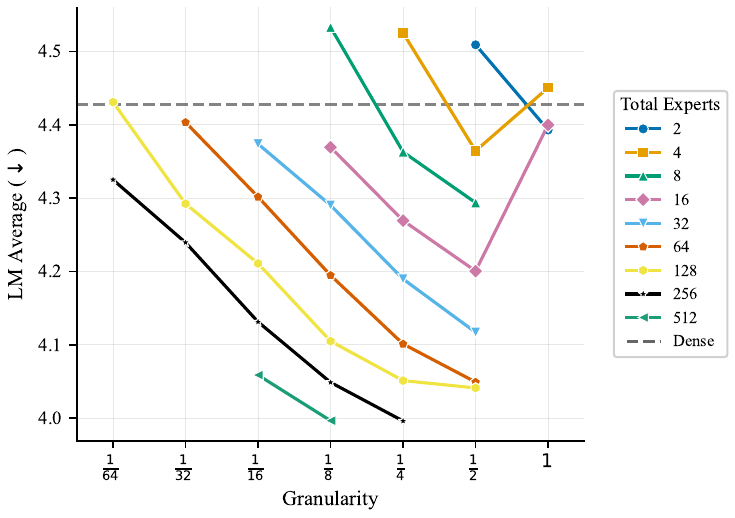}
        \end{subfigure}
        \begin{subfigure}[t]{0.33\textwidth}
            \centering
            \includegraphics[width=\linewidth]{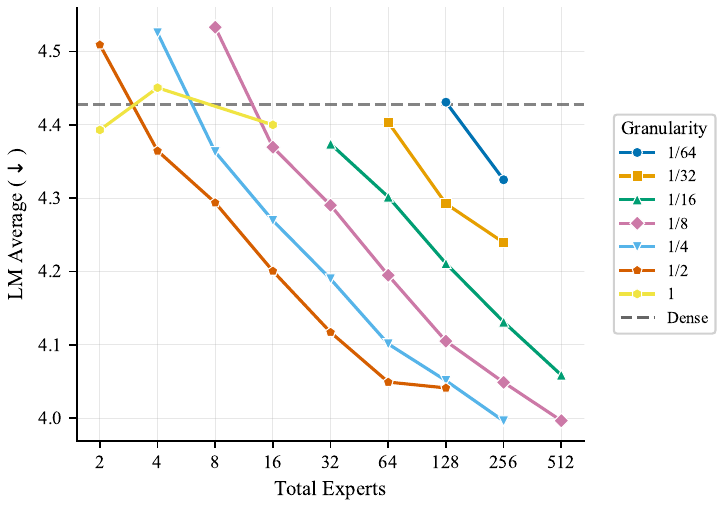}
        \end{subfigure}
        \begin{subfigure}[t]{0.33\textwidth}
            \centering
            \includegraphics[width=\linewidth]{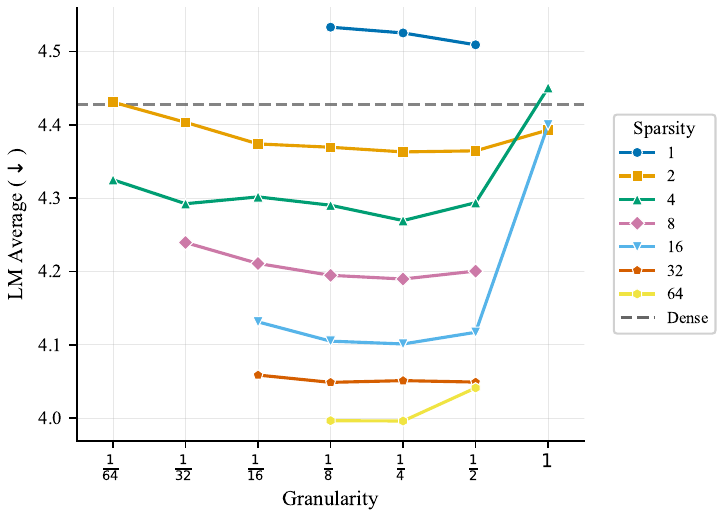}
        \end{subfigure}
        \caption{110M active, 110M - 1.4B total parameters}
    \end{subfigure}
    \begin{subfigure}[t]{\textwidth}
        \begin{subfigure}[t]{0.33\textwidth}
            \centering
            \includegraphics[width=\linewidth]{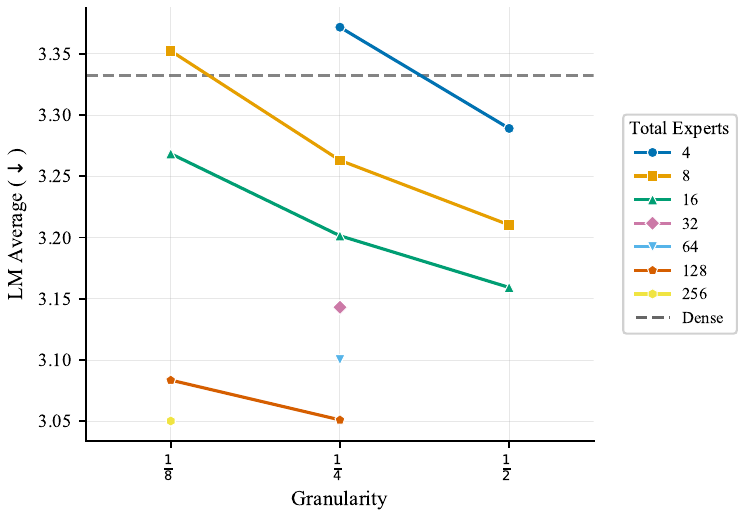}
        \end{subfigure}
        \begin{subfigure}[t]{0.33\textwidth}
            \centering
            \includegraphics[width=\linewidth]{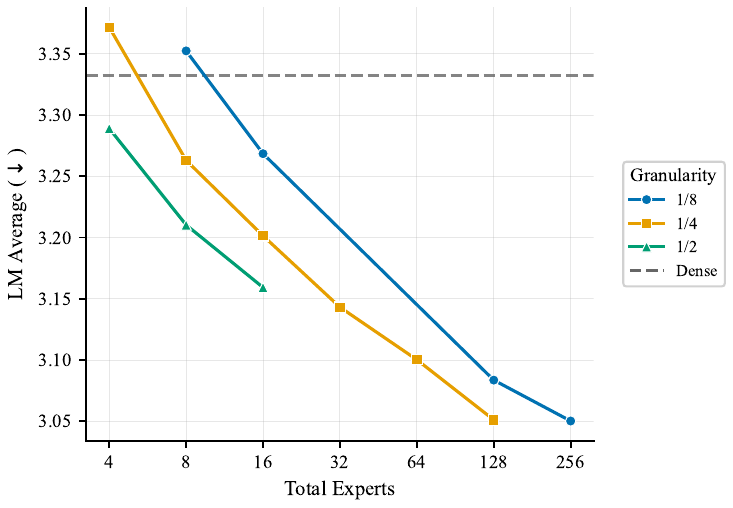}
        \end{subfigure}
        \begin{subfigure}[t]{0.33\textwidth}
            \centering
            \includegraphics[width=\linewidth]{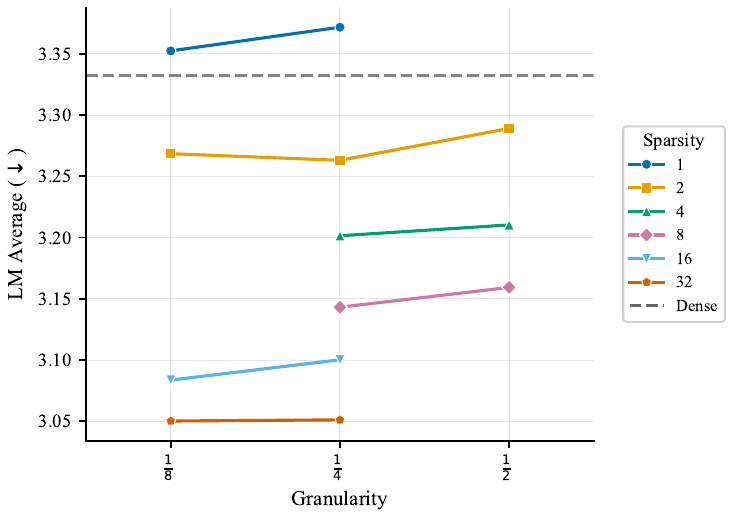}
        \end{subfigure}
        \caption{300M active, 300M - 6.6B total parameters}
    \end{subfigure}
    \caption{
    \textbf{Increasing inactive expert parameters via expert size (left) or total count (center) improves performance in MoEs (\S\ref{sec:expt_main}).} This effect is seen both when holding total number of experts fixed (left) and when holding expert granularity fixed (center). In general, increasing total parameters results in improved performance. Additional model scales in Appendix \S\ref{app:expt_experts}. \textbf{Optimal tradeoff between expert count and granularity varies in MoEs (right). (\S\ref{sec:expt_main})}
    At each activation sparsity $s$ (equivalently, at each total parameter count), the optimal (total expert count, expert granularity) configuration varies. As $s$ increases, optimal expert granularity remains nearly fixed, suggesting that sparsity should be scaled up primarily by increasing total expert count $n$, while maintaining a near constant, slowly increasing expert granularity $g$. Additional model scales in Appendix \S\ref{app:expt_experts}.
    }
    \label{fig:moe_count_and_granularity}
\end{figure*}

%% file: fig_tex/4_1_gran_4_2_het.tex
\begin{figure}[!t]
    \centering
    \begin{minipage}{0.47\textwidth}
        \centering
        \includegraphics[width=\textwidth]{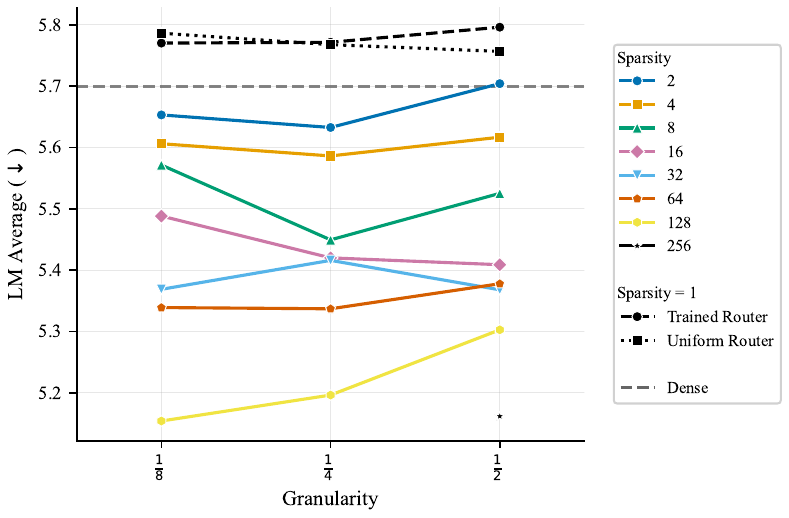}
        \caption{ \textbf{MoE performance improvements are not due to granularity alone (\S\ref{sec:expt_main})}. To further disentangle the effects of granularity from the sparsity afforded by MoE LMs, we investigate configurations with activation sparsity $s=1$ and granularity $g\in\{\frac{1}{2}, \frac{1}{4} , \frac{1}{8}\}$. In other words, all parameters are activated, as in the dense Transformer baseline, but the FFN mechanism is split into 2, 4, or 8 components. We find that all of these settings substantially underperform the dense baseline, indicating that granularity without sparsity is not sufficient.
        }
        \label{fig:dense_ablation}
    \end{minipage}\hfill
    \begin{minipage}{0.47\textwidth}
        \centering
        \includegraphics[width=\linewidth]{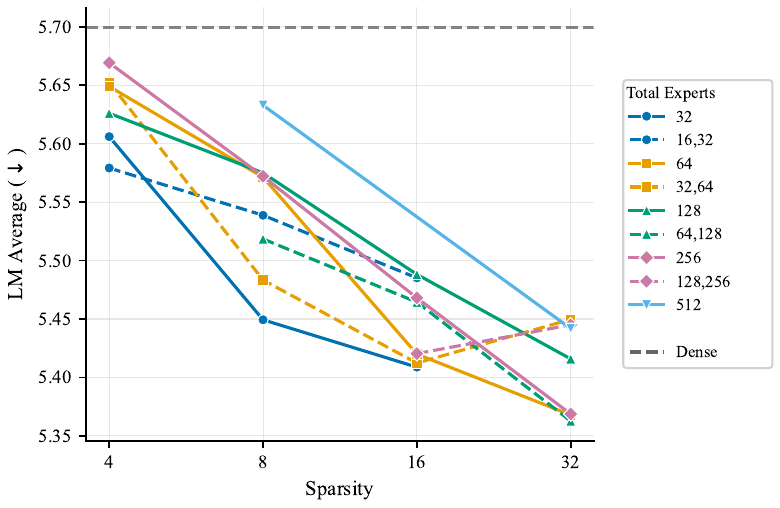}
        \caption{
        \textbf{Heterogeneity of expert size alone does not improve MoE performance (\S\ref{sec:expt_hetgen}).} To explore the potential benefits of their architectural flexibility, we compare heterogeneous MoEs (indicated by dotted lines) to active- and total-parameter-matched homogeneous MoEs. Heterogeneity alone does not result in performance gains, as, at each activation sparsity $s$, heterogeneous MoEs with $n_1, n_2 = a, b$ lie between or near the 2 closest homogeneous MoEs, with $n=a$ and with $n=b$. Additional model scales in Appendix \S\ref{app:expt_experts}.
        }
    \label{fig:heterogeneous}
    \end{minipage}
\end{figure}

%% file: fig_tex/4_2_gen.tex
\begin{figure*}[!t]
    \centering
    \begin{subfigure}[t]{0.98\textwidth}
         \centering
         \includegraphics[width=\linewidth]{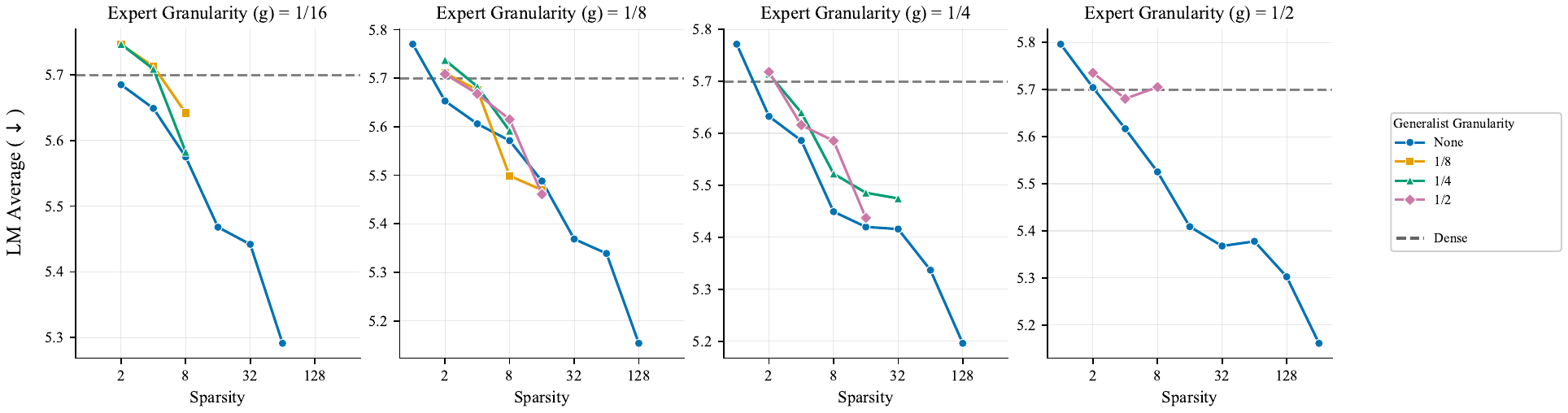}
         \caption{50M active, 50M - 930M total parameters}
    \end{subfigure}       
    \caption{
    \textbf{The inclusion of a generalist consistently degrades performance in homogeneous MoEs (\S\ref{sec:expt_hetgen}).}
    We train MoE LMs which consist of some routed experts with granularity $g$, as well as a generalist with granularity $g_{gen}\in \{\frac{1}{2}, \frac{1}{4}, \frac{1}{8}\} $. We compare to settings with no generalist, only routed experts with granularity $g$. In all settings and configurations, the addition of any granularity generalist results in comparable or degraded performance. Additional model scales in Appendix \S\ref{app:expt_experts}.}
    \label{fig:generalist}
\end{figure*}

%% file: fig_tex/4_3_dropless.tex
\begin{figure*}[!t]
    \centering
    \begin{subfigure}[]{0.46\textwidth}
        \centering
        \includegraphics[width=\linewidth]{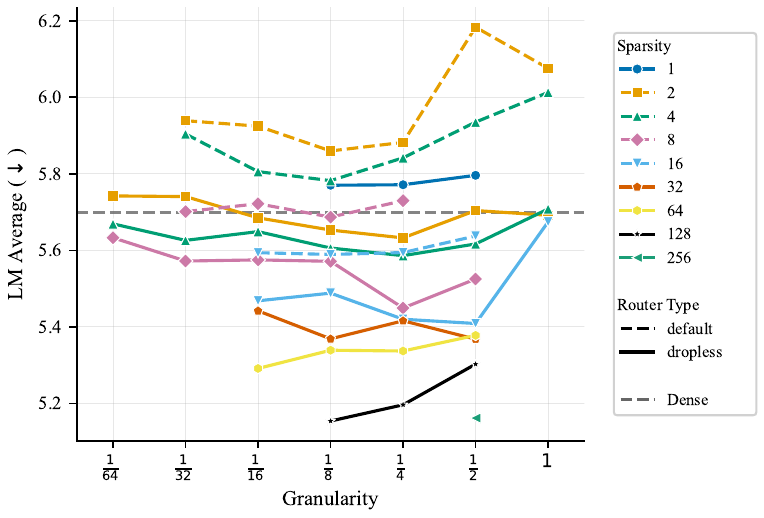}
        \caption{50M active, 50M - 930M total parameters}
    \end{subfigure}
    \hspace{1em}
    \begin{subfigure}[]{0.46\textwidth}
        \centering
        \includegraphics[width=\linewidth]{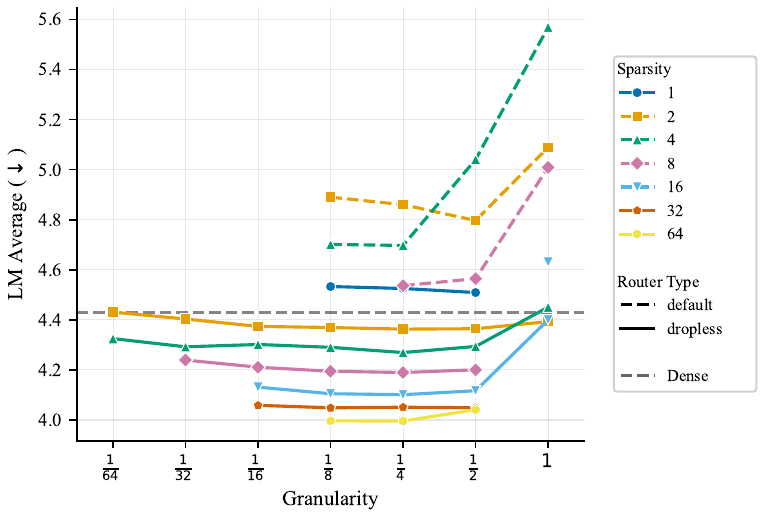}
        \caption{110M active, 110M - 1.4B total parameters}
    \end{subfigure}

    \caption{\textbf{Dropless routing outperforms default routing (\S\ref{sec:expt_router}).}
    We compare dropless routing to the default setting, which allow tokens to be dropped. Across all scales, we find that dropless routing outperforms or performs comparably to default routing. Additional model scales in Appendix \S\ref{app:expt_experts}.}
    \label{fig:dropless}
\end{figure*}

%% file: fig_tex/4_3_lb.tex
\begin{figure*}[!t]
    \centering
    \begin{subfigure}[]{\textwidth}
        \centering
        \includegraphics[width=0.46\linewidth]{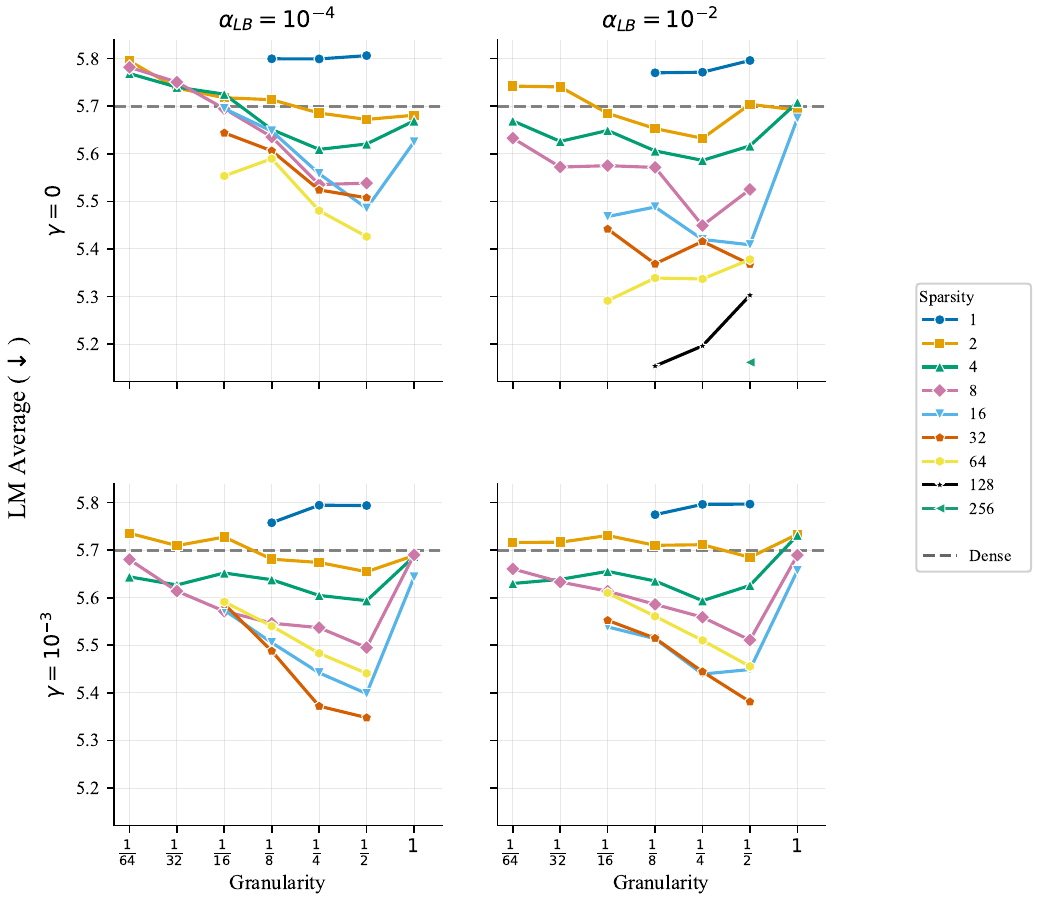}
        \hspace{1em}
        \includegraphics[width=0.46\linewidth]{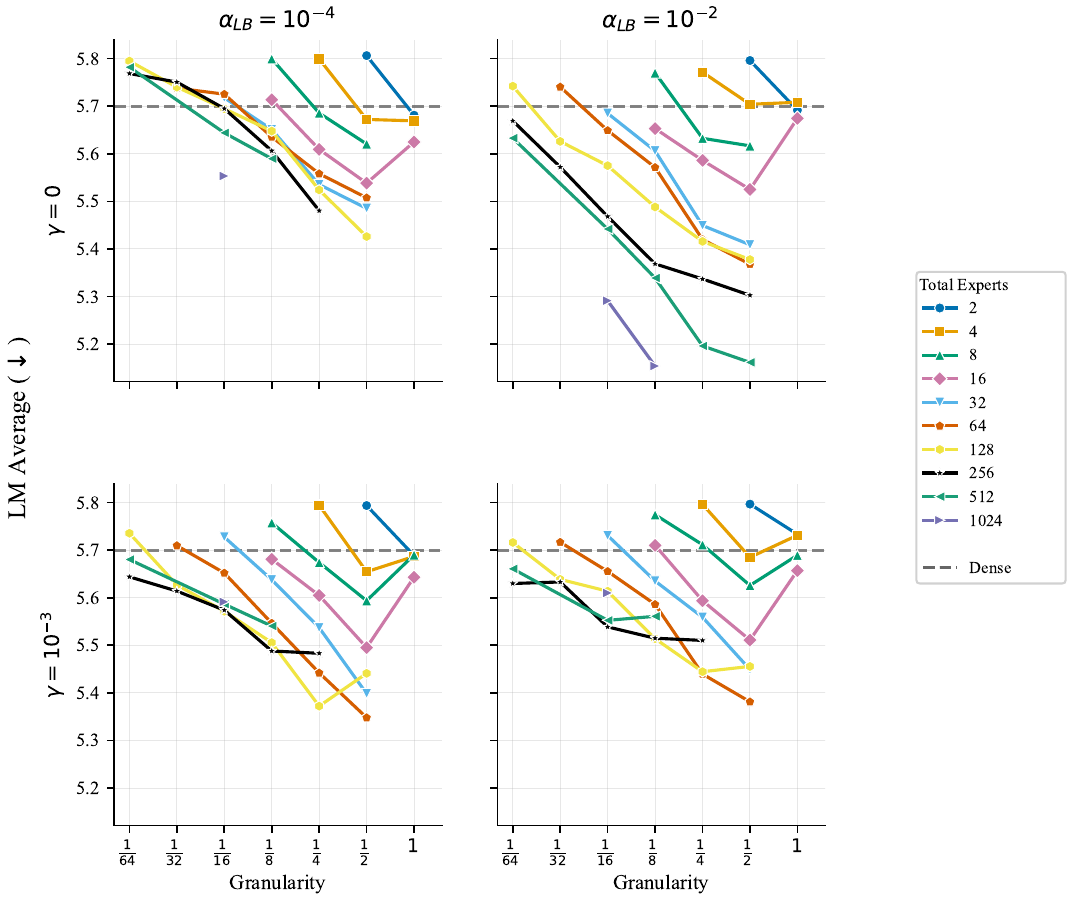}
        \caption{50M active, 50M - 930M total parameters}
    \end{subfigure}
    \caption{\textbf{Load balancing mechanisms must be tuned correctly (\S\ref{sec:expt_router}).}
    We consider load balancing loss weight $\alpha_{LB} \in \{\num{1e-2}, \num{1e-4}\}$ and loss-free load balancing with bias $\gamma\in\{0, \num{1e-3}\}$ ($\gamma=0$ indicates no loss-free mechanism). Results show that poorly chosen hyperparameters, such as high bias $\gamma = 1e-3$ with total experts $n\geq 512$, may impair performance. However, all settings other than $(\alpha_{LB}=\num{1e-2}, \gamma=\num{1e-3})$ perform comparably at $n \leq 512$, suggesting that a wide range of load balancing settings achieve near-optimal results. Additional model scales in Appendix \S\ref{app:expt_experts}.
    }
    \label{fig:lb}
\end{figure*}

%% file: tables/hyperparams.tex
\begin{table}[h]
    \centering
\begin{tabular}{lc}
\toprule
 Hyperparameter & Value \\ \midrule
Vocabulary size & 50K \\
Batch Size &    512	\\
Sequence Length & 2048	\\
Learning Rate & 4e-4  \\
Encoder-Decoder Weight Sharing & No \\
Feedforward Dimension & 4 x hidden dimension \\
LR Schedule & \textbf{Cosine Decay}, Warmup-Stable-Decay \\
LR Warmup & 50 steps \\
End LR & 0.1 x Peak LR \\
Optimizer & Adam $(\beta_1, \beta_2 = 0.9, 0.95)$ \\
Nonlinearity & SwiGLU \\
MoE Z-loss & 1e-3 \\
Load Balancing Loss Weight & \textbf{1e-2}, 1e-4 \\
Loss-Free Load Balancing Bias $\gamma$ & \textbf{0}, 1e-3 \\
MoE Token Dropping & \textbf{Dropless}, Default \\
MoE Routing Choice & Token Choice \\
\bottomrule
\end{tabular}
    \caption{
    Hyperparameter details for models trained in \S\ref{sec:expt}.  Multiple values indicate that we ablate over those values in \S\ref{sec:expt_router} or additional results shown in the appendix. Values in bold are used in \S\ref{sec:expt_main}-\ref{sec:expt_hetgen}}
\label{tab:hparams} 
\end{table}

%% file: tables/param_counts.tex
\begin{table}[!ht]
    \centering
    \footnotesize
\begin{tabular}{lcccccccccc}
\toprule
\multirow{3}{*}{\textbf{Name}} & \multirow{3}{*}{\textbf{Layers}} &  \multirow{3}{*}{\shortstack[c]{\textbf{Model} \\ \textbf{Dim}}}& \multirow{3}{*}{\shortstack[c]{\textbf{Attention}\\ \textbf{Heads}}} & \textbf{MoE} & \textbf{Active} & \textbf{Total} & \multirow{3}{*}{\shortstack[c]{\textbf{Active} \\ \textbf{$N$}}} & \multirow{3}{*}{\shortstack[c]{\textbf{Total} \\ \textbf{$N$}}}
\\
& & & & \textbf{Activation} & \textbf{Non-embedding} & \textbf{Non-embedding} &  & \\
& & & & \textbf{Sparsity} & \textbf{$N$} & \textbf{$N$} &  & \\
\midrule
10M     & 3     & 48   & 3     & 1     &	110.9K     & 110.9K    & 9.7M  & 9.7M \\
&	&	&	& 2 	&	110.9K	&	193.9K	&	9.7M	&	9.8M	\\
&	&	&	& 4 	&	110.9K	&	359.8K	&	9.7M	&	10.0M	\\
&	&	&	& 8 	&	110.9K	&	691.5K	&	9.7M	&	10.3M	\\
&	&	&	& 16	&	110.9K	&	1.4M	&	9.7M	&	11.0M	\\
&	&	&	& 32 	&	110.9K	&	2.7M	&	9.7M	&	12.3M	\\
&	&	&	& 64	&	110.9K	&	5.3M	&	9.7M	&	15.0M	\\
\midrule
20M     & 4     & 96   & 4     & 1     & 590.7K & 590.7K    & 19.8M	& 19.8M \\
&	&	&	& 2 	&	590.7K	&	1.0M	&	19.8M	&	20.3M	\\
&	&	&	& 4 	&	590.7K	&	1.9M	&	19.8M	&	21.2M	\\
&	&	&	& 8 	&	590.7K	&	3.7M	&	19.8M	&	22.9M	\\
&	&	&	& 16 	&	590.7K	&	7.2M	&	19.8M	&	26.5M	\\
&	&	&	& 32 	&	590.7K	&	14.3M	&	19.8M	&	33.6M	\\
&	&	&	& 64 	&	590.7K	&	28.5M	&	19.8M	&	47.7M	\\
\midrule
50M     & 5     & 240   & 6     & 1     & 4.6M	 & 4.6M  & 52.7M     & 52.7M \\
&	&	&	& 2 	&	4.6M	&	8.1M	&	52.7M	&	56.2M	\\
&	&	&	& 4 	&	4.6M	&	15.0M	&	52.7M	&	63.1M	\\
&	&	&	& 8  	&	4.6M	&	28.8M	&	52.7M	&	76.9M	\\
&	&	&	& 16 	&	4.6M	&	56.5M	&	52.7M	&	104.6M	\\
&	&	&	& 32 	&	4.6M	&	111.7M	&	52.7M	&	159.9M	\\
&	&	&	& 64 	&	4.6M	&	222.3M	&	52.7M	&	270.5M	\\
&	&	&	& 128	&	4.6M	&	443.5M	&	52.7M	&	491.7M	\\
&	&	&	& 256	&	4.6M	&	885.9M	&	52.7M	&	934.0M	\\
\midrule
80M     & 8 	& 336	& 7	     & 1	&	14.5M	&	14.5M	&	81.8M	&	81.8M	\\	
&	&	&	& 2       &	14.5M	&	25.3M	&	81.8M	&	92.7M	\\	
&	&	&	& 4       &	14.5M	&	47.0M	&	81.8M	&	114.4M	\\	
&	&	&	& 8       &	14.5M	&	90.3M	&	81.8M	&	157.7M	\\	
&	&	&	& 16      &	14.5M	&	177.0M	&	81.8M	&	244.4M	\\	
&	&	&	& 32	 &	14.5M	&	350.4M	&	81.8M	&	417.8M	\\	
&	&	&	& 64	&	14.5M	&	697.3M	&	81.8M	&	764.6M	\\	
\midrule
110M     & 9     & 432   & 9     & 1     &	26.9M	&	26.9M	&	113.5M	&	113.5M		\\
&	&	&	& 2      &	26.9M	&	47.0M	&	113.5M	&	133.7M		\\
&	&	&	& 4      &	26.9M	&	87.3M	&	113.5M	&	174.0M		\\
&	&	&	&  8     &	26.9M	&	168.0M	&	113.5M	&	254.6M		\\
&	&	&	&  16     &	26.9M	&	329.2M	&	113.5M	&	415.9M		\\
&	&	&	&  32     &	26.9M	&	651.7M	&	113.5M	&	738.3M		\\
&	&	&	&  64     &	26.9M	&	1.3B	&	113.5M	&	1.4B		\\
\midrule
200M     & 10    & 640   & 10    & 1    &	65.5M	&	65.5M	&	193.9M	&	193.9M		\\
&	&	&	& 2      &	65.5M	&	114.7M	&	193.9M	&	243.1M		\\
&	&	&	&  4     &	65.5M	&	213.0M	&	193.9M	&	341.4M		\\
&	&	&	& 8      &	65.5M	&	409.6M	&	193.9M	&	538.0M		\\
&	&	&	& 16      &	65.5M	&	802.8M	&	193.9M	&	931.2M		\\
&	&	&	& 32      &	65.5M	&	1.6B	&	193.9M	&	1.7B		\\
&	&	&	& 64    &	65.5M	&	3.2B	&	193.9M	&	3.3B	\\
\midrule
300M &	12	& 832	& 13	 &      1     &	132.9M	&	132.9M	&	299.8M	&	299.8M		\\
&	&	&	&  2     &	132.9M	&	232.6M	&	299.8M	&	399.5M		\\
&	&	&	&  4     &	132.9M	&	432.0M	&	299.8M	&	598.8M		\\
&	&	&	&      8       &	132.9M	&	830.7M	&	299.8M	&	997.6M	\\
&	&	&	& 16     &	132.9M	&	1.6B	&	299.8M	&	1.8B	\\
&	&	&	& 32     &	132.9M	&	3.2B	&	299.8M	&	3.4B	\\
&	&	&	& 64     &	132.9M	&	6.4B	&	299.8M	&	6.6B	\\

\bottomrule
\end{tabular}
    \caption{
    \textbf{Architecture Details and Parameter Counts }
    }
\label{tab:param_counts} 
\end{table}

%% file: tables/hgn_configs.tex
\begin{figure*}
\center
\begin{tikzpicture}[scale=1]

    \def\activePairs{
        1/1, 2/0.5, 4/0.25, 8/0.125, 
        2/1, 4/0.5, 8/0.25, 16/0.125, 32/0.0625, 64/0.03125, 128/0.015625, 
        4/1, 8/0.5, 16/0.25, 32/0.125, 64/0.0625, 128/0.03125, 256/0.015625, 
        8/1, 16/0.5, 32/0.25, 64/0.125, 128/0.0625, 256/0.03125, 512/0.015625, 
        16/1, 32/0.5, 64/0.25, 128/0.125, 256/0.0625, 512/0.03125,
        64/0.5, 128/0.25, 256/0.125, 512/0.0625, 
        128/0.5, 256/0.25, 512/0.125, 1024/0.0625, 
        256/0.5, 512/0.25, 1024/0.125, 
        512/0.5
    }
    \def\granValues{1, 0.5, 0.25, 0.125, 0.0625, 0.03125, 0.015625}
    \def\expertValues{1, 2, 4, 8, 16, 32, 64, 128, 256, 512, 1024}

    \node[rotate=90, font=\bfseries] at (-1.5, 3.5) {Granularity ($g$)};
    \node[font=\bfseries] at (5.5, -1.5) {Number of Experts ($N$)};

    \draw[step=1cm, gray!20, very thin] (0,0) grid (11,7);

    \foreach \gv [count=\yi] in \granValues {
        \foreach \ev [count=\xi] in \expertValues {
            
            \foreach \activeE/\activeG in \activePairs {
                \pgfmathparse{(\ev == \activeE && \gv == \activeG) ? 1 : 0}
                
                \ifnum\pgfmathresult=1
                    \pgfmathsetmacro{\result}{\ev * \gv}
                    \pgfmathsetmacro{\myshade}{ln(\result+1.2)*18}
                    
                    \fill[blue!\myshade!white] (\xi-1, 7-\yi) rectangle (\xi, 8-\yi);
                    
                    \node[font=\tiny] at (\xi-0.5, 7.5-\yi) {
                        \pgfmathprintnumber[fixed, precision=2, zerofill=false]{\result}
                    };
                \fi
            }
        }
    }

    \draw[thick] (0,0) rectangle (11,7);

    \foreach \ylab [count=\i] in {1/64, 1/32, 1/16, 1/8, 1/4, 1/2, 1} {
        \node[anchor=east] at (-0.2, \i-0.5) {$\ylab$};
    }

    \foreach \xlab [count=\i] in {1, 2, 4, 8, 16, 32, 64, 128, 256, 512, 1024} {
        \node[rotate=45, anchor=north east] at (\i-0.5, -0.2) {\xlab};
    }
\end{tikzpicture}
\caption{\textbf{Homogeneous MoE configurations represented in \S\ref{sec:expt} Figure~\ref{fig:moe_count_and_granularity}}}
\label{fig:hgn_configs}
\end{figure*}

%% file: tables/het_configs.tex
\begin{table}[!ht]
    \centering
\begin{tabular}{cccc}
\toprule
\textbf{Granularities}   & \textbf{Active Expert Counts}    &  \textbf{Total Expert Counts}     & \textbf{Activation Sparsity}   \\
\textbf{$(g_1, g_2)$}   & \textbf{ $(k_1, k_2)$}    &  \textbf{$(n_1, n_2)$}     & \textbf{$(s)$}   \\
\midrule
($\frac{1}{2}, \frac{1}{4}$)    & (1, 2)    & & \\
& & (4, 8) & 4 \\
& & (8, 16) & 8 \\
& & (16, 32) & 16 \\
& & (32, 64) & 32 \\
($\frac{1}{4}, \frac{1}{8}$)    & (2, 4)    &  & \\
& & (8, 16) & 4 \\
& & (16, 32) & 8\\
& & (32, 64) & 16 \\
& & (64, 128) & 32 \\
($\frac{1}{8}, \frac{1}{16}$)   & (4, 8)    &   & \\
& & (16, 32) & 4 \\
& & (32, 64) & 8 \\
& & (64, 128) & 16 \\
& & (128, 256) & 32 \\
($\frac{1}{16}, \frac{1}{32}$)  & (8, 16) &  & \\
& & (16, 32) & 2 \\
& & (32, 64) & 4 \\
& & (64, 128) & 8 \\
& & (128, 256) & 16 \\
\bottomrule
\end{tabular}
    \caption{\textbf{Heterogeneous MoE configurations (\S\ref{sec:expt_hetgen}).}. We show the configurations used in the heterogeneous MoE experiments. 
    }
\label{tab:het_configs} 
\end{table}

%% file: tables/random_seeds.tex
\begin{table}[!ht]
    \centering
    \small
\begin{tabular}{llcccc}
\toprule
& \multirow{2}{*}{\textbf{Dataset}}   & \multicolumn{2}{c}{\textbf{Dense}}   & \multicolumn{2}{c}{\textbf{MoE ($g=\frac{1}{2},n=64$)}}    \\
& & CE Loss (Mean) & Std. Dev. & CE Loss (Mean) & Std. Dev. \\
\midrule
\parbox[t]{2mm}{\multirow{12}{*}{\rotatebox[origin=c]{90}{\centering Language Modeling}}} & \textsc{C4}	 & 5.27 & .01 & 5.57 & .01 \\
& \textsc{Dolma} Books	 & 5.55 & .01 & 5.82 & .01 \\
& \textsc{Dolma} common-crawl	 & 5.30 & .01 & 5.60 & .01 \\
& \textsc{Dolma} pes2o	 & 5.06 & .02 & 5.41 & .01 \\
& \textsc{Dolma} reddit	 & 4.99 & .00 & 5.26 & .00 \\
& \textsc{Dolma} stack	 & 5.90 & .06 & 6.27 & .05 \\
& \textsc{Dolma} wiki	 & 5.19 & .02 & 5.52 & .01 \\
& \textsc{ICE}	 & 5.58 & .02 & 5.90 & .02 \\
& \textsc{M2D2 S2ORC}	 & 5.24 & .01 & 5.49 & .01 \\
& \textsc{The Pile}	 & 5.27 & .03 & 5.59 & .02 \\
& \textsc{WikiText-103} 	 & 5.85 & .03 & 6.28 & .03 \\
\cmidrule{2-6}
& Average LM	 & 5.38 & .02 & 5.70 & .02 \\
\midrule
\parbox[t]{2mm}{\multirow{7}{*}{\rotatebox[origin=c]{90}{\centering Downstream}}} & BoolQ	 & 3.39 & .17 & 3.40 & .08 \\
& HellaSwag	 & 1.04 & .00 & 1.11 & .00 \\
& MMLU Humanities	 & 4.81 & .34 & 4.92 & .37 \\
& MMLU Other	 & 4.09 & .30 & 3.93 & .20 \\
& MMLU Social Sciences (5-shot)	 & 4.32 & .43 & 4.39 & .17 \\
& MMLU STEM (5-shot)	 & 3.97 & .30 & 4.02 & .20 \\
\cmidrule{2-6}
& Average Downstream	 & 3.60 & .24 & 3.63 & .16 \\
\bottomrule
\end{tabular}
    \caption{\textbf{Random Seed Sensitivity (\S\ref{sec:expt}).}
    Means and standard deviations of cross-entropy loss on all language modeling and downstream tasks, across 5 random seeds at 50M active parameter scale.
    }
\label{tab:random_seeds} 
\end{table}

%% file: fig_tex/hist_plots.tex
\begin{figure*}[ht]
    \centering
    \begin{subfigure}[t]{\textwidth}
        \centering
        \includegraphics[width=\linewidth]{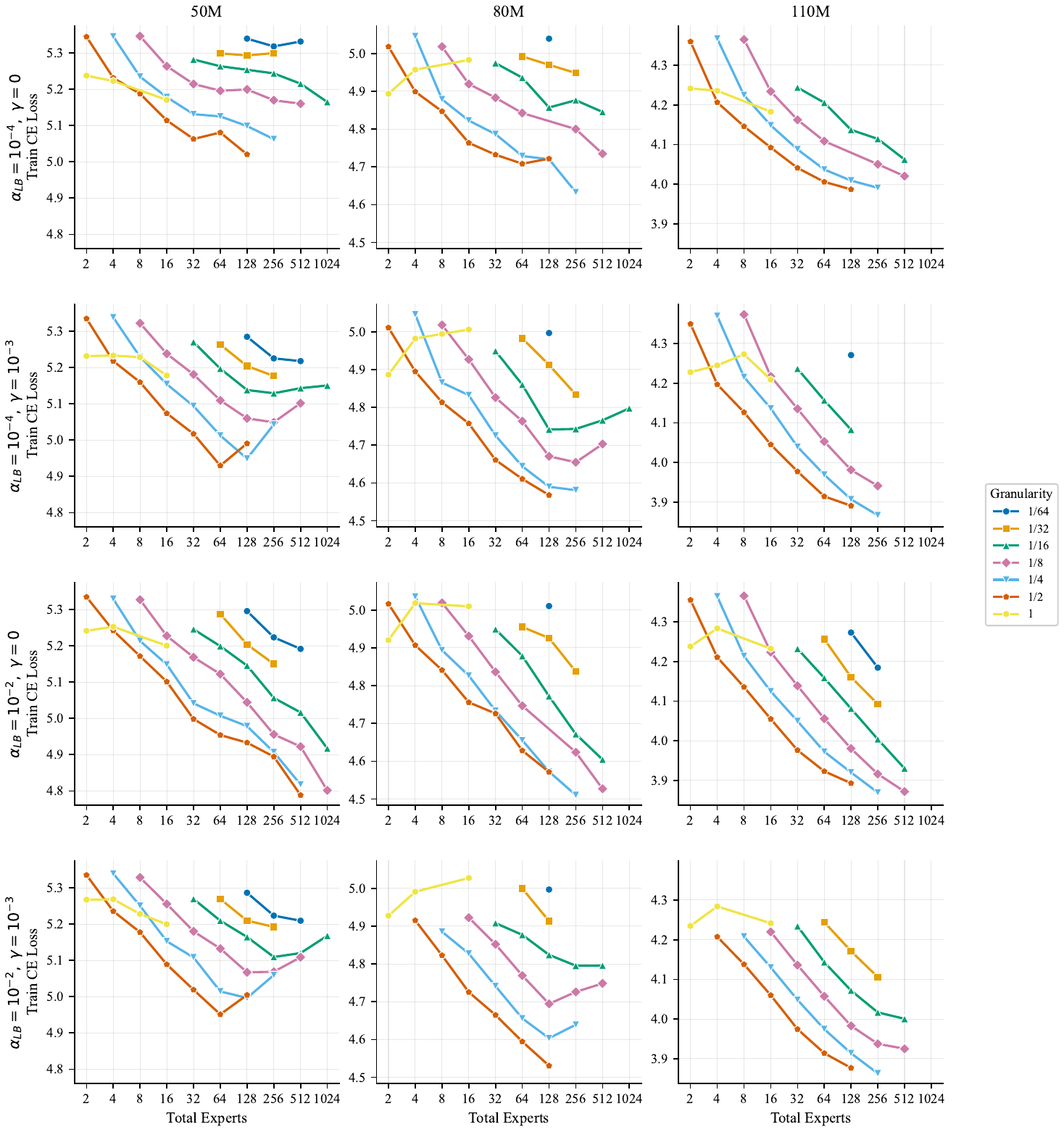}
        \caption{50M - 110M active, 50M - 1.4B total parameters}
    \end{subfigure}
\end{figure*}

\clearpage  

\begin{figure*}[ht]
    \addtocounter{figure}{-1}
    \centering
    \begin{subfigure}[t]{\textwidth}
        \addtocounter{subfigure}{1}
        \centering
        \includegraphics[width=0.7\linewidth]{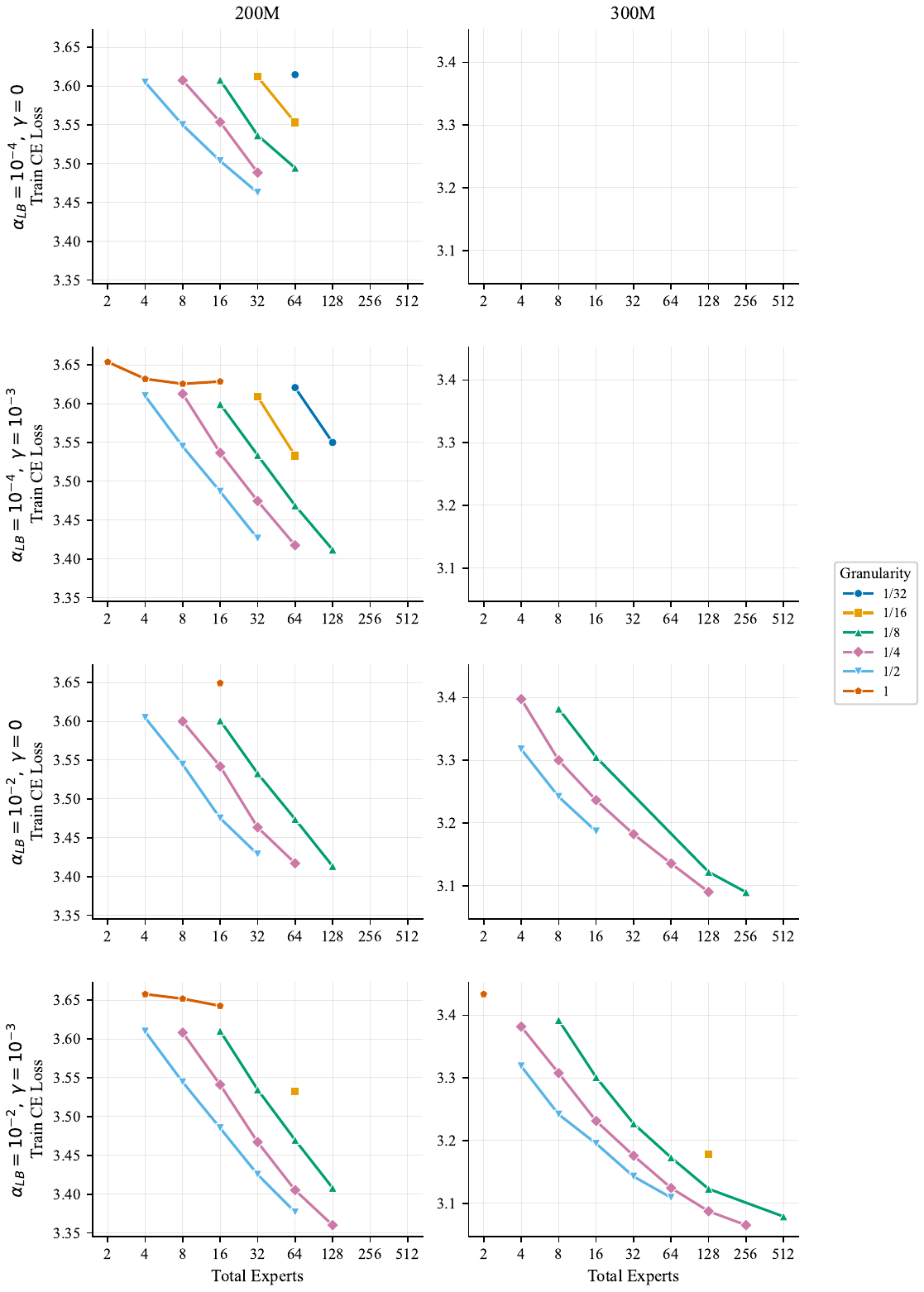}
        \caption{200M - 300M active, 200M - 6.6B total parameters}
    \end{subfigure}
    \caption{
    \textbf{Train Cross-Entropy Loss (\S\ref{sec:expt_main}).} We plot the train cross-entropy loss, averaged over the final 50 steps, at each of 5 active parameter model sizes. For each active parameter count (column), we show all four load balancing settings 
    with $(\alpha_{LB} \in \{\num{1e-4}, \num{1e-2}\}, \gamma \in \{0, \num{1e-3}\})$. The trends seen in cross-entropy loss on train data closely follow those seen in validation data cross-entropy loss (Figure~\ref{fig:lb}, Figure~\ref{fig:lm_avg_lb}).
    }
    \label{fig:train_loss}
\end{figure*}

\begin{figure*}[ht]
    \centering
        \begin{subfigure}[t]{\textwidth}
        \centering
        \includegraphics[width=\linewidth]{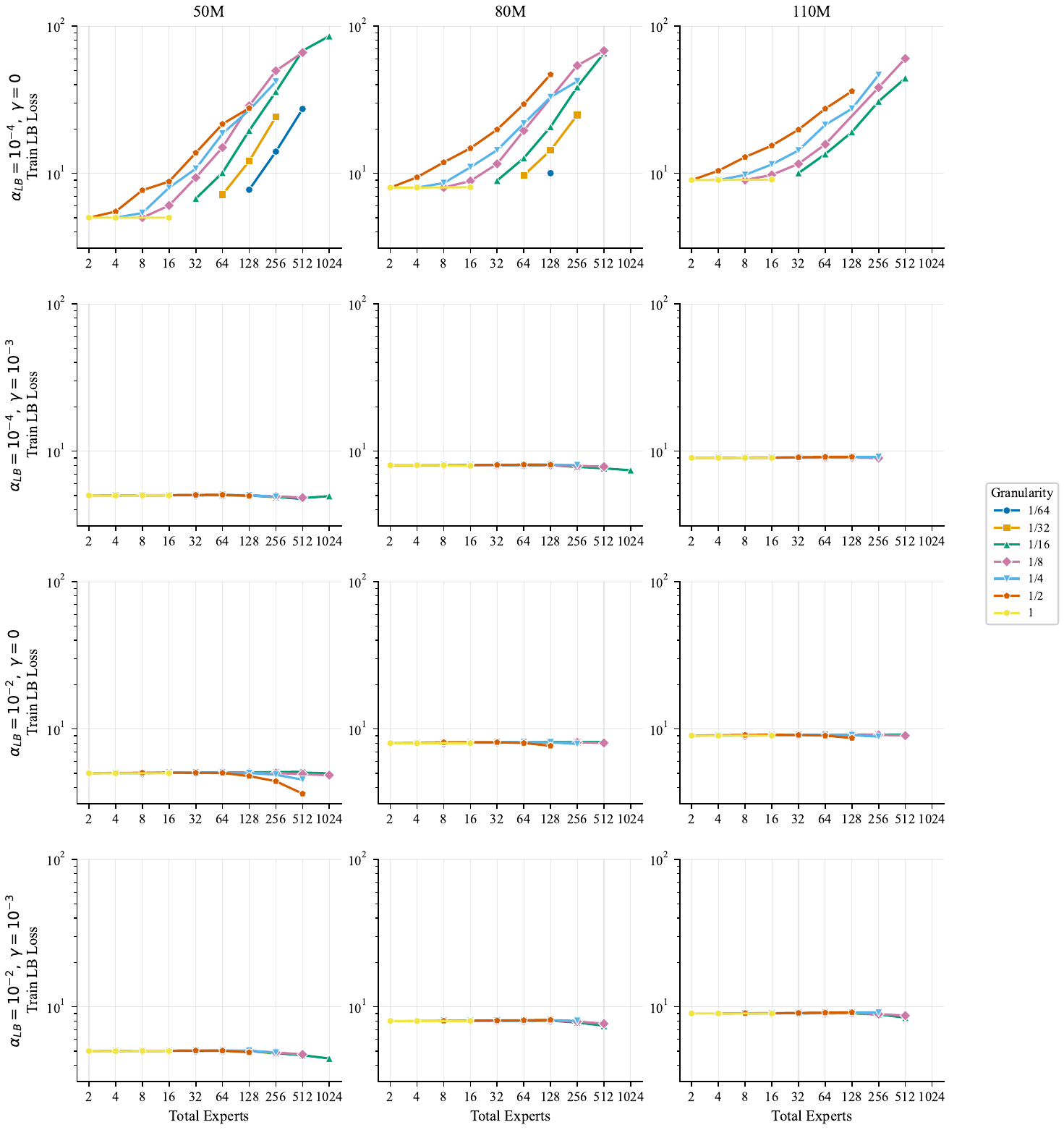}
        \caption{50M - 110M active, 50M - 1.4B total parameters}
    \end{subfigure}
\end{figure*}

\clearpage  

\begin{figure*}[ht]
    \addtocounter{figure}{-1}
    \centering
    \begin{subfigure}[t]{\textwidth}
        \addtocounter{subfigure}{1}
        \centering
        \includegraphics[width=0.7\linewidth]{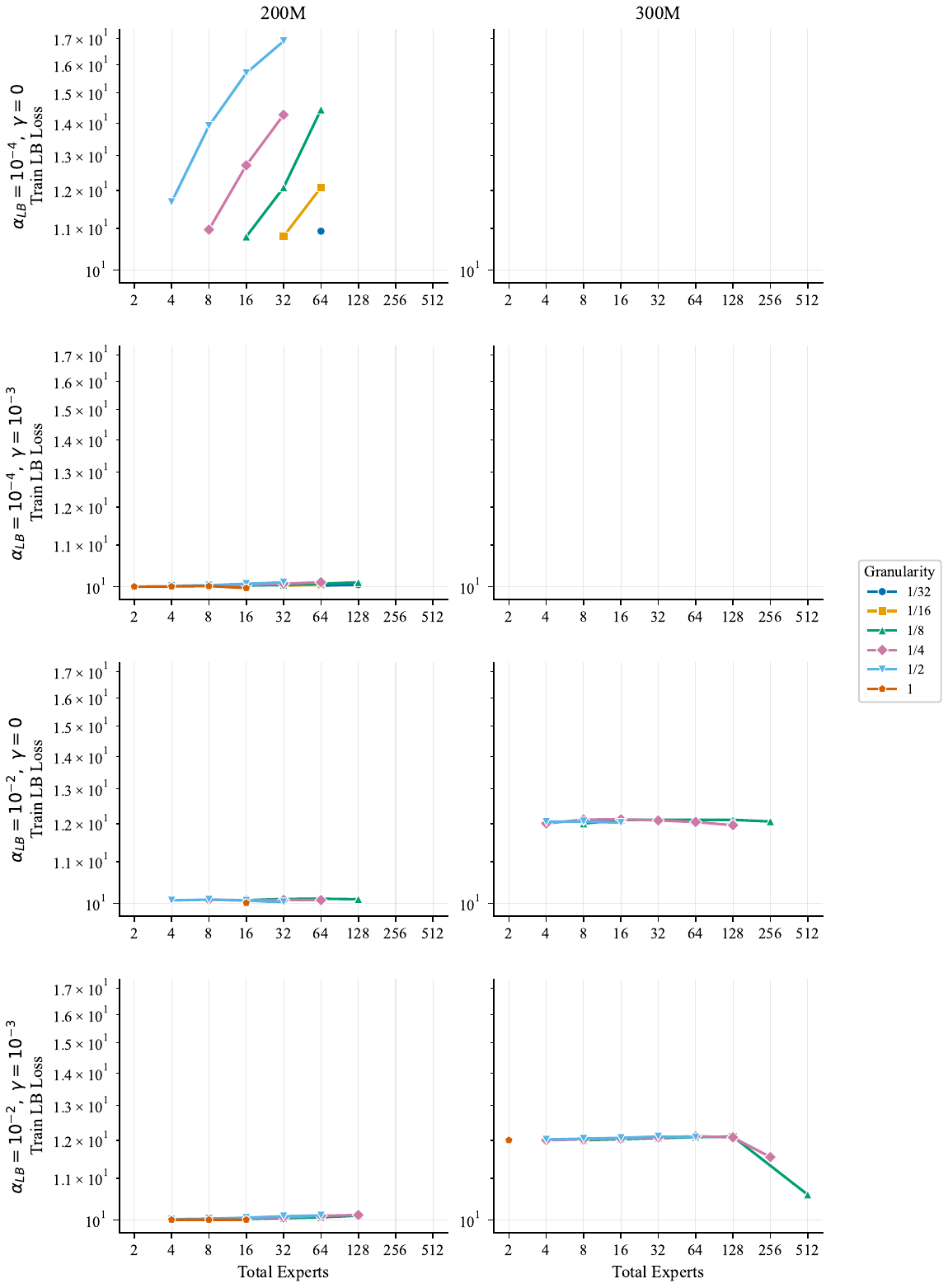}
        \caption{200M - 300M active, 200M - 6.6B total parameters}
    \end{subfigure}
    \caption{
    \textbf{Load Balancing Loss (\S\ref{sec:expt_router}).} We plot the train load balancing loss as defined in \S\ref{sec:bg}, averaged over the final 50 steps, at each of 5 active parameter model sizes. For each active parameter count (column), we show all four load balancing settings 
    with $(\alpha_{LB} \in \{\num{1e-4}, \num{1e-2}\}, \gamma \in \{0, \num{1e-3}\})$. Across all model scales, $(\alpha_{LB} = \num{1e-4}, \gamma  = 0)$ results in higher load balancing loss overall.
    }
    \label{fig:lb_loss}
\end{figure*}

\begin{figure*}[ht]
    \centering
        \begin{subfigure}[t]{\textwidth}
        \centering
        \includegraphics[width=\linewidth]{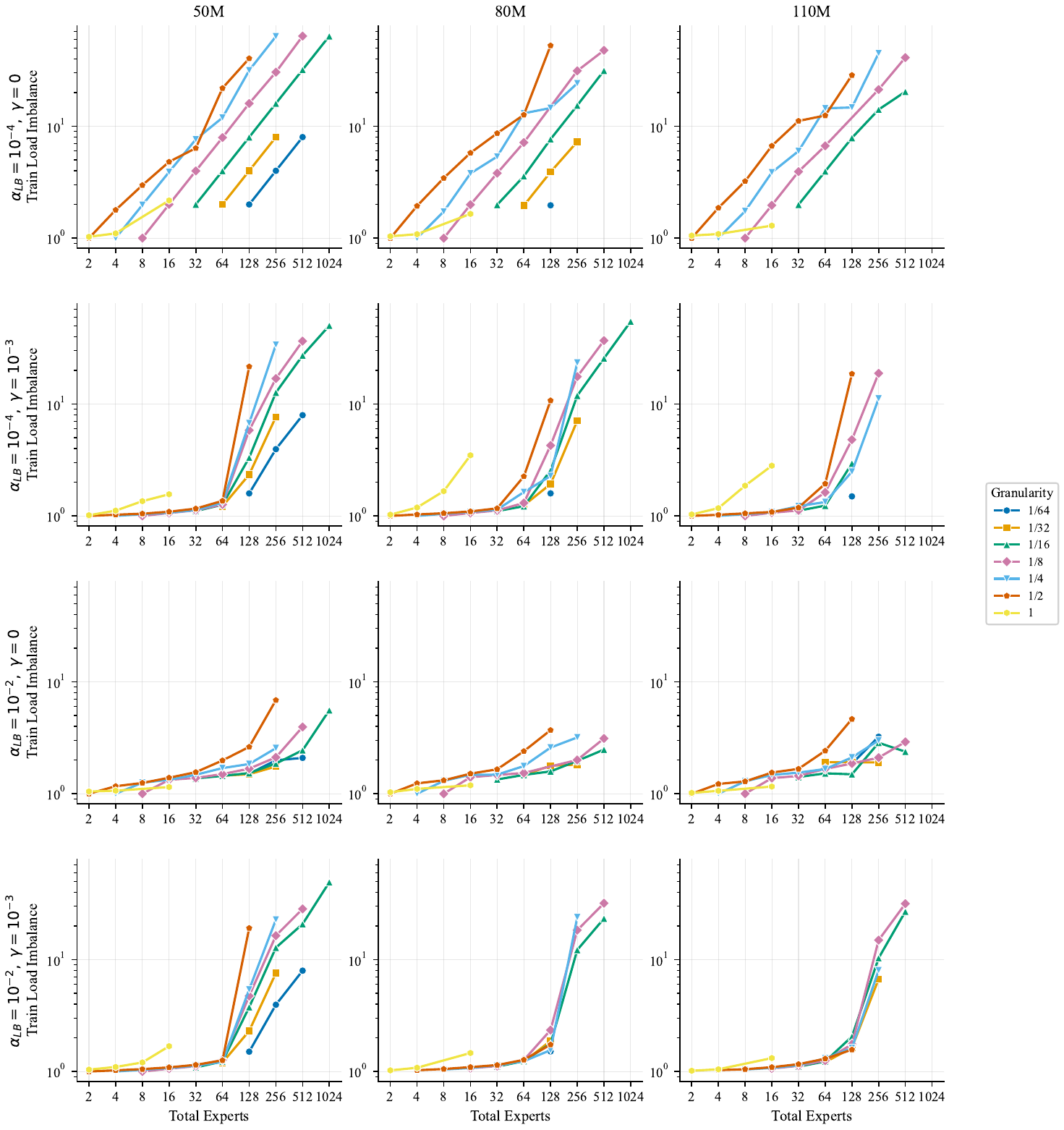}
        \caption{50M - 110M active, 50M - 1.4B total parameters}
    \end{subfigure}
\end{figure*}

\clearpage 

\begin{figure*}[ht]
    \addtocounter{figure}{-1}
    \centering
    \begin{subfigure}[t]{\textwidth}
        \addtocounter{subfigure}{1}
        \centering
        \includegraphics[width=0.7\linewidth]{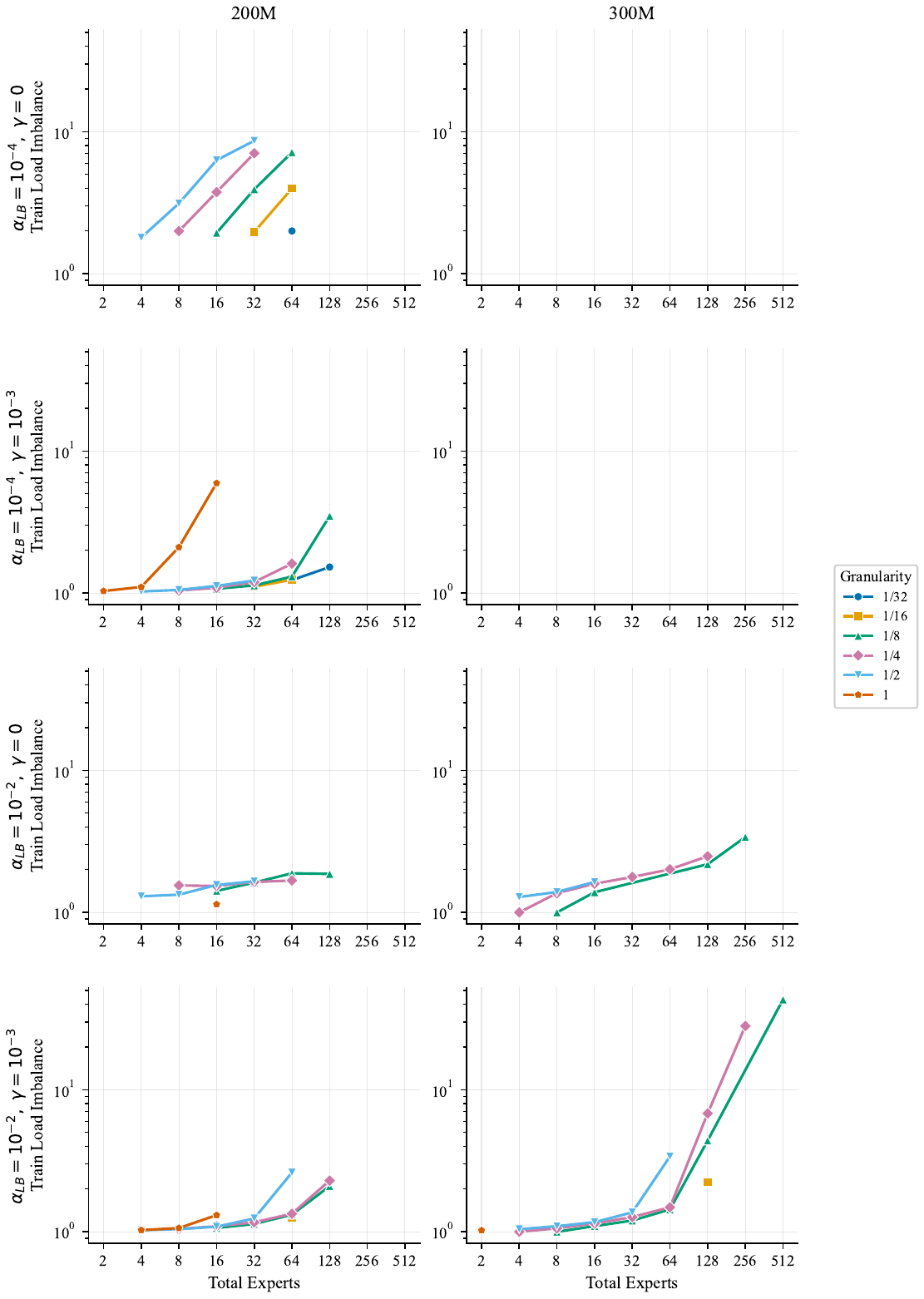}
        \caption{200M - 300M active, 200M - 6.6B total parameters}
    \end{subfigure}
    \caption{
    \textbf{Load Imbalance (\S\ref{sec:expt_router}).} We define load imbalance as the ratio between the maximum and mean expert loads in a batch. We plot the train load imbalance, averaged over the final 50 steps, at each of 5 active parameter model sizes. For each active parameter count (column), we show all four load balancing settings 
    with $(\alpha_{LB} \in \{\num{1e-4}, \num{1e-2}\}, \gamma \in \{0, \num{1e-3}\})$. Across all model scales, $(\alpha_{LB} = \num{1e-4}, \gamma  = 0)$ results in higher imbalance overall, and using the loss free load balancing mechanism with $\gamma = \num{1e-3}$ results in drastically increased load imbalance as total expert count increases.
    }
    \label{fig:load_imbalance}
\end{figure*}

%% file: fig_tex/10M20M.tex
\begin{figure*}[ht]
    \centering
    \begin{subfigure}[t]{\textwidth}
        \begin{subfigure}[t]{0.33\textwidth}
            \centering
            \caption*{\scriptsize Fixed total experts (n)}
            \includegraphics[width=\linewidth]{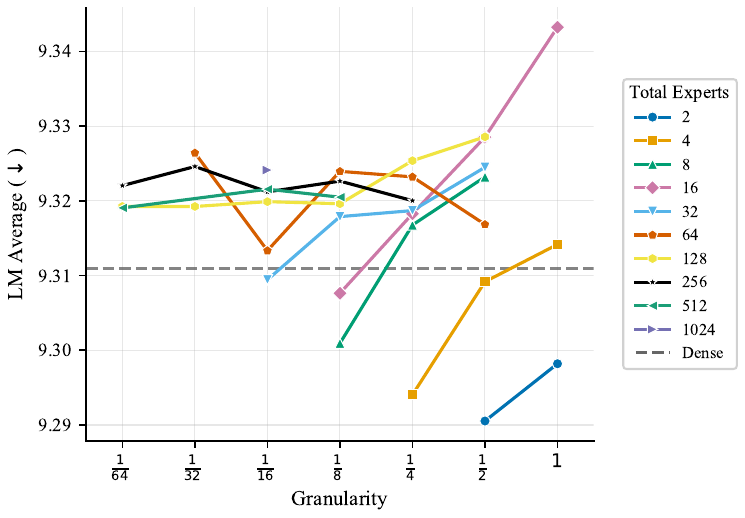}
        \end{subfigure}
        \begin{subfigure}[t]{0.33\textwidth}
            \centering
            \caption*{\scriptsize Fixed granularity (g)}
            
            \includegraphics[width=\linewidth]{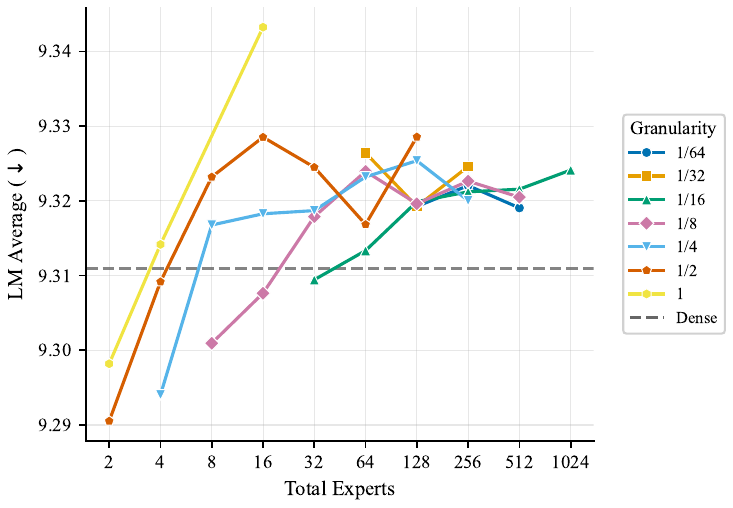}
        \end{subfigure}
        \begin{subfigure}[t]{0.33\textwidth}
            \centering
            \caption*{\scriptsize Fixed activation sparsity (s)}
            \includegraphics[width=\linewidth]{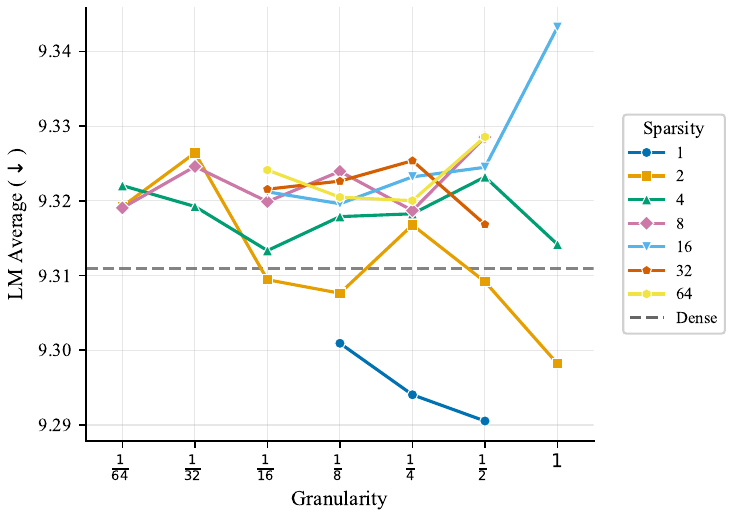}
        \end{subfigure}
        \caption{10M active, 10M - 15M total parameters}
    \end{subfigure}
    \par\bigskip
    \begin{subfigure}[t]{\textwidth}
        \begin{subfigure}[t]{0.33\textwidth}
            \centering
            \includegraphics[width=\linewidth]{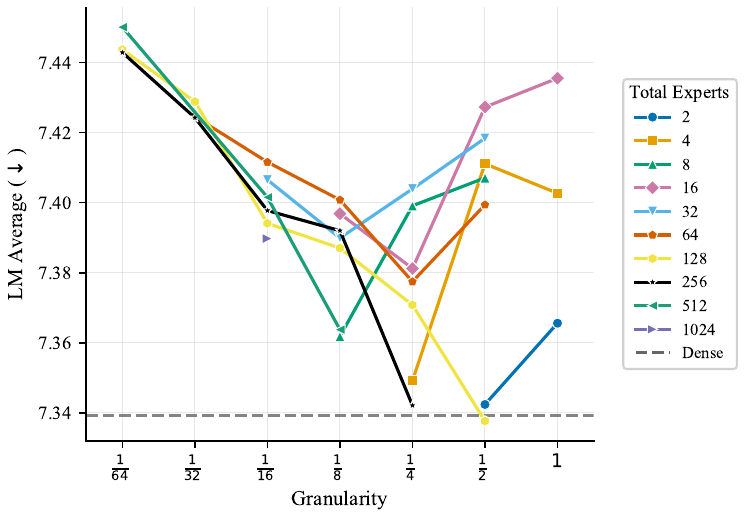}
        \end{subfigure}
        \begin{subfigure}[t]{0.33\textwidth}
            \centering
            \includegraphics[width=\linewidth]{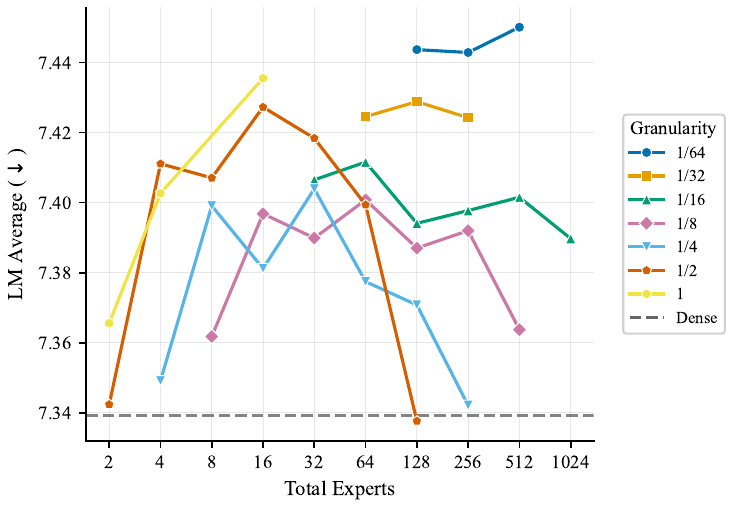}
        \end{subfigure}
        \begin{subfigure}[t]{0.33\textwidth}
            \centering
            \includegraphics[width=\linewidth]{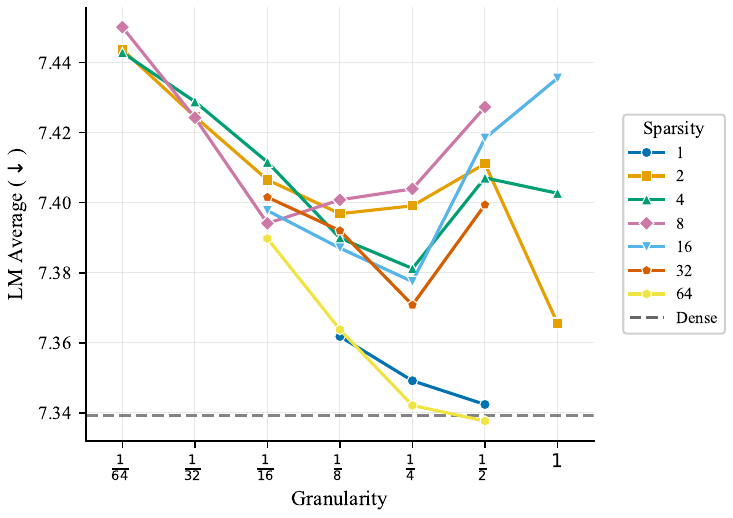}
        \end{subfigure}
        \caption{20M active, 20M - 48M total parameters}
    \end{subfigure}
    \caption{
    \textbf{At 10-20M scale, MoEs underperform dense baselines at all expert (count, granularity) configurations tested.} 
    }
    \label{fig:10M20M_experts}
\end{figure*}

\begin{figure*}[ht]
    \centering
    \begin{subfigure}[t]{\textwidth}
        \begin{subfigure}[t]{0.33\textwidth}
            \centering
            \caption*{\scriptsize Fixed total experts (n)}
            \includegraphics[width=\linewidth]{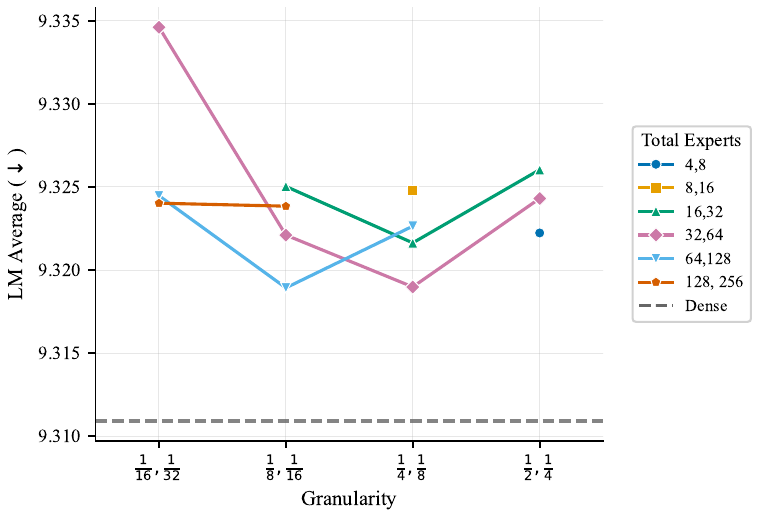}
        \end{subfigure}
        \begin{subfigure}[t]{0.33\textwidth}
            \centering
            \caption*{\scriptsize Fixed granularity (g)}
            
            \includegraphics[width=\linewidth]{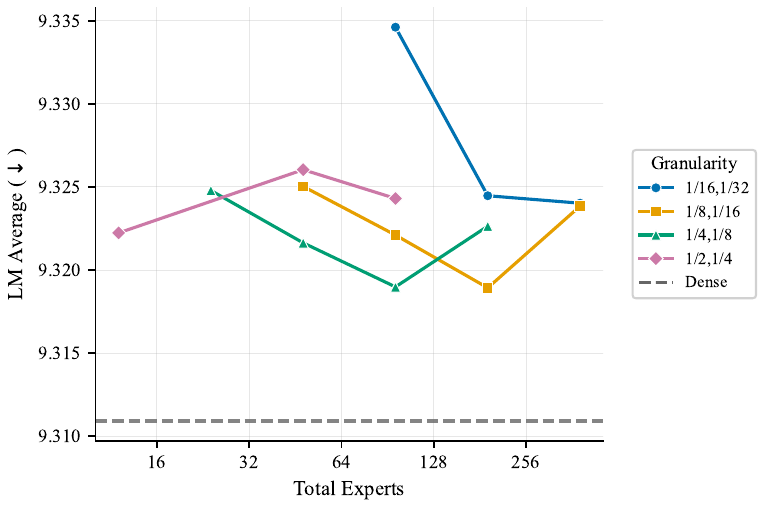}
        \end{subfigure}
        \begin{subfigure}[t]{0.33\textwidth}
            \centering
            \caption*{\scriptsize Fixed activation sparsity (s)}
            \includegraphics[width=\linewidth]{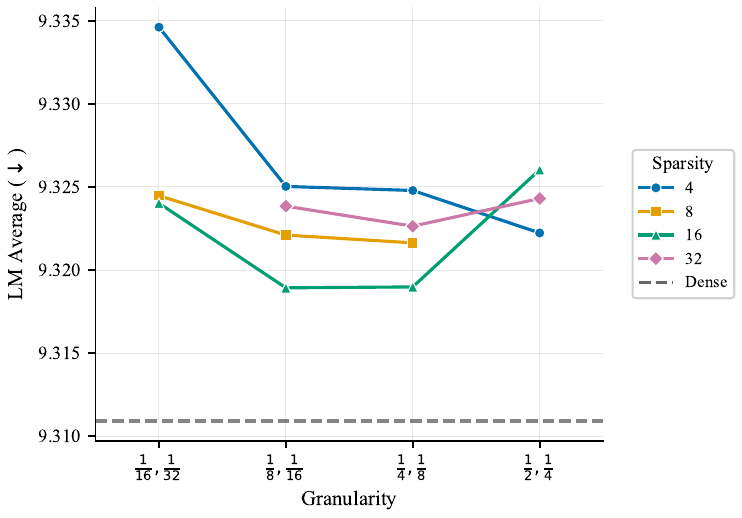}
        \end{subfigure}
        \caption{10M active, 10M - 15M total parameters}
    \end{subfigure}
    \par\bigskip
    \begin{subfigure}[t]{\textwidth}
        \begin{subfigure}[t]{0.33\textwidth}
            \centering
            \includegraphics[width=\linewidth]{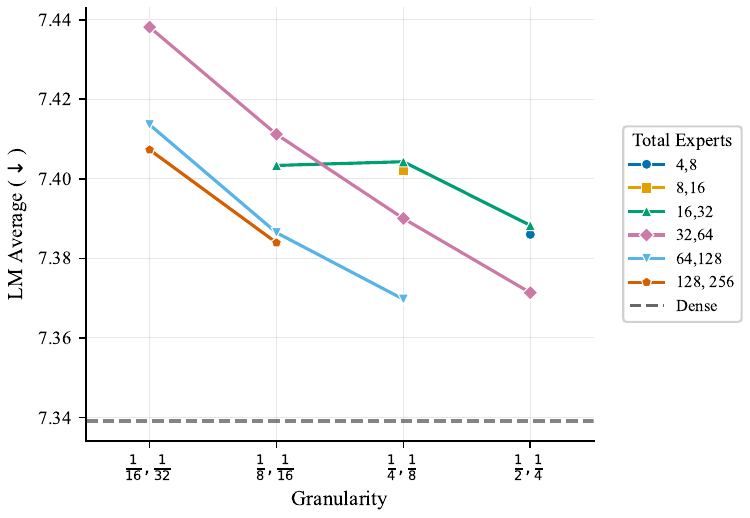}
        \end{subfigure}
        \begin{subfigure}[t]{0.33\textwidth}
            \centering
            \includegraphics[width=\linewidth]{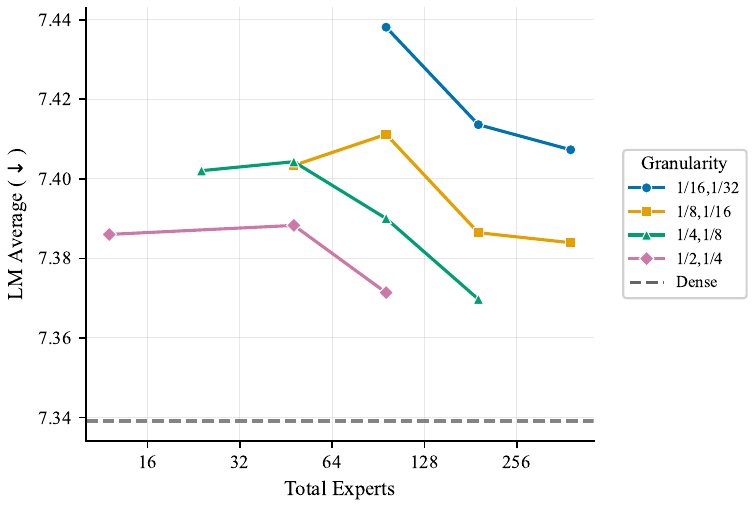}
        \end{subfigure}
        \begin{subfigure}[t]{0.33\textwidth}
            \centering
            \includegraphics[width=\linewidth]{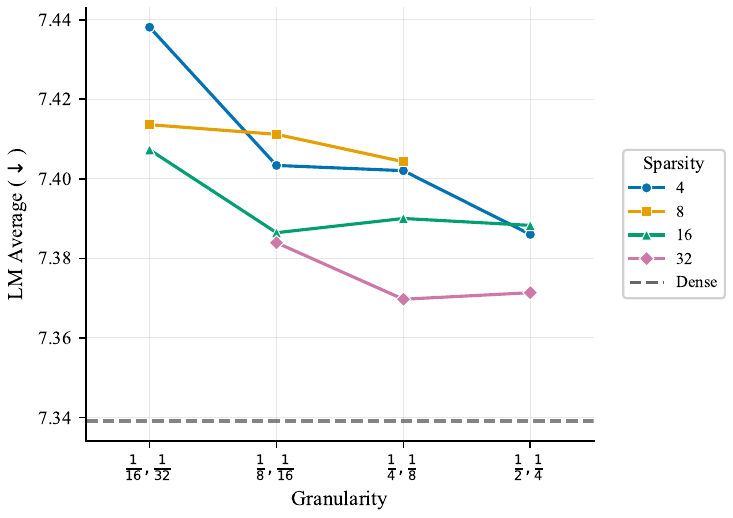}
        \end{subfigure}
        \caption{20M active, 20M - 48M total parameters}
    \end{subfigure}
    \caption{
    \textbf{At 10-20M scale, MoEs do not improve over dense baselines, regardless of expert heterogeneity.} 
    }
    \label{fig:10M20M_het}
\end{figure*}

\begin{figure*}[ht]
    \centering
    \begin{subfigure}[t]{1.0\textwidth}
        \centering
        \includegraphics[width=\linewidth]{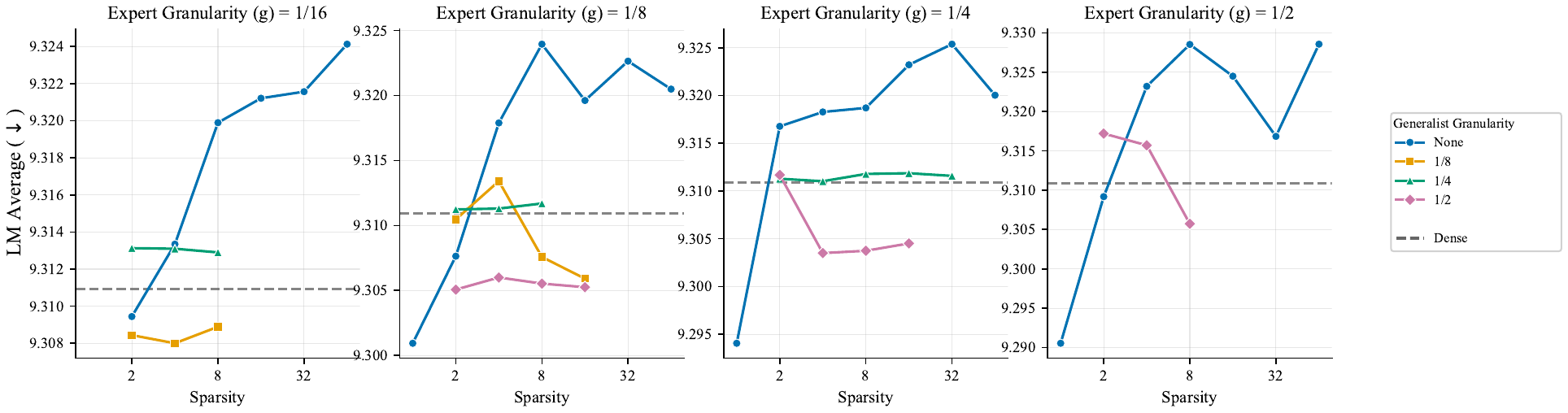}
        \caption{10M active, 10M - 15M total parameters}
    \end{subfigure}
    \par\bigskip
    \begin{subfigure}[t]{1.0\textwidth}
        \centering
        \includegraphics[width=\linewidth]{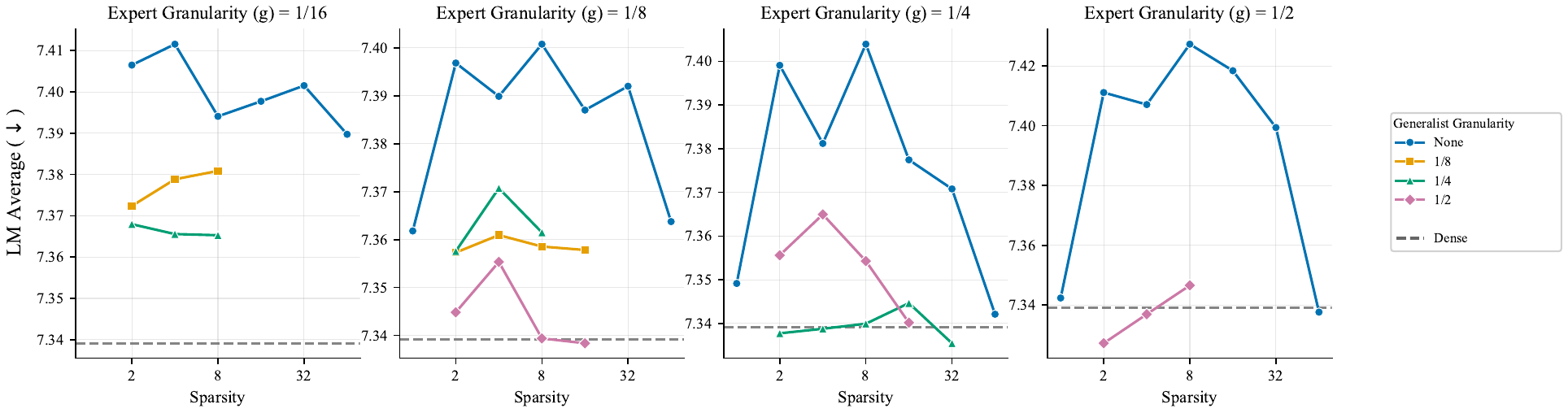}
        \caption{20M active, 20M - 48M total parameters}
    \end{subfigure}
    \caption{
    \textbf{At 10-20M scale, MoEs do not improve over dense baselines regardless of generalist inclusion.} 
    }
    \label{fig:10M20M_gen}
\end{figure*}

\begin{figure*}[ht]
    \centering
    \begin{subfigure}[t]{0.65\textwidth}
        \centering
        \includegraphics[width=\linewidth]{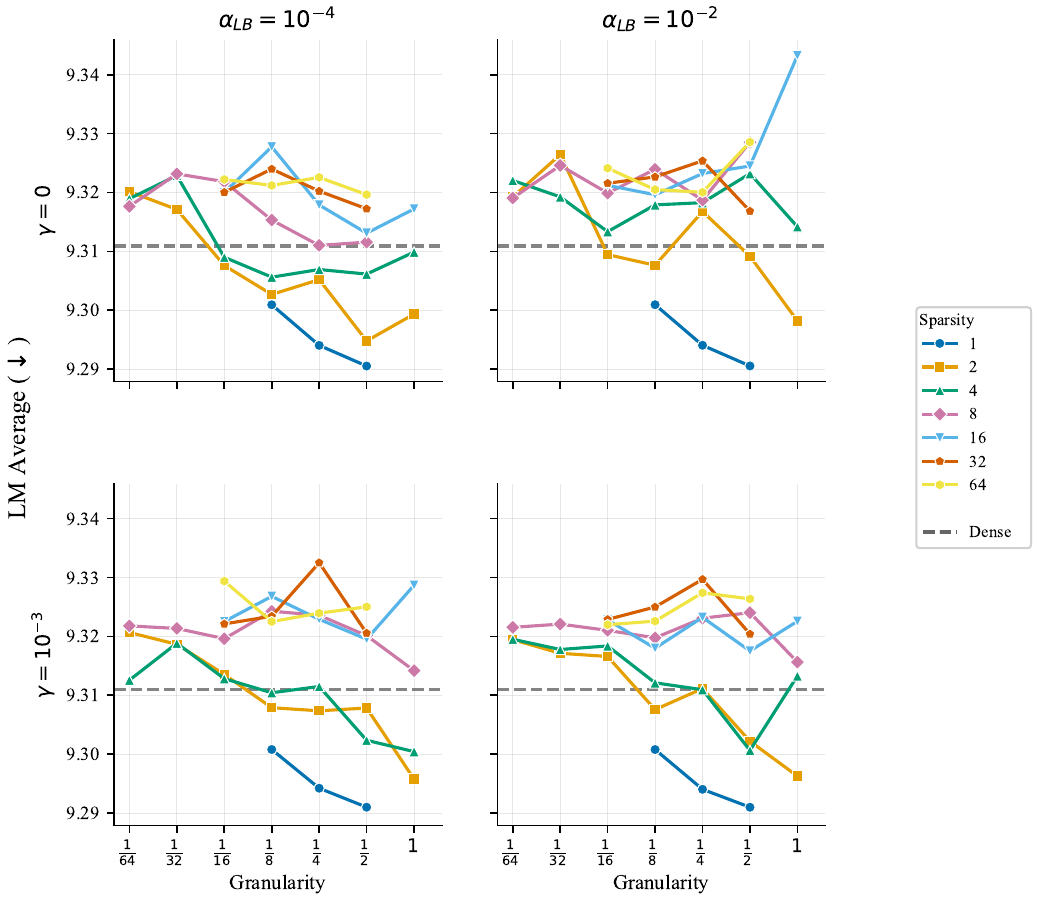}
        \caption{10M active, 10M - 15M total parameters}
    \end{subfigure}
    \par\bigskip
    \begin{subfigure}[t]{0.65\textwidth}
        \centering
        \includegraphics[width=\linewidth]{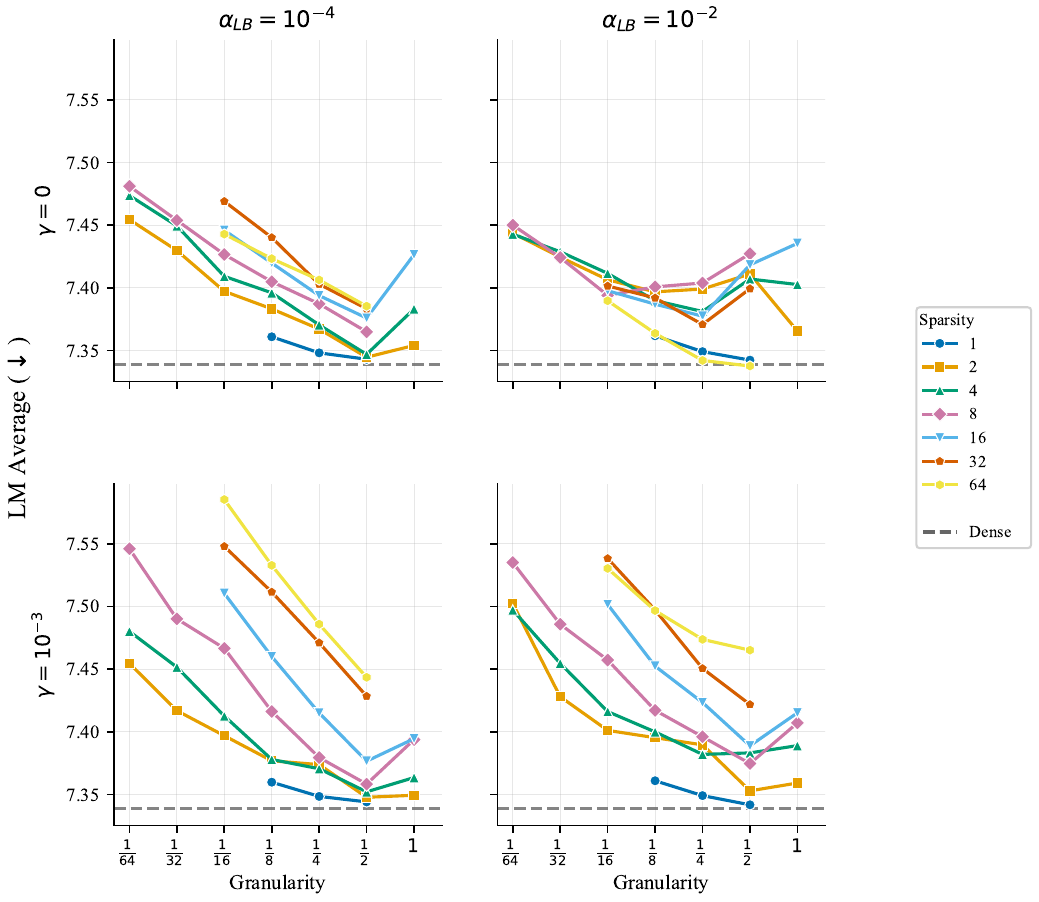}
        \caption{20M active, 20M - 48M total parameters}
    \end{subfigure}
    \caption{
    \textbf{At 10-20M scale, MoEs do not improve over dense baselines regardless of load balancing mechanisms.} 
    }
    \label{fig:10M20M_lb}
\end{figure*}

\begin{figure*}[ht]
    \centering
    \begin{subfigure}[t]{0.7\textwidth}
        \centering
        \includegraphics[width=\linewidth]{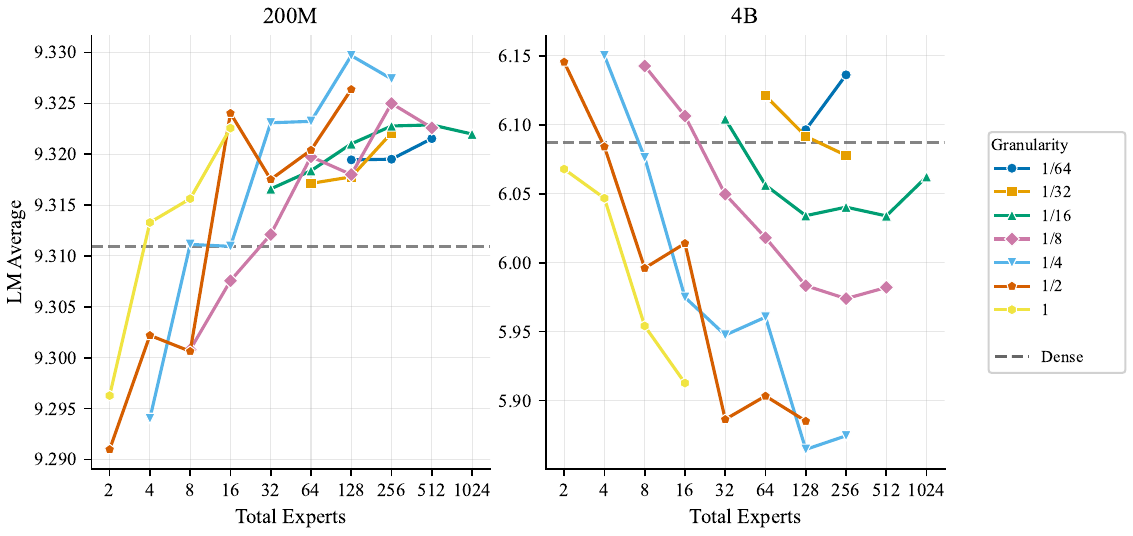}
    \end{subfigure}
    \caption{
    \textbf{MoEs outperform dense models, even at 10M active (10M - 15M total) parameters, given sufficient compute.} At 10M active parameter scale, MoE LMs underperform dense counterparts, if pretrained with Chinchilla-optimal tokens (200M total tokens). However, when pretraining with a 20 times greater data budget (4B total tokens), MoEs outperform dense models, exhibiting trends more similar to those seen in our 50M models, which are trained to a comparable compute budget.
    }
    \label{fig:10M_more_data}
\end{figure*}

\begin{figure*}[ht]
    \centering
    \begin{subfigure}[t]{0.7\textwidth}
        \centering
        \includegraphics[width=\linewidth]{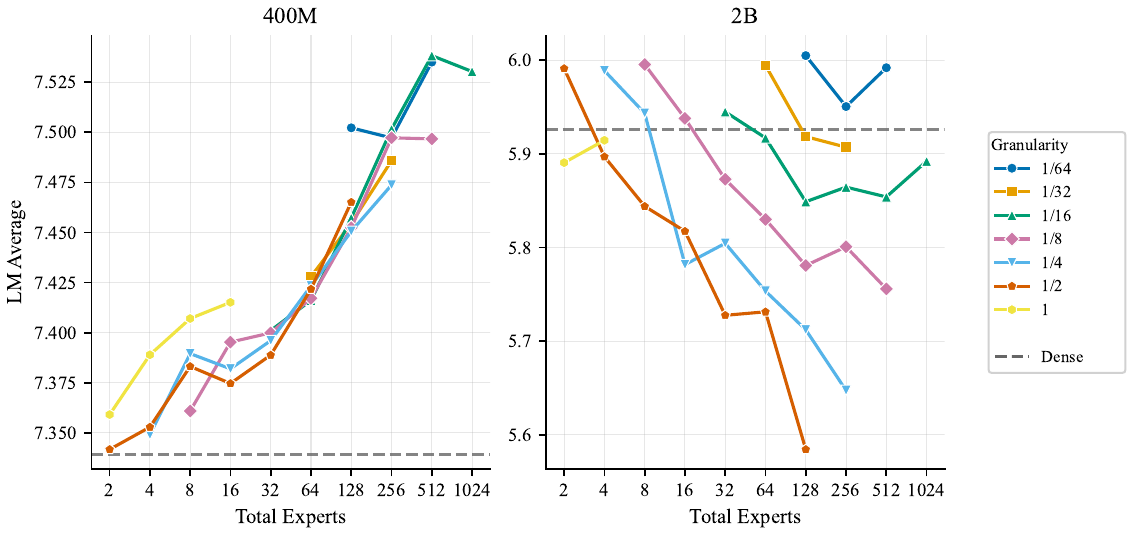}
    \end{subfigure}
    \caption{
    \textbf{MoEs outperform dense models, even at 20M active (20M - 48M total) parameters, given sufficient compute.} At 20M active parameter scale, MoE LMs underperform dense counterparts, if pretrained with Chinchilla-optimal tokens (400M total tokens). However, when pretraining with a 5 times greater data budget (2B total tokens), MoEs outperform dense models, exhibiting trends more similar to those seen in our 50M models, which are trained to a comparable compute budget.
    }
    \label{fig:20M_more_data}
\end{figure*}

%% file: fig_tex/lm_avg.tex
\begin{figure*}[!ht]
    \centering
        \begin{subfigure}[t]{\textwidth}
        \begin{subfigure}[t]{0.33\textwidth}
            \centering
            \caption*{\scriptsize Fixed total experts (n)}
            \includegraphics[width=\linewidth]{figures/lm_avg/hgn_gxn_50M_off_0.01.pdf}
        \end{subfigure}
        \begin{subfigure}[t]{0.33\textwidth}
            \centering
            \caption*{\scriptsize Fixed granularity (g)}
            \includegraphics[width=\linewidth]{figures/lm_avg/hgn_nxg_50M_off_0.01.pdf}
        \end{subfigure}
        \begin{subfigure}[t]{0.33\textwidth}
            \centering
            \caption*{\scriptsize Fixed activation sparsity (s)}
            \includegraphics[width=\linewidth]{figures/lm_avg/hgn_gxs_50M_off_0.01.pdf}
        \end{subfigure}
        \caption{50M active, 50M - 930M total parameters}
    \end{subfigure}
\par\bigskip\bigskip
    \begin{subfigure}[t]{\textwidth}
        \begin{subfigure}[t]{0.33\textwidth}
            \centering
            \includegraphics[width=\linewidth]{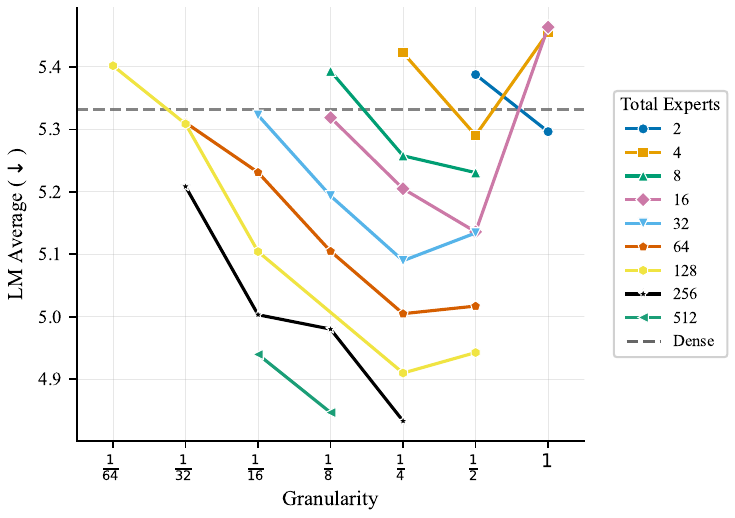}
        \end{subfigure}
        \begin{subfigure}[t]{0.33\textwidth}
            \centering
            \includegraphics[width=\linewidth]{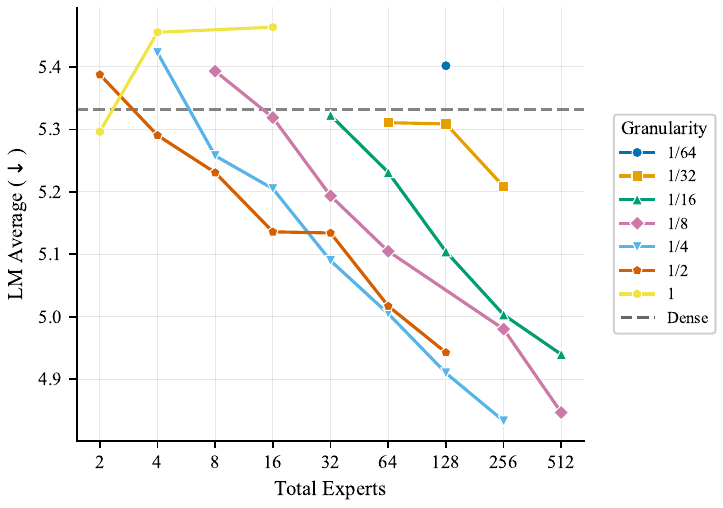}
        \end{subfigure}
        \begin{subfigure}[t]{0.33\textwidth}
            \centering
            \includegraphics[width=\linewidth]{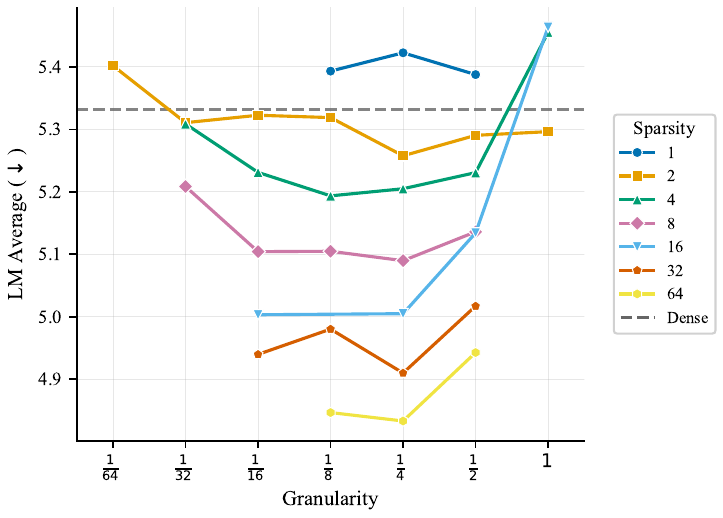}
        \end{subfigure}
        \caption{80M active, 80M - 765M total parameters}
    \end{subfigure}
    \par\bigskip\bigskip
        \begin{subfigure}[t]{\textwidth}
        \begin{subfigure}[t]{0.33\textwidth}
            \centering
            \includegraphics[width=\linewidth]{figures/lm_avg/hgn_gxn_110M_off_0.01.pdf}
        \end{subfigure}
        \begin{subfigure}[t]{0.33\textwidth}
            \centering
            \includegraphics[width=\linewidth]{figures/lm_avg/hgn_nxg_110M_off_0.01.pdf}
        \end{subfigure}
        \begin{subfigure}[t]{0.33\textwidth}
            \centering
            \includegraphics[width=\linewidth]{figures/lm_avg/hgn_gxs_110M_off_0.01.pdf}
        \end{subfigure}
        \caption{110M active, 110M - 1.4B total parameters}
    \end{subfigure}
    \end{figure*}

\clearpage

\begin{figure*}[!ht]
        \addtocounter{figure}{-1}
    \begin{subfigure}[t]{\textwidth}
        \addtocounter{subfigure}{3}
        \begin{subfigure}[t]{0.33\textwidth}
            \centering
            \caption*{\scriptsize Fixed total experts (n)}
            \includegraphics[width=\linewidth]{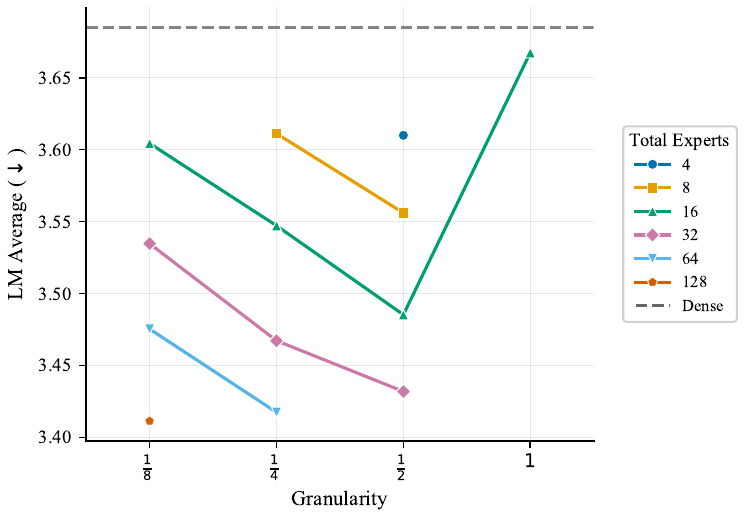}
        \end{subfigure}
        \begin{subfigure}[t]{0.33\textwidth}
            \centering
            \caption*{\scriptsize Fixed granularity (g)}
            \includegraphics[width=\linewidth]{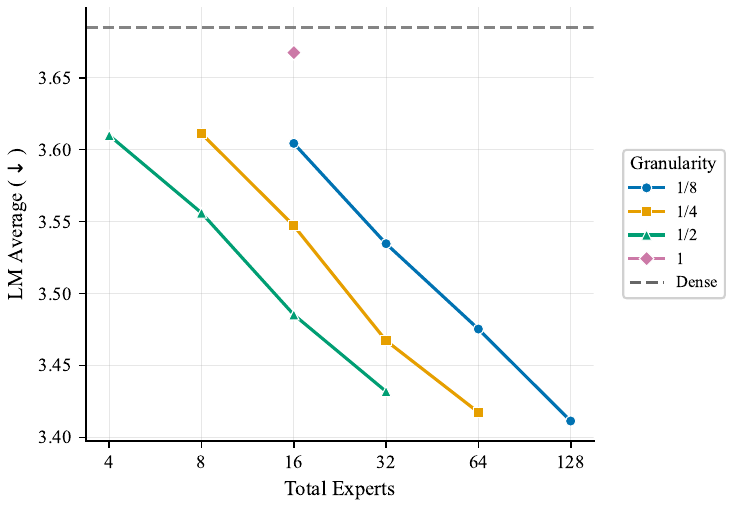}
        \end{subfigure}
        \begin{subfigure}[t]{0.33\textwidth}
            \centering
            \caption*{\scriptsize Fixed activation sparsity (s)}
            \includegraphics[width=\linewidth]{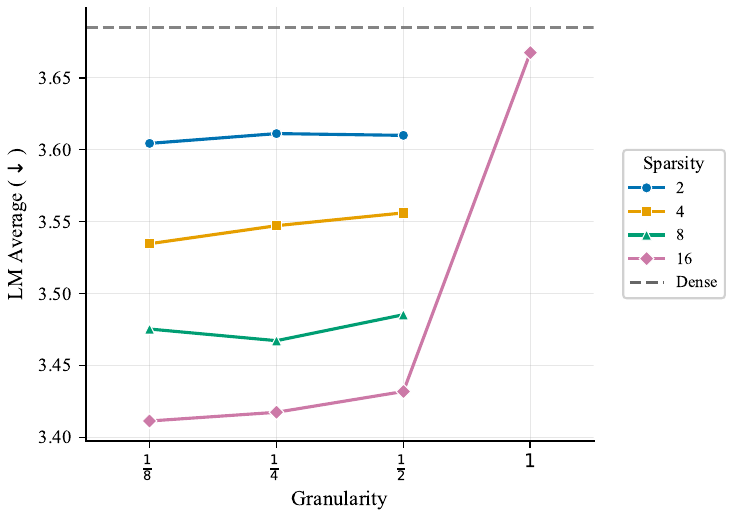}
        \end{subfigure}
        \caption{200M active, 200M - 3.3B total parameters}
    \end{subfigure}
    \par\bigskip\bigskip
        \begin{subfigure}[t]{\textwidth}
        \begin{subfigure}[t]{0.33\textwidth}
            \centering
            \includegraphics[width=\linewidth]{figures/lm_avg/hgn_gxn_300M_off_0.01.pdf}
        \end{subfigure}
        \begin{subfigure}[t]{0.33\textwidth}
            \centering
            \includegraphics[width=\linewidth]{figures/lm_avg/hgn_nxg_300M_off_0.01.pdf}
        \end{subfigure}
        \begin{subfigure}[t]{0.33\textwidth}
            \centering
            \includegraphics[width=\linewidth]{figures/lm_avg/hgn_gxs_300M_off_0.01.pdf}
        \end{subfigure}
        \caption{300M active, 300M - 6.6B total parameters}
    \end{subfigure}

    \caption{
    \textbf{Increasing inactive expert parameters via expert size (left) or total count (center) improves performance in MoEs (\S\ref{sec:expt_main}).} This effect is seen both when holding total number of experts fixed (left) and when holding expert granularity fixed (center). In general, increasing total parameters results in improved performance.  \textbf{Optimal tradeoff between expert count and granularity varies in MoEs (right). (\S\ref{sec:expt_main})}
    At each activation sparsity $s$ (equivalently, at each total parameter count), the optimal (total expert count, expert granularity) configuration varies. As $s$ increases, optimal expert granularity remains nearly fixed, suggesting that sparsity should be scaled up primarily by increasing total expert count $n$, while maintaining a near constant, slowly increasing expert granularity $g$. 
    }
    \label{fig:lm_avg_experts}
\end{figure*}

\begin{figure*}[!ht]
    \centering
    
    \begin{subfigure}[t]{0.46\textwidth}
        \centering
        \includegraphics[width=\linewidth]{figures/lm_avg/het_hgn_sxn_50M_off_0.01.pdf}
        \caption{50M active, 50M - 930M total parameters}
    \end{subfigure}
    \vspace{1em}
    \begin{subfigure}[t]{0.46\textwidth}
        \centering
        \includegraphics[width=\linewidth]{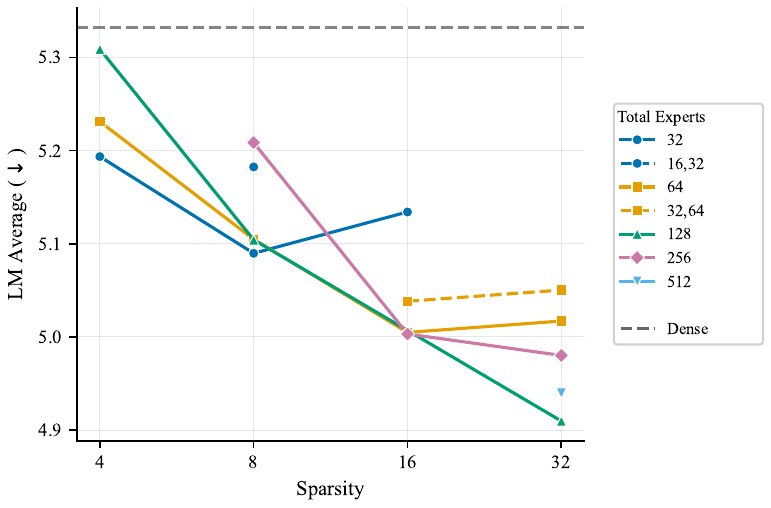}
        \caption{80M active, 80M - 765M total parameters}
    \end{subfigure}
    \caption{
    \textbf{Heterogeneity of expert size alone does not improve MoE performance (\S\ref{sec:expt_hetgen}).} To explore the potential benefits of their architectural flexibility, we compare heterogeneous MoEs (indicated by dotted lines) to active- and total-parameter-matched homogeneous MoEs. Heterogeneity alone does not result in performance gains, as, at each activation sparsity $s$, heterogeneous MoEs with $n_1, n_2 = a, b$ lie between or near the 2 closest homogeneous MoEs, with $n=a$ and with $n=b$.
    }
    \label{fig:lm_avg_het}
\end{figure*}

\begin{figure*}[!ht]
    \centering
    
    \begin{subfigure}[t]{1.0\textwidth}
        \centering
        \includegraphics[width=\linewidth]{figures/lm_avg/hgn_nxgenxs_50M_off_0.01.pdf}
        \caption{50M active, 50M - 930M total parameters}
    \end{subfigure}
    \par\bigskip\bigskip
    \begin{subfigure}[t]{1.0\textwidth}
        \centering
        \includegraphics[width=\linewidth]{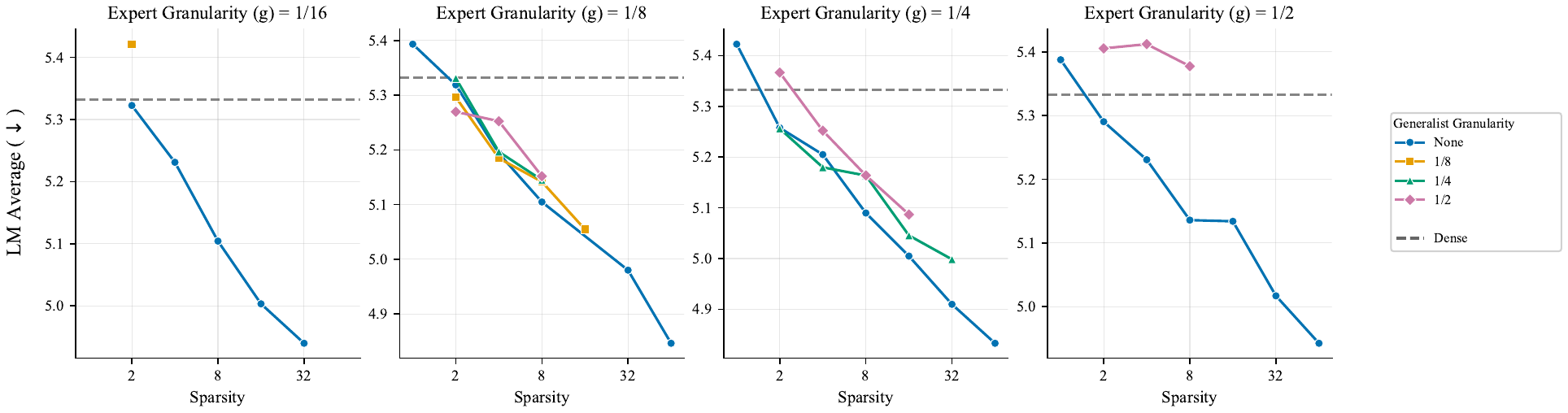}
        \caption{80M active, 80M - 765M total parameters}
    \end{subfigure}
    \par\bigskip\bigskip
    \begin{subfigure}[t]{1.0\textwidth}
        \centering
        \includegraphics[width=\linewidth]{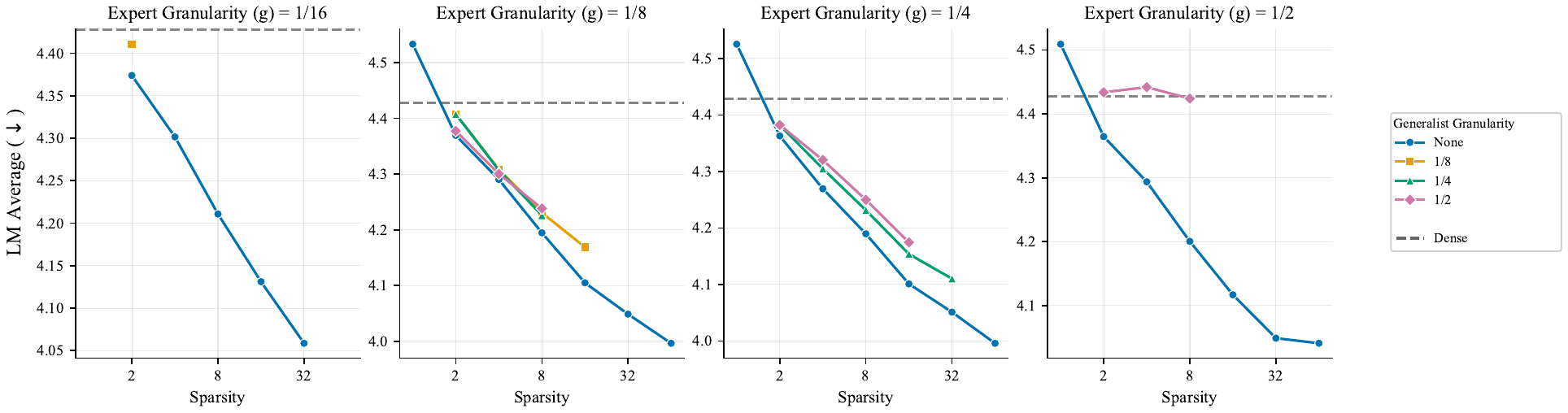}
        \caption{110M active, 110M - 1.4B total parameters}
    \end{subfigure}
    \caption{
    \textbf{The inclusion of a generalist consistently degrades performance in homogeneous MoEs (\S\ref{sec:expt_hetgen}).}
    We train MoE LMs which consist of some routed experts with granularity $g$, as well as a generalist with granularity $g_{gen}\in \{\frac{1}{2}, \frac{1}{4}, \frac{1}{8}\} $. We compare to settings with no generalist, only routed experts with granularity $g$. In all settings and configurations, the addition of any granularity generalist results in comparable or degraded performance. 
    }
    \label{fig:lm_avg_gen}
\end{figure*}

\begin{figure*}[ht]
    \centering
    \begin{subfigure}[t]{1.0\textwidth}
        \centering
        \includegraphics[width=\linewidth]{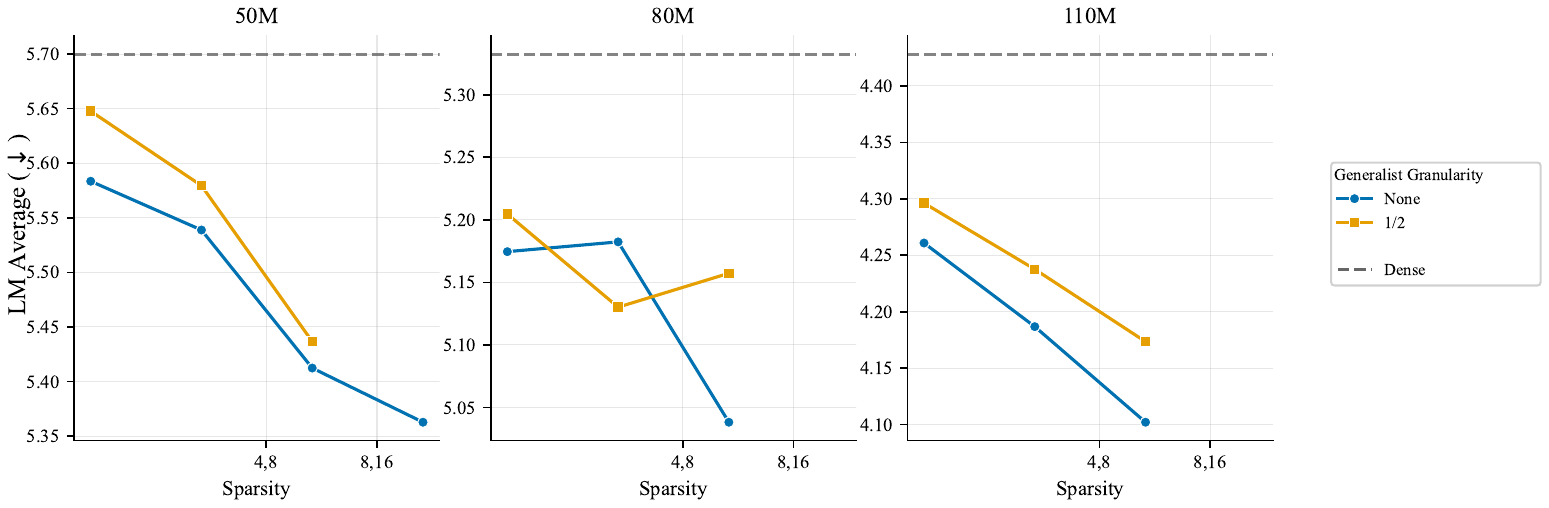}
    \end{subfigure}
    \caption{
    \textbf{The inclusion of a generalist consistently degrades performance in heterogeneous MoEs (\S\ref{sec:expt_hetgen}).}
    We train heterogeneous MoE LMs which consist of  routed experts with granularity $g_1, g_2$, as well as a generalist with granularity $g_{gen} = \frac{1}{2}$. We compare to settings with no generalist. In all settings and configurations, the addition of a generalist results in comparable or degraded performance. 
    }
    \label{fig:lm_avg_hetgen}
\end{figure*}

\begin{figure*}[ht]
    \centering
    \begin{subfigure}[t]{\textwidth}
        \centering
        \begin{subfigure}[t]{0.45\textwidth}
            \includegraphics[width=\linewidth]{figures/lm_avg/default_v_dropless_hgn_gxs_50M_off_0.01.pdf}
            \caption{50M active, 50M - 930M total parameters}
        \end{subfigure}
    \hspace{1em}
        \begin{subfigure}[t]{0.45\textwidth}
            \centering
            \includegraphics[width=\linewidth]{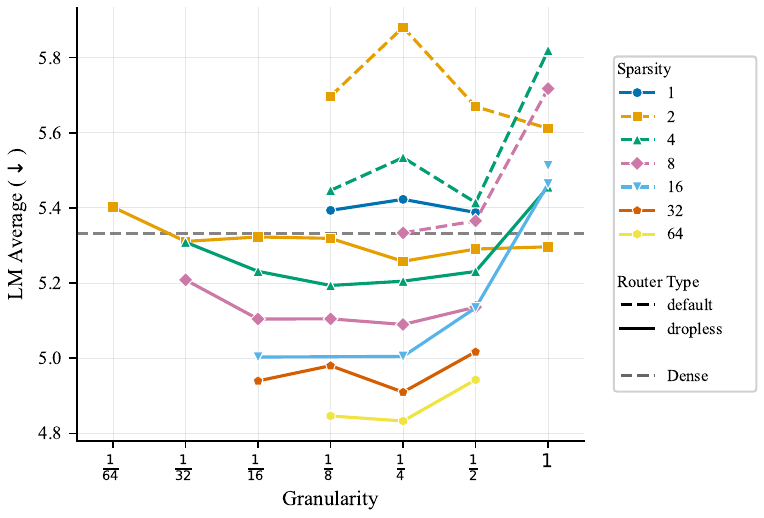}
            \caption{80M active, 80M - 765M total parameters}
        \end{subfigure}
    \end{subfigure}

    \par\bigskip\bigskip
    \begin{subfigure}[t]{0.45\textwidth}
        \centering
        \includegraphics[width=\linewidth]{figures/lm_avg/default_v_dropless_hgn_gxs_110M_off_0.01.pdf}
        \caption{110M active, 110M - 1.4B total parameters}
    \end{subfigure}
    \caption{
    \textbf{Dropless routing outperforms default routing (\S\ref{sec:expt_router}).}
    We compare dropless routing to the default setting, which allow tokens to be dropped. Across all scales, we find that dropless routing outperforms or performs comparably to default routing. 
    }
    \label{fig:lm_avg_dropless}
\end{figure*}

\begin{figure*}[ht]
    \centering
    \begin{subfigure}[t]{0.45\textwidth}
        \centering
        \includegraphics[width=\linewidth]{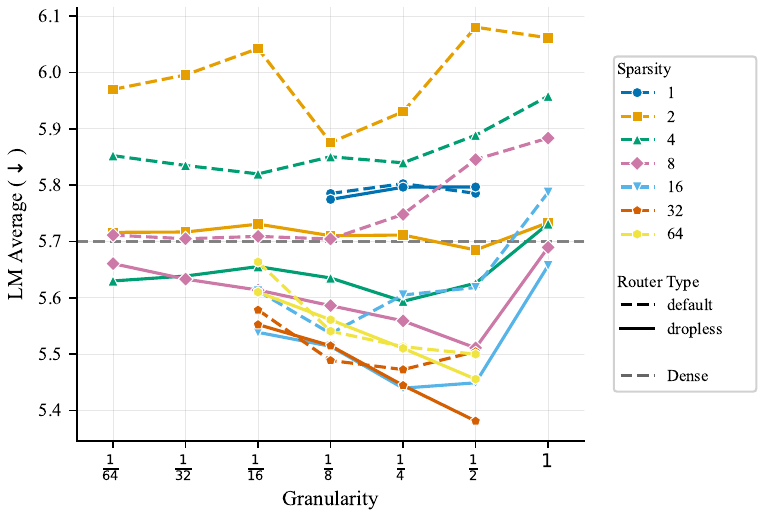}
        \caption{50M active, 50M - 930M total parameters}
    \end{subfigure}
    \hspace{1em}
    \begin{subfigure}[t]{0.45\textwidth}
        \centering
        \includegraphics[width=\linewidth]{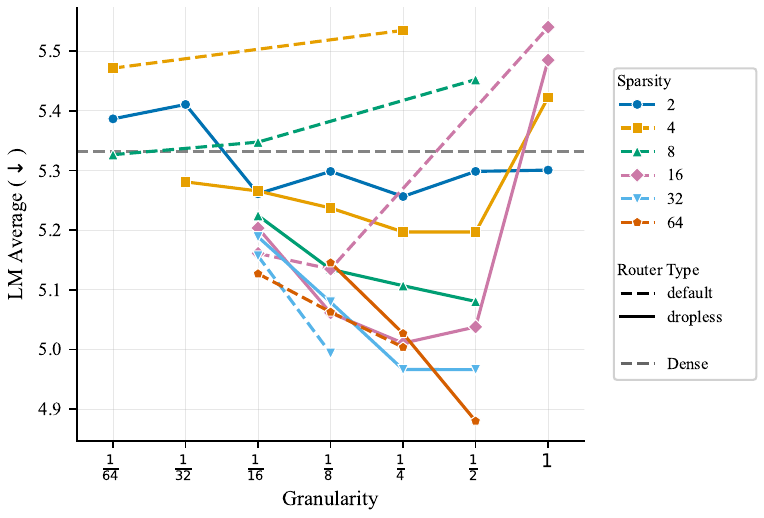}
        \caption{80M active, 80M - 765M total parameters}
    \end{subfigure}
    \caption{
    \textbf{Dropless routing, with bias $\gamma=\num{1e-3}$ (\S\ref{sec:expt_router}).} 
    As in Figure~\ref{fig:lm_avg_dropless}, we compare dropless routing to the default setting, which allow tokens to be dropped. Across all scales, we find that dropless routing outperforms or performs comparably to default routing. We see here with additional higher sparsity default routing runs that as sparsity increases, default routing performance approaches that of dropless routing.
    }
    \label{fig:lm_avg_dropless_with_lf}
\end{figure*}

\begin{figure*}[ht]
    \centering
    \begin{subfigure}[]{\textwidth}
        \centering
        \begin{subfigure}[]{0.46\textwidth}
        \caption*{\scriptsize Fixed activation sparsity (s)}
        \includegraphics[width=\linewidth]{figures/lm_avg/lb_sweep_hgn_gxs_50M.pdf}
        \end{subfigure}
        \hspace{1em}
        \begin{subfigure}[]{0.46\textwidth}
        \caption*{\scriptsize Fixed total experts (n)}
        \includegraphics[width=\linewidth]{figures/lm_avg/lb_sweep_hgn_gxn_50M.pdf}
        \end{subfigure}
        \caption{50M active, 50M - 930M total parameters}
    \end{subfigure}
    \par\bigskip\bigskip
    \begin{subfigure}[]{\textwidth}
        \centering
        \includegraphics[width=0.46\linewidth]{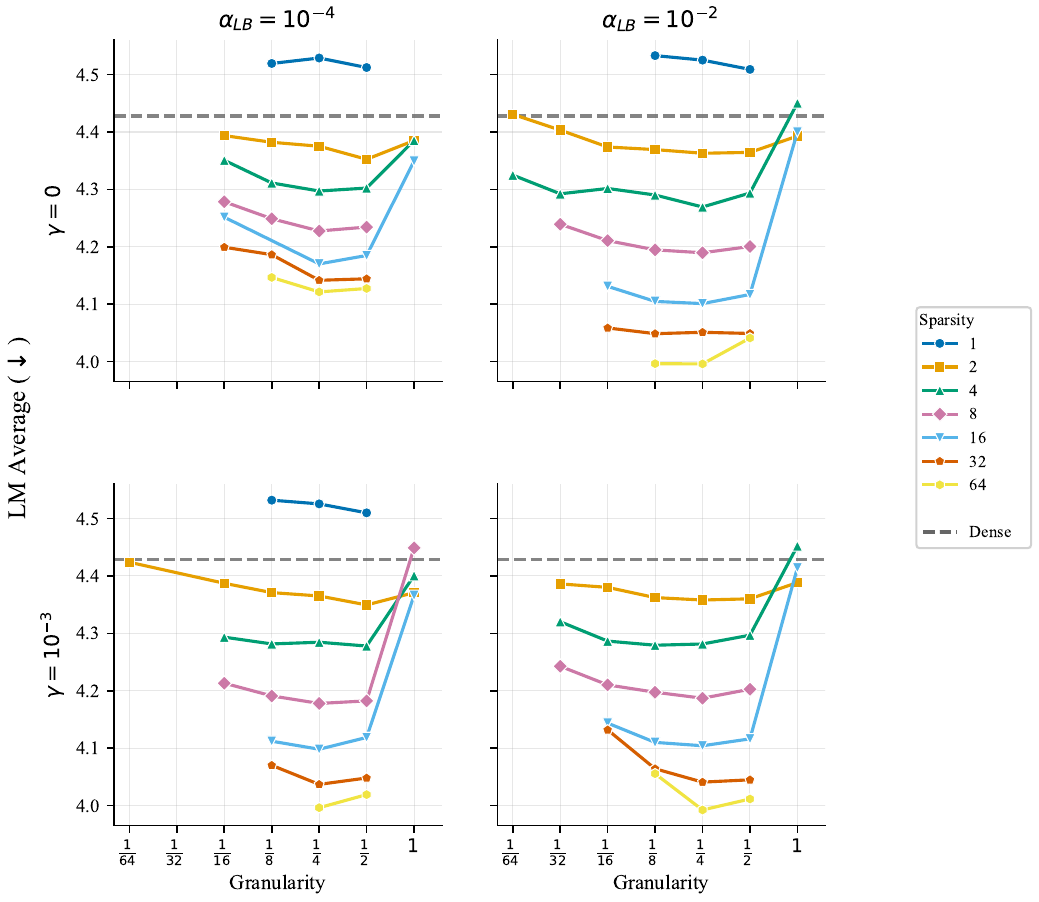}
        \hspace{1em}
        \includegraphics[width=0.46\linewidth]{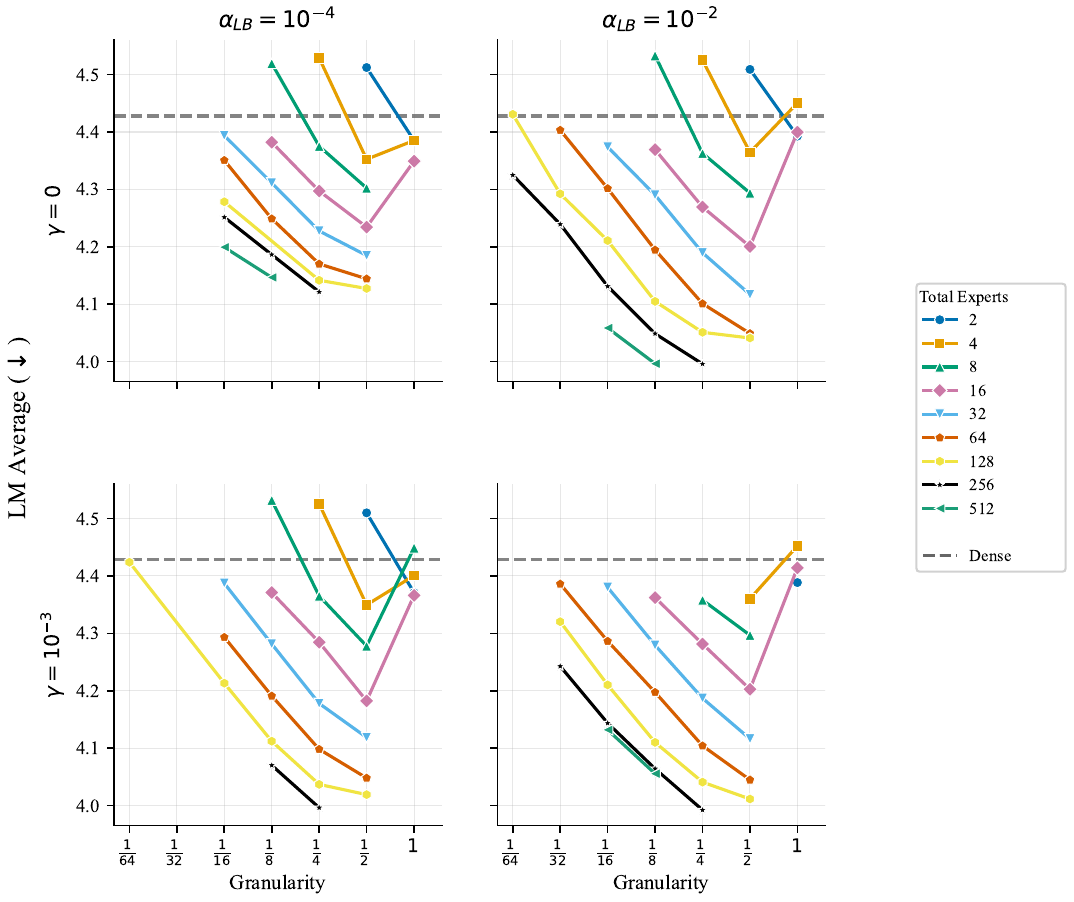}
        \caption{80M active, 80M - 765M total parameters}
    \end{subfigure}
    \par\bigskip\bigskip
    \begin{subfigure}[t]{\textwidth}
        \centering
        \includegraphics[width=0.46\linewidth]{figures/lm_avg/lb_sweep_hgn_gxs_110M.pdf}
        \hspace{1em}
        \includegraphics[width=0.46\linewidth]{figures/lm_avg/lb_sweep_hgn_gxn_110M.pdf}
        \caption{110M active, 110M - 1.4B total parameters}
    \end{subfigure}

    \end{figure*} 

\clearpage 

\begin{figure*}[ht]
    \addtocounter{figure}{-1}
    \centering
    
    \begin{subfigure}[t]{\textwidth}
        \centering
        \begin{subfigure}[]{0.46\textwidth}
        \caption*{\scriptsize Fixed activation sparsity (s)}
        \includegraphics[width=\linewidth]{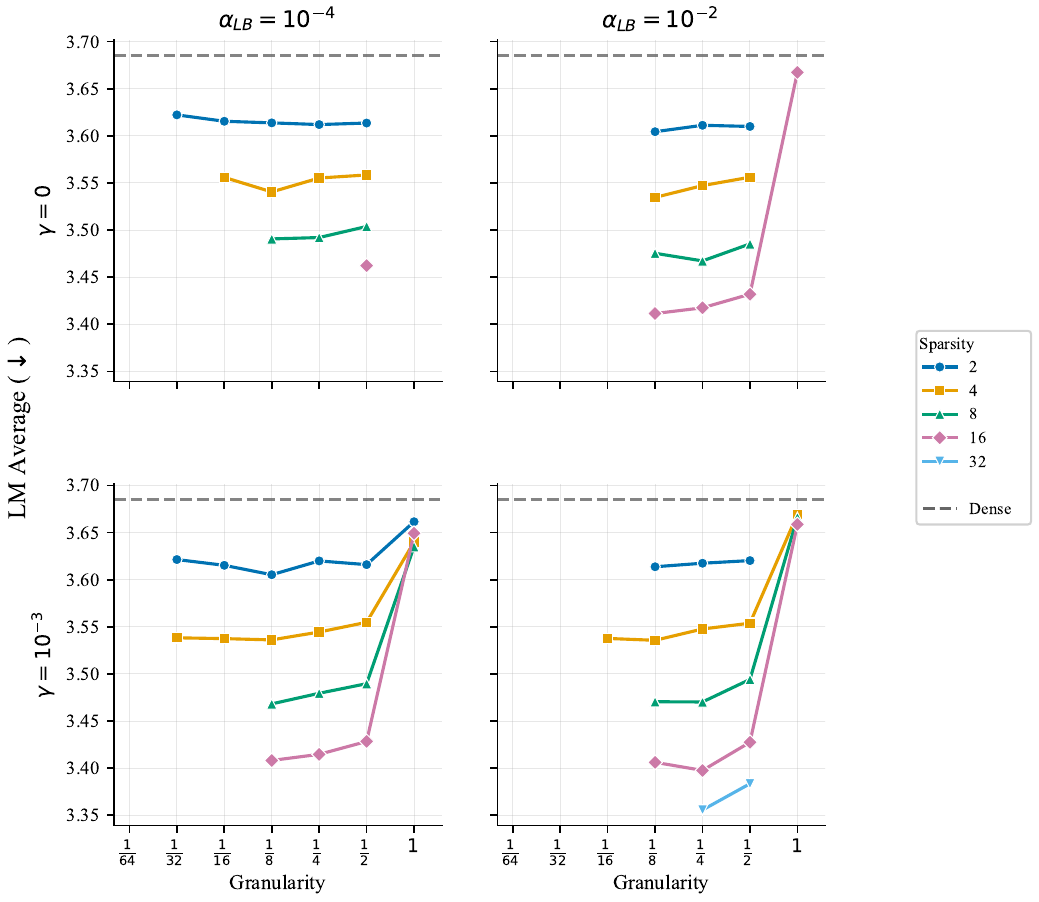}
        \end{subfigure}
        \hspace{1em}
        \begin{subfigure}[]{0.46\textwidth}
        \caption*{\scriptsize Fixed total experts (n)}
        \includegraphics[width=\linewidth]{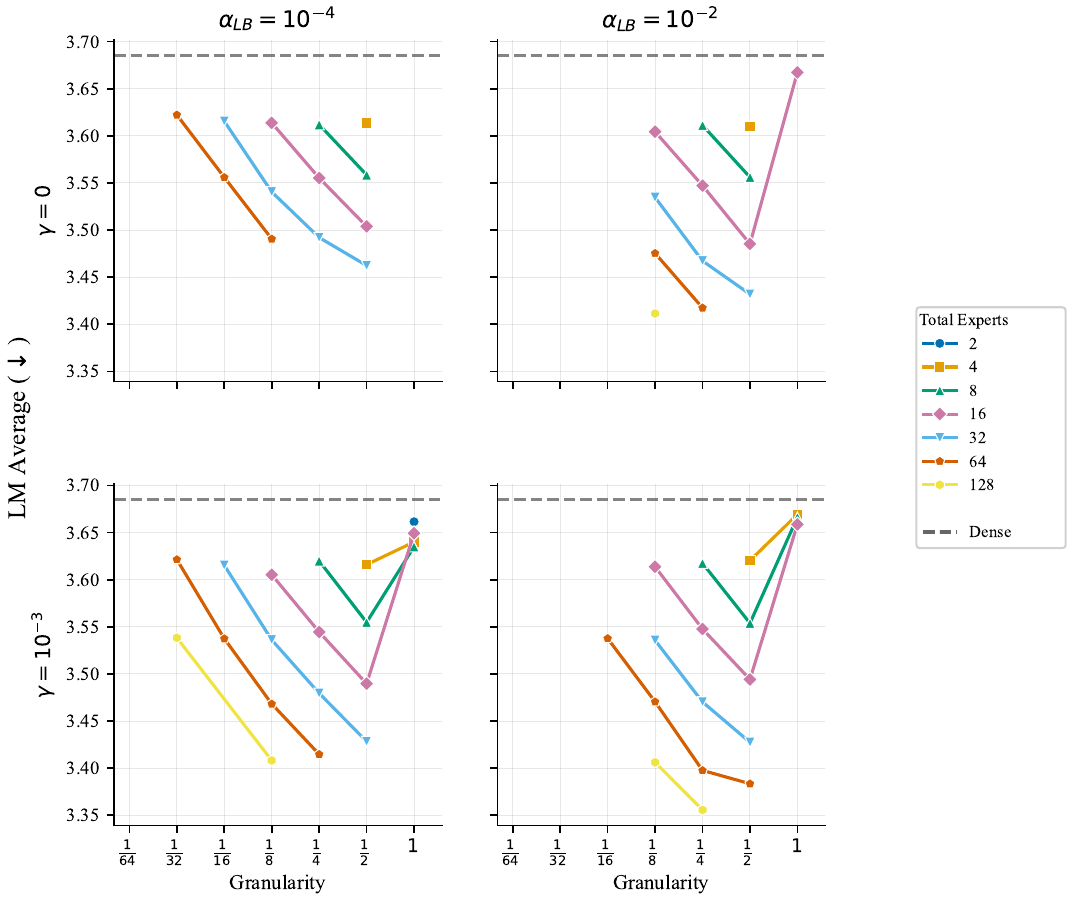}
        \end{subfigure}
        \caption{200M active, 200M - 3.3B total parameters}
    \end{subfigure}
    \par\bigskip\bigskip
    \begin{subfigure}[t]{\textwidth}
        \centering
        \includegraphics[width=0.3\linewidth]{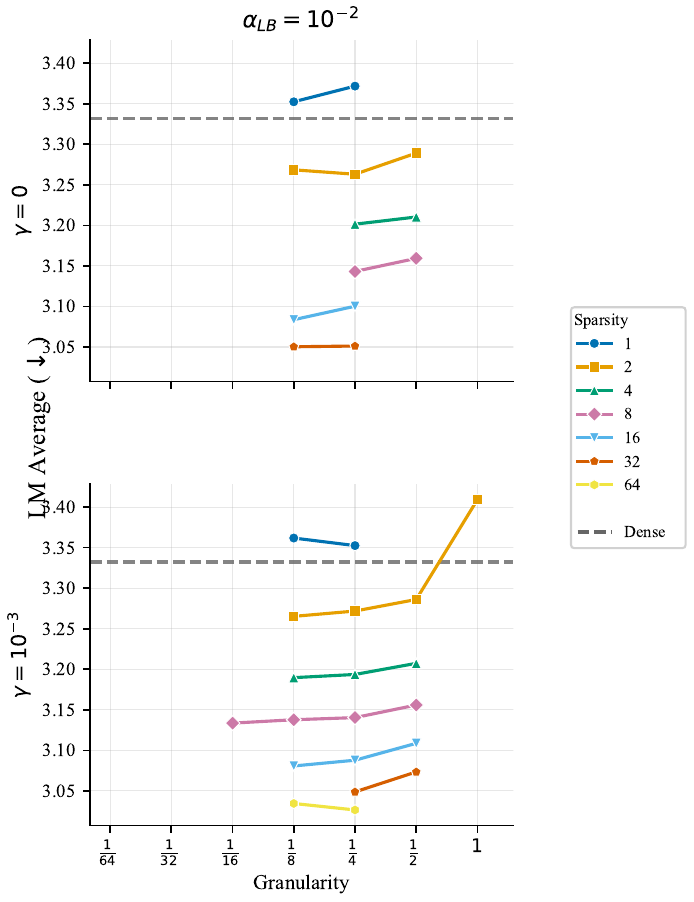}
        \hspace{1em}
        \includegraphics[width=0.3\linewidth]{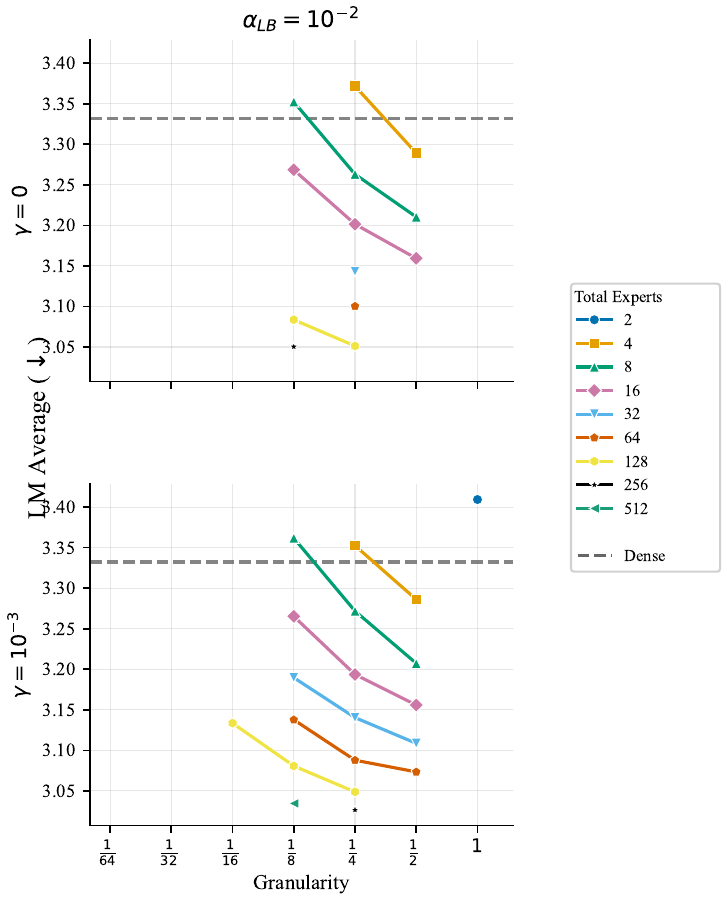}
        \caption{300M active, 300M - 6.6B total parameters}
    \end{subfigure}
    \caption{
    \textbf{Load balancing mechanisms must be tuned correctly (\S\ref{sec:expt_router}).}
    We consider load balancing loss weight $\alpha_{LB} \in \{\num{1e-2}, \num{1e-4}\}$ and loss-free load balancing with bias $\gamma\in\{0, \num{1e-3}\}$ ($\gamma=0$ indicates no loss-free mechanism). Results show that poorly chosen hyperparameters, such as high bias $\gamma = 1e-3$ with total experts $n\geq 512$, may impair performance. However, all settings other than $(\alpha_{LB}=\num{1e-2}, \gamma=\num{1e-3})$ perform comparably for $n \leq 512$, suggesting that a wide range of load balancing settings achieve near-optimal performance. 
    }
    \label{fig:lm_avg_lb}
\end{figure*}

%% file: fig_tex/downstream/hellaswag_acc.tex
\begin{figure*}[!ht]
\centering
\begin{subfigure}[]{\textwidth}
    \centering
    \includegraphics[width=0.46\linewidth]{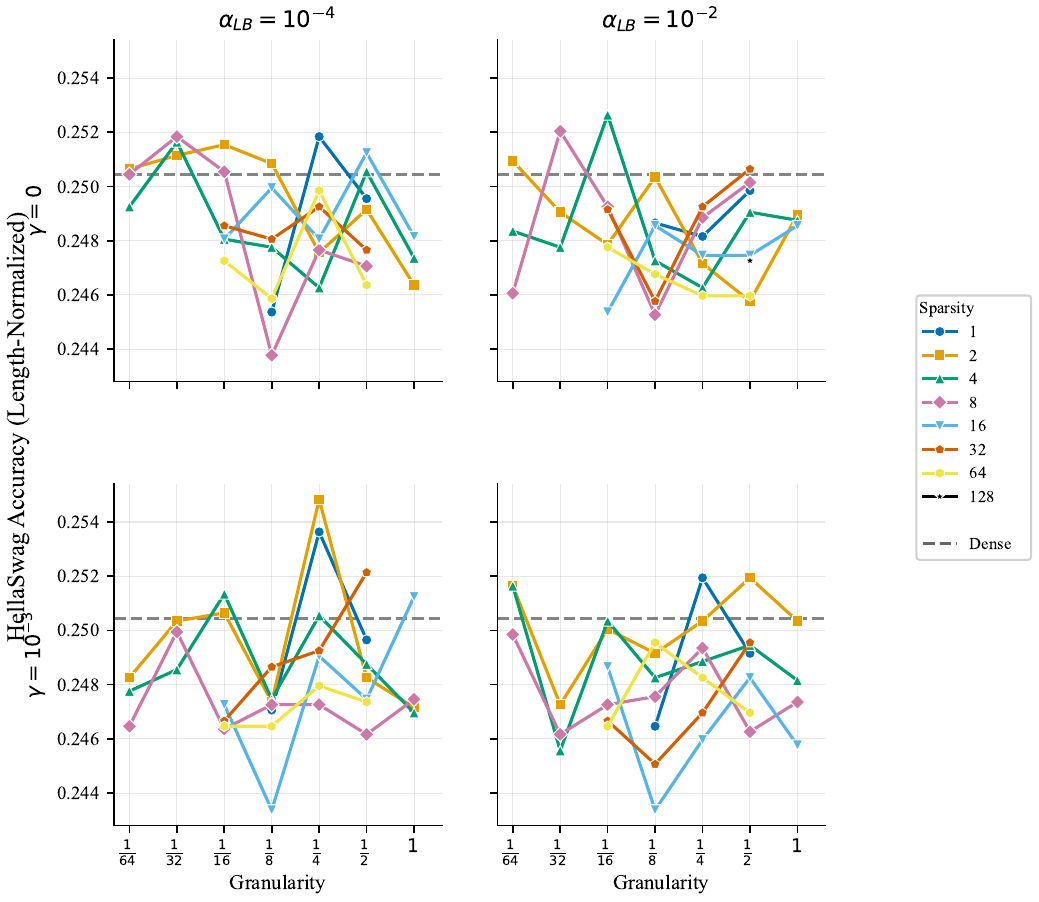}
    \hspace{1em}
    \includegraphics[width=0.46\linewidth]{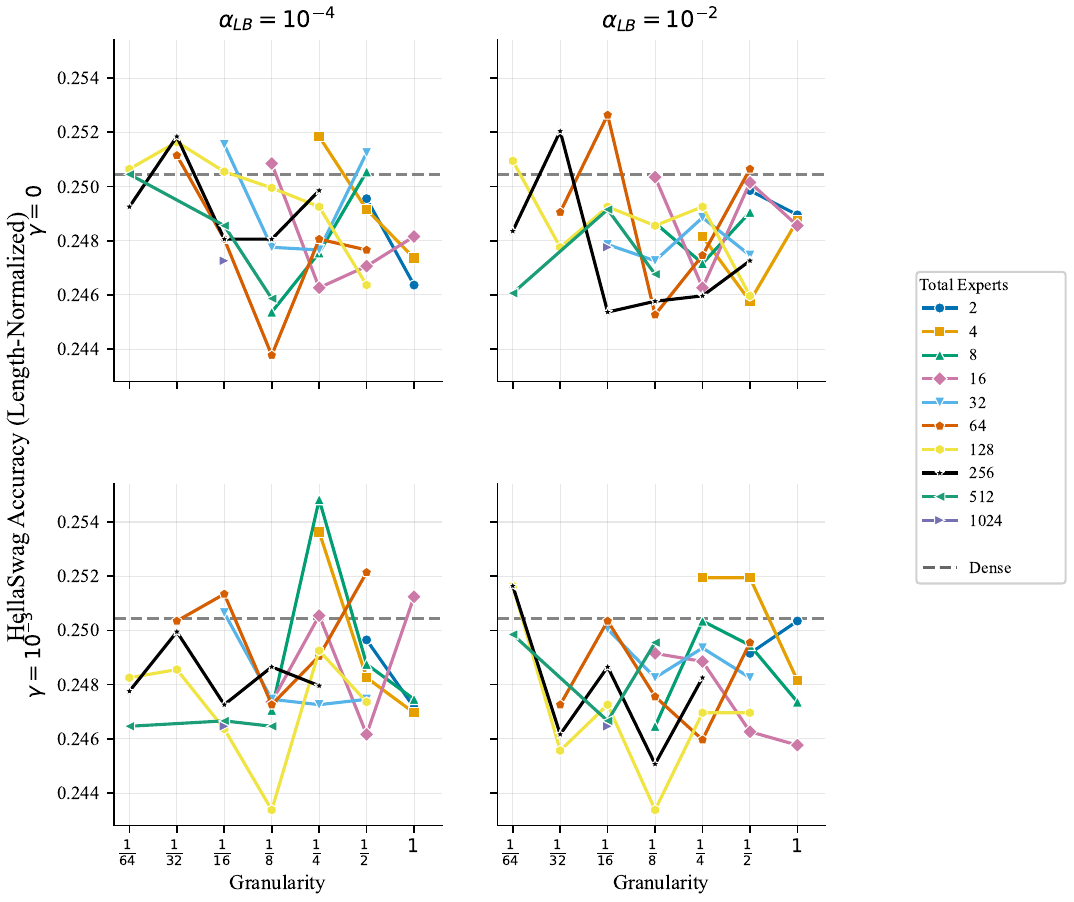}
    \caption{50M active, 50M - 930M total parameters}
\end{subfigure}
\par\bigskip\bigskip
\begin{subfigure}[]{\textwidth}
    \centering
    \includegraphics[width=0.46\linewidth]{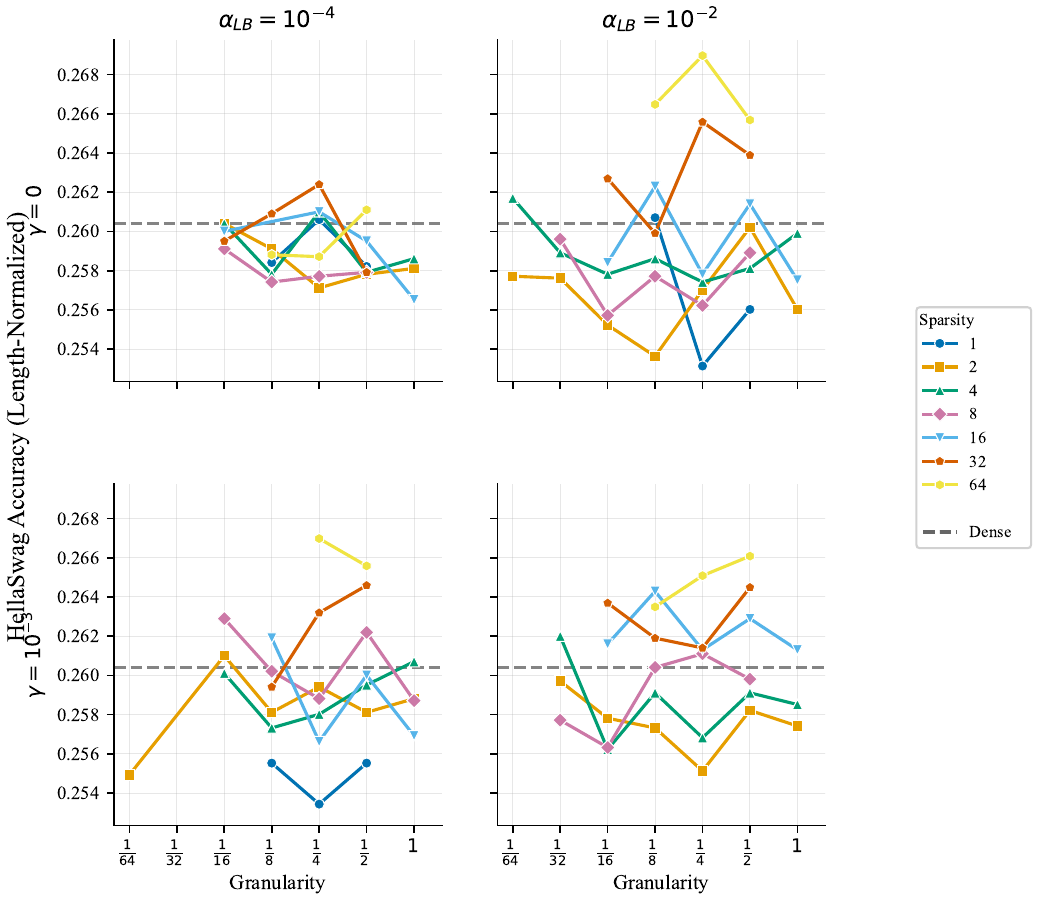}
    \hspace{1em}
    \includegraphics[width=0.46\linewidth]{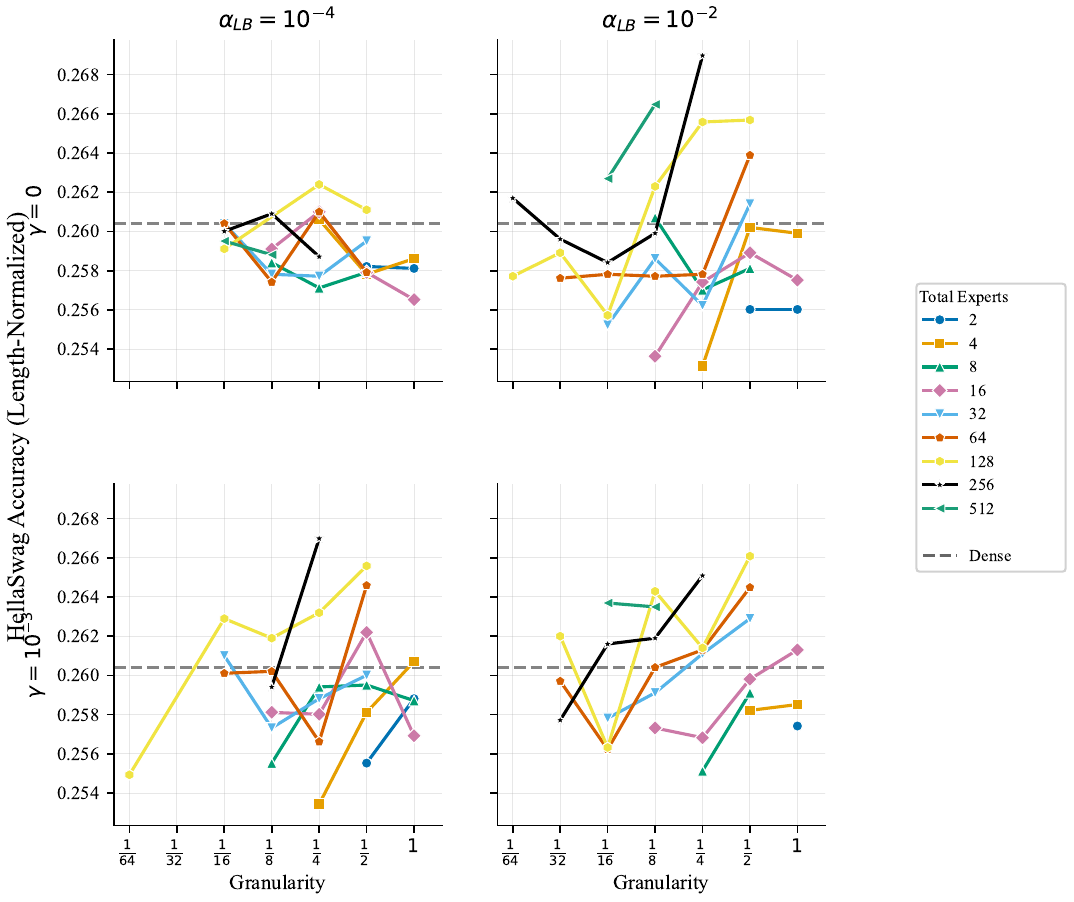}
    \caption{80M active, 80M - 765M total parameters}
\end{subfigure}
\par\bigskip\bigskip
\begin{subfigure}[t]{\textwidth}
    \centering
    \includegraphics[width=0.46\linewidth]{figures/hellaswag_acc/lb_sweep_hgn_gxs_110M.pdf}
    \hspace{1em}
    \includegraphics[width=0.46\linewidth]{figures/hellaswag_acc/lb_sweep_hgn_gxn_110M.pdf}
    \caption{110M active, 110M - 1.4B total parameters}
\end{subfigure}
\end{figure*} 

\clearpage  

\begin{figure*}[ht]
    \addtocounter{figure}{-1}
    \centering
    \begin{subfigure}[t]{\textwidth}
        \centering
        \includegraphics[width=0.46\linewidth]{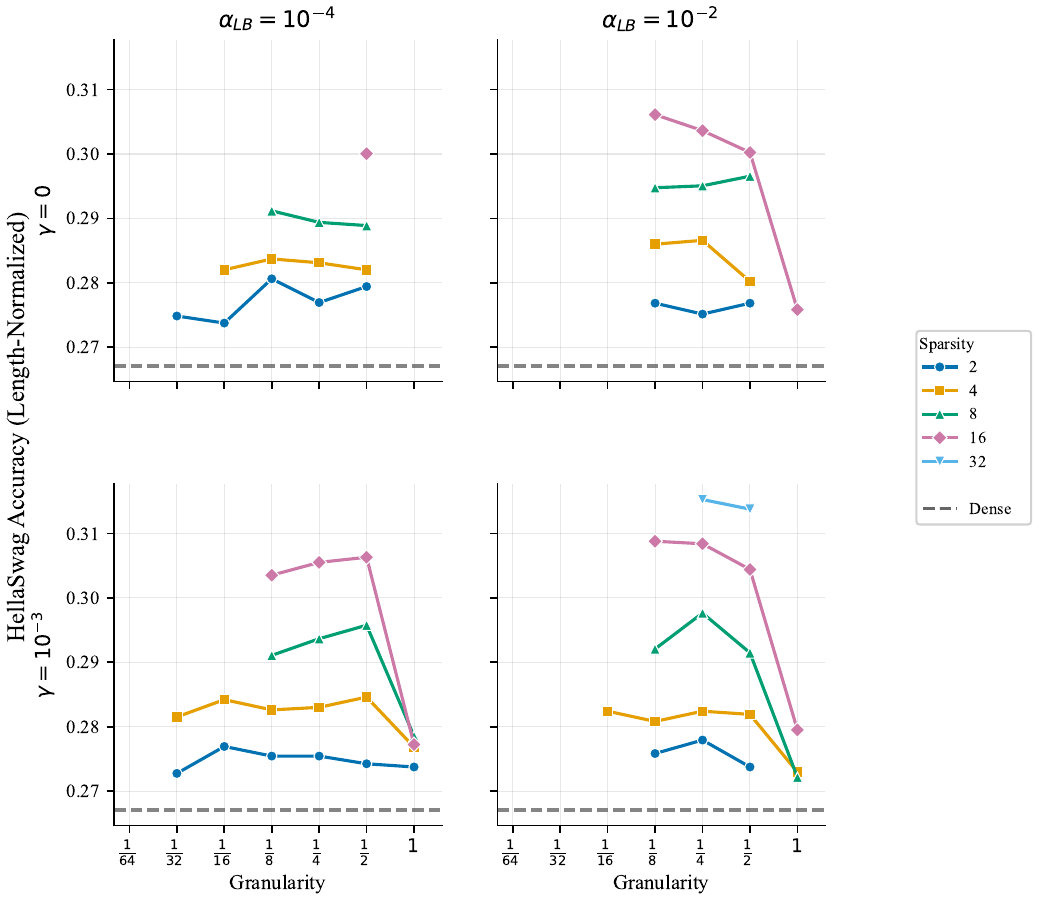}
        \hspace{1em}
        \includegraphics[width=0.46\linewidth]{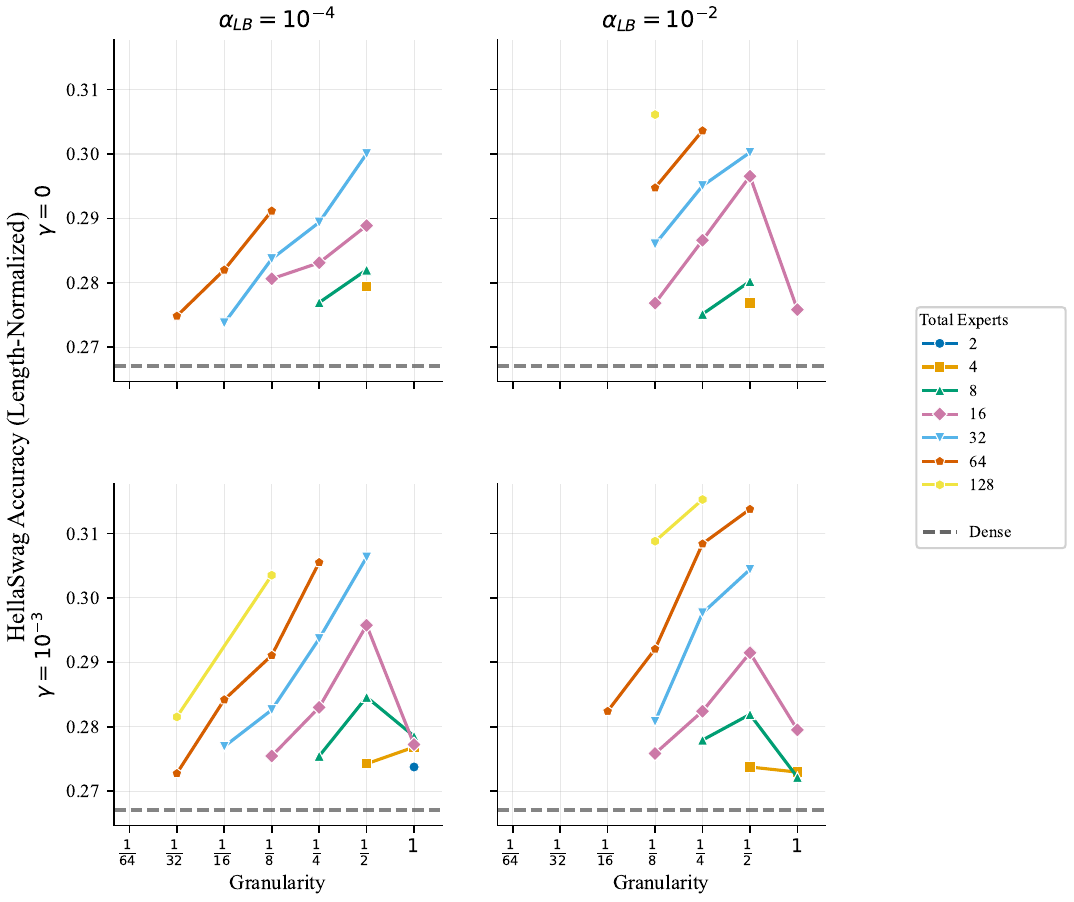}
        \caption{200M active, 200M - 3.3B total parameters}
    \end{subfigure}
    \par\bigskip\bigskip
    \begin{subfigure}[t]{\textwidth}
        \centering
        \includegraphics[width=0.3\linewidth]{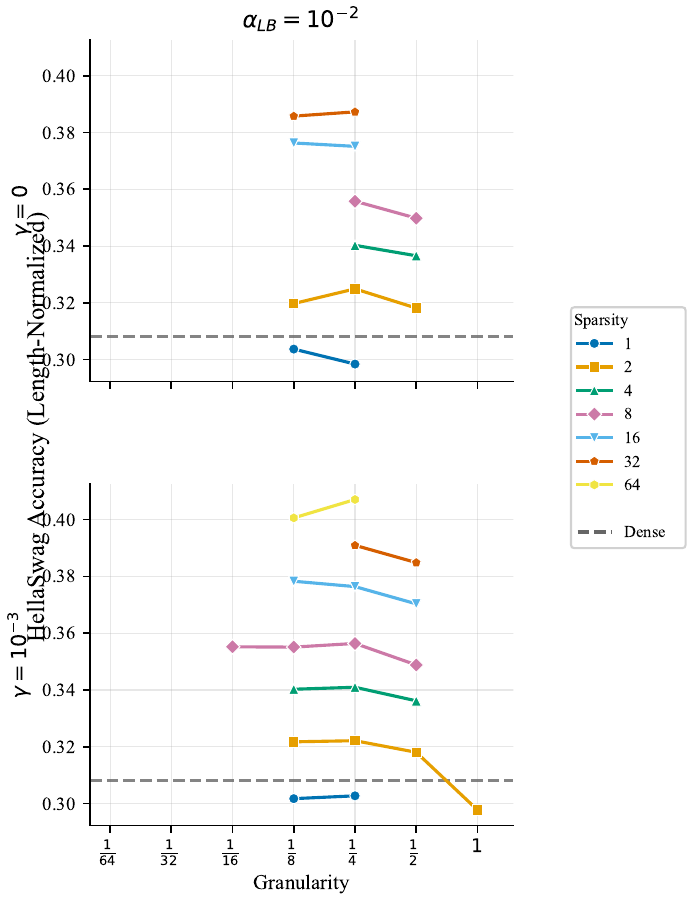}
        \hspace{1em}
        \includegraphics[width=0.3\linewidth]{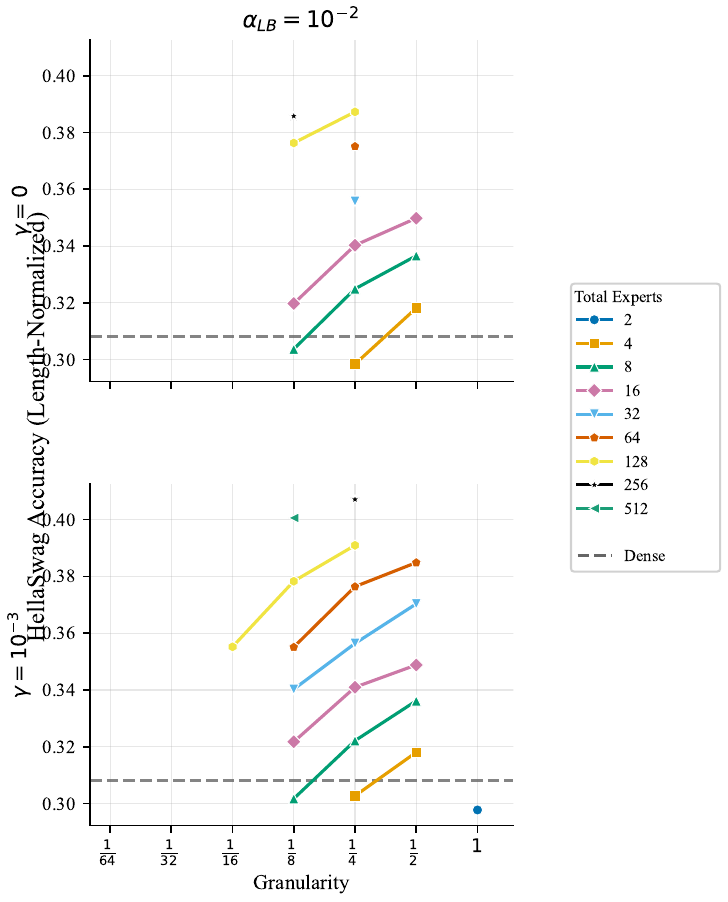}
        \caption{300M active, 300M - 6.6B total parameters}
    \end{subfigure}
    \caption{
    \textbf{At sufficiently large compute scales, MoEs performance on Hellaswag accuracy mirrors cross-entropy loss (\S\ref{sec:expt_router}).} 
    }
    \label{fig:hellaswag_acc_lb}
\end{figure*}

%% file: fig_tex/lm/c4.tex
\begin{figure*}[!ht]
    \centering
        \begin{subfigure}[t]{\textwidth}
        \begin{subfigure}[t]{0.33\textwidth}
            \centering
            \caption*{\scriptsize Fixed total experts (n)}
            \includegraphics[width=\linewidth]{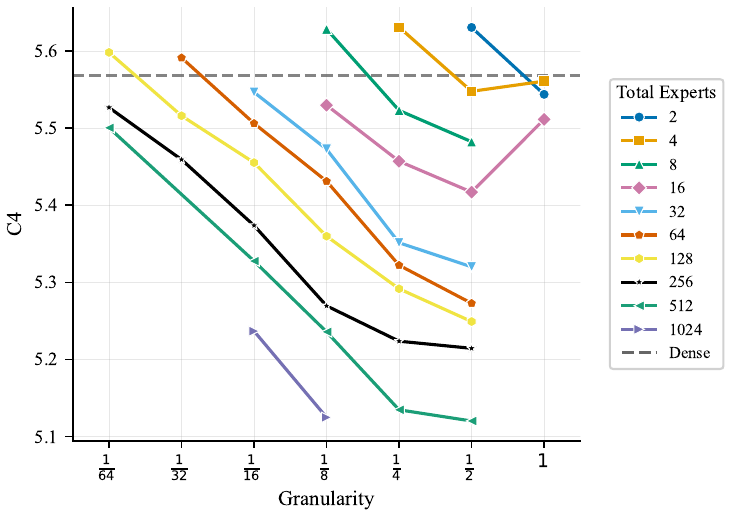}
        \end{subfigure}
        \begin{subfigure}[t]{0.33\textwidth}
            \centering
            \caption*{\scriptsize Fixed granularity (g)}
            \includegraphics[width=\linewidth]{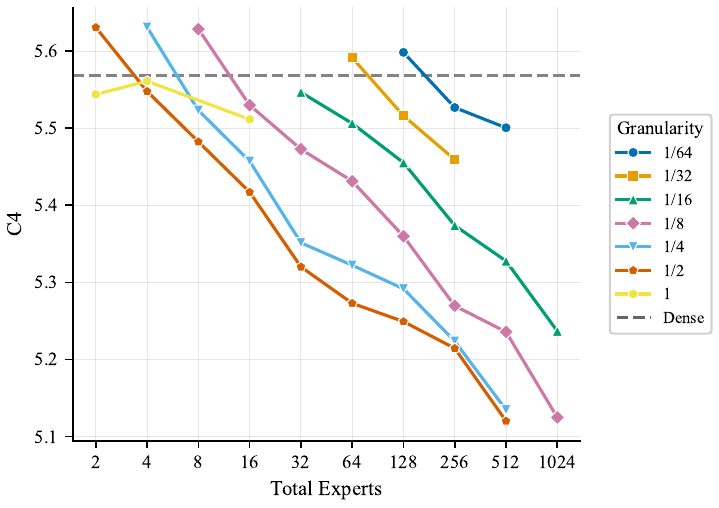}
        \end{subfigure}
        \begin{subfigure}[t]{0.33\textwidth}
            \centering
            \caption*{\scriptsize Fixed activation sparsity (s)}
            \includegraphics[width=\linewidth]{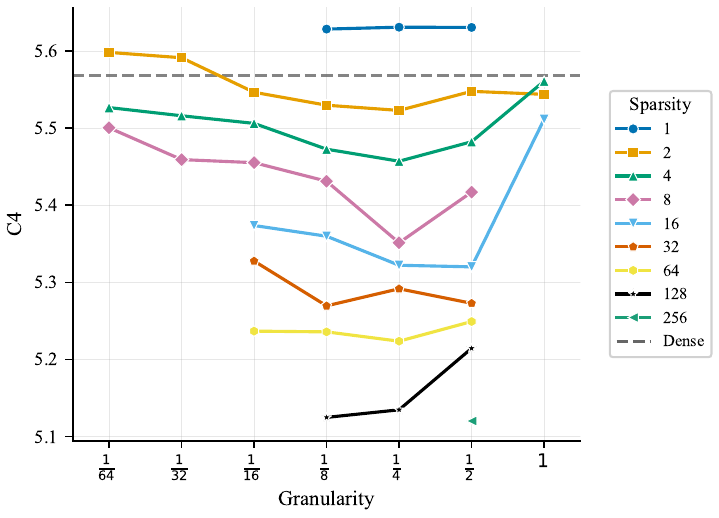}
        \end{subfigure}
        \caption{50M active, 50M - 930M total parameters}
    \end{subfigure}
\par\bigskip\bigskip
    \begin{subfigure}[t]{\textwidth}
        \begin{subfigure}[t]{0.33\textwidth}
            \centering
            \includegraphics[width=\linewidth]{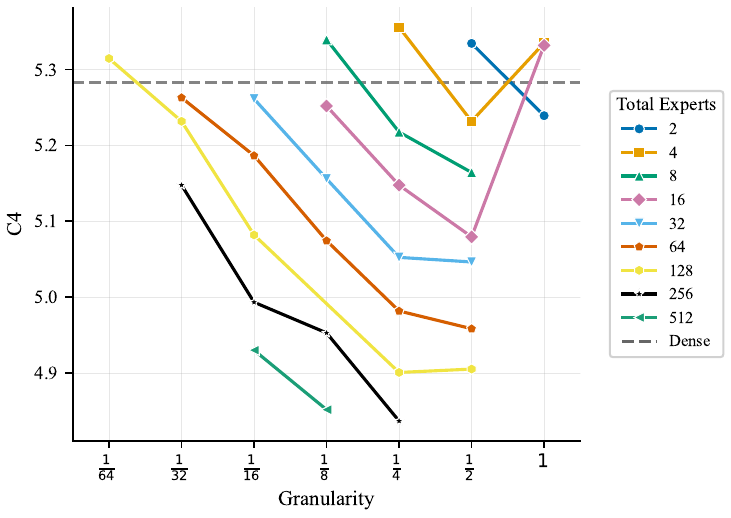}
        \end{subfigure}
        \begin{subfigure}[t]{0.33\textwidth}
            \centering
            \includegraphics[width=\linewidth]{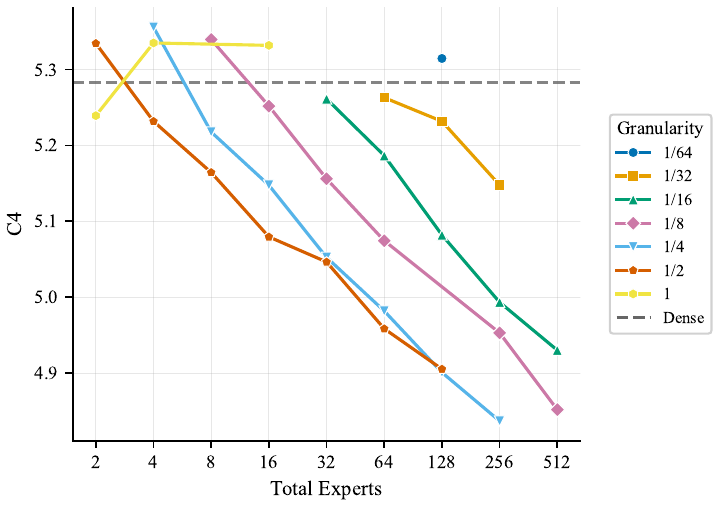}
        \end{subfigure}
        \begin{subfigure}[t]{0.33\textwidth}
            \centering
            \includegraphics[width=\linewidth]{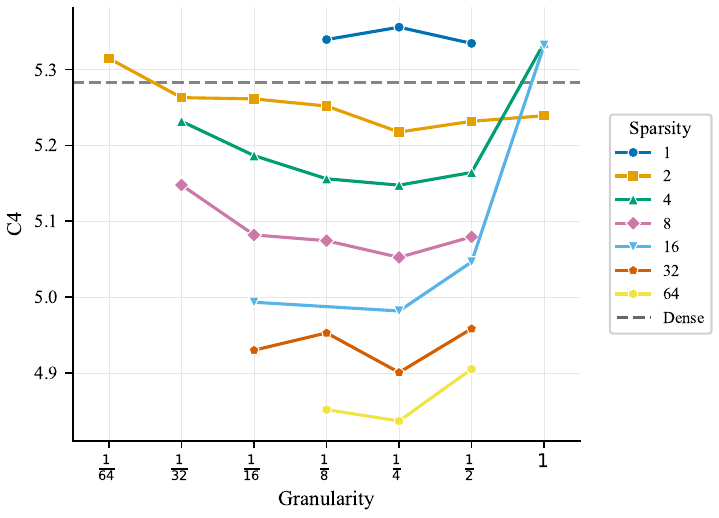}
        \end{subfigure}
        \caption{80M active, 80M - 765M total parameters}
    \end{subfigure}
    \par\bigskip\bigskip
        \begin{subfigure}[t]{\textwidth}
        \begin{subfigure}[t]{0.33\textwidth}
            \centering
            \includegraphics[width=\linewidth]{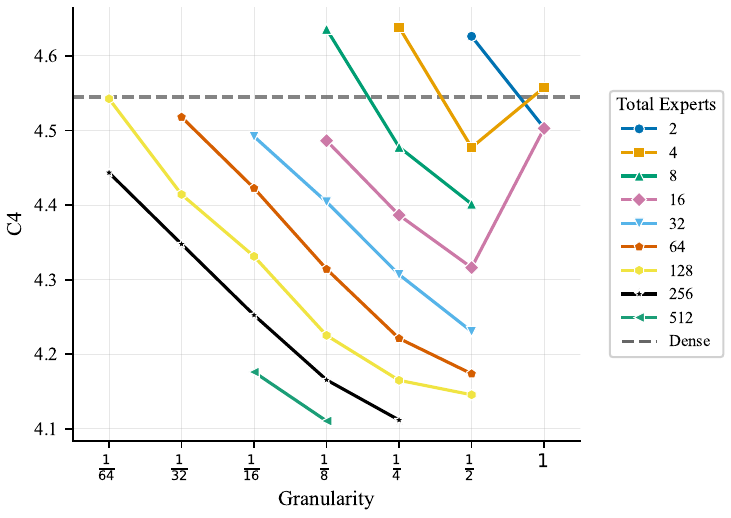}
        \end{subfigure}
        \begin{subfigure}[t]{0.33\textwidth}
            \centering
            \includegraphics[width=\linewidth]{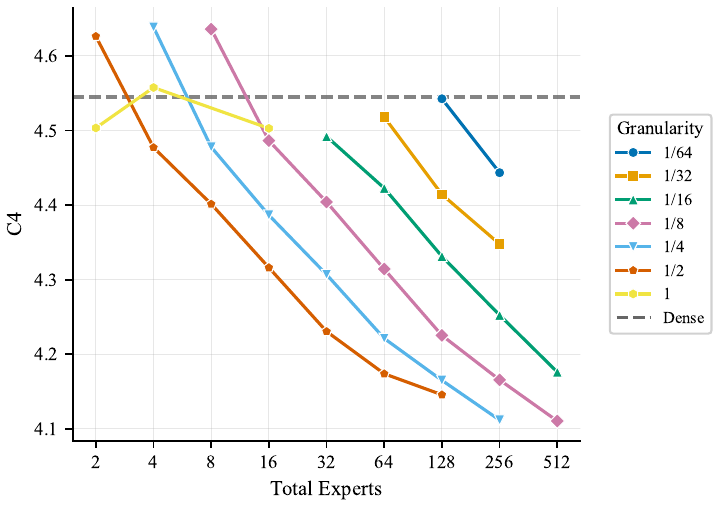}
        \end{subfigure}
        \begin{subfigure}[t]{0.33\textwidth}
            \centering
            \includegraphics[width=\linewidth]{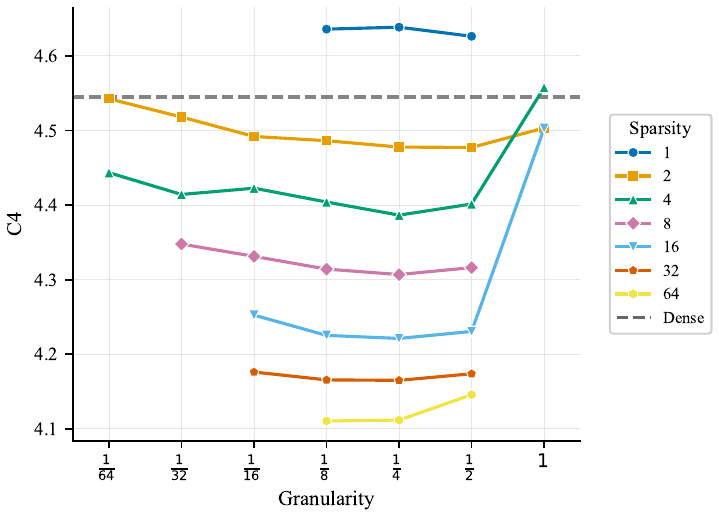}
        \end{subfigure}
        \caption{110M active, 110M - 1.4B total parameters}
    \end{subfigure}
    \end{figure*}

\clearpage  

\begin{figure*}[!ht]
        \addtocounter{figure}{-1}
    \begin{subfigure}[t]{\textwidth}
        \addtocounter{subfigure}{3}
        \begin{subfigure}[t]{0.33\textwidth}
            \centering
            \caption*{\scriptsize Fixed total experts (n)}
            \includegraphics[width=\linewidth]{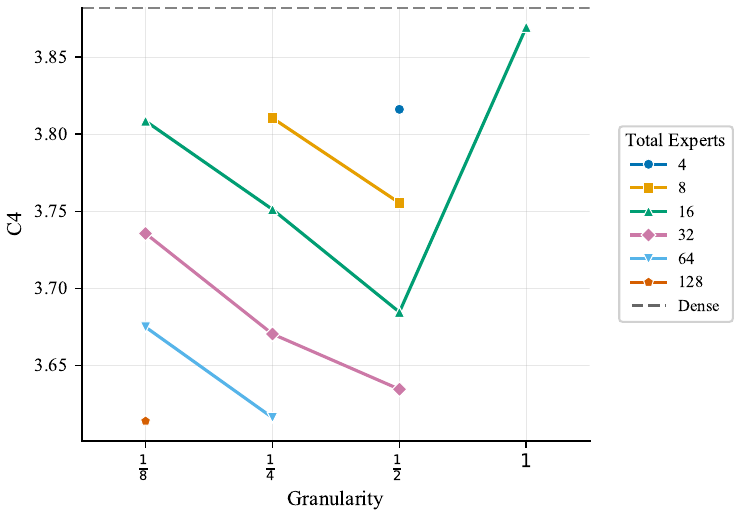}
        \end{subfigure}
        \begin{subfigure}[t]{0.33\textwidth}
            \centering
            \caption*{\scriptsize Fixed granularity (g)}
            \includegraphics[width=\linewidth]{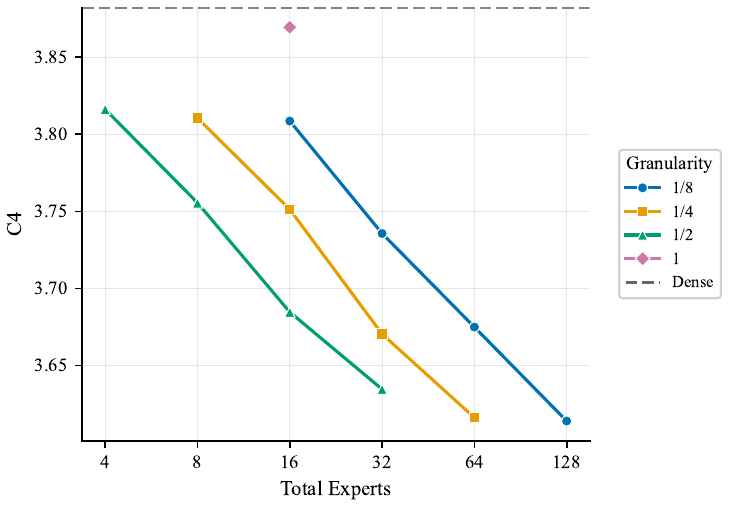}
        \end{subfigure}
        \begin{subfigure}[t]{0.33\textwidth}
            \centering
            \caption*{\scriptsize Fixed activation sparsity (s)}
            \includegraphics[width=\linewidth]{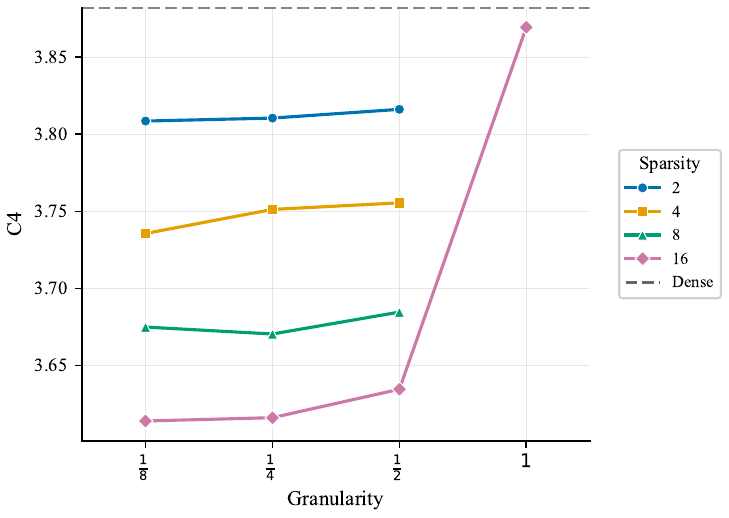}
        \end{subfigure}
        \caption{200M active, 200M - 3.3B total parameters}
    \end{subfigure}
    \par\bigskip\bigskip
        \begin{subfigure}[t]{\textwidth}
        \begin{subfigure}[t]{0.33\textwidth}
            \centering
            \includegraphics[width=\linewidth]{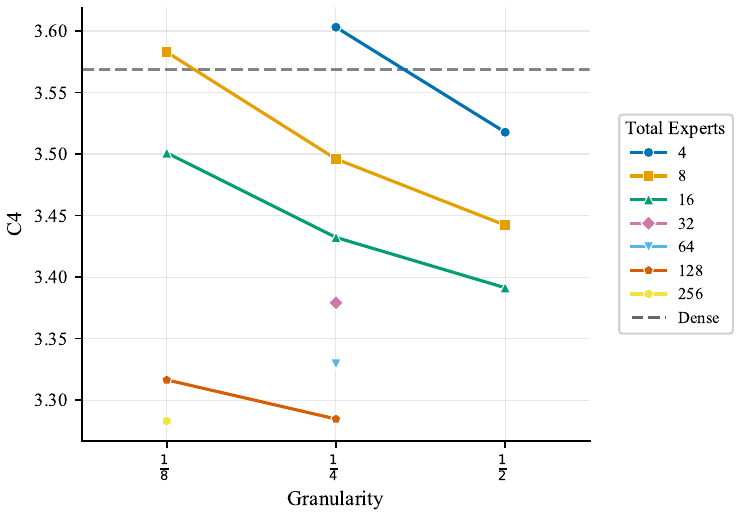}
        \end{subfigure}
        \begin{subfigure}[t]{0.33\textwidth}
            \centering
            \includegraphics[width=\linewidth]{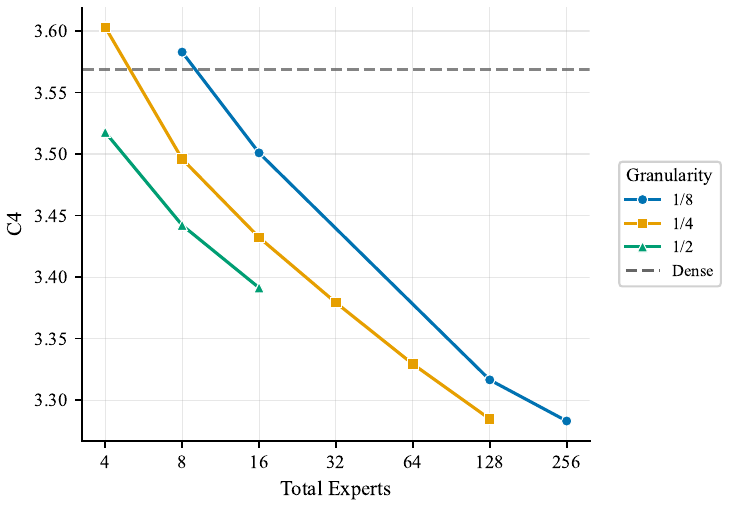}
        \end{subfigure}
        \begin{subfigure}[t]{0.33\textwidth}
            \centering
            \includegraphics[width=\linewidth]{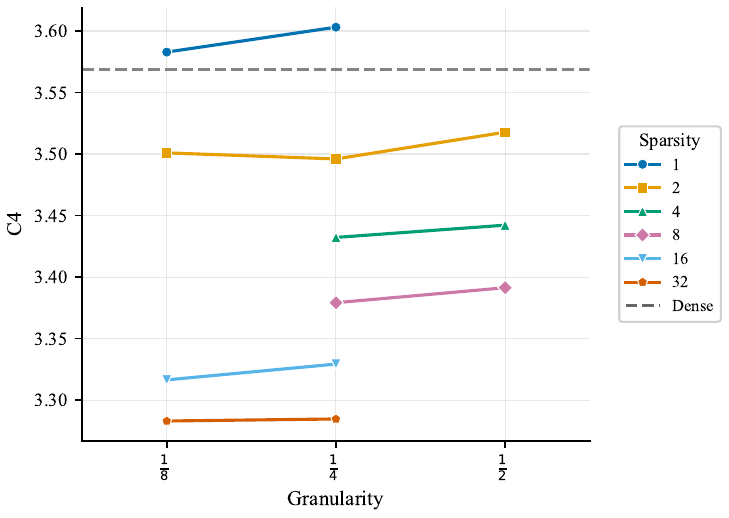}
        \end{subfigure}
        \caption{300M active, 300M - 6.6B total parameters}
    \end{subfigure}

    \caption{
    \textbf{Increasing inactive expert parameters via expert size (left) or total count (center) improves performance in MoEs (\S\ref{sec:expt_main}).} This effect is seen both when holding total number of experts fixed (left) and when holding expert granularity fixed (center). In general, increasing total parameters results in improved performance.  \textbf{Optimal tradeoff between expert count and granularity varies in MoEs (right). (\S\ref{sec:expt_main})}
    At each activation sparsity $s$ (equivalently, at each total parameter count), the optimal (total expert count, expert granularity) configuration varies. As $s$ increases, optimal expert granularity remains nearly fixed, suggesting that sparsity should be scaled up primarily by increasing total expert count $n$, while maintaining a near constant, slowly increasing expert granularity $g$. 
    }
    \label{fig:c4_experts}
\end{figure*}

\begin{figure*}[!ht]
    \centering
    
    \begin{subfigure}[t]{0.46\textwidth}
        \centering
        \includegraphics[width=\linewidth]{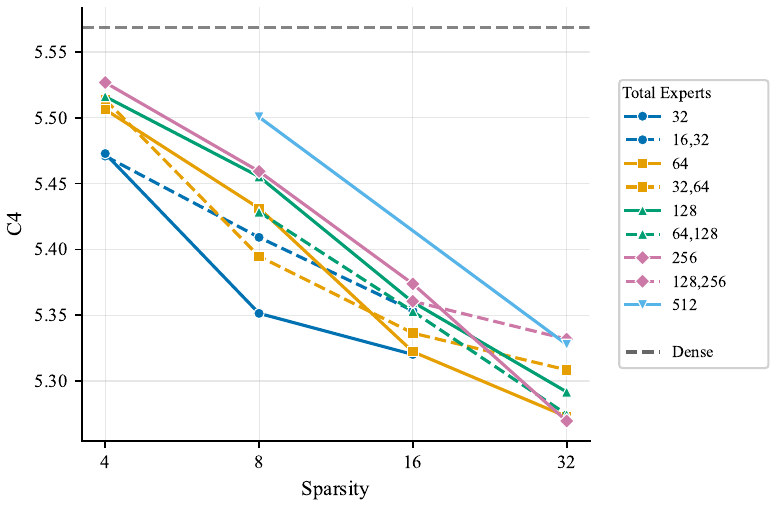}
        \caption{50M active, 50M - 930M total parameters}
    \end{subfigure}
    \vspace{1em}
    \begin{subfigure}[t]{0.46\textwidth}
        \centering
        \includegraphics[width=\linewidth]{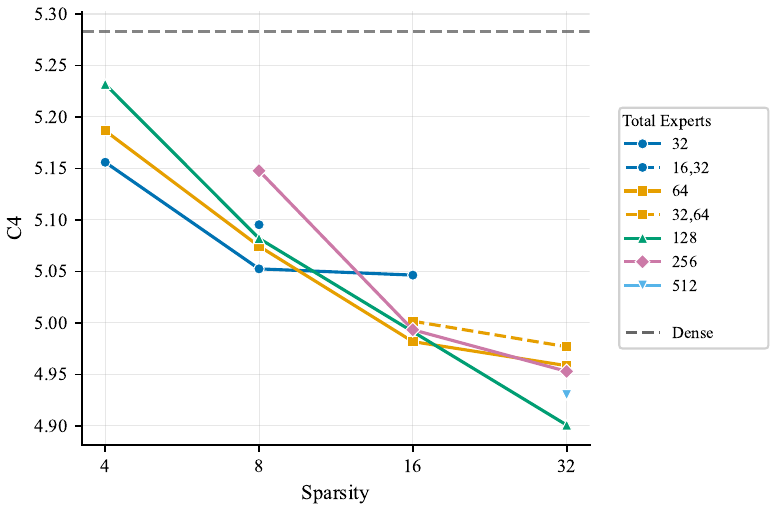}
        \caption{80M active, 80M - 765M total parameters}
    \end{subfigure}
    \caption{
    \textbf{Heterogeneity of expert size alone does not improve MoE performance (\S\ref{sec:expt_hetgen}).} To explore the potential benefits of their architectural flexibility, we compare heterogeneous MoEs (indicated by dotted lines) to active- and total-parameter-matched homogeneous MoEs. Heterogeneity alone does not result in performance gains, as, at each activation sparsity $s$, heterogeneous MoEs with $n_1, n_2 = a, b$ lie between or near the 2 closest homogeneous MoEs, with $n=a$ and with $n=b$.
    }
    \label{fig:c4_het}
\end{figure*}

\begin{figure*}[!ht]
    \centering
    
    \begin{subfigure}[t]{1.0\textwidth}
        \centering
        \includegraphics[width=\linewidth]{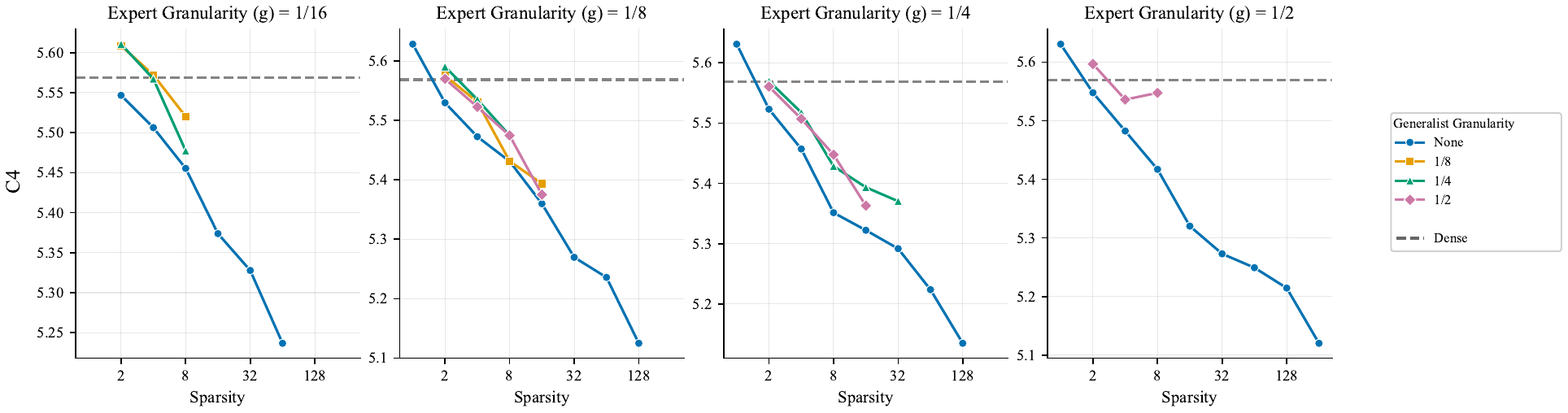}
        \caption{50M active, 50M - 930M total parameters}
    \end{subfigure}
    \par\bigskip\bigskip
    \begin{subfigure}[t]{1.0\textwidth}
        \centering
        \includegraphics[width=\linewidth]{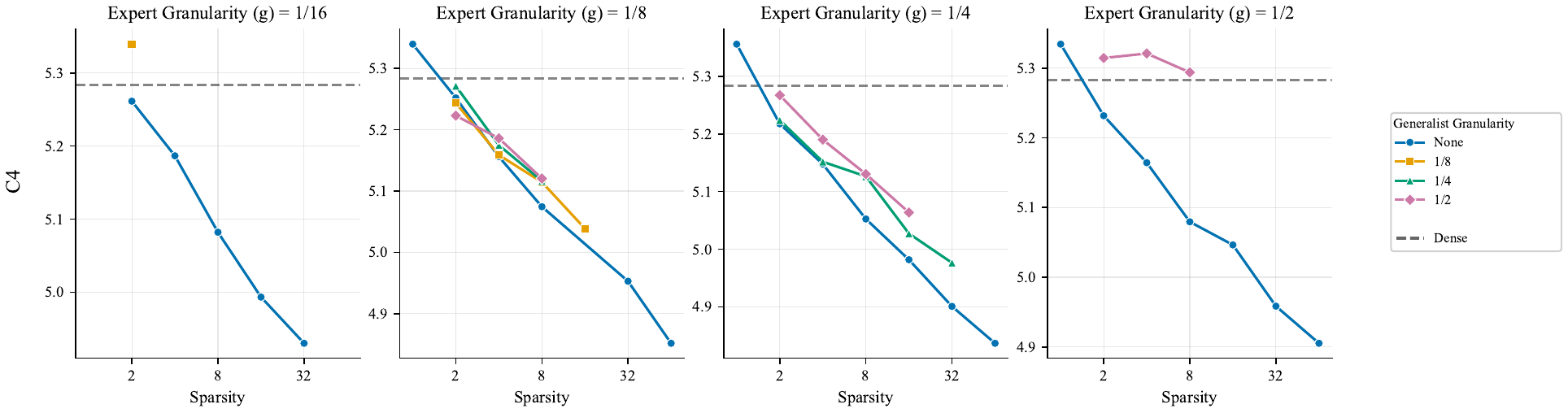}
        \caption{80M active, 80M - 765M total parameters}
    \end{subfigure}
    \par\bigskip\bigskip
    \begin{subfigure}[t]{1.0\textwidth}
        \centering
        \includegraphics[width=\linewidth]{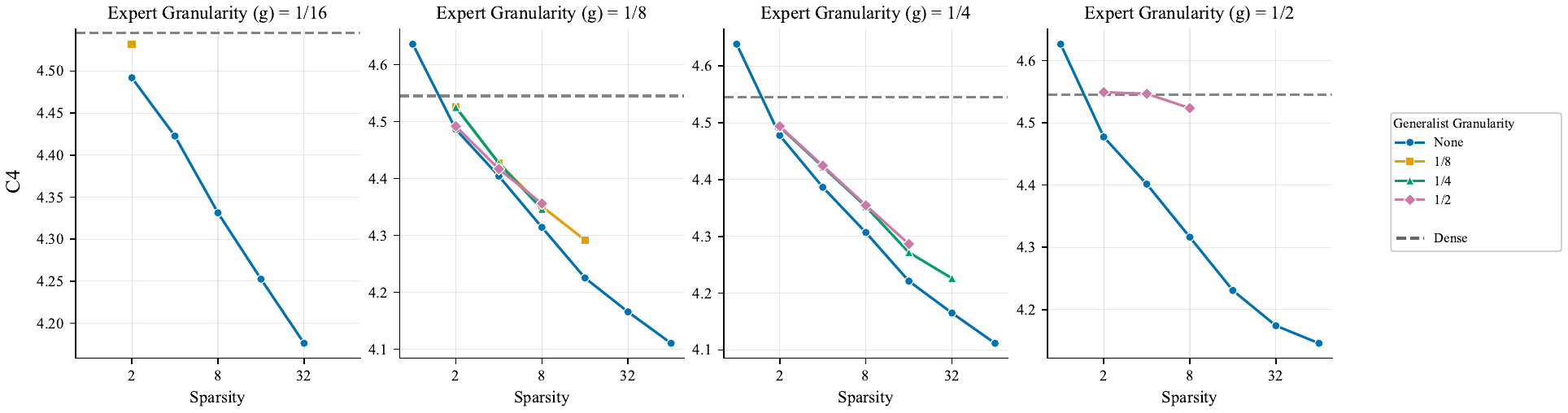}
        \caption{110M active, 110M - 1.4B total parameters}
    \end{subfigure}
    \caption{
    \textbf{The inclusion of a generalist consistently degrades performance in homogeneous MoEs (\S\ref{sec:expt_hetgen}).}
    We train MoE LMs which consist of some routed experts with granularity $g$, as well as a generalist with granularity $g_{gen}\in \{\frac{1}{2}, \frac{1}{4}, \frac{1}{8}\} $. We compare to settings with no generalist, only routed experts with granularity $g$. In all settings and configurations, the addition of any granularity generalist results in comparable or degraded performance. 
    }
    \label{fig:c4_gen}
\end{figure*}

\begin{figure*}[ht]
    \centering
    \begin{subfigure}[t]{1.0\textwidth}
        \centering
        \includegraphics[width=\linewidth]{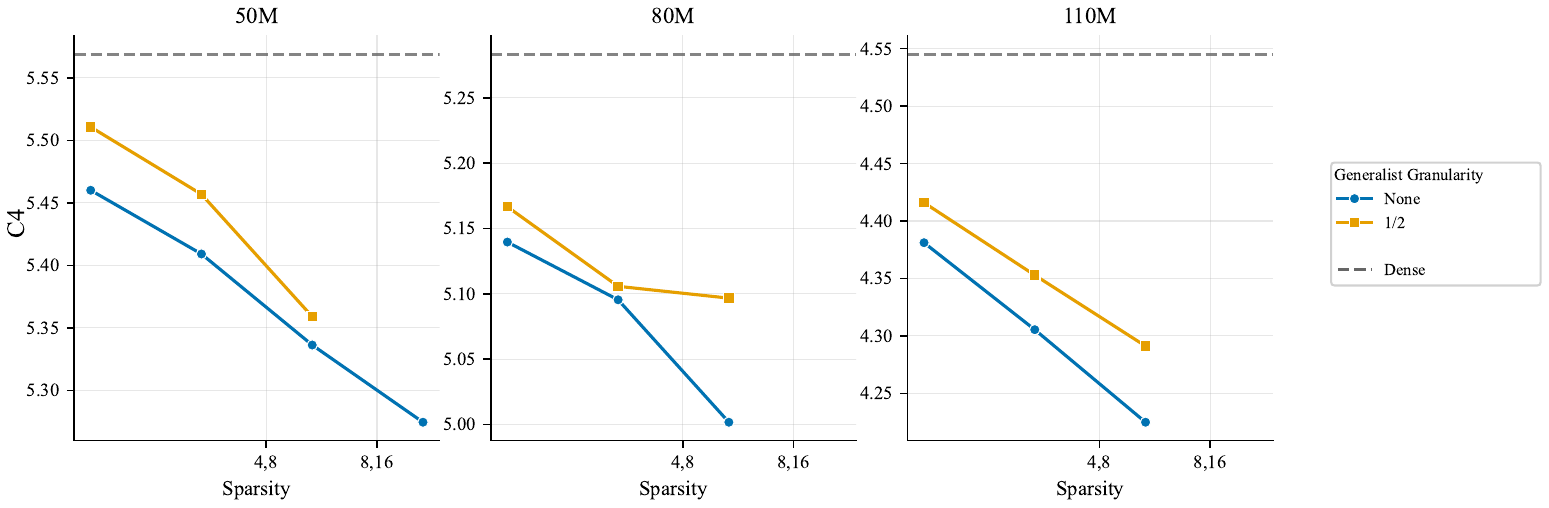}
    \end{subfigure}
    \caption{
    \textbf{The inclusion of a generalist consistently degrades performance in heterogeneous MoEs (\S\ref{sec:expt_hetgen}).}
    We train heterogeneous MoE LMs which consist of  routed experts with granularity $g_1, g_2$, as well as a generalist with granularity $g_{gen} = \frac{1}{2}$. We compare to settings with no generalist. In all settings and configurations, the addition of a generalist results in comparable or degraded performance. 
    }
    \label{fig:c4_hetgen}
\end{figure*}

\begin{figure*}[ht]
    \centering
    \begin{subfigure}[t]{\textwidth}
        \centering
        \begin{subfigure}[t]{0.45\textwidth}
            \includegraphics[width=\linewidth]{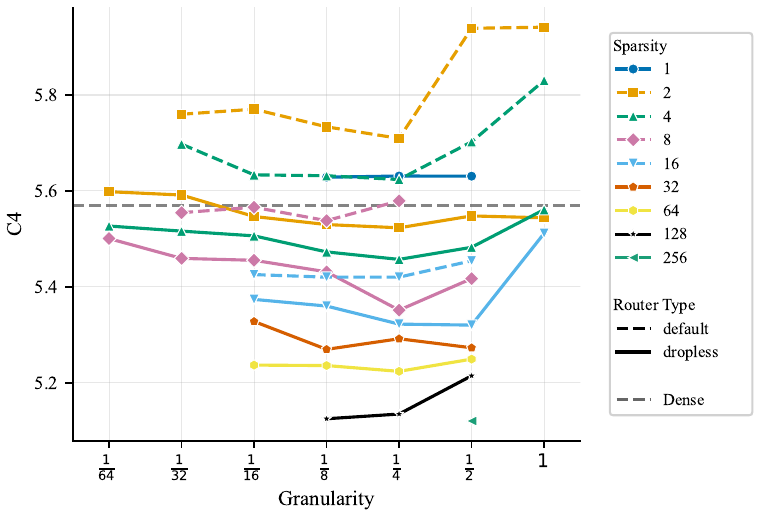}
            \caption{50M active, 50M - 930M total parameters}
        \end{subfigure}
    \hspace{1em}
        \begin{subfigure}[t]{0.45\textwidth}
            \centering
            \includegraphics[width=\linewidth]{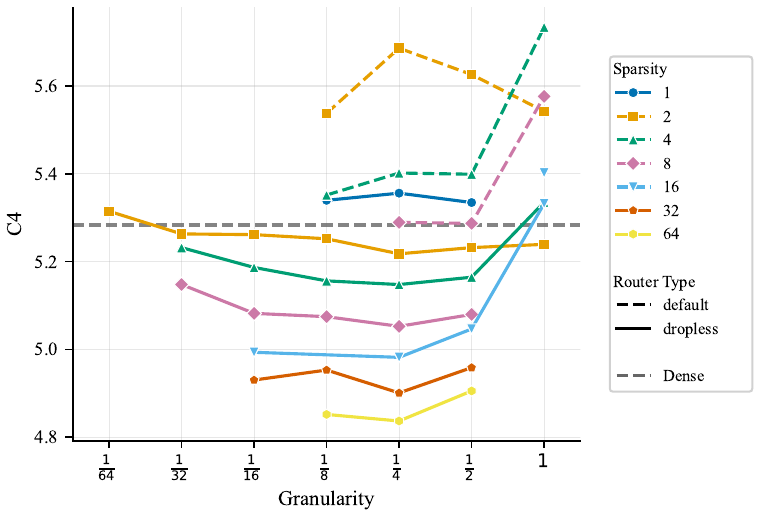}
            \caption{80M active, 80M - 765M total parameters}
        \end{subfigure}
    \end{subfigure}

    \par\bigskip\bigskip
    \begin{subfigure}[t]{0.45\textwidth}
        \centering
        \includegraphics[width=\linewidth]{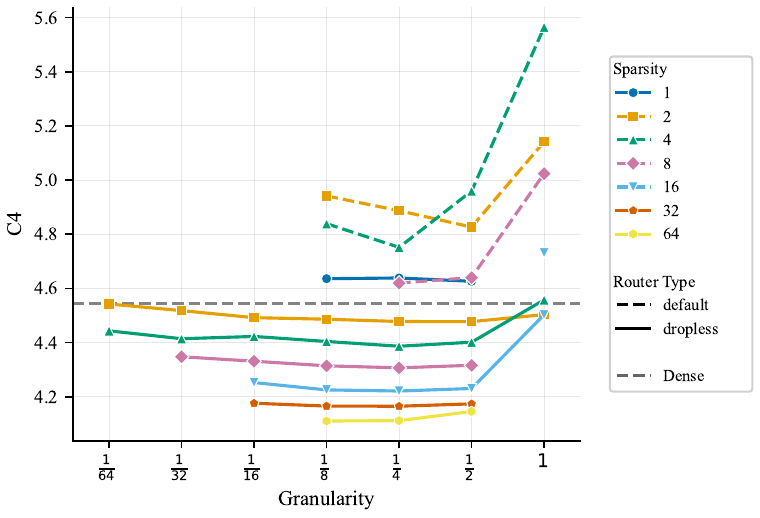}
        \caption{110M active, 110M - 1.4B total parameters}
    \end{subfigure}
    \caption{ 
    \textbf{Dropless routing outperforms default routing (\S\ref{sec:expt_router}).}
    We compare dropless routing to the default setting, which allow tokens to be dropped. Across all scales, we find that dropless routing outperforms or performs comparably to default routing. 
    }
    \label{fig:c4_dropless}
\end{figure*}

\begin{figure*}[ht]
    \centering
    \begin{subfigure}[t]{0.45\textwidth}
        \centering
        \includegraphics[width=\linewidth]{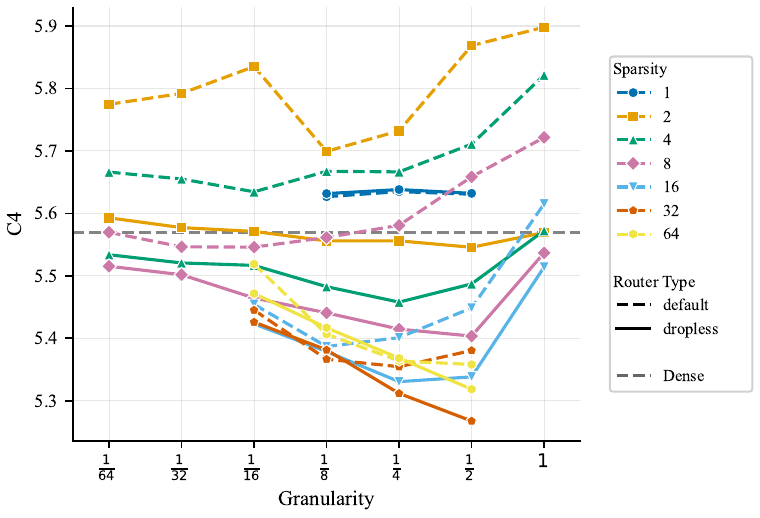}
        \caption{50M active, 50M - 930M total parameters}
    \end{subfigure}
    \hspace{1em}
    \begin{subfigure}[t]{0.45\textwidth}
        \centering
        \includegraphics[width=\linewidth]{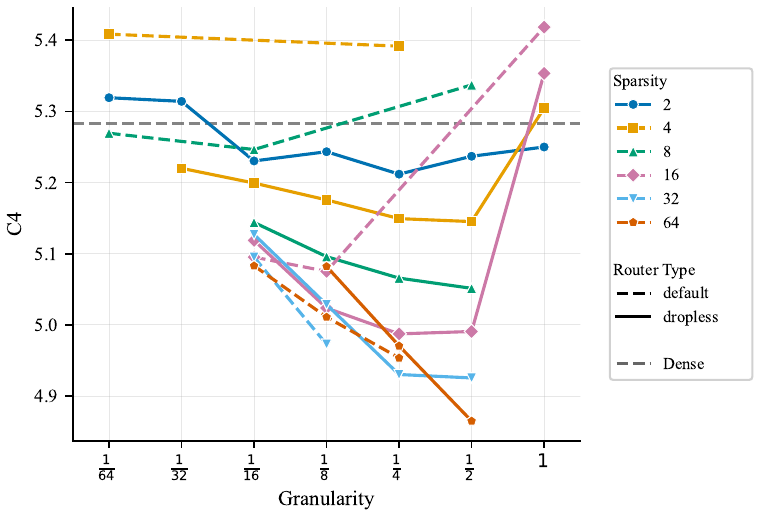}
        \caption{80M active, 80M - 765M total parameters}
    \end{subfigure}
    \caption{
    \textbf{Dropless routing, with bias $\gamma=\num{1e-3}$ (\S\ref{sec:expt_router}).} 
    As in Figure~\ref{fig:lm_avg_dropless}, we compare dropless routing to the default setting, which allow tokens to be dropped. Across all scales, we find that dropless routing outperforms or performs comparably to default routing. We see here with additional higher sparsity default routing runs that as sparsity increases, default routing performance approaches that of dropless routing.
    }
    \label{fig:c4_dropless_with_lf}
\end{figure*}

\begin{figure*}[ht]
    \centering
    \begin{subfigure}[]{\textwidth}
        \centering
        \includegraphics[width=0.46\linewidth]{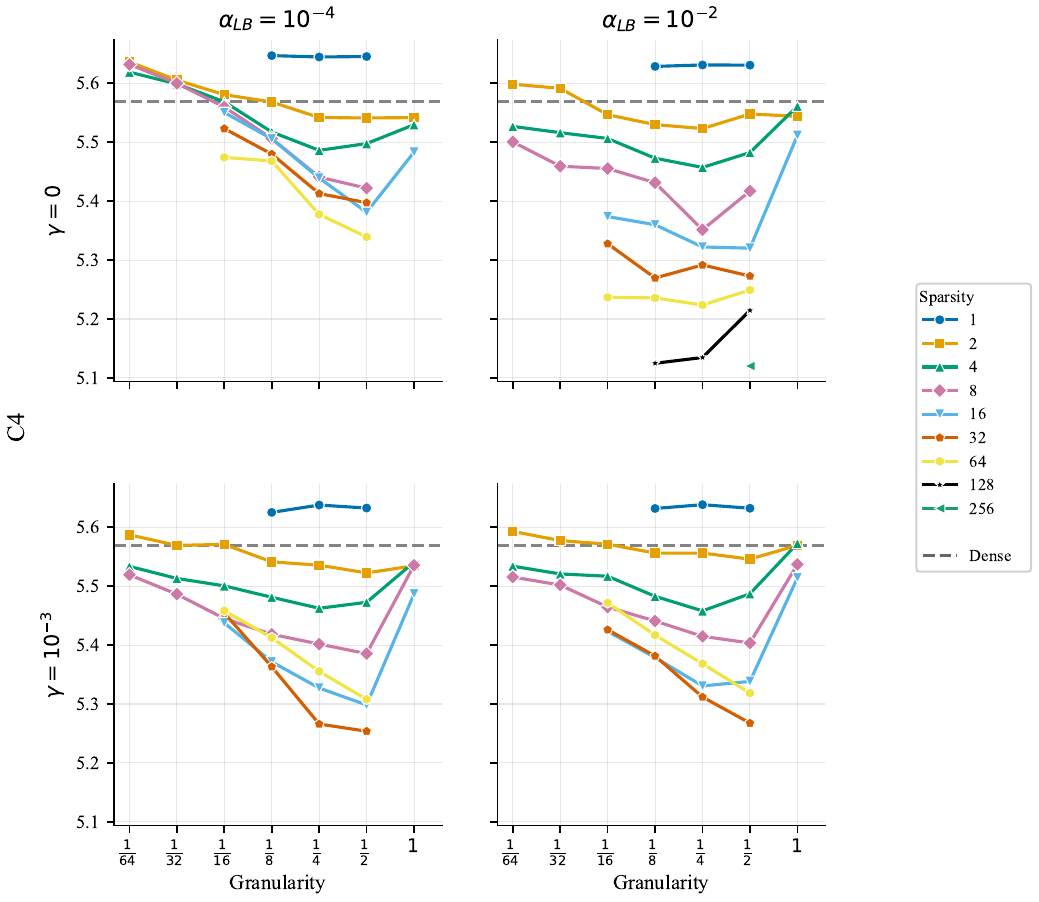}
        \hspace{1em}
        \includegraphics[width=0.46\linewidth]{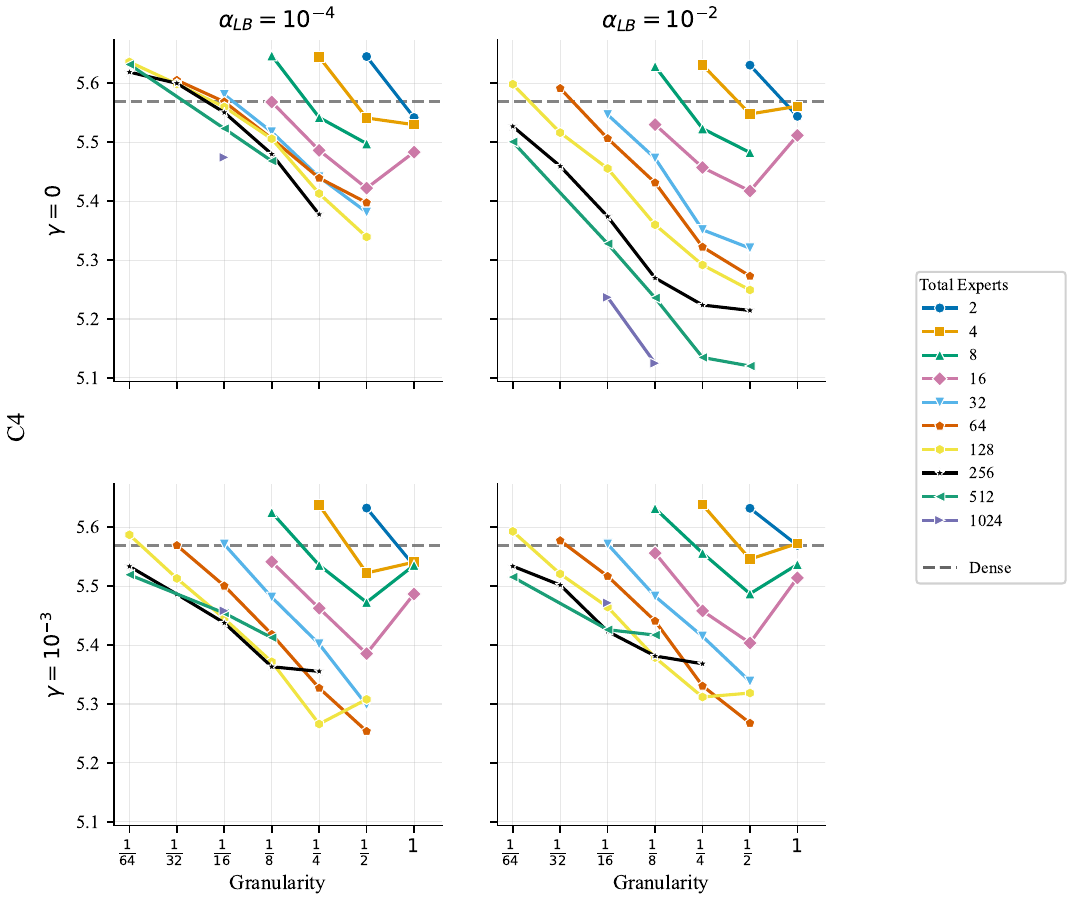}
        \caption{50M active, 50M - 930M total parameters}
    \end{subfigure}
    \par\bigskip\bigskip
    \begin{subfigure}[]{\textwidth}
        \centering
        \includegraphics[width=0.46\linewidth]{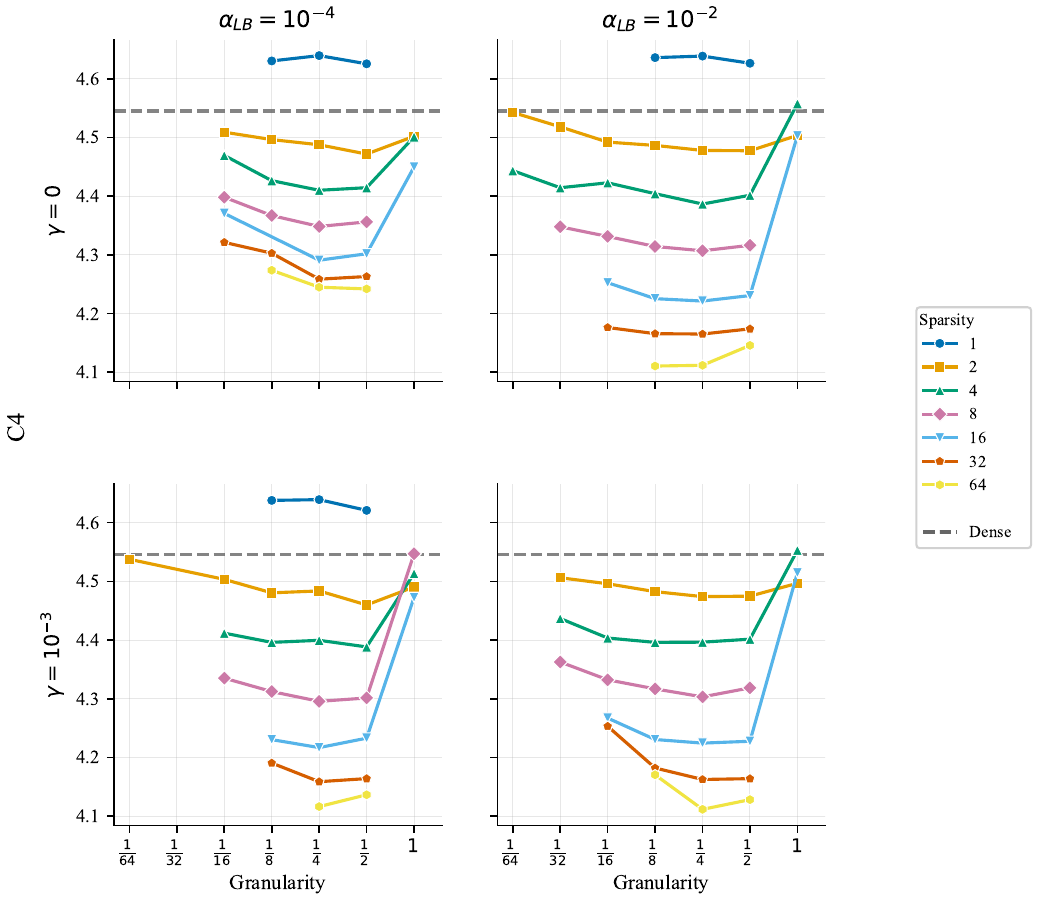}
        \hspace{1em}
        \includegraphics[width=0.46\linewidth]{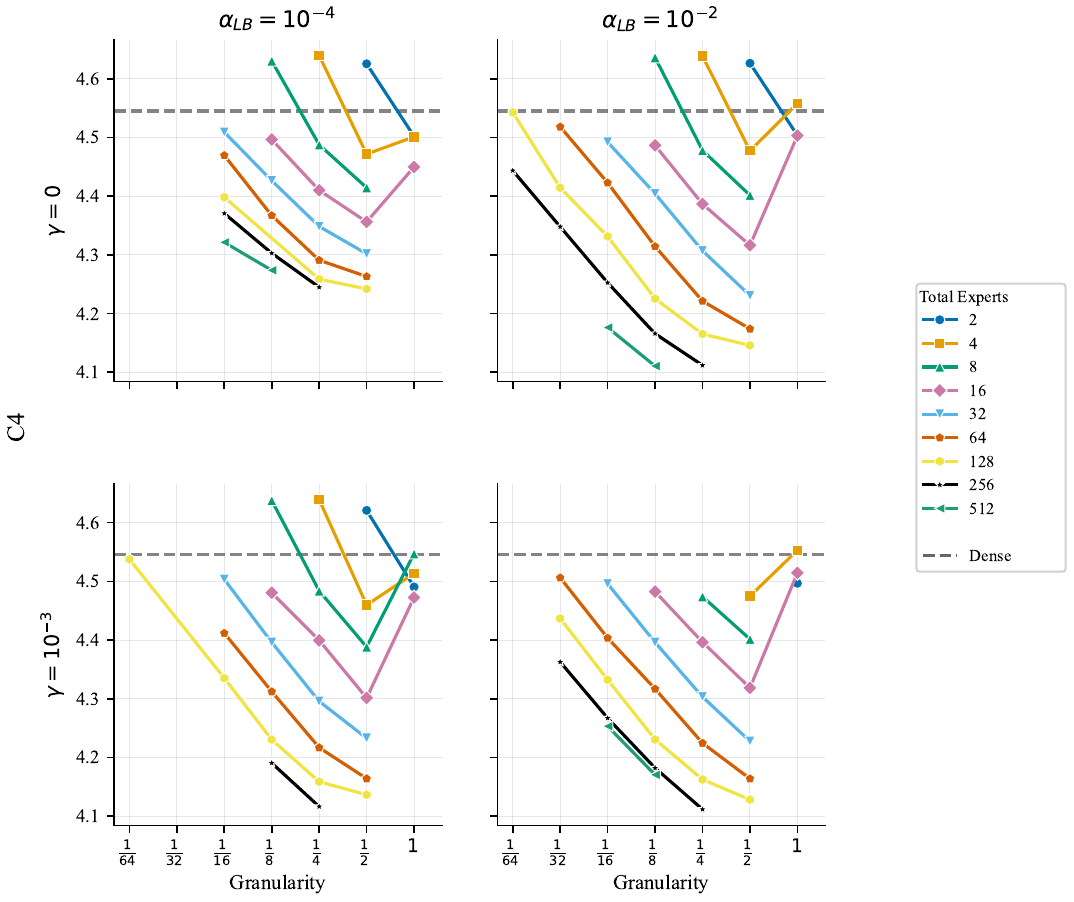}
        \caption{80M active, 80M - 765M total parameters}
    \end{subfigure}
    \par\bigskip\bigskip
    \begin{subfigure}[t]{\textwidth}
        \centering
        \includegraphics[width=0.46\linewidth]{figures/lm/c4_en-validation/ce_loss/lb_sweep_hgn_gxs_110M.pdf}
        \hspace{1em}
        \includegraphics[width=0.46\linewidth]{figures/lm/c4_en-validation/ce_loss/lb_sweep_hgn_gxn_110M.pdf}
        \caption{110M active, 110M - 1.4B total parameters}
    \end{subfigure}

    \end{figure*} 

\clearpage  

\begin{figure*}[ht]
    \addtocounter{figure}{-1}
    \centering
    \begin{subfigure}[t]{\textwidth}
        \centering
        \includegraphics[width=0.46\linewidth]{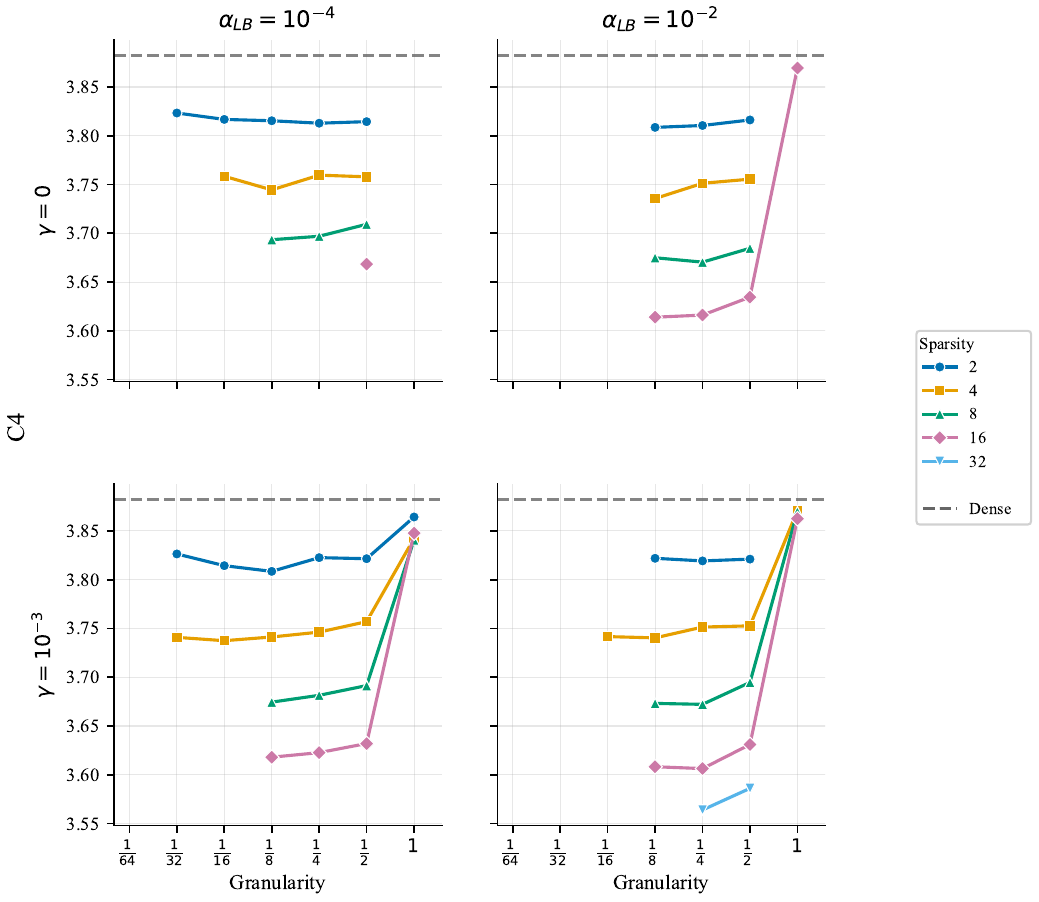}
        \hspace{1em}
        \includegraphics[width=0.46\linewidth]{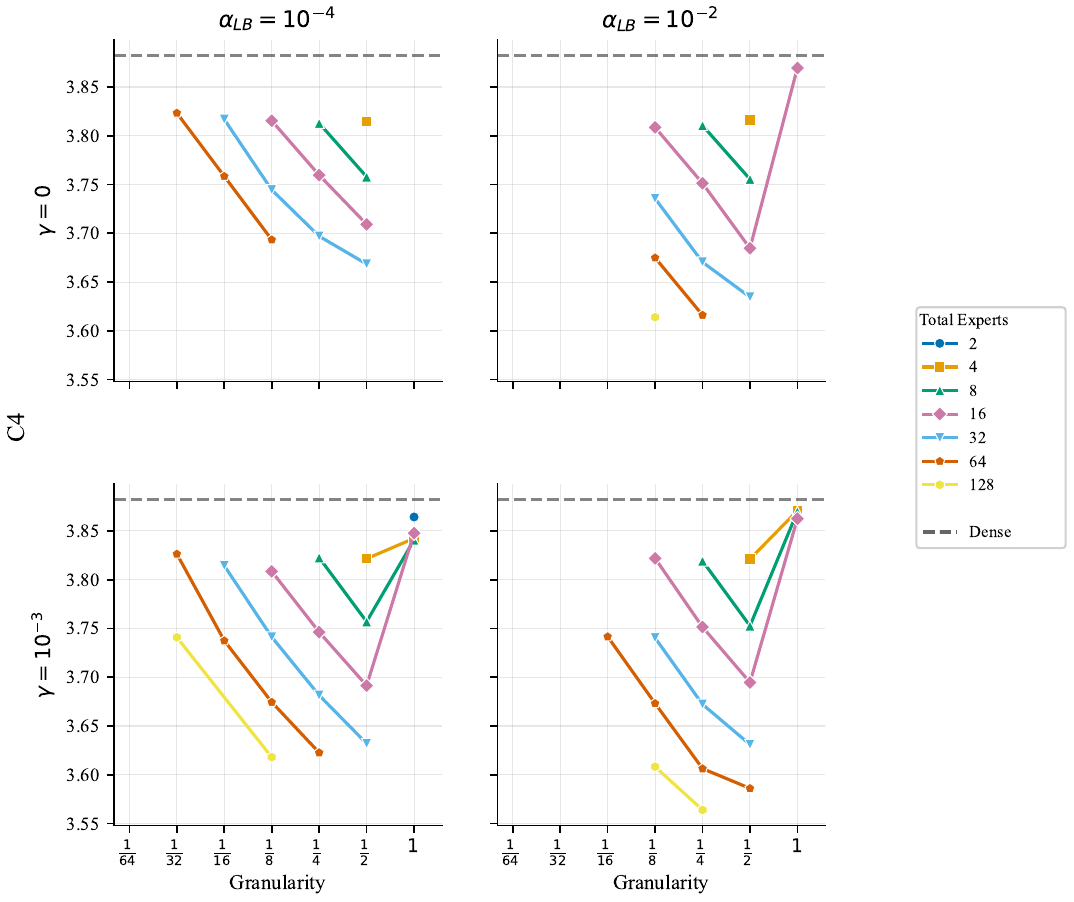}
        \caption{200M active, 200M - 3.3B total parameters}
    \end{subfigure}
    \par\bigskip\bigskip
    \begin{subfigure}[t]{\textwidth}
        \centering
        \includegraphics[width=0.3\linewidth]{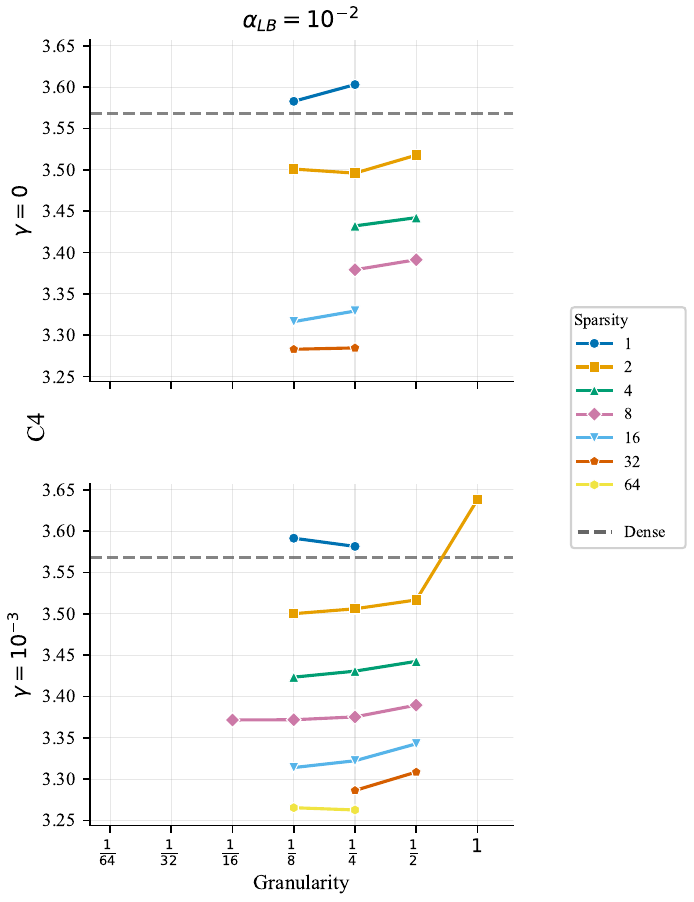}
        \hspace{1em}
        \includegraphics[width=0.3\linewidth]{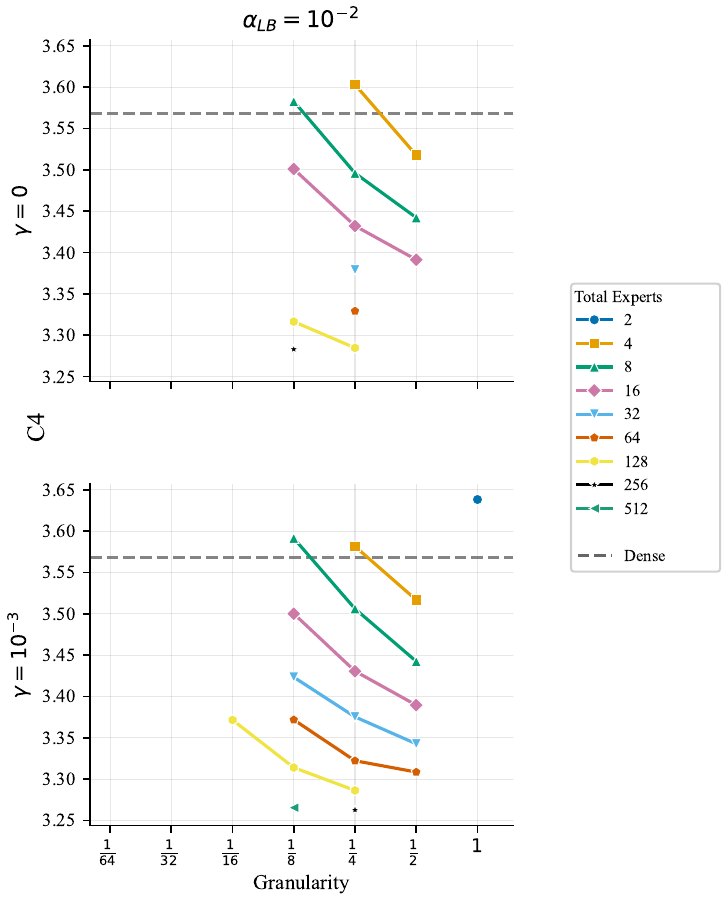}
        \caption{300M active, 300M - 6.6B total parameters}
    \end{subfigure}

    \caption{
    \textbf{Load balancing mechanisms must be tuned correctly (\S\ref{sec:expt_router}).}
    We consider load balancing loss weight $\alpha_{LB} \in \{\num{1e-2}, \num{1e-4}\}$ and loss-free load balancing with bias $\gamma\in\{0, \num{1e-3}\}$ ($\gamma=0$ indicates no loss-free mechanism). Results show that poorly chosen hyperparameters, such as high bias $\gamma = 1e-3$ with total experts $n\geq 512$, may impair performance. However, all settings other than $(\alpha_{LB}=\num{1e-2}, \gamma=\num{1e-3})$ perform comparably for $n \leq 512$, suggesting that a wide range of load balancing settings achieve near-optimal performance. 
    }
    \label{fig:c4_lb}
\end{figure*}

%% file: fig_tex/lm/dolma_books.tex
\begin{figure*}[!ht]
    \centering
        \begin{subfigure}[t]{\textwidth}
        \begin{subfigure}[t]{0.33\textwidth}
            \centering
            \caption*{\scriptsize Fixed total experts (n)}
            \includegraphics[width=\linewidth]{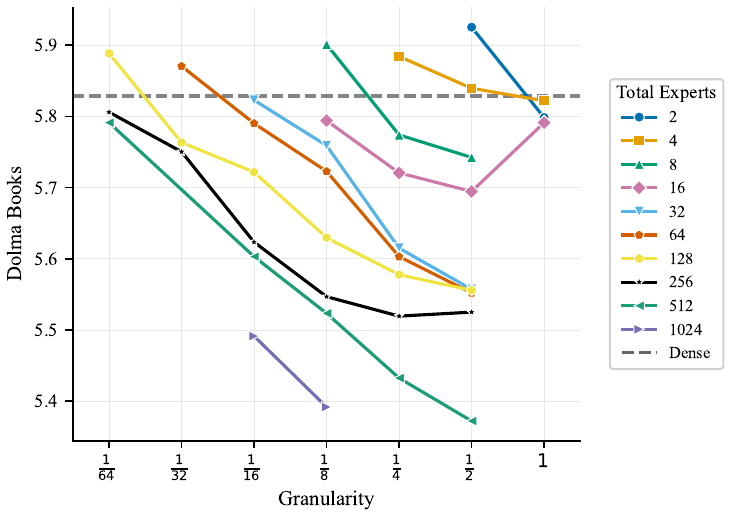}
        \end{subfigure}
        \begin{subfigure}[t]{0.33\textwidth}
            \centering
            \caption*{\scriptsize Fixed granularity (g)}
            \includegraphics[width=\linewidth]{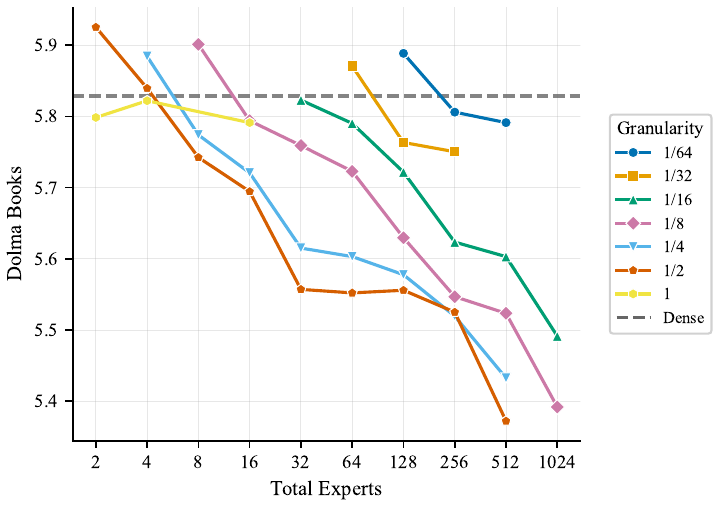}
        \end{subfigure}
        \begin{subfigure}[t]{0.33\textwidth}
            \centering
            \caption*{\scriptsize Fixed activation sparsity (s)}
            \includegraphics[width=\linewidth]{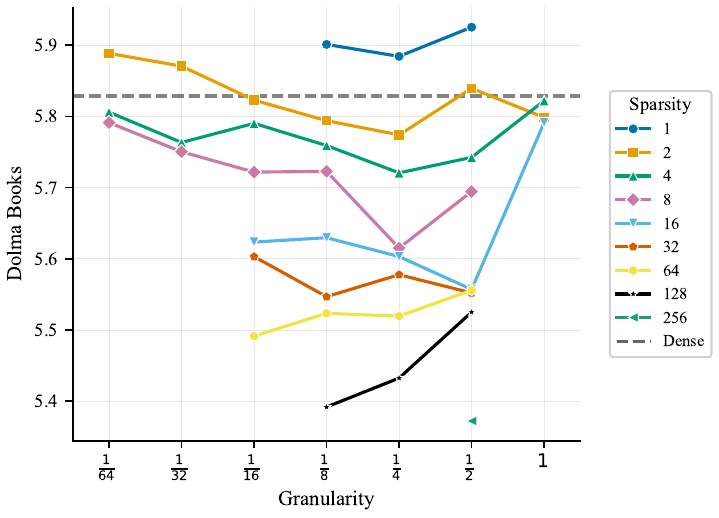}
        \end{subfigure}
        \caption{50M active, 50M - 930M total parameters}
    \end{subfigure}
\par\bigskip\bigskip
    \begin{subfigure}[t]{\textwidth}
        \begin{subfigure}[t]{0.33\textwidth}
            \centering
            \includegraphics[width=\linewidth]{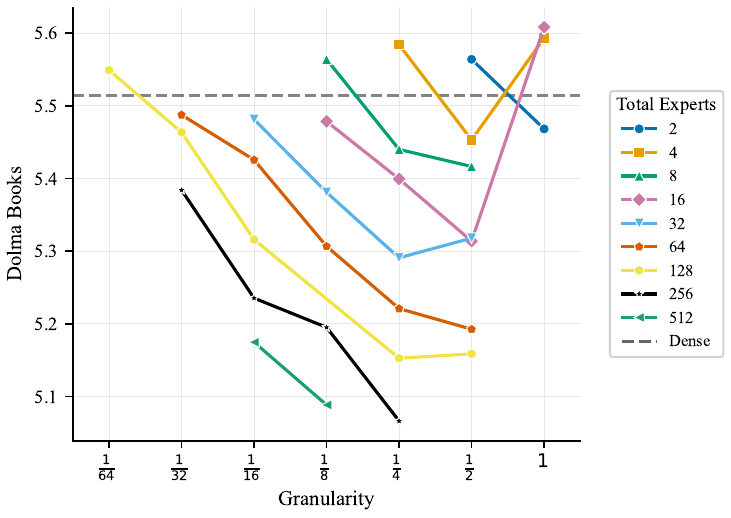}
        \end{subfigure}
        \begin{subfigure}[t]{0.33\textwidth}
            \centering
            \includegraphics[width=\linewidth]{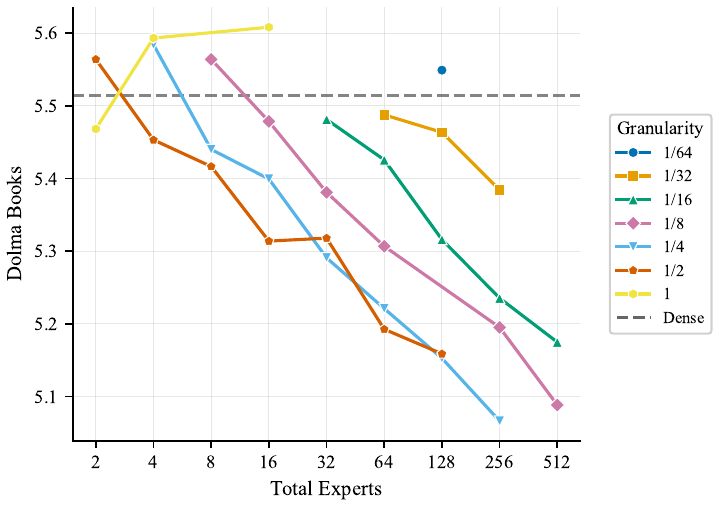}
        \end{subfigure}
        \begin{subfigure}[t]{0.33\textwidth}
            \centering
            \includegraphics[width=\linewidth]{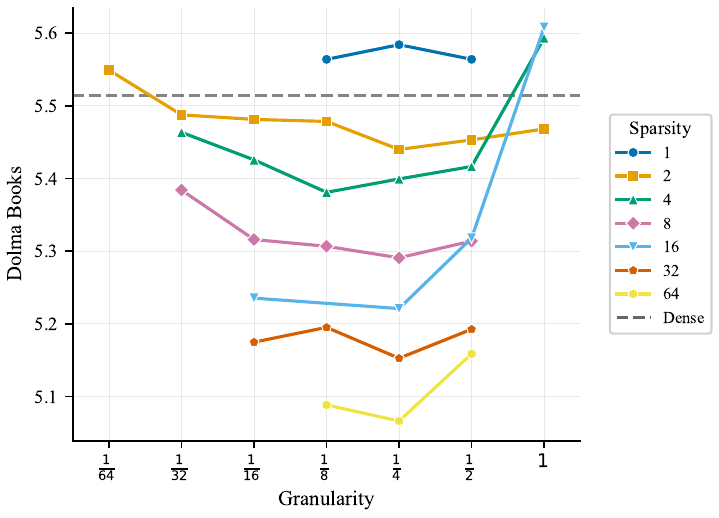}
        \end{subfigure}
        \caption{80M active, 80M - 765M total parameters}
    \end{subfigure}
    \par\bigskip\bigskip
        \begin{subfigure}[t]{\textwidth}
        \begin{subfigure}[t]{0.33\textwidth}
            \centering
            \includegraphics[width=\linewidth]{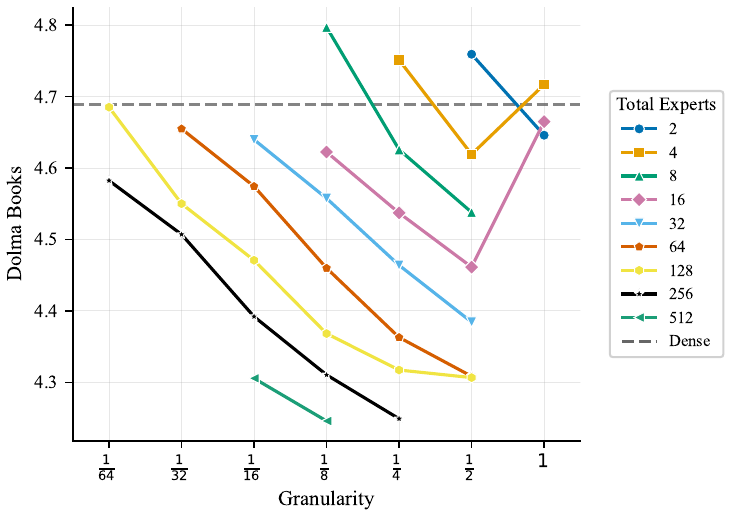}
        \end{subfigure}
        \begin{subfigure}[t]{0.33\textwidth}
            \centering
            \includegraphics[width=\linewidth]{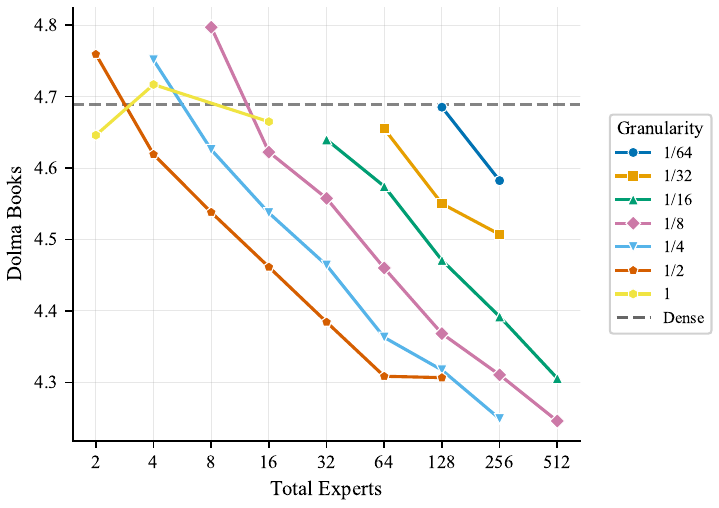}
        \end{subfigure}
        \begin{subfigure}[t]{0.33\textwidth}
            \centering
            \includegraphics[width=\linewidth]{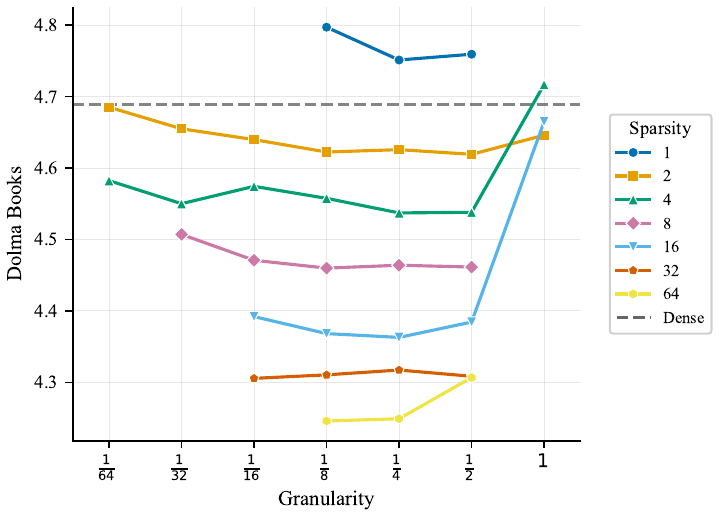}
        \end{subfigure}
        \caption{110M active, 110M - 1.4B total parameters}
    \end{subfigure}
    \end{figure*}

\clearpage  

\begin{figure*}[!ht]
        \addtocounter{figure}{-1}
    \begin{subfigure}[t]{\textwidth}
        \addtocounter{subfigure}{3}
        \begin{subfigure}[t]{0.33\textwidth}
            \centering
            \caption*{\scriptsize Fixed total experts (n)}
            \includegraphics[width=\linewidth]{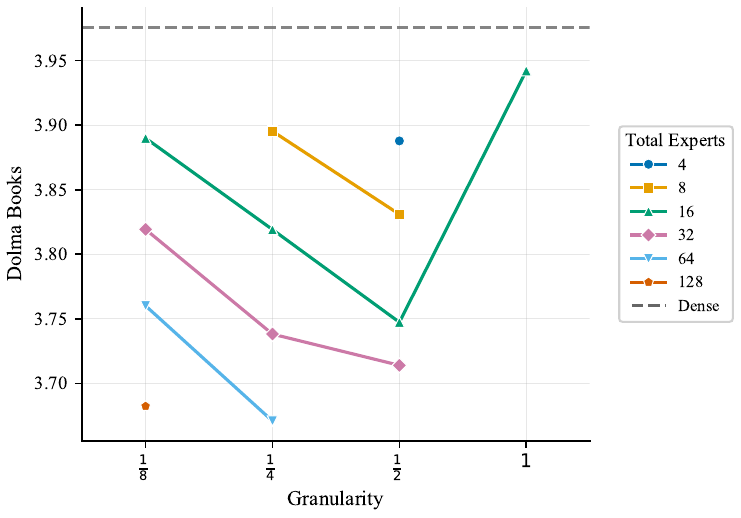}
        \end{subfigure}
        \begin{subfigure}[t]{0.33\textwidth}
            \centering
            \caption*{\scriptsize Fixed granularity (g)}
            \includegraphics[width=\linewidth]{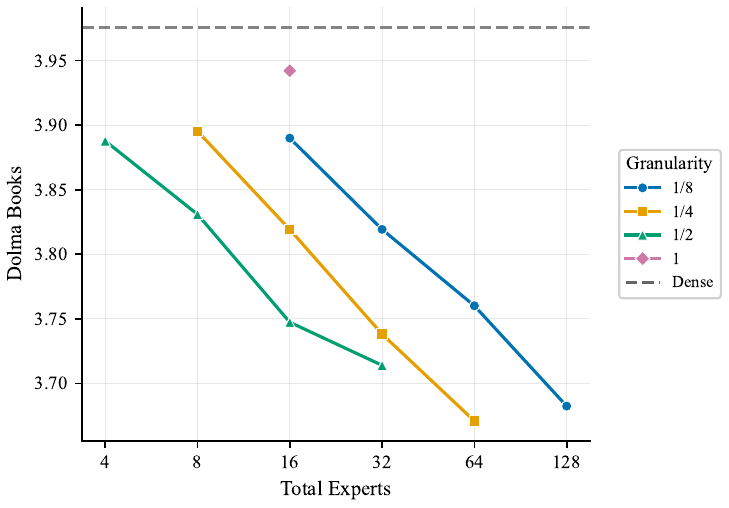}
        \end{subfigure}
        \begin{subfigure}[t]{0.33\textwidth}
            \centering
            \caption*{\scriptsize Fixed activation sparsity (s)}
            \includegraphics[width=\linewidth]{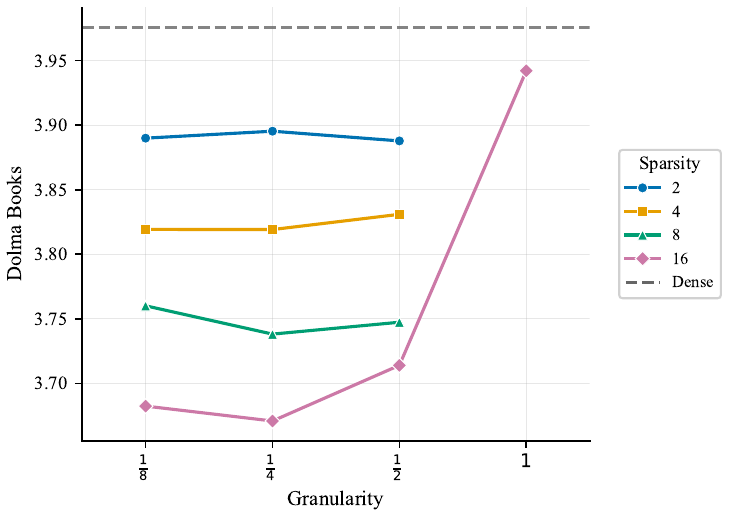}
        \end{subfigure}
        \caption{200M active, 200M - 3.3B total parameters}
    \end{subfigure}
    \par\bigskip\bigskip
        \begin{subfigure}[t]{\textwidth}
        \begin{subfigure}[t]{0.33\textwidth}
            \centering
            \includegraphics[width=\linewidth]{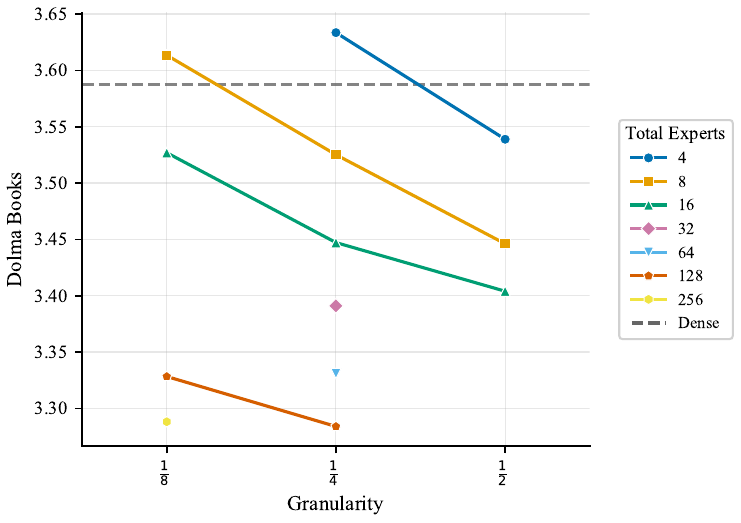}
        \end{subfigure}
        \begin{subfigure}[t]{0.33\textwidth}
            \centering
            \includegraphics[width=\linewidth]{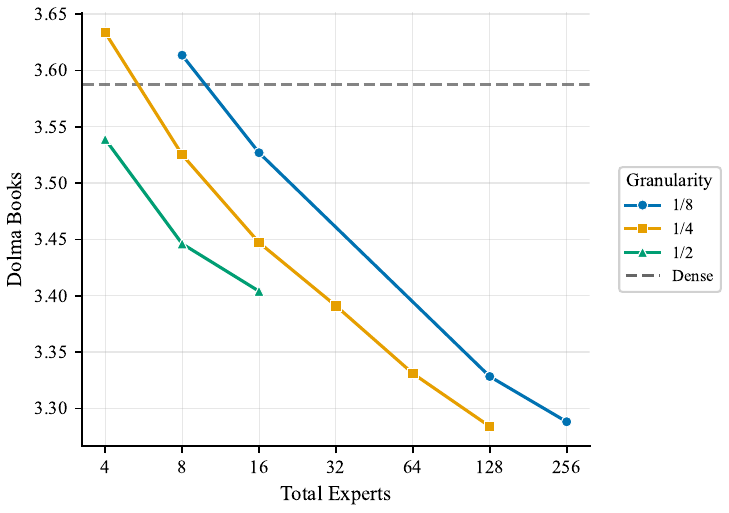}
        \end{subfigure}
        \begin{subfigure}[t]{0.33\textwidth}
            \centering
            \includegraphics[width=\linewidth]{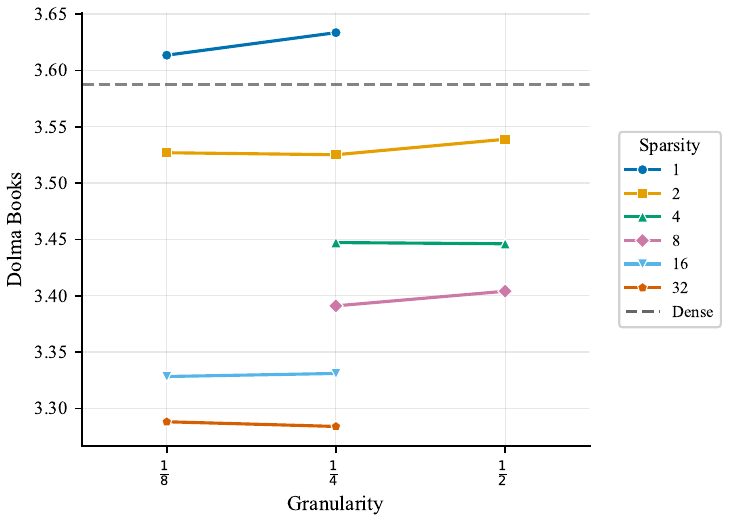}
        \end{subfigure}
        \caption{300M active, 300M - 6.6B total parameters}
    \end{subfigure}

    \caption{
    \textbf{Increasing inactive expert parameters via expert size (left) or total count (center) improves performance in MoEs (\S\ref{sec:expt_main}).} This effect is seen both when holding total number of experts fixed (left) and when holding expert granularity fixed (center). In general, increasing total parameters results in improved performance.  \textbf{Optimal tradeoff between expert count and granularity varies in MoEs (right). (\S\ref{sec:expt_main})}
    At each activation sparsity $s$ (equivalently, at each total parameter count), the optimal (total expert count, expert granularity) configuration varies. As $s$ increases, optimal expert granularity remains nearly fixed, suggesting that sparsity should be scaled up primarily by increasing total expert count $n$, while maintaining a near constant, slowly increasing expert granularity $g$. 
    }
    \label{fig:dolma_books_experts}
\end{figure*}

\begin{figure*}[!ht]
    \centering
    
    \begin{subfigure}[t]{0.46\textwidth}
        \centering
        \includegraphics[width=\linewidth]{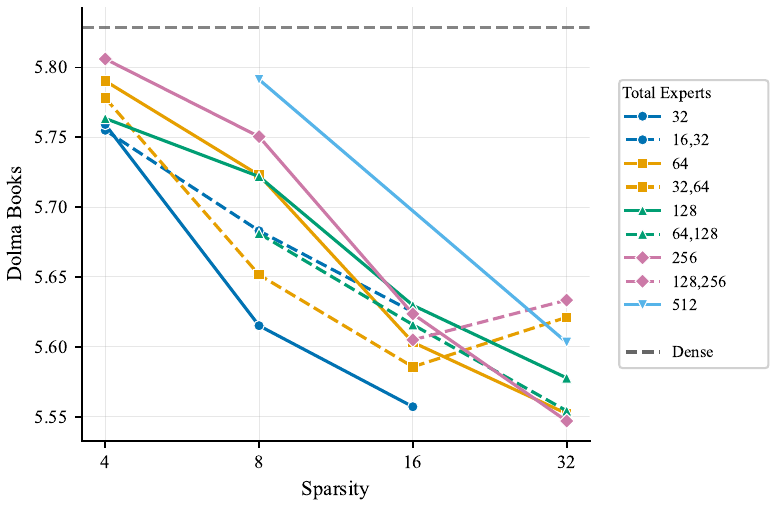}
        \caption{50M active, 50M - 930M total parameters}
    \end{subfigure}
    \vspace{1em}
    \begin{subfigure}[t]{0.46\textwidth}
        \centering
        \includegraphics[width=\linewidth]{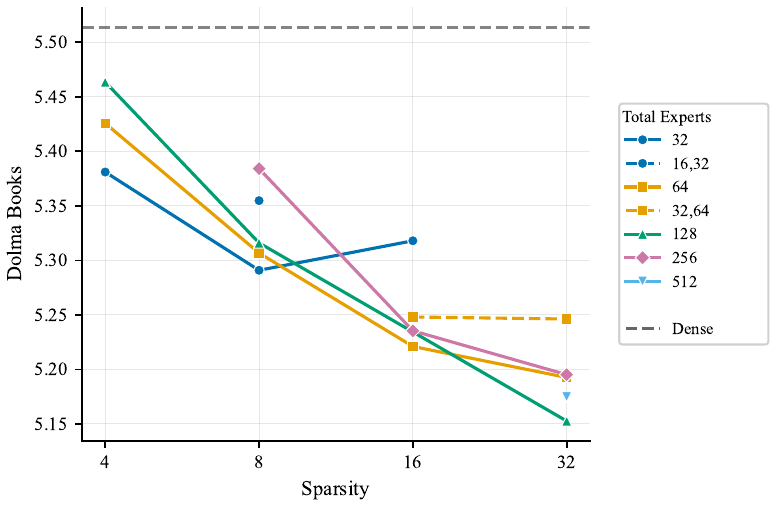}
        \caption{80M active, 80M - 765M total parameters}
    \end{subfigure}
    \caption{
    \textbf{Heterogeneity of expert size alone does not improve MoE performance (\S\ref{sec:expt_hetgen}).} To explore the potential benefits of their architectural flexibility, we compare heterogeneous MoEs (indicated by dotted lines) to active- and total-parameter-matched homogeneous MoEs. Heterogeneity alone does not result in performance gains, as, at each activation sparsity $s$, heterogeneous MoEs with $n_1, n_2 = a, b$ lie between or near the 2 closest homogeneous MoEs, with $n=a$ and with $n=b$.
    }
    \label{fig:dolma_books_het}
\end{figure*}

\begin{figure*}[!ht]
    \centering
    
    \begin{subfigure}[t]{1.0\textwidth}
        \centering
        \includegraphics[width=\linewidth]{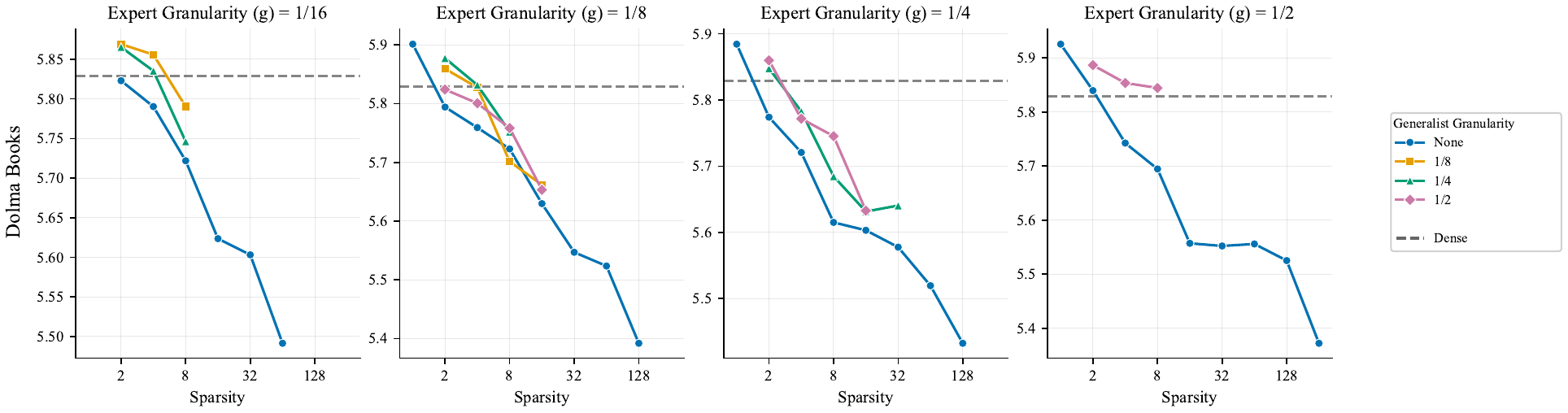}
        \caption{50M active, 50M - 930M total parameters}
    \end{subfigure}
    \par\bigskip\bigskip
    \begin{subfigure}[t]{1.0\textwidth}
        \centering
        \includegraphics[width=\linewidth]{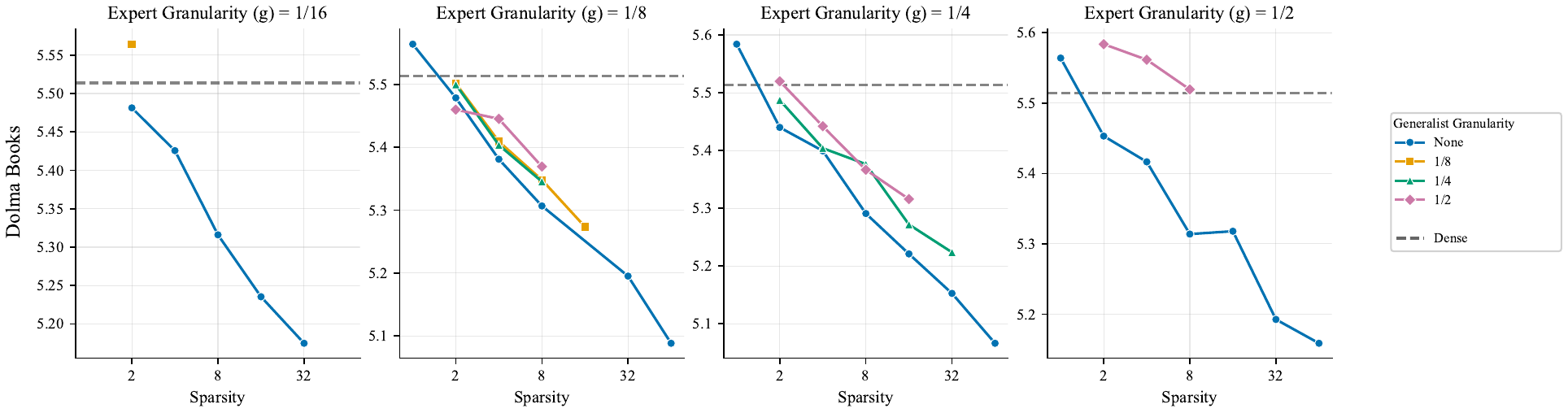}
        \caption{80M active, 80M - 765M total parameters}
    \end{subfigure}
    \par\bigskip\bigskip
    \begin{subfigure}[t]{1.0\textwidth}
        \centering
        \includegraphics[width=\linewidth]{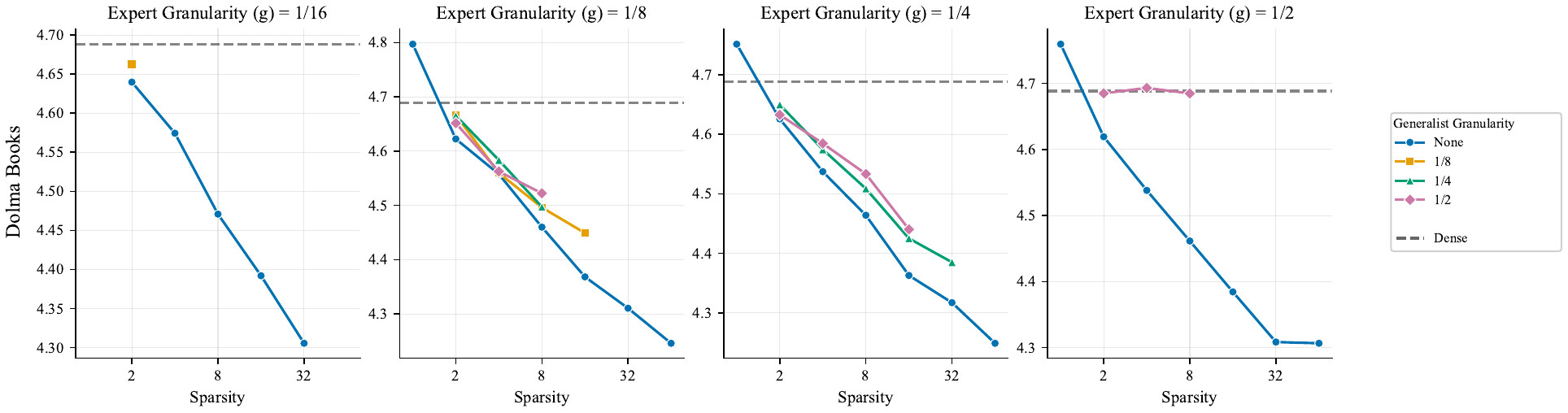}
        \caption{110M active, 110M - 1.4B total parameters}
    \end{subfigure}
    \caption{
    \textbf{The inclusion of a generalist consistently degrades performance in homogeneous MoEs (\S\ref{sec:expt_hetgen}).}
    We train MoE LMs which consist of some routed experts with granularity $g$, as well as a generalist with granularity $g_{gen}\in \{\frac{1}{2}, \frac{1}{4}, \frac{1}{8}\} $. We compare to settings with no generalist, only routed experts with granularity $g$. In all settings and configurations, the addition of any granularity generalist results in comparable or degraded performance. 
    }
    \label{fig:dolma_books_gen}
\end{figure*}

\begin{figure*}[ht]
    \centering
    \begin{subfigure}[t]{1.0\textwidth}
        \centering
        \includegraphics[width=\linewidth]{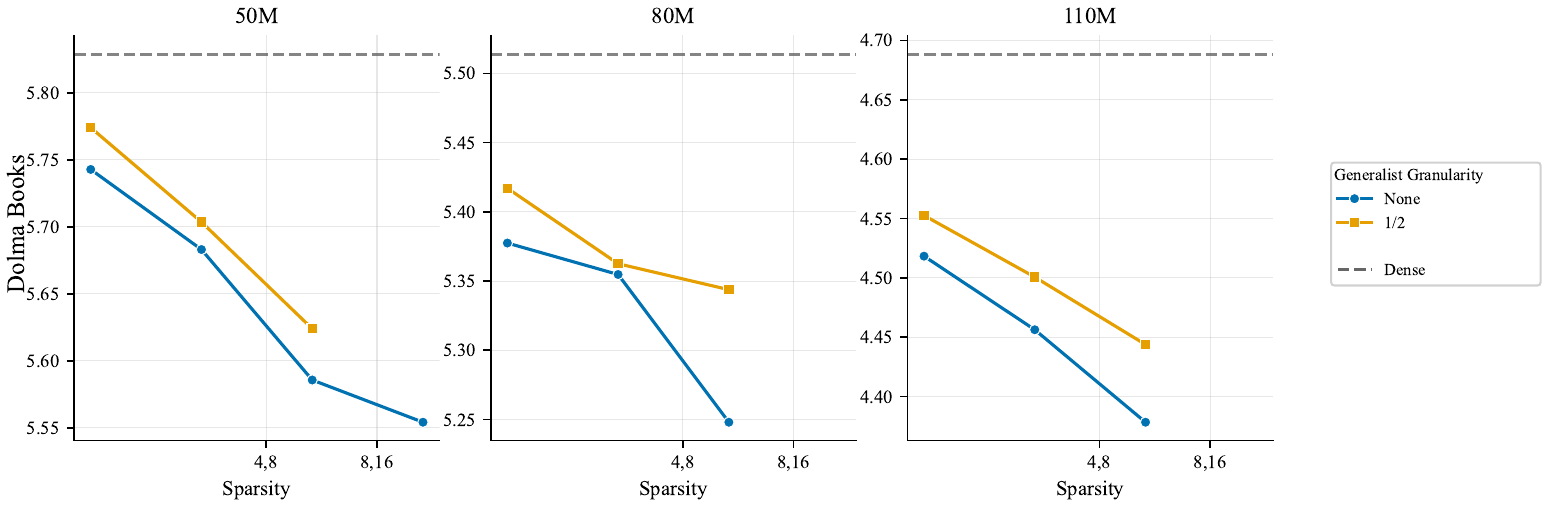}
    \end{subfigure}
    \caption{
    \textbf{The inclusion of a generalist consistently degrades performance in heterogeneous MoEs (\S\ref{sec:expt_hetgen}).}
    We train heterogeneous MoE LMs which consist of  routed experts with granularity $g_1, g_2$, as well as a generalist with granularity $g_{gen} = \frac{1}{2}$. We compare to settings with no generalist. In all settings and configurations, the addition of a generalist results in comparable or degraded performance. 
    }
    \label{fig:dolma_books_hetgen}
\end{figure*}

\begin{figure*}[ht]
    \centering
    \begin{subfigure}[t]{\textwidth}
        \centering
        \begin{subfigure}[t]{0.45\textwidth}
            \includegraphics[width=\linewidth]{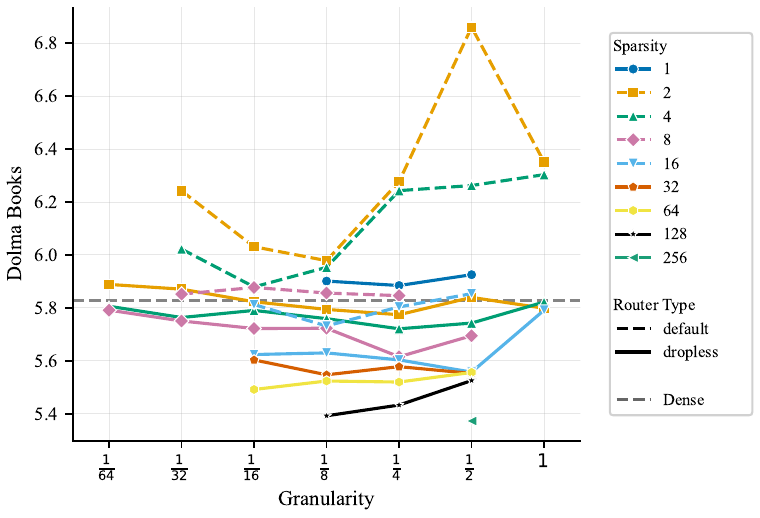}
            \caption{50M active, 50M - 930M total parameters}
        \end{subfigure}
    \hspace{1em}
        \begin{subfigure}[t]{0.45\textwidth}
            \centering
            \includegraphics[width=\linewidth]{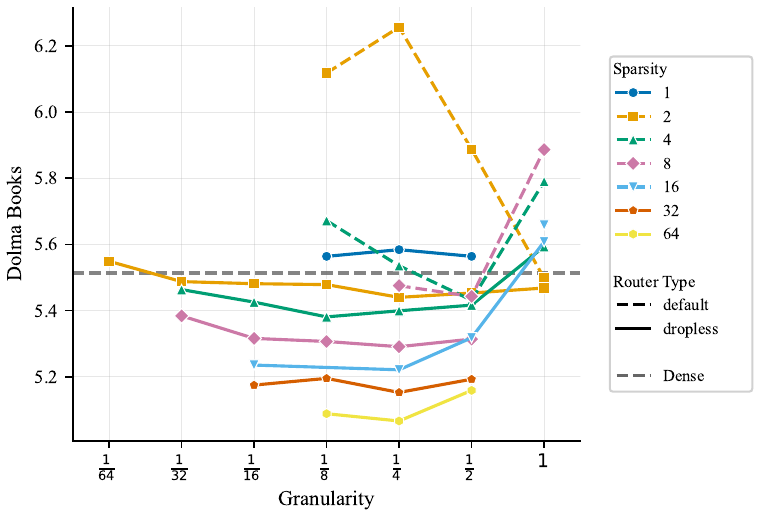}
            \caption{80M active, 80M - 765M total parameters}
        \end{subfigure}
    \end{subfigure}

    \par\bigskip\bigskip
    \begin{subfigure}[t]{0.45\textwidth}
        \centering
        \includegraphics[width=\linewidth]{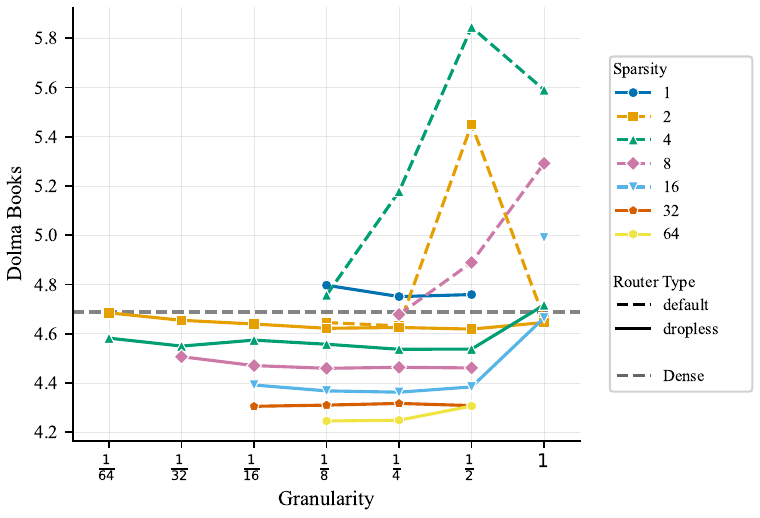}
        \caption{110M active, 110M - 1.4B total parameters}
    \end{subfigure}
    \caption{ 
    \textbf{Dropless routing outperforms default routing (\S\ref{sec:expt_router}).}
    We compare dropless routing to the default setting, which allow tokens to be dropped. Across all scales, we find that dropless routing outperforms or performs comparably to default routing. 
    }
    \label{fig:dolma_books_dropless}
\end{figure*}

\begin{figure*}[ht]
    \centering
    \begin{subfigure}[t]{0.45\textwidth}
        \centering
        \includegraphics[width=\linewidth]{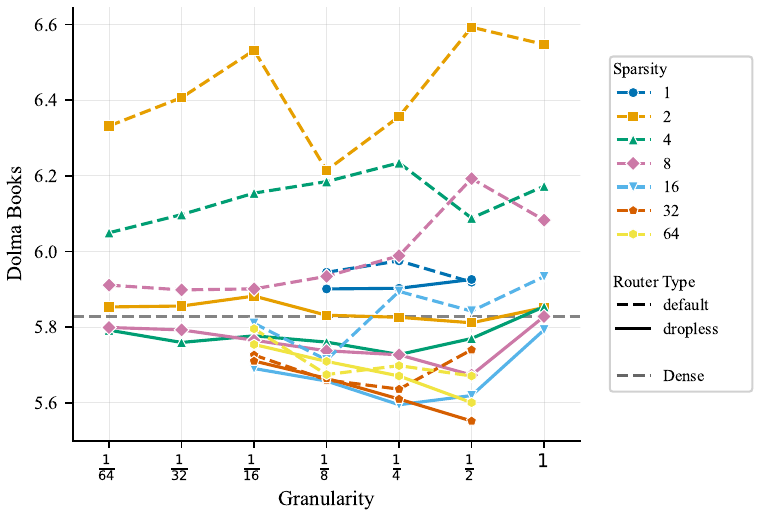}
        \caption{50M active, 50M - 930M total parameters}
    \end{subfigure}
    \hspace{1em}
    \begin{subfigure}[t]{0.45\textwidth}
        \centering
        \includegraphics[width=\linewidth]{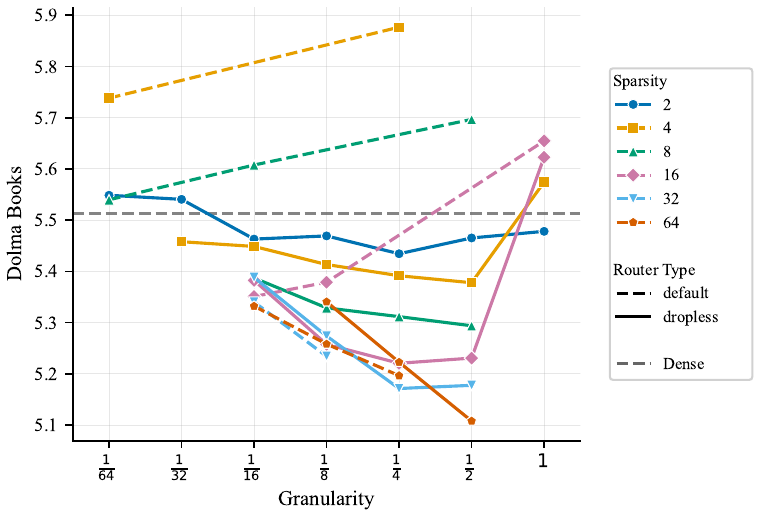}
        \caption{80M active, 80M - 765M total parameters}
    \end{subfigure}
    \caption{
    \textbf{Dropless routing, with bias $\gamma=\num{1e-3}$ (\S\ref{sec:expt_router}).} 
    As in Figure~\ref{fig:lm_avg_dropless}, we compare dropless routing to the default setting, which allow tokens to be dropped. Across all scales, we find that dropless routing outperforms or performs comparably to default routing. We see here with additional higher sparsity default routing runs that as sparsity increases, default routing performance approaches that of dropless routing.
    }
    \label{fig:dolma_books_dropless_with_lf}
\end{figure*}

\begin{figure*}[ht]
    \centering
    \begin{subfigure}[]{\textwidth}
        \centering
        \includegraphics[width=0.46\linewidth]{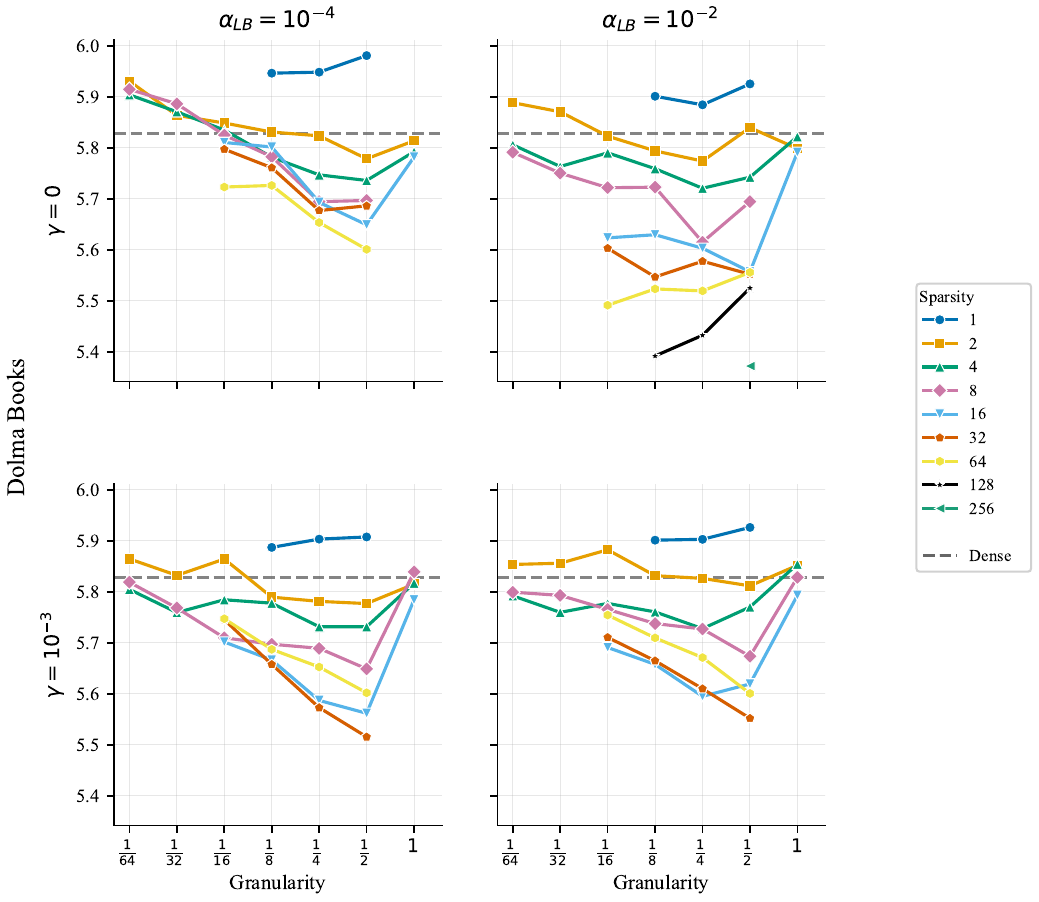}
        \hspace{1em}
        \includegraphics[width=0.46\linewidth]{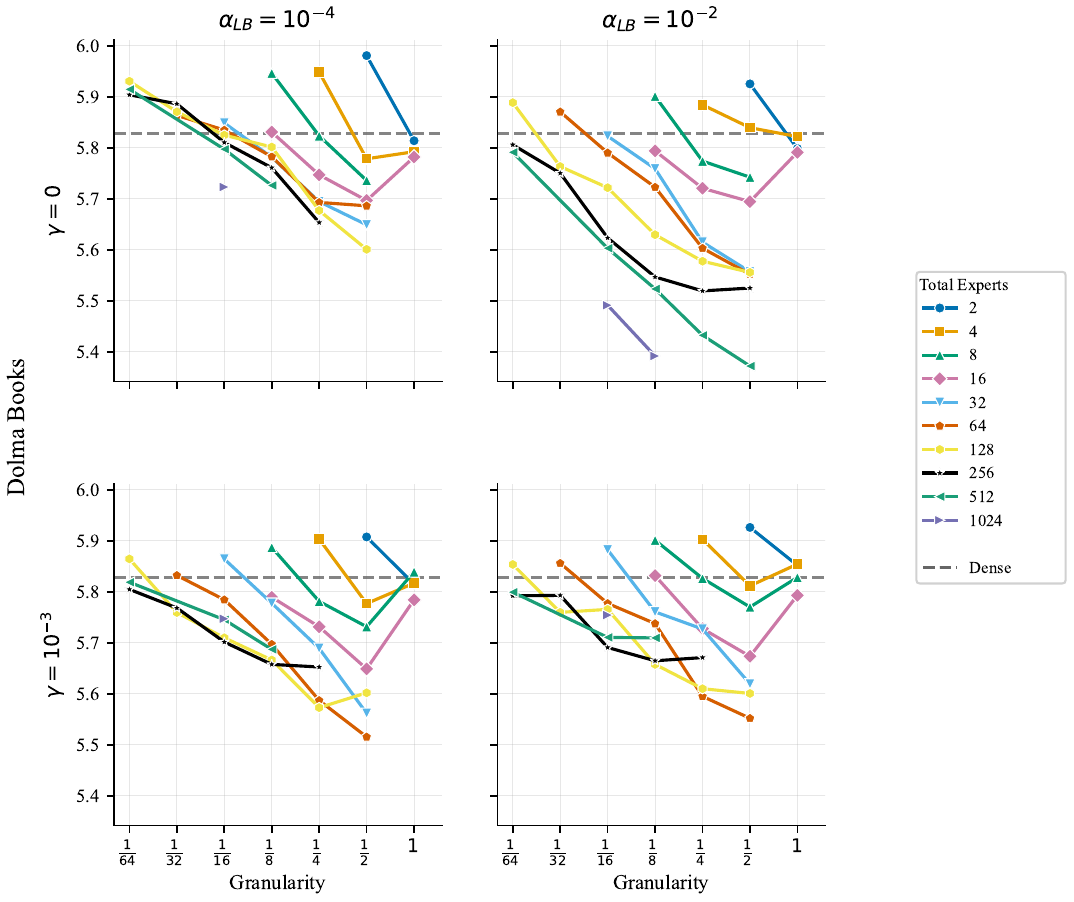}
        \caption{50M active, 50M - 930M total parameters}
    \end{subfigure}
    \par\bigskip\bigskip
    \begin{subfigure}[]{\textwidth}
        \centering
        \includegraphics[width=0.46\linewidth]{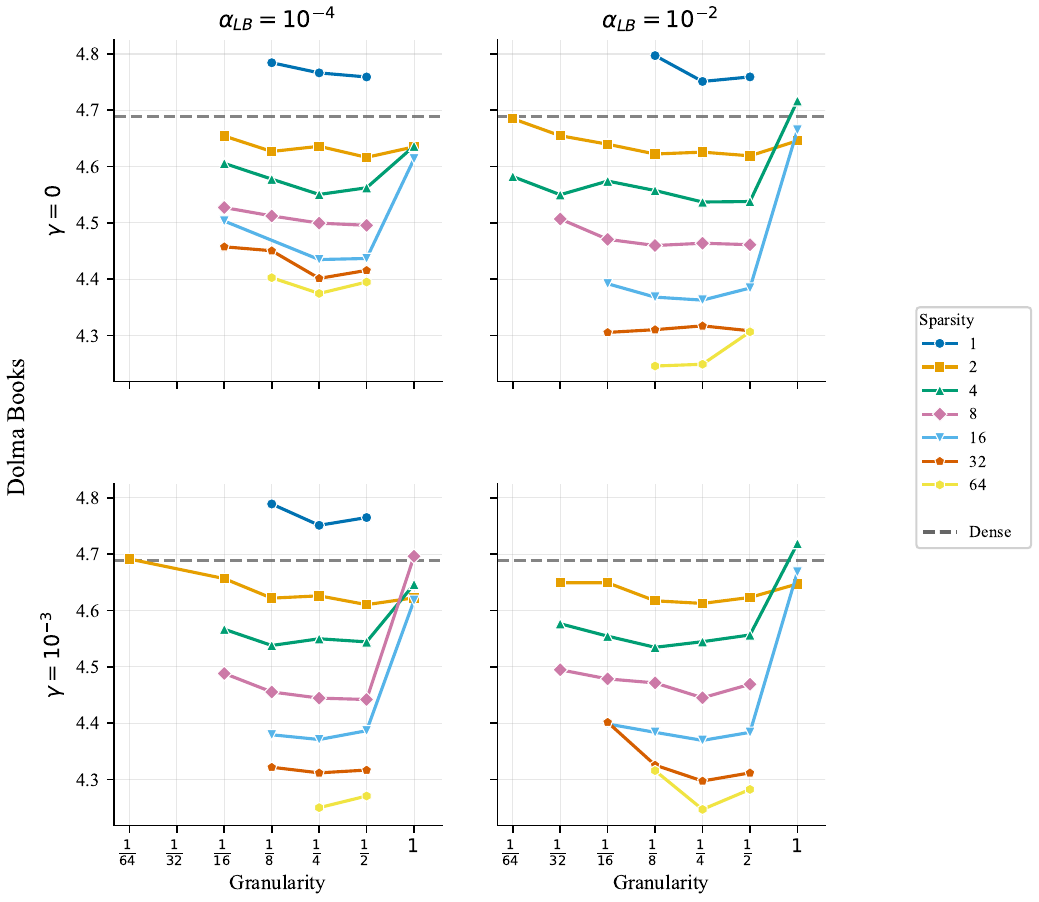}
        \hspace{1em}
        \includegraphics[width=0.46\linewidth]{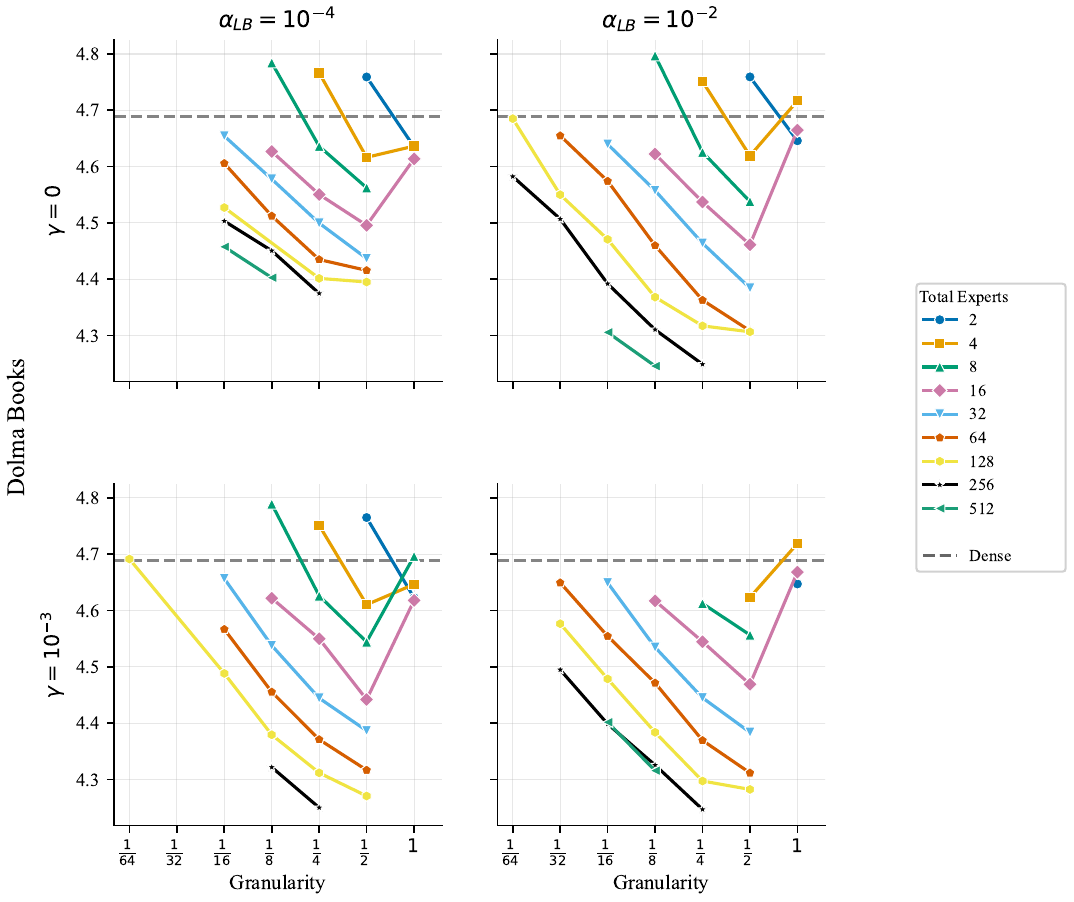}
        \caption{80M active, 80M - 765M total parameters}
    \end{subfigure}
    \par\bigskip\bigskip
    \begin{subfigure}[t]{\textwidth}
        \centering
        \includegraphics[width=0.46\linewidth]{figures/lm/dolma_books-validation/ce_loss/lb_sweep_hgn_gxs_110M.pdf}
        \hspace{1em}
        \includegraphics[width=0.46\linewidth]{figures/lm/dolma_books-validation/ce_loss/lb_sweep_hgn_gxn_110M.pdf}
        \caption{110M active, 110M - 1.4B total parameters}
    \end{subfigure}

    \end{figure*} 

\clearpage  

\begin{figure*}[ht]
    \addtocounter{figure}{-1}
    \centering
    \begin{subfigure}[t]{\textwidth}
        \centering
        \includegraphics[width=0.46\linewidth]{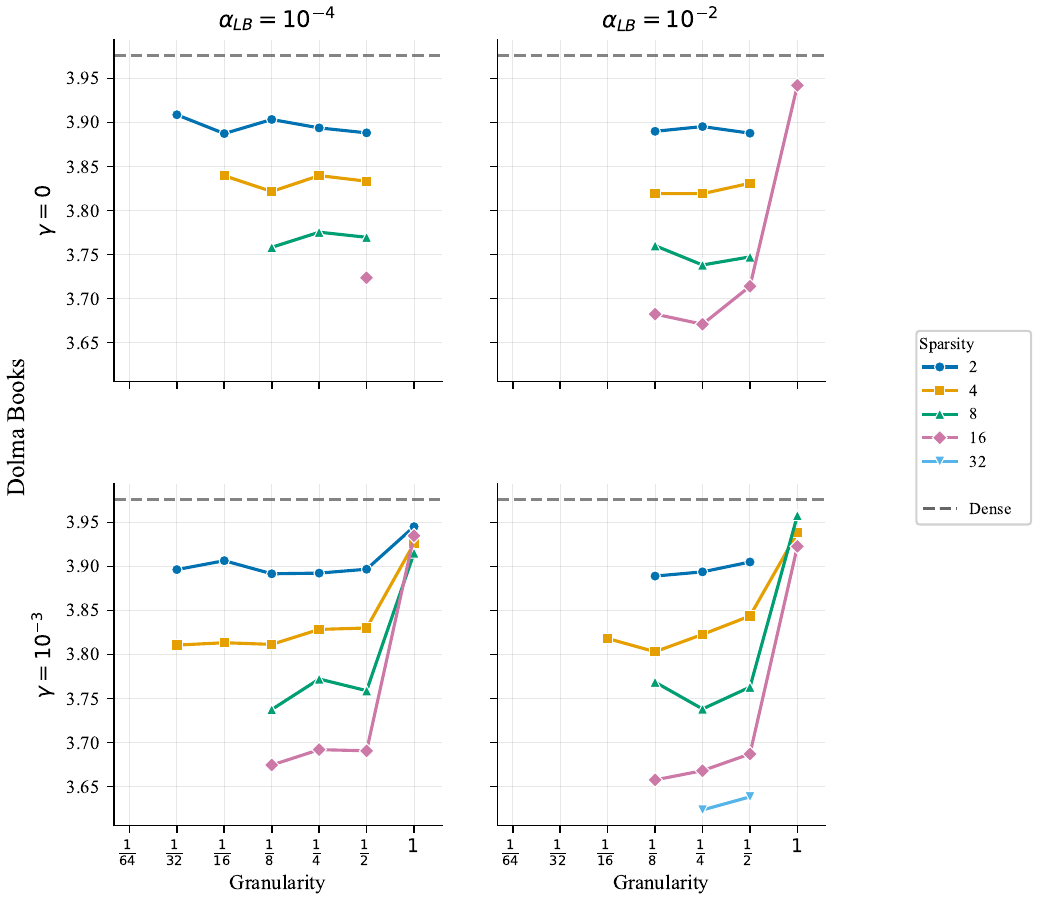}
        \hspace{1em}
        \includegraphics[width=0.46\linewidth]{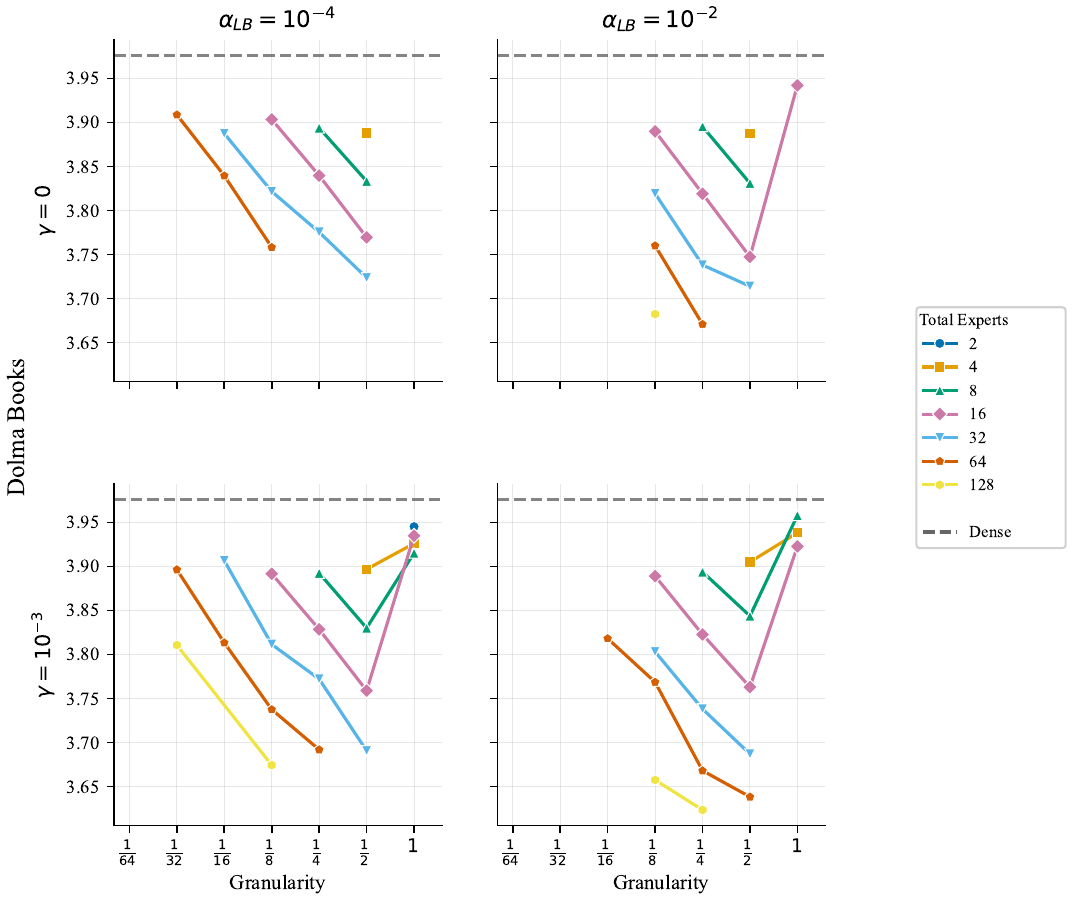}
        \caption{200M active, 200M - 3.3B total parameters}
    \end{subfigure}
    \par\bigskip\bigskip
    \begin{subfigure}[t]{\textwidth}
        \centering
        \includegraphics[width=0.3\linewidth]{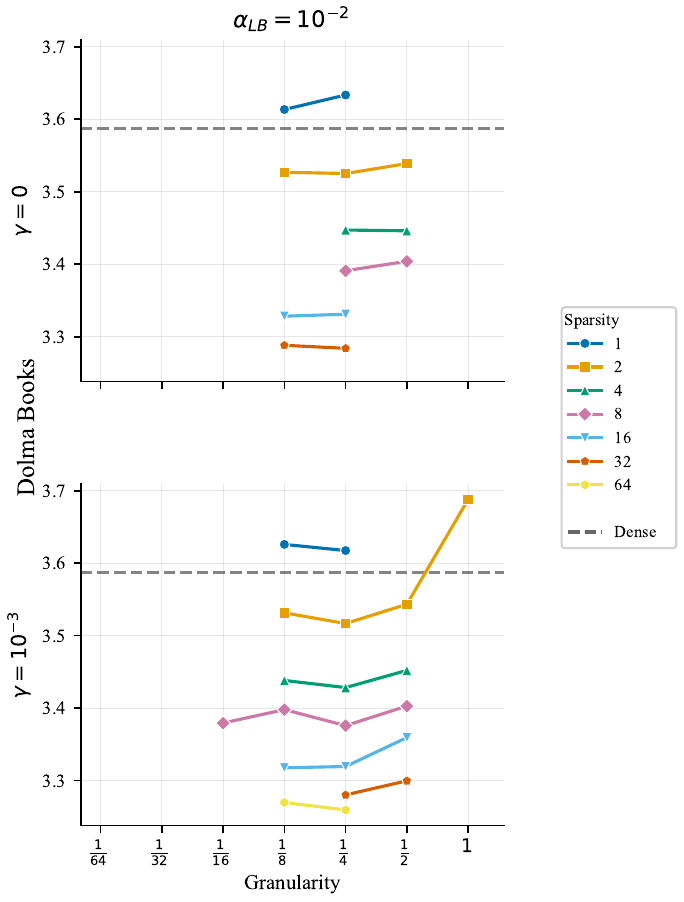}
        \hspace{1em}
        \includegraphics[width=0.3\linewidth]{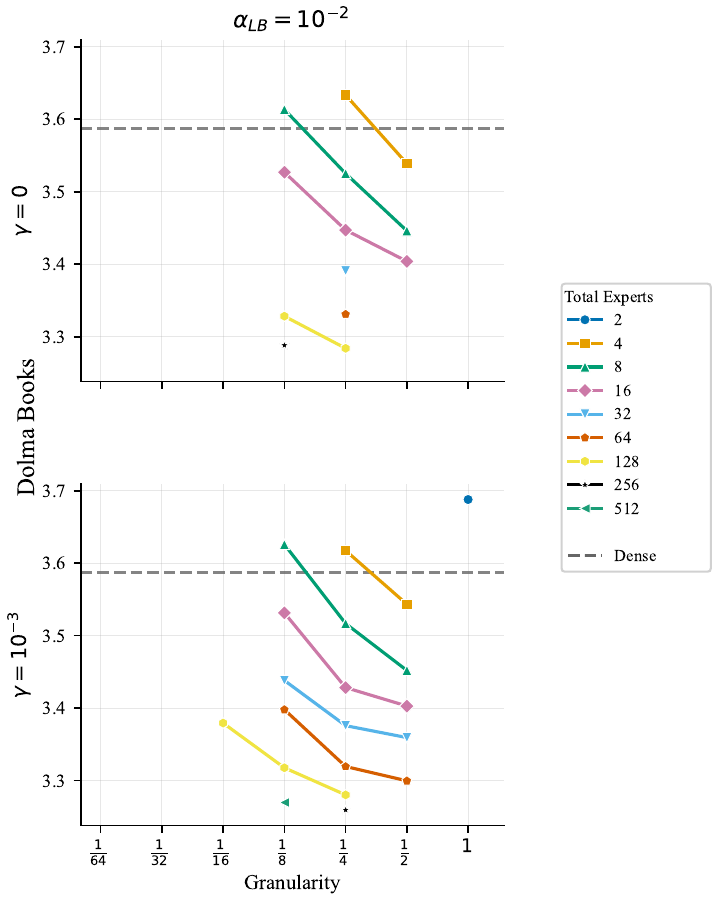}
        \caption{300M active, 300M - 6.6B total parameters}
    \end{subfigure}

    \caption{
    \textbf{Load balancing mechanisms must be tuned correctly (\S\ref{sec:expt_router}).}
    We consider load balancing loss weight $\alpha_{LB} \in \{\num{1e-2}, \num{1e-4}\}$ and loss-free load balancing with bias $\gamma\in\{0, \num{1e-3}\}$ ($\gamma=0$ indicates no loss-free mechanism). Results show that poorly chosen hyperparameters, such as high bias $\gamma = 1e-3$ with total experts $n\geq 512$, may impair performance. However, all settings other than $(\alpha_{LB}=\num{1e-2}, \gamma=\num{1e-3})$ perform comparably for $n \leq 512$, suggesting that a wide range of load balancing settings achieve near-optimal performance. 
    }
    \label{fig:dolma_books_lb}
\end{figure*}

%% file: fig_tex/lm/dolma_common_crawl.tex
\begin{figure*}[!ht]
    \centering
        \begin{subfigure}[t]{\textwidth}
        \begin{subfigure}[t]{0.33\textwidth}
            \centering
            \caption*{\scriptsize Fixed total experts (n)}
            \includegraphics[width=\linewidth]{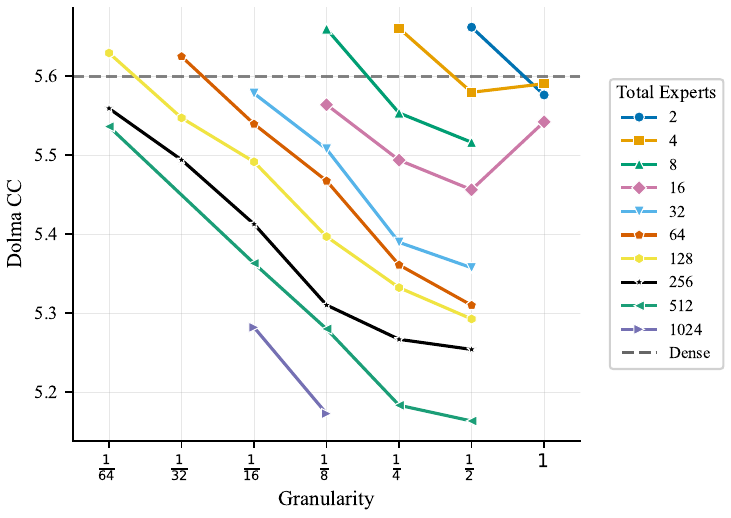}
        \end{subfigure}
        \begin{subfigure}[t]{0.33\textwidth}
            \centering
            \caption*{\scriptsize Fixed granularity (g)}
            \includegraphics[width=\linewidth]{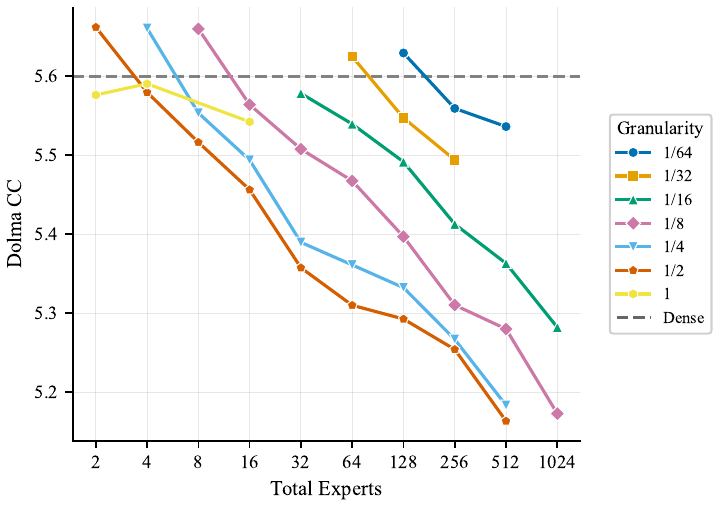}
        \end{subfigure}
        \begin{subfigure}[t]{0.33\textwidth}
            \centering
            \caption*{\scriptsize Fixed activation sparsity (s)}
            \includegraphics[width=\linewidth]{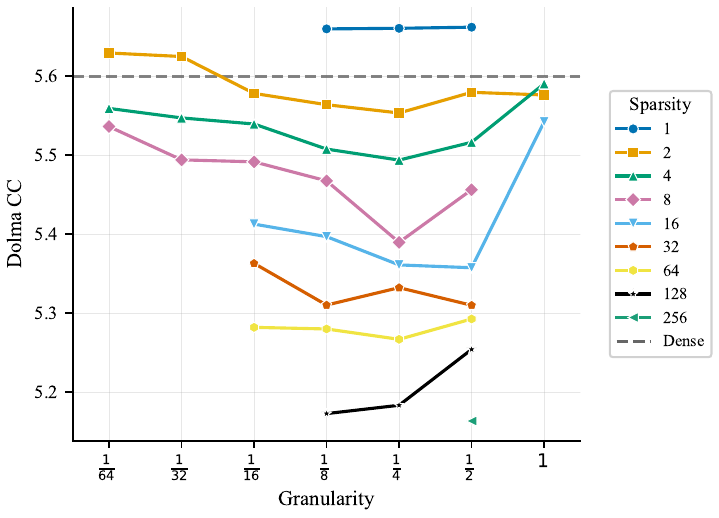}
        \end{subfigure}
        \caption{50M active, 50M - 930M total parameters}
    \end{subfigure}
\par\bigskip\bigskip
    \begin{subfigure}[t]{\textwidth}
        \begin{subfigure}[t]{0.33\textwidth}
            \centering
            \includegraphics[width=\linewidth]{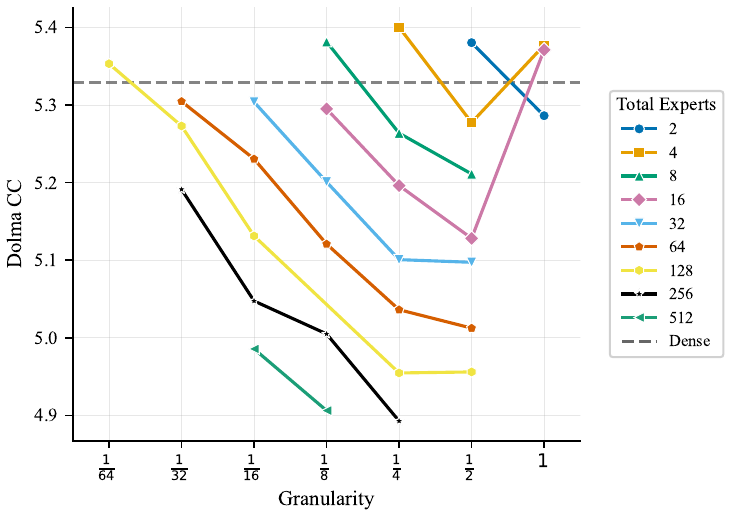}
        \end{subfigure}
        \begin{subfigure}[t]{0.33\textwidth}
            \centering
            \includegraphics[width=\linewidth]{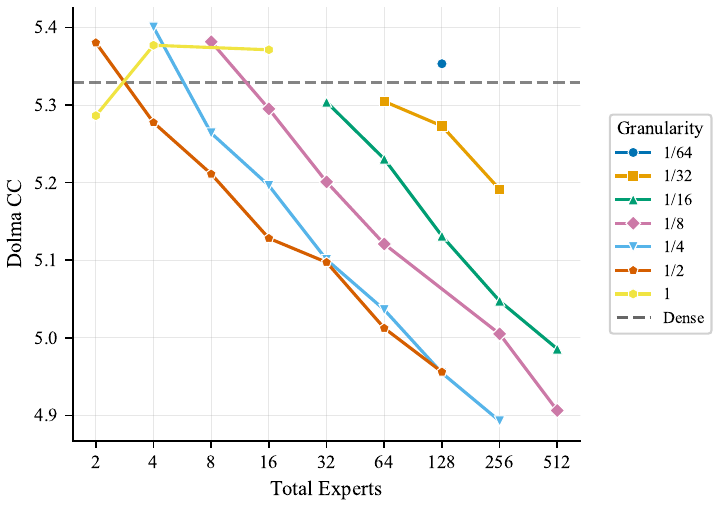}
        \end{subfigure}
        \begin{subfigure}[t]{0.33\textwidth}
            \centering
            \includegraphics[width=\linewidth]{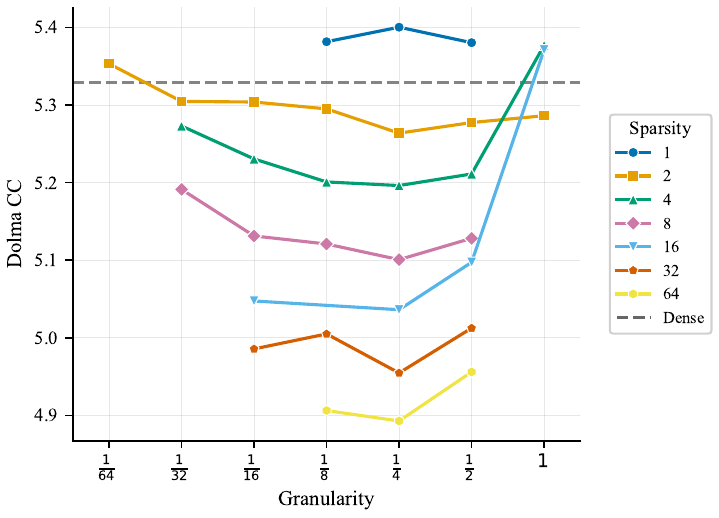}
        \end{subfigure}
        \caption{80M active, 80M - 765M total parameters}
    \end{subfigure}
    \par\bigskip\bigskip
        \begin{subfigure}[t]{\textwidth}
        \begin{subfigure}[t]{0.33\textwidth}
            \centering
            \includegraphics[width=\linewidth]{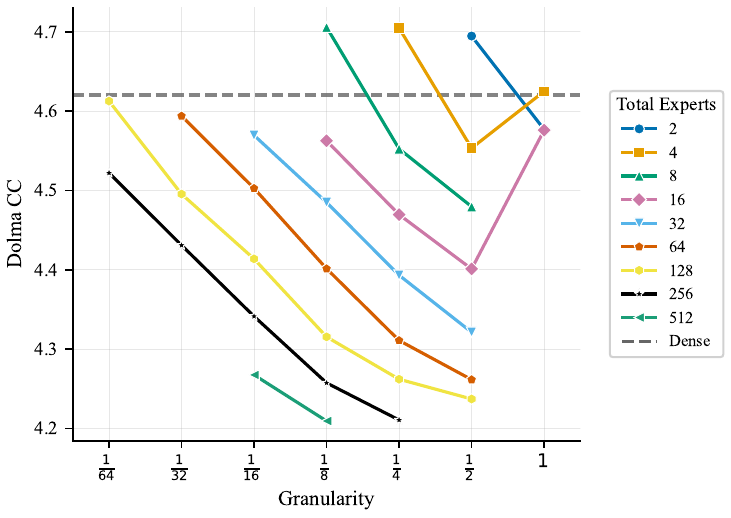}
        \end{subfigure}
        \begin{subfigure}[t]{0.33\textwidth}
            \centering
            \includegraphics[width=\linewidth]{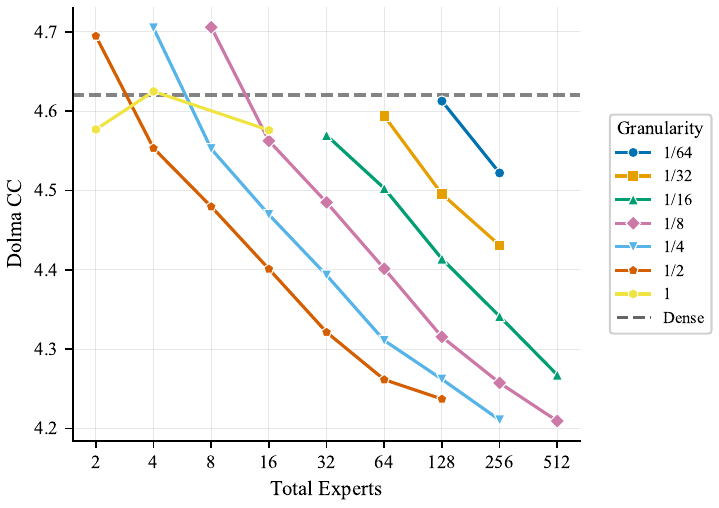}
        \end{subfigure}
        \begin{subfigure}[t]{0.33\textwidth}
            \centering
            \includegraphics[width=\linewidth]{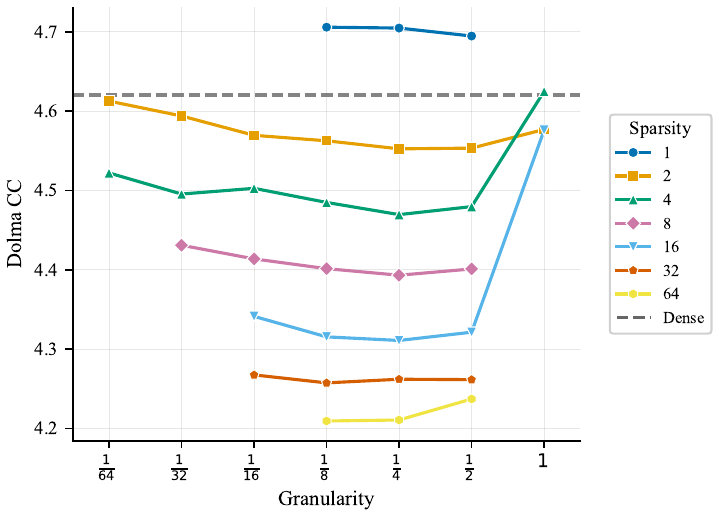}
        \end{subfigure}
        \caption{110M active, 110M - 1.4B total parameters}
    \end{subfigure}
    \end{figure*}

\clearpage  

\begin{figure*}[!ht]
        \addtocounter{figure}{-1}
    \begin{subfigure}[t]{\textwidth}
        \addtocounter{subfigure}{3}
        \begin{subfigure}[t]{0.33\textwidth}
            \centering
            \caption*{\scriptsize Fixed total experts (n)}
            \includegraphics[width=\linewidth]{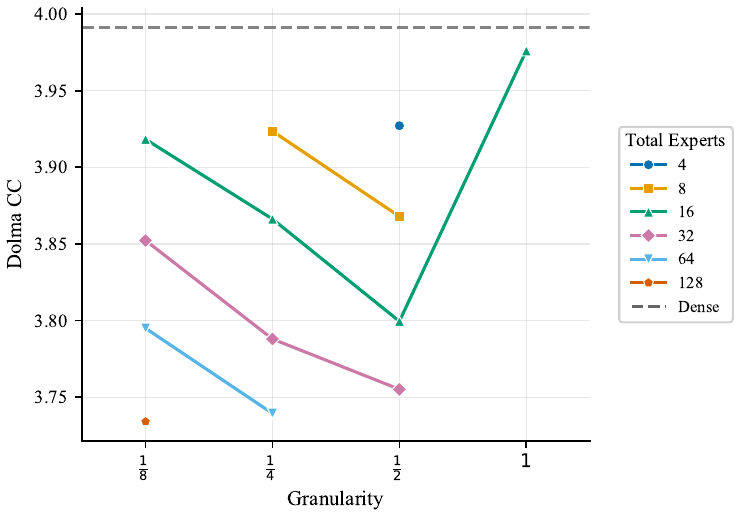}
        \end{subfigure}
        \begin{subfigure}[t]{0.33\textwidth}
            \centering
            \caption*{\scriptsize Fixed granularity (g)}
            \includegraphics[width=\linewidth]{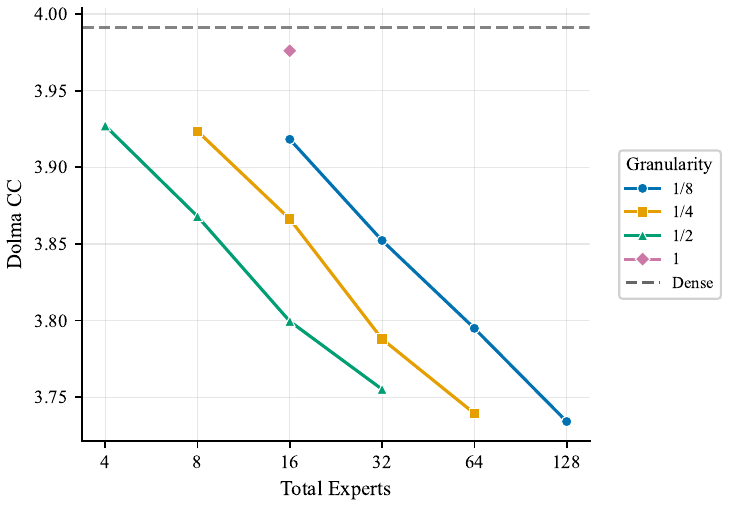}
        \end{subfigure}
        \begin{subfigure}[t]{0.33\textwidth}
            \centering
            \caption*{\scriptsize Fixed activation sparsity (s)}
            \includegraphics[width=\linewidth]{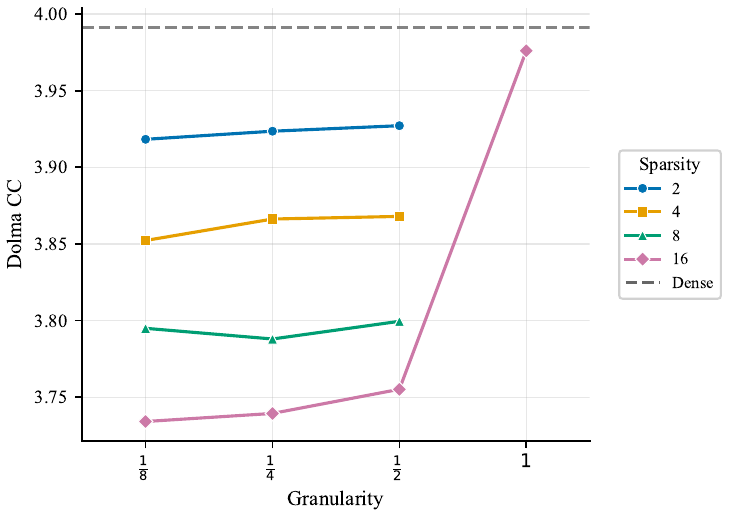}
        \end{subfigure}
        \caption{200M active, 200M - 3.3B total parameters}
    \end{subfigure}
    \par\bigskip\bigskip
        \begin{subfigure}[t]{\textwidth}
        \begin{subfigure}[t]{0.33\textwidth}
            \centering
            \includegraphics[width=\linewidth]{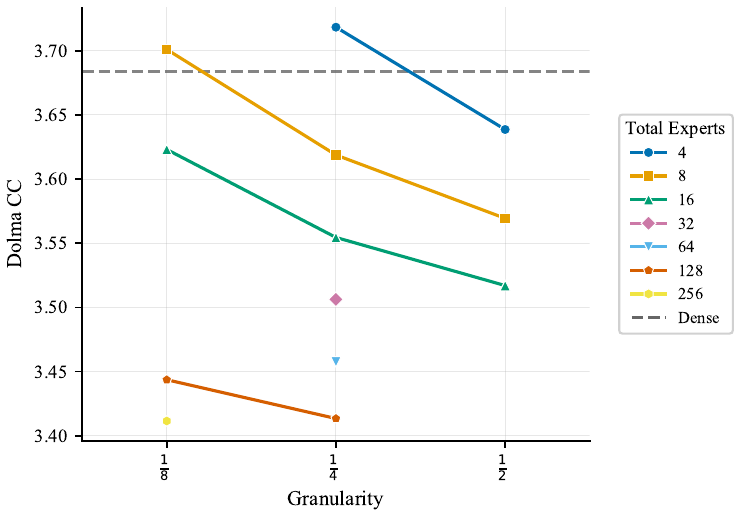}
        \end{subfigure}
        \begin{subfigure}[t]{0.33\textwidth}
            \centering
            \includegraphics[width=\linewidth]{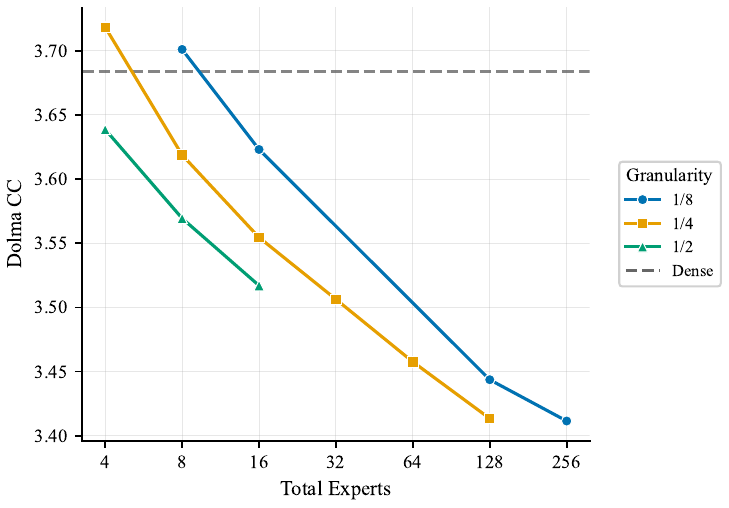}
        \end{subfigure}
        \begin{subfigure}[t]{0.33\textwidth}
            \centering
            \includegraphics[width=\linewidth]{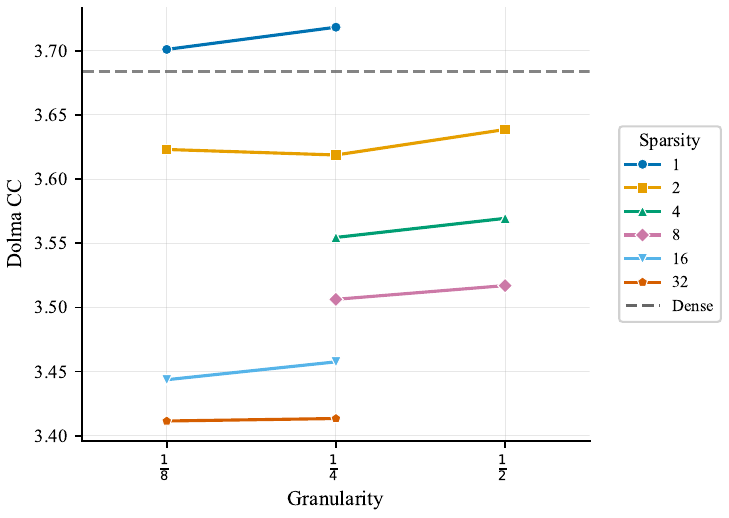}
        \end{subfigure}
        \caption{300M active, 300M - 6.6B total parameters}
    \end{subfigure}

    \caption{
    \textbf{Increasing inactive expert parameters via expert size (left) or total count (center) improves performance in MoEs (\S\ref{sec:expt_main}).} This effect is seen both when holding total number of experts fixed (left) and when holding expert granularity fixed (center). In general, increasing total parameters results in improved performance.  \textbf{Optimal tradeoff between expert count and granularity varies in MoEs (right). (\S\ref{sec:expt_main})}
    At each activation sparsity $s$ (equivalently, at each total parameter count), the optimal (total expert count, expert granularity) configuration varies. As $s$ increases, optimal expert granularity remains nearly fixed, suggesting that sparsity should be scaled up primarily by increasing total expert count $n$, while maintaining a near constant, slowly increasing expert granularity $g$. 
    }
    \label{fig:dolma_common_crawl_experts}
\end{figure*}

\begin{figure*}[!ht]
    \centering
    
    \begin{subfigure}[t]{0.46\textwidth}
        \centering
        \includegraphics[width=\linewidth]{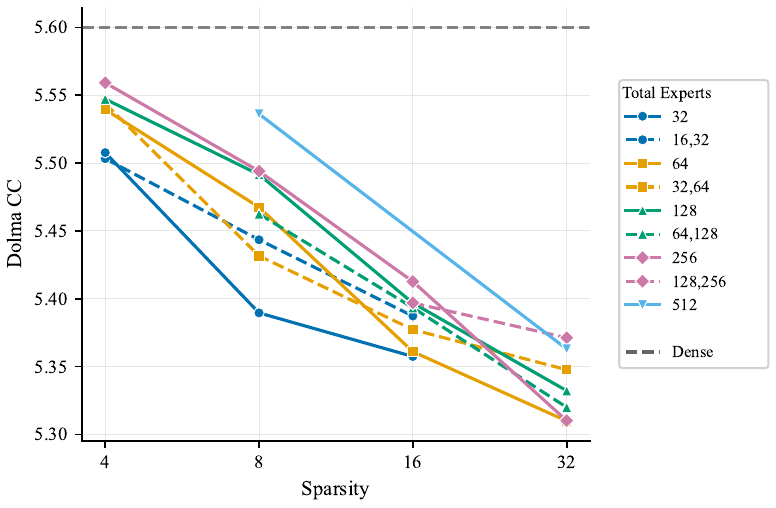}
        \caption{50M active, 50M - 930M total parameters}
    \end{subfigure}
    \vspace{1em}
    \begin{subfigure}[t]{0.46\textwidth}
        \centering
        \includegraphics[width=\linewidth]{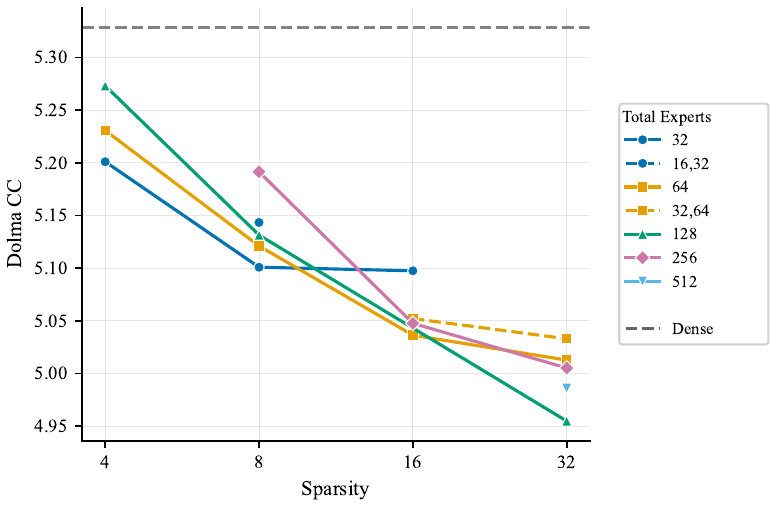}
        \caption{80M active, 80M - 765M total parameters}
    \end{subfigure}
    \caption{
    \textbf{Heterogeneity of expert size alone does not improve MoE performance (\S\ref{sec:expt_hetgen}).} To explore the potential benefits of their architectural flexibility, we compare heterogeneous MoEs (indicated by dotted lines) to active- and total-parameter-matched homogeneous MoEs. Heterogeneity alone does not result in performance gains, as, at each activation sparsity $s$, heterogeneous MoEs with $n_1, n_2 = a, b$ lie between or near the 2 closest homogeneous MoEs, with $n=a$ and with $n=b$.
    }
    \label{fig:dolma_common_crawl_het}
\end{figure*}

\begin{figure*}[!ht]
    \centering
    
    \begin{subfigure}[t]{1.0\textwidth}
        \centering
        \includegraphics[width=\linewidth]{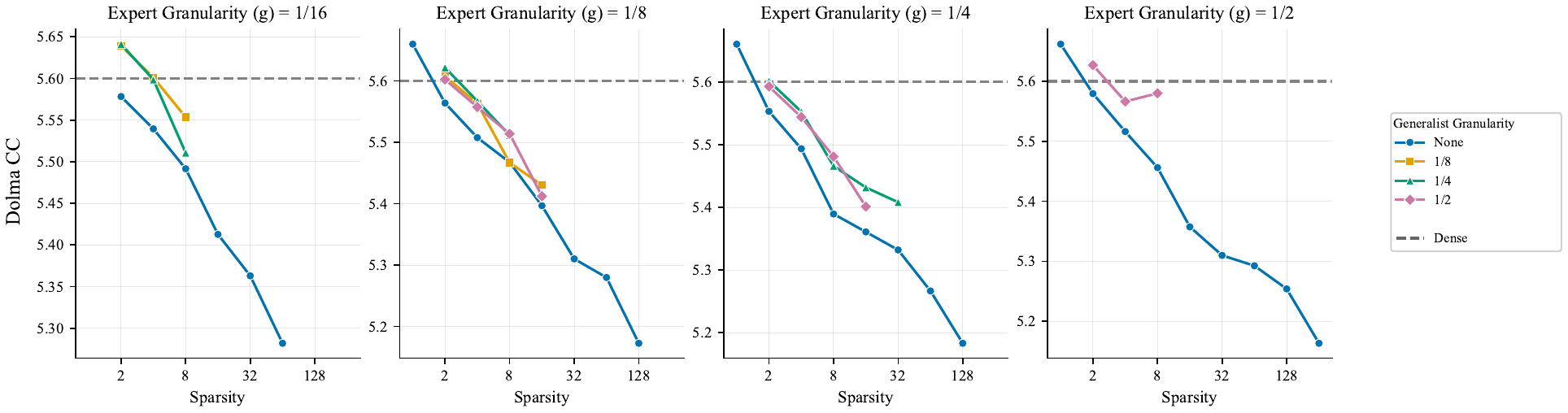}
        \caption{50M active, 50M - 930M total parameters}
    \end{subfigure}
    \par\bigskip\bigskip
    \begin{subfigure}[t]{1.0\textwidth}
        \centering
        \includegraphics[width=\linewidth]{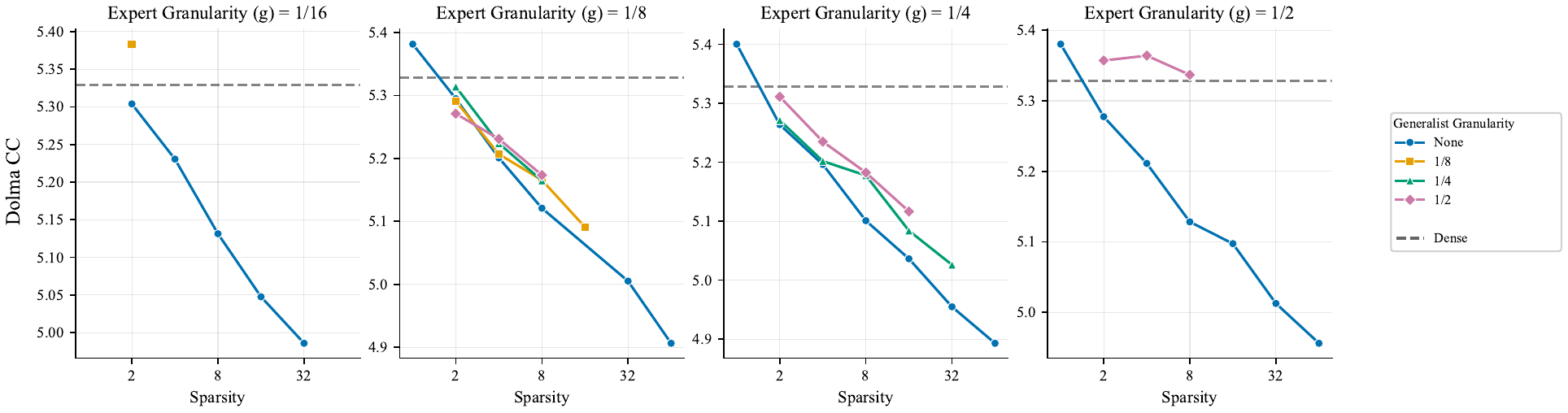}
        \caption{80M active, 80M - 765M total parameters}
    \end{subfigure}
    \par\bigskip\bigskip
    \begin{subfigure}[t]{1.0\textwidth}
        \centering
        \includegraphics[width=\linewidth]{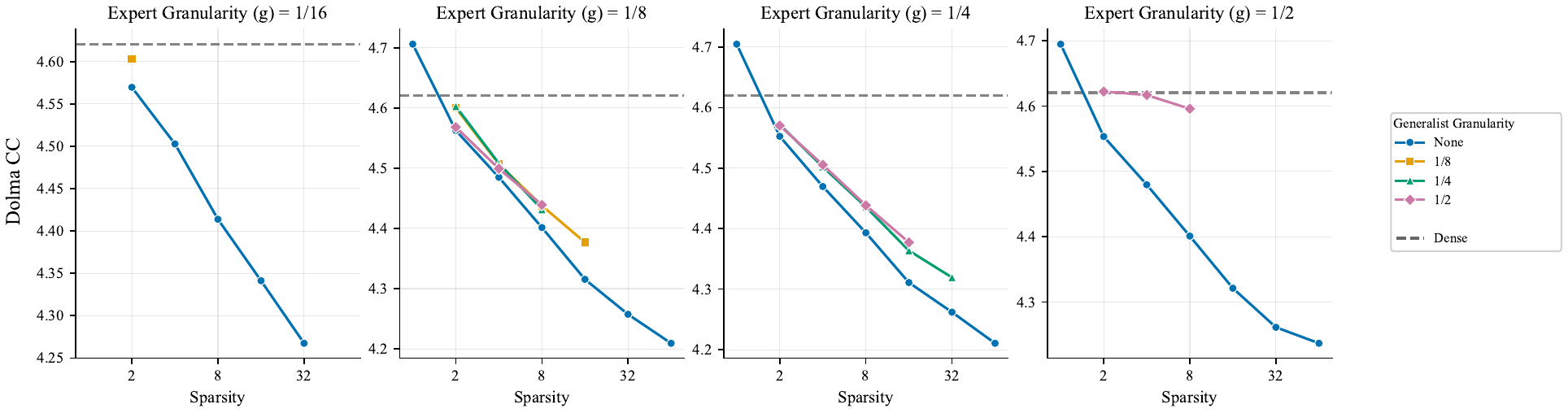}
        \caption{110M active, 110M - 1.4B total parameters}
    \end{subfigure}
    \caption{
    \textbf{The inclusion of a generalist consistently degrades performance in homogeneous MoEs (\S\ref{sec:expt_hetgen}).}
    We train MoE LMs which consist of some routed experts with granularity $g$, as well as a generalist with granularity $g_{gen}\in \{\frac{1}{2}, \frac{1}{4}, \frac{1}{8}\} $. We compare to settings with no generalist, only routed experts with granularity $g$. In all settings and configurations, the addition of any granularity generalist results in comparable or degraded performance. 
    }
    \label{fig:dolma_common_crawl_gen}
\end{figure*}

\begin{figure*}[ht]
    \centering
    \begin{subfigure}[t]{1.0\textwidth}
        \centering
        \includegraphics[width=\linewidth]{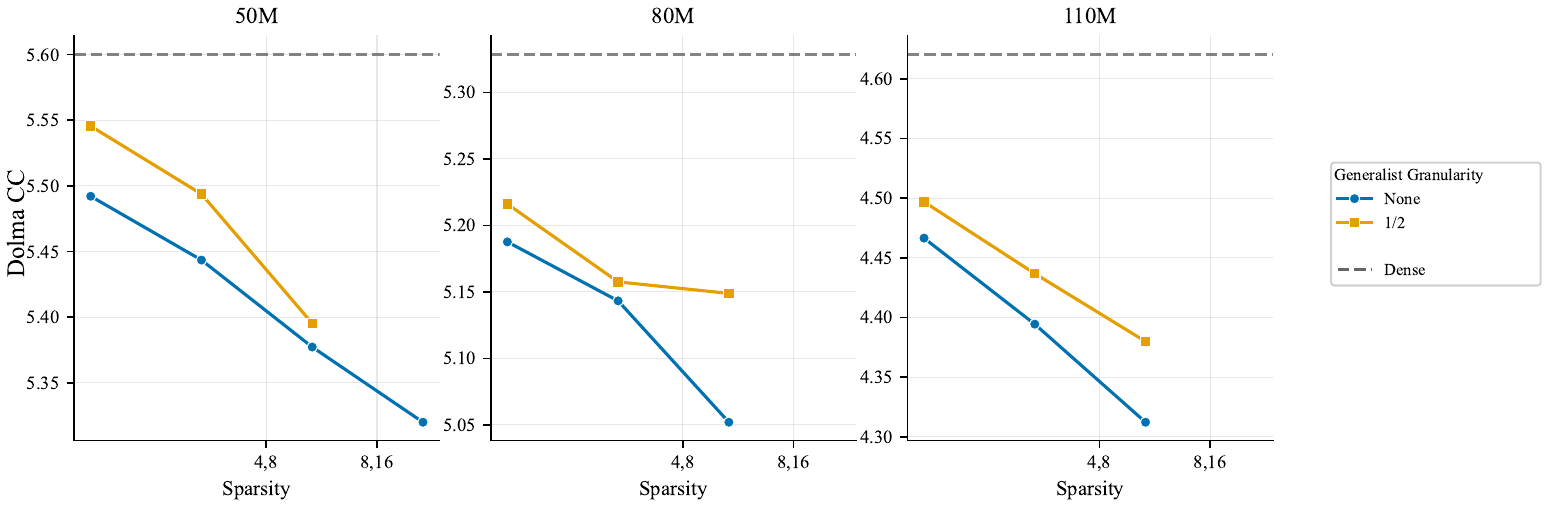}
    \end{subfigure}
    \caption{
    \textbf{The inclusion of a generalist consistently degrades performance in heterogeneous MoEs (\S\ref{sec:expt_hetgen}).}
    We train heterogeneous MoE LMs which consist of  routed experts with granularity $g_1, g_2$, as well as a generalist with granularity $g_{gen} = \frac{1}{2}$. We compare to settings with no generalist. In all settings and configurations, the addition of a generalist results in comparable or degraded performance. 
    }
    \label{fig:dolma_common_crawl_hetgen}
\end{figure*}

\begin{figure*}[ht]
    \centering
    \begin{subfigure}[t]{\textwidth}
        \centering
        \begin{subfigure}[t]{0.45\textwidth}
            \includegraphics[width=\linewidth]{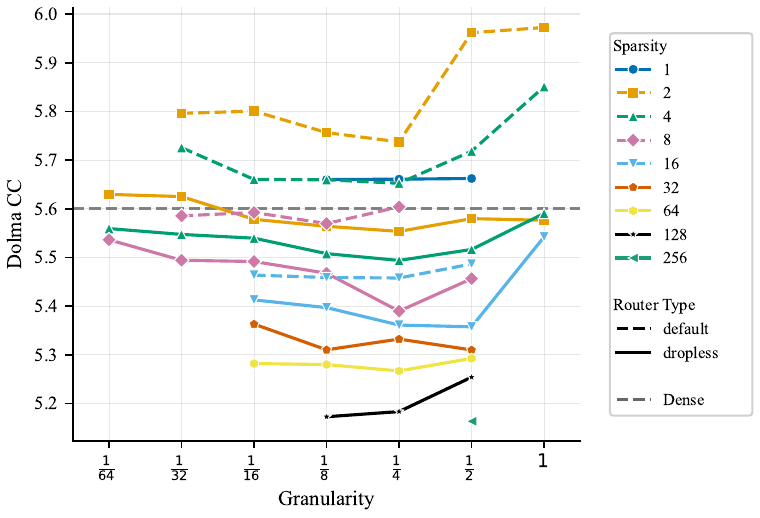}
            \caption{50M active, 50M - 930M total parameters}
        \end{subfigure}
    \hspace{1em}
        \begin{subfigure}[t]{0.45\textwidth}
            \centering
            \includegraphics[width=\linewidth]{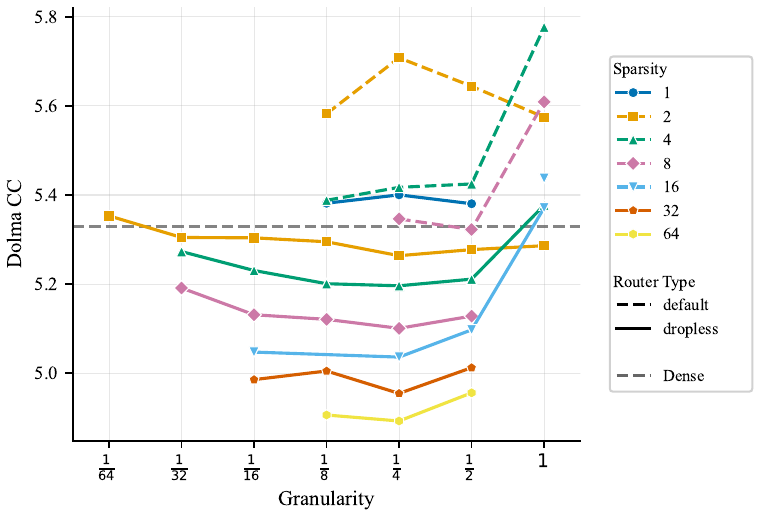}
            \caption{80M active, 80M - 765M total parameters}
        \end{subfigure}
    \end{subfigure}

    \par\bigskip\bigskip
    \begin{subfigure}[t]{0.45\textwidth}
        \centering
        \includegraphics[width=\linewidth]{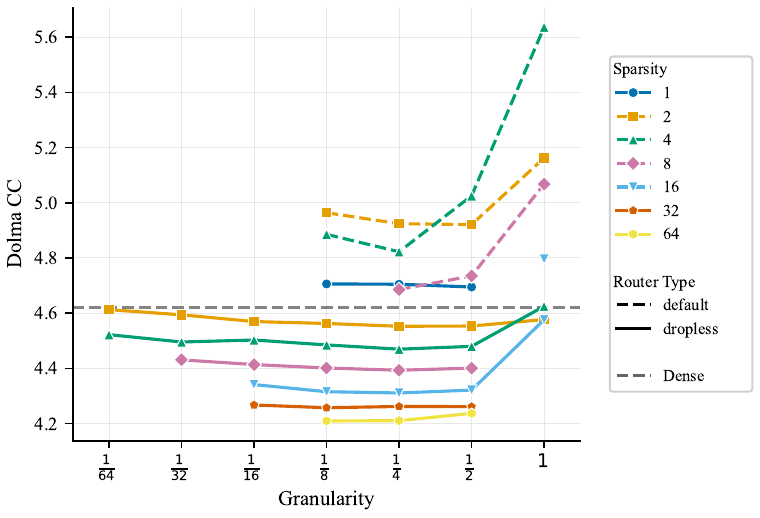}
        \caption{110M active, 110M - 1.4B total parameters}
    \end{subfigure}
    \caption{ 
    \textbf{Dropless routing outperforms default routing (\S\ref{sec:expt_router}).}
    We compare dropless routing to the default setting, which allow tokens to be dropped. Across all scales, we find that dropless routing outperforms or performs comparably to default routing. 
    }
    \label{fig:dolma_common_crawl_dropless}
\end{figure*}

\begin{figure*}[ht]
    \centering
    \begin{subfigure}[t]{0.45\textwidth}
        \centering
        \includegraphics[width=\linewidth]{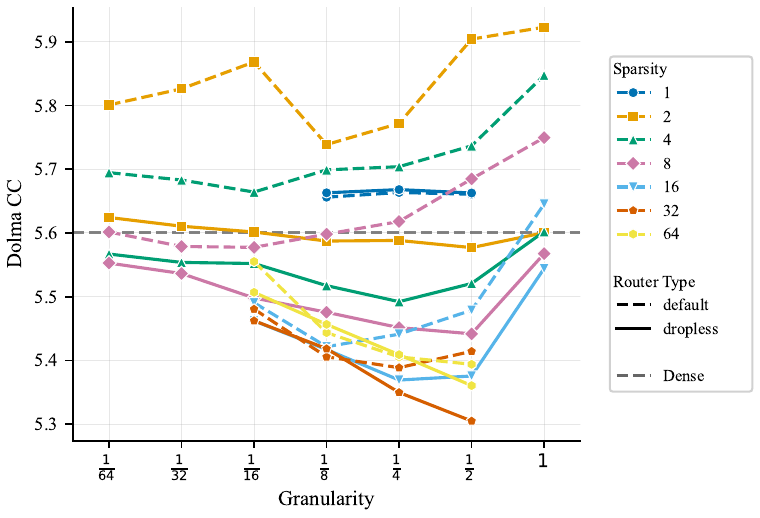}
        \caption{50M active, 50M - 930M total parameters}
    \end{subfigure}
    \hspace{1em}
    \begin{subfigure}[t]{0.45\textwidth}
        \centering
        \includegraphics[width=\linewidth]{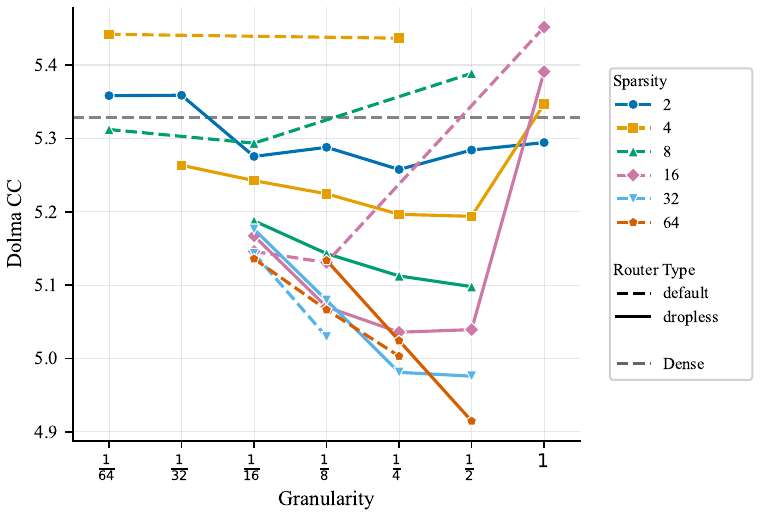}
        \caption{80M active, 80M - 765M total parameters}
    \end{subfigure}
    \caption{
    \textbf{Dropless routing, with bias $\gamma=\num{1e-3}$ (\S\ref{sec:expt_router}).} 
    As in Figure~\ref{fig:lm_avg_dropless}, we compare dropless routing to the default setting, which allow tokens to be dropped. Across all scales, we find that dropless routing outperforms or performs comparably to default routing. We see here with additional higher sparsity default routing runs that as sparsity increases, default routing performance approaches that of dropless routing.
    }
    \label{fig:dolma_common_crawl_dropless_with_lf}
\end{figure*}

\begin{figure*}[ht]
    \centering
    \begin{subfigure}[]{\textwidth}
        \centering
        \includegraphics[width=0.46\linewidth]{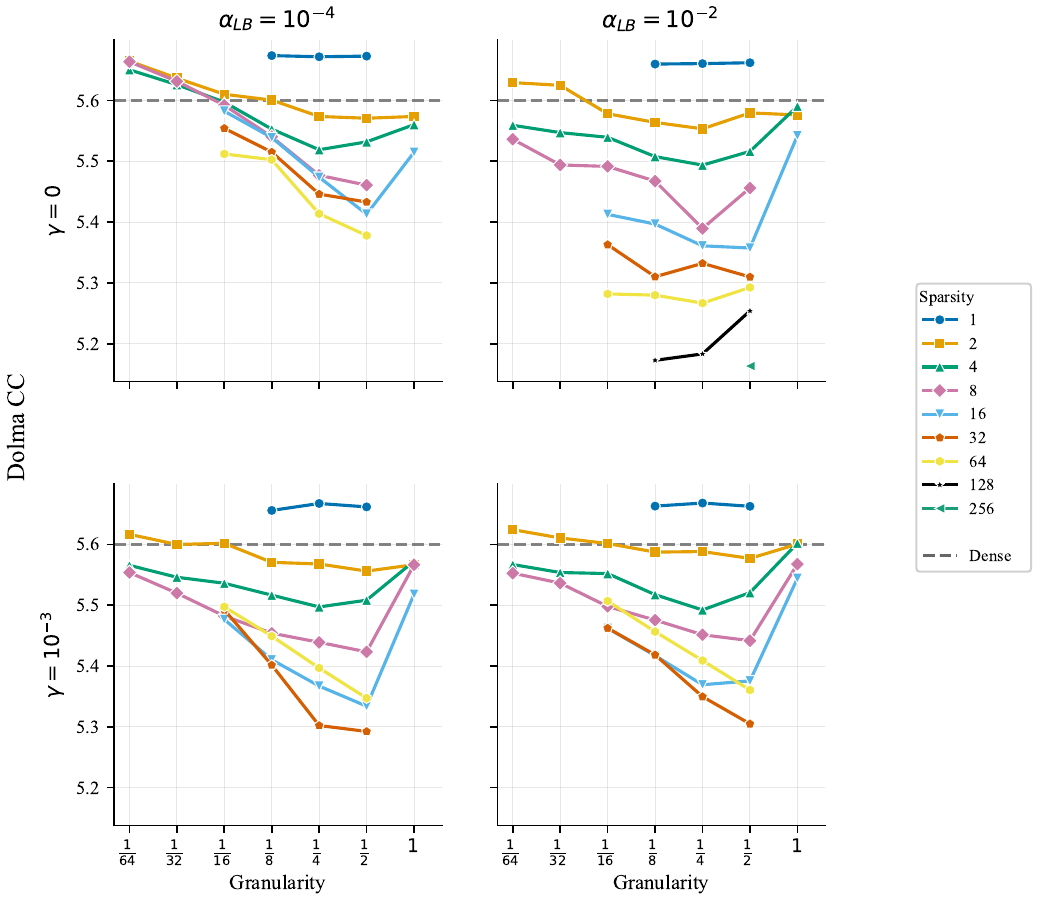}
        \hspace{1em}
        \includegraphics[width=0.46\linewidth]{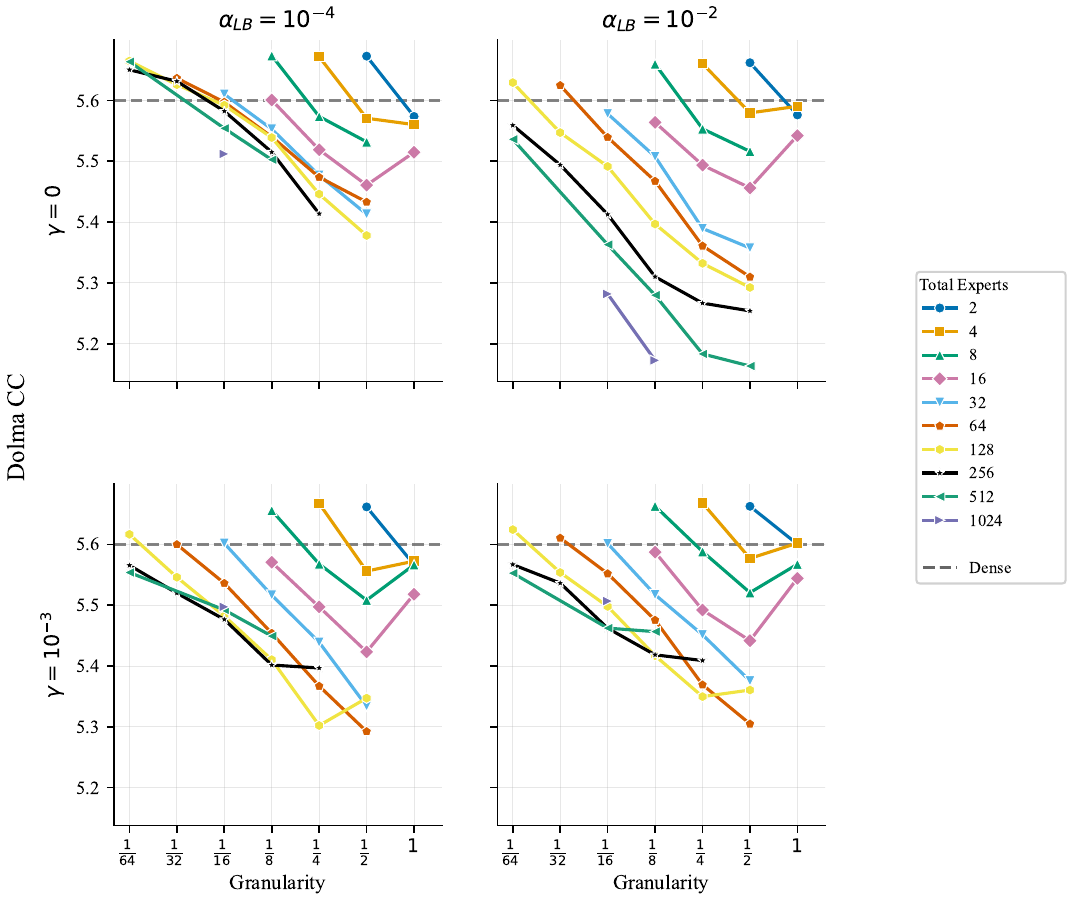}
        \caption{50M active, 50M - 930M total parameters}
    \end{subfigure}
    \par\bigskip\bigskip
    \begin{subfigure}[]{\textwidth}
        \centering
        \includegraphics[width=0.46\linewidth]{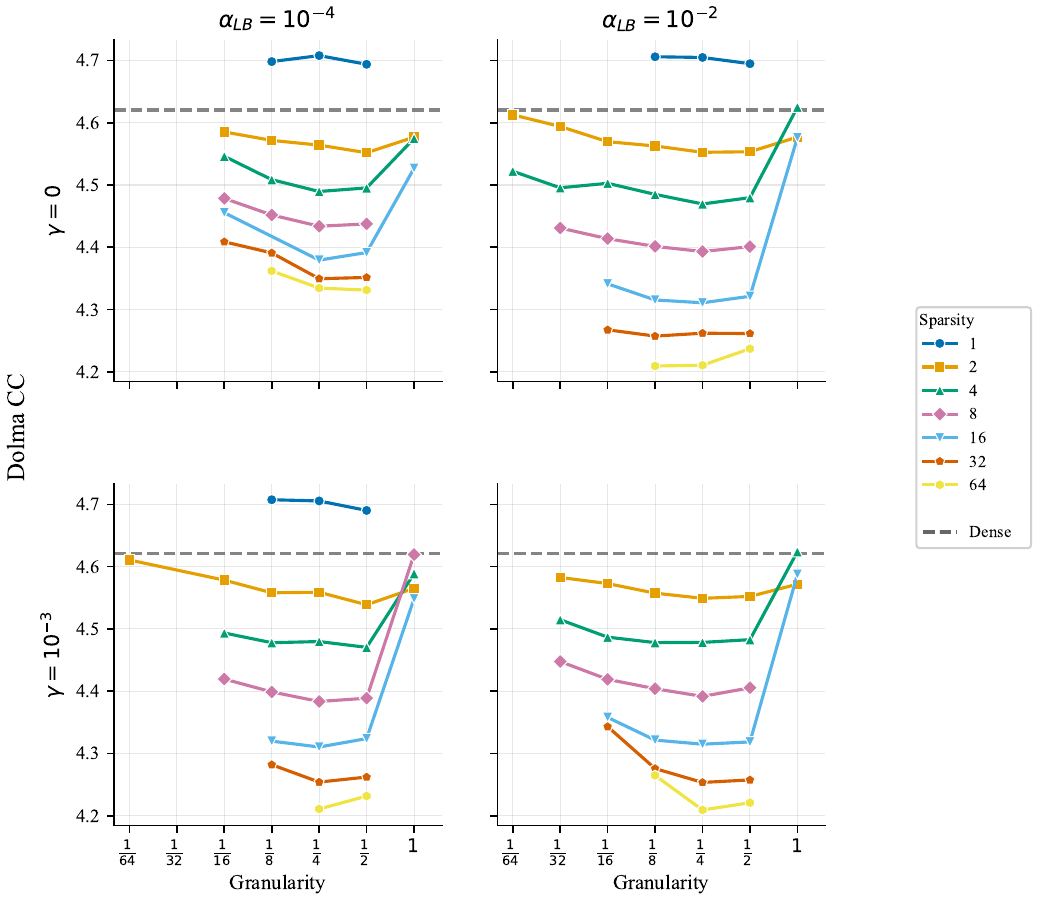}
        \hspace{1em}
        \includegraphics[width=0.46\linewidth]{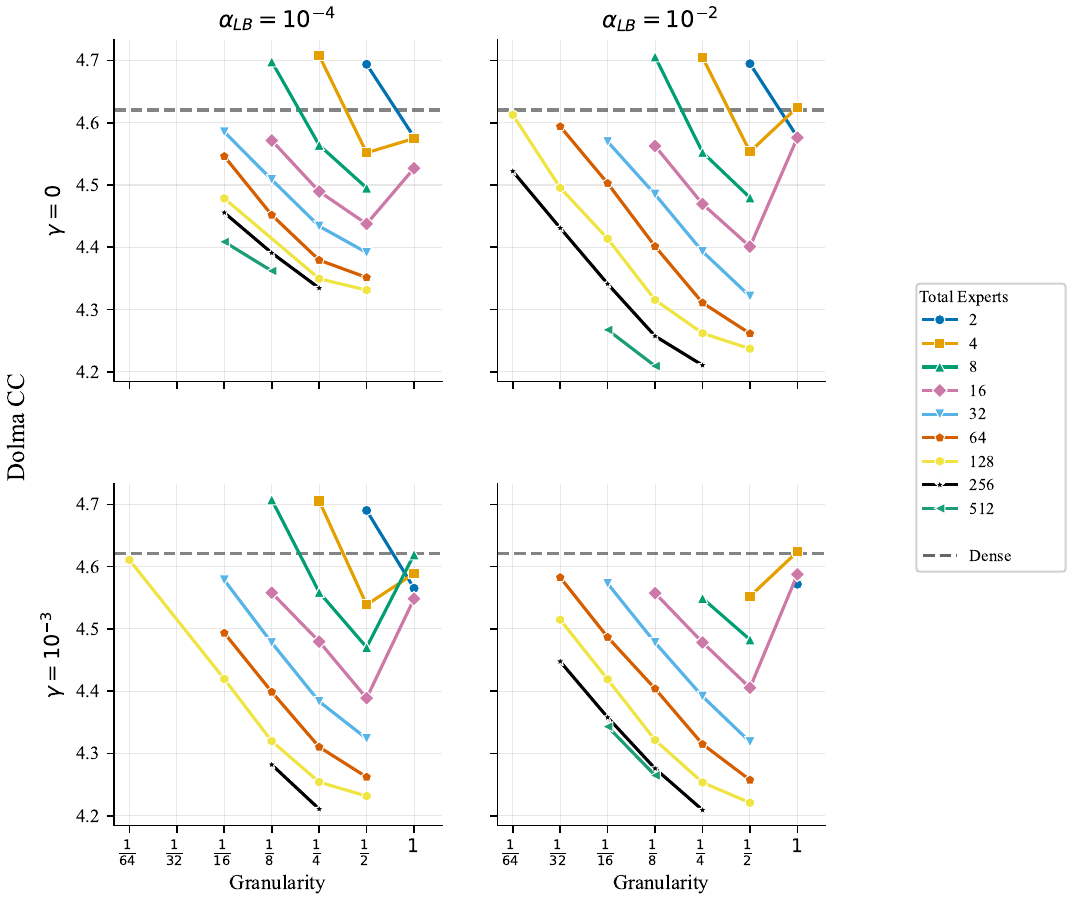}
        \caption{80M active, 80M - 765M total parameters}
    \end{subfigure}
    \par\bigskip\bigskip
    \begin{subfigure}[t]{\textwidth}
        \centering
        \includegraphics[width=0.46\linewidth]{figures/lm/dolma_common-crawl-validation/ce_loss/lb_sweep_hgn_gxs_110M.pdf}
        \hspace{1em}
        \includegraphics[width=0.46\linewidth]{figures/lm/dolma_common-crawl-validation/ce_loss/lb_sweep_hgn_gxn_110M.pdf}
        \caption{110M active, 110M - 1.4B total parameters}
    \end{subfigure}

    \end{figure*} 

\clearpage  

\begin{figure*}[ht]
    \addtocounter{figure}{-1}
    \centering
    \begin{subfigure}[t]{\textwidth}
        \centering
        \includegraphics[width=0.46\linewidth]{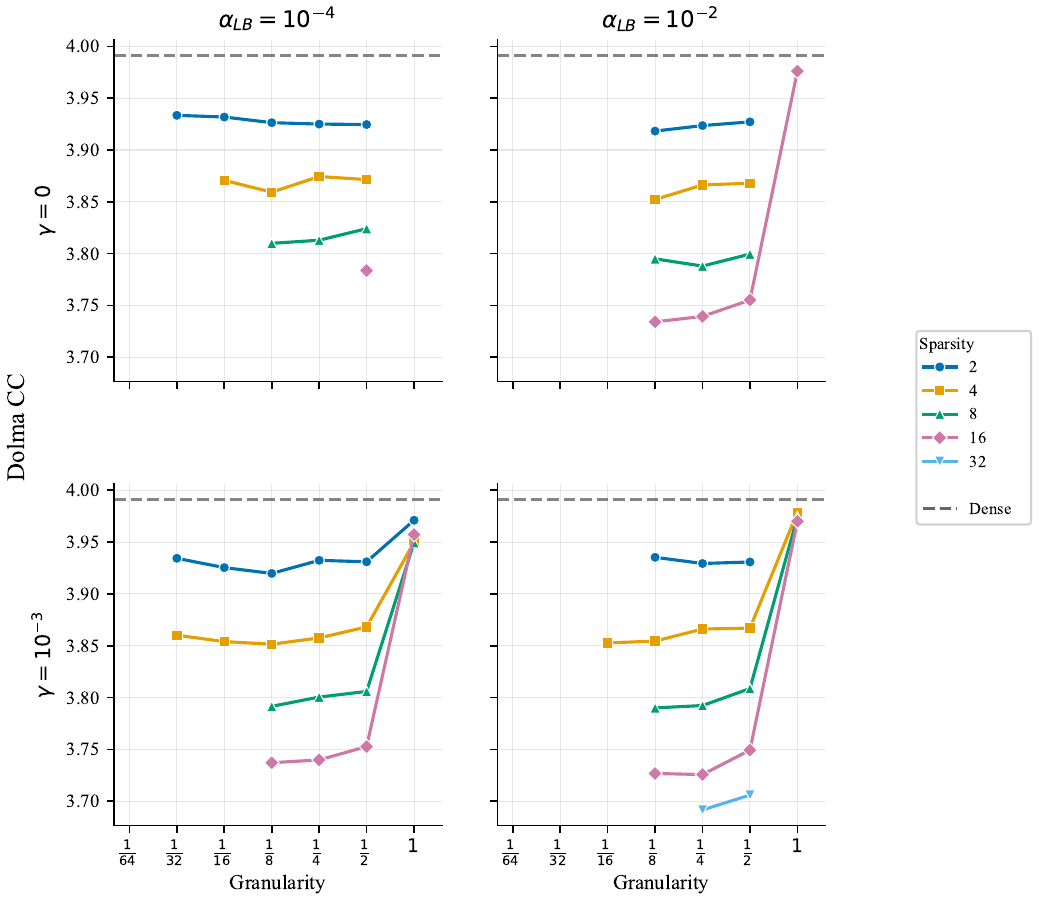}
        \hspace{1em}
        \includegraphics[width=0.46\linewidth]{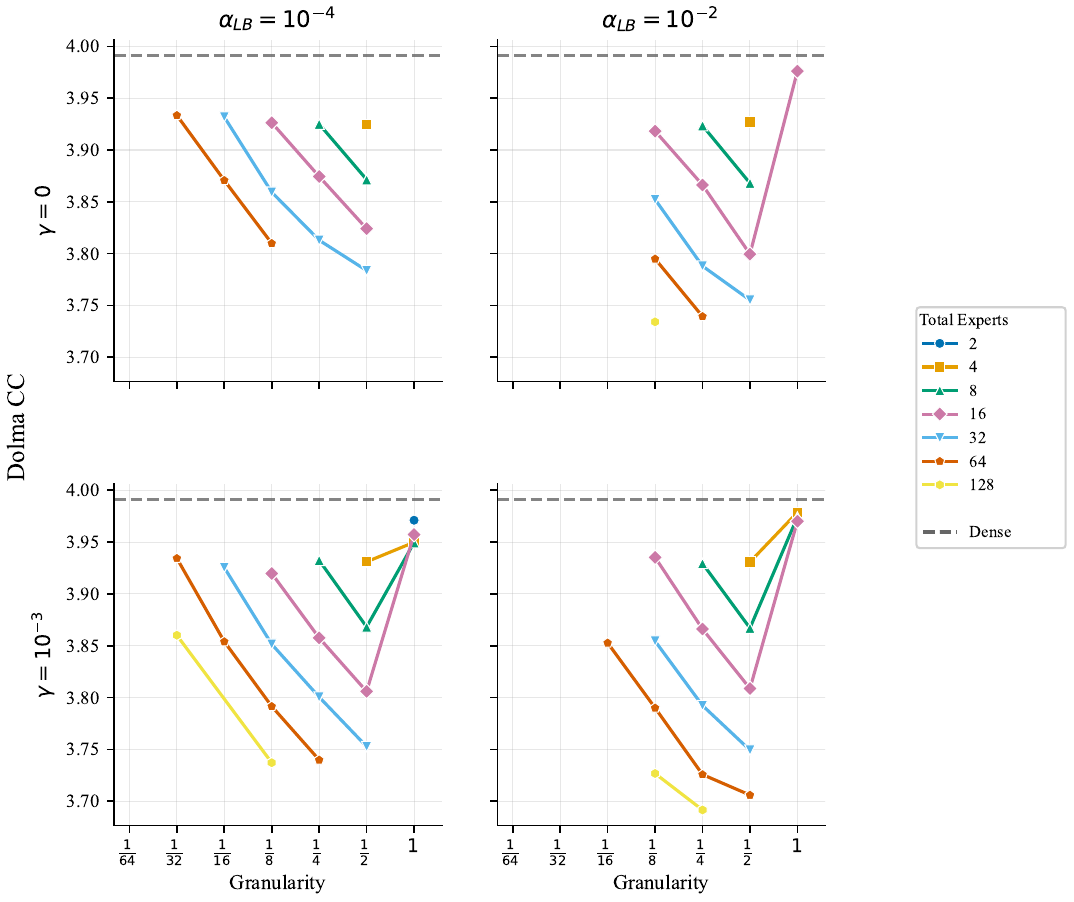}
        \caption{200M active, 200M - 3.3B total parameters}
    \end{subfigure}
    \par\bigskip\bigskip
    \begin{subfigure}[t]{\textwidth}
        \centering
        \includegraphics[width=0.3\linewidth]{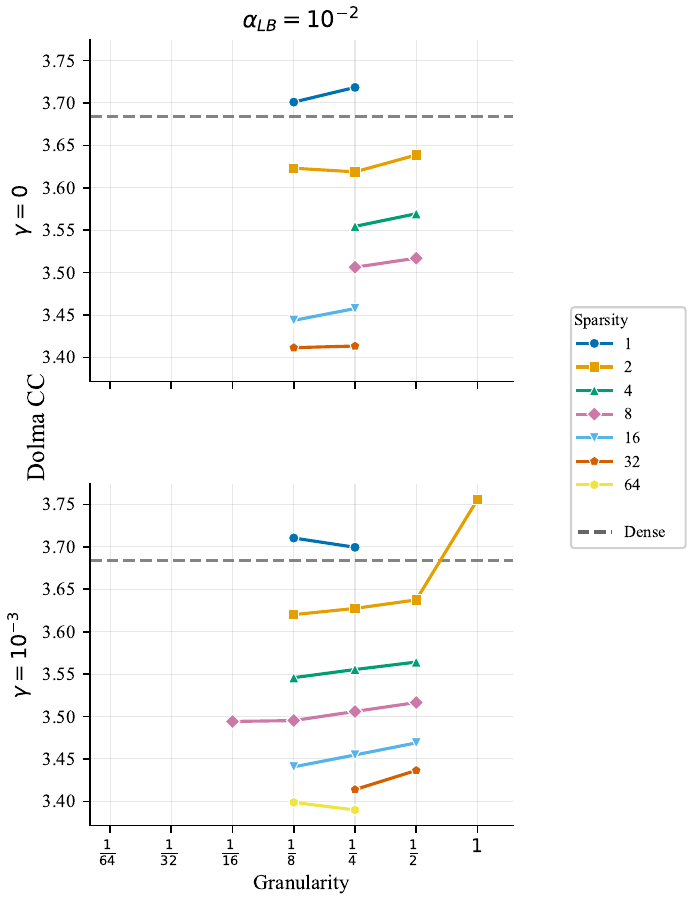}
        \hspace{1em}
        \includegraphics[width=0.3\linewidth]{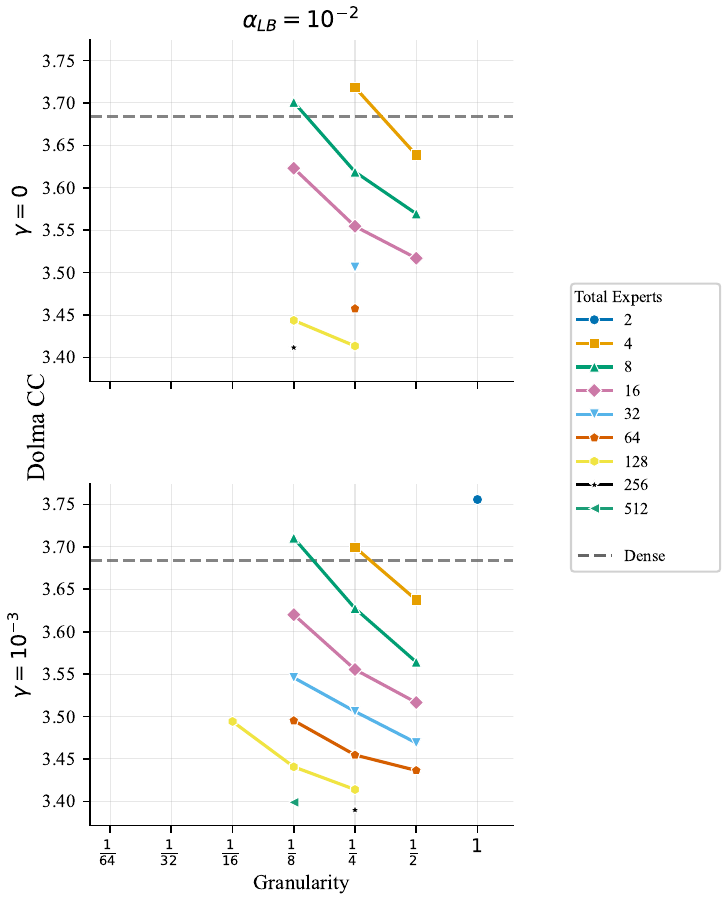}
        \caption{300M active, 300M - 6.6B total parameters}
    \end{subfigure}

    \caption{
    \textbf{Load balancing mechanisms must be tuned correctly (\S\ref{sec:expt_router}).}
    We consider load balancing loss weight $\alpha_{LB} \in \{\num{1e-2}, \num{1e-4}\}$ and loss-free load balancing with bias $\gamma\in\{0, \num{1e-3}\}$ ($\gamma=0$ indicates no loss-free mechanism). Results show that poorly chosen hyperparameters, such as high bias $\gamma = 1e-3$ with total experts $n\geq 512$, may impair performance. However, all settings other than $(\alpha_{LB}=\num{1e-2}, \gamma=\num{1e-3})$ perform comparably for $n \leq 512$, suggesting that a wide range of load balancing settings achieve near-optimal performance. 
    }
    \label{fig:dolma_common_crawl_lb}
\end{figure*}

%% file: fig_tex/lm/dolma_pes2o.tex
\begin{figure*}[!ht]
    \centering
        \begin{subfigure}[t]{\textwidth}
        \begin{subfigure}[t]{0.33\textwidth}
            \centering
            \caption*{\scriptsize Fixed total experts (n)}
            \includegraphics[width=\linewidth]{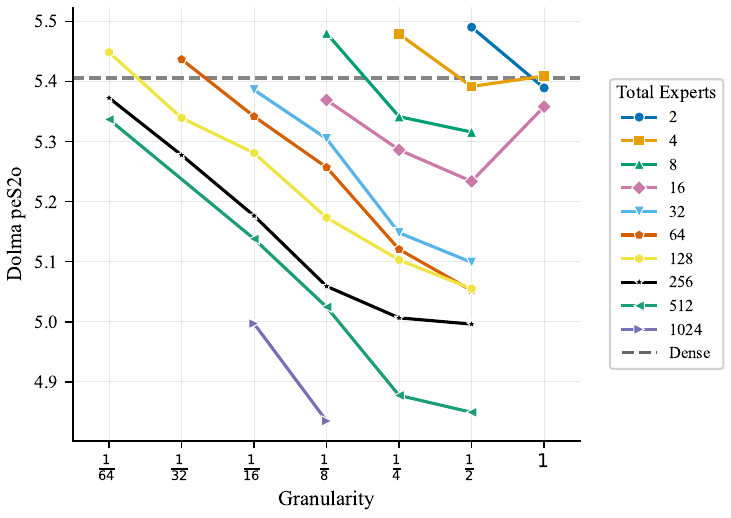}
        \end{subfigure}
        \begin{subfigure}[t]{0.33\textwidth}
            \centering
            \caption*{\scriptsize Fixed granularity (g)}
            \includegraphics[width=\linewidth]{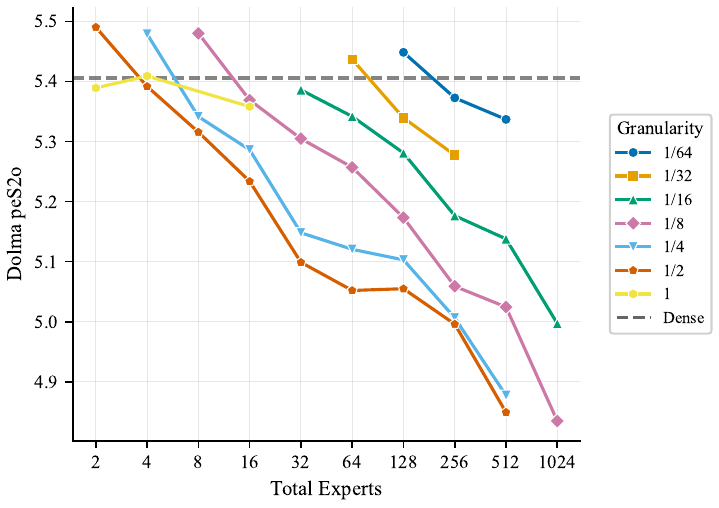}
        \end{subfigure}
        \begin{subfigure}[t]{0.33\textwidth}
            \centering
            \caption*{\scriptsize Fixed activation sparsity (s)}
            \includegraphics[width=\linewidth]{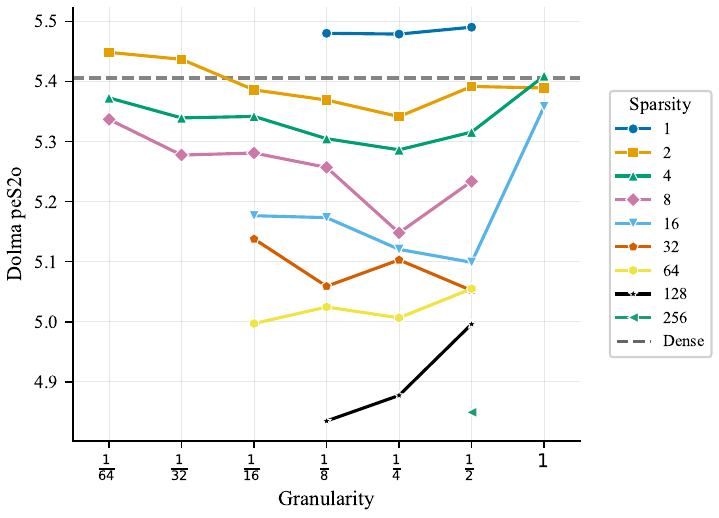}
        \end{subfigure}
        \caption{50M active, 50M - 930M total parameters}
    \end{subfigure}
\par\bigskip\bigskip
    \begin{subfigure}[t]{\textwidth}
        \begin{subfigure}[t]{0.33\textwidth}
            \centering
            \includegraphics[width=\linewidth]{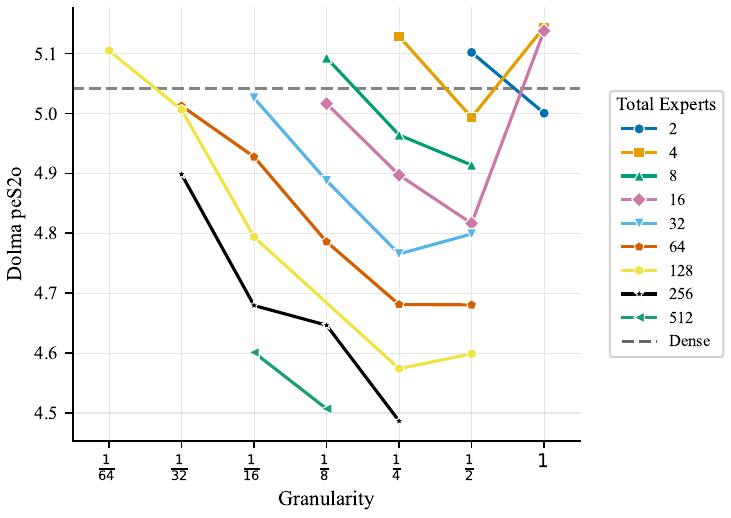}
        \end{subfigure}
        \begin{subfigure}[t]{0.33\textwidth}
            \centering
            \includegraphics[width=\linewidth]{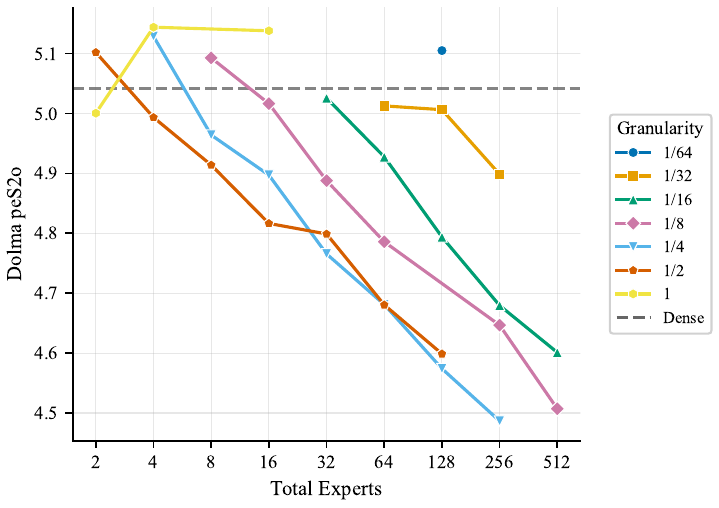}
        \end{subfigure}
        \begin{subfigure}[t]{0.33\textwidth}
            \centering
            \includegraphics[width=\linewidth]{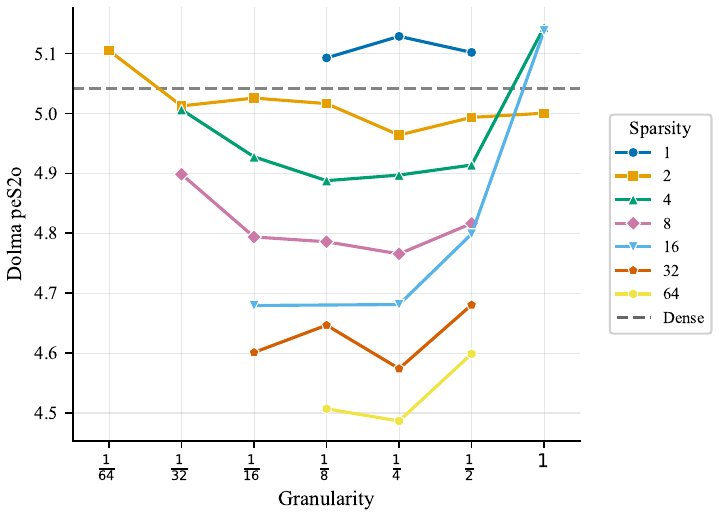}
        \end{subfigure}
        \caption{80M active, 80M - 765M total parameters}
    \end{subfigure}
    \par\bigskip\bigskip
        \begin{subfigure}[t]{\textwidth}
        \begin{subfigure}[t]{0.33\textwidth}
            \centering
            \includegraphics[width=\linewidth]{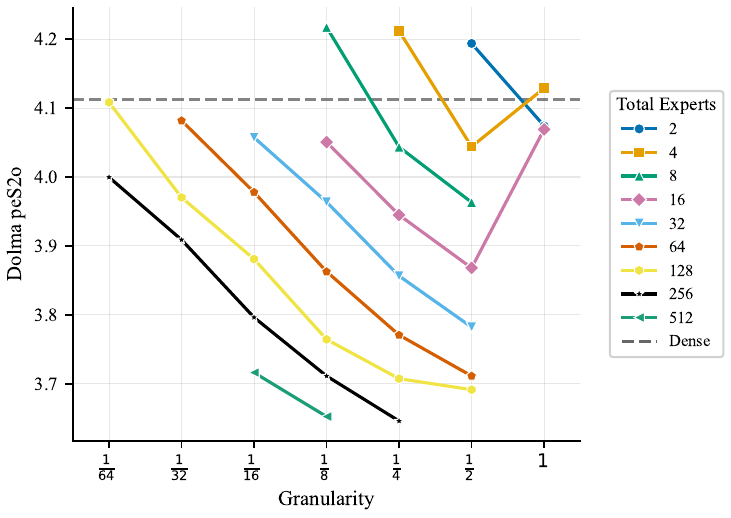}
        \end{subfigure}
        \begin{subfigure}[t]{0.33\textwidth}
            \centering
            \includegraphics[width=\linewidth]{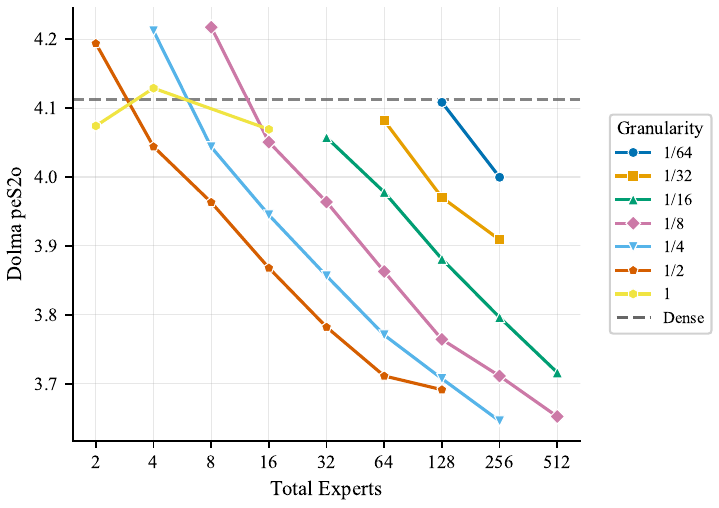}
        \end{subfigure}
        \begin{subfigure}[t]{0.33\textwidth}
            \centering
            \includegraphics[width=\linewidth]{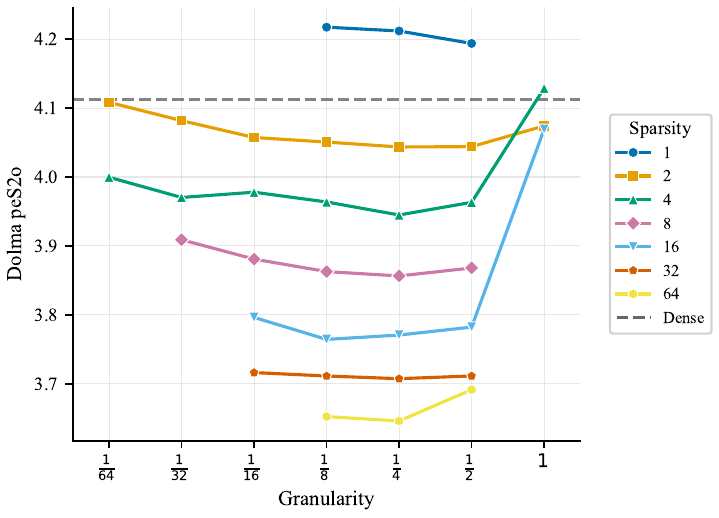}
        \end{subfigure}
        \caption{110M active, 110M - 1.4B total parameters}
    \end{subfigure}
    \end{figure*}

\clearpage  

\begin{figure*}[!ht]
        \addtocounter{figure}{-1}
    \begin{subfigure}[t]{\textwidth}
        \addtocounter{subfigure}{3}
        \begin{subfigure}[t]{0.33\textwidth}
            \centering
            \caption*{\scriptsize Fixed total experts (n)}
            \includegraphics[width=\linewidth]{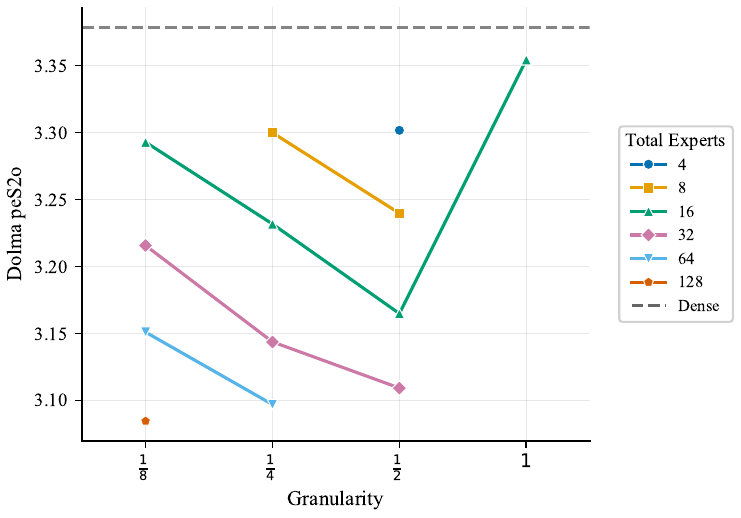}
        \end{subfigure}
        \begin{subfigure}[t]{0.33\textwidth}
            \centering
            \caption*{\scriptsize Fixed granularity (g)}
            \includegraphics[width=\linewidth]{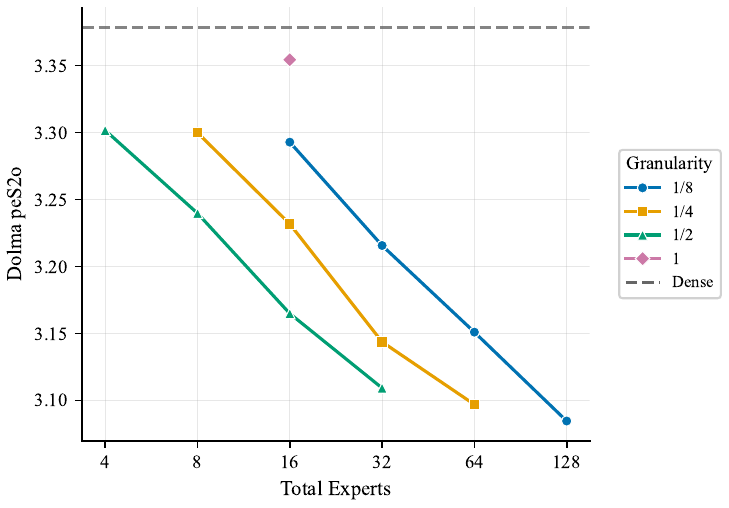}
        \end{subfigure}
        \begin{subfigure}[t]{0.33\textwidth}
            \centering
            \caption*{\scriptsize Fixed activation sparsity (s)}
            \includegraphics[width=\linewidth]{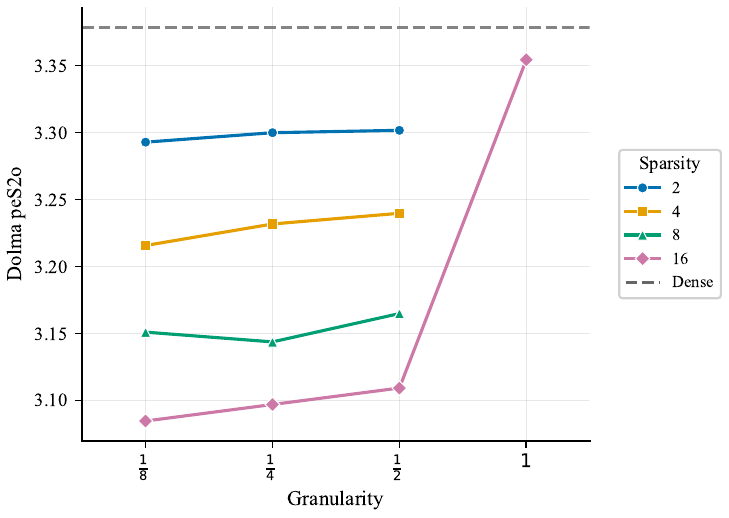}
        \end{subfigure}
        \caption{200M active, 200M - 3.3B total parameters}
    \end{subfigure}
    \par\bigskip\bigskip
        \begin{subfigure}[t]{\textwidth}
        \begin{subfigure}[t]{0.33\textwidth}
            \centering
            \includegraphics[width=\linewidth]{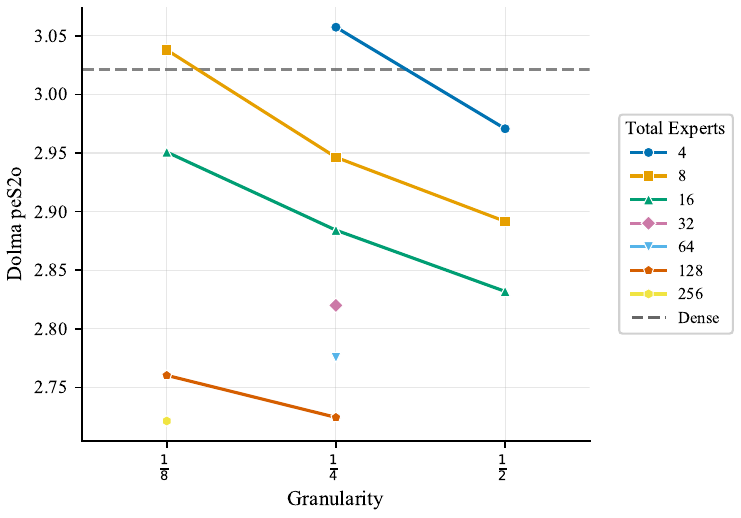}
        \end{subfigure}
        \begin{subfigure}[t]{0.33\textwidth}
            \centering
            \includegraphics[width=\linewidth]{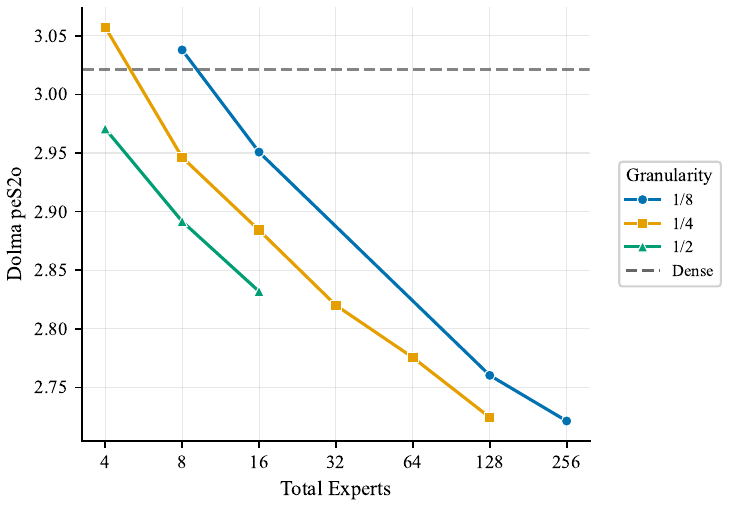}
        \end{subfigure}
        \begin{subfigure}[t]{0.33\textwidth}
            \centering
            \includegraphics[width=\linewidth]{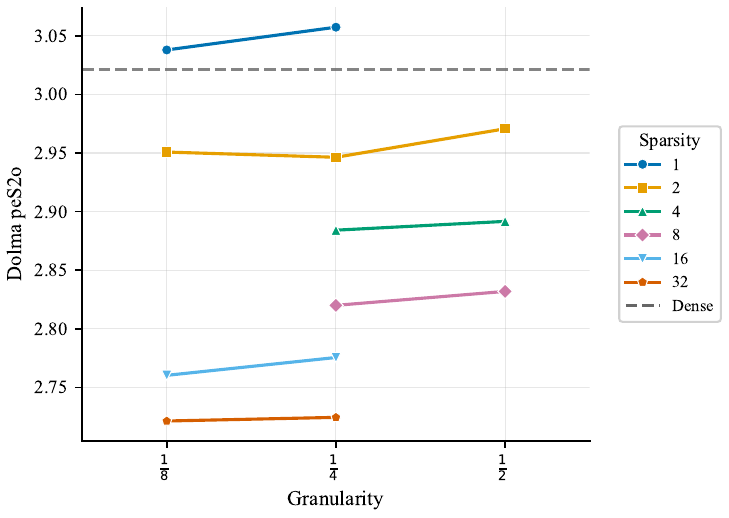}
        \end{subfigure}
        \caption{300M active, 300M - 6.6B total parameters}
    \end{subfigure}

    \caption{
    \textbf{Increasing inactive expert parameters via expert size (left) or total count (center) improves performance in MoEs (\S\ref{sec:expt_main}).} This effect is seen both when holding total number of experts fixed (left) and when holding expert granularity fixed (center). In general, increasing total parameters results in improved performance.  \textbf{Optimal tradeoff between expert count and granularity varies in MoEs (right). (\S\ref{sec:expt_main})}
    At each activation sparsity $s$ (equivalently, at each total parameter count), the optimal (total expert count, expert granularity) configuration varies. As $s$ increases, optimal expert granularity remains nearly fixed, suggesting that sparsity should be scaled up primarily by increasing total expert count $n$, while maintaining a near constant, slowly increasing expert granularity $g$. 
    }
    \label{fig:dolma_pes2o_experts}
\end{figure*}

\begin{figure*}[!ht]
    \centering
    
    \begin{subfigure}[t]{0.46\textwidth}
        \centering
        \includegraphics[width=\linewidth]{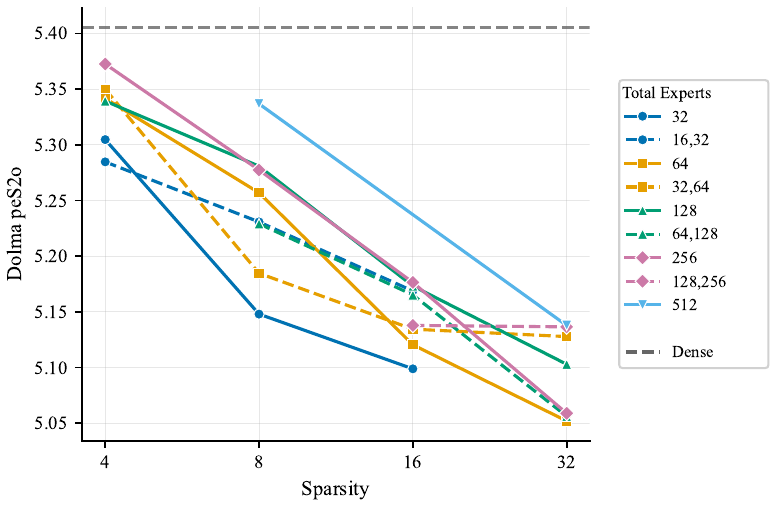}
        \caption{50M active, 50M - 930M total parameters}
    \end{subfigure}
    \vspace{1em}
    \begin{subfigure}[t]{0.46\textwidth}
        \centering
        \includegraphics[width=\linewidth]{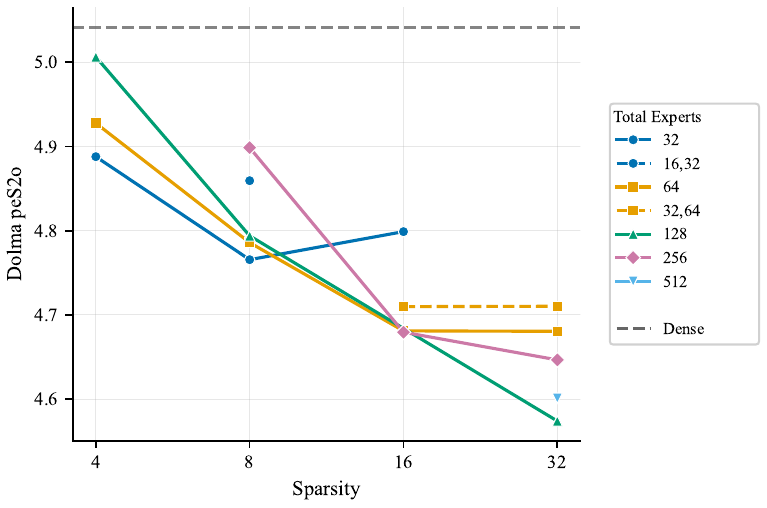}
        \caption{80M active, 80M - 765M total parameters}
    \end{subfigure}
    \caption{
    \textbf{Heterogeneity of expert size alone does not improve MoE performance (\S\ref{sec:expt_hetgen}).} To explore the potential benefits of their architectural flexibility, we compare heterogeneous MoEs (indicated by dotted lines) to active- and total-parameter-matched homogeneous MoEs. Heterogeneity alone does not result in performance gains, as, at each activation sparsity $s$, heterogeneous MoEs with $n_1, n_2 = a, b$ lie between or near the 2 closest homogeneous MoEs, with $n=a$ and with $n=b$.
    }
    \label{fig:dolma_pes2o_het}
\end{figure*}

\begin{figure*}[!ht]
    \centering
    
    \begin{subfigure}[t]{1.0\textwidth}
        \centering
        \includegraphics[width=\linewidth]{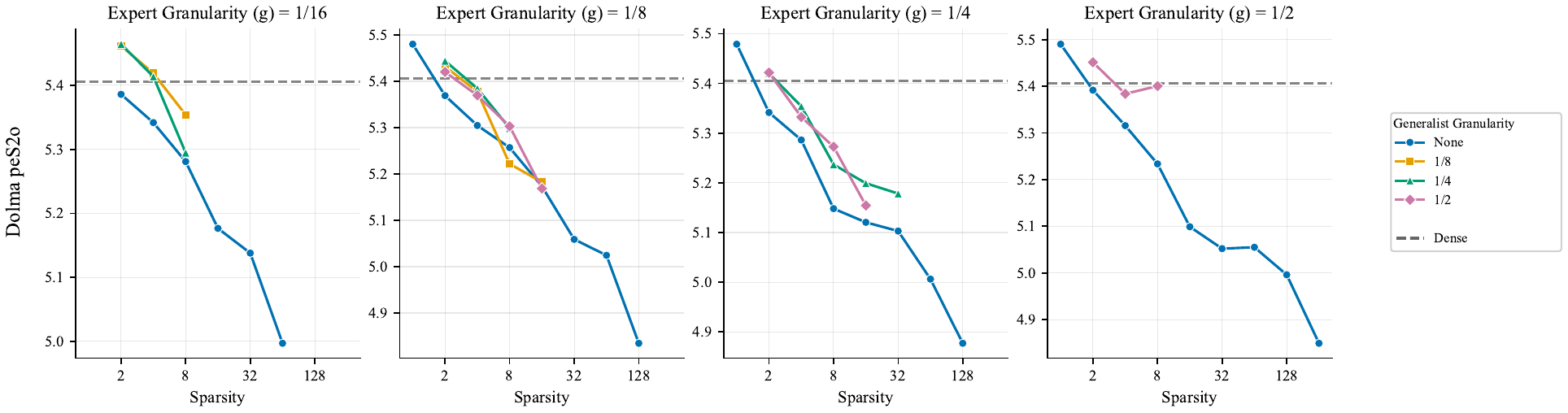}
        \caption{50M active, 50M - 930M total parameters}
    \end{subfigure}
    \par\bigskip\bigskip
    \begin{subfigure}[t]{1.0\textwidth}
        \centering
        \includegraphics[width=\linewidth]{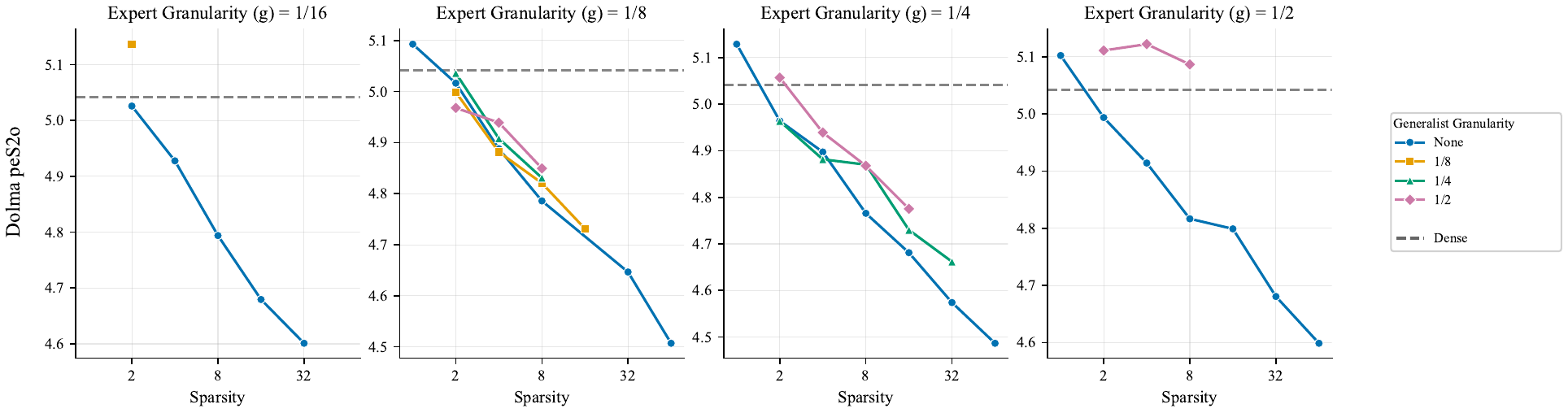}
        \caption{80M active, 80M - 765M total parameters}
    \end{subfigure}
    \par\bigskip\bigskip
    \begin{subfigure}[t]{1.0\textwidth}
        \centering
        \includegraphics[width=\linewidth]{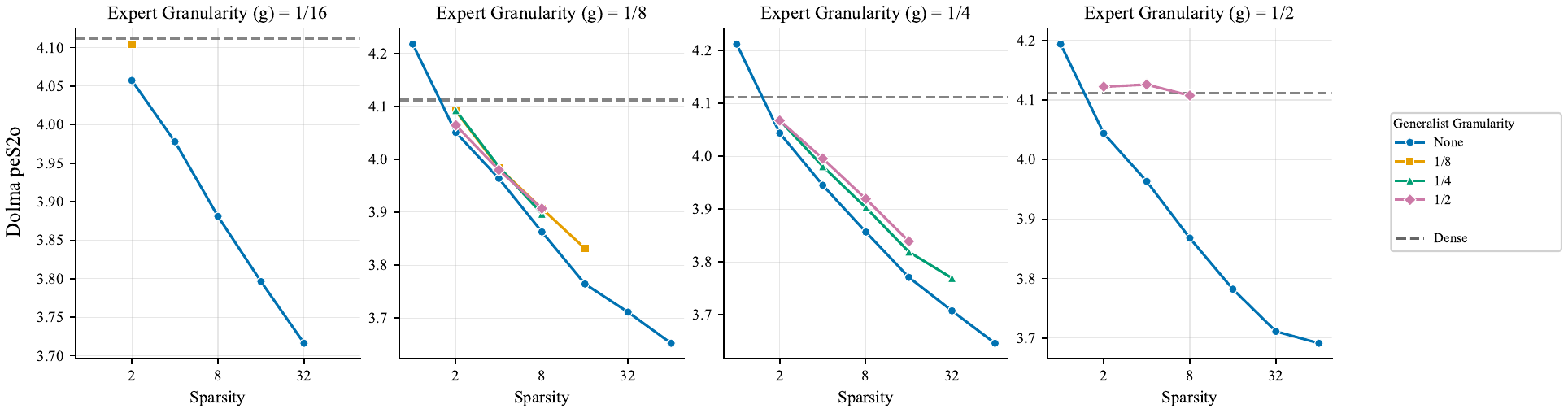}
        \caption{110M active, 110M - 1.4B total parameters}
    \end{subfigure}
    \caption{
    \textbf{The inclusion of a generalist consistently degrades performance in homogeneous MoEs (\S\ref{sec:expt_hetgen}).}
    We train MoE LMs which consist of some routed experts with granularity $g$, as well as a generalist with granularity $g_{gen}\in \{\frac{1}{2}, \frac{1}{4}, \frac{1}{8}\} $. We compare to settings with no generalist, only routed experts with granularity $g$. In all settings and configurations, the addition of any granularity generalist results in comparable or degraded performance. 
    }
    \label{fig:dolma_pes2o_gen}
\end{figure*}

\begin{figure*}[ht]
    \centering
    \begin{subfigure}[t]{1.0\textwidth}
        \centering
        \includegraphics[width=\linewidth]{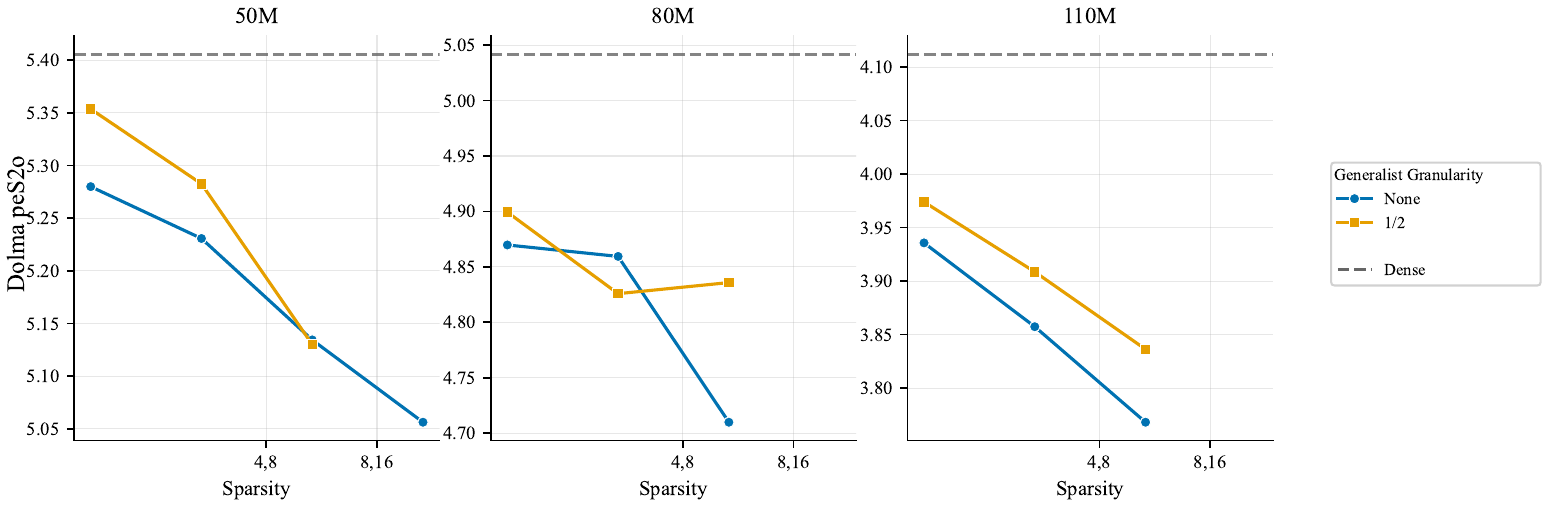}
    \end{subfigure}
    \caption{
    \textbf{The inclusion of a generalist consistently degrades performance in heterogeneous MoEs (\S\ref{sec:expt_hetgen}).}
    We train heterogeneous MoE LMs which consist of  routed experts with granularity $g_1, g_2$, as well as a generalist with granularity $g_{gen} = \frac{1}{2}$. We compare to settings with no generalist. In all settings and configurations, the addition of a generalist results in comparable or degraded performance. 
    }
    \label{fig:dolma_pes2o_hetgen}
\end{figure*}

\begin{figure*}[ht]
    \centering
    \begin{subfigure}[t]{\textwidth}
        \centering
        \begin{subfigure}[t]{0.45\textwidth}
            \includegraphics[width=\linewidth]{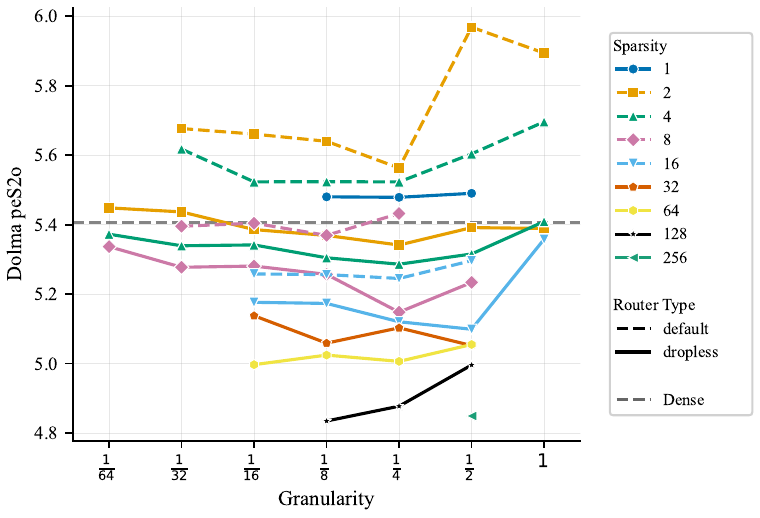}
            \caption{50M active, 50M - 930M total parameters}
        \end{subfigure}
    \hspace{1em}
        \begin{subfigure}[t]{0.45\textwidth}
            \centering
            \includegraphics[width=\linewidth]{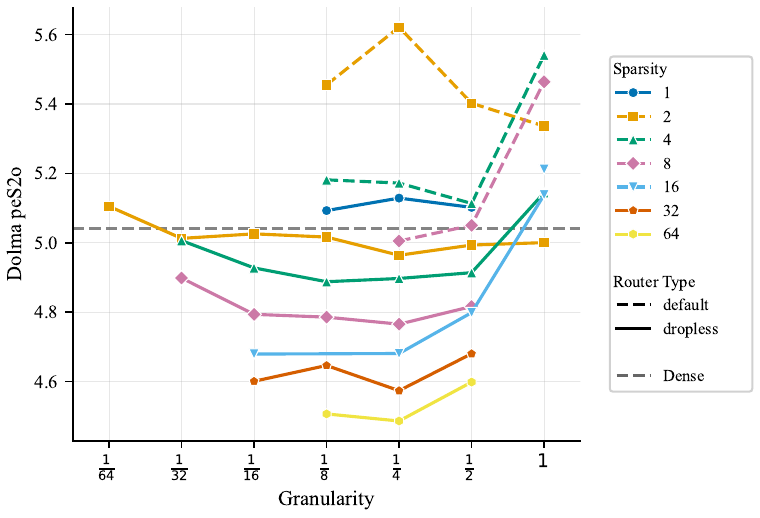}
            \caption{80M active, 80M - 765M total parameters}
        \end{subfigure}
    \end{subfigure}

    \par\bigskip\bigskip
    \begin{subfigure}[t]{0.45\textwidth}
        \centering
        \includegraphics[width=\linewidth]{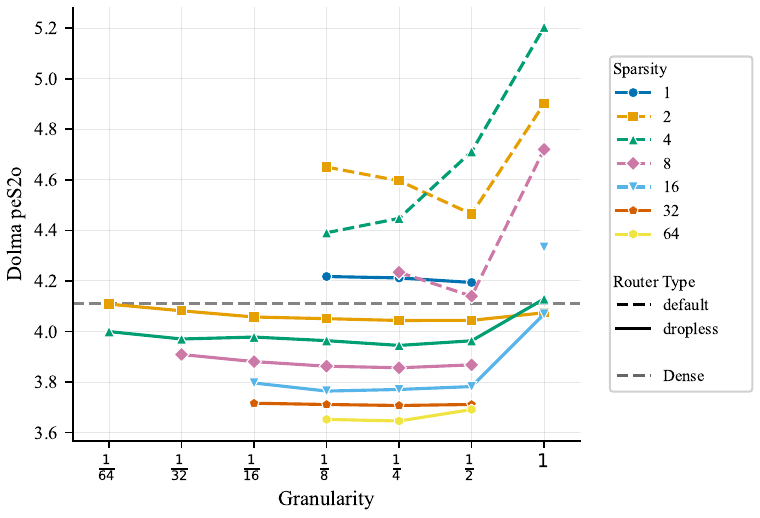}
        \caption{110M active, 110M - 1.4B total parameters}
    \end{subfigure}
    \caption{ 
    \textbf{Dropless routing outperforms default routing (\S\ref{sec:expt_router}).}
    We compare dropless routing to the default setting, which allow tokens to be dropped. Across all scales, we find that dropless routing outperforms or performs comparably to default routing. 
    }
    \label{fig:dolma_pes2o_dropless}
\end{figure*}

\begin{figure*}[ht]
    \centering
    \begin{subfigure}[t]{0.45\textwidth}
        \centering
        \includegraphics[width=\linewidth]{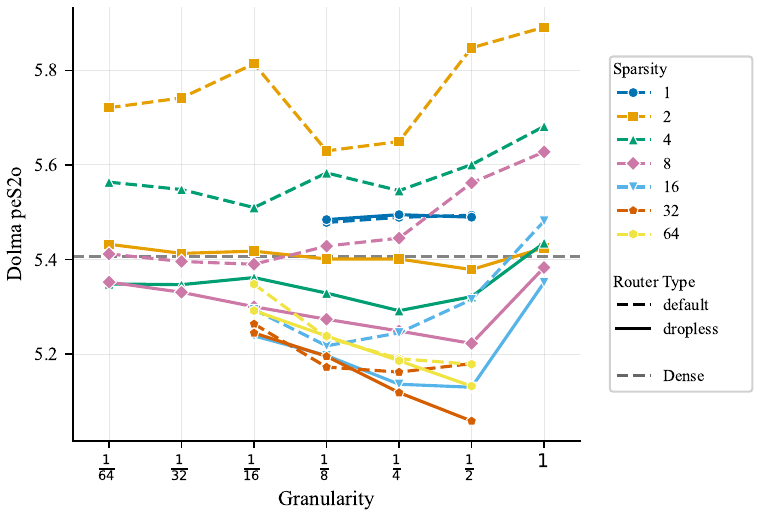}
        \caption{50M active, 50M - 930M total parameters}
    \end{subfigure}
    \hspace{1em}
    \begin{subfigure}[t]{0.45\textwidth}
        \centering
        \includegraphics[width=\linewidth]{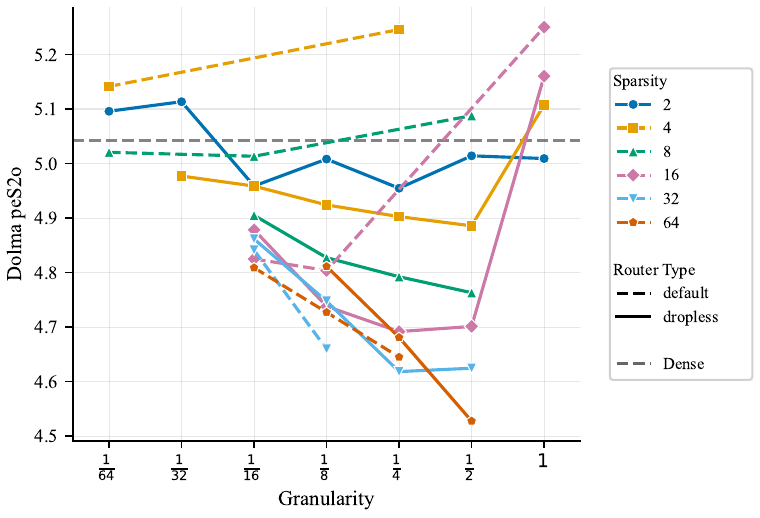}
        \caption{80M active, 80M - 765M total parameters}
    \end{subfigure}
    \caption{
    \textbf{Dropless routing, with bias $\gamma=\num{1e-3}$ (\S\ref{sec:expt_router}).} 
    As in Figure~\ref{fig:lm_avg_dropless}, we compare dropless routing to the default setting, which allow tokens to be dropped. Across all scales, we find that dropless routing outperforms or performs comparably to default routing. We see here with additional higher sparsity default routing runs that as sparsity increases, default routing performance approaches that of dropless routing.
    }
    \label{fig:dolma_pes2o_dropless_with_lf}
\end{figure*}

\begin{figure*}[ht]
    \centering
    \begin{subfigure}[]{\textwidth}
        \centering
        \includegraphics[width=0.46\linewidth]{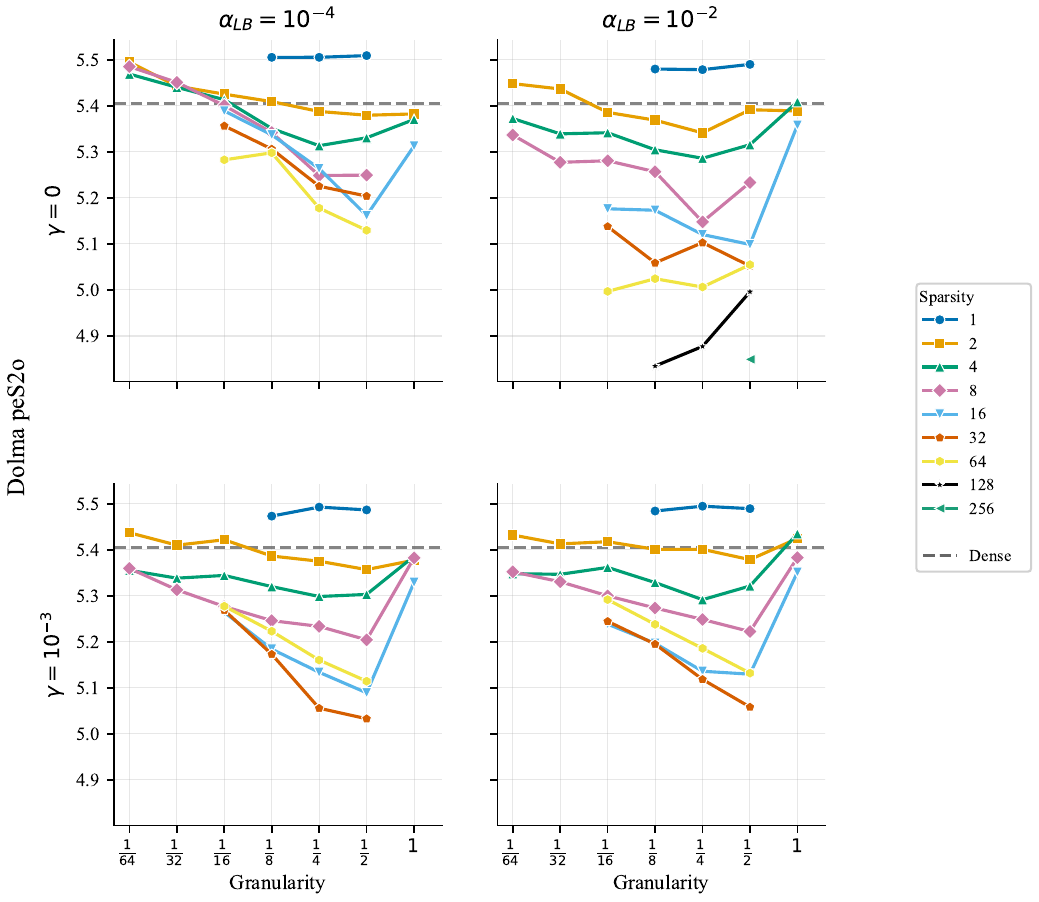}
        \hspace{1em}
        \includegraphics[width=0.46\linewidth]{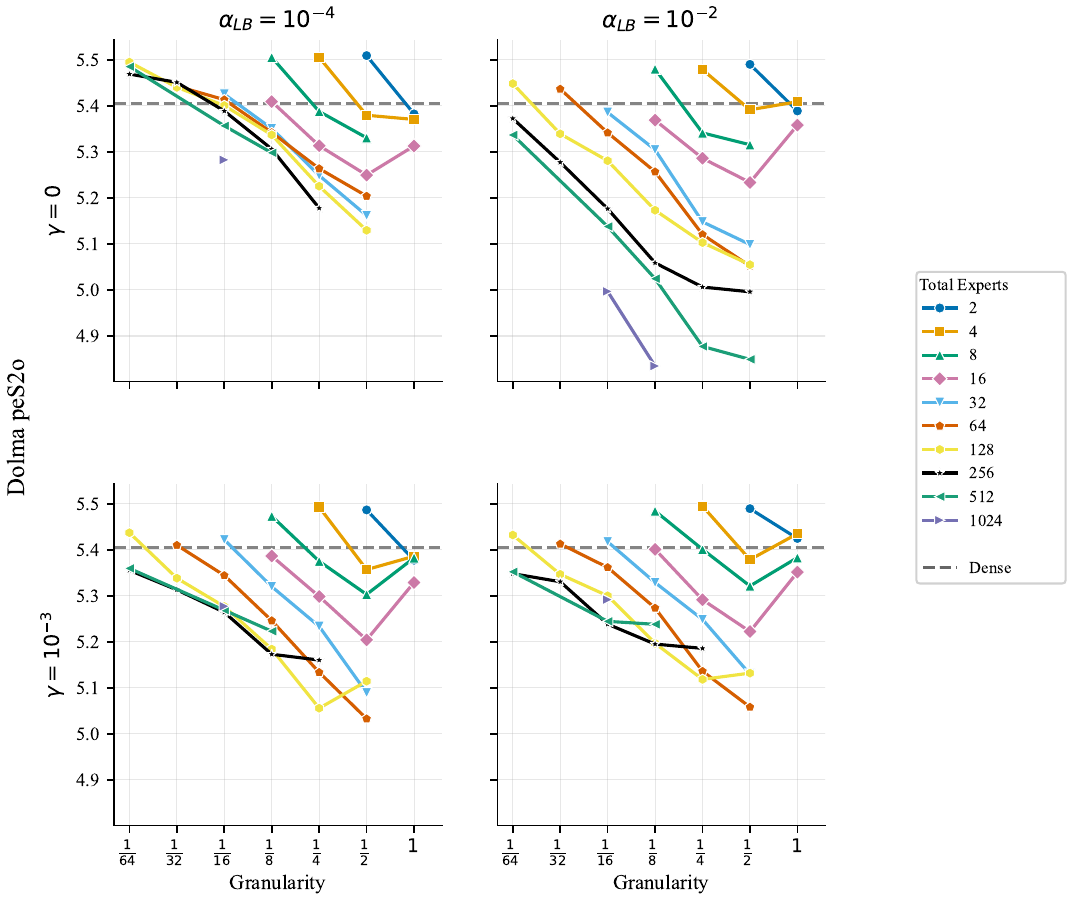}
        \caption{50M active, 50M - 930M total parameters}
    \end{subfigure}
    \par\bigskip\bigskip
    \begin{subfigure}[]{\textwidth}
        \centering
        \includegraphics[width=0.46\linewidth]{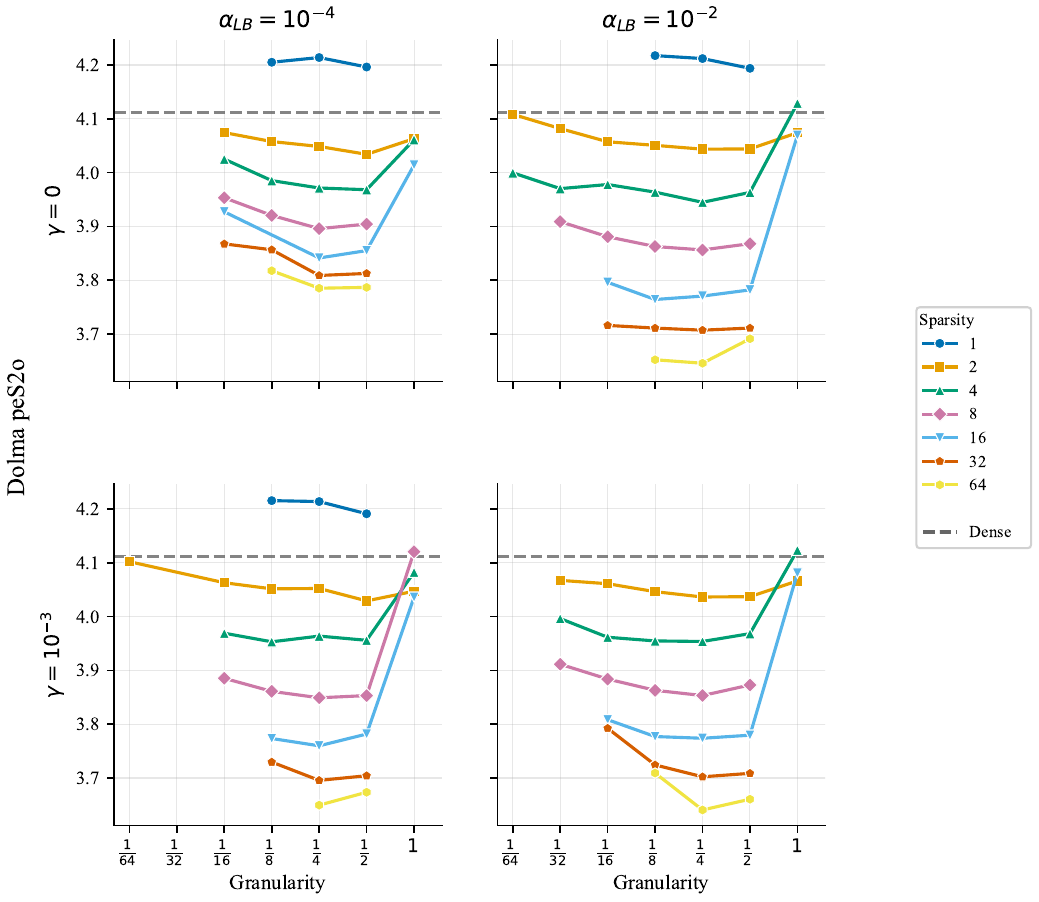}
        \hspace{1em}
        \includegraphics[width=0.46\linewidth]{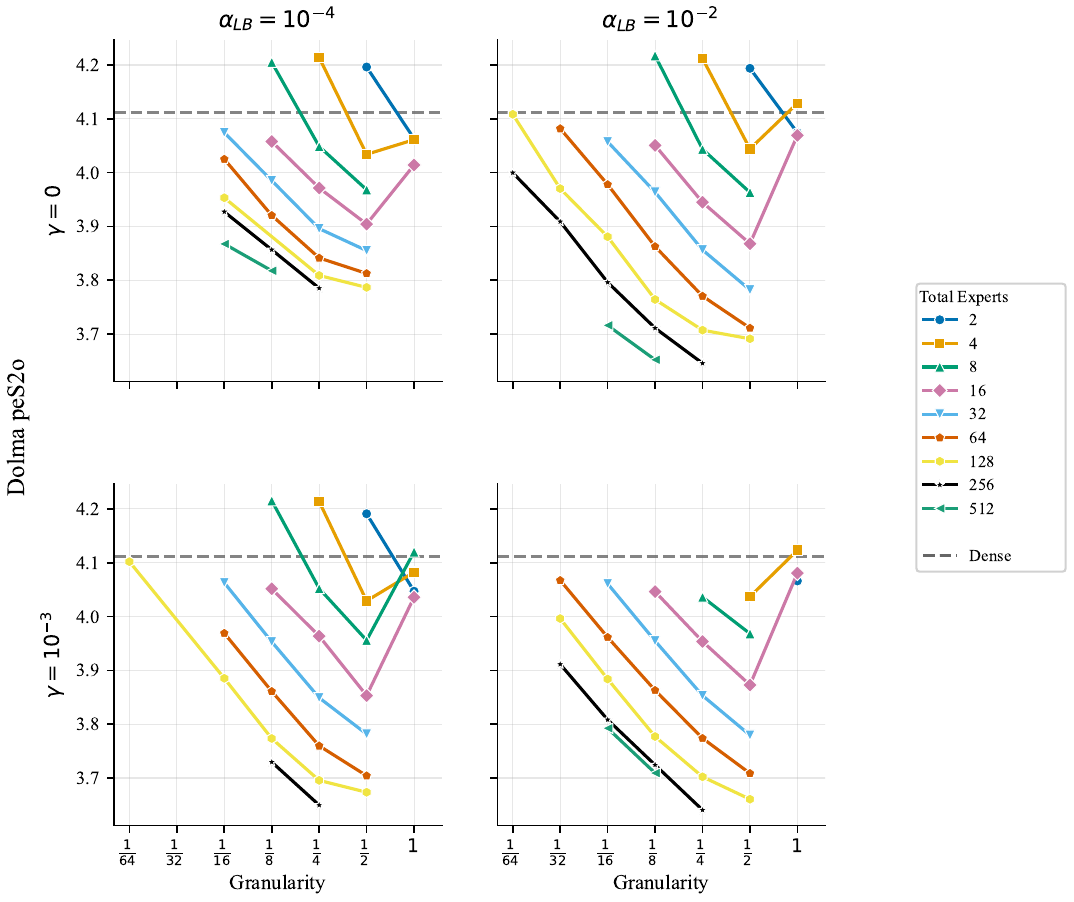}
        \caption{80M active, 80M - 765M total parameters}
    \end{subfigure}
    \par\bigskip\bigskip
    \begin{subfigure}[t]{\textwidth}
        \centering
        \includegraphics[width=0.46\linewidth]{figures/lm/dolma_pes2o-validation/ce_loss/lb_sweep_hgn_gxs_110M.pdf}
        \hspace{1em}
        \includegraphics[width=0.46\linewidth]{figures/lm/dolma_pes2o-validation/ce_loss/lb_sweep_hgn_gxn_110M.pdf}
        \caption{110M active, 110M - 1.4B total parameters}
    \end{subfigure}

    \end{figure*} 

\clearpage  

\begin{figure*}[ht]
    \addtocounter{figure}{-1}
    \centering
    \begin{subfigure}[t]{\textwidth}
        \centering
        \includegraphics[width=0.46\linewidth]{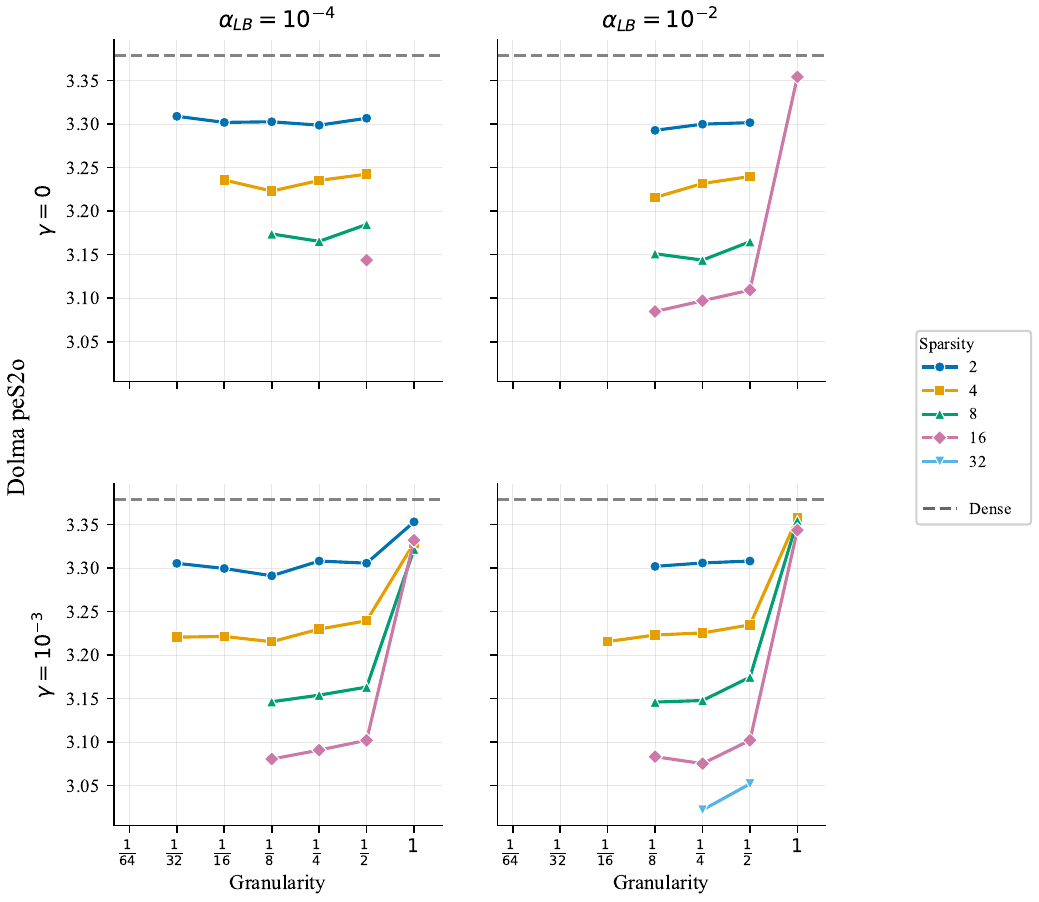}
        \hspace{1em}
        \includegraphics[width=0.46\linewidth]{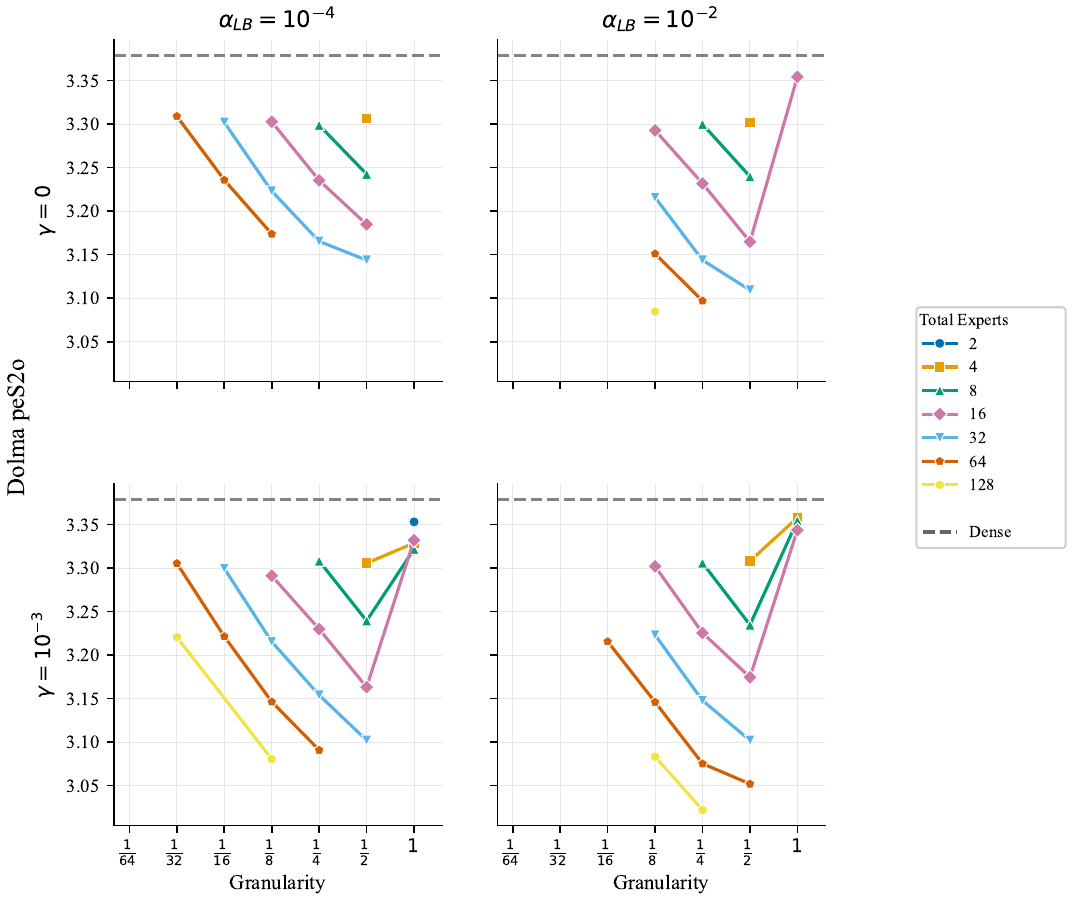}
        \caption{200M active, 200M - 3.3B total parameters}
    \end{subfigure}
    \par\bigskip\bigskip
    \begin{subfigure}[t]{\textwidth}
        \centering
        \includegraphics[width=0.3\linewidth]{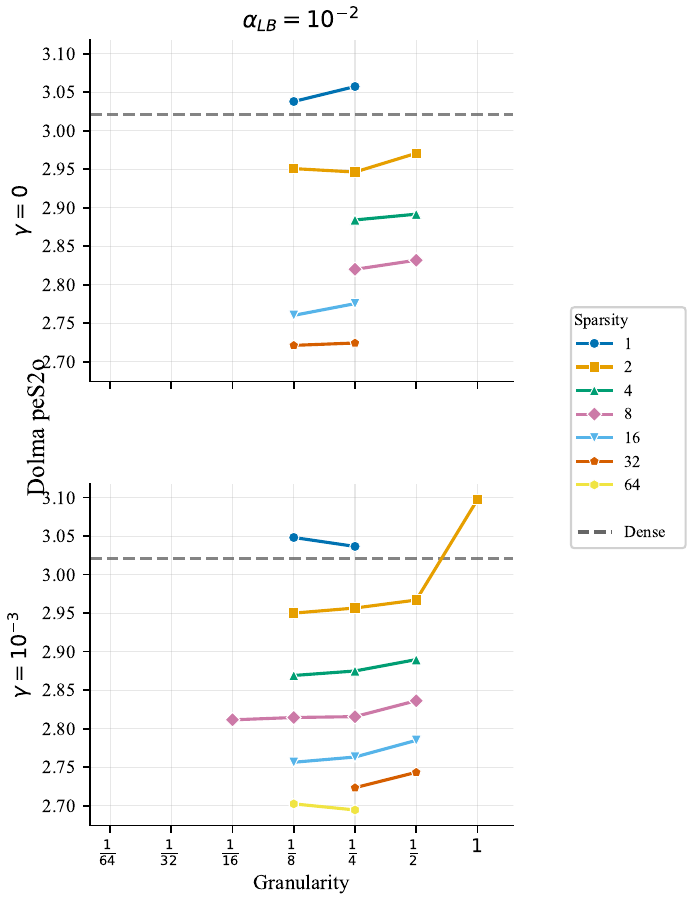}
        \hspace{1em}
        \includegraphics[width=0.3\linewidth]{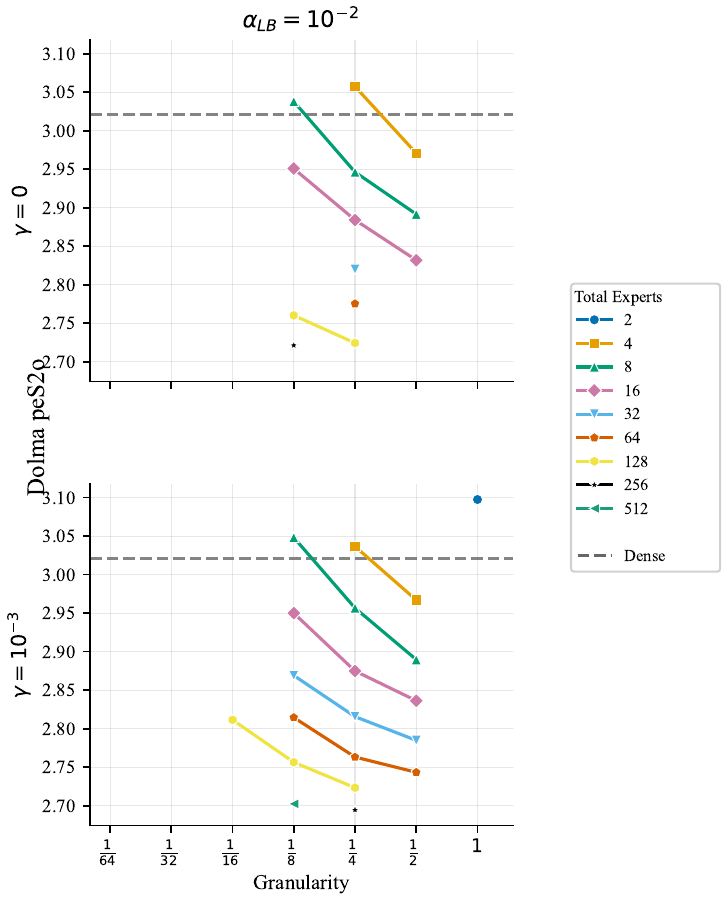}
        \caption{300M active, 300M - 6.6B total parameters}
    \end{subfigure}

    \caption{
    \textbf{Load balancing mechanisms must be tuned correctly (\S\ref{sec:expt_router}).}
    We consider load balancing loss weight $\alpha_{LB} \in \{\num{1e-2}, \num{1e-4}\}$ and loss-free load balancing with bias $\gamma\in\{0, \num{1e-3}\}$ ($\gamma=0$ indicates no loss-free mechanism). Results show that poorly chosen hyperparameters, such as high bias $\gamma = 1e-3$ with total experts $n\geq 512$, may impair performance. However, all settings other than $(\alpha_{LB}=\num{1e-2}, \gamma=\num{1e-3})$ perform comparably for $n \leq 512$, suggesting that a wide range of load balancing settings achieve near-optimal performance. 
    }
    \label{fig:dolma_pes2o_lb}
\end{figure*}

%% file: fig_tex/lm/dolma_reddit.tex
\begin{figure*}[!ht]
    \centering
        \begin{subfigure}[t]{\textwidth}
        \begin{subfigure}[t]{0.33\textwidth}
            \centering
            \caption*{\scriptsize Fixed total experts (n)}
            \includegraphics[width=\linewidth]{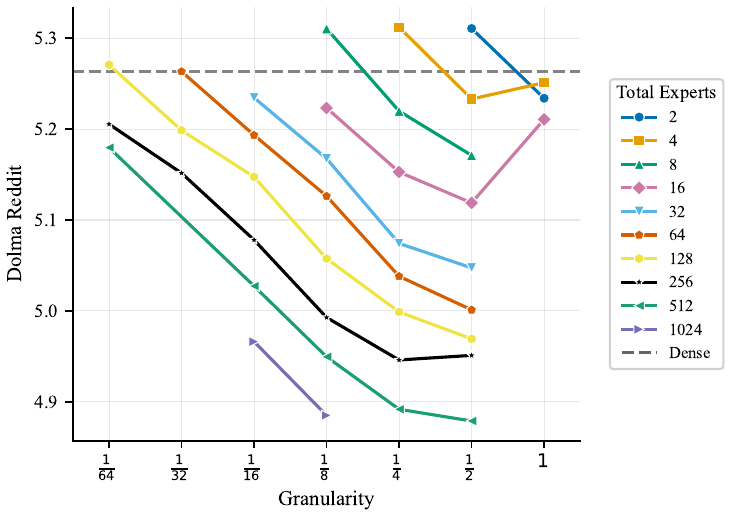}
        \end{subfigure}
        \begin{subfigure}[t]{0.33\textwidth}
            \centering
            \caption*{\scriptsize Fixed granularity (g)}
            \includegraphics[width=\linewidth]{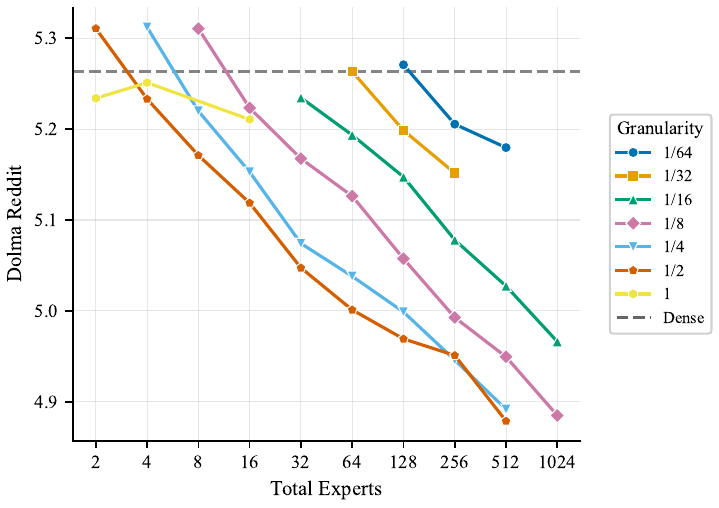}
        \end{subfigure}
        \begin{subfigure}[t]{0.33\textwidth}
            \centering
            \caption*{\scriptsize Fixed activation sparsity (s)}
            \includegraphics[width=\linewidth]{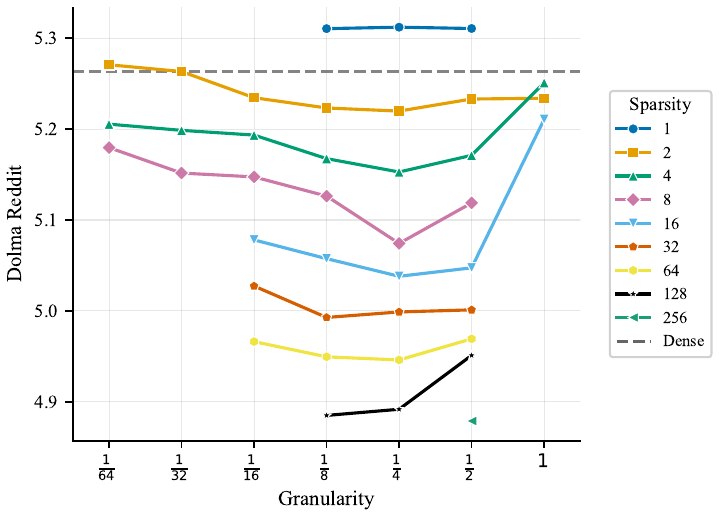}
        \end{subfigure}
        \caption{50M active, 50M - 930M total parameters}
    \end{subfigure}
\par\bigskip\bigskip
    \begin{subfigure}[t]{\textwidth}
        \begin{subfigure}[t]{0.33\textwidth}
            \centering
            \includegraphics[width=\linewidth]{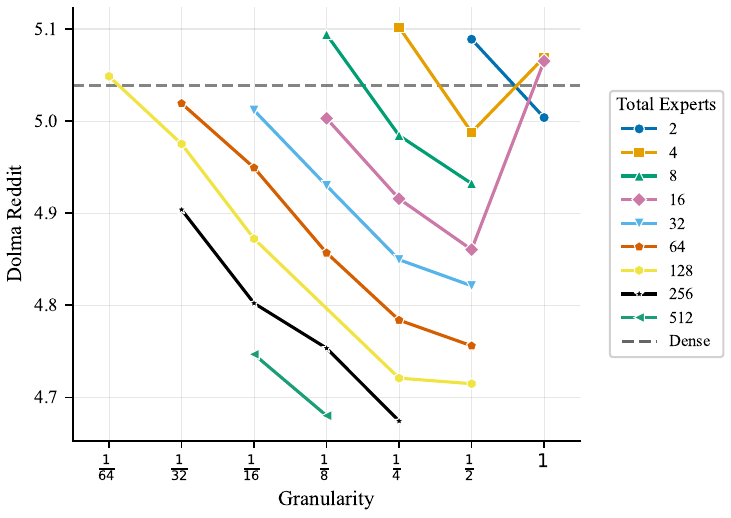}
        \end{subfigure}
        \begin{subfigure}[t]{0.33\textwidth}
            \centering
            \includegraphics[width=\linewidth]{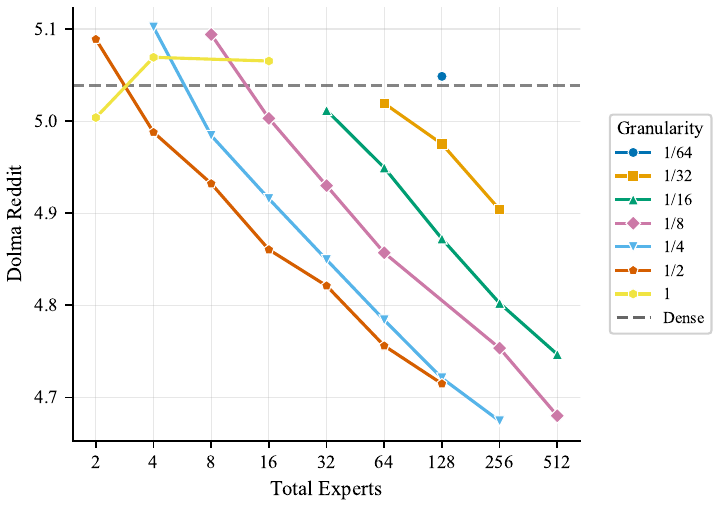}
        \end{subfigure}
        \begin{subfigure}[t]{0.33\textwidth}
            \centering
            \includegraphics[width=\linewidth]{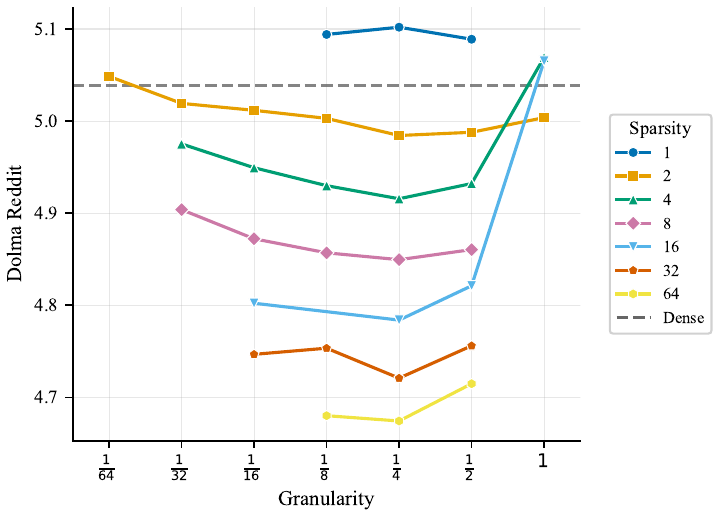}
        \end{subfigure}
        \caption{80M active, 80M - 765M total parameters}
    \end{subfigure}
    \par\bigskip\bigskip
        \begin{subfigure}[t]{\textwidth}
        \begin{subfigure}[t]{0.33\textwidth}
            \centering
            \includegraphics[width=\linewidth]{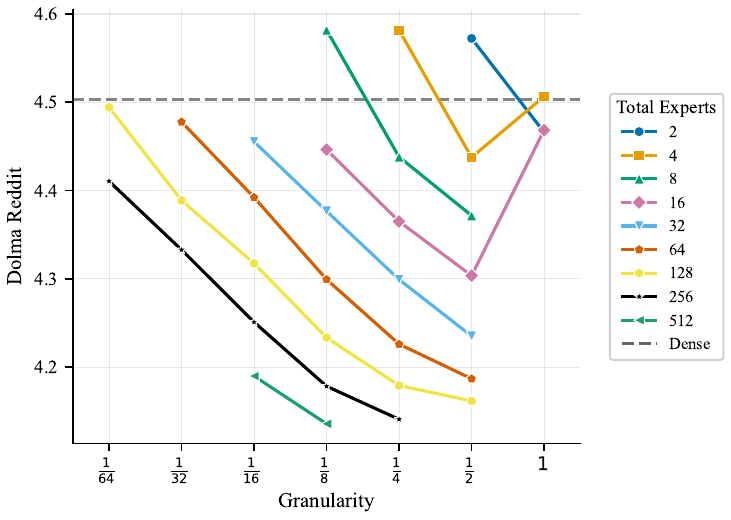}
        \end{subfigure}
        \begin{subfigure}[t]{0.33\textwidth}
            \centering
            \includegraphics[width=\linewidth]{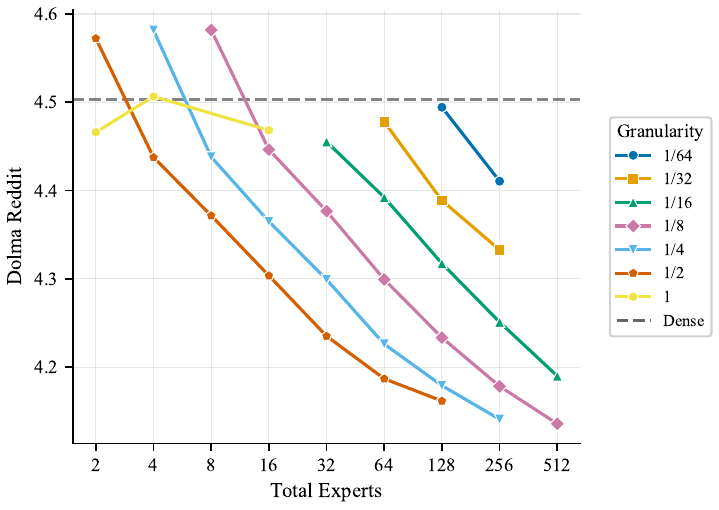}
        \end{subfigure}
        \begin{subfigure}[t]{0.33\textwidth}
            \centering
            \includegraphics[width=\linewidth]{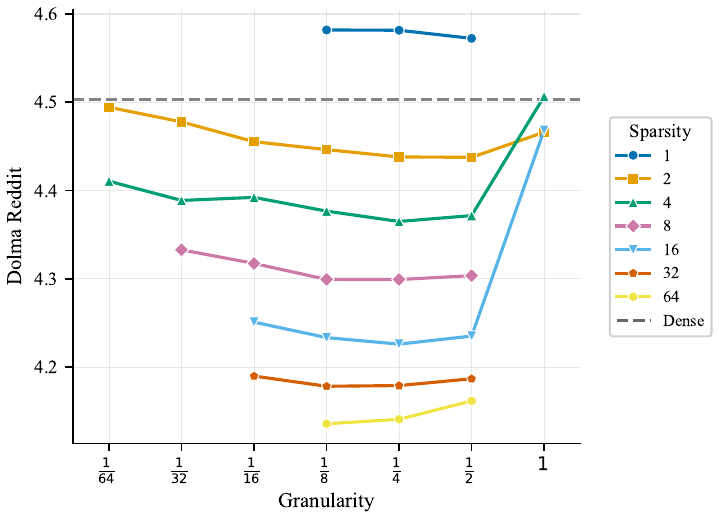}
        \end{subfigure}
        \caption{110M active, 110M - 1.4B total parameters}
    \end{subfigure}
    \end{figure*}

\clearpage  

\begin{figure*}[!ht]
        \addtocounter{figure}{-1}
    \begin{subfigure}[t]{\textwidth}
        \addtocounter{subfigure}{3}
        \begin{subfigure}[t]{0.33\textwidth}
            \centering
            \caption*{\scriptsize Fixed total experts (n)}
            \includegraphics[width=\linewidth]{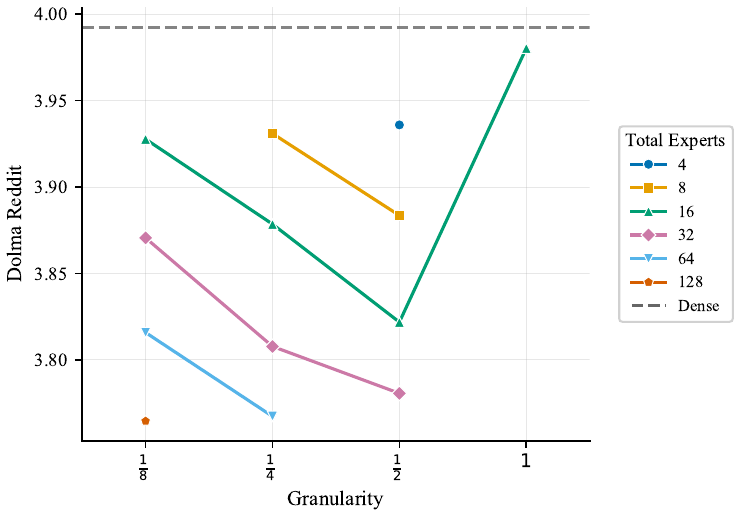}
        \end{subfigure}
        \begin{subfigure}[t]{0.33\textwidth}
            \centering
            \caption*{\scriptsize Fixed granularity (g)}
            \includegraphics[width=\linewidth]{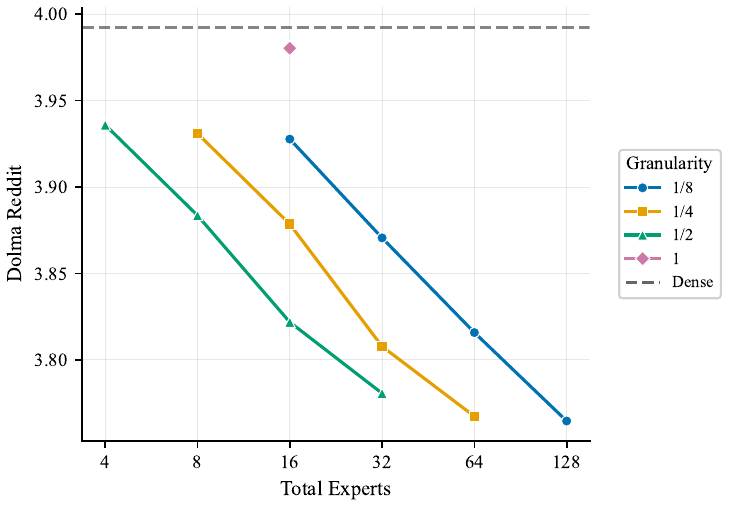}
        \end{subfigure}
        \begin{subfigure}[t]{0.33\textwidth}
            \centering
            \caption*{\scriptsize Fixed activation sparsity (s)}
            \includegraphics[width=\linewidth]{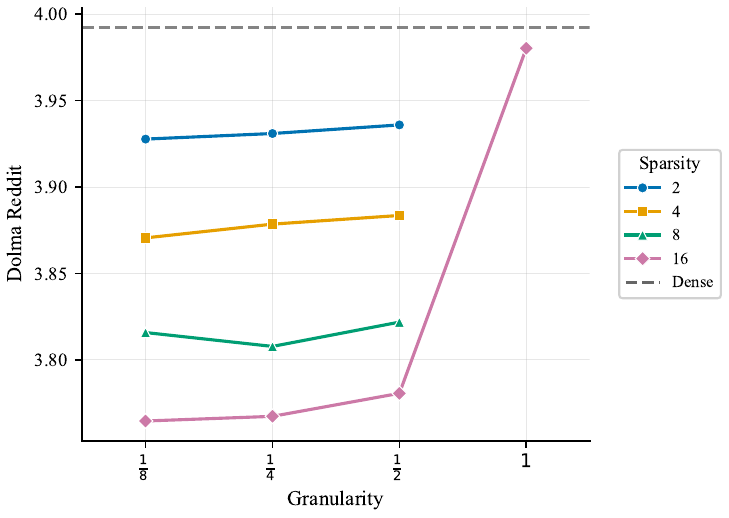}
        \end{subfigure}
        \caption{200M active, 200M - 3.3B total parameters}
    \end{subfigure}
    \par\bigskip\bigskip
        \begin{subfigure}[t]{\textwidth}
        \begin{subfigure}[t]{0.33\textwidth}
            \centering
            \includegraphics[width=\linewidth]{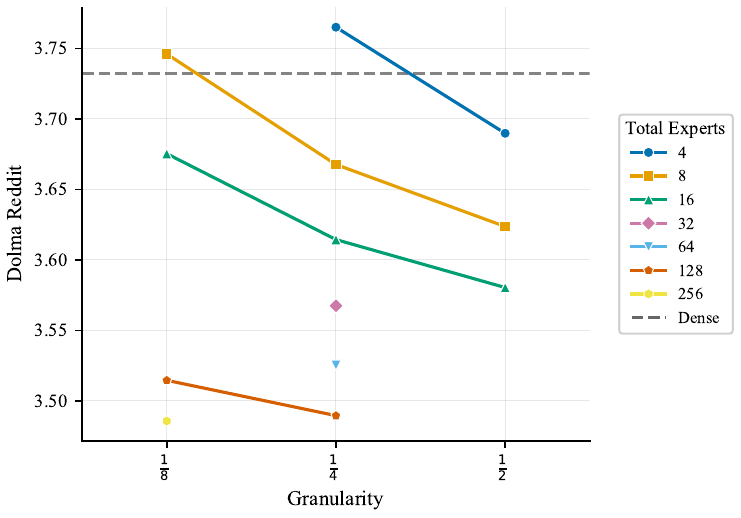}
        \end{subfigure}
        \begin{subfigure}[t]{0.33\textwidth}
            \centering
            \includegraphics[width=\linewidth]{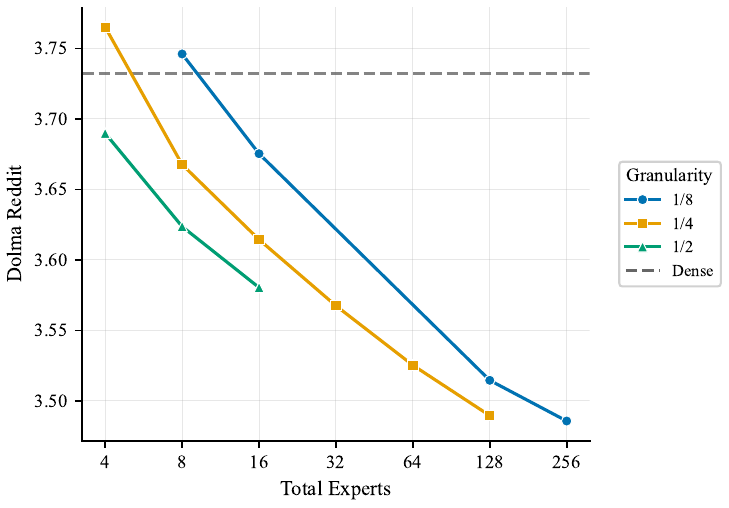}
        \end{subfigure}
        \begin{subfigure}[t]{0.33\textwidth}
            \centering
            \includegraphics[width=\linewidth]{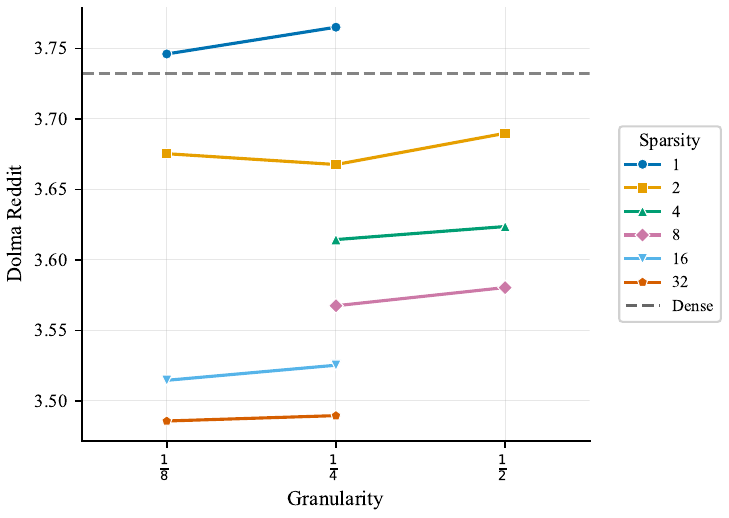}
        \end{subfigure}
        \caption{300M active, 300M - 6.6B total parameters}
    \end{subfigure}

    \caption{
    \textbf{Increasing inactive expert parameters via expert size (left) or total count (center) improves performance in MoEs (\S\ref{sec:expt_main}).} This effect is seen both when holding total number of experts fixed (left) and when holding expert granularity fixed (center). In general, increasing total parameters results in improved performance.  \textbf{Optimal tradeoff between expert count and granularity varies in MoEs (right). (\S\ref{sec:expt_main})}
    At each activation sparsity $s$ (equivalently, at each total parameter count), the optimal (total expert count, expert granularity) configuration varies. As $s$ increases, optimal expert granularity remains nearly fixed, suggesting that sparsity should be scaled up primarily by increasing total expert count $n$, while maintaining a near constant, slowly increasing expert granularity $g$. 
    }
    \label{fig:dolma_reddit_experts}
\end{figure*}

\begin{figure*}[!ht]
    \centering
    
    \begin{subfigure}[t]{0.46\textwidth}
        \centering
        \includegraphics[width=\linewidth]{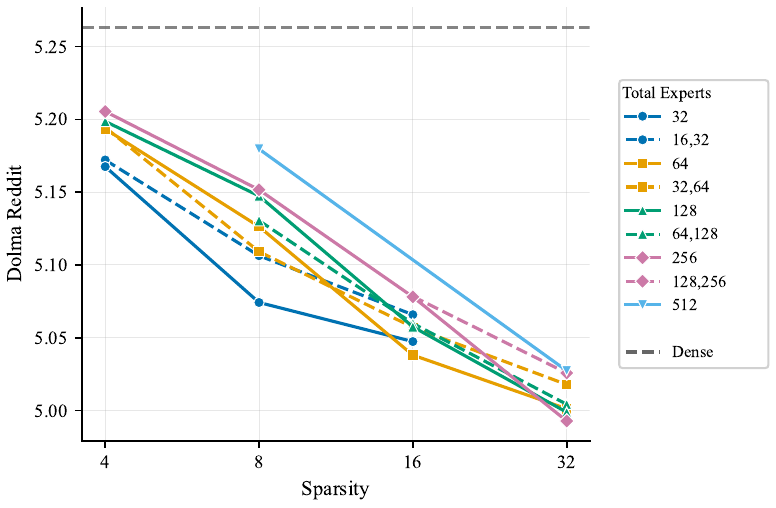}
        \caption{50M active, 50M - 930M total parameters}
    \end{subfigure}
    \vspace{1em}
    \begin{subfigure}[t]{0.46\textwidth}
        \centering
        \includegraphics[width=\linewidth]{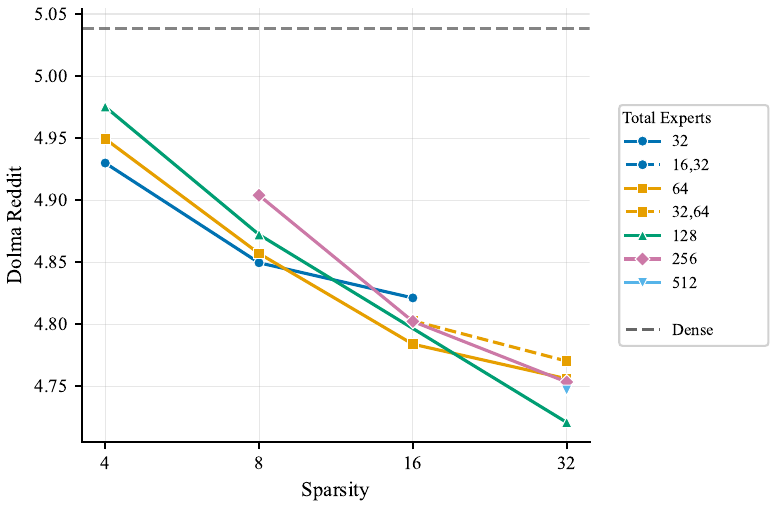}
        \caption{80M active, 80M - 765M total parameters}
    \end{subfigure}
    \caption{
    \textbf{Heterogeneity of expert size alone does not improve MoE performance (\S\ref{sec:expt_hetgen}).} To explore the potential benefits of their architectural flexibility, we compare heterogeneous MoEs (indicated by dotted lines) to active- and total-parameter-matched homogeneous MoEs. Heterogeneity alone does not result in performance gains, as, at each activation sparsity $s$, heterogeneous MoEs with $n_1, n_2 = a, b$ lie between or near the 2 closest homogeneous MoEs, with $n=a$ and with $n=b$.
    }
    \label{fig:dolma_reddit_het}
\end{figure*}

\begin{figure*}[!ht]
    \centering
    
    \begin{subfigure}[t]{1.0\textwidth}
        \centering
        \includegraphics[width=\linewidth]{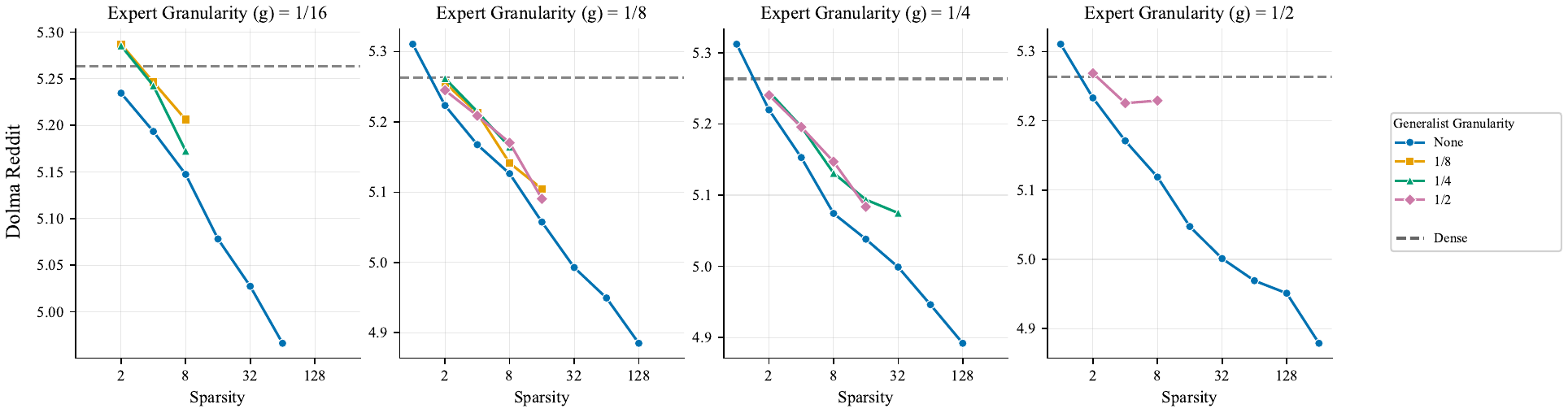}
        \caption{50M active, 50M - 930M total parameters}
    \end{subfigure}
    \par\bigskip\bigskip
    \begin{subfigure}[t]{1.0\textwidth}
        \centering
        \includegraphics[width=\linewidth]{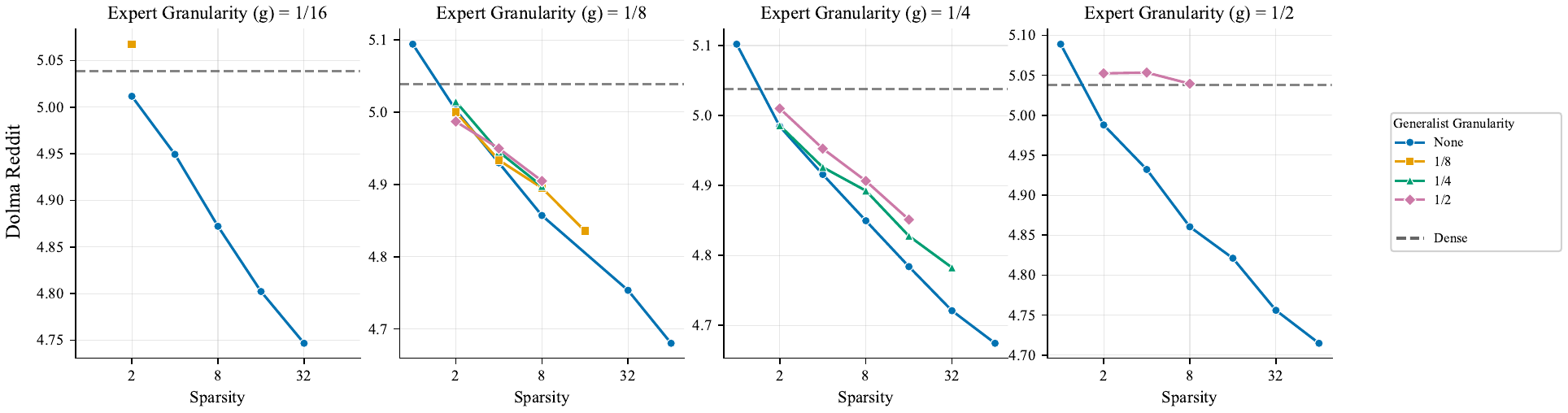}
        \caption{80M active, 80M - 765M total parameters}
    \end{subfigure}
    \par\bigskip\bigskip
    \begin{subfigure}[t]{1.0\textwidth}
        \centering
        \includegraphics[width=\linewidth]{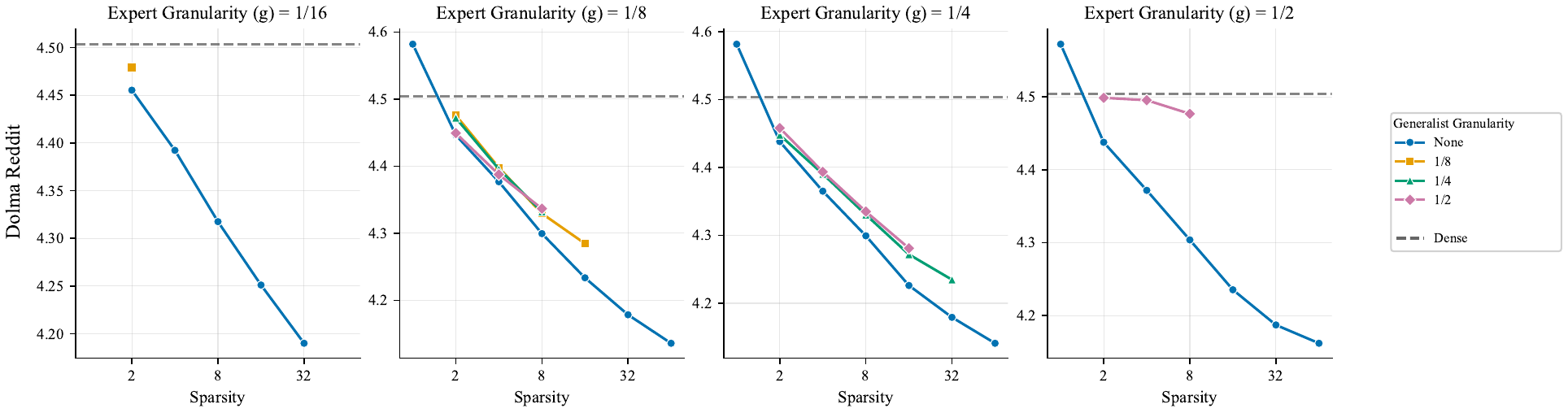}
        \caption{110M active, 110M - 1.4B total parameters}
    \end{subfigure}
    \caption{
    \textbf{The inclusion of a generalist consistently degrades performance in homogeneous MoEs (\S\ref{sec:expt_hetgen}).}
    We train MoE LMs which consist of some routed experts with granularity $g$, as well as a generalist with granularity $g_{gen}\in \{\frac{1}{2}, \frac{1}{4}, \frac{1}{8}\} $. We compare to settings with no generalist, only routed experts with granularity $g$. In all settings and configurations, the addition of any granularity generalist results in comparable or degraded performance. 
    }
    \label{fig:dolma_reddit_gen}
\end{figure*}

\begin{figure*}[ht]
    \centering
    \begin{subfigure}[t]{1.0\textwidth}
        \centering
        \includegraphics[width=\linewidth]{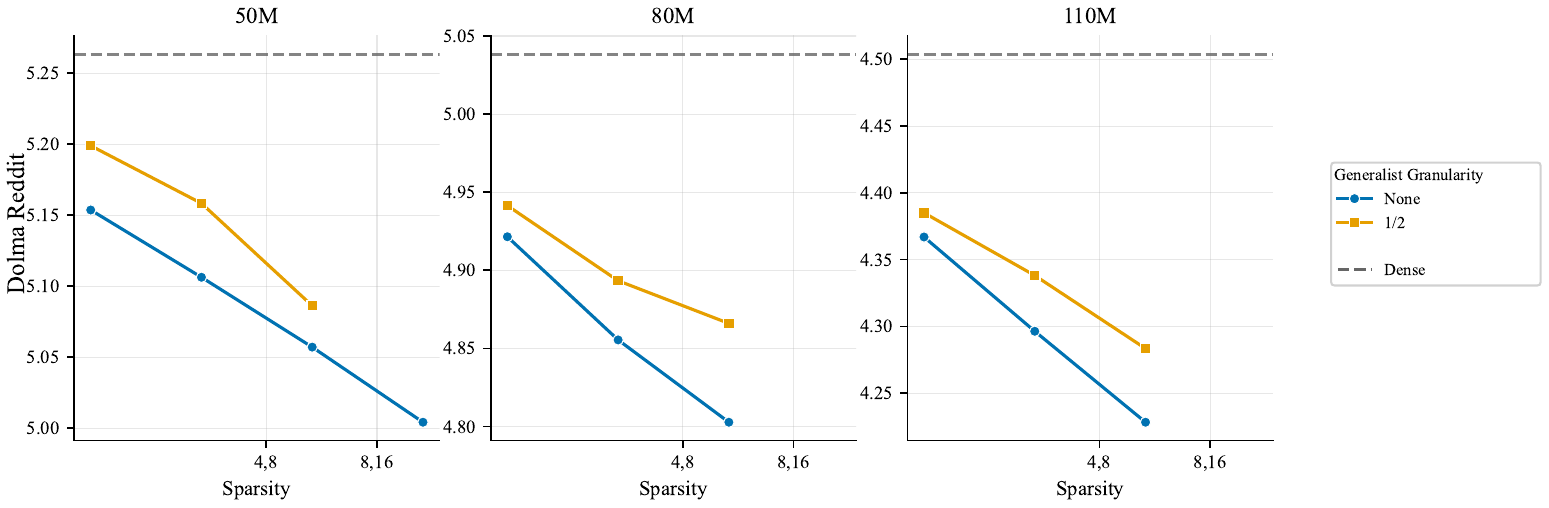}
    \end{subfigure}
    \caption{
    \textbf{The inclusion of a generalist consistently degrades performance in heterogeneous MoEs (\S\ref{sec:expt_hetgen}).}
    We train heterogeneous MoE LMs which consist of  routed experts with granularity $g_1, g_2$, as well as a generalist with granularity $g_{gen} = \frac{1}{2}$. We compare to settings with no generalist. In all settings and configurations, the addition of a generalist results in comparable or degraded performance. 
    }
    \label{fig:dolma_reddit_hetgen}
\end{figure*}

\begin{figure*}[ht]
    \centering
    \begin{subfigure}[t]{\textwidth}
        \centering
        \begin{subfigure}[t]{0.45\textwidth}
            \includegraphics[width=\linewidth]{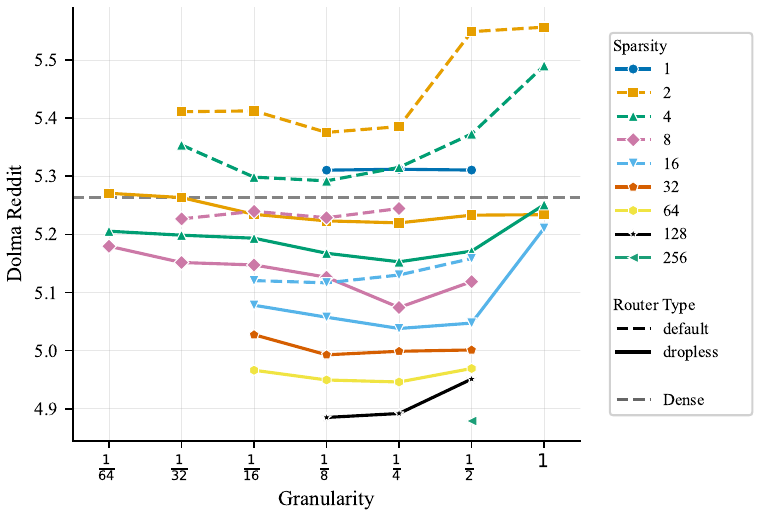}
            \caption{50M active, 50M - 930M total parameters}
        \end{subfigure}
    \hspace{1em}
        \begin{subfigure}[t]{0.45\textwidth}
            \centering
            \includegraphics[width=\linewidth]{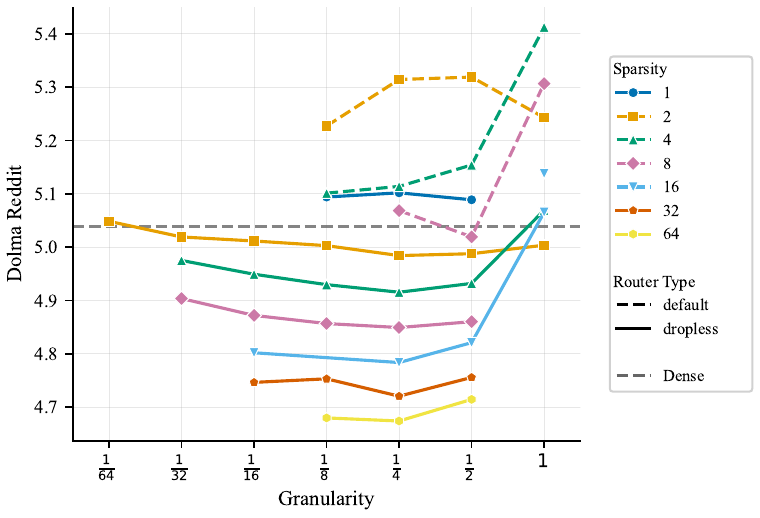}
            \caption{80M active, 80M - 765M total parameters}
        \end{subfigure}
    \end{subfigure}

    \par\bigskip\bigskip
    \begin{subfigure}[t]{0.45\textwidth}
        \centering
        \includegraphics[width=\linewidth]{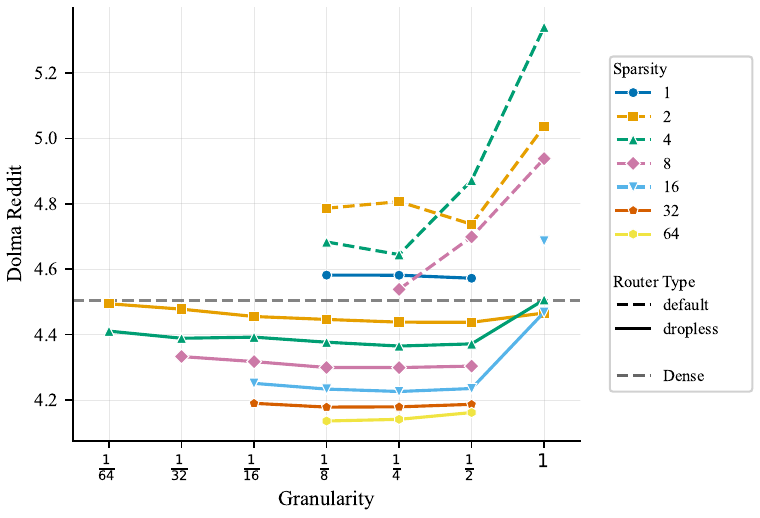}
        \caption{110M active, 110M - 1.4B total parameters}
    \end{subfigure}
    \caption{ 
    \textbf{Dropless routing outperforms default routing (\S\ref{sec:expt_router}).}
    We compare dropless routing to the default setting, which allow tokens to be dropped. Across all scales, we find that dropless routing outperforms or performs comparably to default routing. 
    }
    \label{fig:dolma_reddit_dropless}
\end{figure*}

\begin{figure*}[ht]
    \centering
    \begin{subfigure}[t]{0.45\textwidth}
        \centering
        \includegraphics[width=\linewidth]{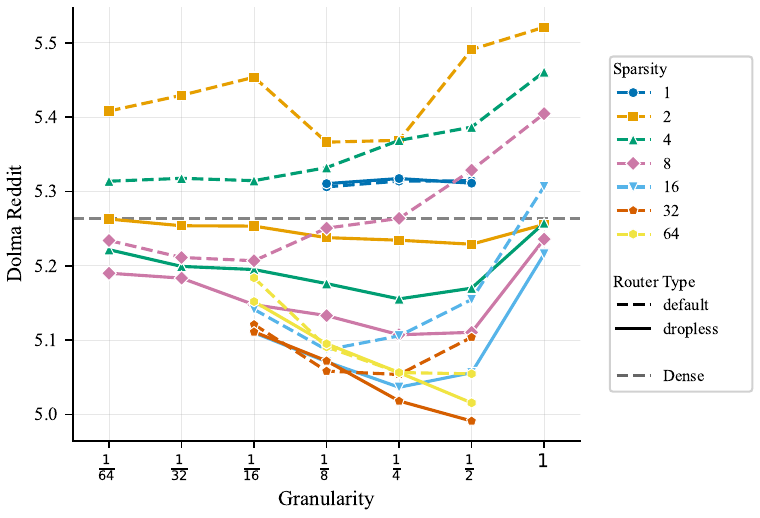}
        \caption{50M active, 50M - 930M total parameters}
    \end{subfigure}
    \hspace{1em}
    \begin{subfigure}[t]{0.45\textwidth}
        \centering
        \includegraphics[width=\linewidth]{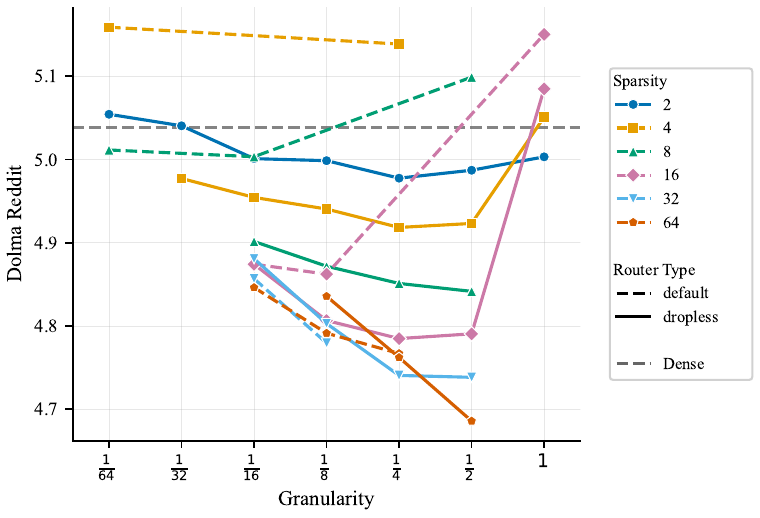}
        \caption{80M active, 80M - 765M total parameters}
    \end{subfigure}
    \caption{
    \textbf{Dropless routing, with bias $\gamma=\num{1e-3}$ (\S\ref{sec:expt_router}).} 
    As in Figure~\ref{fig:lm_avg_dropless}, we compare dropless routing to the default setting, which allow tokens to be dropped. Across all scales, we find that dropless routing outperforms or performs comparably to default routing. We see here with additional higher sparsity default routing runs that as sparsity increases, default routing performance approaches that of dropless routing.
    }
    \label{fig:dolma_reddit_dropless_with_lf}
\end{figure*}

\begin{figure*}[ht]
    \centering
    \begin{subfigure}[]{\textwidth}
        \centering
        \includegraphics[width=0.46\linewidth]{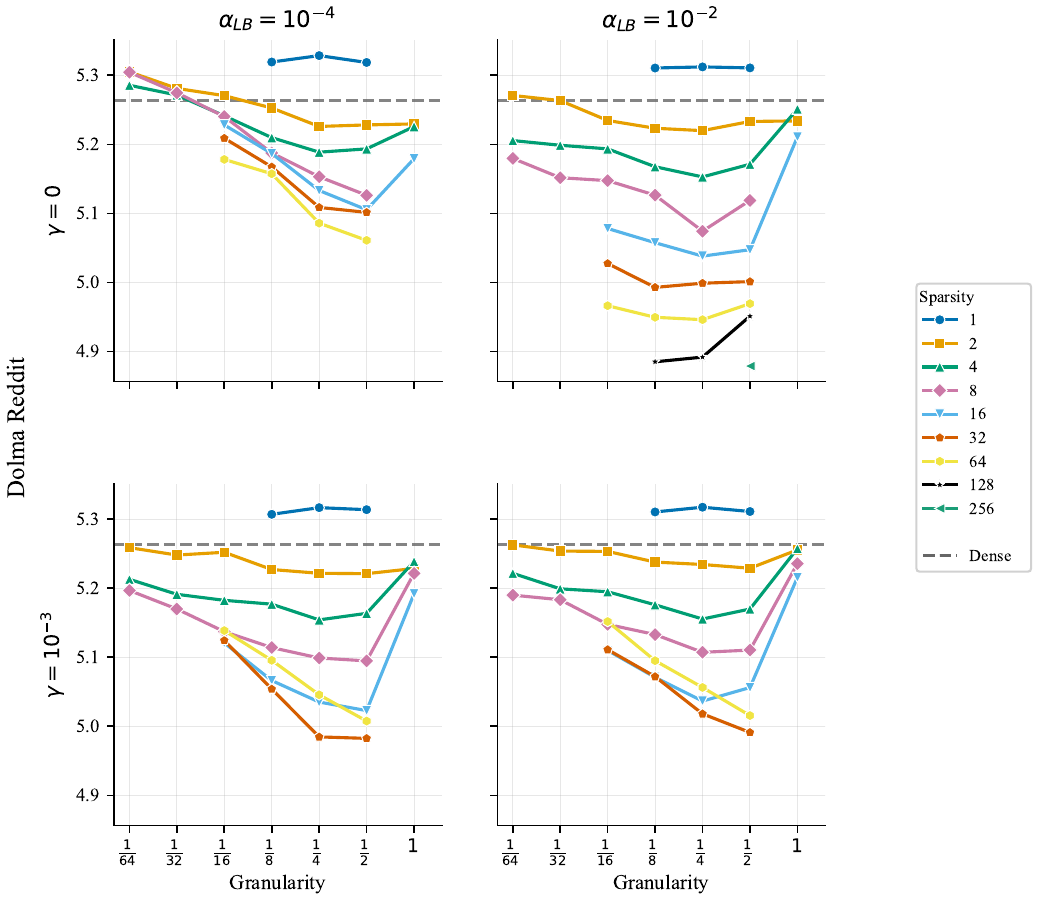}
        \hspace{1em}
        \includegraphics[width=0.46\linewidth]{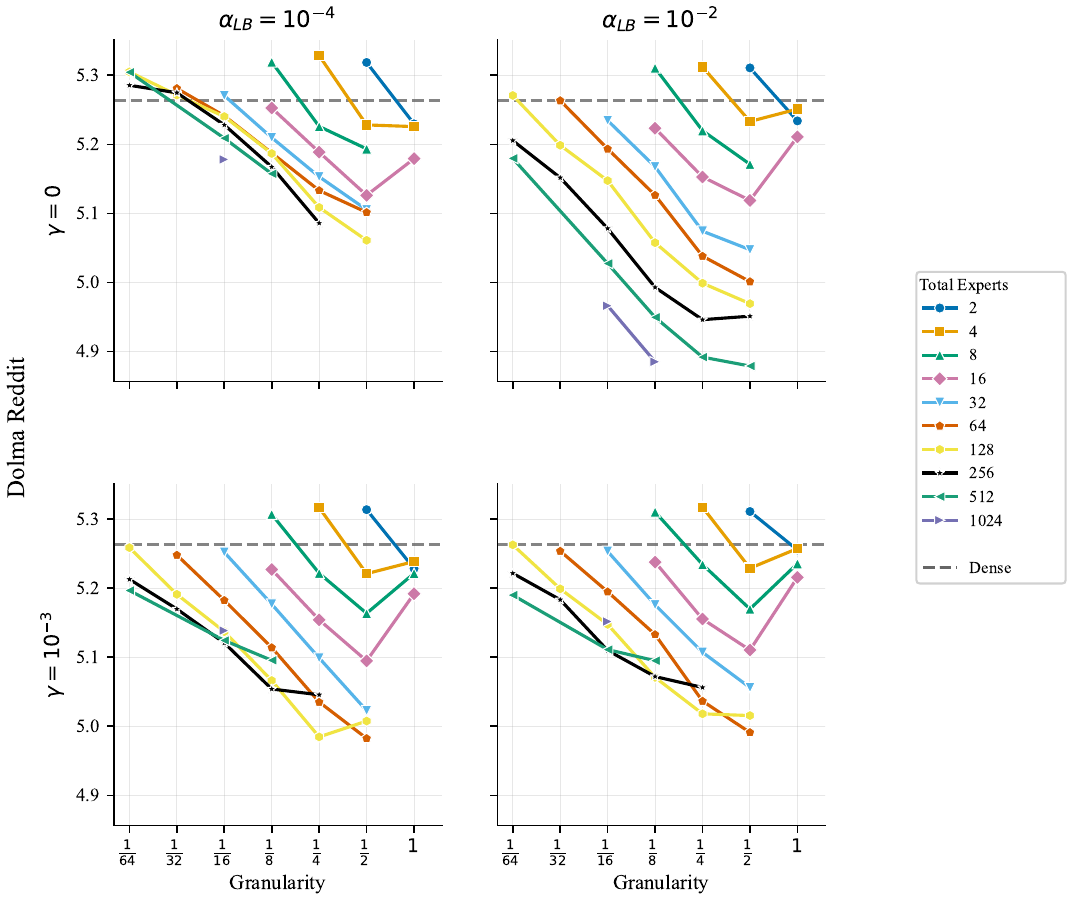}
        \caption{50M active, 50M - 930M total parameters}
    \end{subfigure}
    \par\bigskip\bigskip
    \begin{subfigure}[]{\textwidth}
        \centering
        \includegraphics[width=0.46\linewidth]{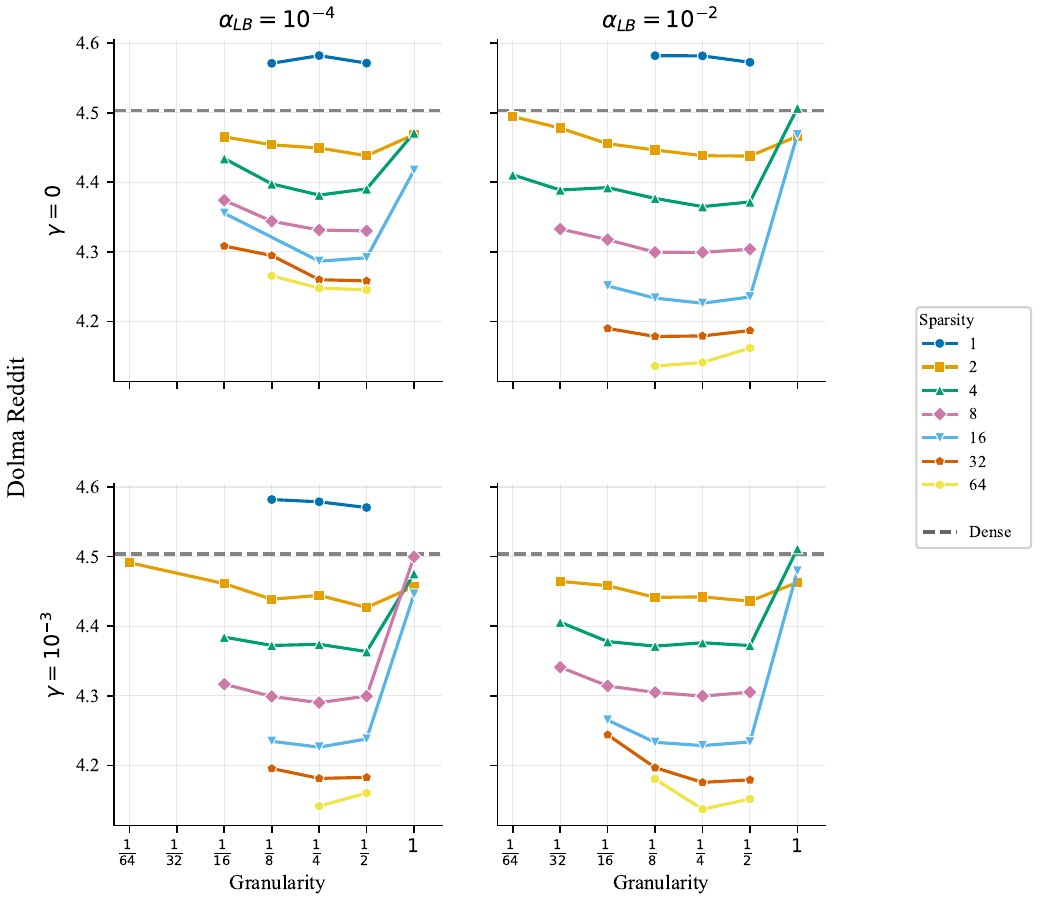}
        \hspace{1em}
        \includegraphics[width=0.46\linewidth]{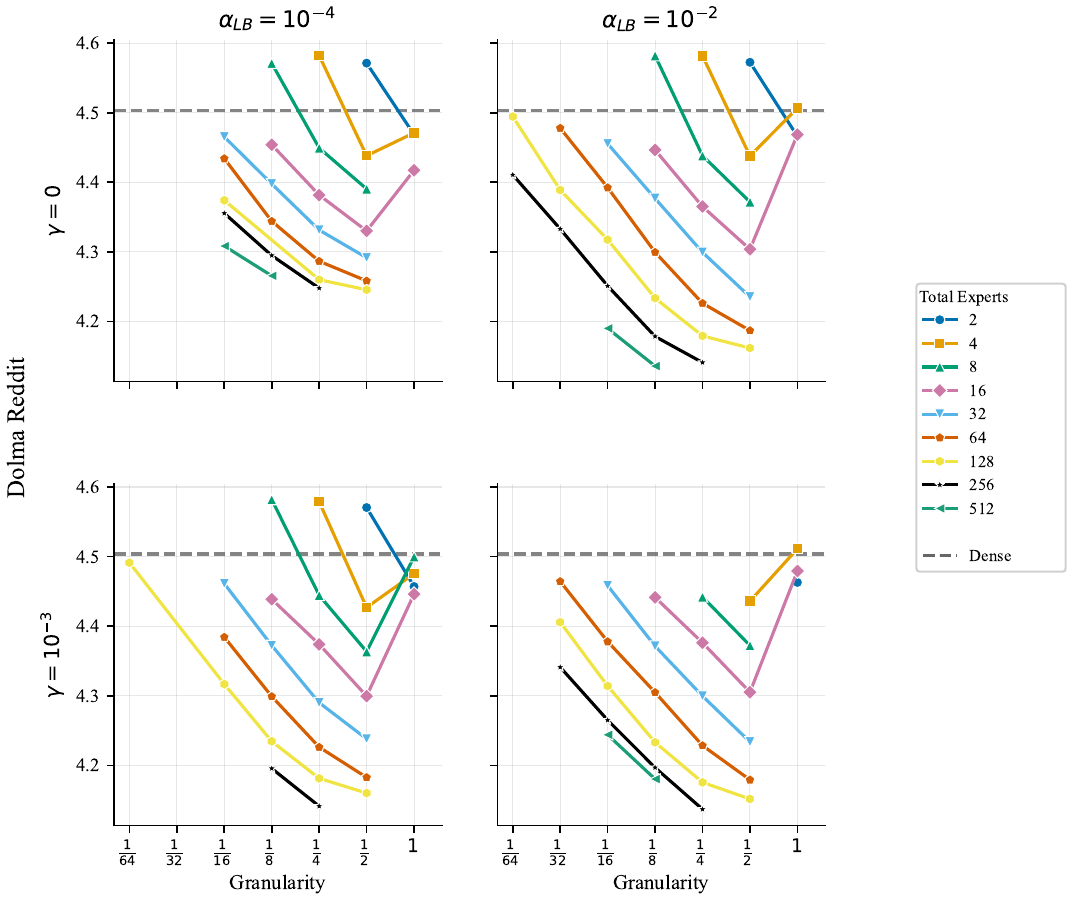}
        \caption{80M active, 80M - 765M total parameters}
    \end{subfigure}
    \par\bigskip\bigskip
    \begin{subfigure}[t]{\textwidth}
        \centering
        \includegraphics[width=0.46\linewidth]{figures/lm/dolma_reddit-validation/ce_loss/lb_sweep_hgn_gxs_110M.pdf}
        \hspace{1em}
        \includegraphics[width=0.46\linewidth]{figures/lm/dolma_reddit-validation/ce_loss/lb_sweep_hgn_gxn_110M.pdf}
        \caption{110M active, 110M - 1.4B total parameters}
    \end{subfigure}

    \end{figure*} 

\clearpage  

\begin{figure*}[ht]
    \addtocounter{figure}{-1}
    \centering
    \begin{subfigure}[t]{\textwidth}
        \centering
        \includegraphics[width=0.46\linewidth]{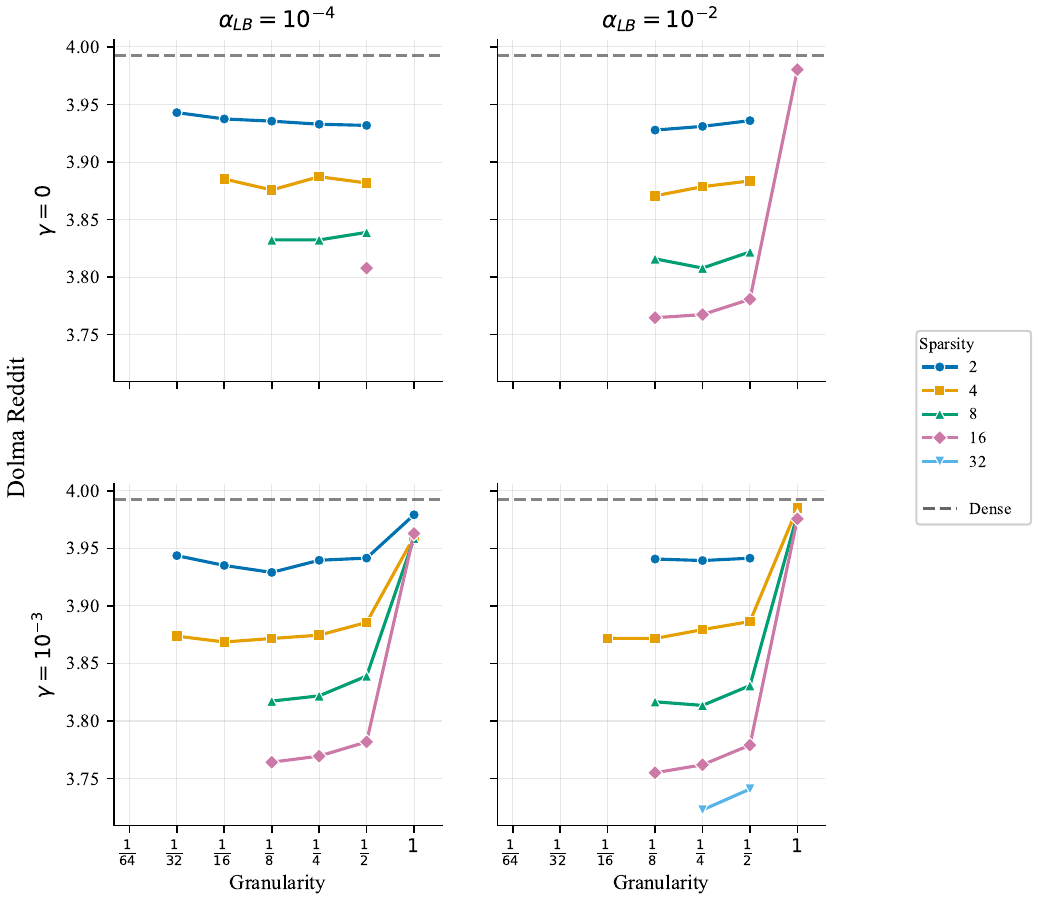}
        \hspace{1em}
        \includegraphics[width=0.46\linewidth]{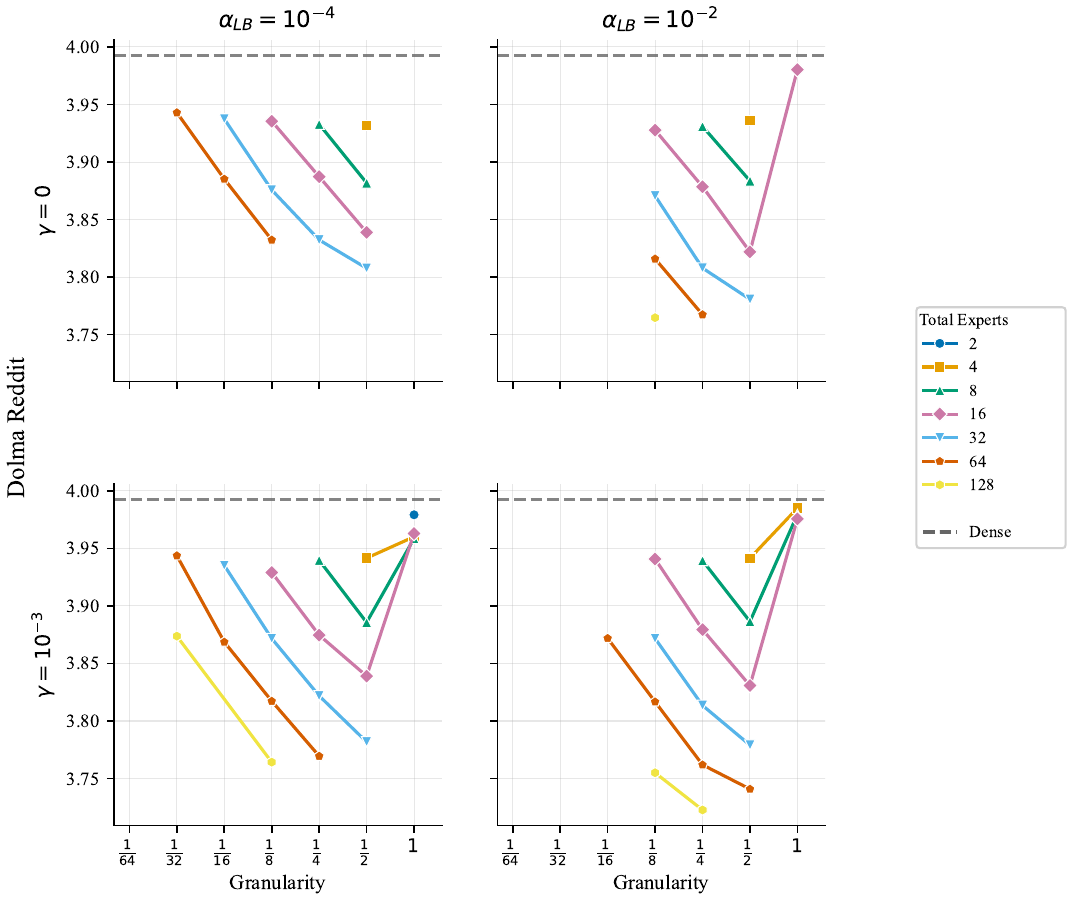}
        \caption{200M active, 200M - 3.3B total parameters}
    \end{subfigure}
    \par\bigskip\bigskip
    \begin{subfigure}[t]{\textwidth}
        \centering
        \includegraphics[width=0.3\linewidth]{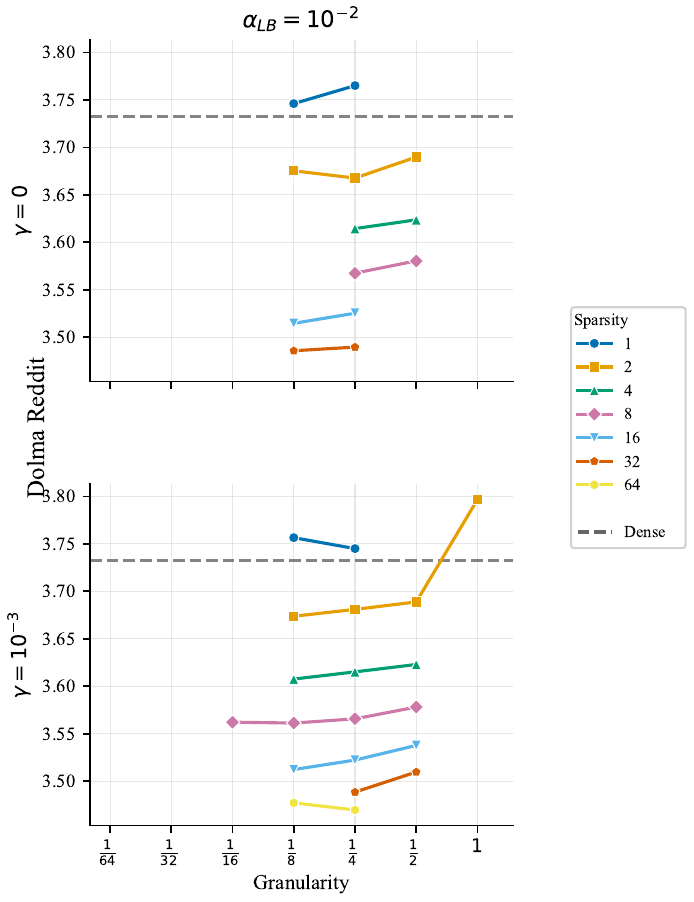}
        \hspace{1em}
        \includegraphics[width=0.3\linewidth]{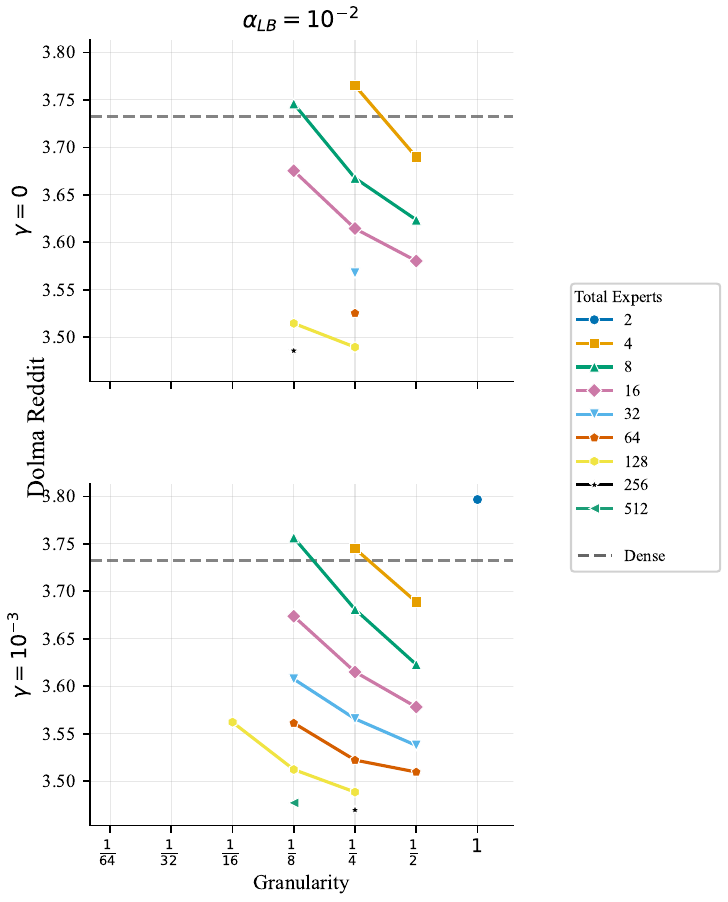}
        \caption{300M active, 300M - 6.6B total parameters}
    \end{subfigure}

    \caption{
    \textbf{Load balancing mechanisms must be tuned correctly (\S\ref{sec:expt_router}).}
    We consider load balancing loss weight $\alpha_{LB} \in \{\num{1e-2}, \num{1e-4}\}$ and loss-free load balancing with bias $\gamma\in\{0, \num{1e-3}\}$ ($\gamma=0$ indicates no loss-free mechanism). Results show that poorly chosen hyperparameters, such as high bias $\gamma = 1e-3$ with total experts $n\geq 512$, may impair performance. However, all settings other than $(\alpha_{LB}=\num{1e-2}, \gamma=\num{1e-3})$ perform comparably for $n \leq 512$, suggesting that a wide range of load balancing settings achieve near-optimal performance. 
    }
    \label{fig:dolma_reddit_lb}
\end{figure*}

%% file: fig_tex/lm/dolma_stack.tex
\begin{figure*}[!ht]
    \centering
        \begin{subfigure}[t]{\textwidth}
        \begin{subfigure}[t]{0.33\textwidth}
            \centering
            \caption*{\scriptsize Fixed total experts (n)}
            \includegraphics[width=\linewidth]{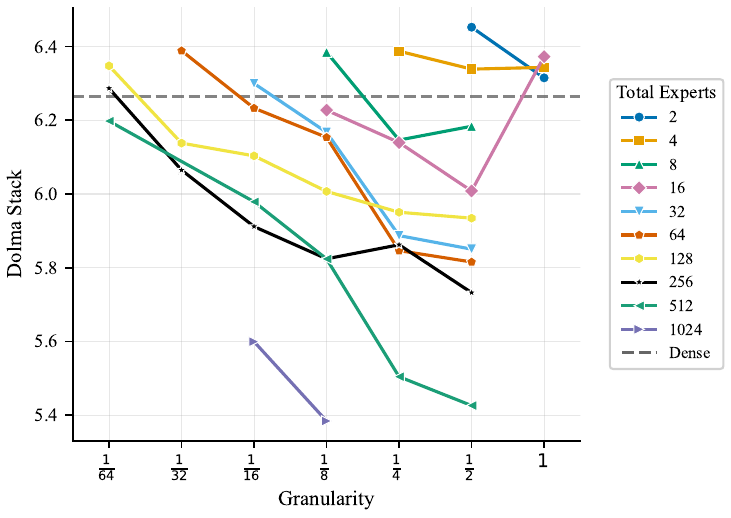}
        \end{subfigure}
        \begin{subfigure}[t]{0.33\textwidth}
            \centering
            \caption*{\scriptsize Fixed granularity (g)}
            \includegraphics[width=\linewidth]{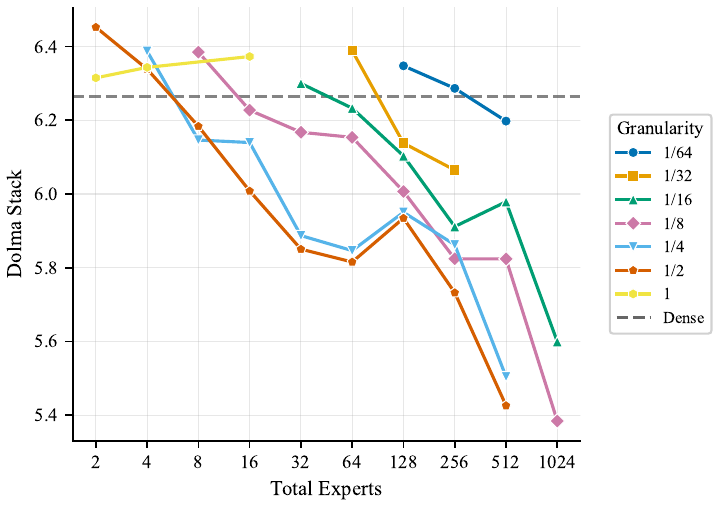}
        \end{subfigure}
        \begin{subfigure}[t]{0.33\textwidth}
            \centering
            \caption*{\scriptsize Fixed activation sparsity (s)}
            \includegraphics[width=\linewidth]{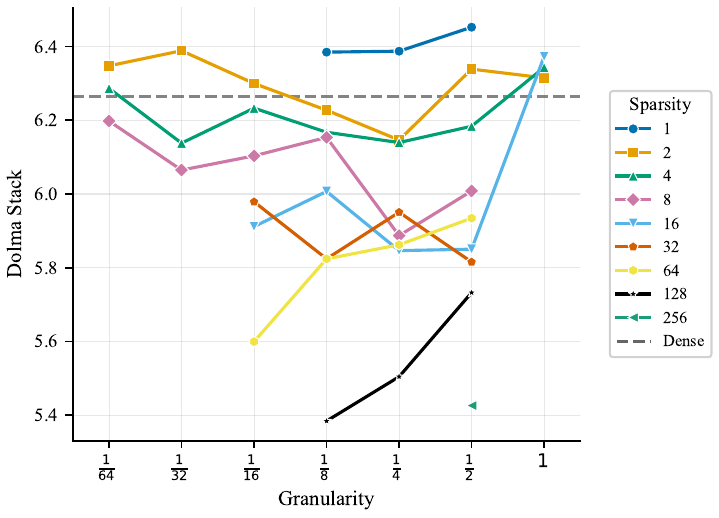}
        \end{subfigure}
        \caption{50M active, 50M - 930M total parameters}
    \end{subfigure}
\par\bigskip\bigskip
    \begin{subfigure}[t]{\textwidth}
        \begin{subfigure}[t]{0.33\textwidth}
            \centering
            \includegraphics[width=\linewidth]{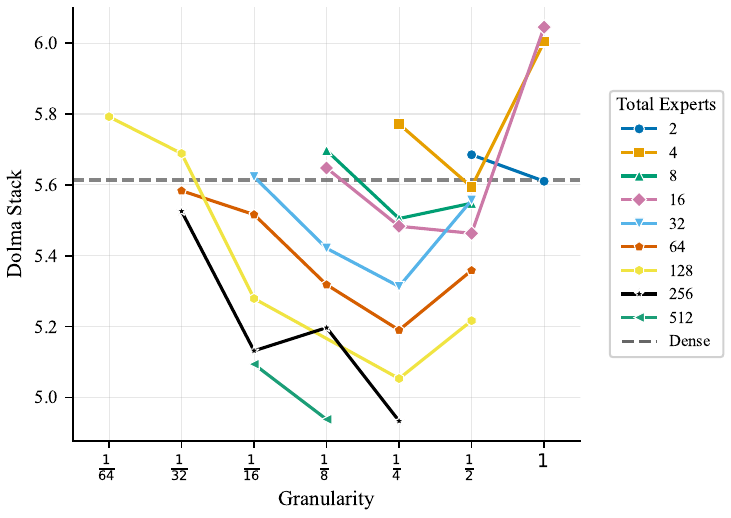}
        \end{subfigure}
        \begin{subfigure}[t]{0.33\textwidth}
            \centering
            \includegraphics[width=\linewidth]{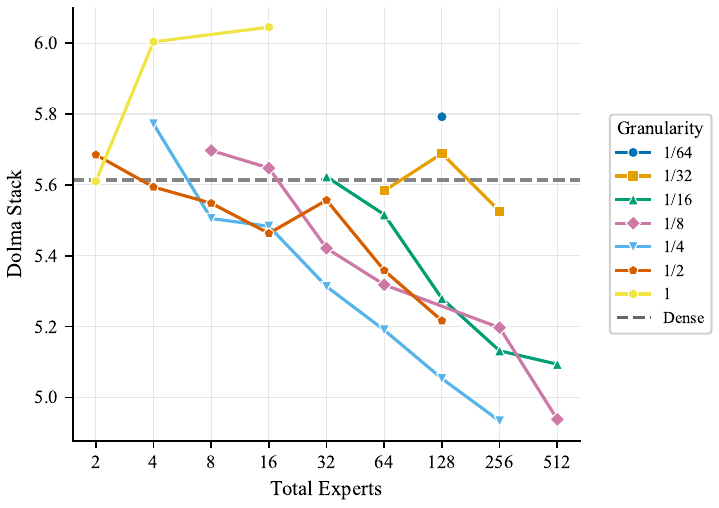}
        \end{subfigure}
        \begin{subfigure}[t]{0.33\textwidth}
            \centering
            \includegraphics[width=\linewidth]{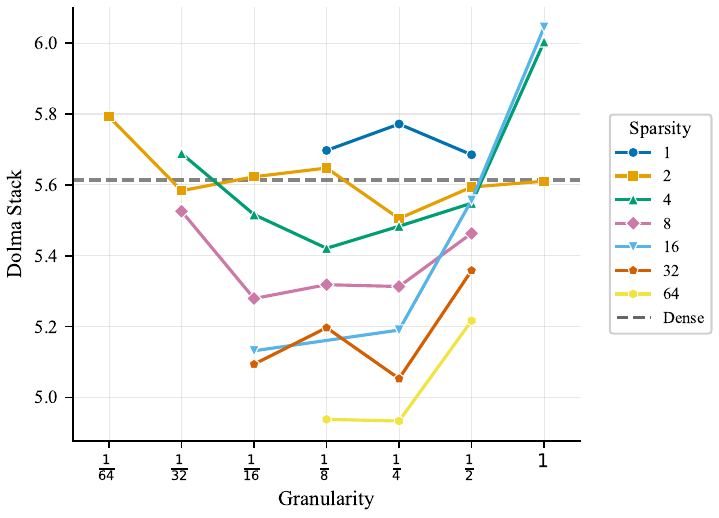}
        \end{subfigure}
        \caption{80M active, 80M - 765M total parameters}
    \end{subfigure}
    \par\bigskip\bigskip
        \begin{subfigure}[t]{\textwidth}
        \begin{subfigure}[t]{0.33\textwidth}
            \centering
            \includegraphics[width=\linewidth]{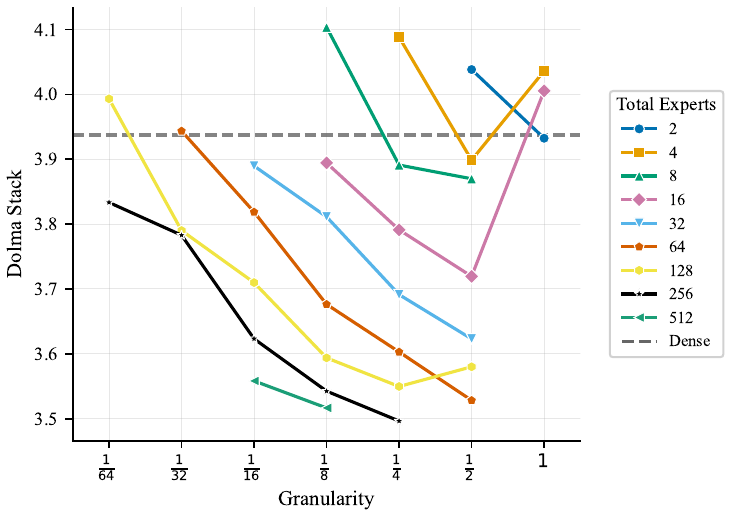}
        \end{subfigure}
        \begin{subfigure}[t]{0.33\textwidth}
            \centering
            \includegraphics[width=\linewidth]{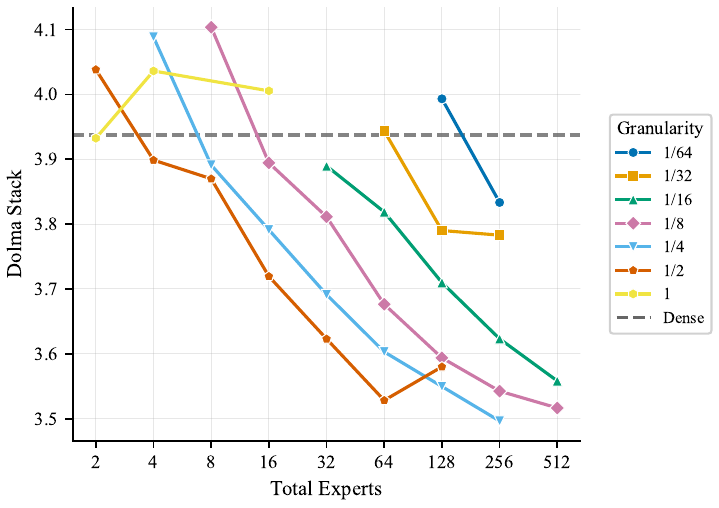}
        \end{subfigure}
        \begin{subfigure}[t]{0.33\textwidth}
            \centering
            \includegraphics[width=\linewidth]{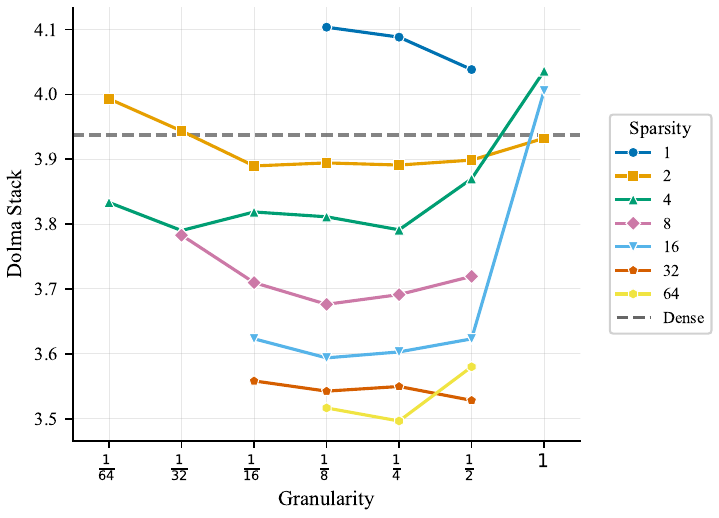}
        \end{subfigure}
        \caption{110M active, 110M - 1.4B total parameters}
    \end{subfigure}
    \end{figure*}

\clearpage  

\begin{figure*}[!ht]
        \addtocounter{figure}{-1}
    \begin{subfigure}[t]{\textwidth}
        \addtocounter{subfigure}{3}
        \begin{subfigure}[t]{0.33\textwidth}
            \centering
            \caption*{\scriptsize Fixed total experts (n)}
            \includegraphics[width=\linewidth]{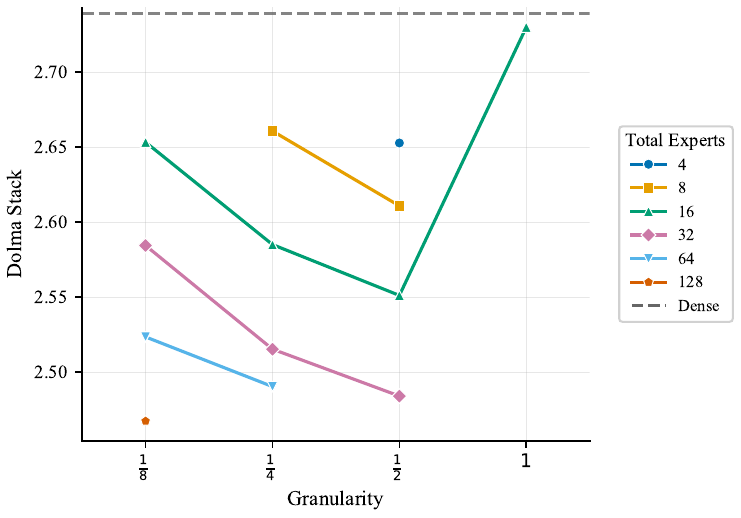}
        \end{subfigure}
        \begin{subfigure}[t]{0.33\textwidth}
            \centering
            \caption*{\scriptsize Fixed granularity (g)}
            \includegraphics[width=\linewidth]{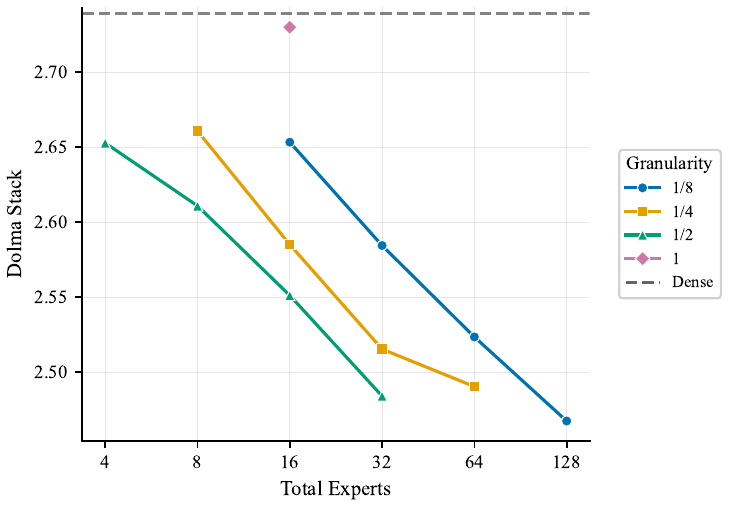}
        \end{subfigure}
        \begin{subfigure}[t]{0.33\textwidth}
            \centering
            \caption*{\scriptsize Fixed activation sparsity (s)}
            \includegraphics[width=\linewidth]{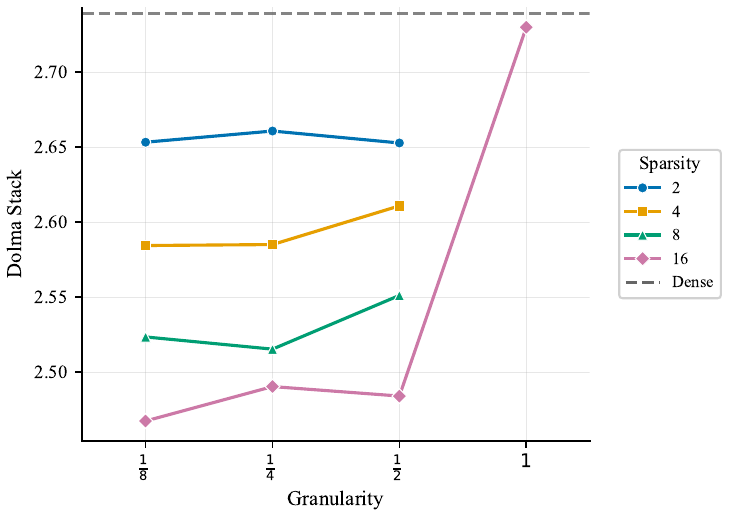}
        \end{subfigure}
        \caption{200M active, 200M - 3.3B total parameters}
    \end{subfigure}
    \par\bigskip\bigskip
        \begin{subfigure}[t]{\textwidth}
        \begin{subfigure}[t]{0.33\textwidth}
            \centering
            \includegraphics[width=\linewidth]{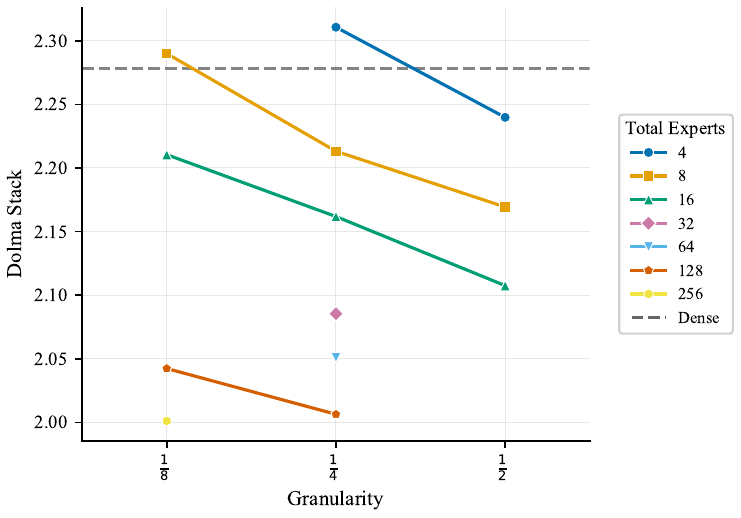}
        \end{subfigure}
        \begin{subfigure}[t]{0.33\textwidth}
            \centering
            \includegraphics[width=\linewidth]{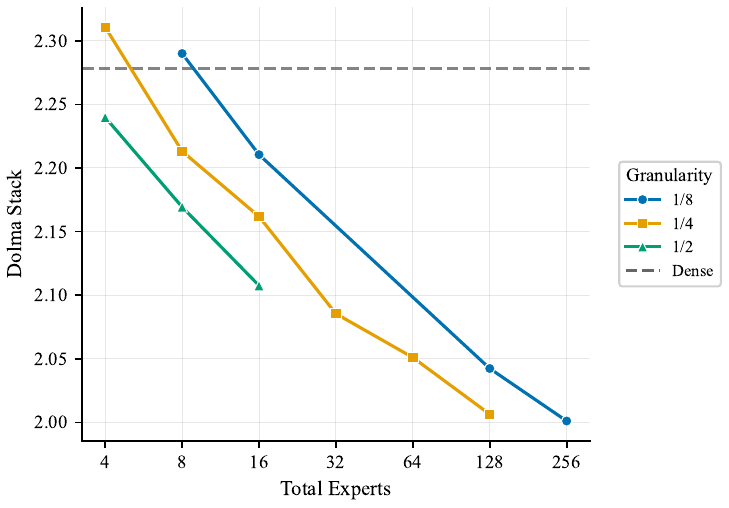}
        \end{subfigure}
        \begin{subfigure}[t]{0.33\textwidth}
            \centering
            \includegraphics[width=\linewidth]{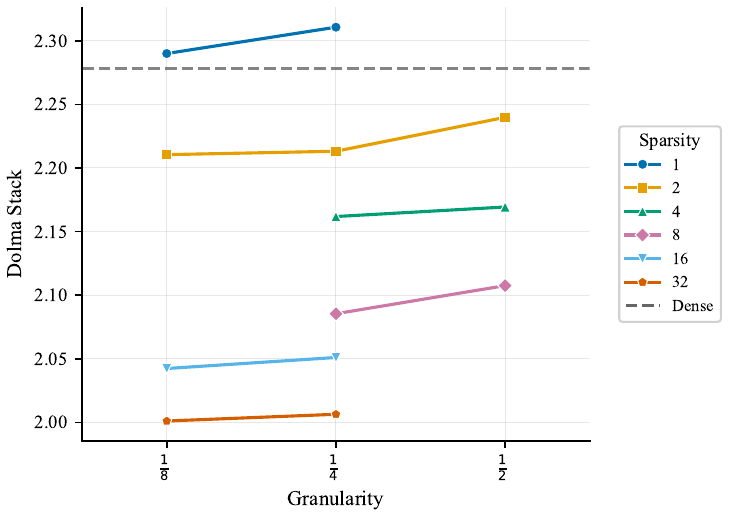}
        \end{subfigure}
        \caption{300M active, 300M - 6.6B total parameters}
    \end{subfigure}

    \caption{
    \textbf{Increasing inactive expert parameters via expert size (left) or total count (center) improves performance in MoEs (\S\ref{sec:expt_main}).} This effect is seen both when holding total number of experts fixed (left) and when holding expert granularity fixed (center). In general, increasing total parameters results in improved performance.  \textbf{Optimal tradeoff between expert count and granularity varies in MoEs (right). (\S\ref{sec:expt_main})}
    At each activation sparsity $s$ (equivalently, at each total parameter count), the optimal (total expert count, expert granularity) configuration varies. As $s$ increases, optimal expert granularity remains nearly fixed, suggesting that sparsity should be scaled up primarily by increasing total expert count $n$, while maintaining a near constant, slowly increasing expert granularity $g$. 
    }
    \label{fig:dolma_stack_experts}
\end{figure*}

\begin{figure*}[!ht]
    \centering
    
    \begin{subfigure}[t]{0.46\textwidth}
        \centering
        \includegraphics[width=\linewidth]{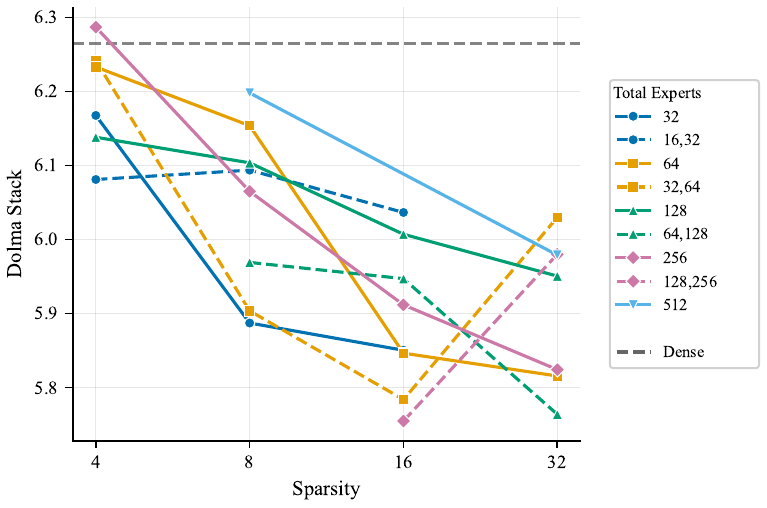}
        \caption{50M active, 50M - 930M total parameters}
    \end{subfigure}
    \vspace{1em}
    \begin{subfigure}[t]{0.46\textwidth}
        \centering
        \includegraphics[width=\linewidth]{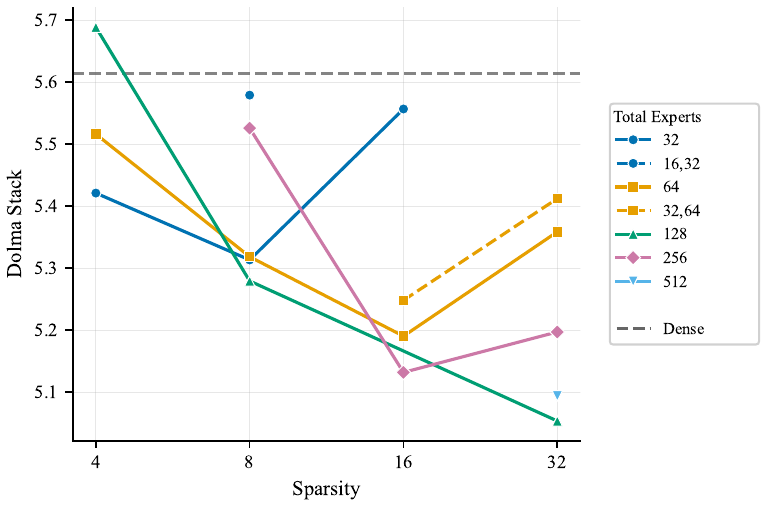}
        \caption{80M active, 80M - 765M total parameters}
    \end{subfigure}
    \caption{
    \textbf{Heterogeneity of expert size alone does not improve MoE performance (\S\ref{sec:expt_hetgen}).} To explore the potential benefits of their architectural flexibility, we compare heterogeneous MoEs (indicated by dotted lines) to active- and total-parameter-matched homogeneous MoEs. Heterogeneity alone does not result in performance gains, as, at each activation sparsity $s$, heterogeneous MoEs with $n_1, n_2 = a, b$ lie between or near the 2 closest homogeneous MoEs, with $n=a$ and with $n=b$.
    }
    \label{fig:dolma_stack_het}
\end{figure*}

\begin{figure*}[!ht]
    \centering
    
    \begin{subfigure}[t]{1.0\textwidth}
        \centering
        \includegraphics[width=\linewidth]{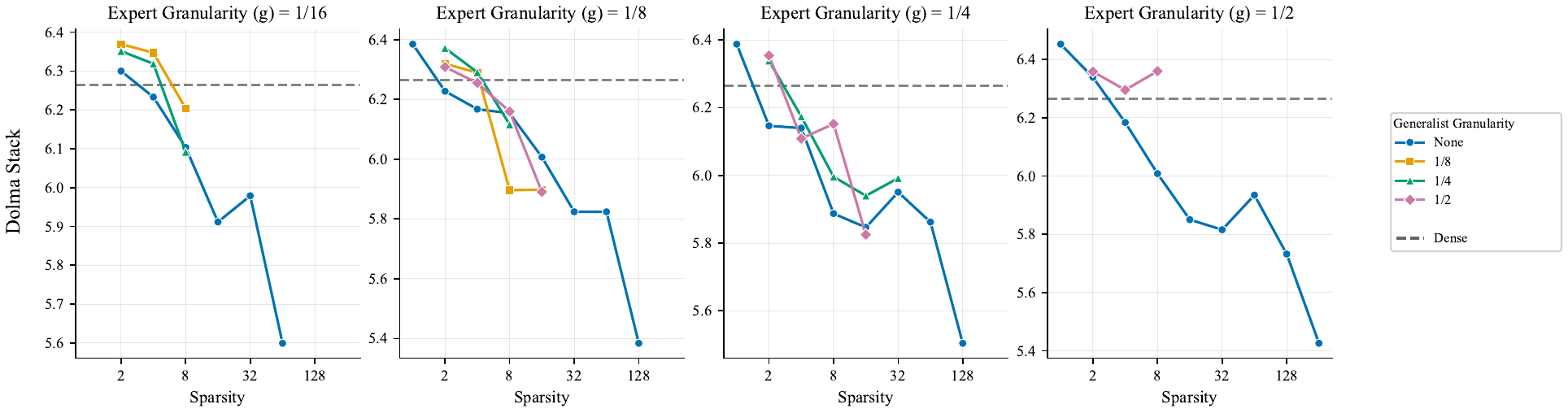}
        \caption{50M active, 50M - 930M total parameters}
    \end{subfigure}
    \par\bigskip\bigskip
    \begin{subfigure}[t]{1.0\textwidth}
        \centering
        \includegraphics[width=\linewidth]{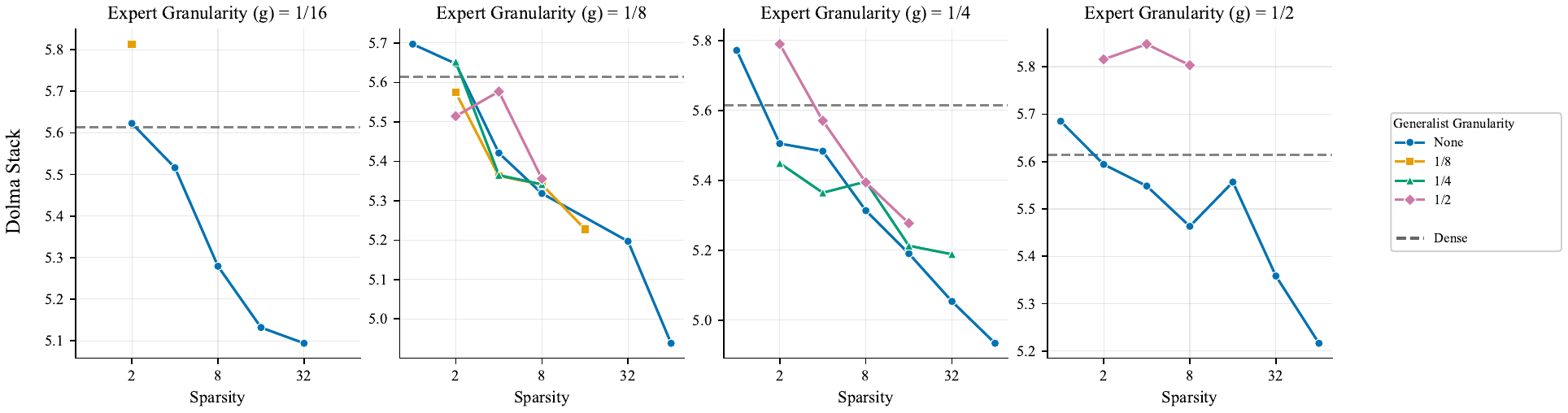}
        \caption{80M active, 80M - 765M total parameters}
    \end{subfigure}
    \par\bigskip\bigskip
    \begin{subfigure}[t]{1.0\textwidth}
        \centering
        \includegraphics[width=\linewidth]{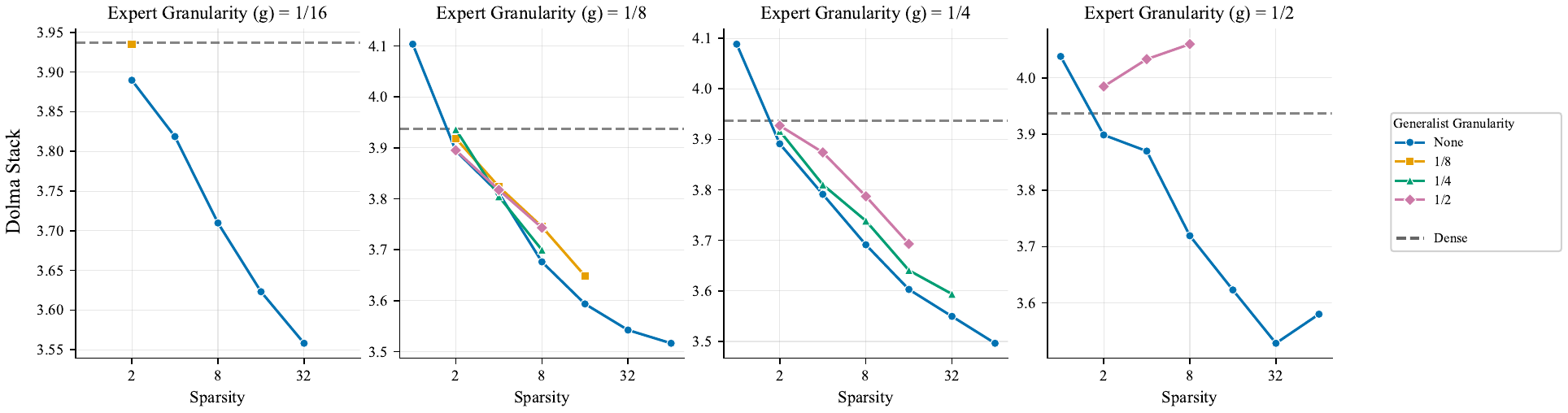}
        \caption{110M active, 110M - 1.4B total parameters}
    \end{subfigure}
    \caption{
    \textbf{The inclusion of a generalist consistently degrades performance in homogeneous MoEs (\S\ref{sec:expt_hetgen}).}
    We train MoE LMs which consist of some routed experts with granularity $g$, as well as a generalist with granularity $g_{gen}\in \{\frac{1}{2}, \frac{1}{4}, \frac{1}{8}\} $. We compare to settings with no generalist, only routed experts with granularity $g$. In all settings and configurations, the addition of any granularity generalist results in comparable or degraded performance. 
    }
    \label{fig:dolma_stack_gen}
\end{figure*}

\begin{figure*}[ht]
    \centering
    \begin{subfigure}[t]{1.0\textwidth}
        \centering
        \includegraphics[width=\linewidth]{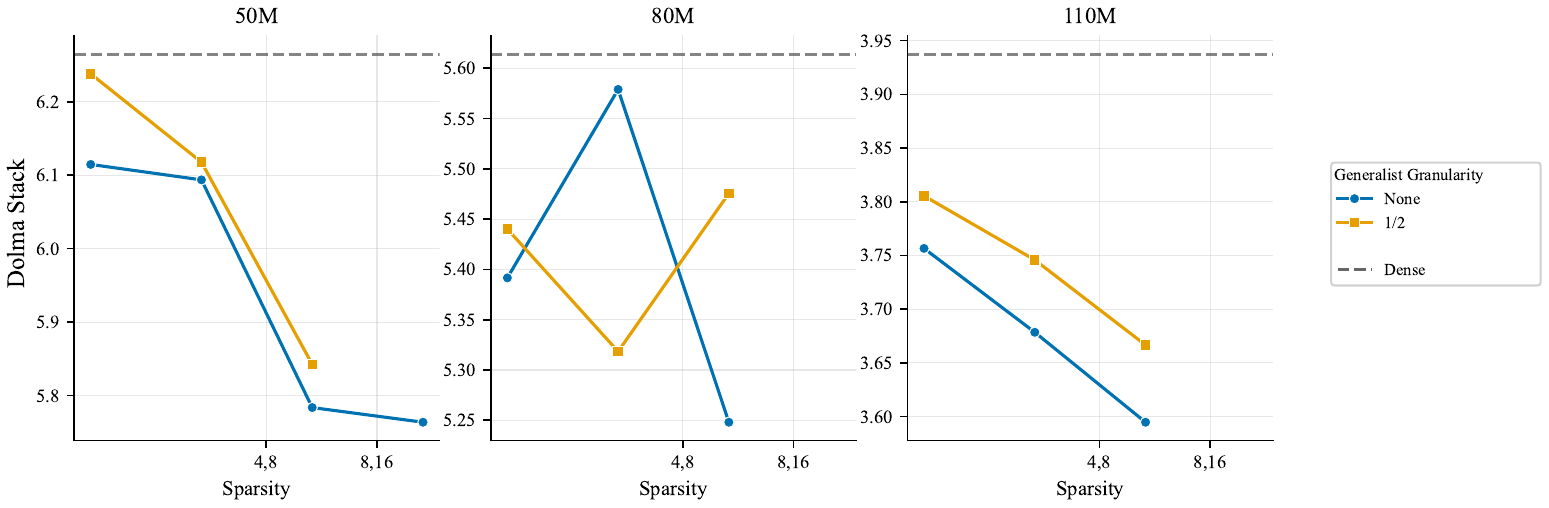}
    \end{subfigure}
    \caption{
    \textbf{The inclusion of a generalist consistently degrades performance in heterogeneous MoEs (\S\ref{sec:expt_hetgen}).}
    We train heterogeneous MoE LMs which consist of  routed experts with granularity $g_1, g_2$, as well as a generalist with granularity $g_{gen} = \frac{1}{2}$. We compare to settings with no generalist. In all settings and configurations, the addition of a generalist results in comparable or degraded performance. 
    }
    \label{fig:dolma_stack_hetgen}
\end{figure*}

\begin{figure*}[ht]
    \centering
    \begin{subfigure}[t]{\textwidth}
        \centering
        \begin{subfigure}[t]{0.45\textwidth}
            \includegraphics[width=\linewidth]{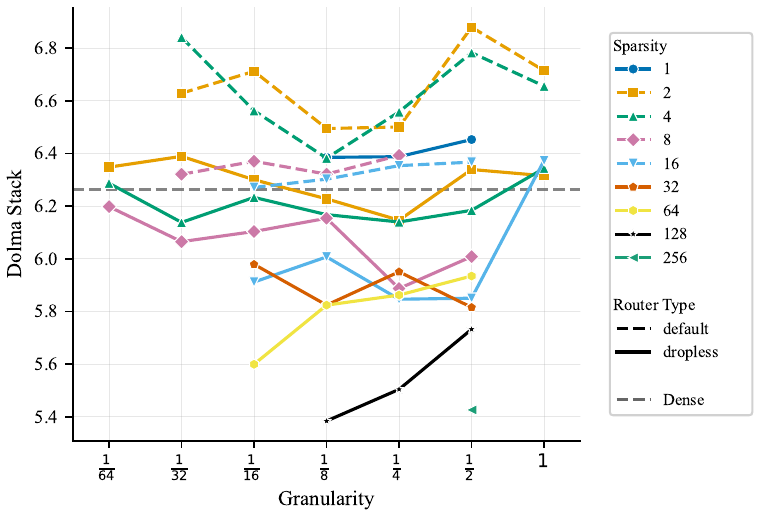}
            \caption{50M active, 50M - 930M total parameters}
        \end{subfigure}
    \hspace{1em}
        \begin{subfigure}[t]{0.45\textwidth}
            \centering
            \includegraphics[width=\linewidth]{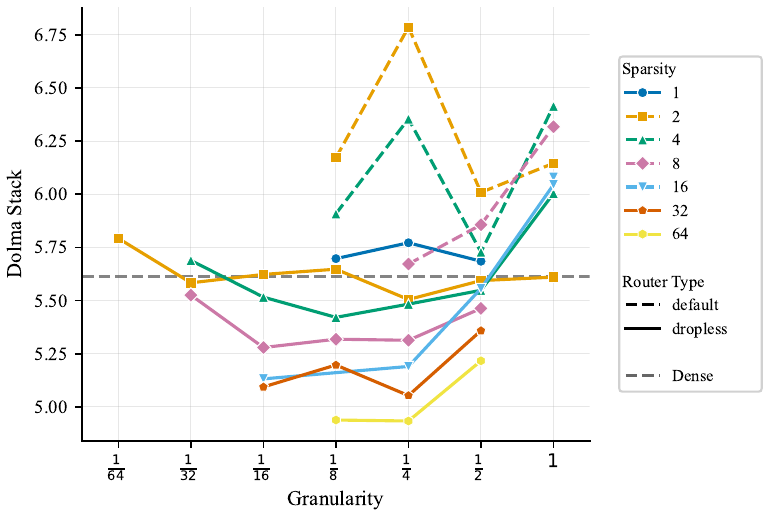}
            \caption{80M active, 80M - 765M total parameters}
        \end{subfigure}
    \end{subfigure}

    \par\bigskip\bigskip
    \begin{subfigure}[t]{0.45\textwidth}
        \centering
        \includegraphics[width=\linewidth]{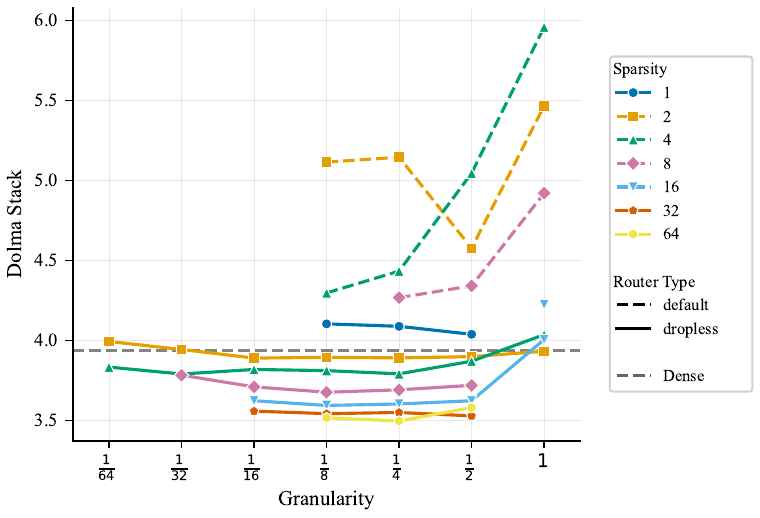}
        \caption{110M active, 110M - 1.4B total parameters}
    \end{subfigure}
    \caption{ 
    \textbf{Dropless routing outperforms default routing (\S\ref{sec:expt_router}).}
    We compare dropless routing to the default setting, which allow tokens to be dropped. Across all scales, we find that dropless routing outperforms or performs comparably to default routing. 
    }
    \label{fig:dolma_stack_dropless}
\end{figure*}

\begin{figure*}[ht]
    \centering
    \begin{subfigure}[t]{0.45\textwidth}
        \centering
        \includegraphics[width=\linewidth]{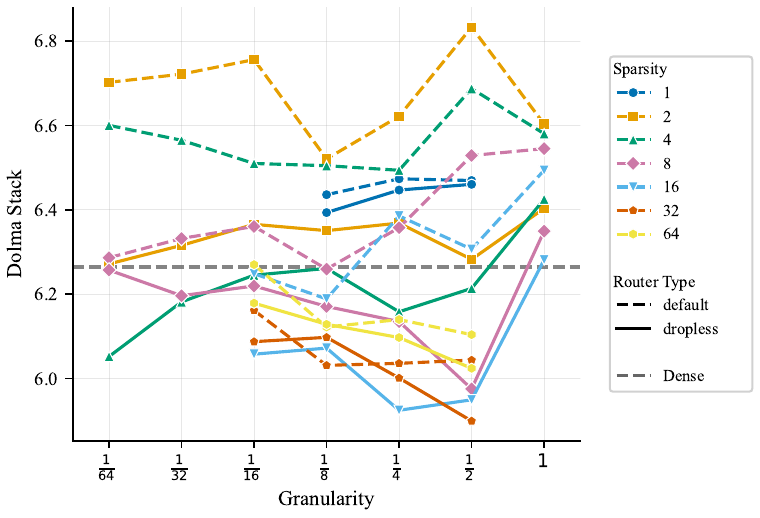}
        \caption{50M active, 50M - 930M total parameters}
    \end{subfigure}
    \hspace{1em}
    \begin{subfigure}[t]{0.45\textwidth}
        \centering
        \includegraphics[width=\linewidth]{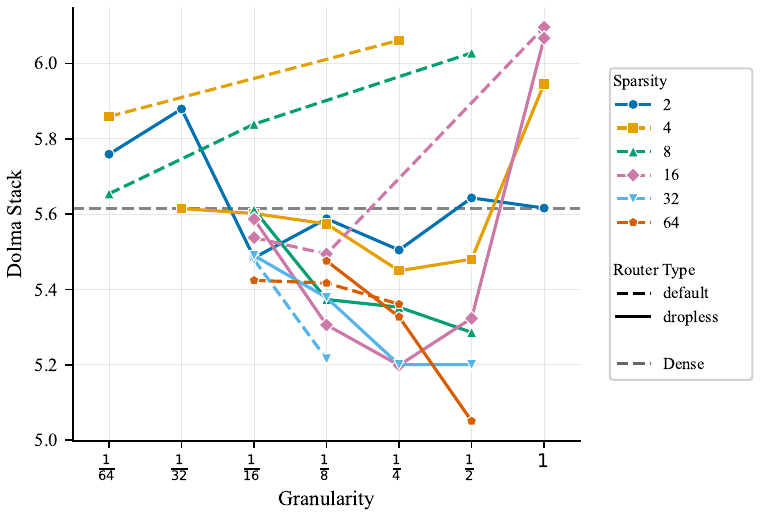}
        \caption{80M active, 80M - 765M total parameters}
    \end{subfigure}
    \caption{
    \textbf{Dropless routing, with bias $\gamma=\num{1e-3}$ (\S\ref{sec:expt_router}).} 
    As in Figure~\ref{fig:lm_avg_dropless}, we compare dropless routing to the default setting, which allow tokens to be dropped. Across all scales, we find that dropless routing outperforms or performs comparably to default routing. We see here with additional higher sparsity default routing runs that as sparsity increases, default routing performance approaches that of dropless routing.
    }
    \label{fig:dolma_stack_dropless_with_lf}
\end{figure*}

\begin{figure*}[ht]
    \centering
    \begin{subfigure}[]{\textwidth}
        \centering
        \includegraphics[width=0.46\linewidth]{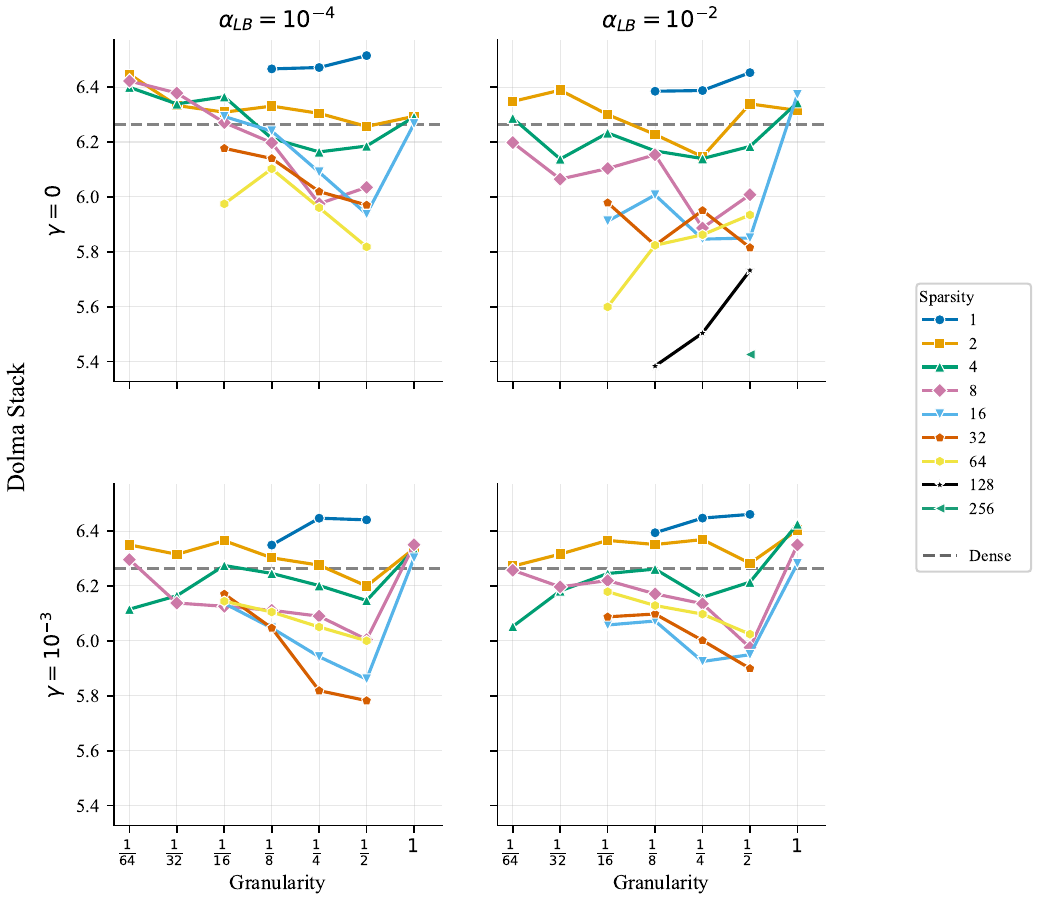}
        \hspace{1em}
        \includegraphics[width=0.46\linewidth]{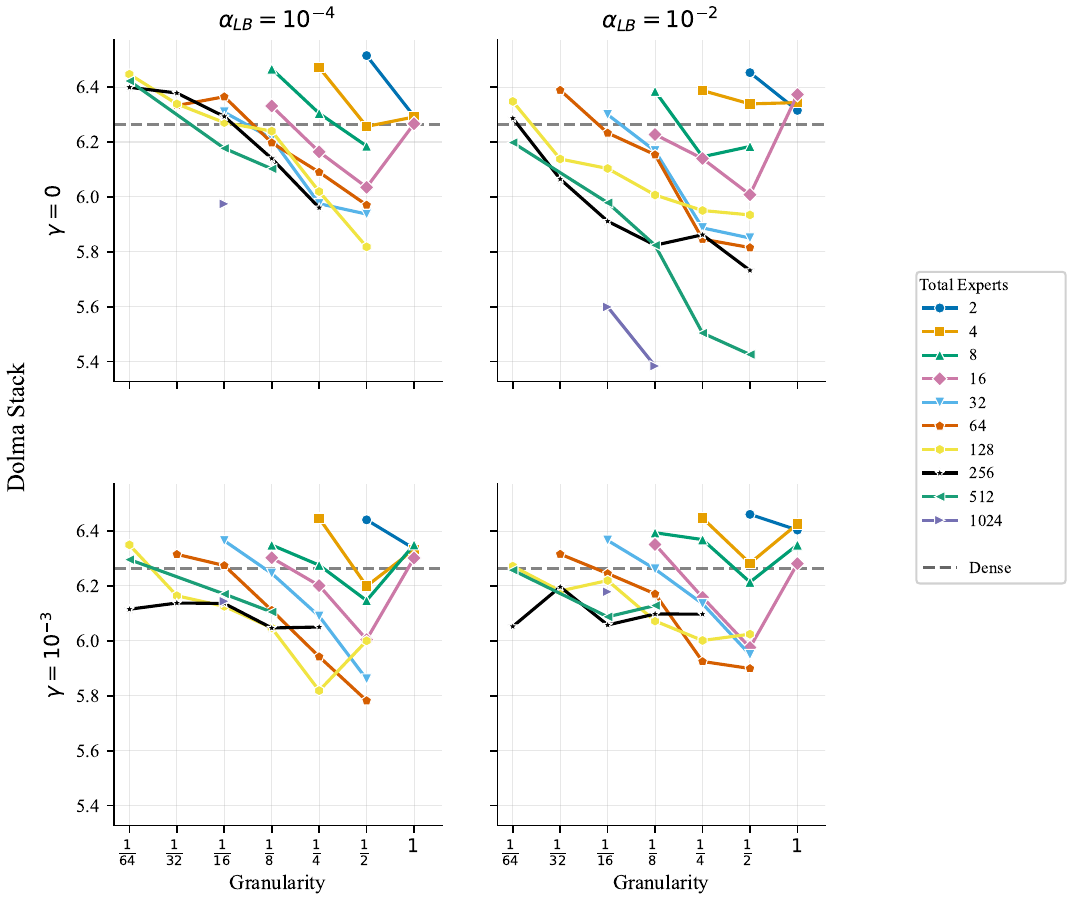}
        \caption{50M active, 50M - 930M total parameters}
    \end{subfigure}
    \par\bigskip\bigskip
    \begin{subfigure}[]{\textwidth}
        \centering
        \includegraphics[width=0.46\linewidth]{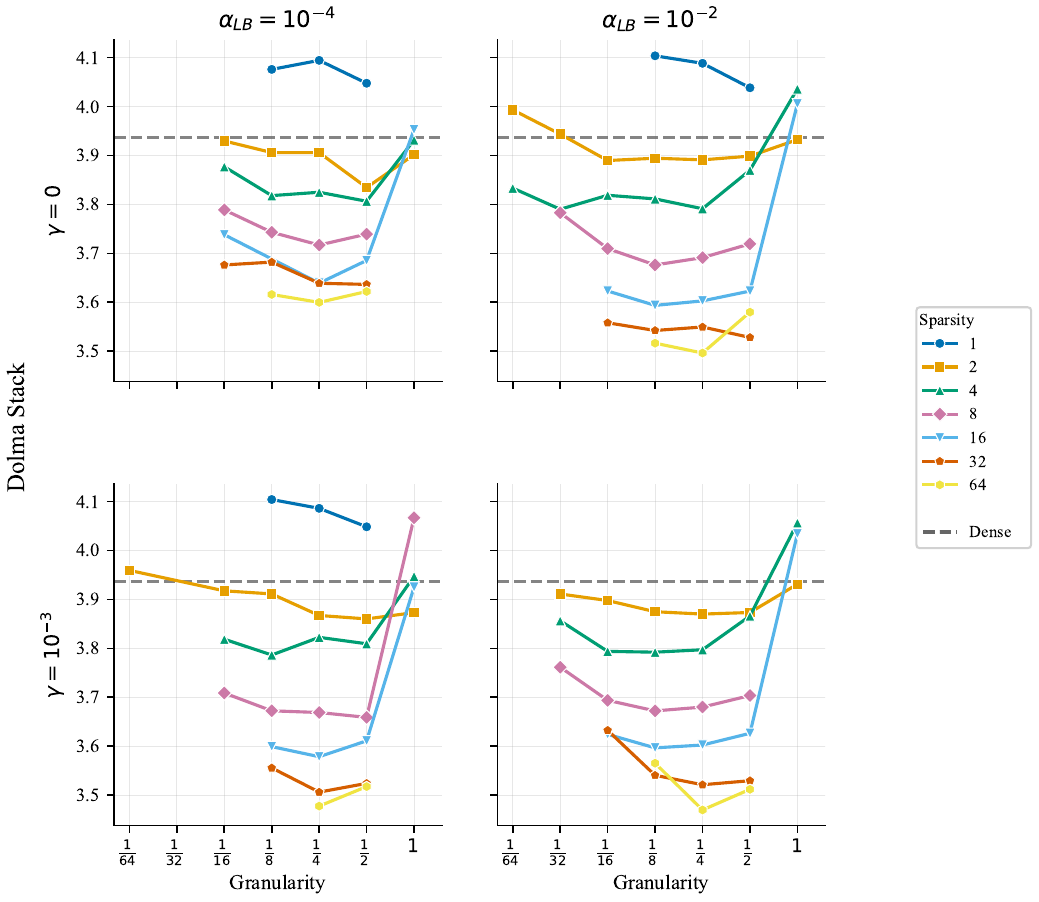}
        \hspace{1em}
        \includegraphics[width=0.46\linewidth]{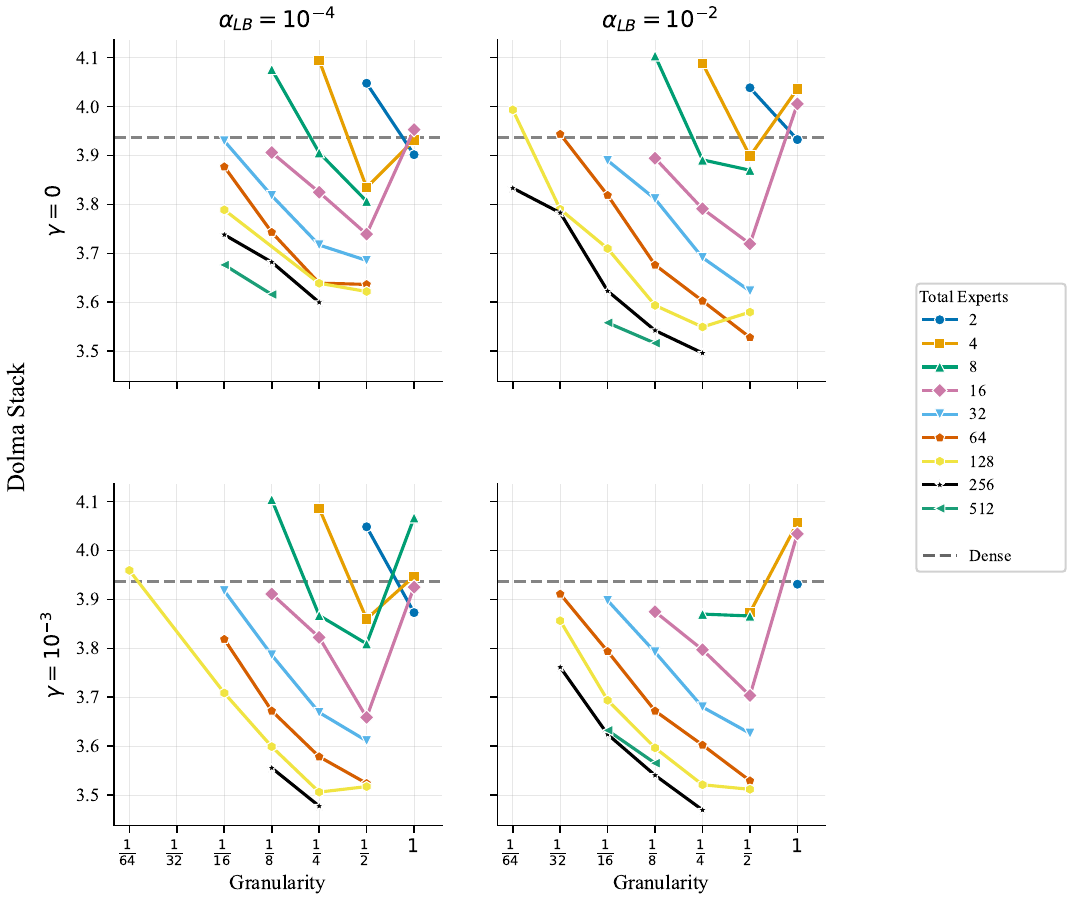}
        \caption{80M active, 80M - 765M total parameters}
    \end{subfigure}
    \par\bigskip\bigskip
    \begin{subfigure}[t]{\textwidth}
        \centering
        \includegraphics[width=0.46\linewidth]{figures/lm/dolma_stack-validation/ce_loss/lb_sweep_hgn_gxs_110M.pdf}
        \hspace{1em}
        \includegraphics[width=0.46\linewidth]{figures/lm/dolma_stack-validation/ce_loss/lb_sweep_hgn_gxn_110M.pdf}
        \caption{110M active, 110M - 1.4B total parameters}
    \end{subfigure}

    \end{figure*} 

\clearpage  

\begin{figure*}[ht]
    \addtocounter{figure}{-1}
    \centering
    \begin{subfigure}[t]{\textwidth}
        \centering
        \includegraphics[width=0.46\linewidth]{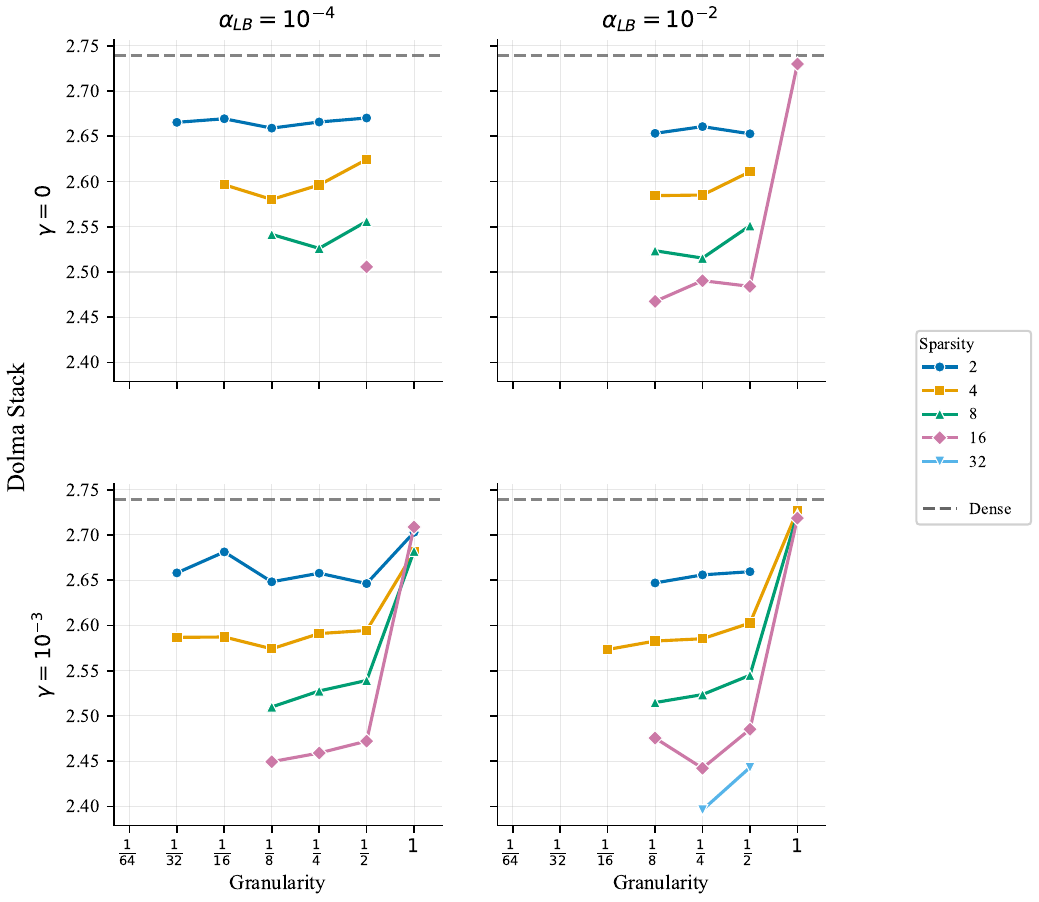}
        \hspace{1em}
        \includegraphics[width=0.46\linewidth]{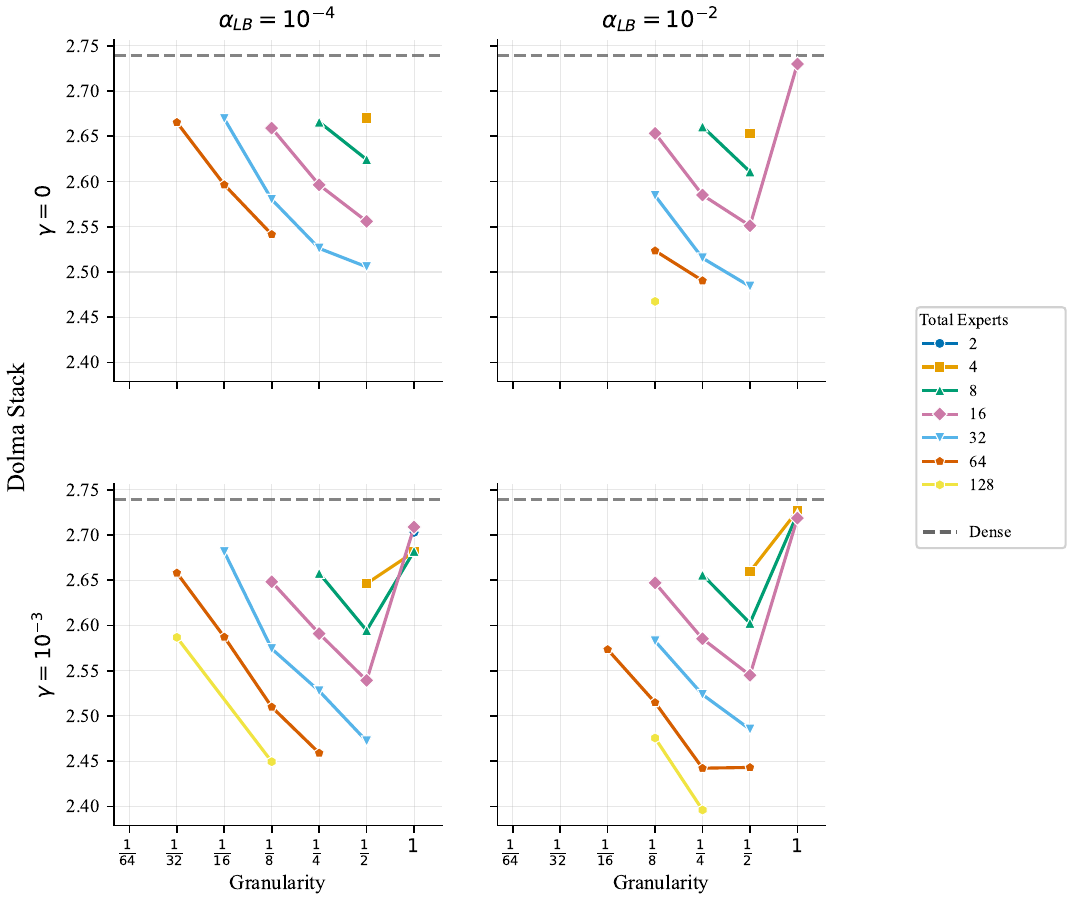}
        \caption{200M active, 200M - 3.3B total parameters}
    \end{subfigure}
    \par\bigskip\bigskip
    \begin{subfigure}[t]{\textwidth}
        \centering
        \includegraphics[width=0.3\linewidth]{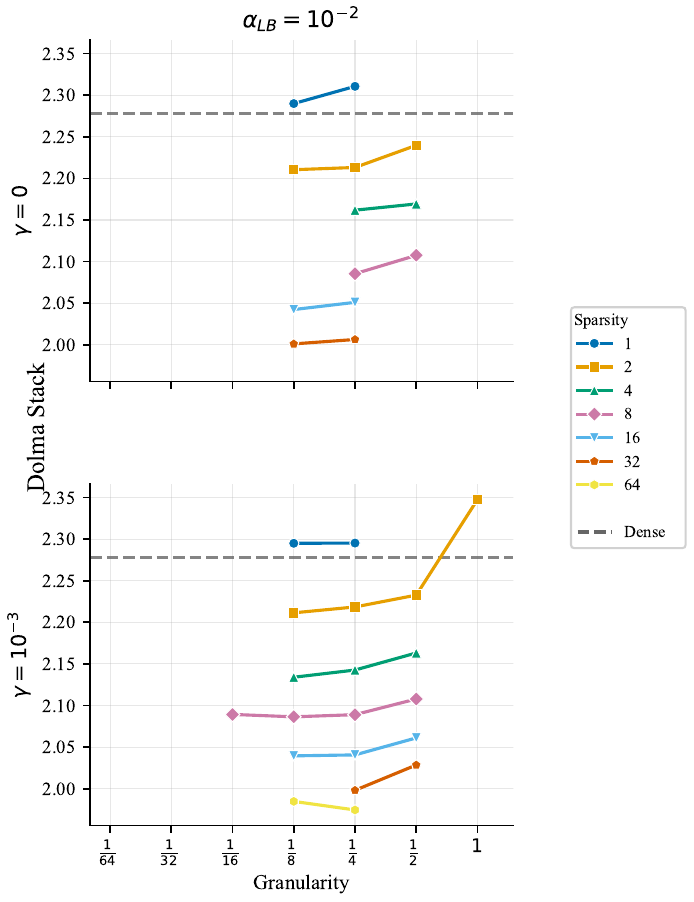}
        \hspace{1em}
        \includegraphics[width=0.3\linewidth]{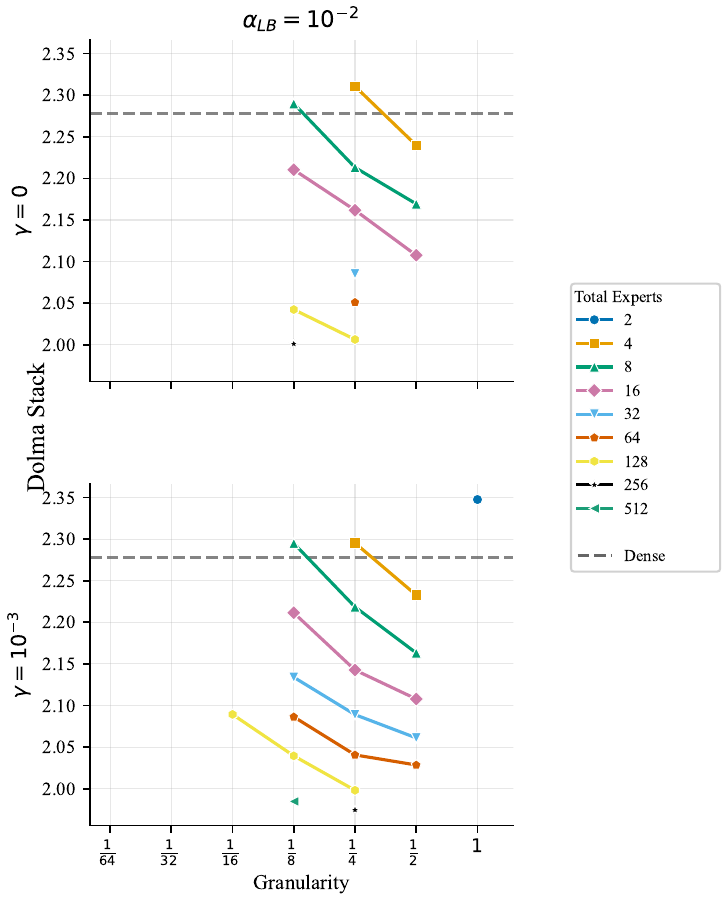}
        \caption{300M active, 300M - 6.6B total parameters}
    \end{subfigure}

    \caption{
    \textbf{Load balancing mechanisms must be tuned correctly (\S\ref{sec:expt_router}).}
    We consider load balancing loss weight $\alpha_{LB} \in \{\num{1e-2}, \num{1e-4}\}$ and loss-free load balancing with bias $\gamma\in\{0, \num{1e-3}\}$ ($\gamma=0$ indicates no loss-free mechanism). Results show that poorly chosen hyperparameters, such as high bias $\gamma = 1e-3$ with total experts $n\geq 512$, may impair performance. However, all settings other than $(\alpha_{LB}=\num{1e-2}, \gamma=\num{1e-3})$ perform comparably for $n \leq 512$, suggesting that a wide range of load balancing settings achieve near-optimal performance. 
    }
    \label{fig:dolma_stack_lb}
\end{figure*}

%% file: fig_tex/lm/dolma_wiki.tex
\begin{figure*}[!ht]
    \centering
        \begin{subfigure}[t]{\textwidth}
        \begin{subfigure}[t]{0.33\textwidth}
            \centering
            \caption*{\scriptsize Fixed total experts (n)}
            \includegraphics[width=\linewidth]{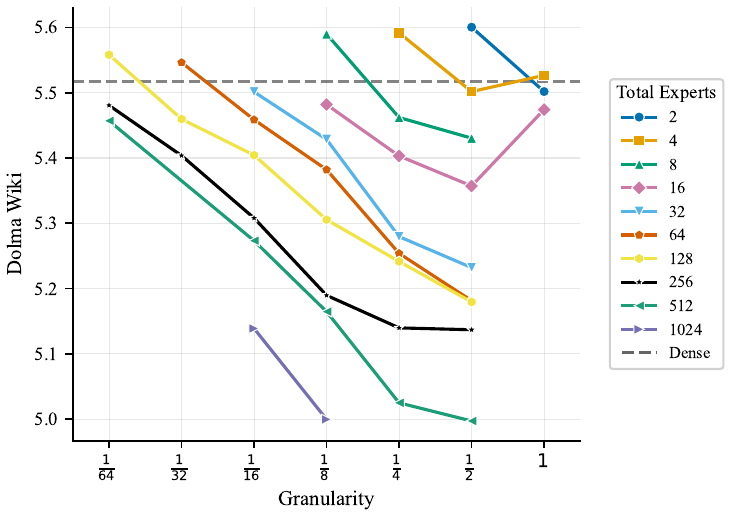}
        \end{subfigure}
        \begin{subfigure}[t]{0.33\textwidth}
            \centering
            \caption*{\scriptsize Fixed granularity (g)}
            \includegraphics[width=\linewidth]{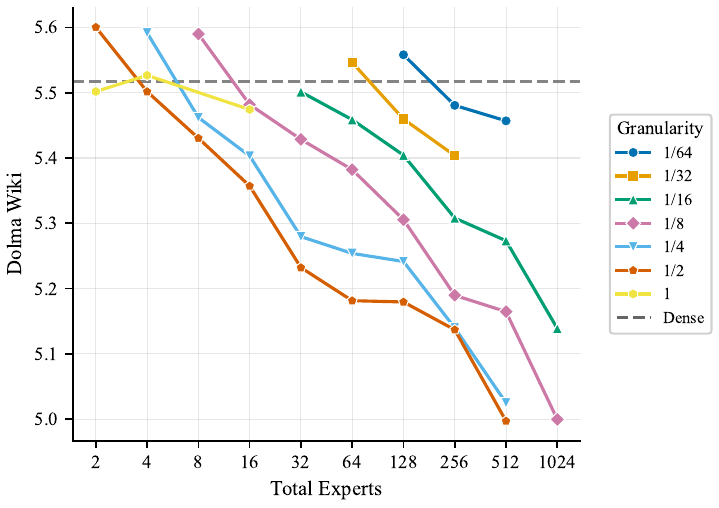}
        \end{subfigure}
        \begin{subfigure}[t]{0.33\textwidth}
            \centering
            \caption*{\scriptsize Fixed activation sparsity (s)}
            \includegraphics[width=\linewidth]{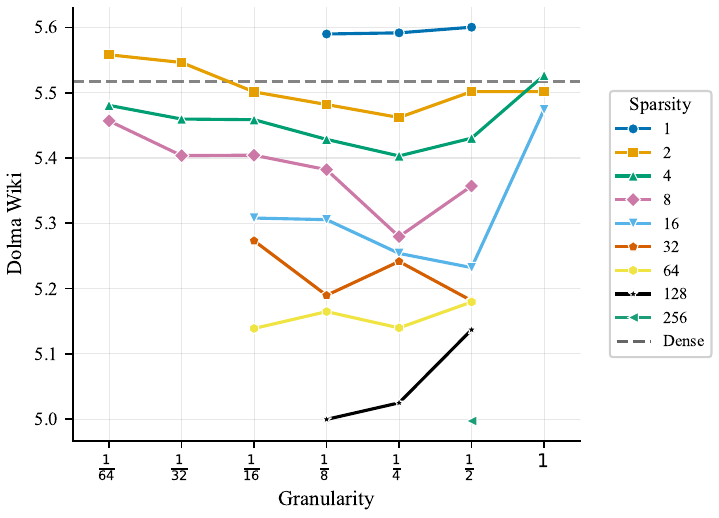}
        \end{subfigure}
        \caption{50M active, 50M - 930M total parameters}
    \end{subfigure}
\par\bigskip\bigskip
    \begin{subfigure}[t]{\textwidth}
        \begin{subfigure}[t]{0.33\textwidth}
            \centering
            \includegraphics[width=\linewidth]{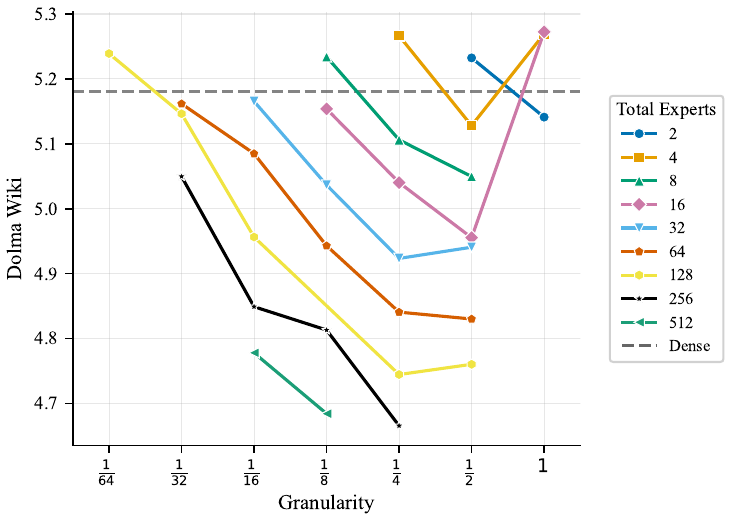}
        \end{subfigure}
        \begin{subfigure}[t]{0.33\textwidth}
            \centering
            \includegraphics[width=\linewidth]{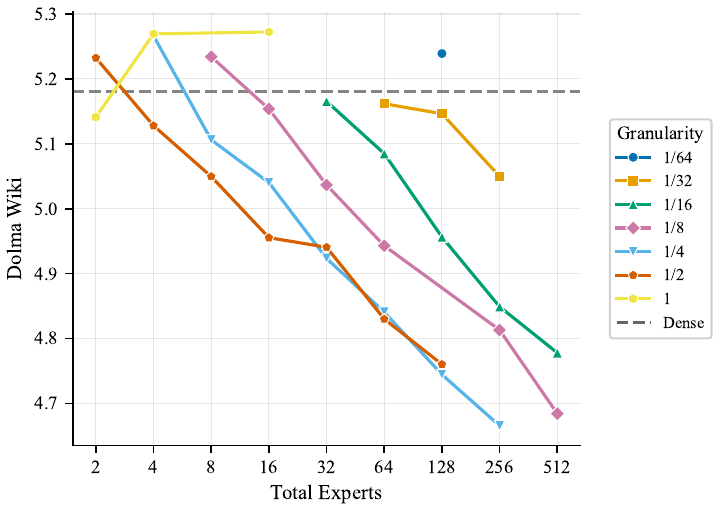}
        \end{subfigure}
        \begin{subfigure}[t]{0.33\textwidth}
            \centering
            \includegraphics[width=\linewidth]{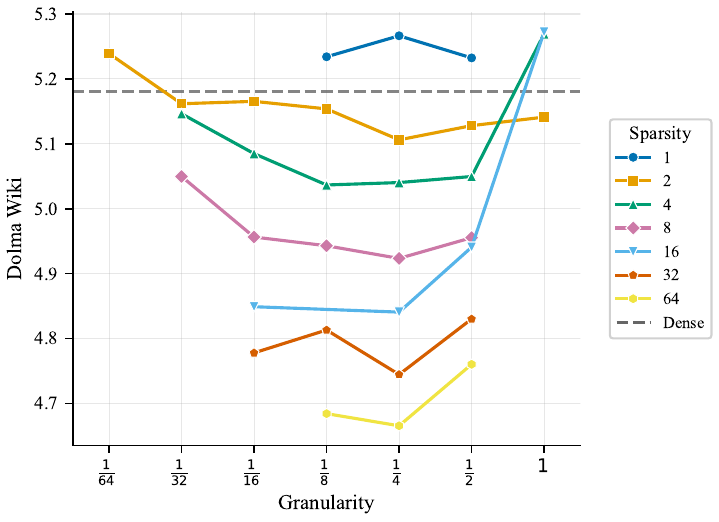}
        \end{subfigure}
        \caption{80M active, 80M - 765M total parameters}
    \end{subfigure}
    \par\bigskip\bigskip
        \begin{subfigure}[t]{\textwidth}
        \begin{subfigure}[t]{0.33\textwidth}
            \centering
            \includegraphics[width=\linewidth]{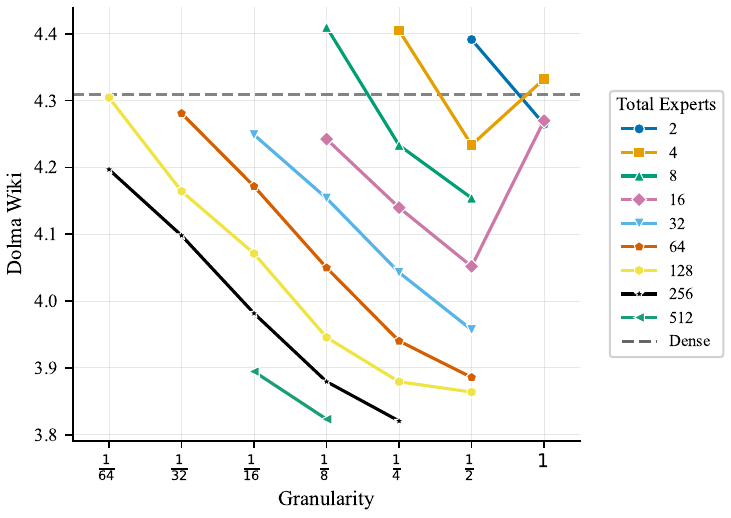}
        \end{subfigure}
        \begin{subfigure}[t]{0.33\textwidth}
            \centering
            \includegraphics[width=\linewidth]{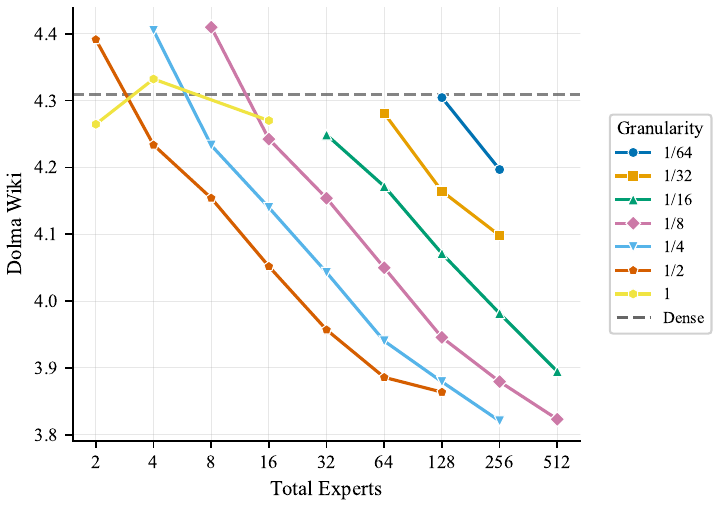}
        \end{subfigure}
        \begin{subfigure}[t]{0.33\textwidth}
            \centering
            \includegraphics[width=\linewidth]{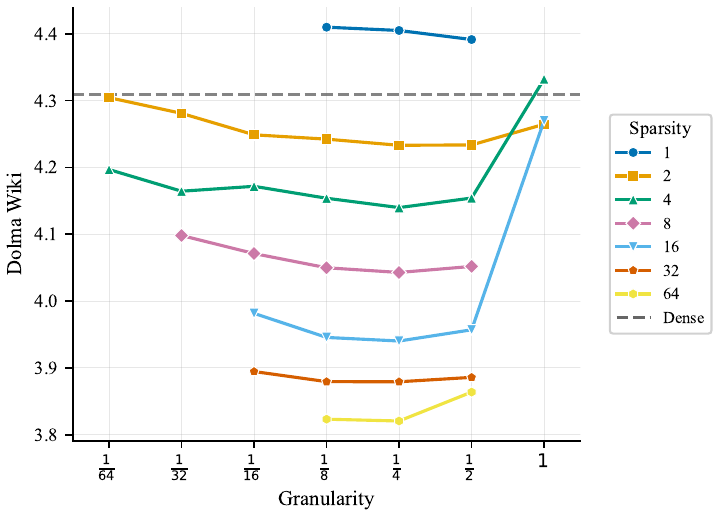}
        \end{subfigure}
        \caption{110M active, 110M - 1.4B total parameters}
    \end{subfigure}
    \end{figure*}

\clearpage  

\begin{figure*}[!ht]
        \addtocounter{figure}{-1}
    \begin{subfigure}[t]{\textwidth}
        \addtocounter{subfigure}{3}
        \begin{subfigure}[t]{0.33\textwidth}
            \centering
            \caption*{\scriptsize Fixed total experts (n)}
            \includegraphics[width=\linewidth]{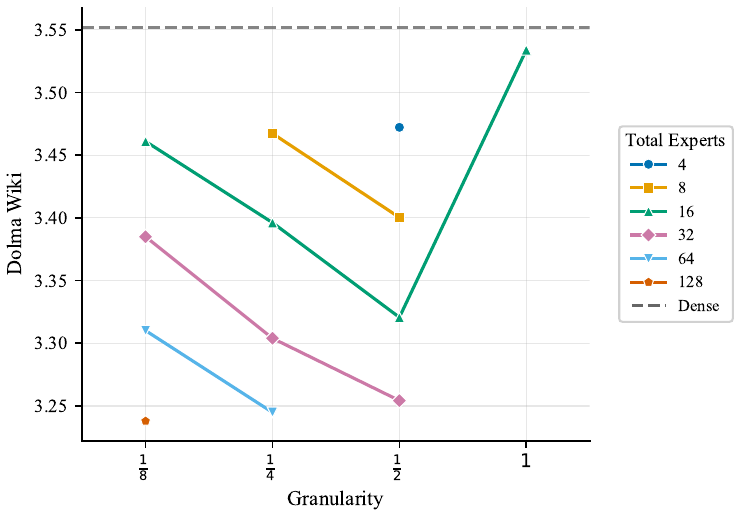}
        \end{subfigure}
        \begin{subfigure}[t]{0.33\textwidth}
            \centering
            \caption*{\scriptsize Fixed granularity (g)}
            \includegraphics[width=\linewidth]{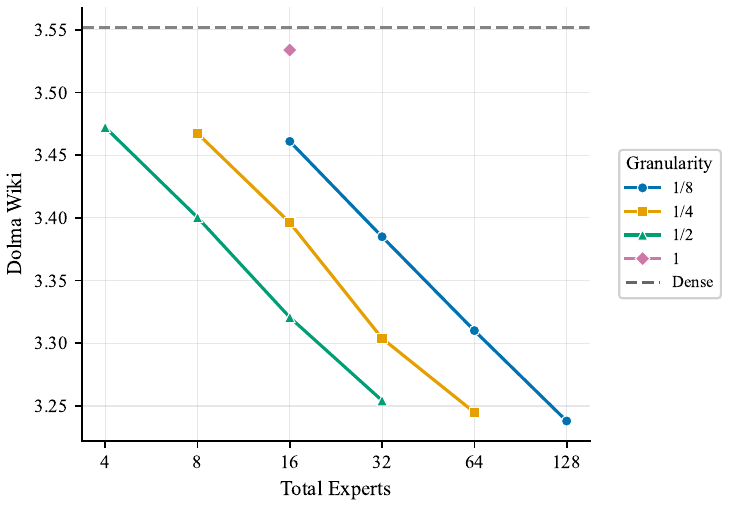}
        \end{subfigure}
        \begin{subfigure}[t]{0.33\textwidth}
            \centering
            \caption*{\scriptsize Fixed activation sparsity (s)}
            \includegraphics[width=\linewidth]{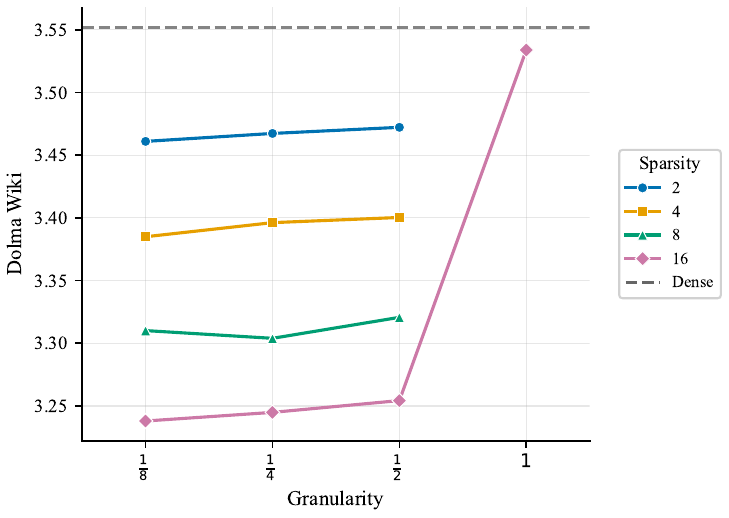}
        \end{subfigure}
        \caption{200M active, 200M - 3.3B total parameters}
    \end{subfigure}
    \par\bigskip\bigskip
        \begin{subfigure}[t]{\textwidth}
        \begin{subfigure}[t]{0.33\textwidth}
            \centering
            \includegraphics[width=\linewidth]{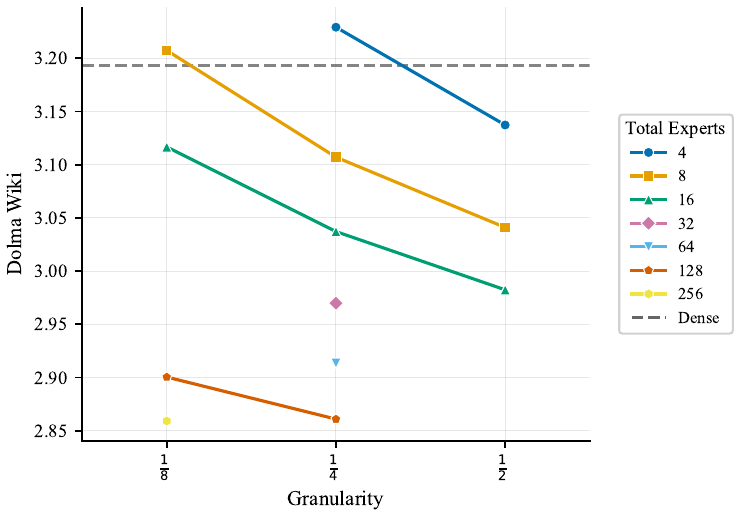}
        \end{subfigure}
        \begin{subfigure}[t]{0.33\textwidth}
            \centering
            \includegraphics[width=\linewidth]{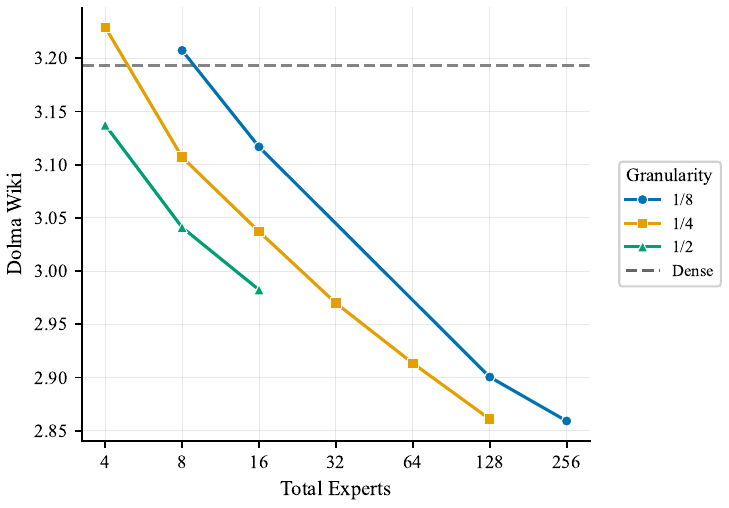}
        \end{subfigure}
        \begin{subfigure}[t]{0.33\textwidth}
            \centering
            \includegraphics[width=\linewidth]{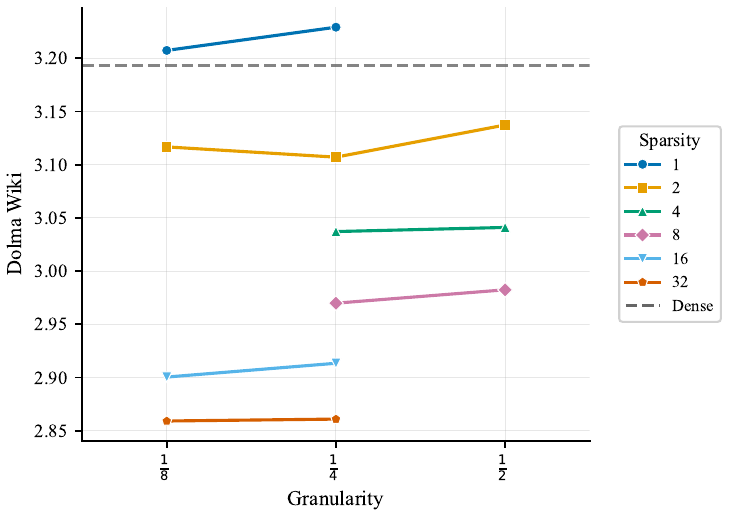}
        \end{subfigure}
        \caption{300M active, 300M - 6.6B total parameters}
    \end{subfigure}

    \caption{
    \textbf{Increasing inactive expert parameters via expert size (left) or total count (center) improves performance in MoEs (\S\ref{sec:expt_main}).} This effect is seen both when holding total number of experts fixed (left) and when holding expert granularity fixed (center). In general, increasing total parameters results in improved performance.  \textbf{Optimal tradeoff between expert count and granularity varies in MoEs (right). (\S\ref{sec:expt_main})}
    At each activation sparsity $s$ (equivalently, at each total parameter count), the optimal (total expert count, expert granularity) configuration varies. As $s$ increases, optimal expert granularity remains nearly fixed, suggesting that sparsity should be scaled up primarily by increasing total expert count $n$, while maintaining a near constant, slowly increasing expert granularity $g$. 
    }
    \label{fig:dolma_wiki_experts}
\end{figure*}

\begin{figure*}[!ht]
    \centering
    
    \begin{subfigure}[t]{0.46\textwidth}
        \centering
        \includegraphics[width=\linewidth]{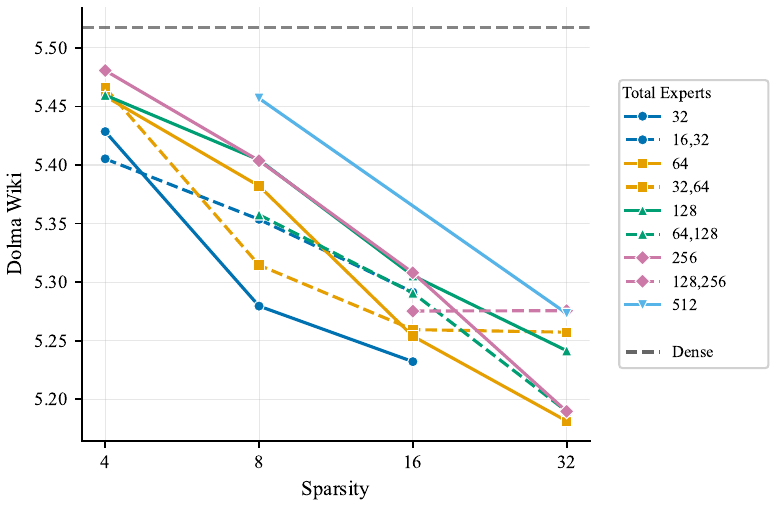}
        \caption{50M active, 50M - 930M total parameters}
    \end{subfigure}
    \vspace{1em}
    \begin{subfigure}[t]{0.46\textwidth}
        \centering
        \includegraphics[width=\linewidth]{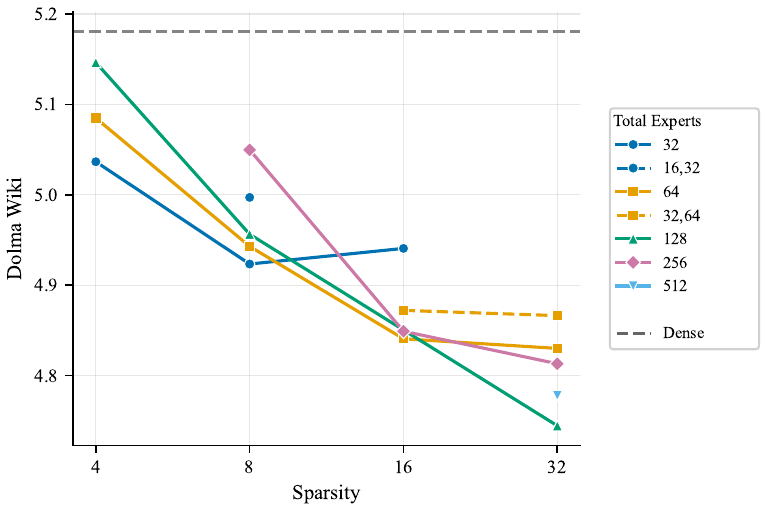}
        \caption{80M active, 80M - 765M total parameters}
    \end{subfigure}
    \caption{
    \textbf{Heterogeneity of expert size alone does not improve MoE performance (\S\ref{sec:expt_hetgen}).} To explore the potential benefits of their architectural flexibility, we compare heterogeneous MoEs (indicated by dotted lines) to active- and total-parameter-matched homogeneous MoEs. Heterogeneity alone does not result in performance gains, as, at each activation sparsity $s$, heterogeneous MoEs with $n_1, n_2 = a, b$ lie between or near the 2 closest homogeneous MoEs, with $n=a$ and with $n=b$.
    }
    \label{fig:dolma_wiki_het}
\end{figure*}

\begin{figure*}[!ht]
    \centering
    
    \begin{subfigure}[t]{1.0\textwidth}
        \centering
        \includegraphics[width=\linewidth]{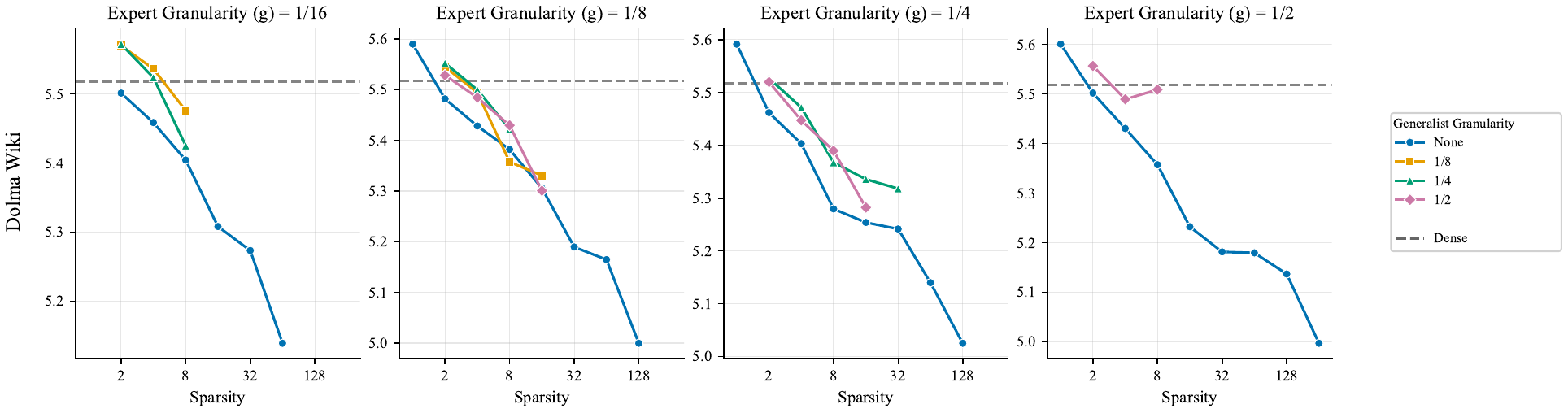}
        \caption{50M active, 50M - 930M total parameters}
    \end{subfigure}
    \par\bigskip\bigskip
    \begin{subfigure}[t]{1.0\textwidth}
        \centering
        \includegraphics[width=\linewidth]{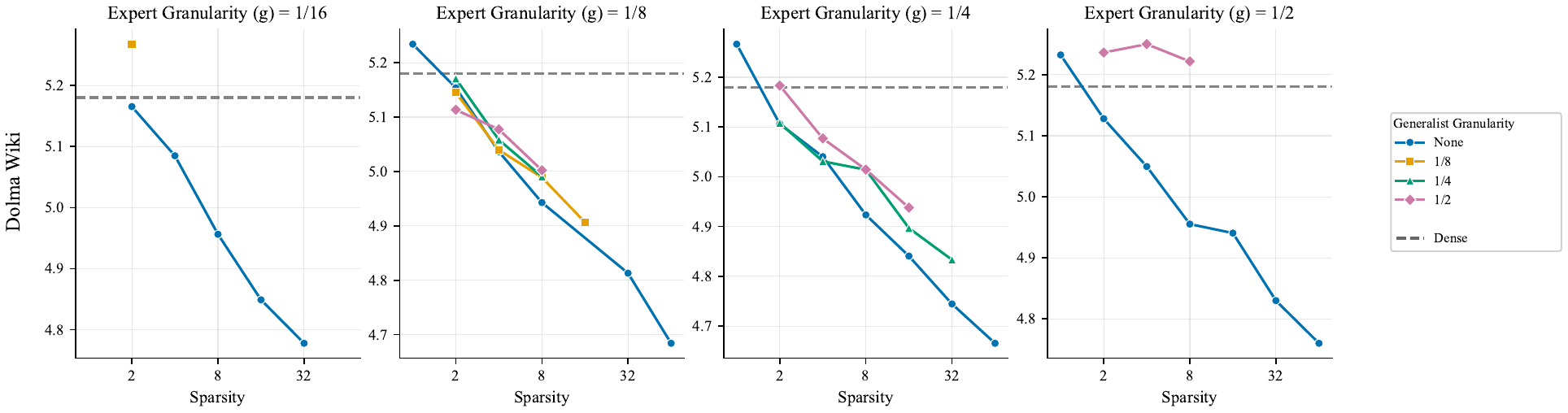}
        \caption{80M active, 80M - 765M total parameters}
    \end{subfigure}
    \par\bigskip\bigskip
    \begin{subfigure}[t]{1.0\textwidth}
        \centering
        \includegraphics[width=\linewidth]{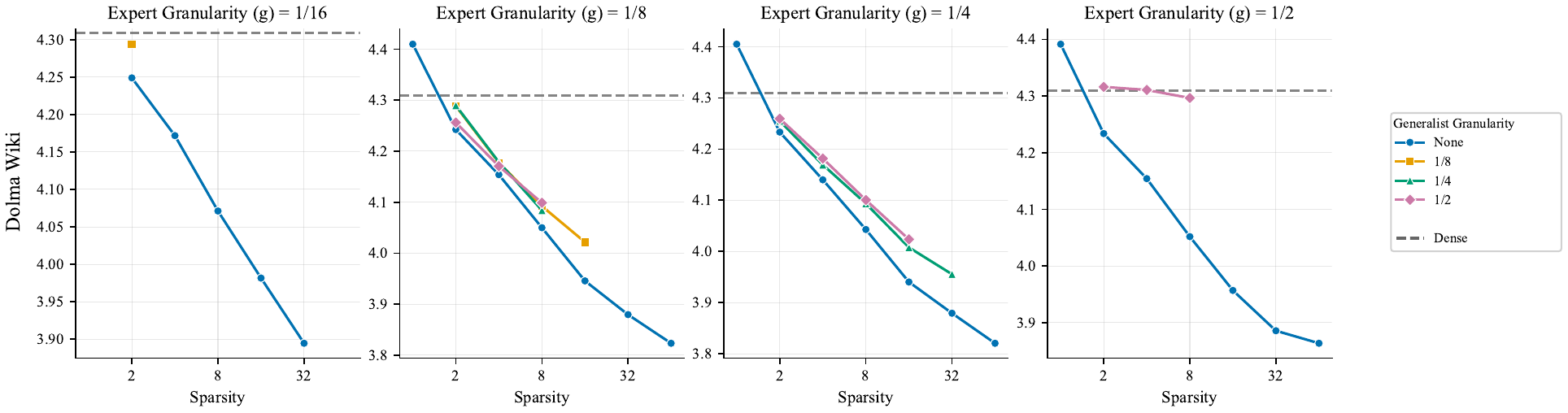}
        \caption{110M active, 110M - 1.4B total parameters}
    \end{subfigure}
    \caption{
    \textbf{The inclusion of a generalist consistently degrades performance in homogeneous MoEs (\S\ref{sec:expt_hetgen}).}
    We train MoE LMs which consist of some routed experts with granularity $g$, as well as a generalist with granularity $g_{gen}\in \{\frac{1}{2}, \frac{1}{4}, \frac{1}{8}\} $. We compare to settings with no generalist, only routed experts with granularity $g$. In all settings and configurations, the addition of any granularity generalist results in comparable or degraded performance. 
    }
    \label{fig:dolma_wiki_gen}
\end{figure*}

\begin{figure*}[ht]
    \centering
    \begin{subfigure}[t]{1.0\textwidth}
        \centering
        \includegraphics[width=\linewidth]{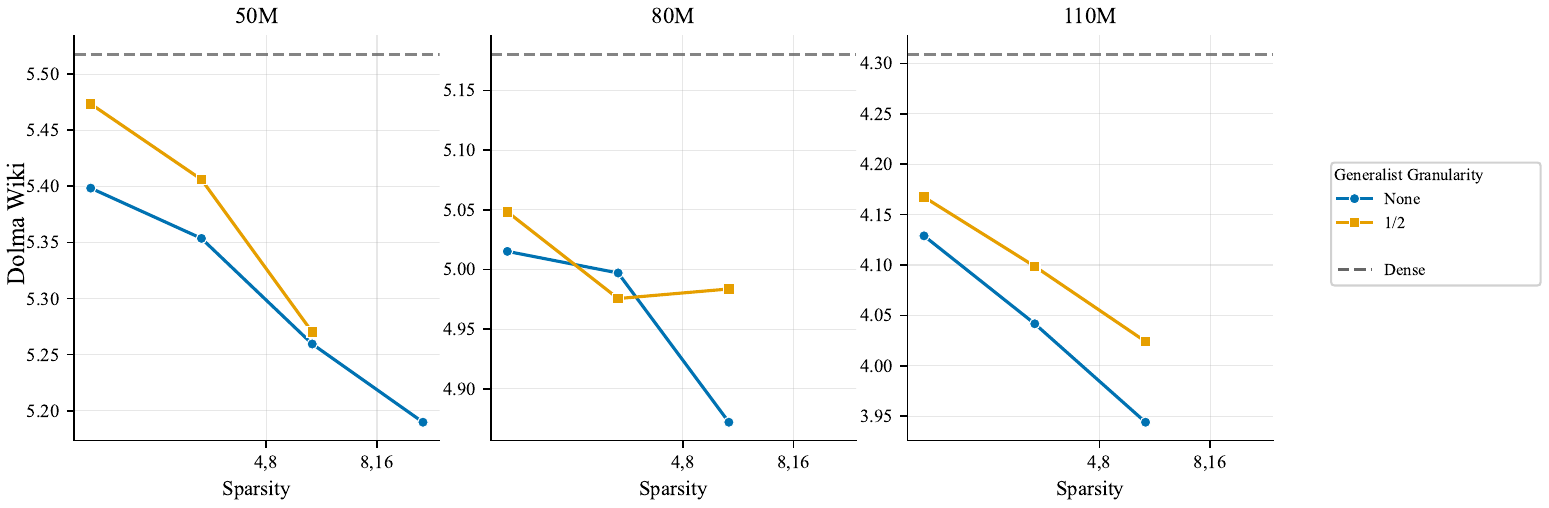}
    \end{subfigure}
    \caption{
    \textbf{The inclusion of a generalist consistently degrades performance in heterogeneous MoEs (\S\ref{sec:expt_hetgen}).}
    We train heterogeneous MoE LMs which consist of  routed experts with granularity $g_1, g_2$, as well as a generalist with granularity $g_{gen} = \frac{1}{2}$. We compare to settings with no generalist. In all settings and configurations, the addition of a generalist results in comparable or degraded performance. 
    }
    \label{fig:dolma_wiki_hetgen}
\end{figure*}

\begin{figure*}[ht]
    \centering
    \begin{subfigure}[t]{\textwidth}
        \centering
        \begin{subfigure}[t]{0.45\textwidth}
            \includegraphics[width=\linewidth]{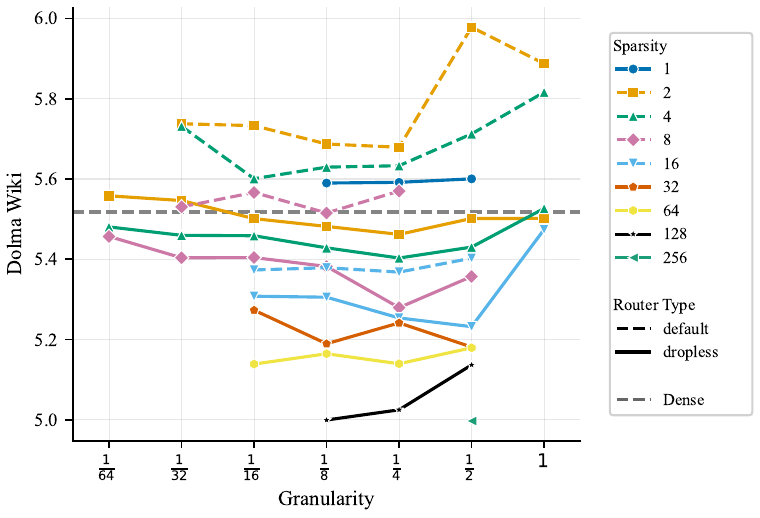}
            \caption{50M active, 50M - 930M total parameters}
        \end{subfigure}
    \hspace{1em}
        \begin{subfigure}[t]{0.45\textwidth}
            \centering
            \includegraphics[width=\linewidth]{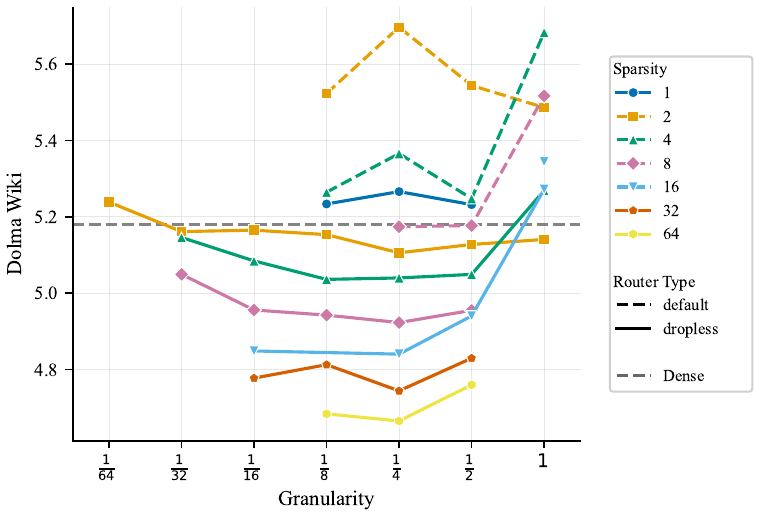}
            \caption{80M active, 80M - 765M total parameters}
        \end{subfigure}
    \end{subfigure}

    \par\bigskip\bigskip
    \begin{subfigure}[t]{0.45\textwidth}
        \centering
        \includegraphics[width=\linewidth]{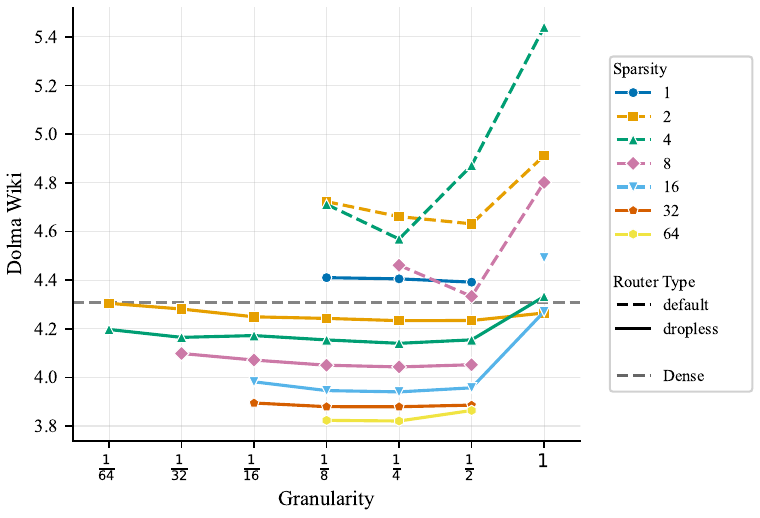}
        \caption{110M active, 110M - 1.4B total parameters}
    \end{subfigure}
    \caption{ 
    \textbf{Dropless routing outperforms default routing (\S\ref{sec:expt_router}).}
    We compare dropless routing to the default setting, which allow tokens to be dropped. Across all scales, we find that dropless routing outperforms or performs comparably to default routing. 
    }
    \label{fig:dolma_wiki_dropless}
\end{figure*}

\begin{figure*}[ht]
    \centering
    \begin{subfigure}[t]{0.45\textwidth}
        \centering
        \includegraphics[width=\linewidth]{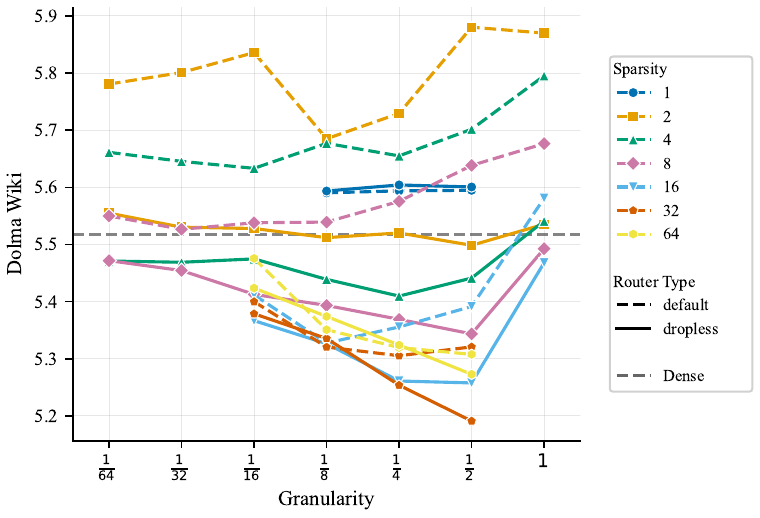}
        \caption{50M active, 50M - 930M total parameters}
    \end{subfigure}
    \hspace{1em}
    \begin{subfigure}[t]{0.45\textwidth}
        \centering
        \includegraphics[width=\linewidth]{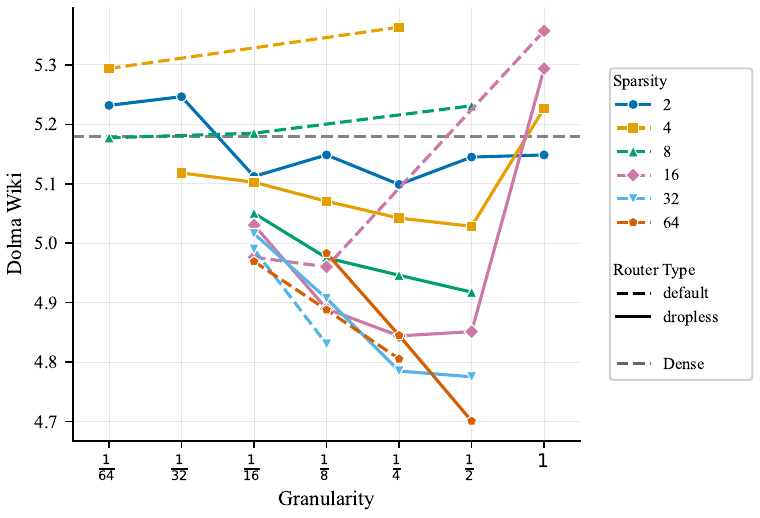}
        \caption{80M active, 80M - 765M total parameters}
    \end{subfigure}
    \caption{
    \textbf{Dropless routing, with bias $\gamma=\num{1e-3}$ (\S\ref{sec:expt_router}).} 
    As in Figure~\ref{fig:lm_avg_dropless}, we compare dropless routing to the default setting, which allow tokens to be dropped. Across all scales, we find that dropless routing outperforms or performs comparably to default routing. We see here with additional higher sparsity default routing runs that as sparsity increases, default routing performance approaches that of dropless routing.
    }
    \label{fig:dolma_wiki_dropless_with_lf}
\end{figure*}

\begin{figure*}[ht]
    \centering
    \begin{subfigure}[]{\textwidth}
        \centering
        \includegraphics[width=0.46\linewidth]{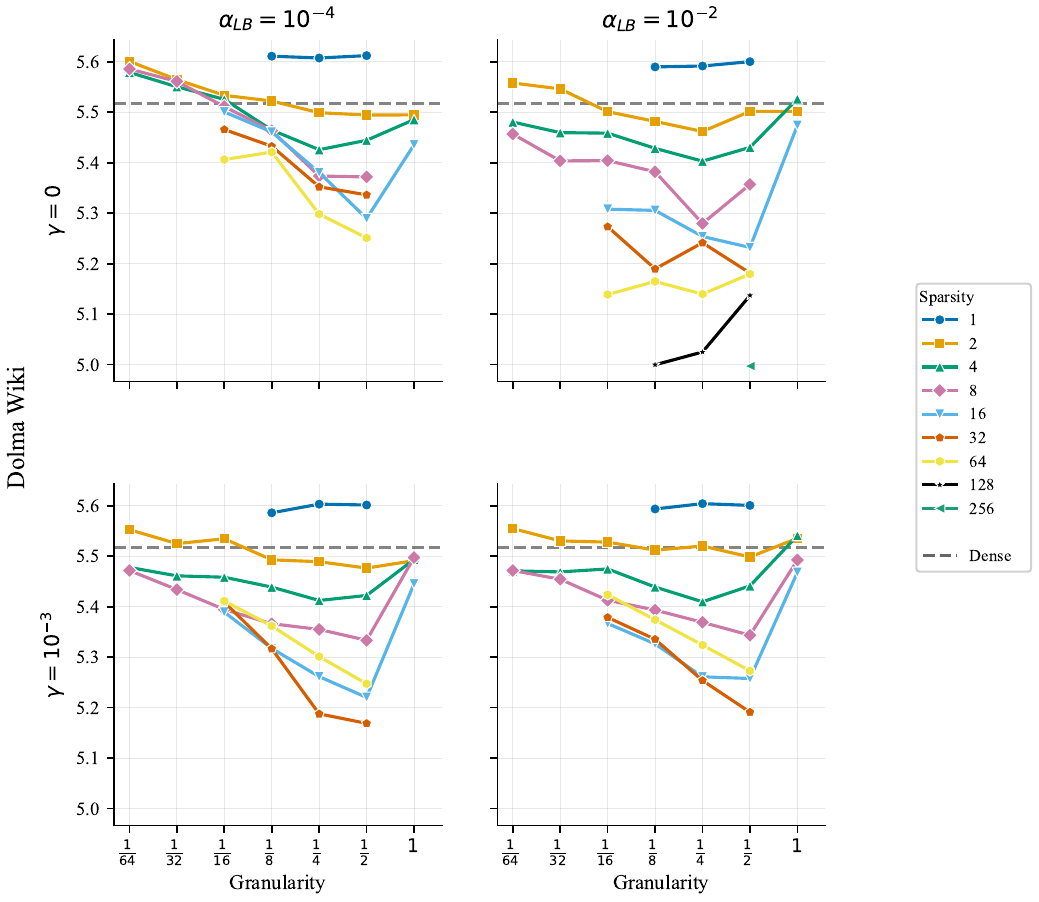}
        \hspace{1em}
        \includegraphics[width=0.46\linewidth]{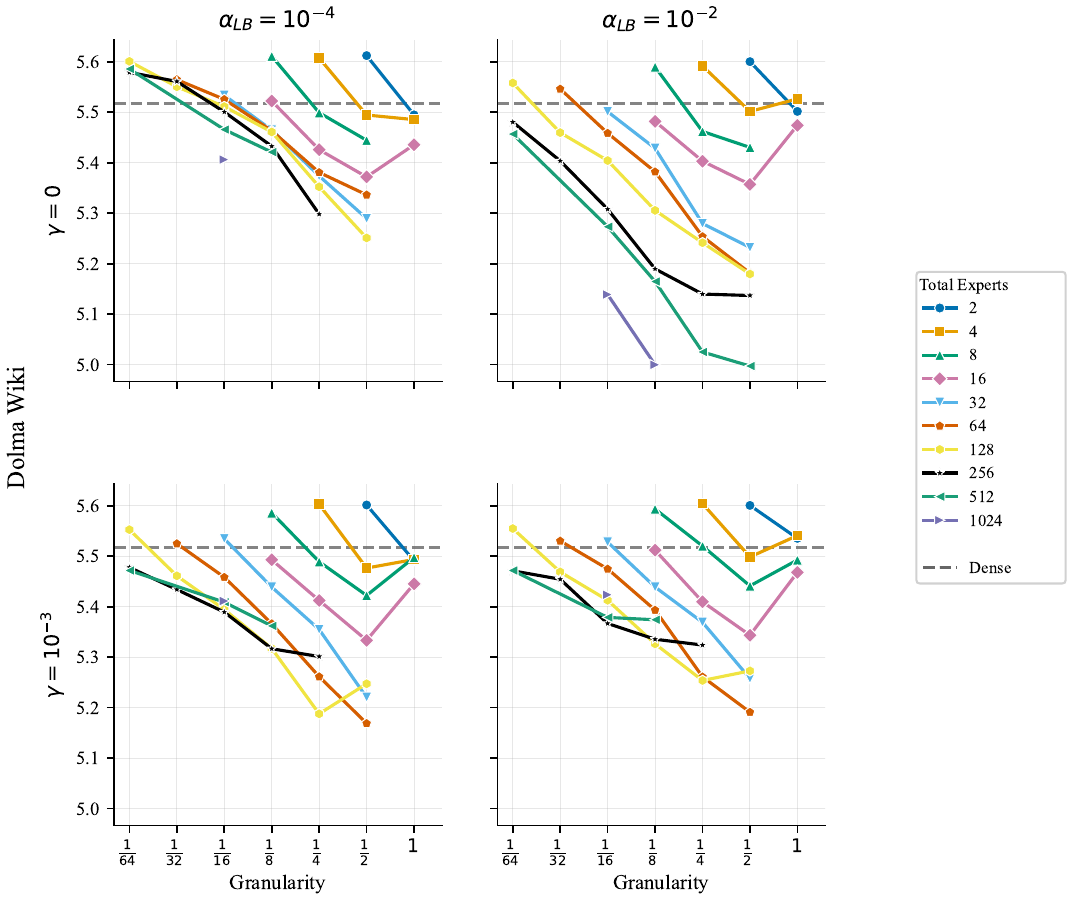}
        \caption{50M active, 50M - 930M total parameters}
    \end{subfigure}
    \par\bigskip\bigskip
    \begin{subfigure}[]{\textwidth}
        \centering
        \includegraphics[width=0.46\linewidth]{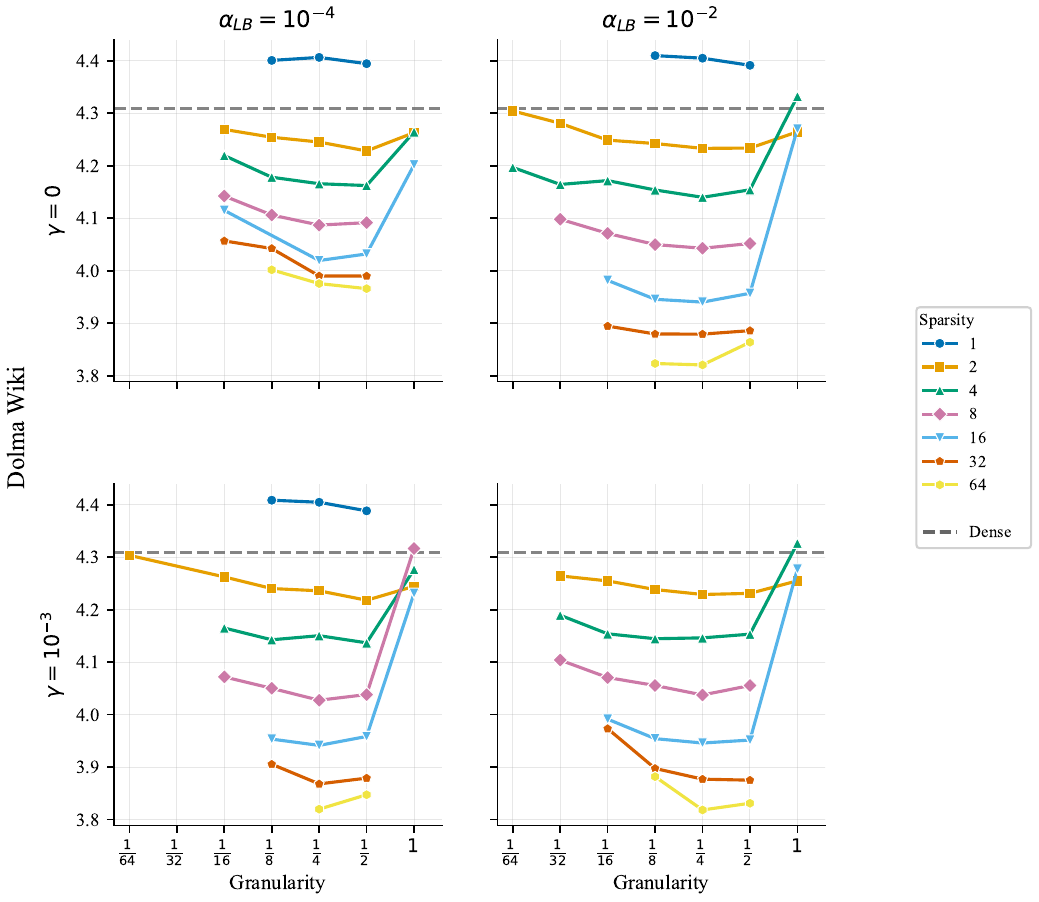}
        \hspace{1em}
        \includegraphics[width=0.46\linewidth]{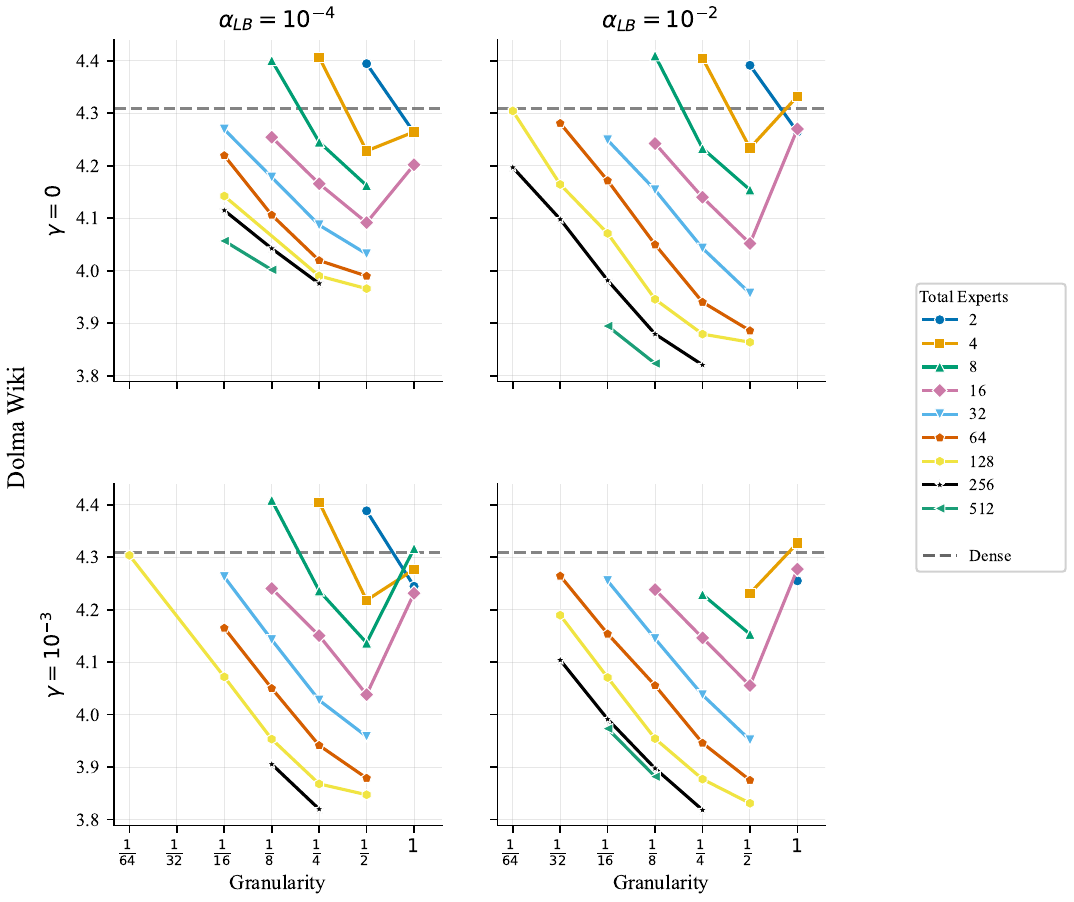}
        \caption{80M active, 80M - 765M total parameters}
    \end{subfigure}
    \par\bigskip\bigskip
    \begin{subfigure}[t]{\textwidth}
        \centering
        \includegraphics[width=0.46\linewidth]{figures/lm/dolma_wiki-validation/ce_loss/lb_sweep_hgn_gxs_110M.pdf}
        \hspace{1em}
        \includegraphics[width=0.46\linewidth]{figures/lm/dolma_wiki-validation/ce_loss/lb_sweep_hgn_gxn_110M.pdf}
        \caption{110M active, 110M - 1.4B total parameters}
    \end{subfigure}

    \end{figure*} 

\clearpage  

\begin{figure*}[ht]
    \addtocounter{figure}{-1}
    \centering
    \begin{subfigure}[t]{\textwidth}
        \centering
        \includegraphics[width=0.46\linewidth]{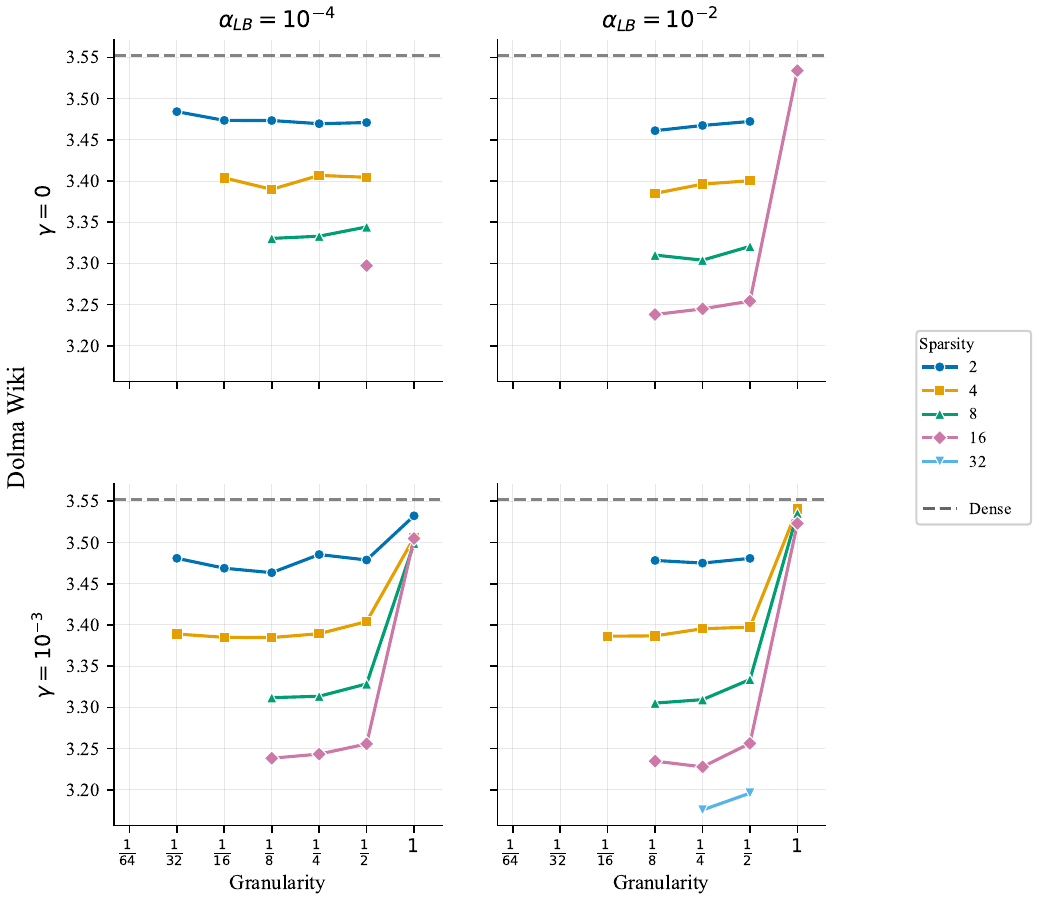}
        \hspace{1em}
        \includegraphics[width=0.46\linewidth]{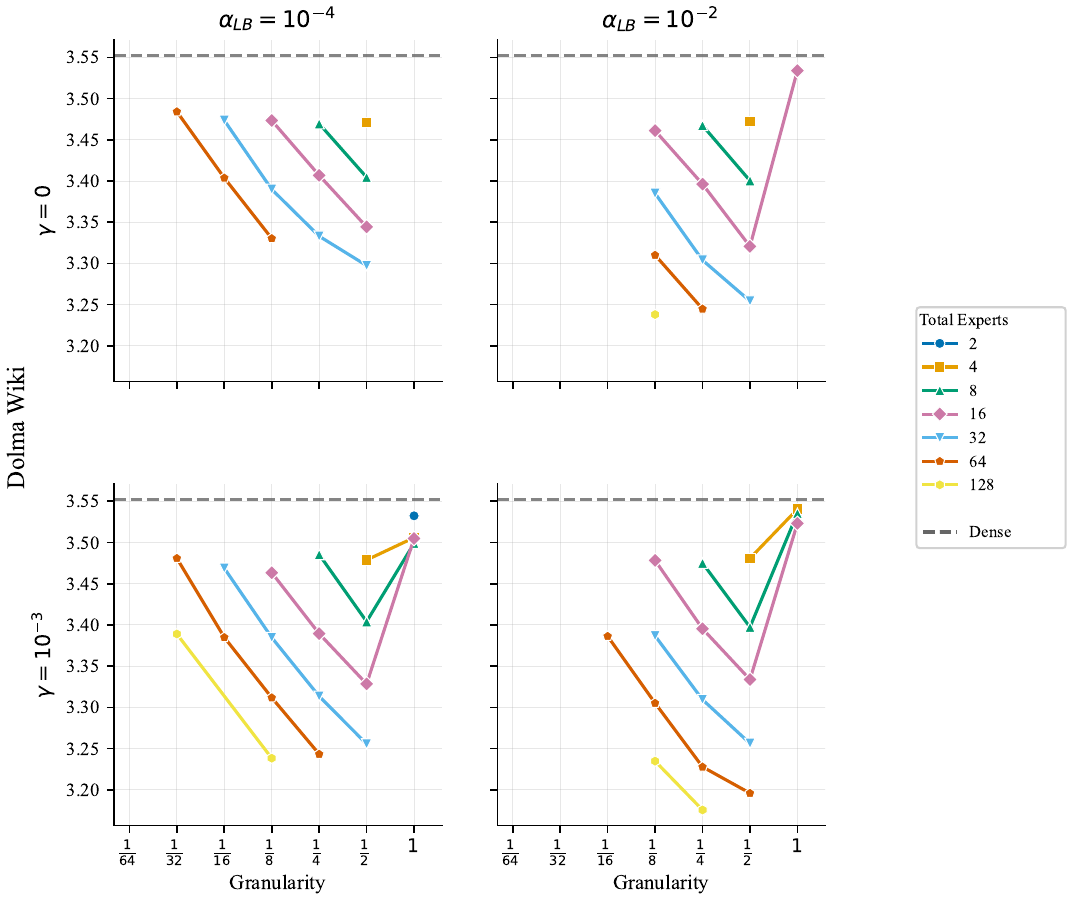}
        \caption{200M active, 200M - 3.3B total parameters}
    \end{subfigure}
    \par\bigskip\bigskip
    \begin{subfigure}[t]{\textwidth}
        \centering
        \includegraphics[width=0.3\linewidth]{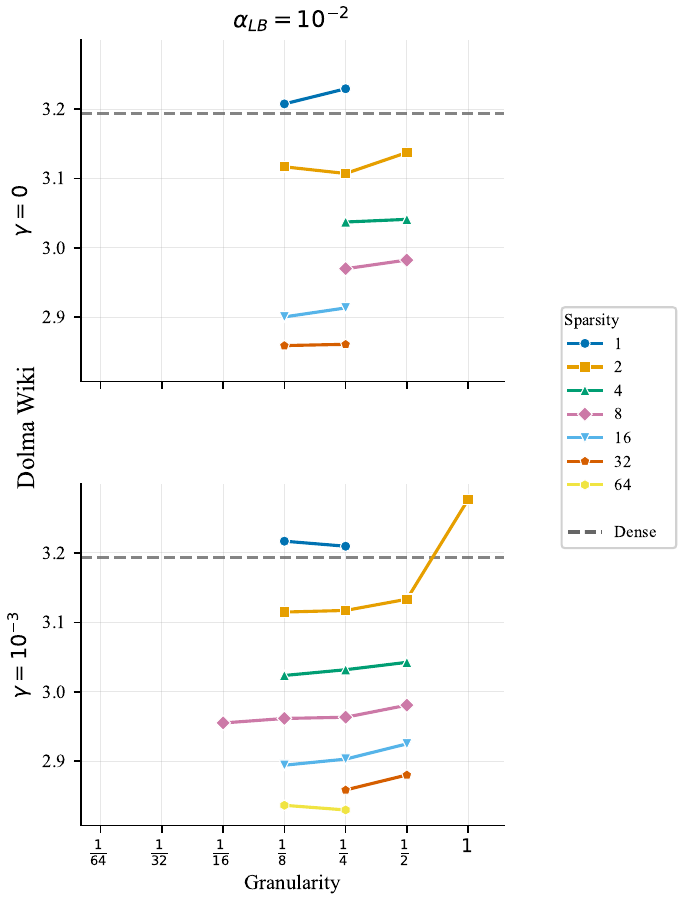}
        \hspace{1em}
        \includegraphics[width=0.3\linewidth]{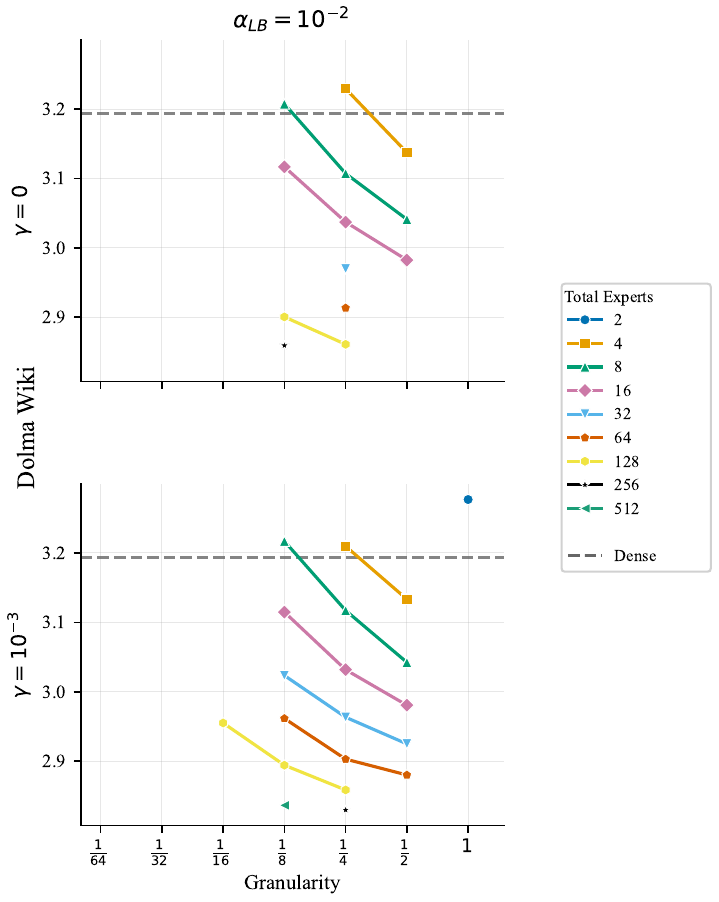}
        \caption{300M active, 300M - 6.6B total parameters}
    \end{subfigure}

    \caption{
    \textbf{Load balancing mechanisms must be tuned correctly (\S\ref{sec:expt_router}).}
    We consider load balancing loss weight $\alpha_{LB} \in \{\num{1e-2}, \num{1e-4}\}$ and loss-free load balancing with bias $\gamma\in\{0, \num{1e-3}\}$ ($\gamma=0$ indicates no loss-free mechanism). Results show that poorly chosen hyperparameters, such as high bias $\gamma = 1e-3$ with total experts $n\geq 512$, may impair performance. However, all settings other than $(\alpha_{LB}=\num{1e-2}, \gamma=\num{1e-3})$ perform comparably for $n \leq 512$, suggesting that a wide range of load balancing settings achieve near-optimal performance. 
    }
    \label{fig:dolma_wiki_lb}
\end{figure*}

%% file: fig_tex/lm/ice.tex
\begin{figure*}[!ht]
    \centering
        \begin{subfigure}[t]{\textwidth}
        \begin{subfigure}[t]{0.33\textwidth}
            \centering
            \caption*{\scriptsize Fixed total experts (n)}
            \includegraphics[width=\linewidth]{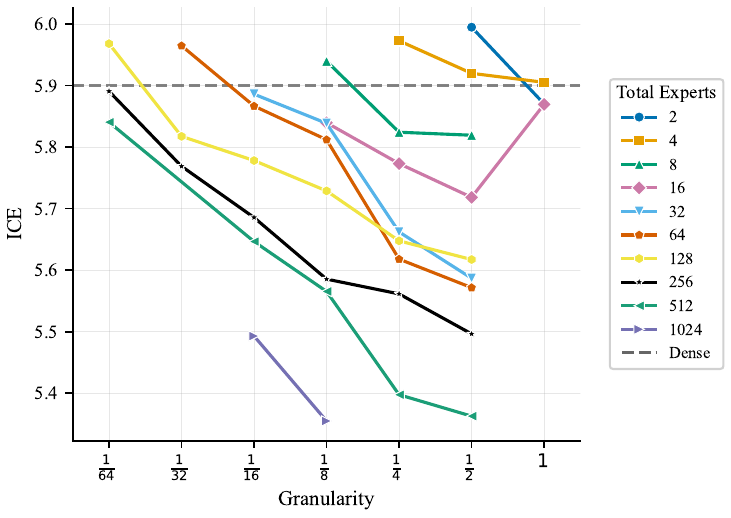}
        \end{subfigure}
        \begin{subfigure}[t]{0.33\textwidth}
            \centering
            \caption*{\scriptsize Fixed granularity (g)}
            \includegraphics[width=\linewidth]{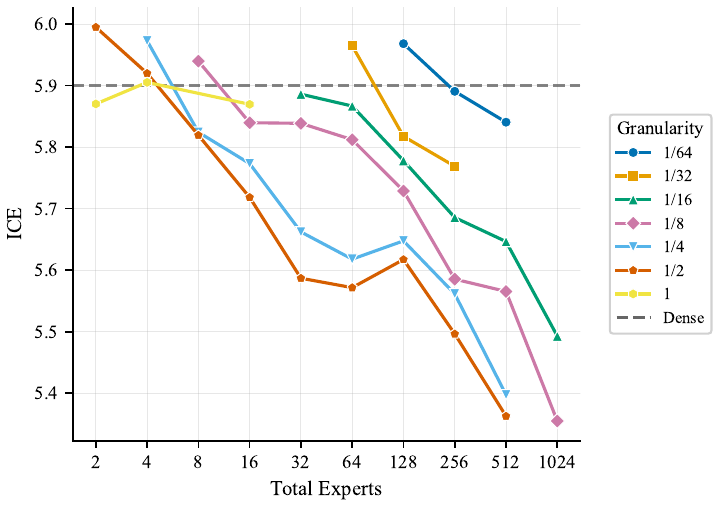}
        \end{subfigure}
        \begin{subfigure}[t]{0.33\textwidth}
            \centering
            \caption*{\scriptsize Fixed activation sparsity (s)}
            \includegraphics[width=\linewidth]{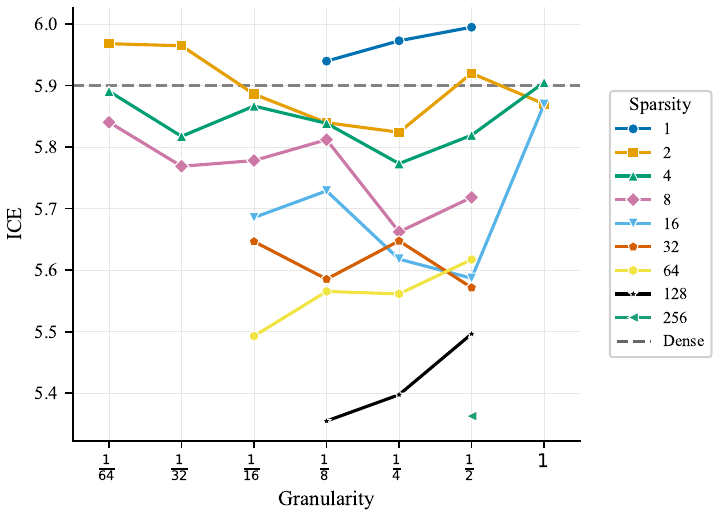}
        \end{subfigure}
        \caption{50M active, 50M - 930M total parameters}
    \end{subfigure}
\par\bigskip\bigskip
    \begin{subfigure}[t]{\textwidth}
        \begin{subfigure}[t]{0.33\textwidth}
            \centering
            \includegraphics[width=\linewidth]{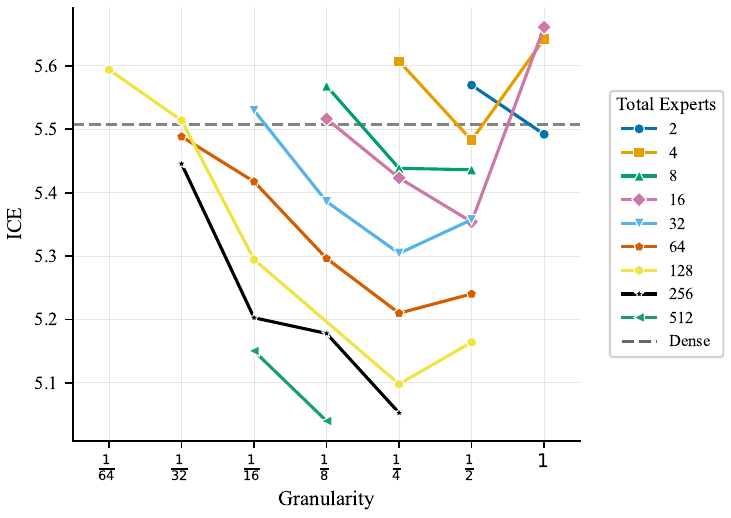}
        \end{subfigure}
        \begin{subfigure}[t]{0.33\textwidth}
            \centering
            \includegraphics[width=\linewidth]{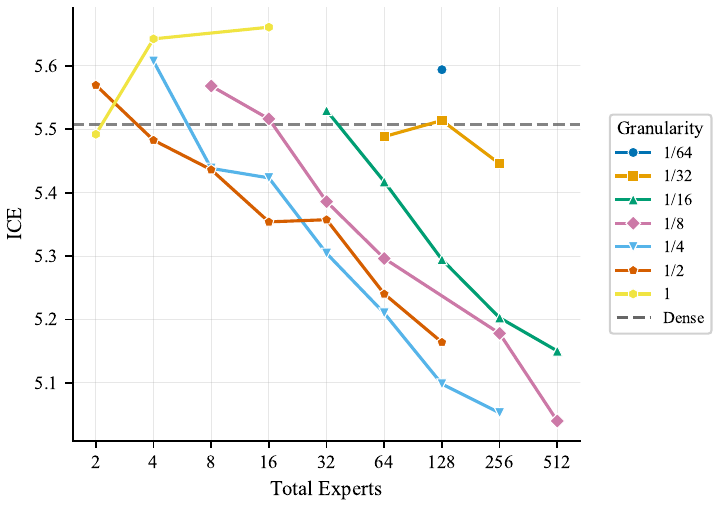}
        \end{subfigure}
        \begin{subfigure}[t]{0.33\textwidth}
            \centering
            \includegraphics[width=\linewidth]{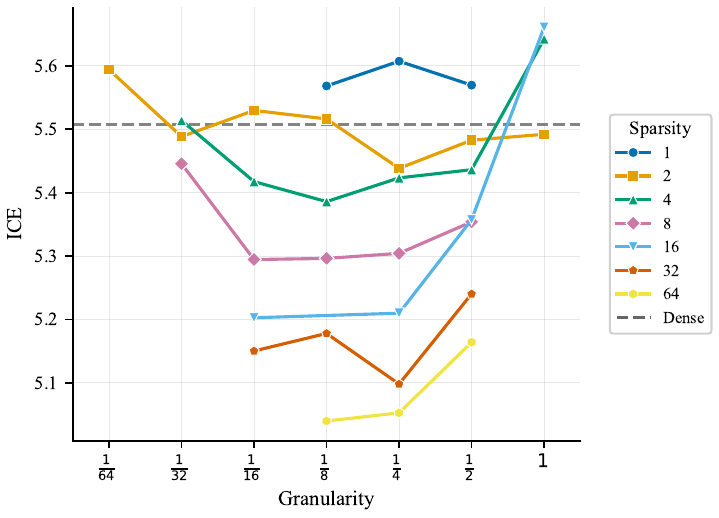}
        \end{subfigure}
        \caption{80M active, 80M - 765M total parameters}
    \end{subfigure}
    \par\bigskip\bigskip
        \begin{subfigure}[t]{\textwidth}
        \begin{subfigure}[t]{0.33\textwidth}
            \centering
            \includegraphics[width=\linewidth]{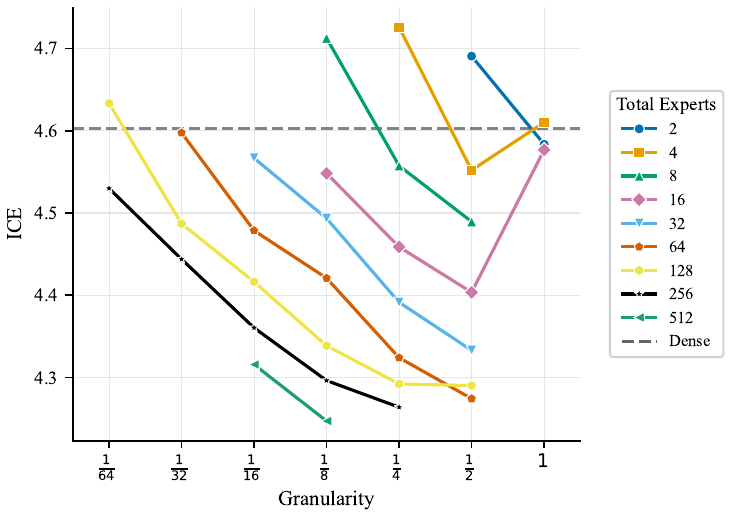}
        \end{subfigure}
        \begin{subfigure}[t]{0.33\textwidth}
            \centering
            \includegraphics[width=\linewidth]{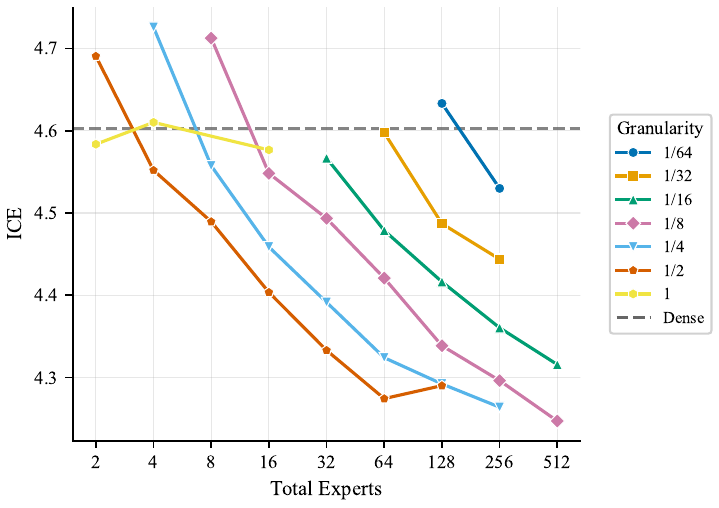}
        \end{subfigure}
        \begin{subfigure}[t]{0.33\textwidth}
            \centering
            \includegraphics[width=\linewidth]{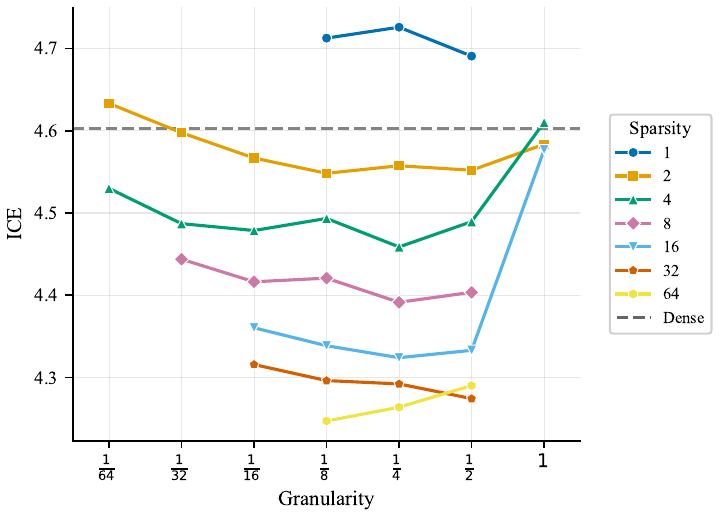}
        \end{subfigure}
        \caption{110M active, 110M - 1.4B total parameters}
    \end{subfigure}
    \end{figure*}

\clearpage  

\begin{figure*}[!ht]
        \addtocounter{figure}{-1}
    \begin{subfigure}[t]{\textwidth}
        \addtocounter{subfigure}{3}
        \begin{subfigure}[t]{0.33\textwidth}
            \centering
            \caption*{\scriptsize Fixed total experts (n)}
            \includegraphics[width=\linewidth]{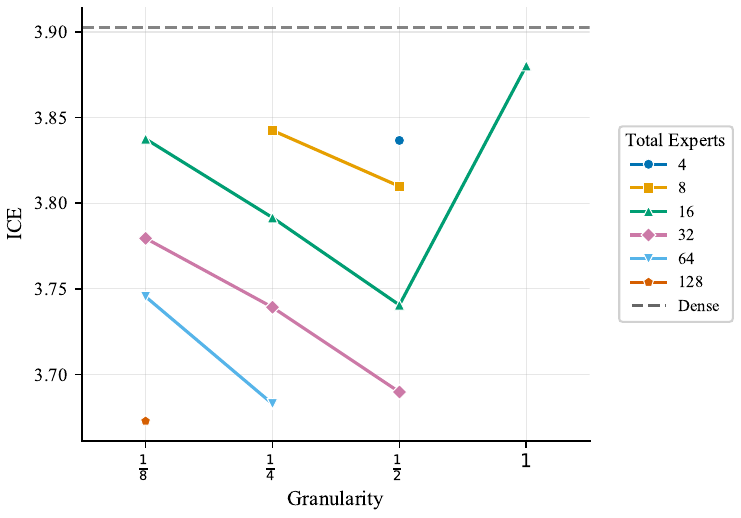}
        \end{subfigure}
        \begin{subfigure}[t]{0.33\textwidth}
            \centering
            \caption*{\scriptsize Fixed granularity (g)}
            \includegraphics[width=\linewidth]{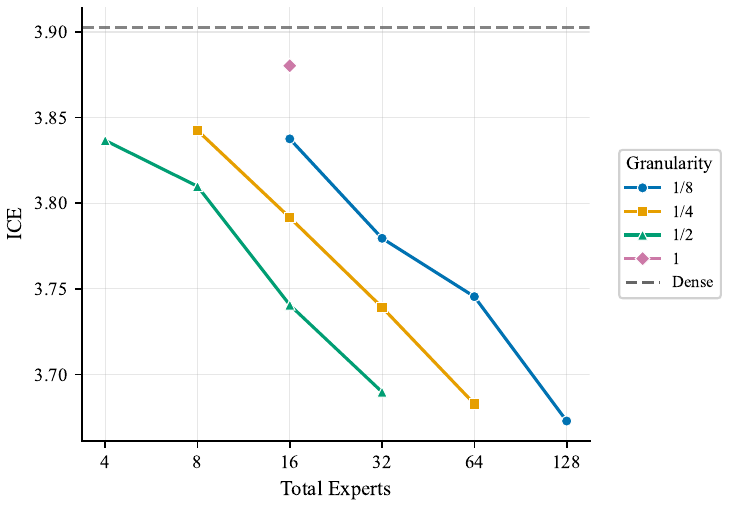}
        \end{subfigure}
        \begin{subfigure}[t]{0.33\textwidth}
            \centering
            \caption*{\scriptsize Fixed activation sparsity (s)}
            \includegraphics[width=\linewidth]{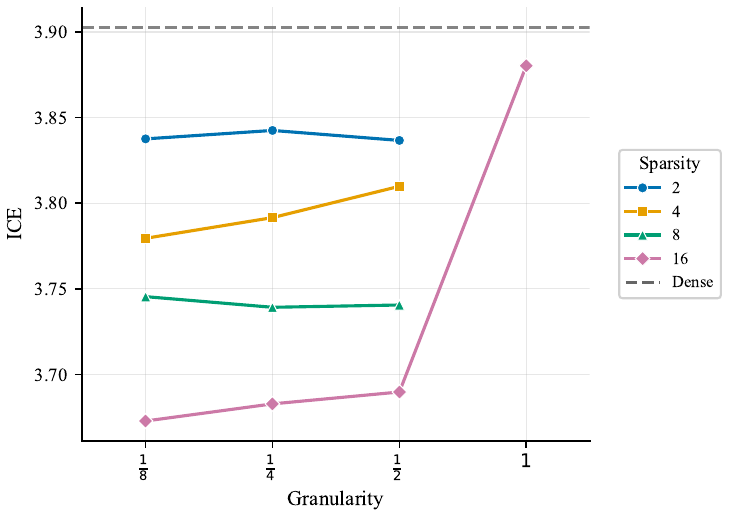}
        \end{subfigure}
        \caption{200M active, 200M - 3.3B total parameters}
    \end{subfigure}
    \par\bigskip\bigskip
        \begin{subfigure}[t]{\textwidth}
        \begin{subfigure}[t]{0.33\textwidth}
            \centering
            \includegraphics[width=\linewidth]{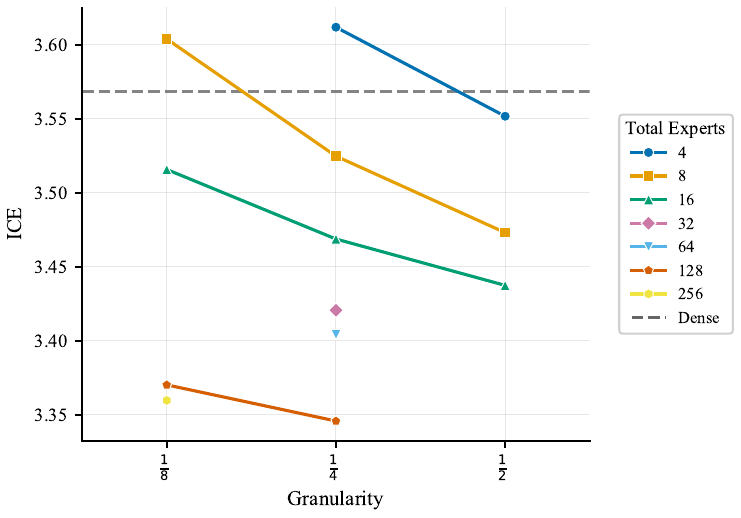}
        \end{subfigure}
        \begin{subfigure}[t]{0.33\textwidth}
            \centering
            \includegraphics[width=\linewidth]{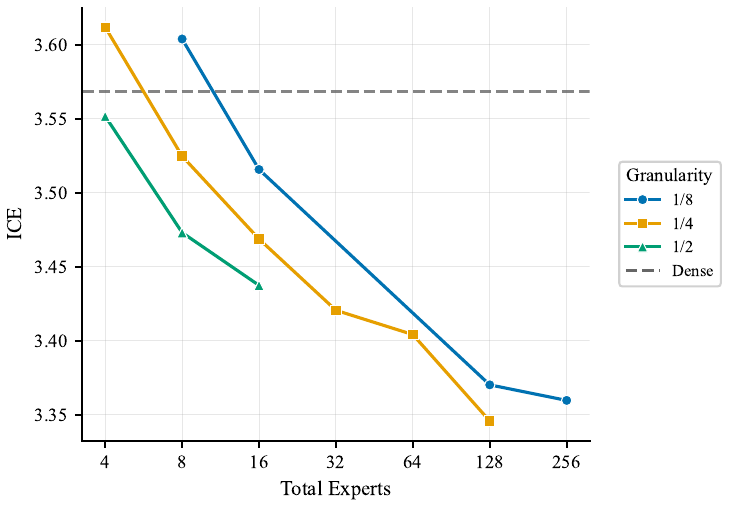}
        \end{subfigure}
        \begin{subfigure}[t]{0.33\textwidth}
            \centering
            \includegraphics[width=\linewidth]{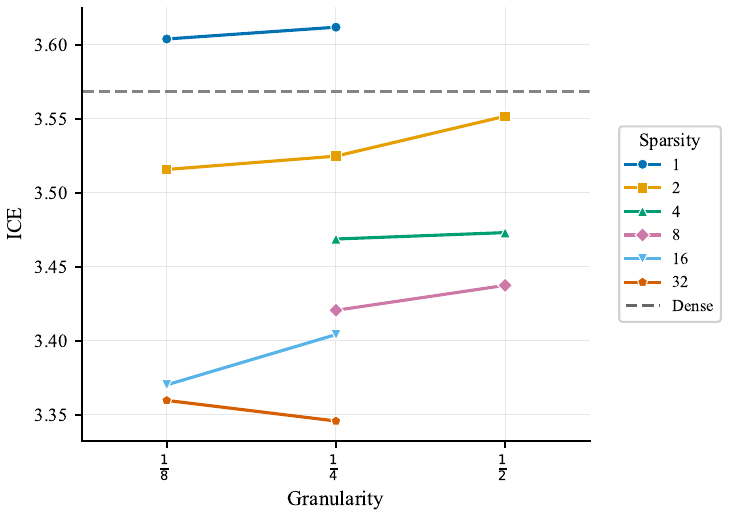}
        \end{subfigure}
        \caption{300M active, 300M - 6.6B total parameters}
    \end{subfigure}

    \caption{
    \textbf{Increasing inactive expert parameters via expert size (left) or total count (center) improves performance in MoEs (\S\ref{sec:expt_main}).} This effect is seen both when holding total number of experts fixed (left) and when holding expert granularity fixed (center). In general, increasing total parameters results in improved performance.  \textbf{Optimal tradeoff between expert count and granularity varies in MoEs (right). (\S\ref{sec:expt_main})}
    At each activation sparsity $s$ (equivalently, at each total parameter count), the optimal (total expert count, expert granularity) configuration varies. As $s$ increases, optimal expert granularity remains nearly fixed, suggesting that sparsity should be scaled up primarily by increasing total expert count $n$, while maintaining a near constant, slowly increasing expert granularity $g$. 
    }
    \label{fig:ice_experts}
\end{figure*}

\begin{figure*}[!ht]
    \centering
    
    \begin{subfigure}[t]{0.46\textwidth}
        \centering
        \includegraphics[width=\linewidth]{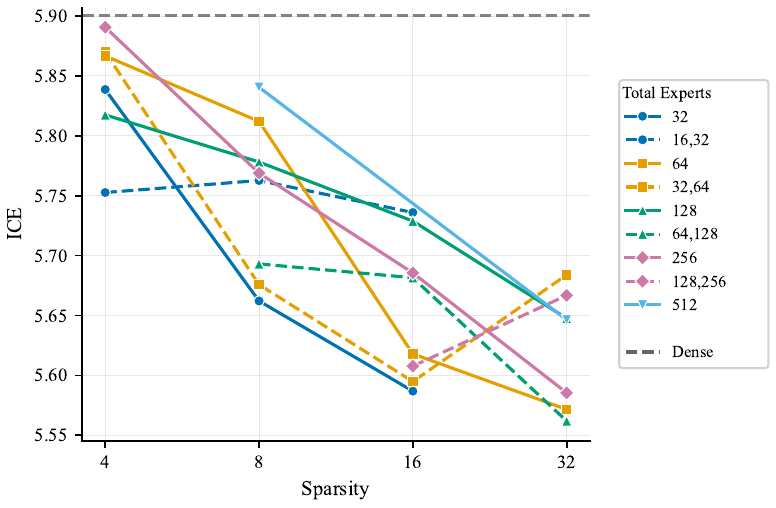}
        \caption{50M active, 50M - 930M total parameters}
    \end{subfigure}
    \vspace{1em}
    \begin{subfigure}[t]{0.46\textwidth}
        \centering
        \includegraphics[width=\linewidth]{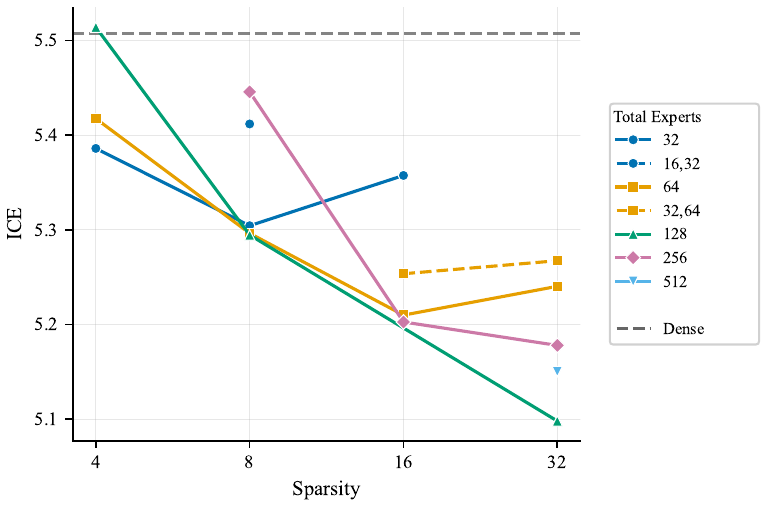}
        \caption{80M active, 80M - 765M total parameters}
    \end{subfigure}
    \caption{
    \textbf{Heterogeneity of expert size alone does not improve MoE performance (\S\ref{sec:expt_hetgen}).} To explore the potential benefits of their architectural flexibility, we compare heterogeneous MoEs (indicated by dotted lines) to active- and total-parameter-matched homogeneous MoEs. Heterogeneity alone does not result in performance gains, as, at each activation sparsity $s$, heterogeneous MoEs with $n_1, n_2 = a, b$ lie between or near the 2 closest homogeneous MoEs, with $n=a$ and with $n=b$.
    }
    \label{fig:ice_het}
\end{figure*}

\begin{figure*}[!ht]
    \centering
    
    \begin{subfigure}[t]{1.0\textwidth}
        \centering
        \includegraphics[width=\linewidth]{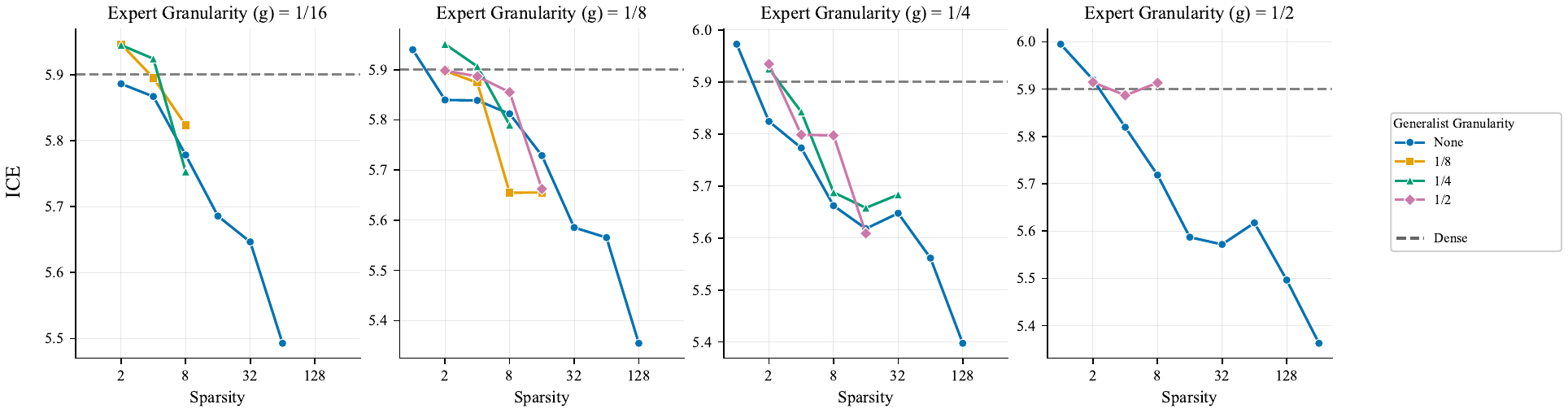}
        \caption{50M active, 50M - 930M total parameters}
    \end{subfigure}
    \par\bigskip\bigskip
    \begin{subfigure}[t]{1.0\textwidth}
        \centering
        \includegraphics[width=\linewidth]{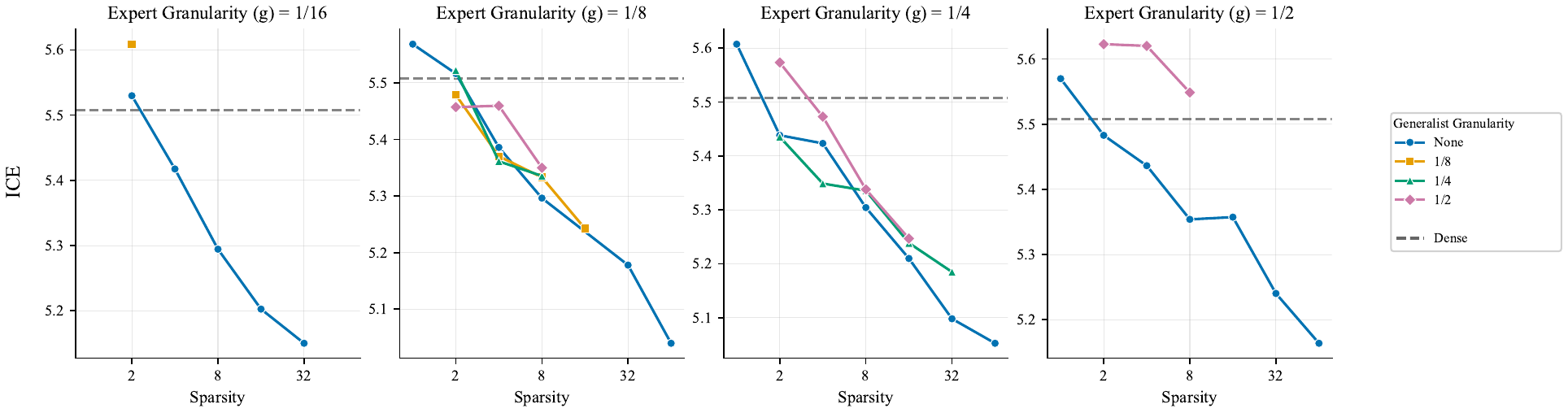}
        \caption{80M active, 80M - 765M total parameters}
    \end{subfigure}
    \par\bigskip\bigskip
    \begin{subfigure}[t]{1.0\textwidth}
        \centering
        \includegraphics[width=\linewidth]{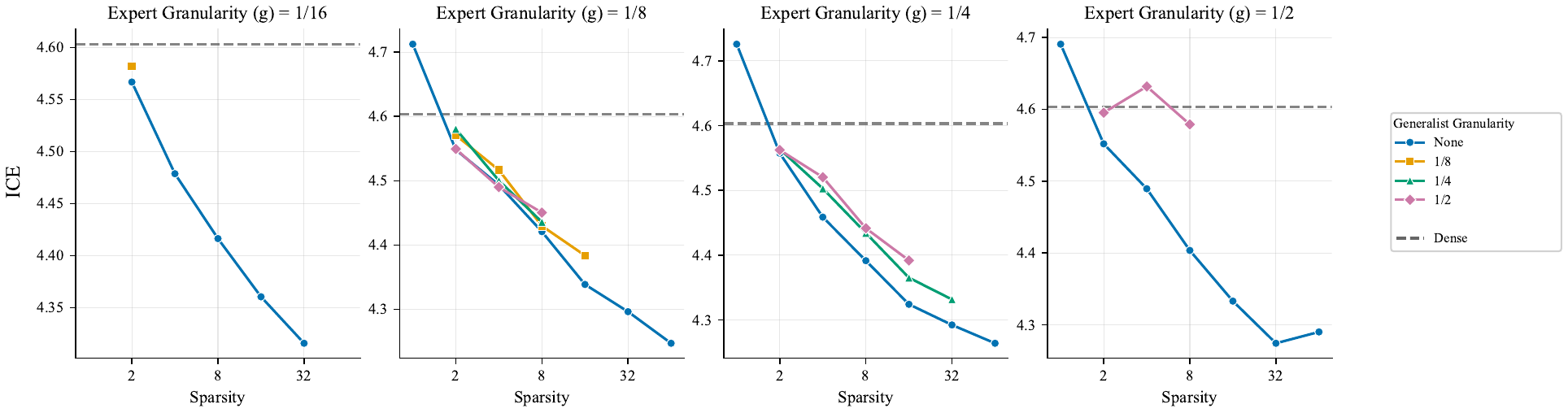}
        \caption{110M active, 110M - 1.4B total parameters}
    \end{subfigure}
    \caption{
    \textbf{The inclusion of a generalist consistently degrades performance in homogeneous MoEs (\S\ref{sec:expt_hetgen}).}
    We train MoE LMs which consist of some routed experts with granularity $g$, as well as a generalist with granularity $g_{gen}\in \{\frac{1}{2}, \frac{1}{4}, \frac{1}{8}\} $. We compare to settings with no generalist, only routed experts with granularity $g$. In all settings and configurations, the addition of any granularity generalist results in comparable or degraded performance. 
    }
    \label{fig:ice_gen}
\end{figure*}

\begin{figure*}[ht]
    \centering
    \begin{subfigure}[t]{1.0\textwidth}
        \centering
        \includegraphics[width=\linewidth]{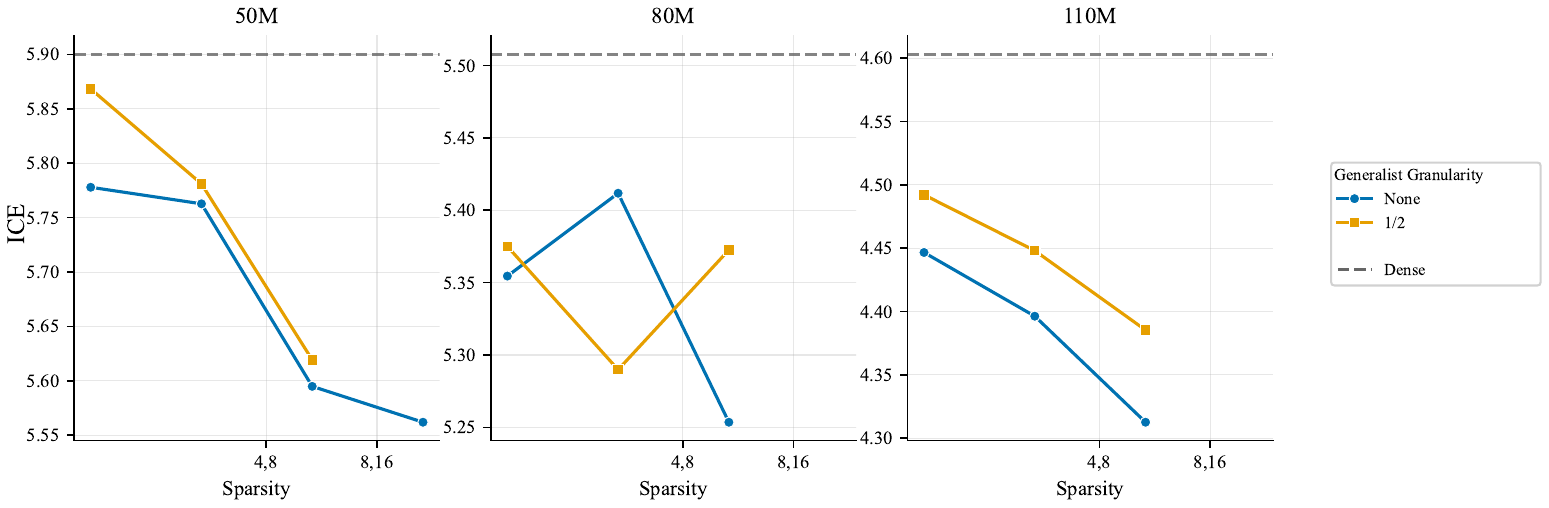}
    \end{subfigure}
    \caption{
    \textbf{The inclusion of a generalist consistently degrades performance in heterogeneous MoEs (\S\ref{sec:expt_hetgen}).}
    We train heterogeneous MoE LMs which consist of  routed experts with granularity $g_1, g_2$, as well as a generalist with granularity $g_{gen} = \frac{1}{2}$. We compare to settings with no generalist. In all settings and configurations, the addition of a generalist results in comparable or degraded performance. 
    }
    \label{fig:ice_hetgen}
\end{figure*}

\begin{figure*}[ht]
    \centering
    \begin{subfigure}[t]{\textwidth}
        \centering
        \begin{subfigure}[t]{0.45\textwidth}
            \includegraphics[width=\linewidth]{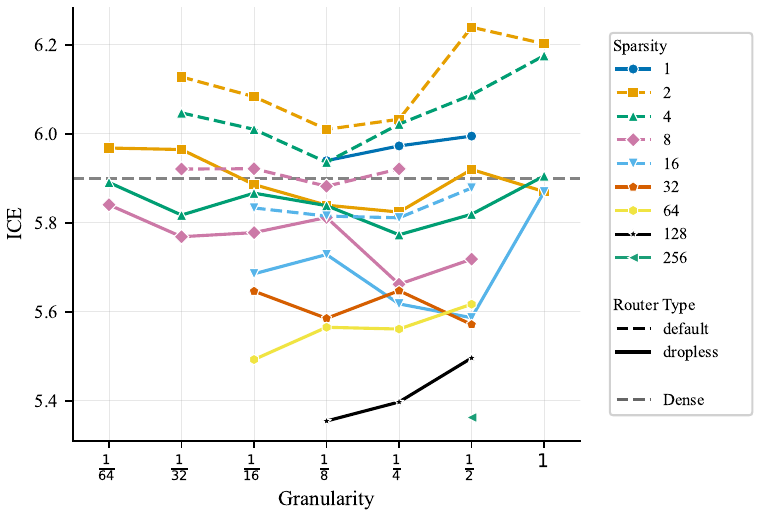}
            \caption{50M active, 50M - 930M total parameters}
        \end{subfigure}
    \hspace{1em}
        \begin{subfigure}[t]{0.45\textwidth}
            \centering
            \includegraphics[width=\linewidth]{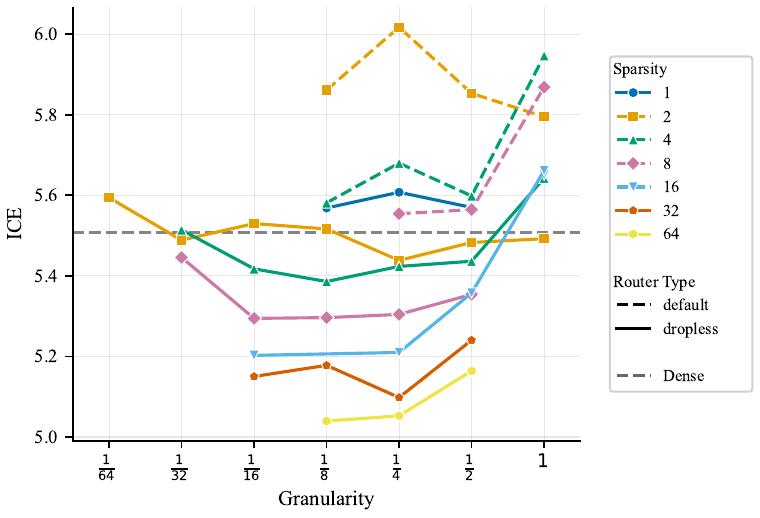}
            \caption{80M active, 80M - 765M total parameters}
        \end{subfigure}
    \end{subfigure}

    \par\bigskip\bigskip
    \begin{subfigure}[t]{0.45\textwidth}
        \centering
        \includegraphics[width=\linewidth]{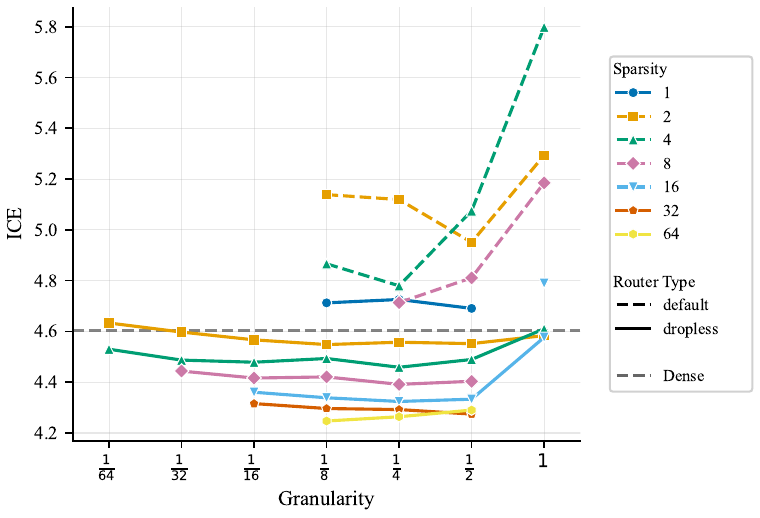}
        \caption{110M active, 110M - 1.4B total parameters}
    \end{subfigure}
    \caption{ 
    \textbf{Dropless routing outperforms default routing (\S\ref{sec:expt_router}).}
    We compare dropless routing to the default setting, which allow tokens to be dropped. Across all scales, we find that dropless routing outperforms or performs comparably to default routing. 
    }
    \label{fig:ice_dropless}
\end{figure*}

\begin{figure*}[ht]
    \centering
    \begin{subfigure}[t]{0.45\textwidth}
        \centering
        \includegraphics[width=\linewidth]{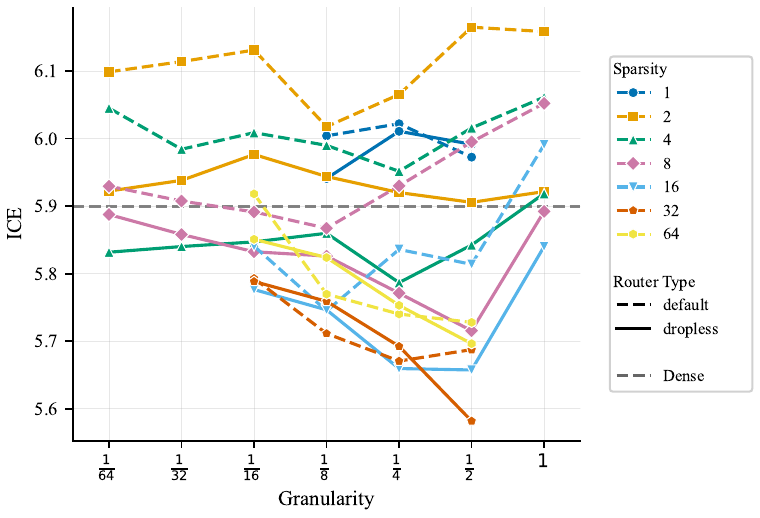}
        \caption{50M active, 50M - 930M total parameters}
    \end{subfigure}
    \hspace{1em}
    \begin{subfigure}[t]{0.45\textwidth}
        \centering
        \includegraphics[width=\linewidth]{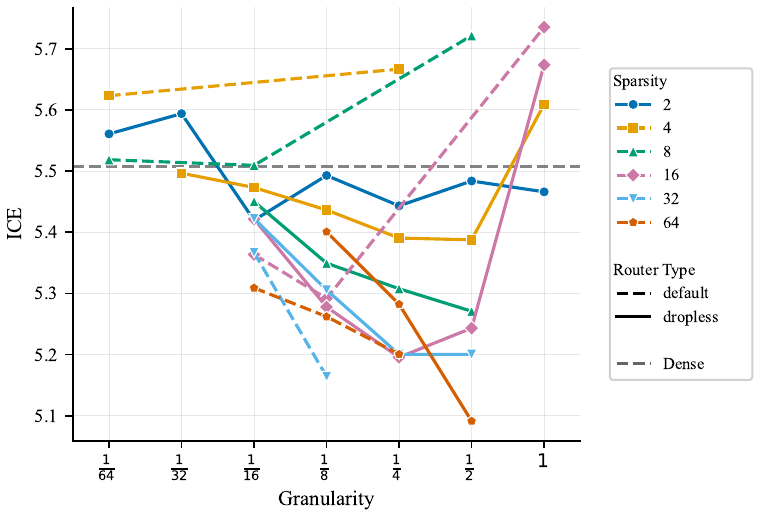}
        \caption{80M active, 80M - 765M total parameters}
    \end{subfigure}
    \caption{
    \textbf{Dropless routing, with bias $\gamma=\num{1e-3}$ (\S\ref{sec:expt_router}).} 
    As in Figure~\ref{fig:lm_avg_dropless}, we compare dropless routing to the default setting, which allow tokens to be dropped. Across all scales, we find that dropless routing outperforms or performs comparably to default routing. We see here with additional higher sparsity default routing runs that as sparsity increases, default routing performance approaches that of dropless routing.
    }
    \label{fig:ice_dropless_with_lf}
\end{figure*}

\begin{figure*}[ht]
    \centering
    \begin{subfigure}[]{\textwidth}
        \centering
        \includegraphics[width=0.46\linewidth]{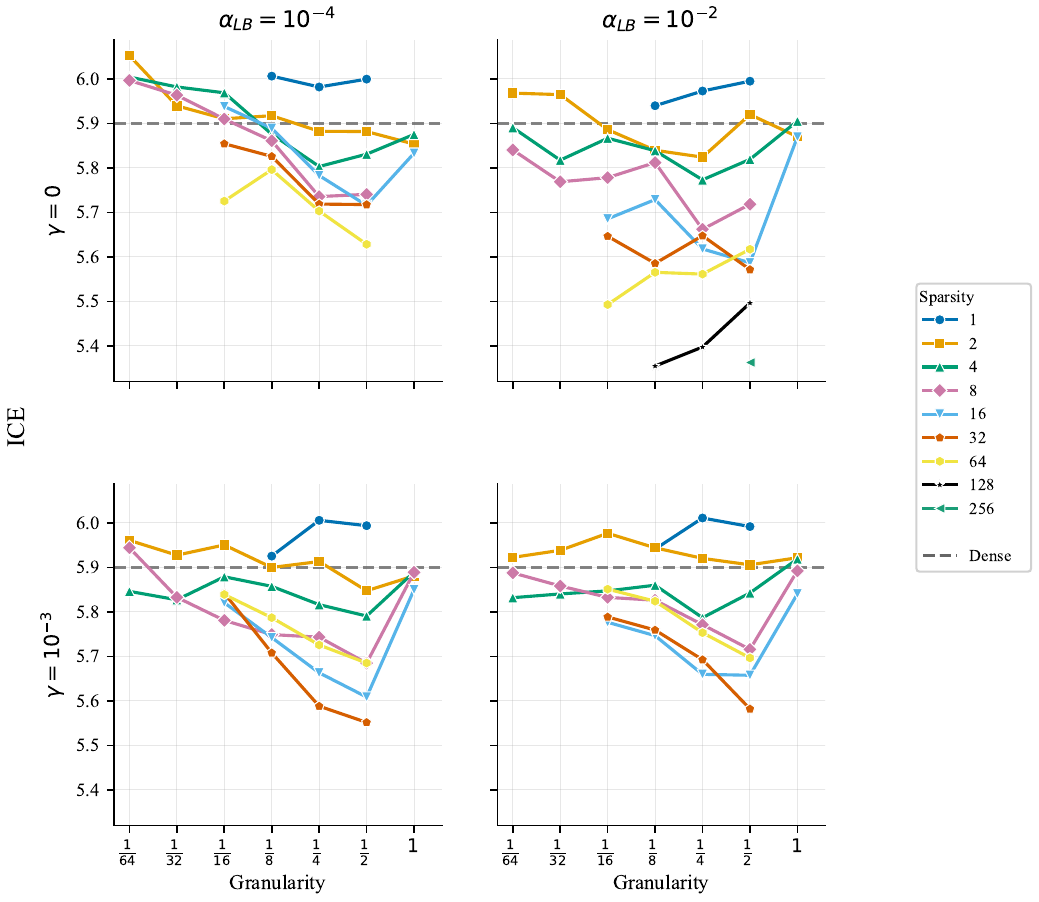}
        \hspace{1em}
        \includegraphics[width=0.46\linewidth]{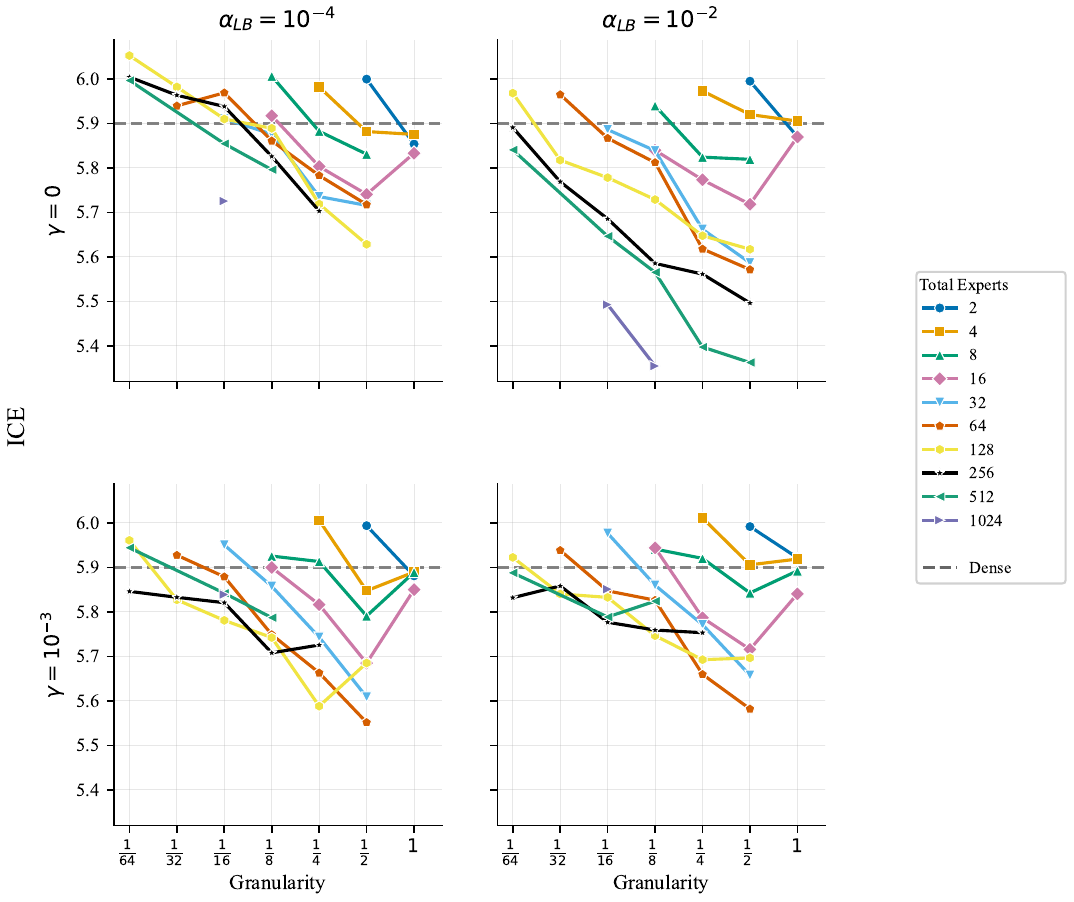}
        \caption{50M active, 50M - 930M total parameters}
    \end{subfigure}
    \par\bigskip\bigskip
    \begin{subfigure}[]{\textwidth}
        \centering
        \includegraphics[width=0.46\linewidth]{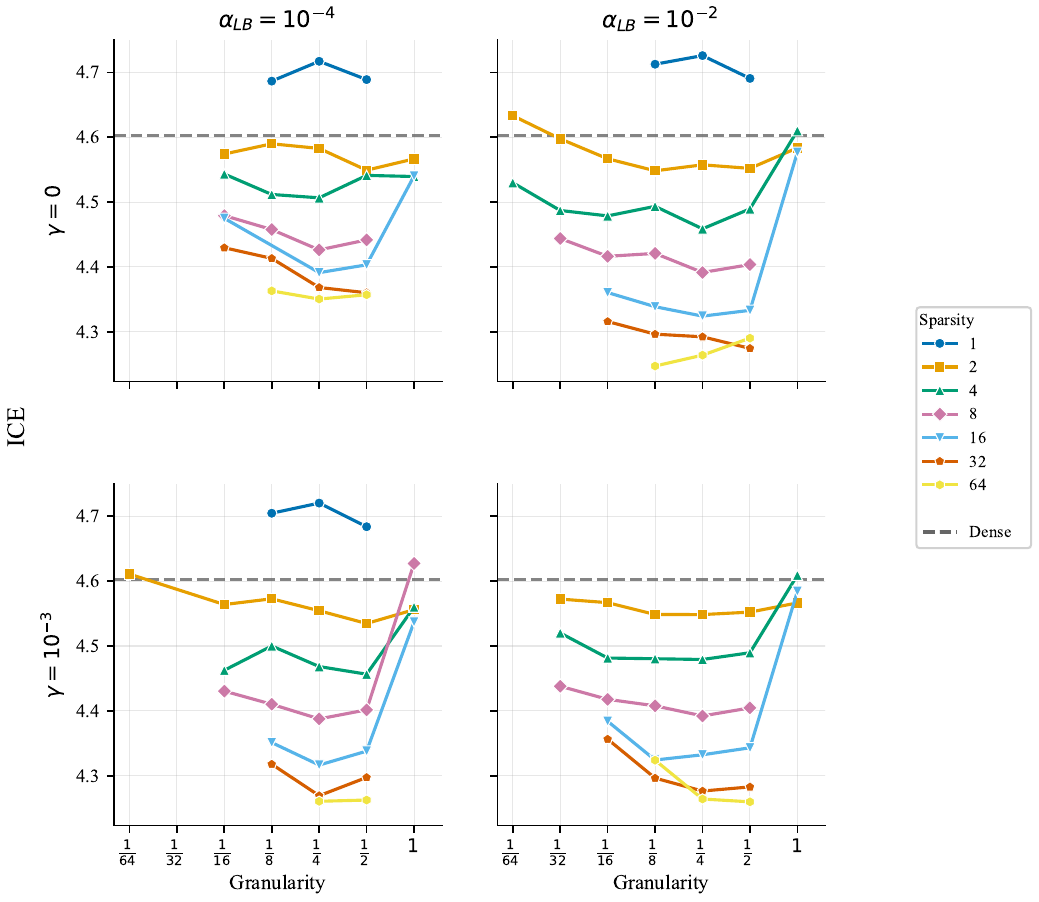}
        \hspace{1em}
        \includegraphics[width=0.46\linewidth]{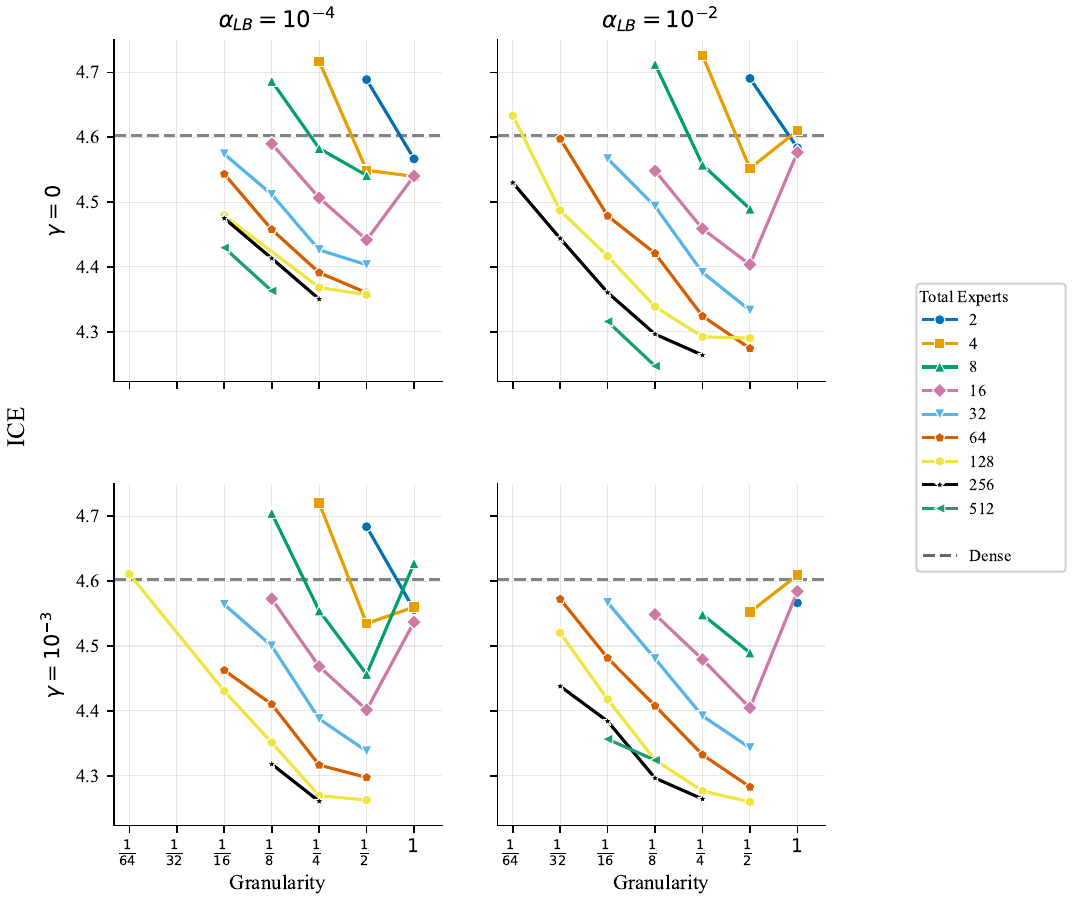}
        \caption{80M active, 80M - 765M total parameters}
    \end{subfigure}
    \par\bigskip\bigskip
    \begin{subfigure}[t]{\textwidth}
        \centering
        \includegraphics[width=0.46\linewidth]{figures/lm/ice-validation/ce_loss/lb_sweep_hgn_gxs_110M.pdf}
        \hspace{1em}
        \includegraphics[width=0.46\linewidth]{figures/lm/ice-validation/ce_loss/lb_sweep_hgn_gxn_110M.pdf}
        \caption{110M active, 110M - 1.4B total parameters}
    \end{subfigure}

    \end{figure*} 

\clearpage  

\begin{figure*}[ht]
    \addtocounter{figure}{-1}
    \centering
    \begin{subfigure}[t]{\textwidth}
        \centering
        \includegraphics[width=0.46\linewidth]{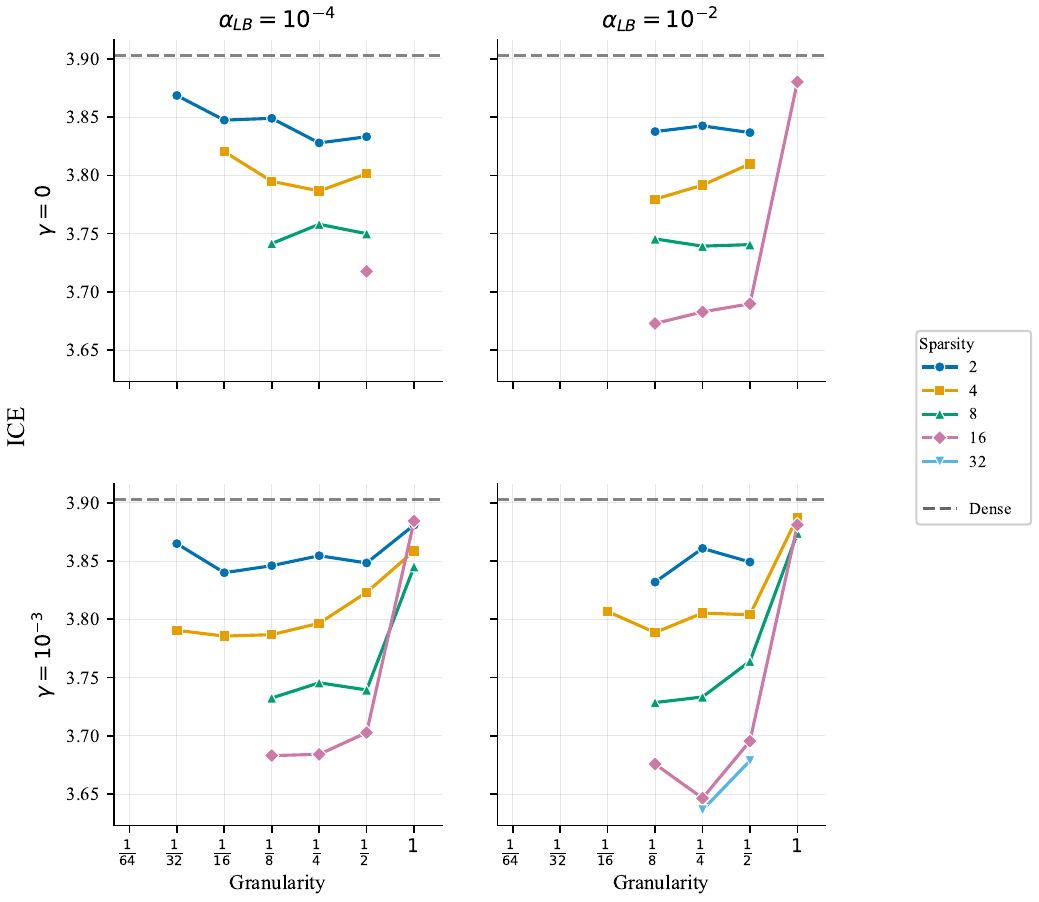}
        \hspace{1em}
        \includegraphics[width=0.46\linewidth]{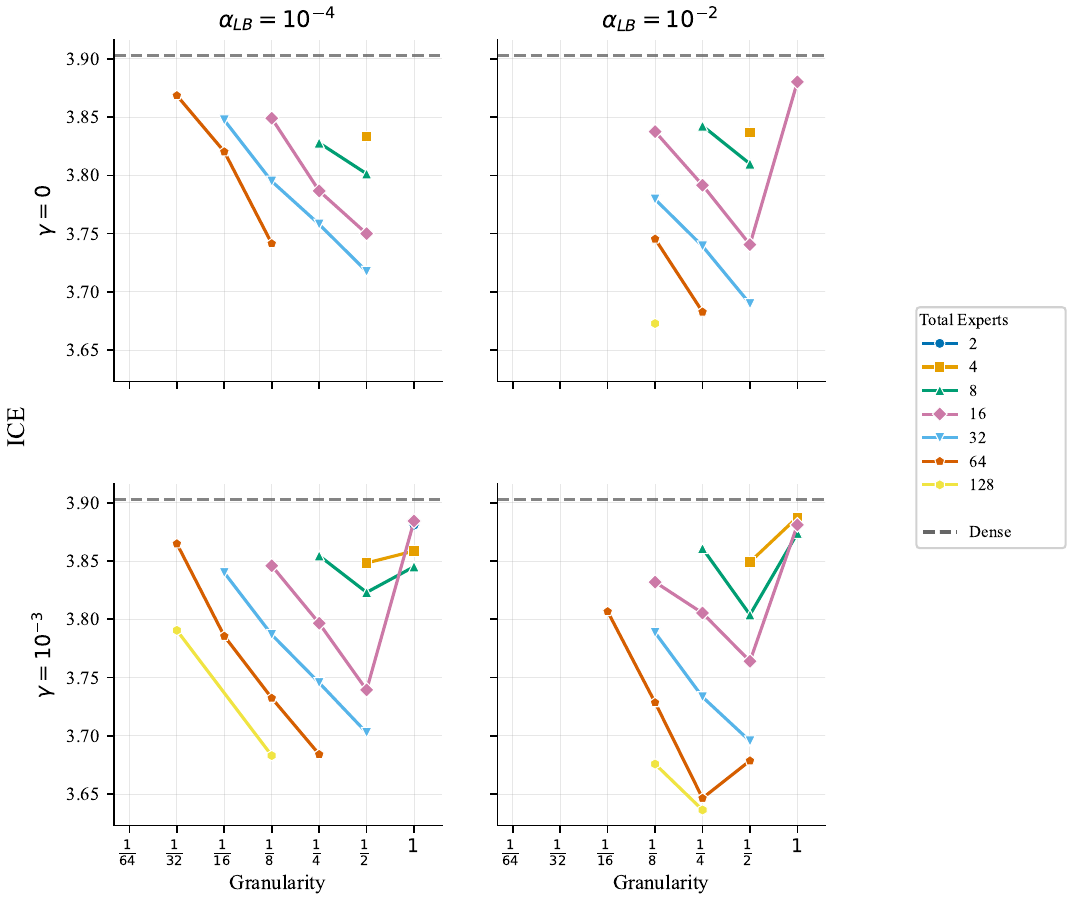}
        \caption{200M active, 200M - 3.3B total parameters}
    \end{subfigure}
    \par\bigskip\bigskip
    \begin{subfigure}[t]{\textwidth}
        \centering
        \includegraphics[width=0.3\linewidth]{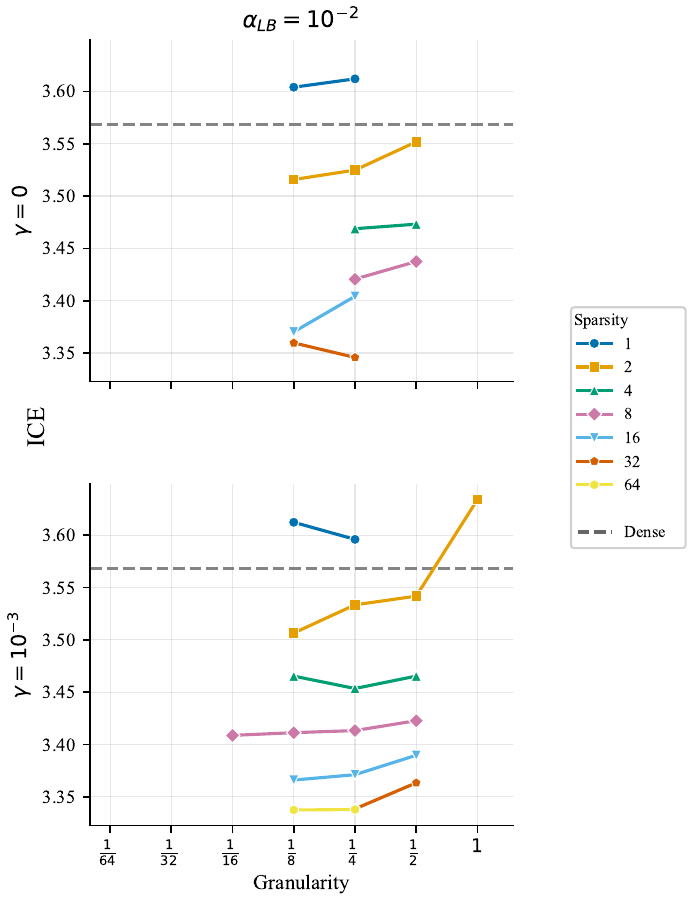}
        \hspace{1em}
        \includegraphics[width=0.3\linewidth]{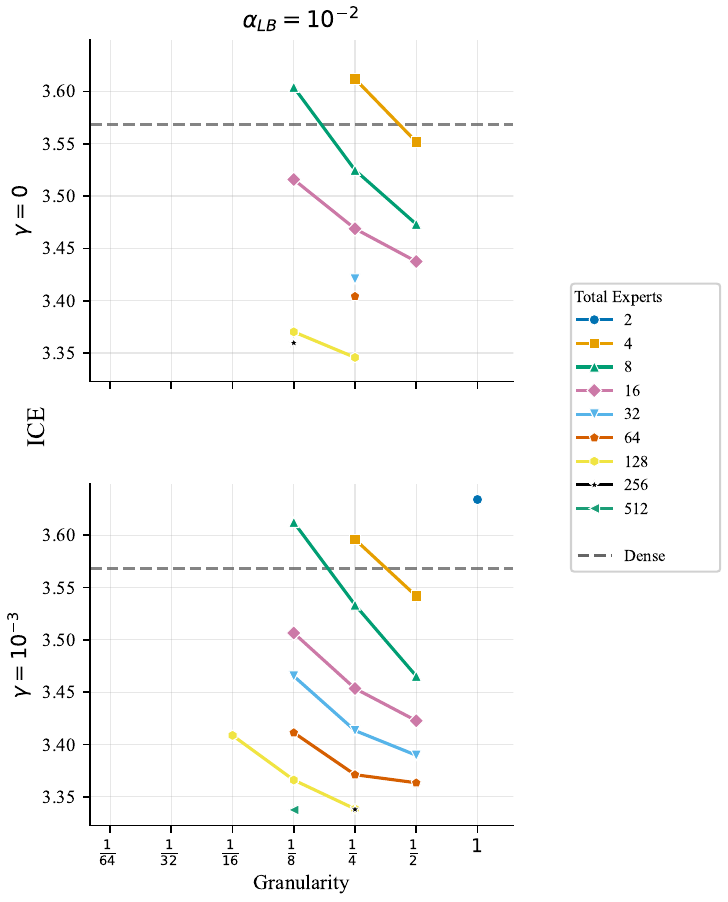}
        \caption{300M active, 300M - 6.6B total parameters}
    \end{subfigure}

    \caption{
    \textbf{Load balancing mechanisms must be tuned correctly (\S\ref{sec:expt_router}).}
    We consider load balancing loss weight $\alpha_{LB} \in \{\num{1e-2}, \num{1e-4}\}$ and loss-free load balancing with bias $\gamma\in\{0, \num{1e-3}\}$ ($\gamma=0$ indicates no loss-free mechanism). Results show that poorly chosen hyperparameters, such as high bias $\gamma = 1e-3$ with total experts $n\geq 512$, may impair performance. However, all settings other than $(\alpha_{LB}=\num{1e-2}, \gamma=\num{1e-3})$ perform comparably for $n \leq 512$, suggesting that a wide range of load balancing settings achieve near-optimal performance. 
    }
    \label{fig:ice_lb}
\end{figure*}

%% file: fig_tex/lm/m2d2_s2orc.tex
\begin{figure*}[!ht]
    \centering
        \begin{subfigure}[t]{\textwidth}
        \begin{subfigure}[t]{0.33\textwidth}
            \centering
            \caption*{\scriptsize Fixed total experts (n)}
            \includegraphics[width=\linewidth]{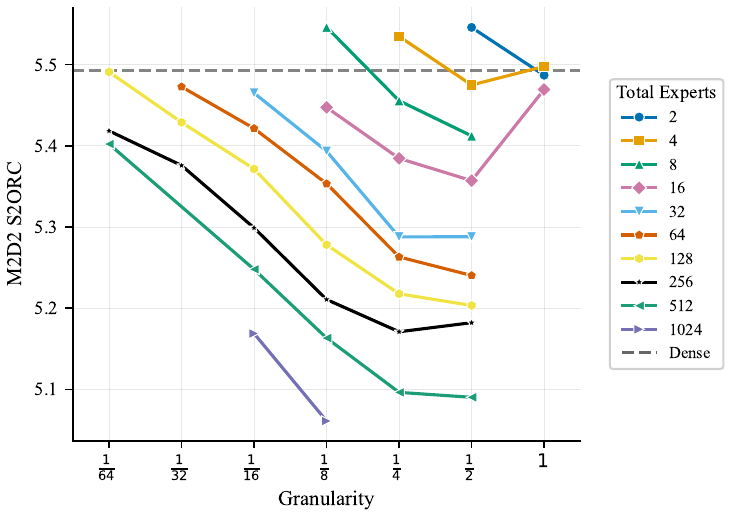}
        \end{subfigure}
        \begin{subfigure}[t]{0.33\textwidth}
            \centering
            \caption*{\scriptsize Fixed granularity (g)}
            \includegraphics[width=\linewidth]{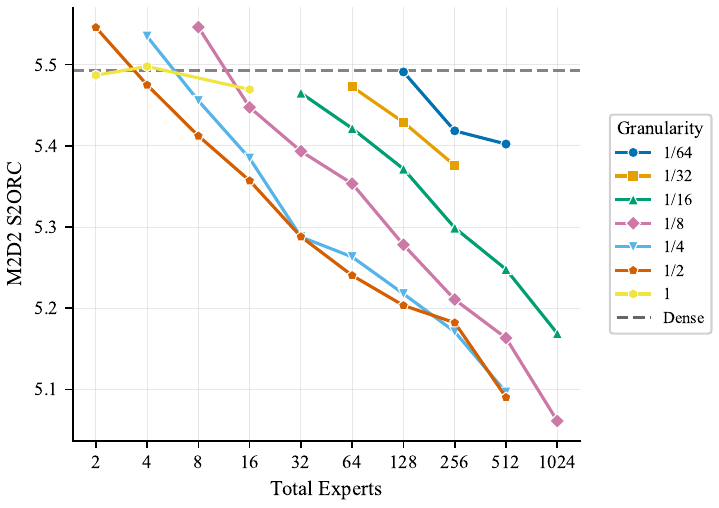}
        \end{subfigure}
        \begin{subfigure}[t]{0.33\textwidth}
            \centering
            \caption*{\scriptsize Fixed activation sparsity (s)}
            \includegraphics[width=\linewidth]{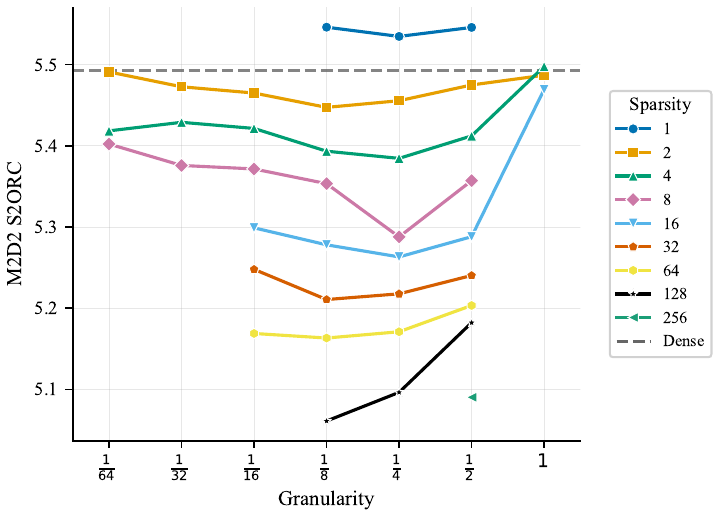}
        \end{subfigure}
        \caption{50M active, 50M - 930M total parameters}
    \end{subfigure}
\par\bigskip\bigskip
    \begin{subfigure}[t]{\textwidth}
        \begin{subfigure}[t]{0.33\textwidth}
            \centering
            \includegraphics[width=\linewidth]{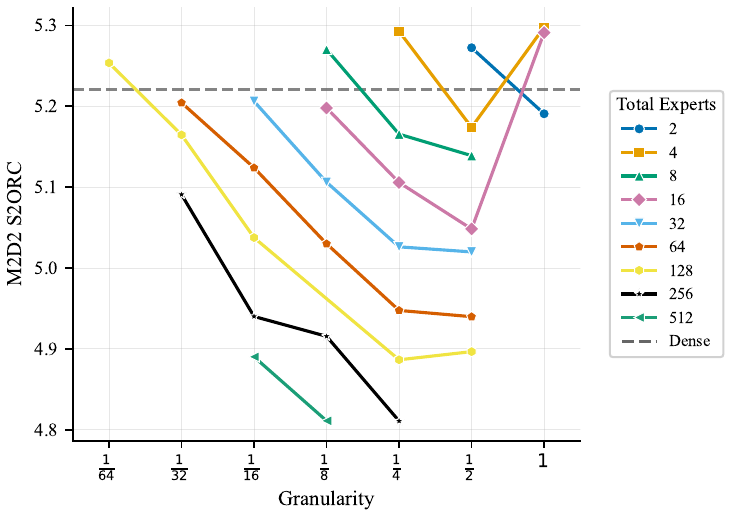}
        \end{subfigure}
        \begin{subfigure}[t]{0.33\textwidth}
            \centering
            \includegraphics[width=\linewidth]{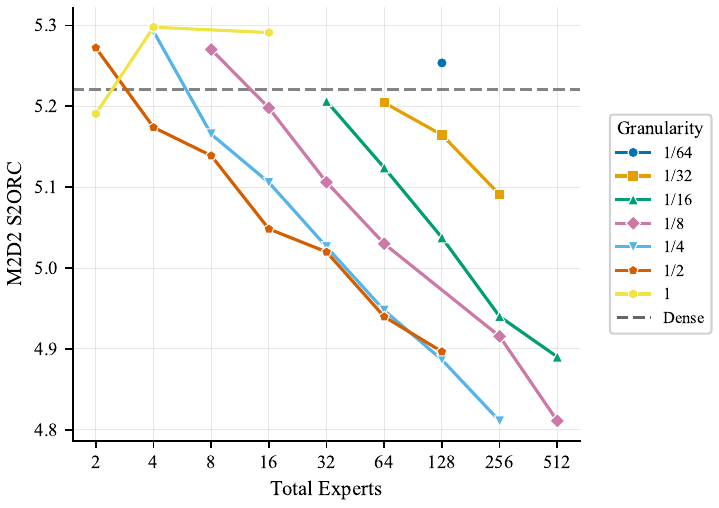}
        \end{subfigure}
        \begin{subfigure}[t]{0.33\textwidth}
            \centering
            \includegraphics[width=\linewidth]{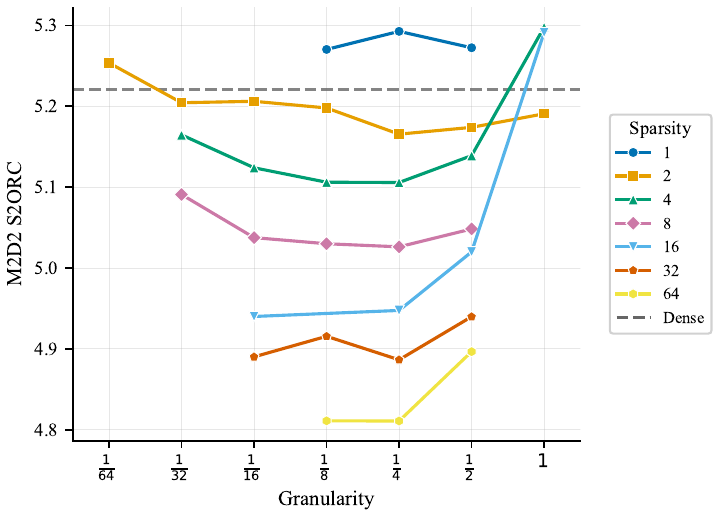}
        \end{subfigure}
        \caption{80M active, 80M - 765M total parameters}
    \end{subfigure}
    \par\bigskip\bigskip
        \begin{subfigure}[t]{\textwidth}
        \begin{subfigure}[t]{0.33\textwidth}
            \centering
            \includegraphics[width=\linewidth]{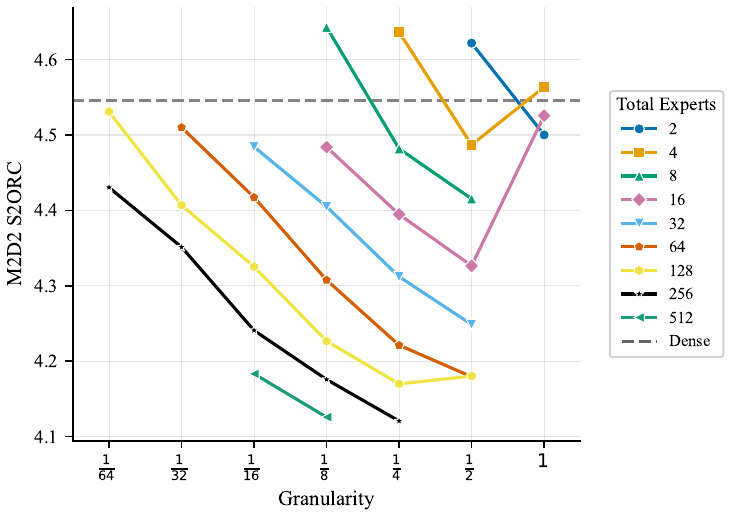}
        \end{subfigure}
        \begin{subfigure}[t]{0.33\textwidth}
            \centering
            \includegraphics[width=\linewidth]{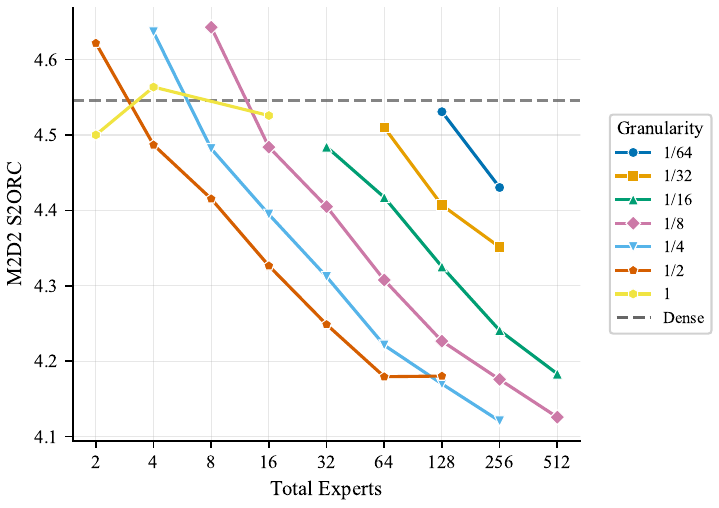}
        \end{subfigure}
        \begin{subfigure}[t]{0.33\textwidth}
            \centering
            \includegraphics[width=\linewidth]{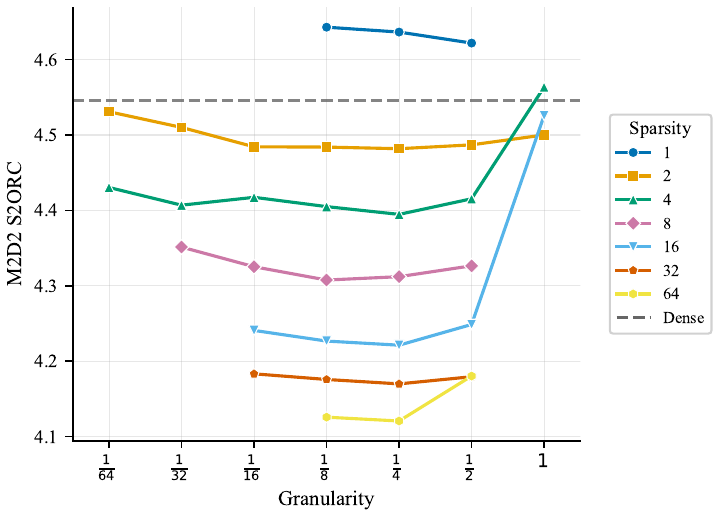}
        \end{subfigure}
        \caption{110M active, 110M - 1.4B total parameters}
    \end{subfigure}
    \end{figure*}

\clearpage  

\begin{figure*}[!ht]
        \addtocounter{figure}{-1}
    \begin{subfigure}[t]{\textwidth}
        \addtocounter{subfigure}{3}
        \begin{subfigure}[t]{0.33\textwidth}
            \centering
            \caption*{\scriptsize Fixed total experts (n)}
            \includegraphics[width=\linewidth]{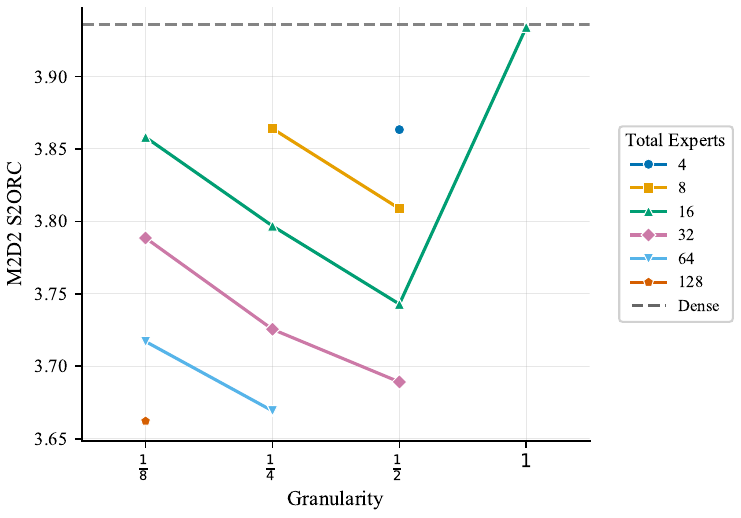}
        \end{subfigure}
        \begin{subfigure}[t]{0.33\textwidth}
            \centering
            \caption*{\scriptsize Fixed granularity (g)}
            \includegraphics[width=\linewidth]{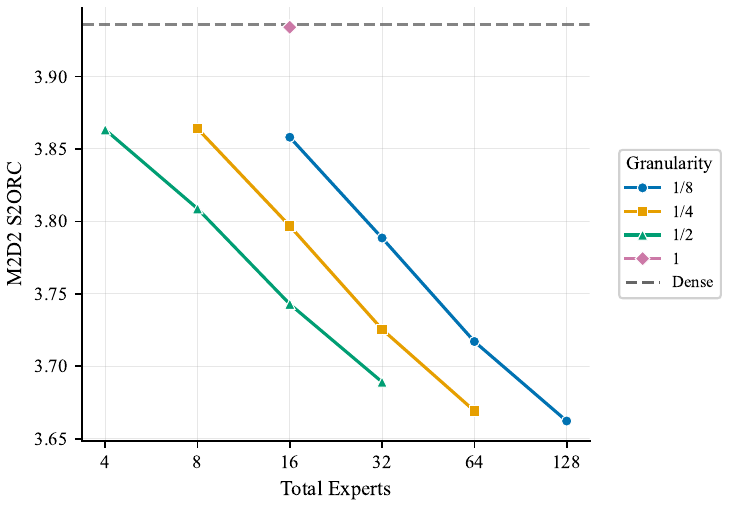}
        \end{subfigure}
        \begin{subfigure}[t]{0.33\textwidth}
            \centering
            \caption*{\scriptsize Fixed activation sparsity (s)}
            \includegraphics[width=\linewidth]{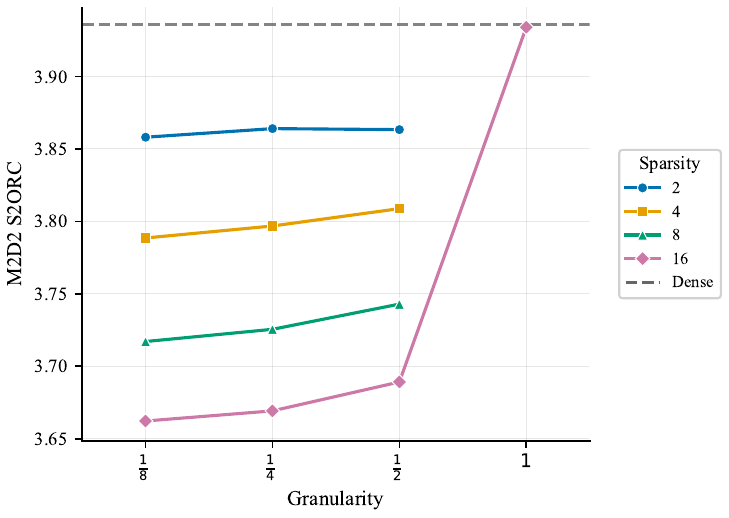}
        \end{subfigure}
        \caption{200M active, 200M - 3.3B total parameters}
    \end{subfigure}
    \par\bigskip\bigskip
        \begin{subfigure}[t]{\textwidth}
        \begin{subfigure}[t]{0.33\textwidth}
            \centering
            \includegraphics[width=\linewidth]{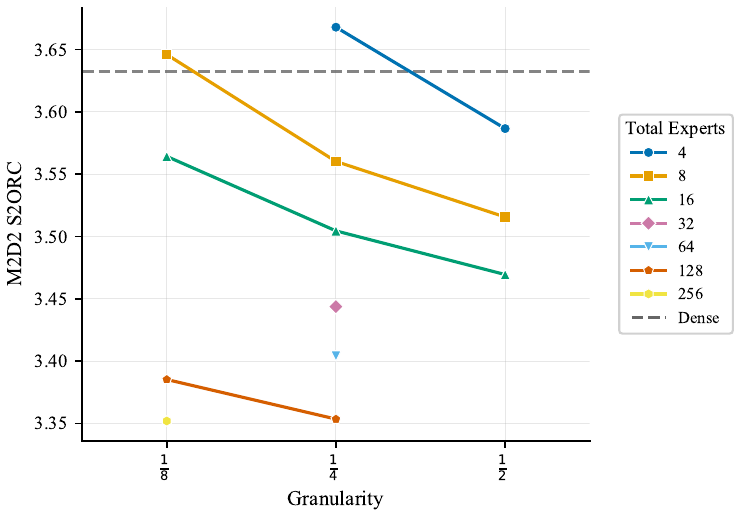}
        \end{subfigure}
        \begin{subfigure}[t]{0.33\textwidth}
            \centering
            \includegraphics[width=\linewidth]{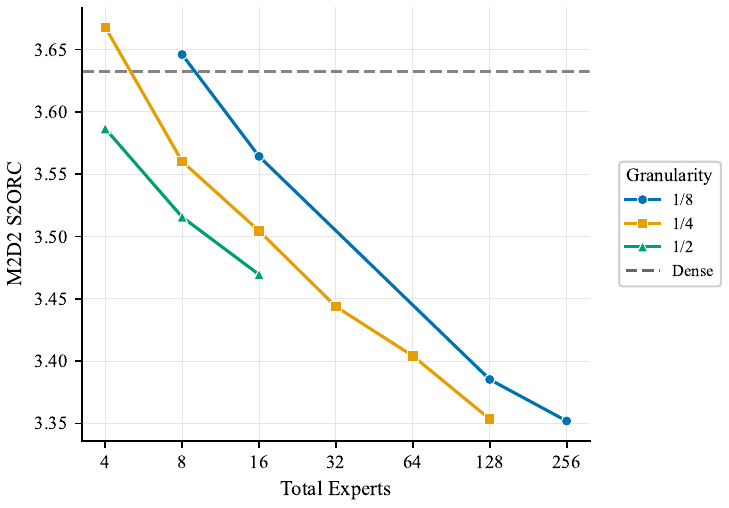}
        \end{subfigure}
        \begin{subfigure}[t]{0.33\textwidth}
            \centering
            \includegraphics[width=\linewidth]{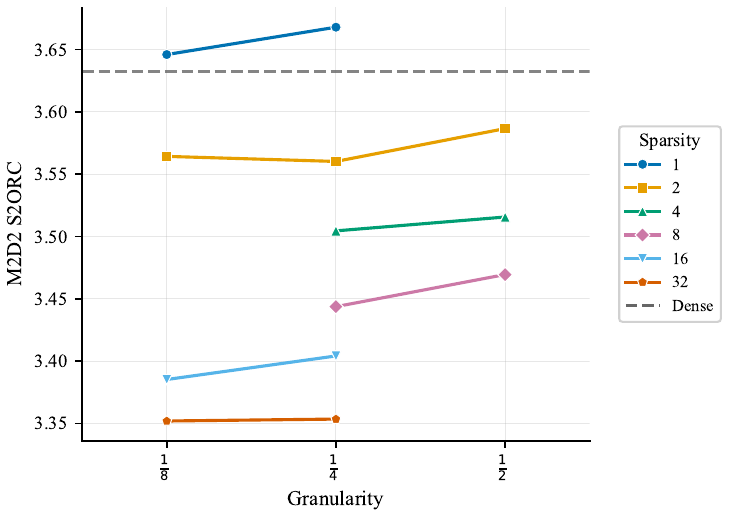}
        \end{subfigure}
        \caption{300M active, 300M - 6.6B total parameters}
    \end{subfigure}

    \caption{
    \textbf{Increasing inactive expert parameters via expert size (left) or total count (center) improves performance in MoEs (\S\ref{sec:expt_main}).} This effect is seen both when holding total number of experts fixed (left) and when holding expert granularity fixed (center). In general, increasing total parameters results in improved performance.  \textbf{Optimal tradeoff between expert count and granularity varies in MoEs (right). (\S\ref{sec:expt_main})}
    At each activation sparsity $s$ (equivalently, at each total parameter count), the optimal (total expert count, expert granularity) configuration varies. As $s$ increases, optimal expert granularity remains nearly fixed, suggesting that sparsity should be scaled up primarily by increasing total expert count $n$, while maintaining a near constant, slowly increasing expert granularity $g$. 
    }
    \label{fig:m2d2_s2orc_experts}
\end{figure*}

\begin{figure*}[!ht]
    \centering
    
    \begin{subfigure}[t]{0.46\textwidth}
        \centering
        \includegraphics[width=\linewidth]{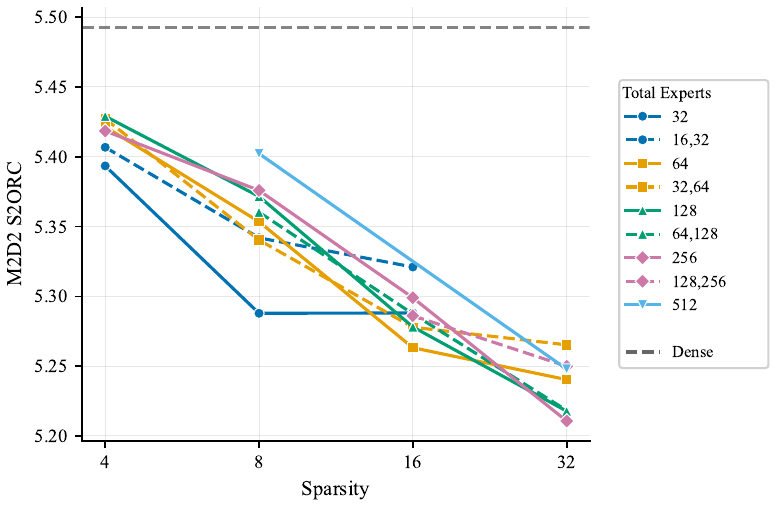}
        \caption{50M active, 50M - 930M total parameters}
    \end{subfigure}
    \vspace{1em}
    \begin{subfigure}[t]{0.46\textwidth}
        \centering
        \includegraphics[width=\linewidth]{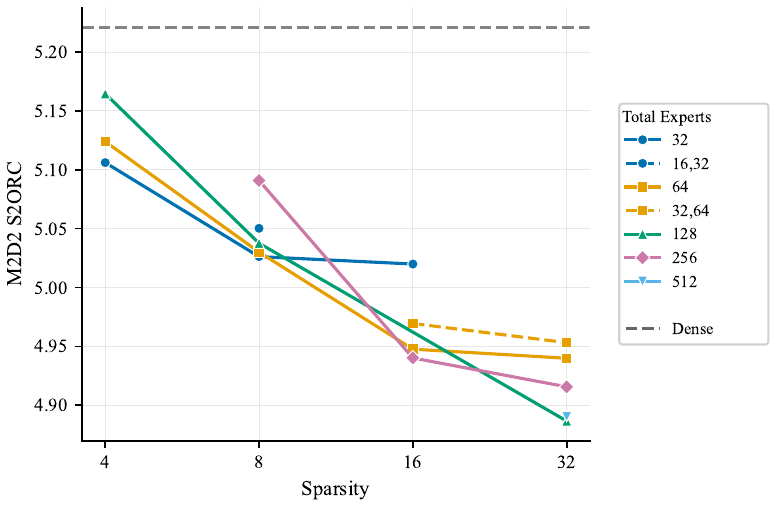}
        \caption{80M active, 80M - 765M total parameters}
    \end{subfigure}
    \caption{
    \textbf{Heterogeneity of expert size alone does not improve MoE performance (\S\ref{sec:expt_hetgen}).} To explore the potential benefits of their architectural flexibility, we compare heterogeneous MoEs (indicated by dotted lines) to active- and total-parameter-matched homogeneous MoEs. Heterogeneity alone does not result in performance gains, as, at each activation sparsity $s$, heterogeneous MoEs with $n_1, n_2 = a, b$ lie between or near the 2 closest homogeneous MoEs, with $n=a$ and with $n=b$.
    }
    \label{fig:m2d2_s2orc_het}
\end{figure*}

\begin{figure*}[!ht]
    \centering
    
    \begin{subfigure}[t]{1.0\textwidth}
        \centering
        \includegraphics[width=\linewidth]{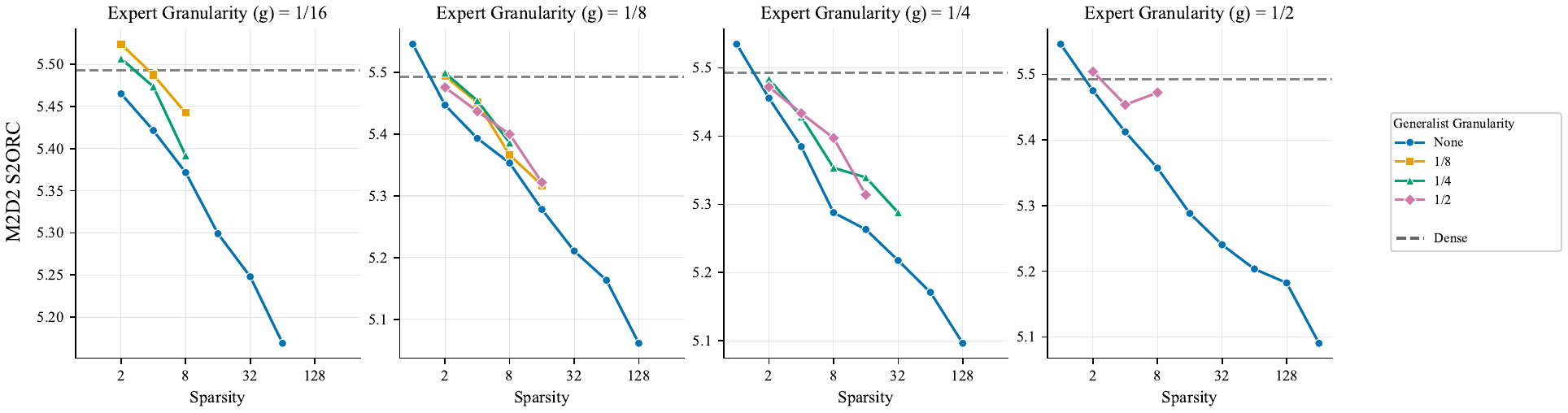}
        \caption{50M active, 50M - 930M total parameters}
    \end{subfigure}
    \par\bigskip\bigskip
    \begin{subfigure}[t]{1.0\textwidth}
        \centering
        \includegraphics[width=\linewidth]{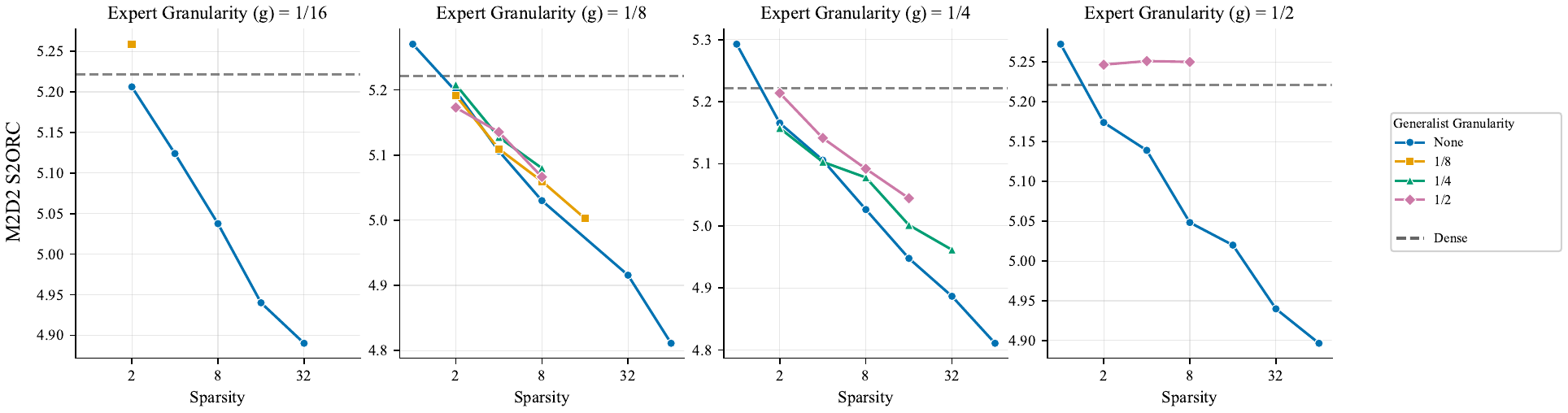}
        \caption{80M active, 80M - 765M total parameters}
    \end{subfigure}
    \par\bigskip\bigskip
    \begin{subfigure}[t]{1.0\textwidth}
        \centering
        \includegraphics[width=\linewidth]{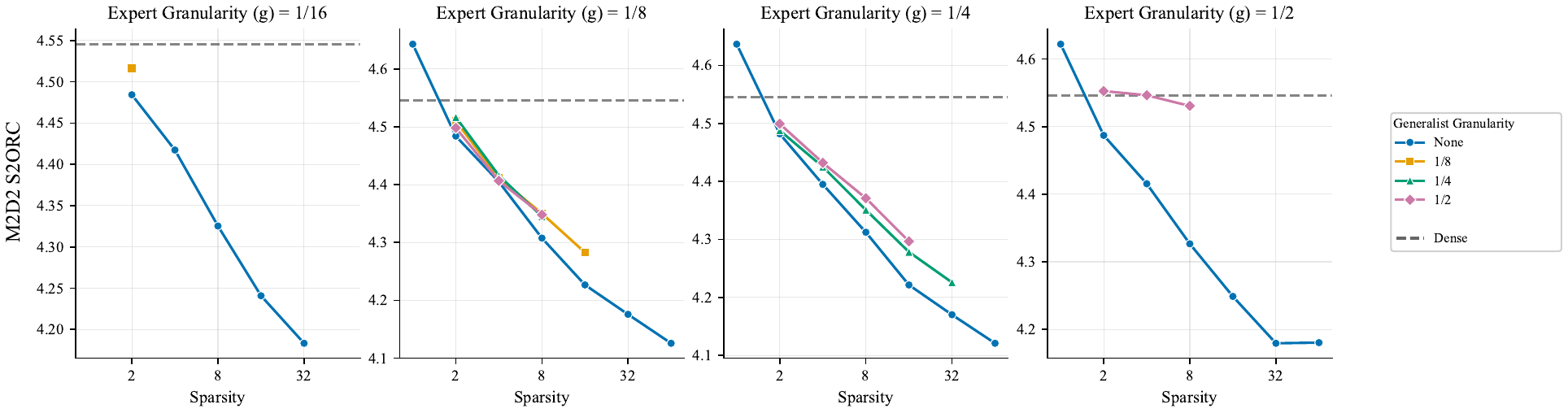}
        \caption{110M active, 110M - 1.4B total parameters}
    \end{subfigure}
    \caption{
    \textbf{The inclusion of a generalist consistently degrades performance in homogeneous MoEs (\S\ref{sec:expt_hetgen}).}
    We train MoE LMs which consist of some routed experts with granularity $g$, as well as a generalist with granularity $g_{gen}\in \{\frac{1}{2}, \frac{1}{4}, \frac{1}{8}\} $. We compare to settings with no generalist, only routed experts with granularity $g$. In all settings and configurations, the addition of any granularity generalist results in comparable or degraded performance. 
    }
    \label{fig:m2d2_s2orc_gen}
\end{figure*}

\begin{figure*}[ht]
    \centering
    \begin{subfigure}[t]{1.0\textwidth}
        \centering
        \includegraphics[width=\linewidth]{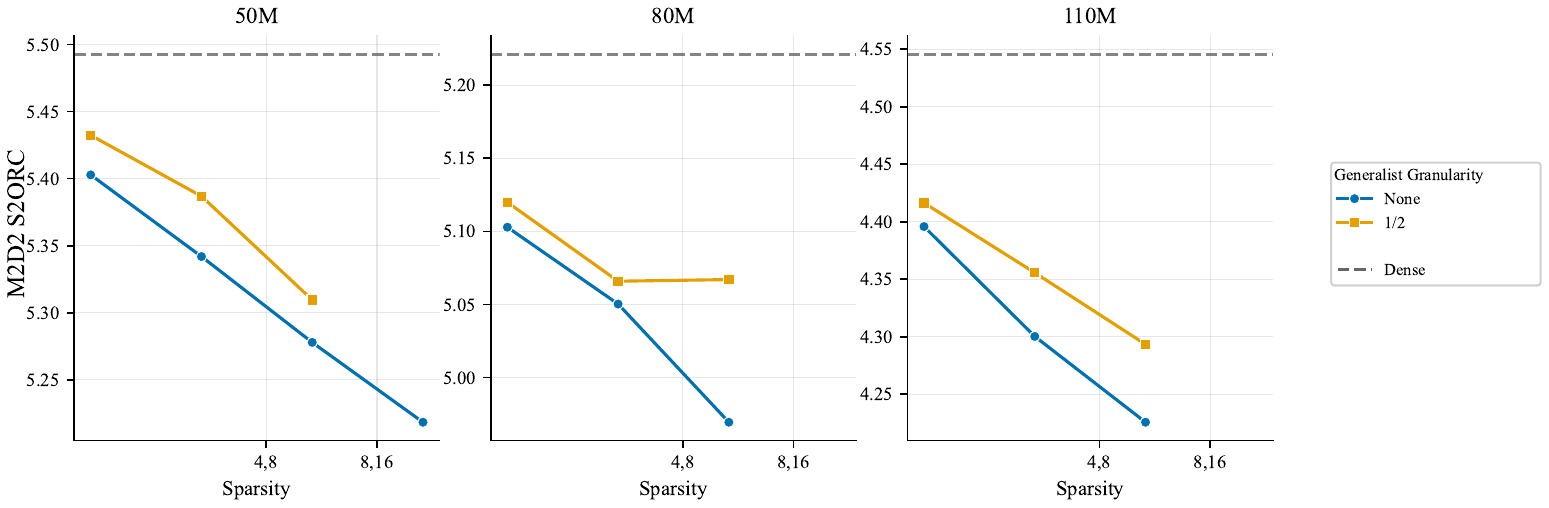}
    \end{subfigure}
    \caption{
    \textbf{The inclusion of a generalist consistently degrades performance in heterogeneous MoEs (\S\ref{sec:expt_hetgen}).}
    We train heterogeneous MoE LMs which consist of  routed experts with granularity $g_1, g_2$, as well as a generalist with granularity $g_{gen} = \frac{1}{2}$. We compare to settings with no generalist. In all settings and configurations, the addition of a generalist results in comparable or degraded performance. 
    }
    \label{fig:m2d2_s2orc_hetgen}
\end{figure*}

\begin{figure*}[ht]
    \centering
    \begin{subfigure}[t]{\textwidth}
        \centering
        \begin{subfigure}[t]{0.45\textwidth}
            \includegraphics[width=\linewidth]{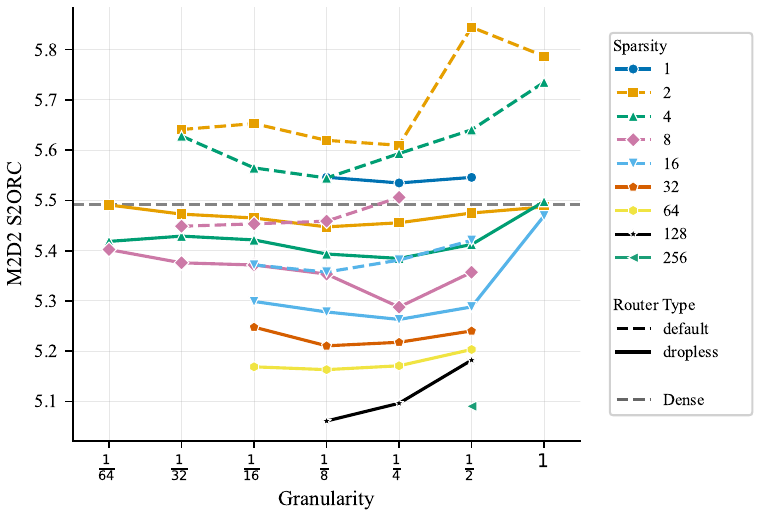}
            \caption{50M active, 50M - 930M total parameters}
        \end{subfigure}
    \hspace{1em}
        \begin{subfigure}[t]{0.45\textwidth}
            \centering
            \includegraphics[width=\linewidth]{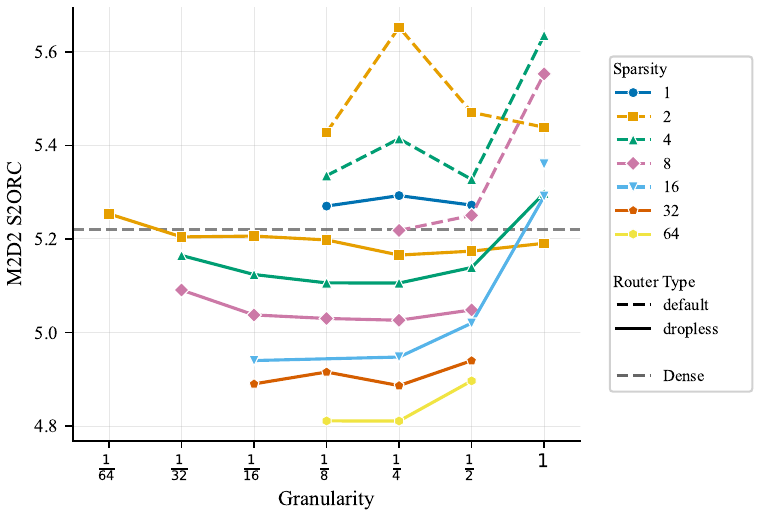}
            \caption{80M active, 80M - 765M total parameters}
        \end{subfigure}
    \end{subfigure}

    \par\bigskip\bigskip
    \begin{subfigure}[t]{0.45\textwidth}
        \centering
        \includegraphics[width=\linewidth]{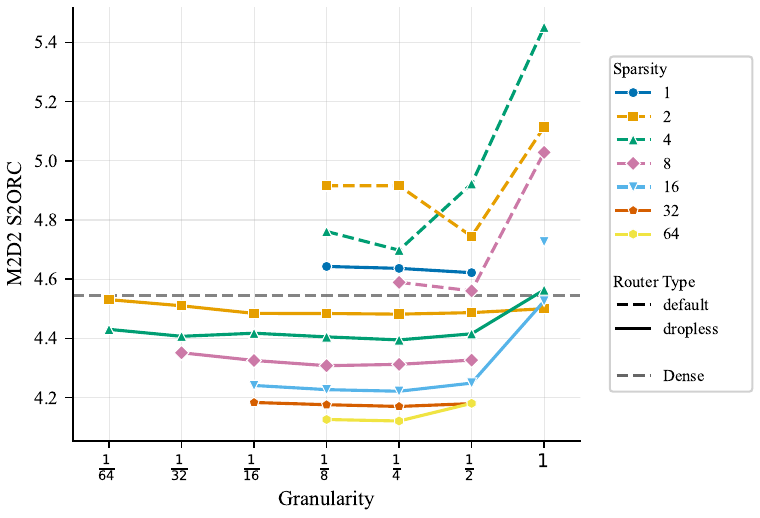}
        \caption{110M active, 110M - 1.4B total parameters}
    \end{subfigure}
    \caption{ 
    \textbf{Dropless routing outperforms default routing (\S\ref{sec:expt_router}).}
    We compare dropless routing to the default setting, which allow tokens to be dropped. Across all scales, we find that dropless routing outperforms or performs comparably to default routing. 
    }
    \label{fig:m2d2_s2orc_dropless}
\end{figure*}

\begin{figure*}[ht]
    \centering
    \begin{subfigure}[t]{0.45\textwidth}
        \centering
        \includegraphics[width=\linewidth]{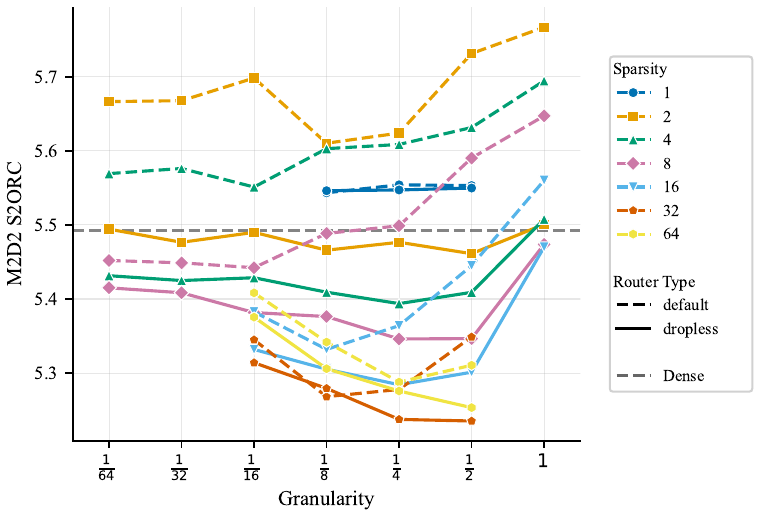}
        \caption{50M active, 50M - 930M total parameters}
    \end{subfigure}
    \hspace{1em}
    \begin{subfigure}[t]{0.45\textwidth}
        \centering
        \includegraphics[width=\linewidth]{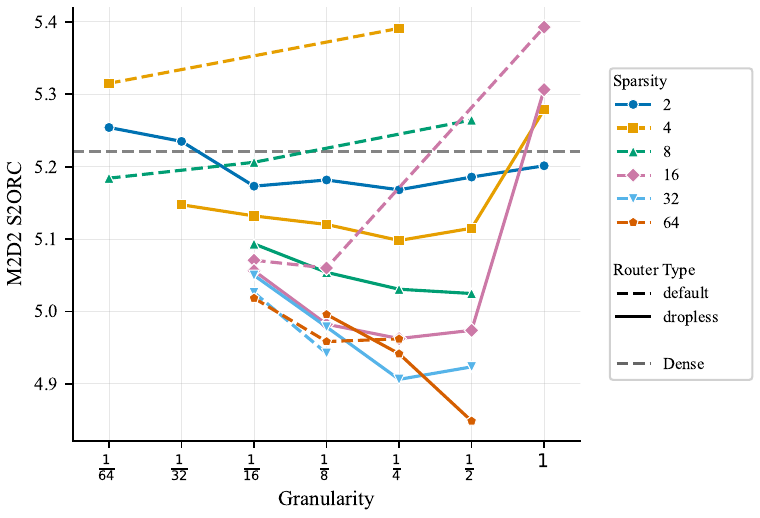}
        \caption{80M active, 80M - 765M total parameters}
    \end{subfigure}
    \caption{
    \textbf{Dropless routing, with bias $\gamma=\num{1e-3}$ (\S\ref{sec:expt_router}).} 
    As in Figure~\ref{fig:lm_avg_dropless}, we compare dropless routing to the default setting, which allow tokens to be dropped. Across all scales, we find that dropless routing outperforms or performs comparably to default routing. We see here with additional higher sparsity default routing runs that as sparsity increases, default routing performance approaches that of dropless routing.
    }
    \label{fig:m2d2_s2orc_dropless_with_lf}
\end{figure*}

\begin{figure*}[ht]
    \centering
    \begin{subfigure}[]{\textwidth}
        \centering
        \includegraphics[width=0.46\linewidth]{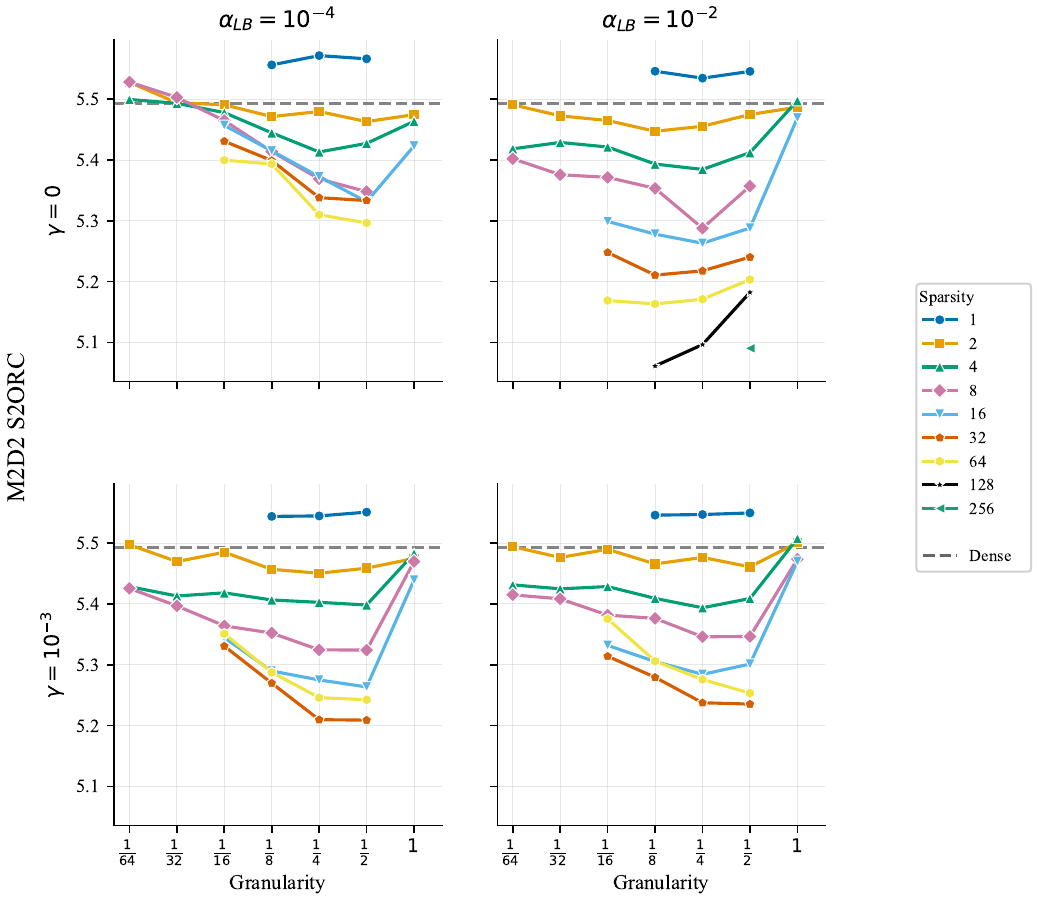}
        \hspace{1em}
        \includegraphics[width=0.46\linewidth]{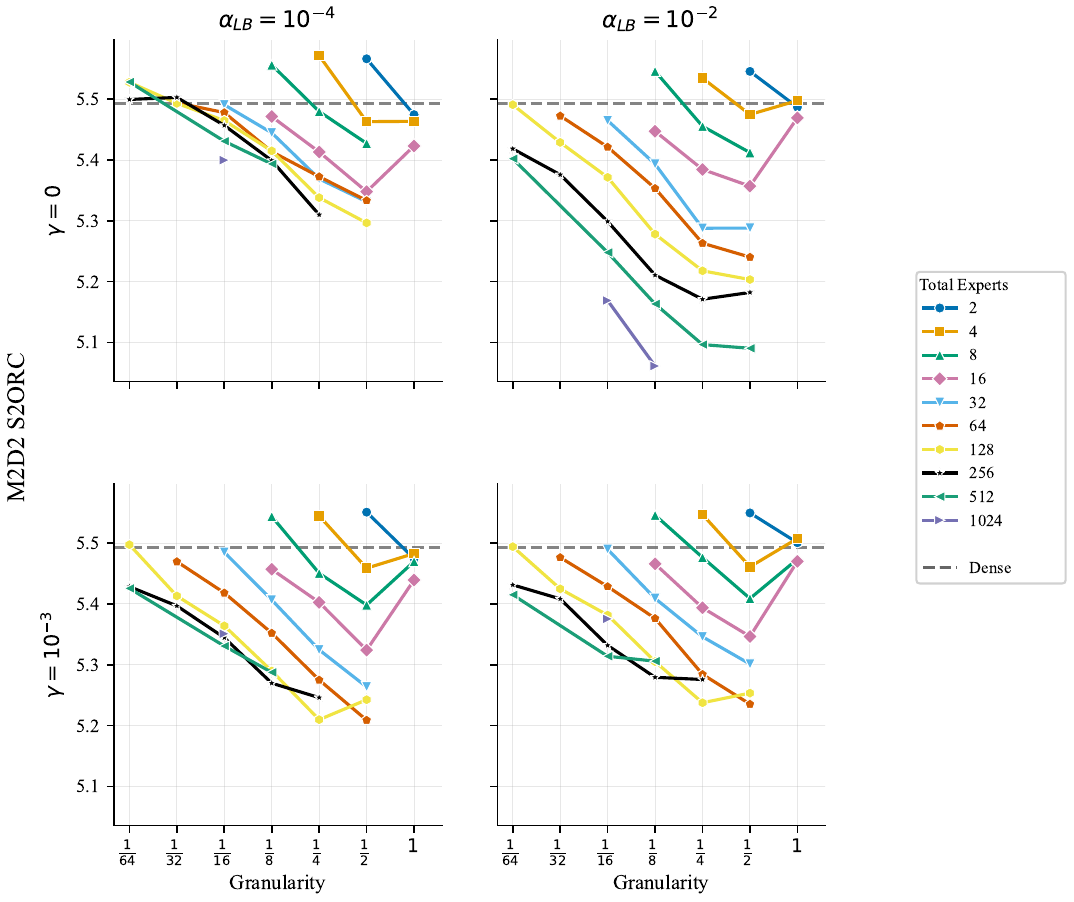}
        \caption{50M active, 50M - 930M total parameters}
    \end{subfigure}
    \par\bigskip\bigskip
    \begin{subfigure}[]{\textwidth}
        \centering
        \includegraphics[width=0.46\linewidth]{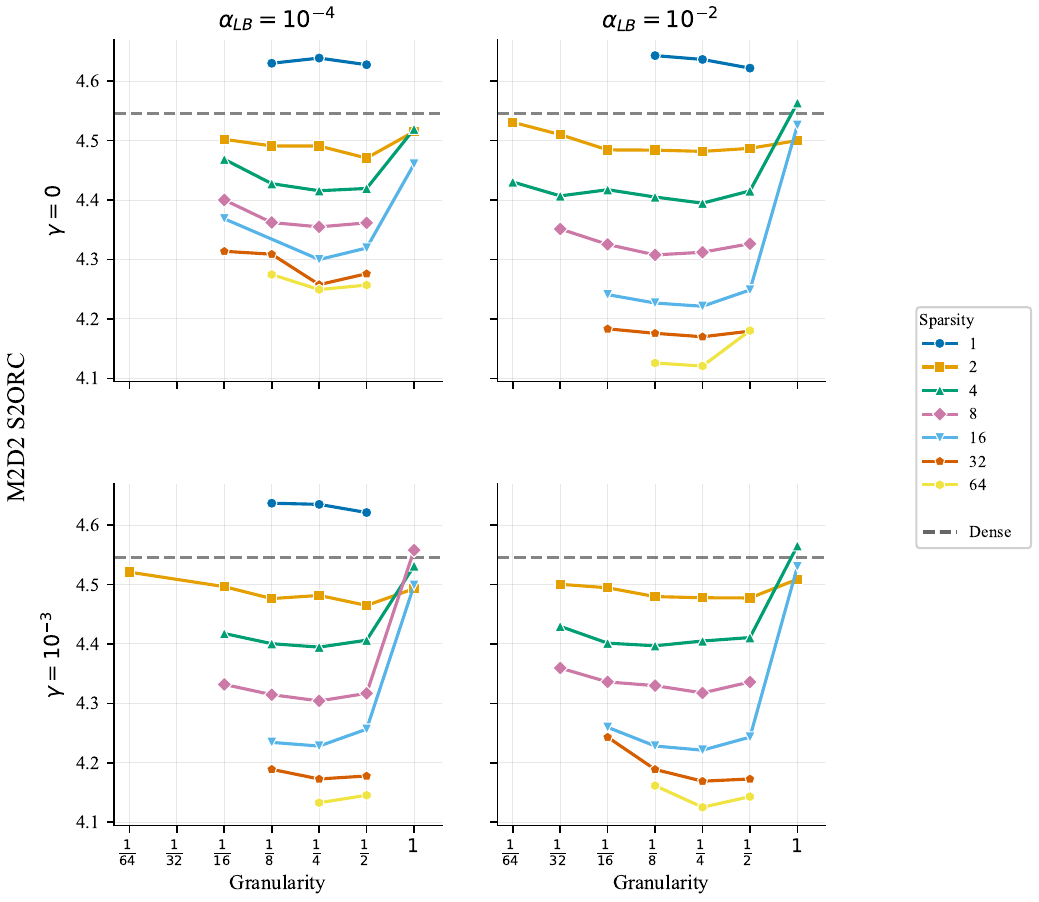}
        \hspace{1em}
        \includegraphics[width=0.46\linewidth]{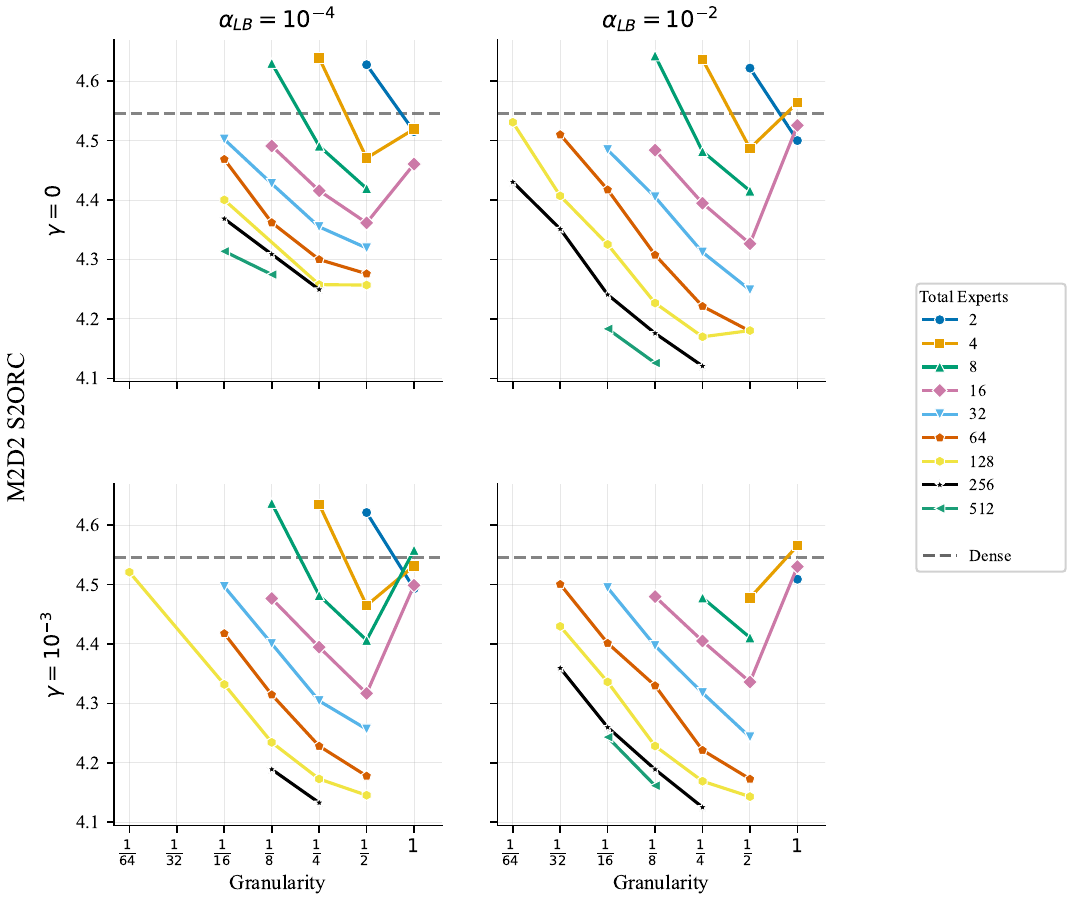}
        \caption{80M active, 80M - 765M total parameters}
    \end{subfigure}
    \par\bigskip\bigskip
    \begin{subfigure}[t]{\textwidth}
        \centering
        \includegraphics[width=0.46\linewidth]{figures/lm/m2d2_s2orc-validation/ce_loss/lb_sweep_hgn_gxs_110M.pdf}
        \hspace{1em}
        \includegraphics[width=0.46\linewidth]{figures/lm/m2d2_s2orc-validation/ce_loss/lb_sweep_hgn_gxn_110M.pdf}
        \caption{110M active, 110M - 1.4B total parameters}
    \end{subfigure}

    \end{figure*} 

\clearpage  

\begin{figure*}[ht]
    \addtocounter{figure}{-1}
    \centering
    \begin{subfigure}[t]{\textwidth}
        \centering
        \includegraphics[width=0.46\linewidth]{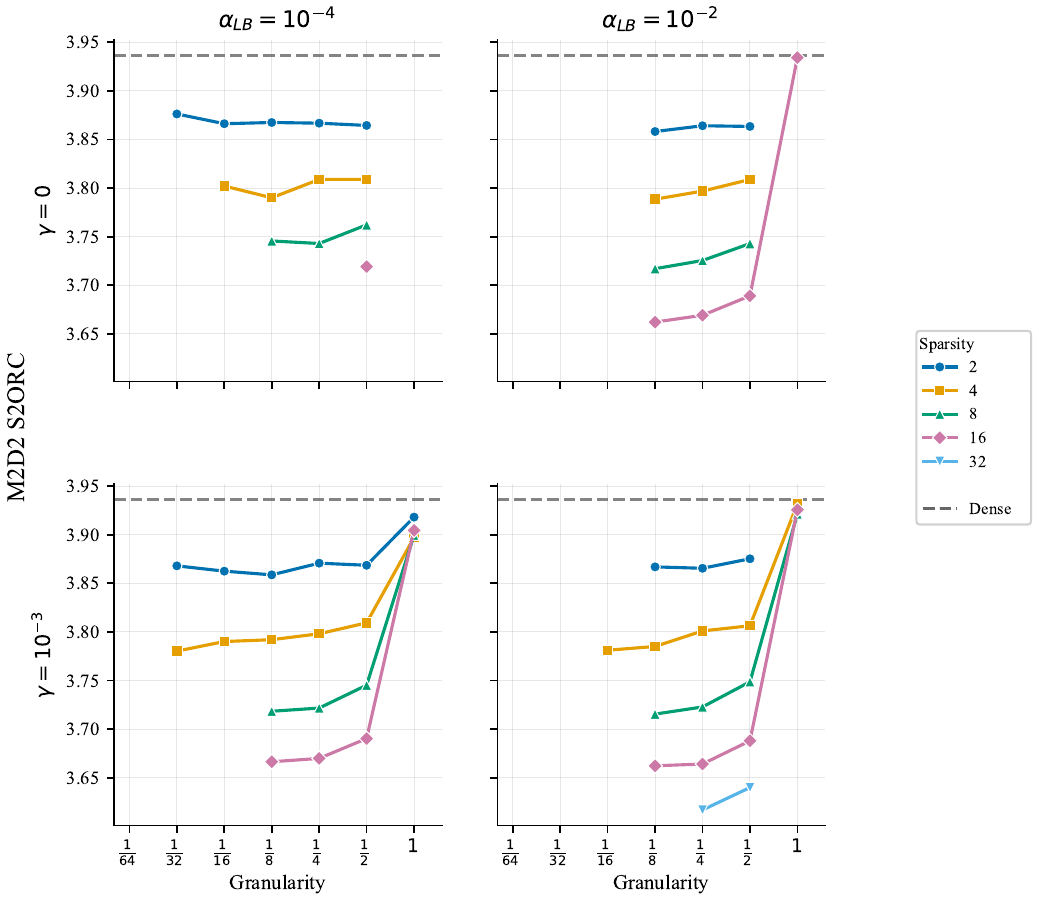}
        \hspace{1em}
        \includegraphics[width=0.46\linewidth]{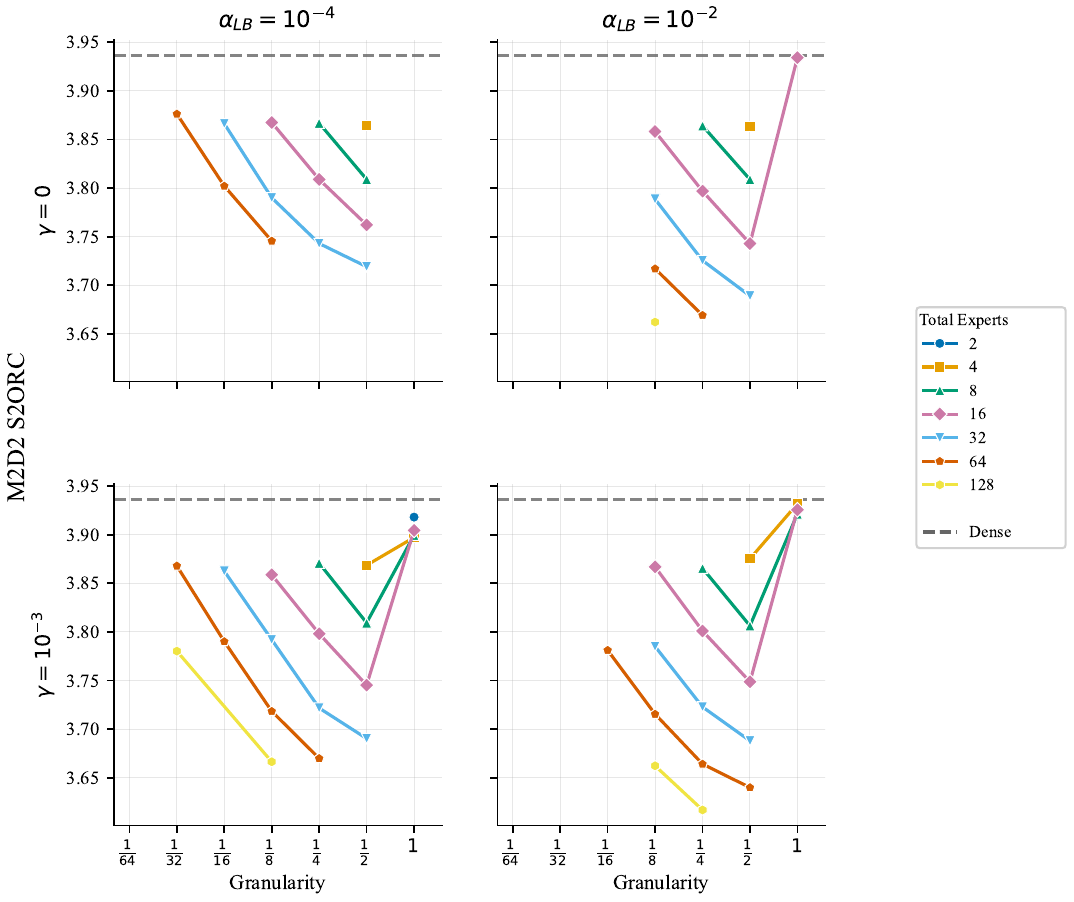}
        \caption{200M active, 200M - 3.3B total parameters}
    \end{subfigure}
    \par\bigskip\bigskip
    \begin{subfigure}[t]{\textwidth}
        \centering
        \includegraphics[width=0.3\linewidth]{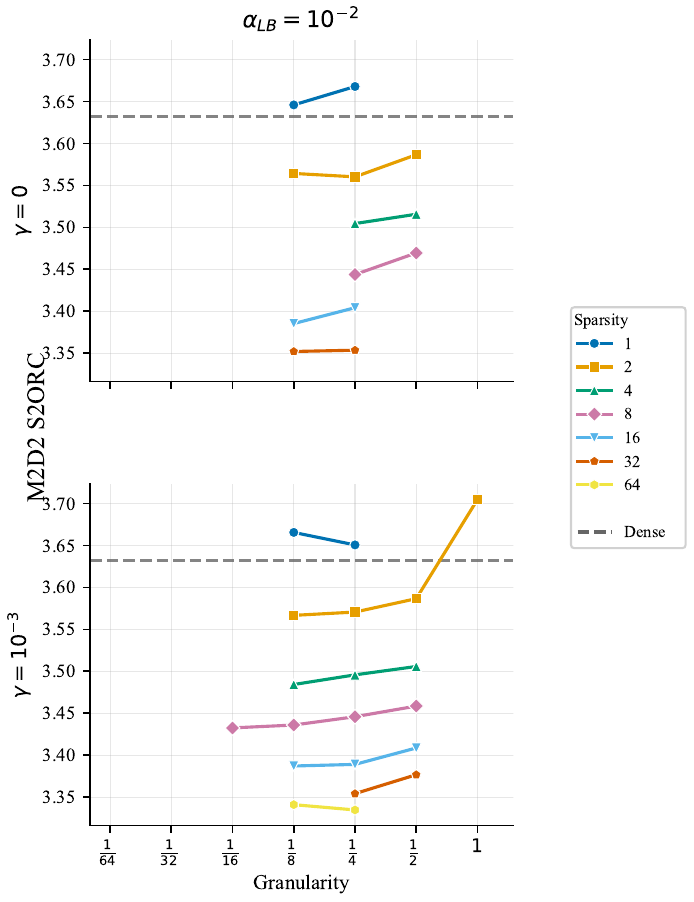}
        \hspace{1em}
        \includegraphics[width=0.3\linewidth]{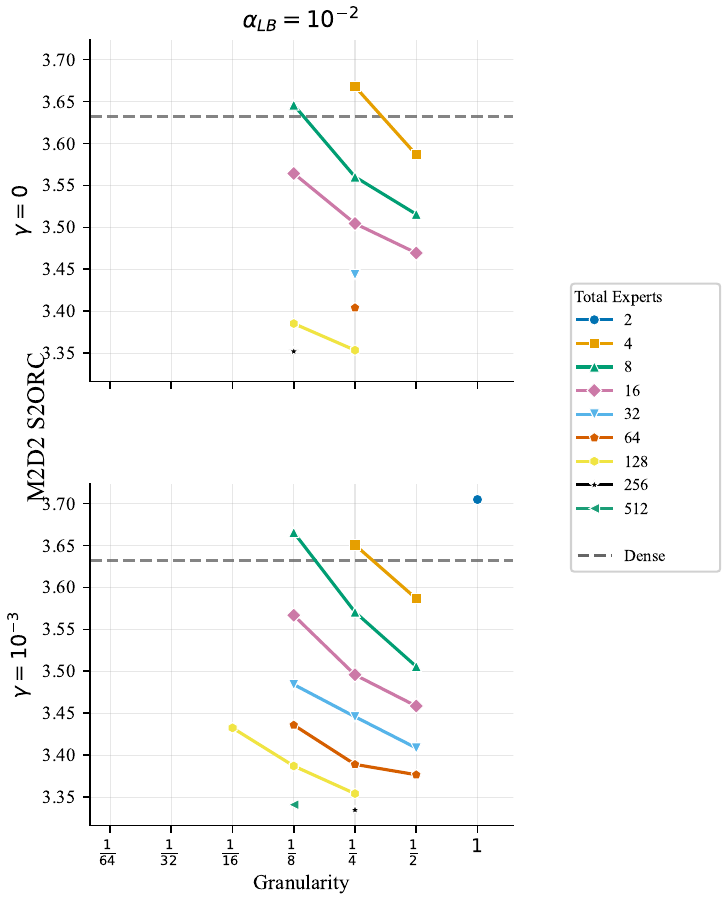}
        \caption{300M active, 300M - 6.6B total parameters}
    \end{subfigure}

    \caption{
    \textbf{Load balancing mechanisms must be tuned correctly (\S\ref{sec:expt_router}).}
    We consider load balancing loss weight $\alpha_{LB} \in \{\num{1e-2}, \num{1e-4}\}$ and loss-free load balancing with bias $\gamma\in\{0, \num{1e-3}\}$ ($\gamma=0$ indicates no loss-free mechanism). Results show that poorly chosen hyperparameters, such as high bias $\gamma = 1e-3$ with total experts $n\geq 512$, may impair performance. However, all settings other than $(\alpha_{LB}=\num{1e-2}, \gamma=\num{1e-3})$ perform comparably for $n \leq 512$, suggesting that a wide range of load balancing settings achieve near-optimal performance. 
    }
    \label{fig:m2d2_s2orc_lb}
\end{figure*}

%% file: fig_tex/lm/pile.tex
\begin{figure*}[!ht]
    \centering
        \begin{subfigure}[t]{\textwidth}
        \begin{subfigure}[t]{0.33\textwidth}
            \centering
            \caption*{\scriptsize Fixed total experts (n)}
            \includegraphics[width=\linewidth]{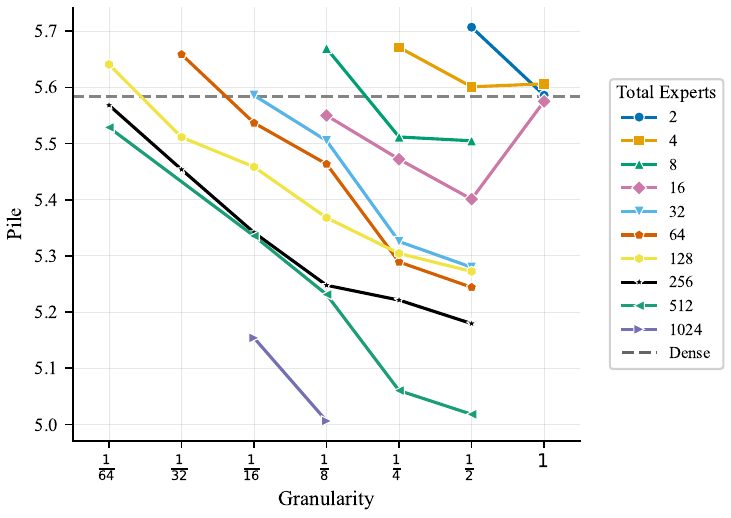}
        \end{subfigure}
        \begin{subfigure}[t]{0.33\textwidth}
            \centering
            \caption*{\scriptsize Fixed granularity (g)}
            \includegraphics[width=\linewidth]{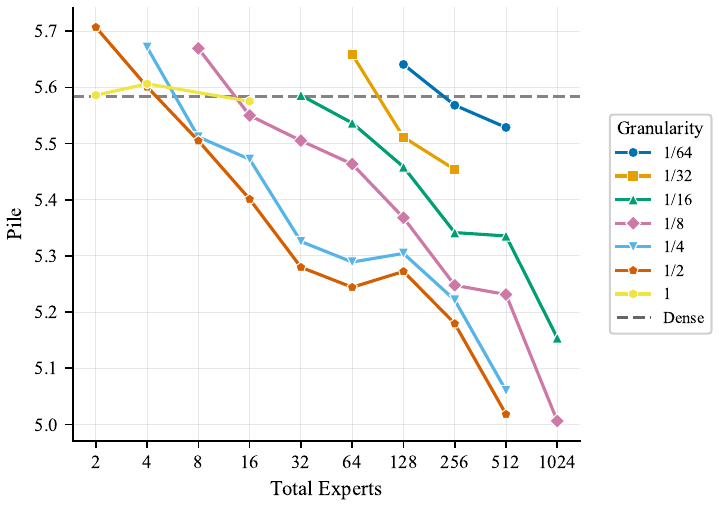}
        \end{subfigure}
        \begin{subfigure}[t]{0.33\textwidth}
            \centering
            \caption*{\scriptsize Fixed activation sparsity (s)}
            \includegraphics[width=\linewidth]{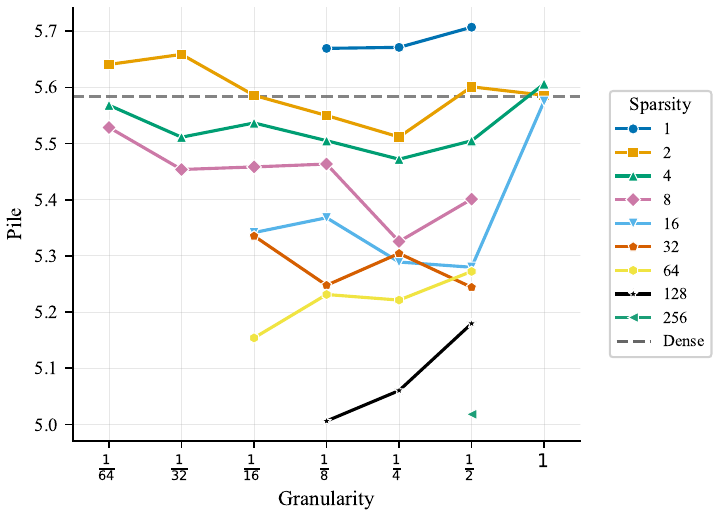}
        \end{subfigure}
        \caption{50M active, 50M - 930M total parameters}
    \end{subfigure}
\par\bigskip\bigskip
    \begin{subfigure}[t]{\textwidth}
        \begin{subfigure}[t]{0.33\textwidth}
            \centering
            \includegraphics[width=\linewidth]{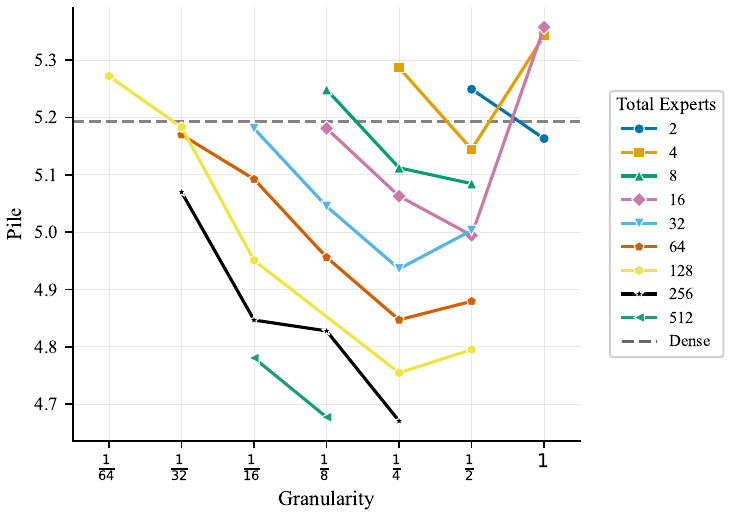}
        \end{subfigure}
        \begin{subfigure}[t]{0.33\textwidth}
            \centering
            \includegraphics[width=\linewidth]{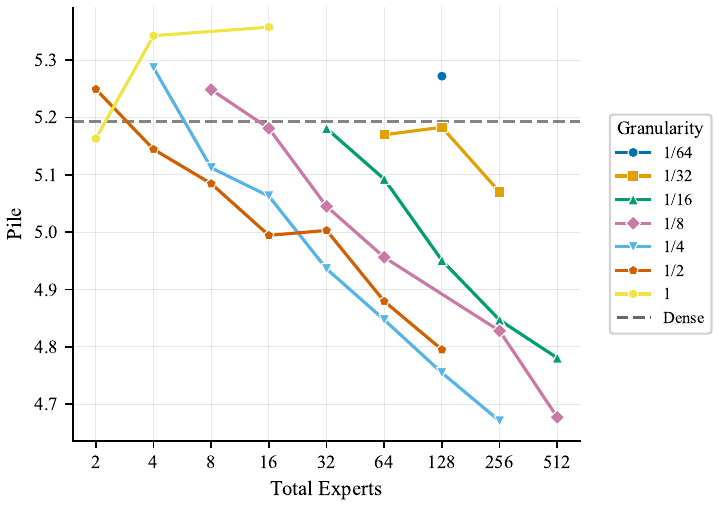}
        \end{subfigure}
        \begin{subfigure}[t]{0.33\textwidth}
            \centering
            \includegraphics[width=\linewidth]{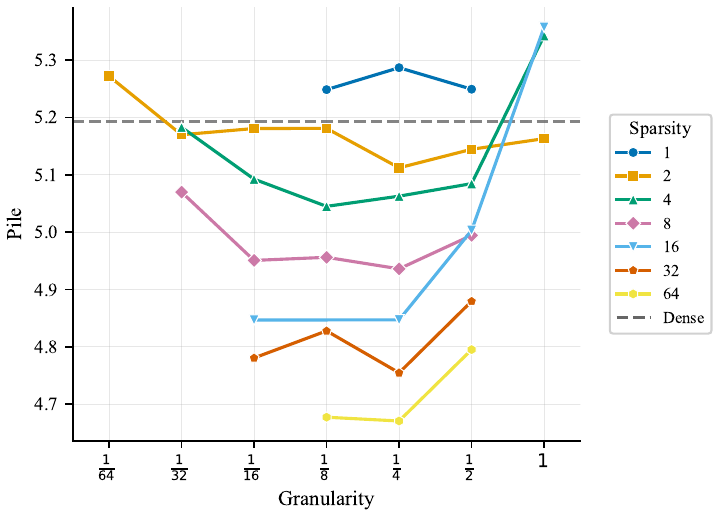}
        \end{subfigure}
        \caption{80M active, 80M - 765M total parameters}
    \end{subfigure}
    \par\bigskip\bigskip
        \begin{subfigure}[t]{\textwidth}
        \begin{subfigure}[t]{0.33\textwidth}
            \centering
            \includegraphics[width=\linewidth]{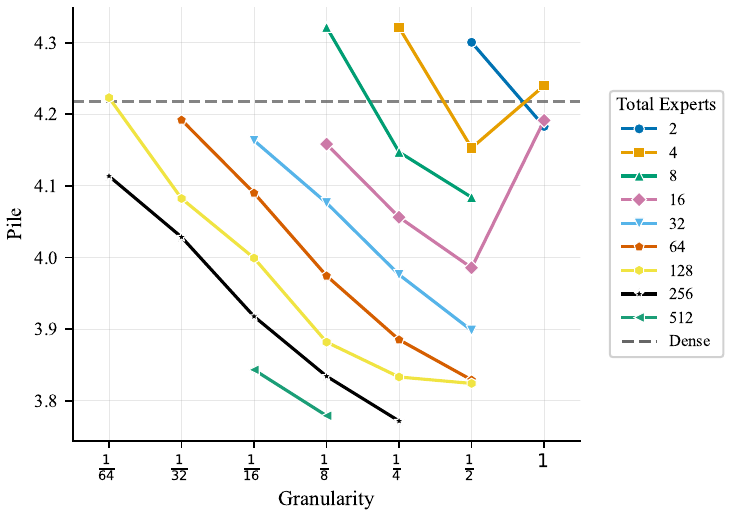}
        \end{subfigure}
        \begin{subfigure}[t]{0.33\textwidth}
            \centering
            \includegraphics[width=\linewidth]{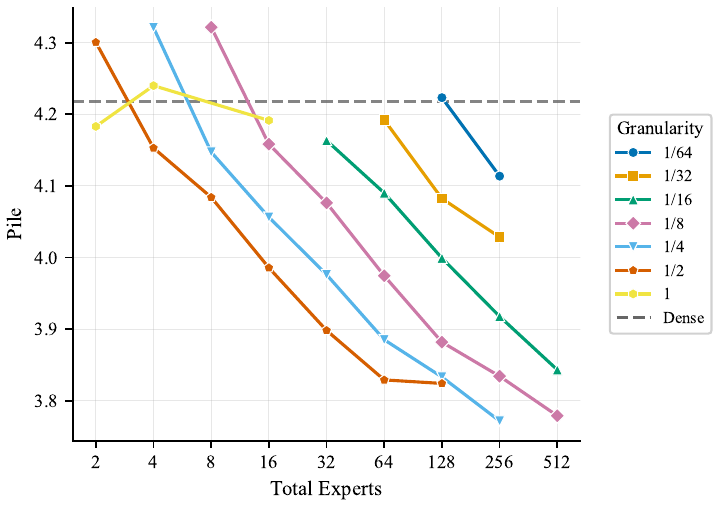}
        \end{subfigure}
        \begin{subfigure}[t]{0.33\textwidth}
            \centering
            \includegraphics[width=\linewidth]{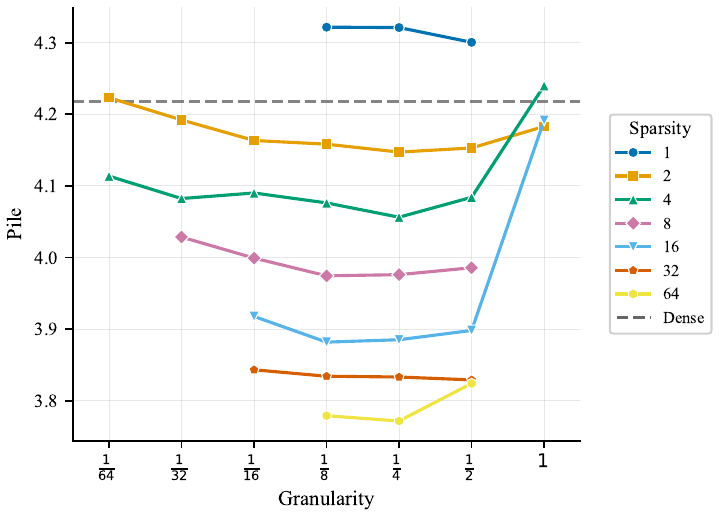}
        \end{subfigure}
        \caption{110M active, 110M - 1.4B total parameters}
    \end{subfigure}
    \end{figure*}

\clearpage  

\begin{figure*}[!ht]
        \addtocounter{figure}{-1}
    \begin{subfigure}[t]{\textwidth}
        \addtocounter{subfigure}{3}
        \begin{subfigure}[t]{0.33\textwidth}
            \centering
            \caption*{\scriptsize Fixed total experts (n)}
            \includegraphics[width=\linewidth]{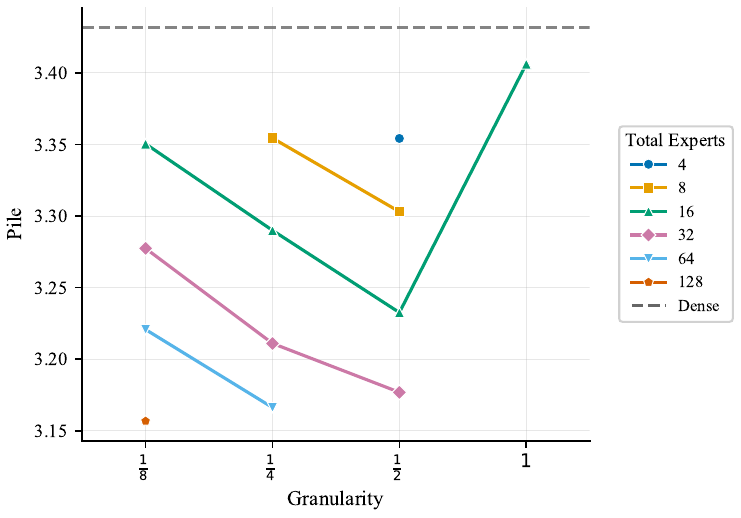}
        \end{subfigure}
        \begin{subfigure}[t]{0.33\textwidth}
            \centering
            \caption*{\scriptsize Fixed granularity (g)}
            \includegraphics[width=\linewidth]{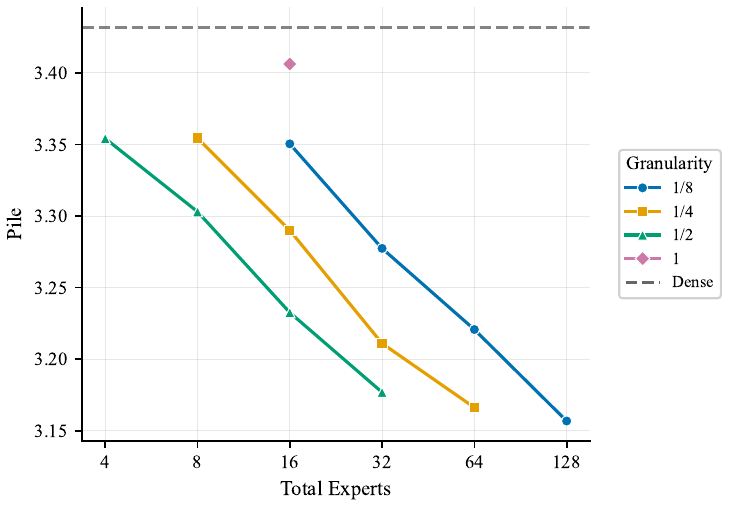}
        \end{subfigure}
        \begin{subfigure}[t]{0.33\textwidth}
            \centering
            \caption*{\scriptsize Fixed activation sparsity (s)}
            \includegraphics[width=\linewidth]{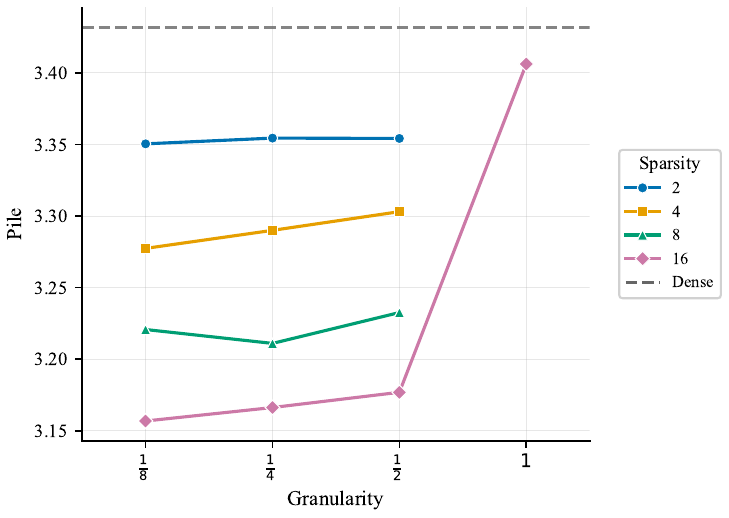}
        \end{subfigure}
        \caption{200M active, 200M - 3.3B total parameters}
    \end{subfigure}
    \par\bigskip\bigskip
        \begin{subfigure}[t]{\textwidth}
        \begin{subfigure}[t]{0.33\textwidth}
            \centering
            \includegraphics[width=\linewidth]{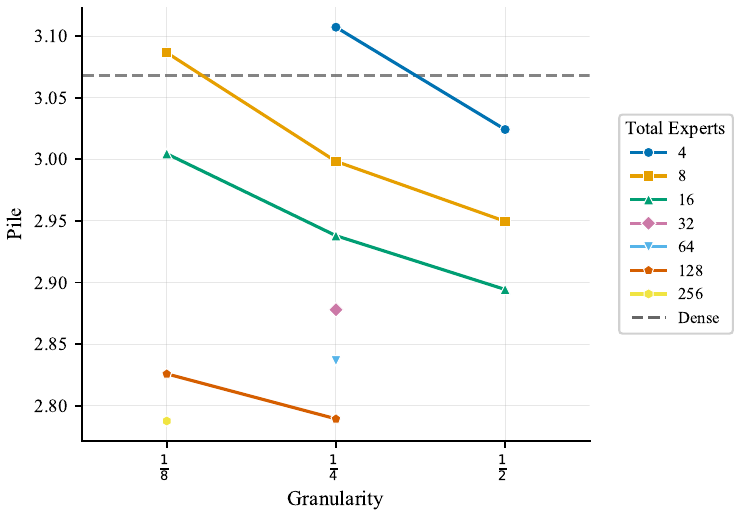}
        \end{subfigure}
        \begin{subfigure}[t]{0.33\textwidth}
            \centering
            \includegraphics[width=\linewidth]{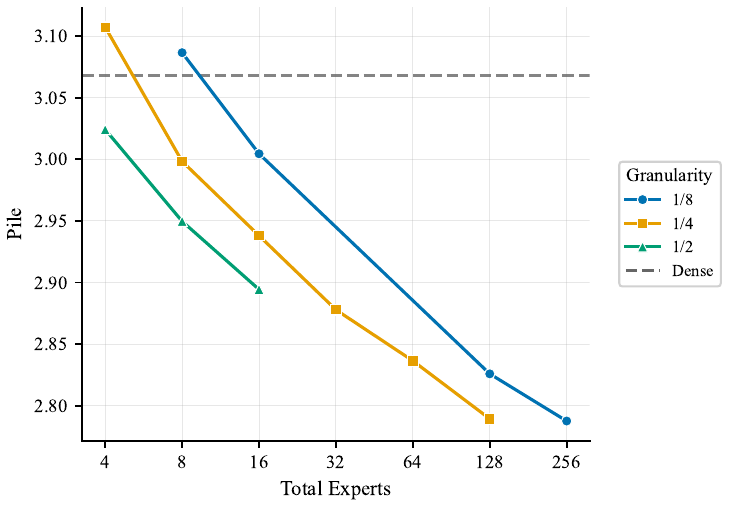}
        \end{subfigure}
        \begin{subfigure}[t]{0.33\textwidth}
            \centering
            \includegraphics[width=\linewidth]{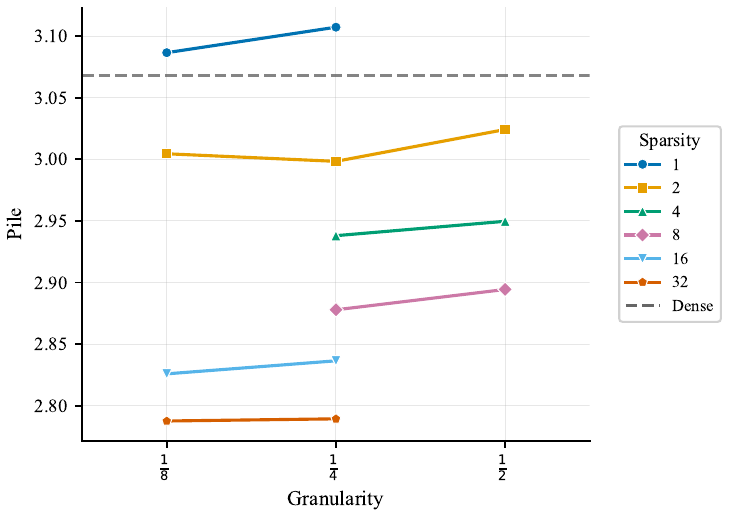}
        \end{subfigure}
        \caption{300M active, 300M - 6.6B total parameters}
    \end{subfigure}

    \caption{
    \textbf{Increasing inactive expert parameters via expert size (left) or total count (center) improves performance in MoEs (\S\ref{sec:expt_main}).} This effect is seen both when holding total number of experts fixed (left) and when holding expert granularity fixed (center). In general, increasing total parameters results in improved performance.  \textbf{Optimal tradeoff between expert count and granularity varies in MoEs (right). (\S\ref{sec:expt_main})}
    At each activation sparsity $s$ (equivalently, at each total parameter count), the optimal (total expert count, expert granularity) configuration varies. As $s$ increases, optimal expert granularity remains nearly fixed, suggesting that sparsity should be scaled up primarily by increasing total expert count $n$, while maintaining a near constant, slowly increasing expert granularity $g$. 
    }
    \label{fig:pile_experts}
\end{figure*}

\begin{figure*}[!ht]
    \centering
    
    \begin{subfigure}[t]{0.46\textwidth}
        \centering
        \includegraphics[width=\linewidth]{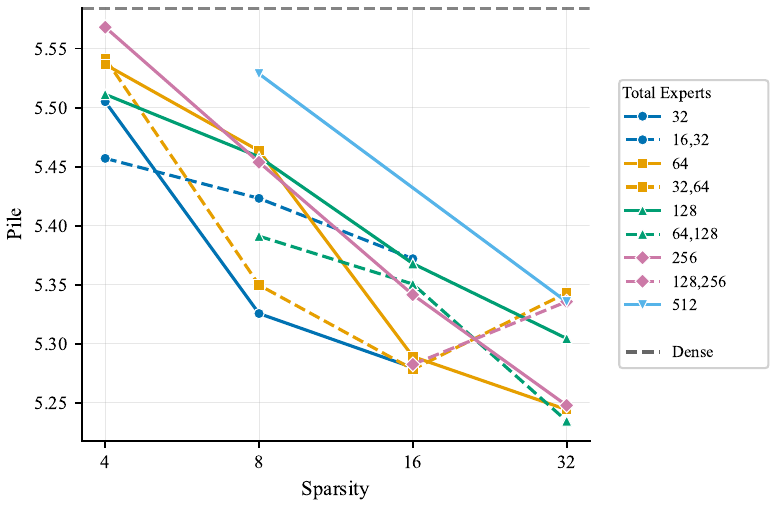}
        \caption{50M active, 50M - 930M total parameters}
    \end{subfigure}
    \vspace{1em}
    \begin{subfigure}[t]{0.46\textwidth}
        \centering
        \includegraphics[width=\linewidth]{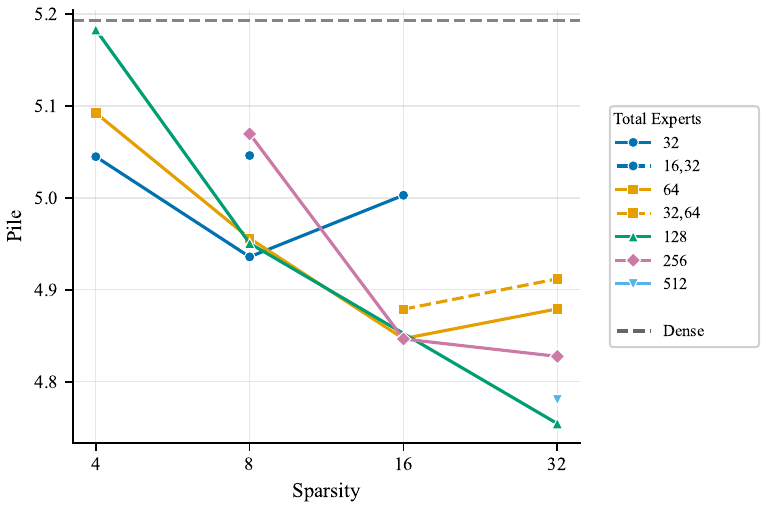}
        \caption{80M active, 80M - 765M total parameters}
    \end{subfigure}
    \caption{
    \textbf{Heterogeneity of expert size alone does not improve MoE performance (\S\ref{sec:expt_hetgen}).} To explore the potential benefits of their architectural flexibility, we compare heterogeneous MoEs (indicated by dotted lines) to active- and total-parameter-matched homogeneous MoEs. Heterogeneity alone does not result in performance gains, as, at each activation sparsity $s$, heterogeneous MoEs with $n_1, n_2 = a, b$ lie between or near the 2 closest homogeneous MoEs, with $n=a$ and with $n=b$.
    }
    \label{fig:pile_het}
\end{figure*}

\begin{figure*}[!ht]
    \centering
    
    \begin{subfigure}[t]{1.0\textwidth}
        \centering
        \includegraphics[width=\linewidth]{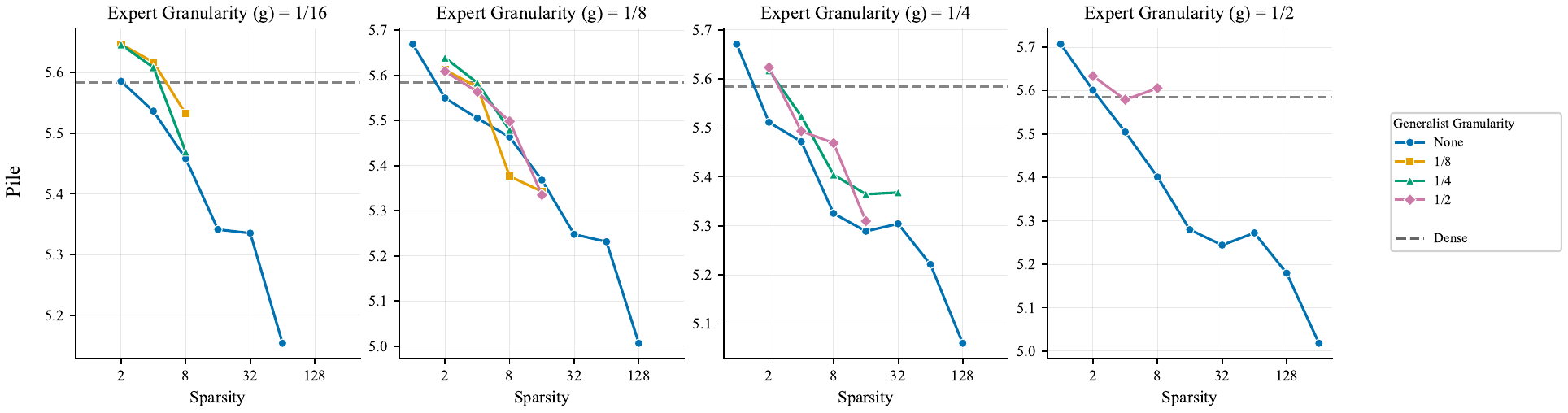}
        \caption{50M active, 50M - 930M total parameters}
    \end{subfigure}
    \par\bigskip\bigskip
    \begin{subfigure}[t]{1.0\textwidth}
        \centering
        \includegraphics[width=\linewidth]{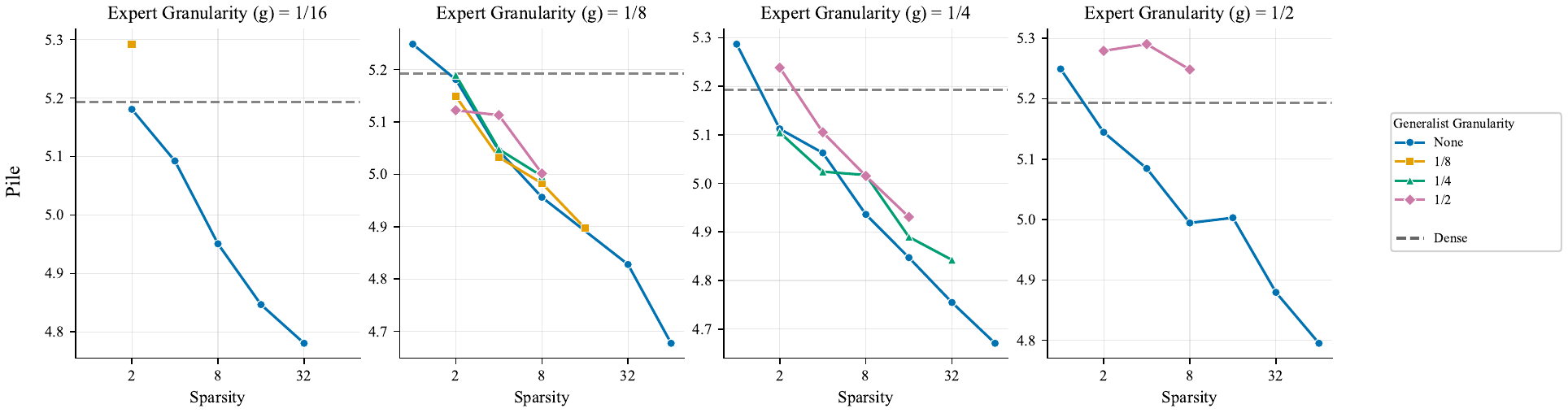}
        \caption{80M active, 80M - 765M total parameters}
    \end{subfigure}
    \par\bigskip\bigskip
    \begin{subfigure}[t]{1.0\textwidth}
        \centering
        \includegraphics[width=\linewidth]{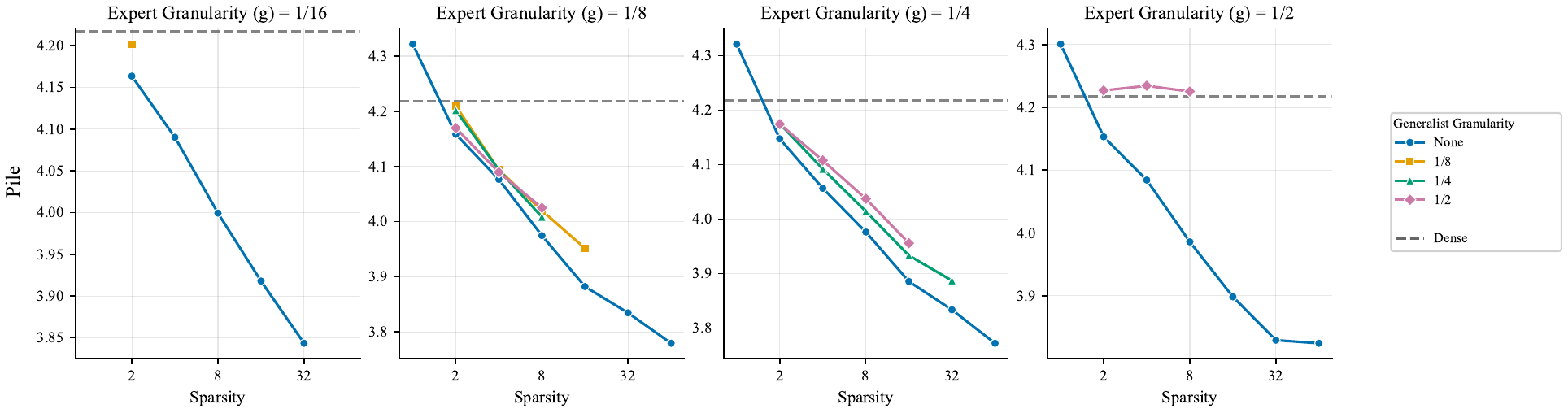}
        \caption{110M active, 110M - 1.4B total parameters}
    \end{subfigure}
    \caption{
    \textbf{The inclusion of a generalist consistently degrades performance in homogeneous MoEs (\S\ref{sec:expt_hetgen}).}
    We train MoE LMs which consist of some routed experts with granularity $g$, as well as a generalist with granularity $g_{gen}\in \{\frac{1}{2}, \frac{1}{4}, \frac{1}{8}\} $. We compare to settings with no generalist, only routed experts with granularity $g$. In all settings and configurations, the addition of any granularity generalist results in comparable or degraded performance. 
    }
    \label{fig:pile_gen}
\end{figure*}

\begin{figure*}[ht]
    \centering
    \begin{subfigure}[t]{1.0\textwidth}
        \centering
        \includegraphics[width=\linewidth]{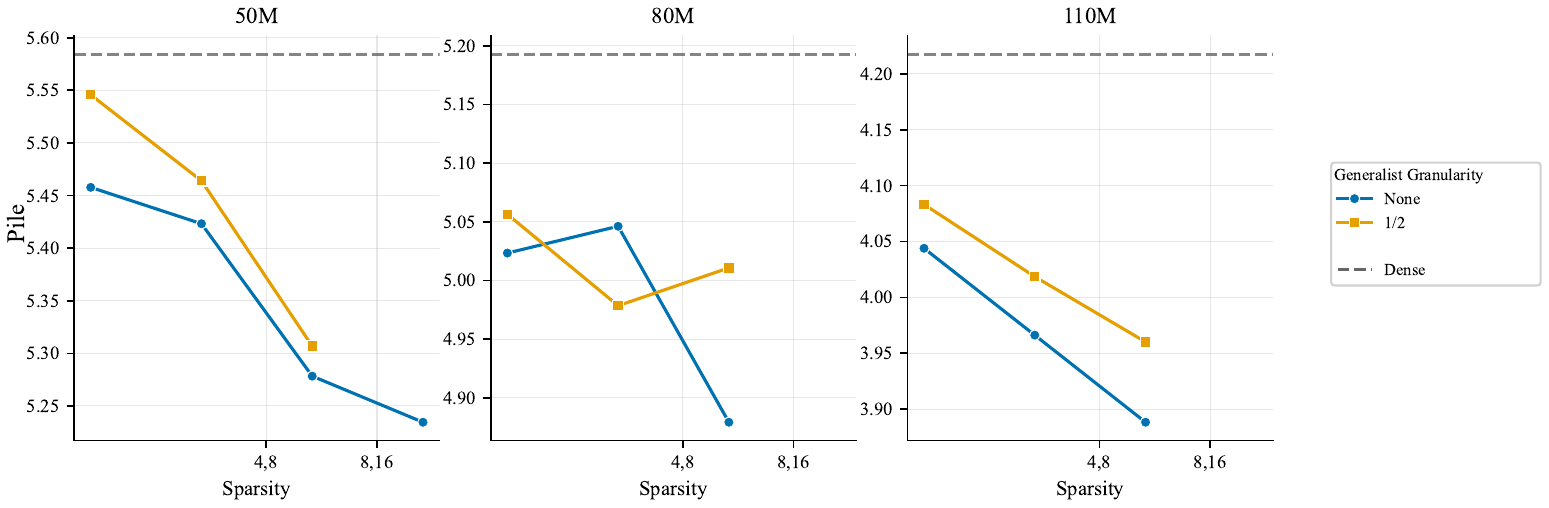}
    \end{subfigure}
    \caption{
    \textbf{The inclusion of a generalist consistently degrades performance in heterogeneous MoEs (\S\ref{sec:expt_hetgen}).}
    We train heterogeneous MoE LMs which consist of  routed experts with granularity $g_1, g_2$, as well as a generalist with granularity $g_{gen} = \frac{1}{2}$. We compare to settings with no generalist. In all settings and configurations, the addition of a generalist results in comparable or degraded performance. 
    }
    \label{fig:pile_hetgen}
\end{figure*}

\begin{figure*}[ht]
    \centering
    \begin{subfigure}[t]{\textwidth}
        \centering
        \begin{subfigure}[t]{0.45\textwidth}
            \includegraphics[width=\linewidth]{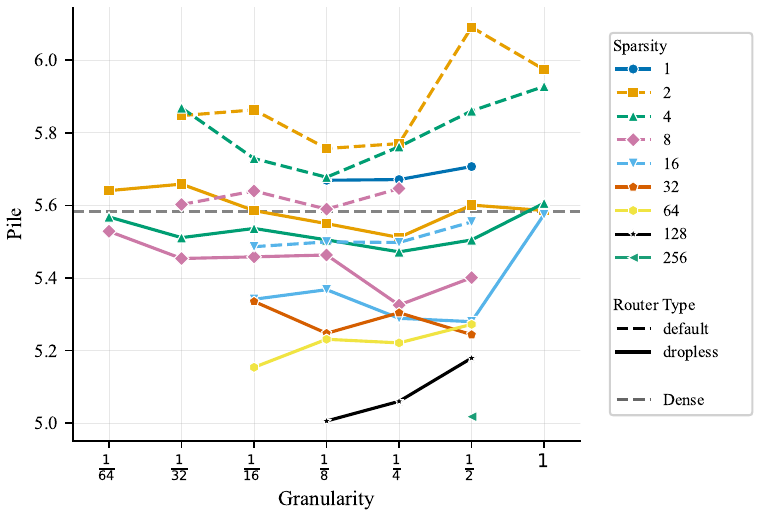}
            \caption{50M active, 50M - 930M total parameters}
        \end{subfigure}
    \hspace{1em}
        \begin{subfigure}[t]{0.45\textwidth}
            \centering
            \includegraphics[width=\linewidth]{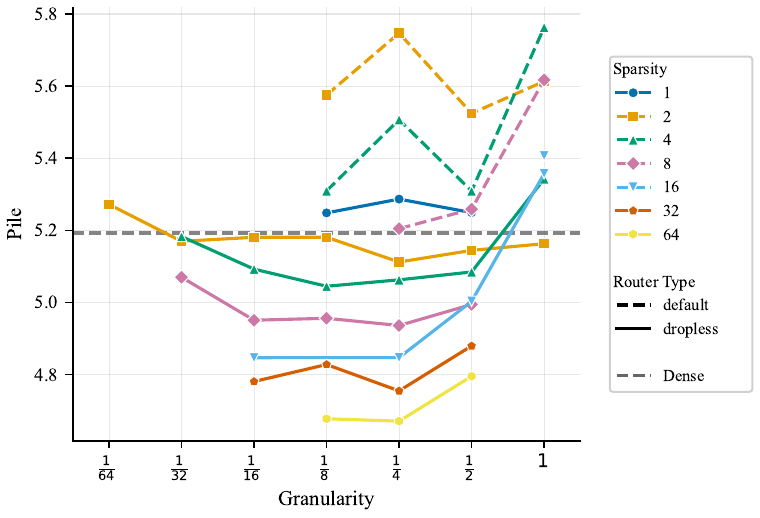}
            \caption{80M active, 80M - 765M total parameters}
        \end{subfigure}
    \end{subfigure}

    \par\bigskip\bigskip
    \begin{subfigure}[t]{0.45\textwidth}
        \centering
        \includegraphics[width=\linewidth]{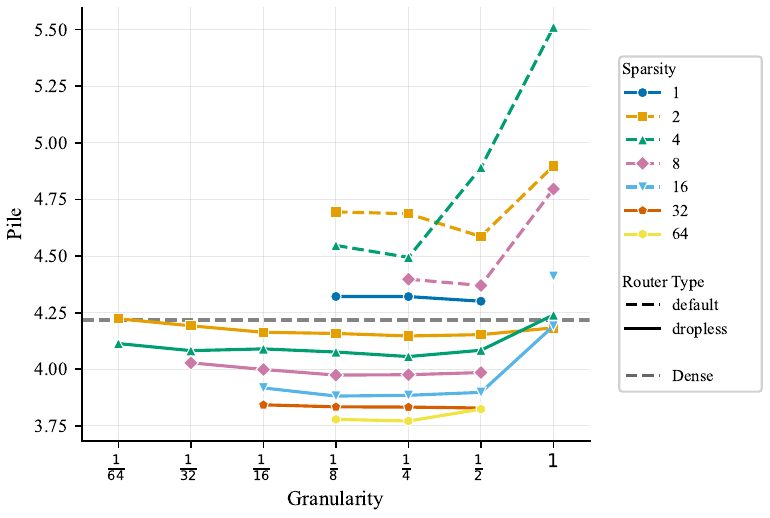}
        \caption{110M active, 110M - 1.4B total parameters}
    \end{subfigure}
    \caption{ 
    \textbf{Dropless routing outperforms default routing (\S\ref{sec:expt_router}).}
    We compare dropless routing to the default setting, which allow tokens to be dropped. Across all scales, we find that dropless routing outperforms or performs comparably to default routing. 
    }
    \label{fig:pile_dropless}
\end{figure*}

\begin{figure*}[ht]
    \centering
    \begin{subfigure}[t]{0.45\textwidth}
        \centering
        \includegraphics[width=\linewidth]{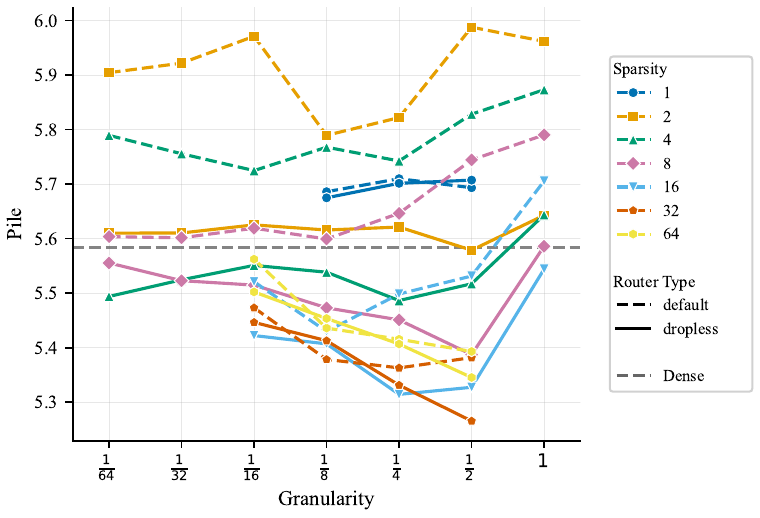}
        \caption{50M active, 50M - 930M total parameters}
    \end{subfigure}
    \hspace{1em}
    \begin{subfigure}[t]{0.45\textwidth}
        \centering
        \includegraphics[width=\linewidth]{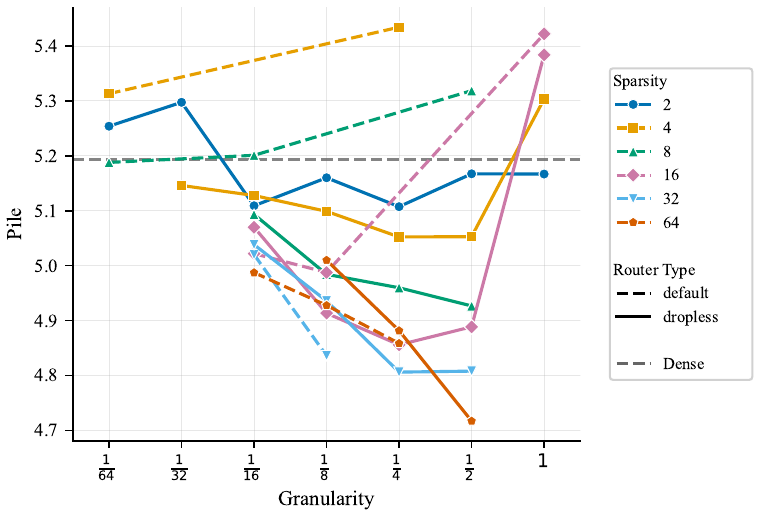}
        \caption{80M active, 80M - 765M total parameters}
    \end{subfigure}
    \caption{
    \textbf{Dropless routing, with bias $\gamma=\num{1e-3}$ (\S\ref{sec:expt_router}).} 
    As in Figure~\ref{fig:lm_avg_dropless}, we compare dropless routing to the default setting, which allow tokens to be dropped. Across all scales, we find that dropless routing outperforms or performs comparably to default routing. We see here with additional higher sparsity default routing runs that as sparsity increases, default routing performance approaches that of dropless routing.
    }
    \label{fig:pile_dropless_with_lf}
\end{figure*}

\begin{figure*}[ht]
    \centering
    \begin{subfigure}[]{\textwidth}
        \centering
        \includegraphics[width=0.46\linewidth]{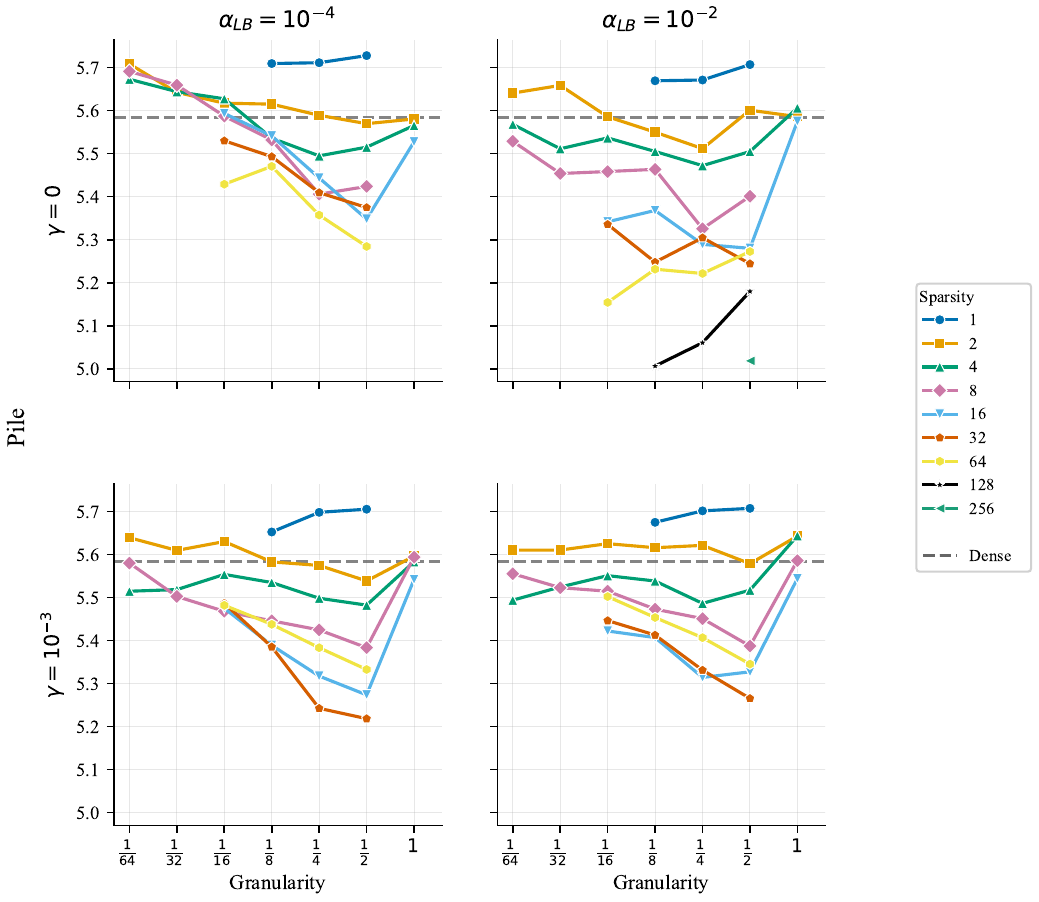}
        \hspace{1em}
        \includegraphics[width=0.46\linewidth]{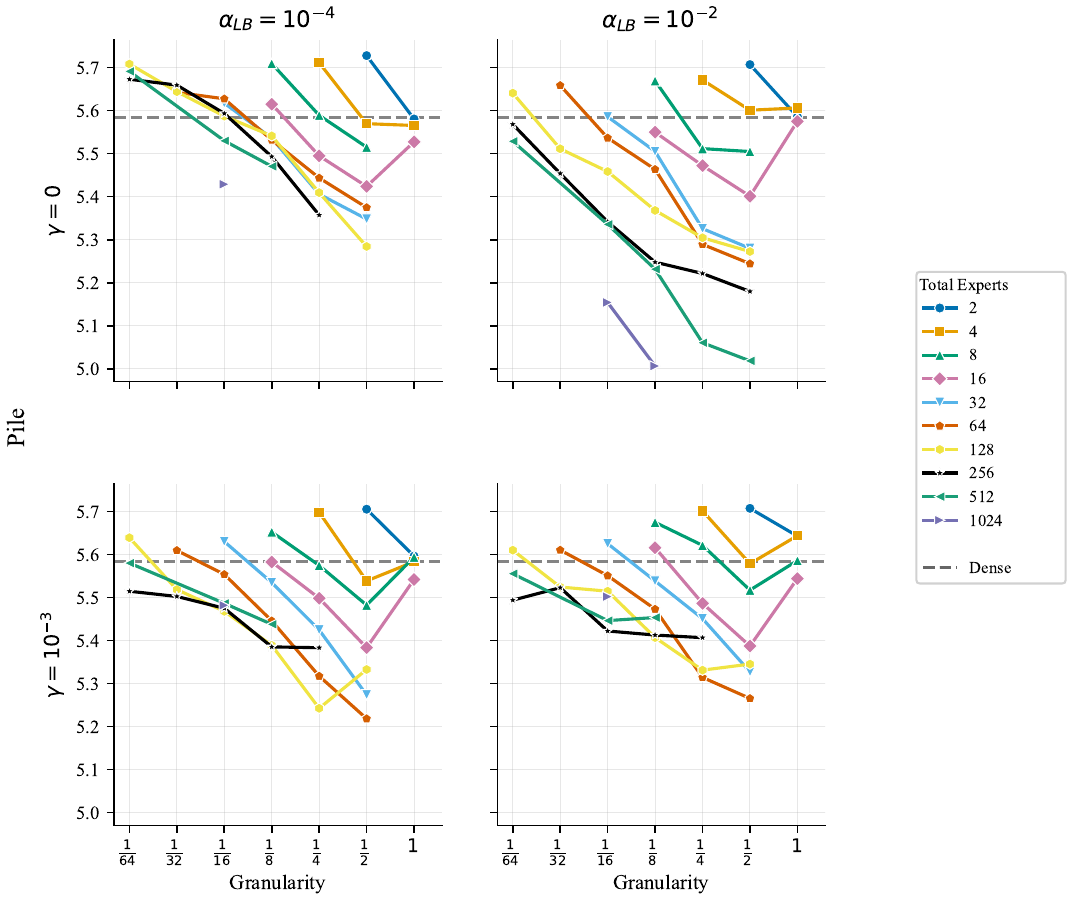}
        \caption{50M active, 50M - 930M total parameters}
    \end{subfigure}
    \par\bigskip\bigskip
    \begin{subfigure}[]{\textwidth}
        \centering
        \includegraphics[width=0.46\linewidth]{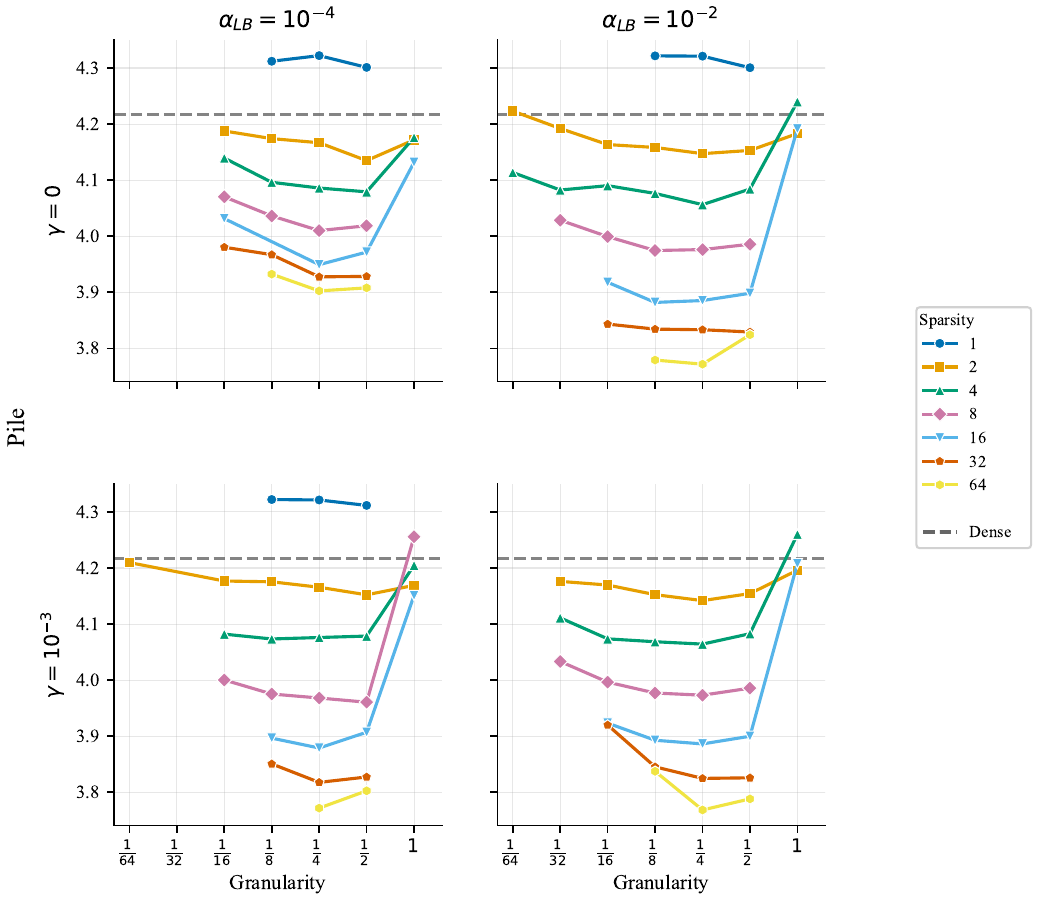}
        \hspace{1em}
        \includegraphics[width=0.46\linewidth]{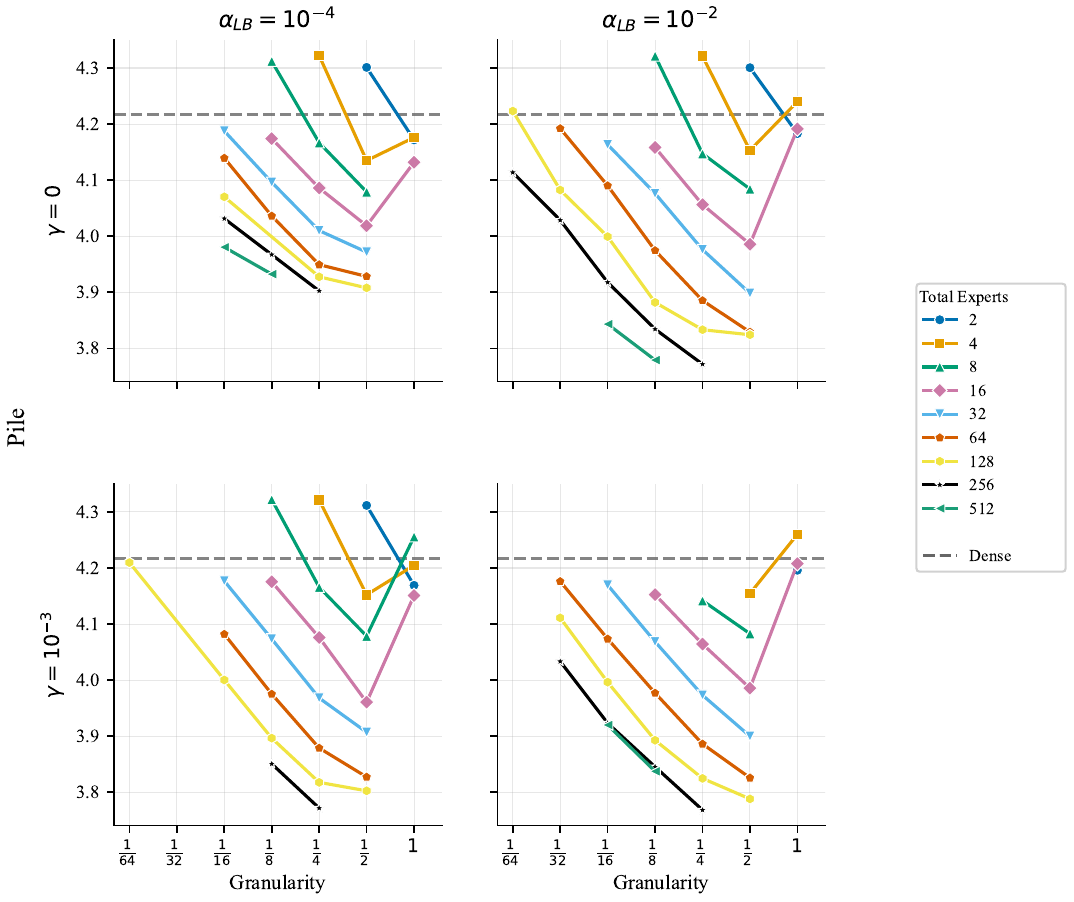}
        \caption{80M active, 80M - 765M total parameters}
    \end{subfigure}
    \par\bigskip\bigskip
    \begin{subfigure}[t]{\textwidth}
        \centering
        \includegraphics[width=0.46\linewidth]{figures/lm/pile-validation/ce_loss/lb_sweep_hgn_gxs_110M.pdf}
        \hspace{1em}
        \includegraphics[width=0.46\linewidth]{figures/lm/pile-validation/ce_loss/lb_sweep_hgn_gxn_110M.pdf}
        \caption{110M active, 110M - 1.4B total parameters}
    \end{subfigure}

    \end{figure*} 

\clearpage  

\begin{figure*}[ht]
    \addtocounter{figure}{-1}
    \centering
    \begin{subfigure}[t]{\textwidth}
        \centering
        \includegraphics[width=0.46\linewidth]{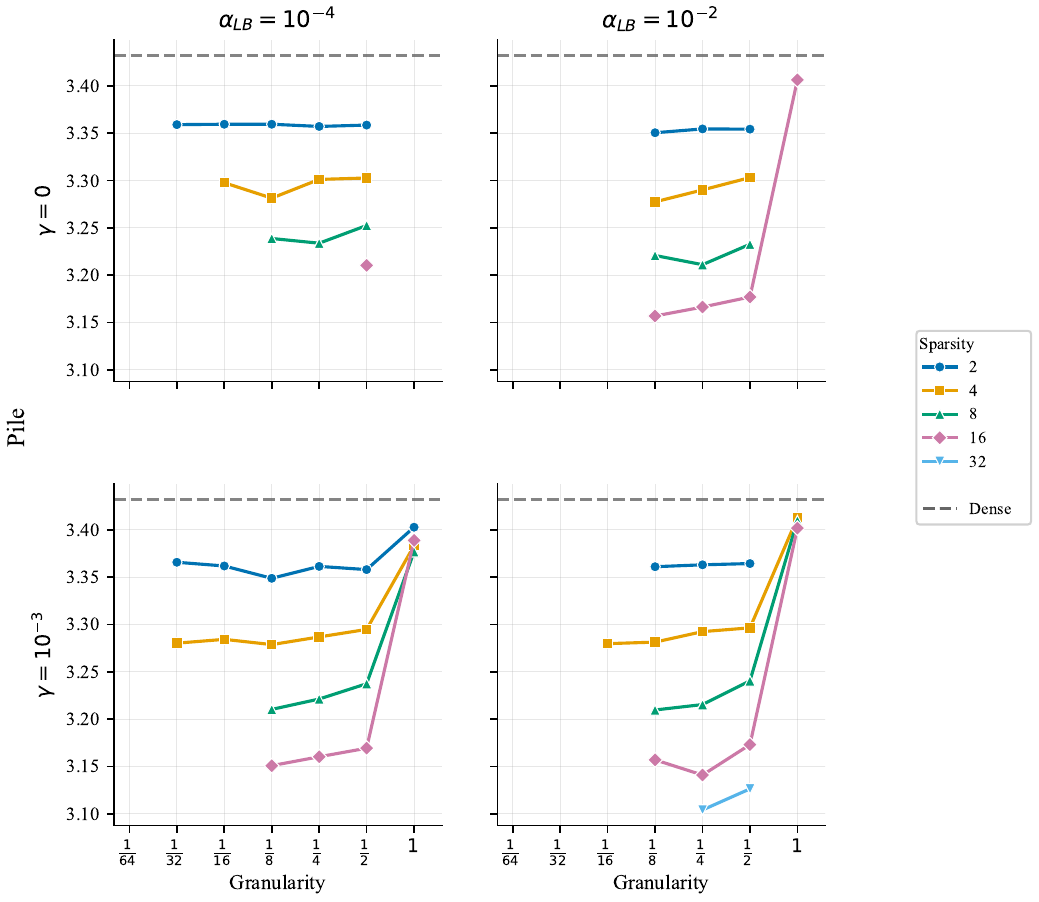}
        \hspace{1em}
        \includegraphics[width=0.46\linewidth]{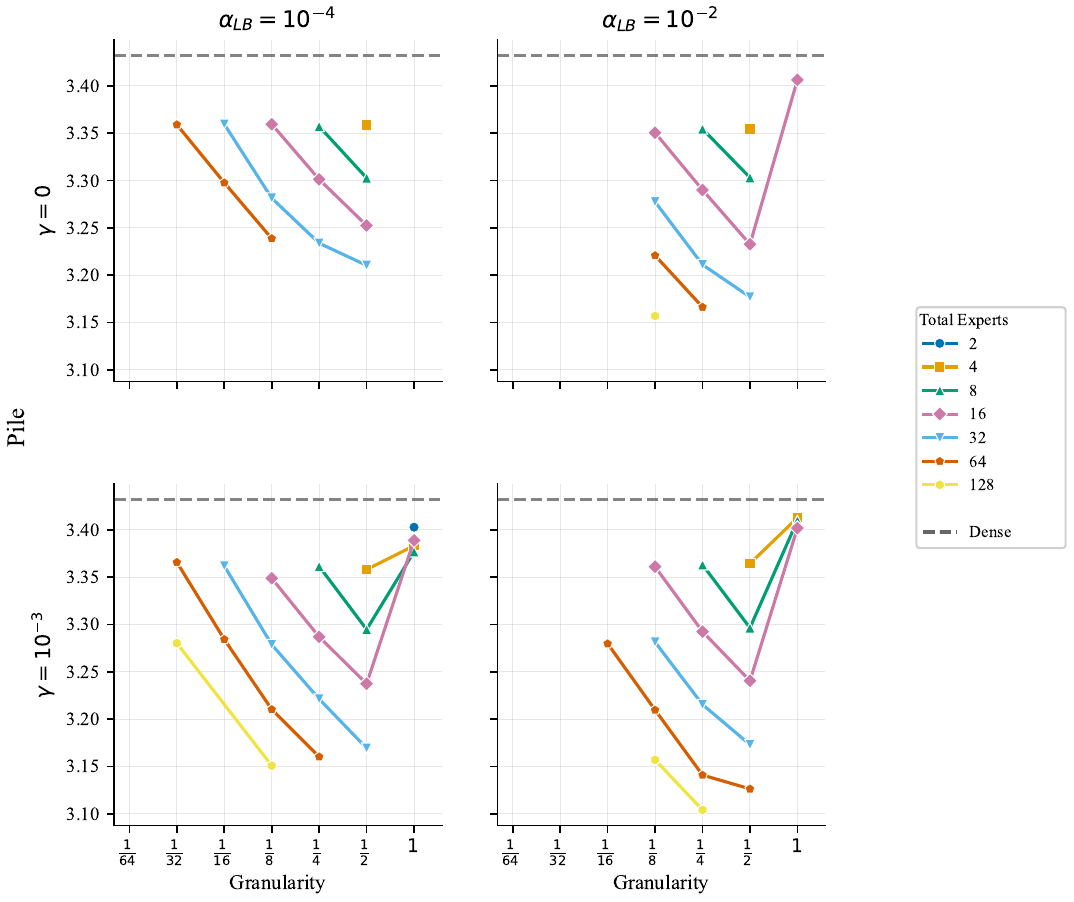}
        \caption{200M active, 200M - 3.3B total parameters}
    \end{subfigure}
    \par\bigskip\bigskip
    \begin{subfigure}[t]{\textwidth}
        \centering
        \includegraphics[width=0.3\linewidth]{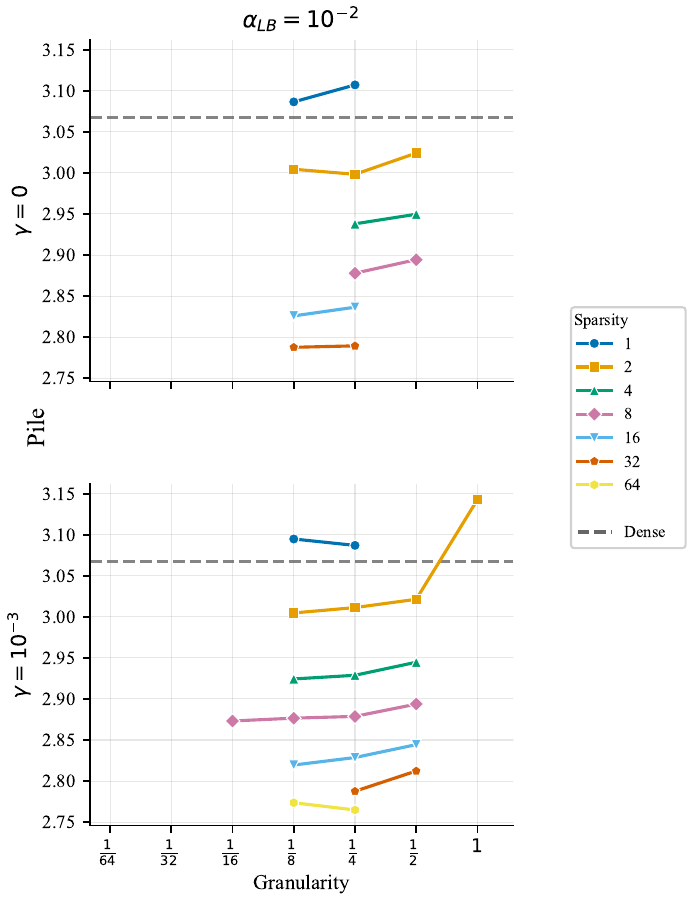}
        \hspace{1em}
        \includegraphics[width=0.3\linewidth]{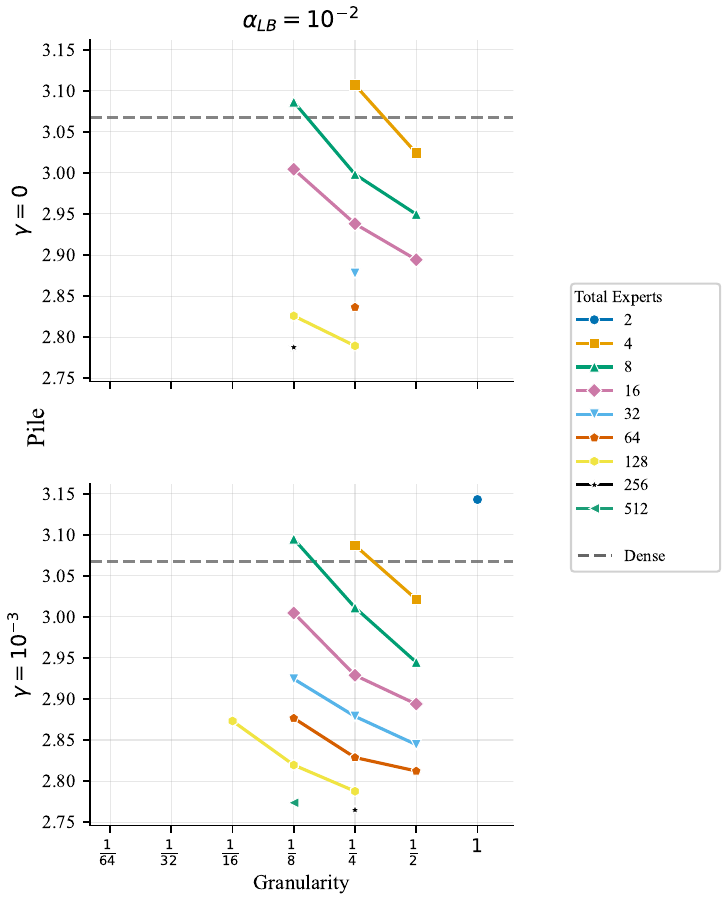}
        \caption{300M active, 300M - 6.6B total parameters}
    \end{subfigure}

    \caption{
    \textbf{Load balancing mechanisms must be tuned correctly (\S\ref{sec:expt_router}).}
    We consider load balancing loss weight $\alpha_{LB} \in \{\num{1e-2}, \num{1e-4}\}$ and loss-free load balancing with bias $\gamma\in\{0, \num{1e-3}\}$ ($\gamma=0$ indicates no loss-free mechanism). Results show that poorly chosen hyperparameters, such as high bias $\gamma = 1e-3$ with total experts $n\geq 512$, may impair performance. However, all settings other than $(\alpha_{LB}=\num{1e-2}, \gamma=\num{1e-3})$ perform comparably for $n \leq 512$, suggesting that a wide range of load balancing settings achieve near-optimal performance. 
    }
    \label{fig:pile_lb}
\end{figure*}

%% file: fig_tex/lm/wikitext_103.tex
\begin{figure*}[!ht]
    \centering
        \begin{subfigure}[t]{\textwidth}
        \begin{subfigure}[t]{0.33\textwidth}
            \centering
            \caption*{\scriptsize Fixed total experts (n)}
            \includegraphics[width=\linewidth]{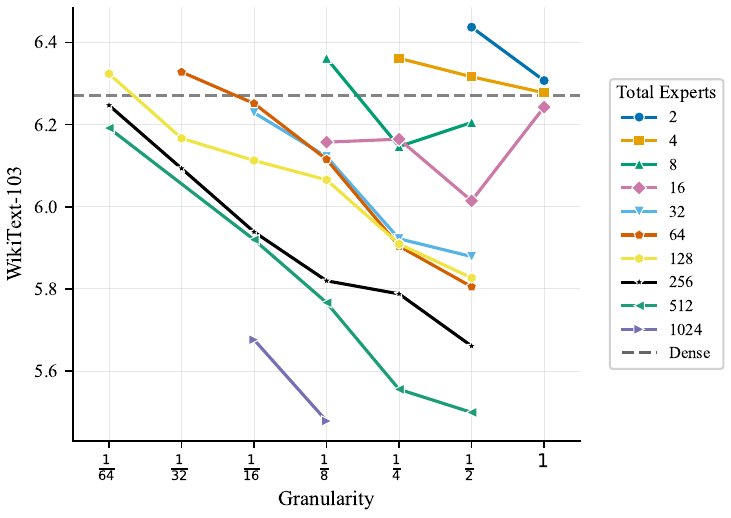}
        \end{subfigure}
        \begin{subfigure}[t]{0.33\textwidth}
            \centering
            \caption*{\scriptsize Fixed granularity (g)}
            \includegraphics[width=\linewidth]{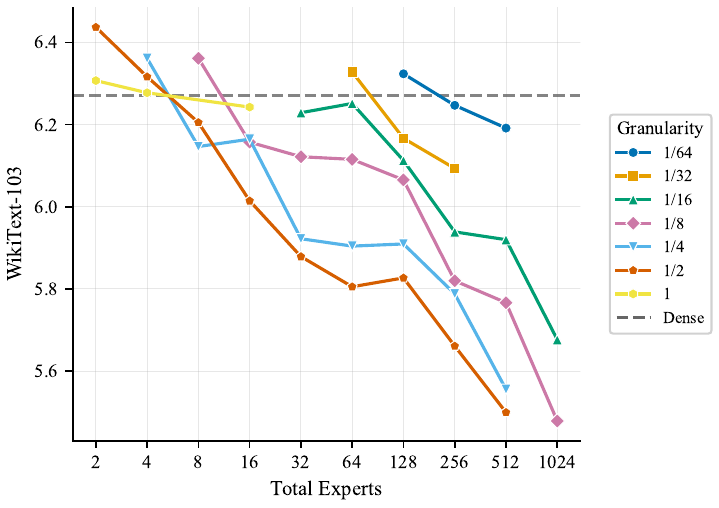}
        \end{subfigure}
        \begin{subfigure}[t]{0.33\textwidth}
            \centering
            \caption*{\scriptsize Fixed activation sparsity (s)}
            \includegraphics[width=\linewidth]{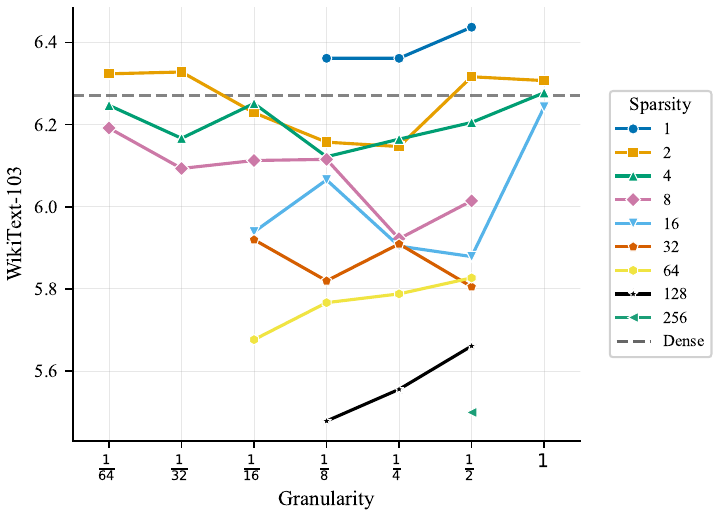}
        \end{subfigure}
        \caption{50M active, 50M - 930M total parameters}
    \end{subfigure}
\par\bigskip\bigskip
    \begin{subfigure}[t]{\textwidth}
        \begin{subfigure}[t]{0.33\textwidth}
            \centering
            \includegraphics[width=\linewidth]{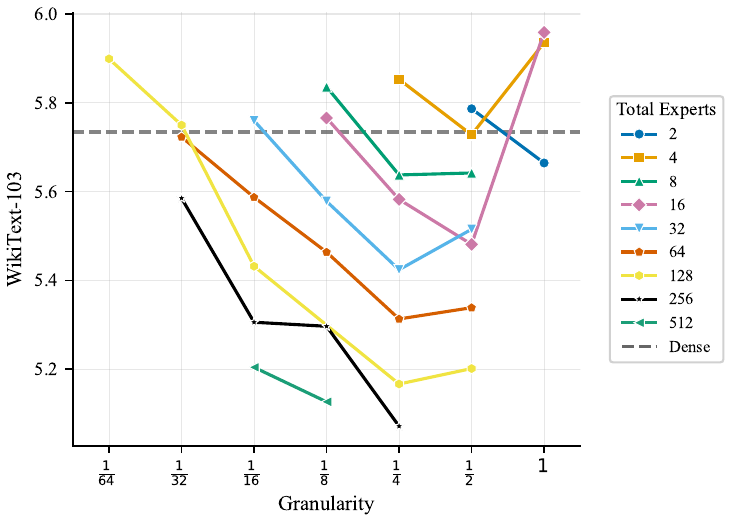}
        \end{subfigure}
        \begin{subfigure}[t]{0.33\textwidth}
            \centering
            \includegraphics[width=\linewidth]{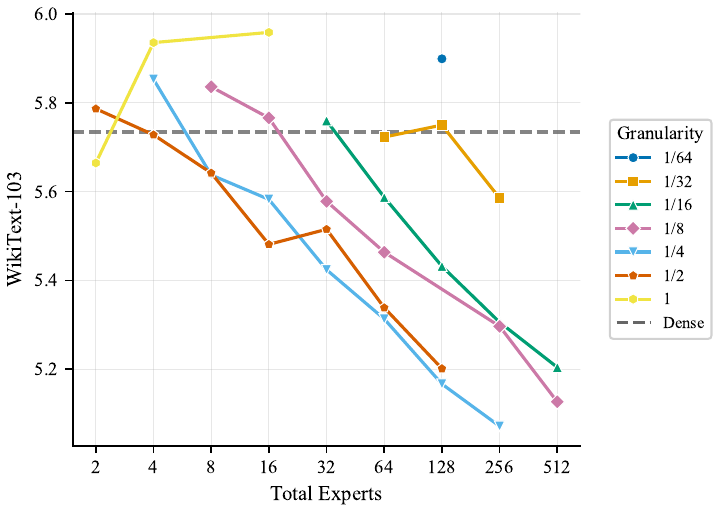}
        \end{subfigure}
        \begin{subfigure}[t]{0.33\textwidth}
            \centering
            \includegraphics[width=\linewidth]{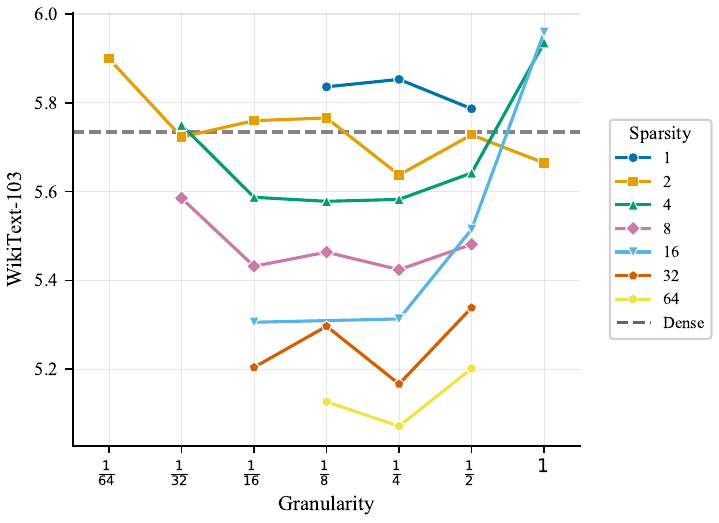}
        \end{subfigure}
        \caption{80M active, 80M - 765M total parameters}
    \end{subfigure}
    \par\bigskip\bigskip
        \begin{subfigure}[t]{\textwidth}
        \begin{subfigure}[t]{0.33\textwidth}
            \centering
            \includegraphics[width=\linewidth]{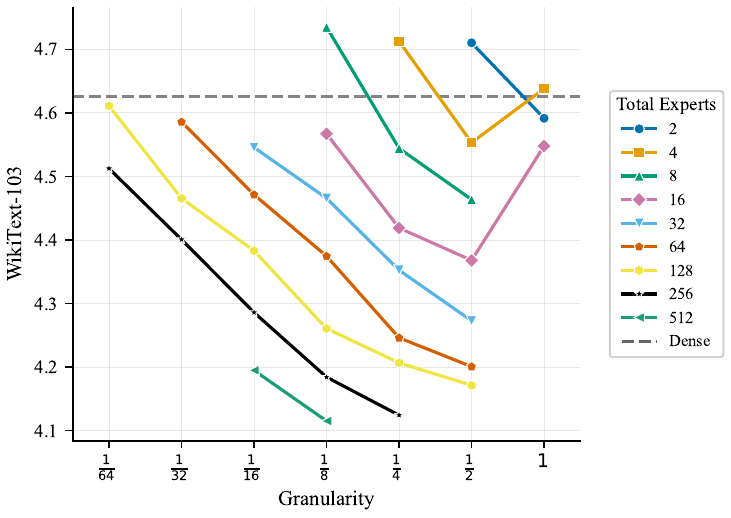}
        \end{subfigure}
        \begin{subfigure}[t]{0.33\textwidth}
            \centering
            \includegraphics[width=\linewidth]{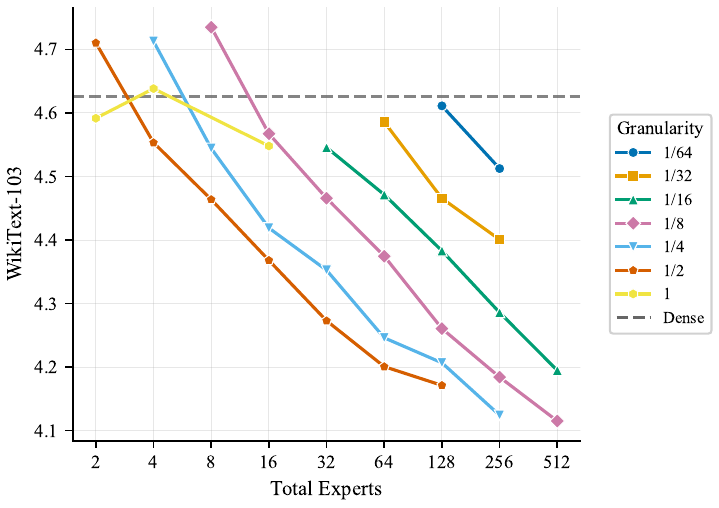}
        \end{subfigure}
        \begin{subfigure}[t]{0.33\textwidth}
            \centering
            \includegraphics[width=\linewidth]{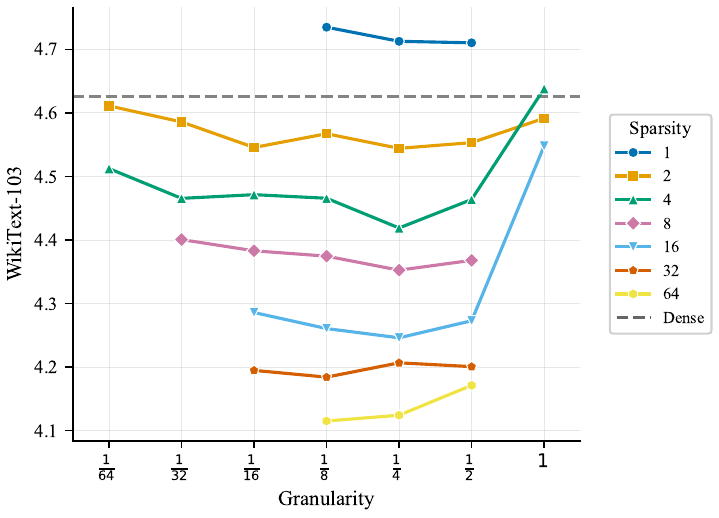}
        \end{subfigure}
        \caption{110M active, 110M - 1.4B total parameters}
    \end{subfigure}
    \end{figure*}

\clearpage  

\begin{figure*}[!ht]
        \addtocounter{figure}{-1}
    \begin{subfigure}[t]{\textwidth}
        \addtocounter{subfigure}{3}
        \begin{subfigure}[t]{0.33\textwidth}
            \centering
            \caption*{\scriptsize Fixed total experts (n)}
            \includegraphics[width=\linewidth]{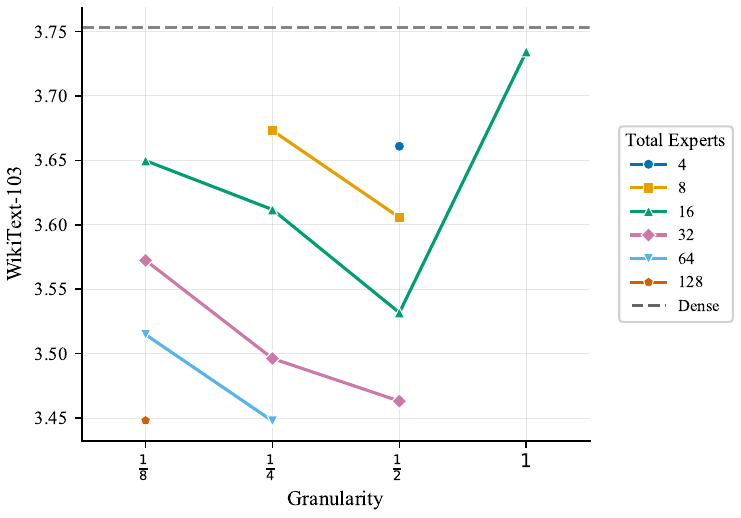}
        \end{subfigure}
        \begin{subfigure}[t]{0.33\textwidth}
            \centering
            \caption*{\scriptsize Fixed granularity (g)}
            \includegraphics[width=\linewidth]{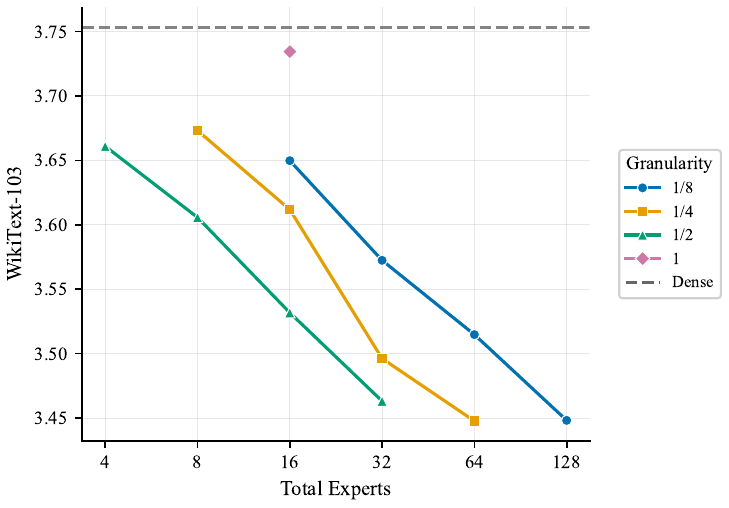}
        \end{subfigure}
        \begin{subfigure}[t]{0.33\textwidth}
            \centering
            \caption*{\scriptsize Fixed activation sparsity (s)}
            \includegraphics[width=\linewidth]{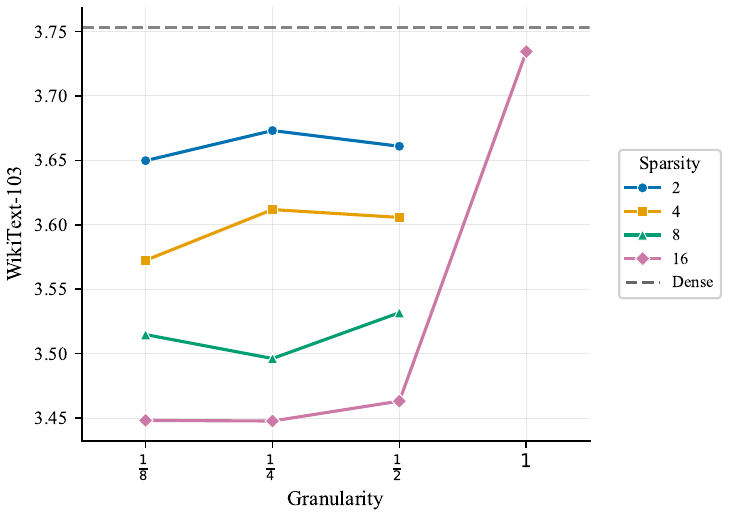}
        \end{subfigure}
        \caption{200M active, 200M - 3.3B total parameters}
    \end{subfigure}
    \par\bigskip\bigskip
        \begin{subfigure}[t]{\textwidth}
        \begin{subfigure}[t]{0.33\textwidth}
            \centering
            \includegraphics[width=\linewidth]{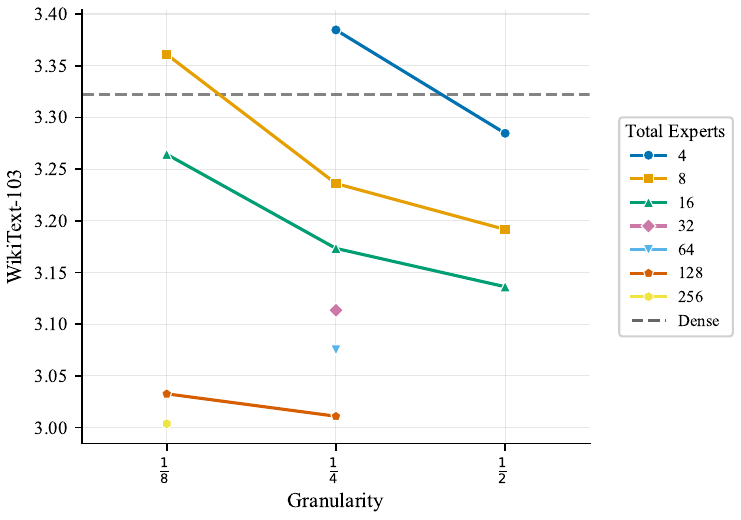}
        \end{subfigure}
        \begin{subfigure}[t]{0.33\textwidth}
            \centering
            \includegraphics[width=\linewidth]{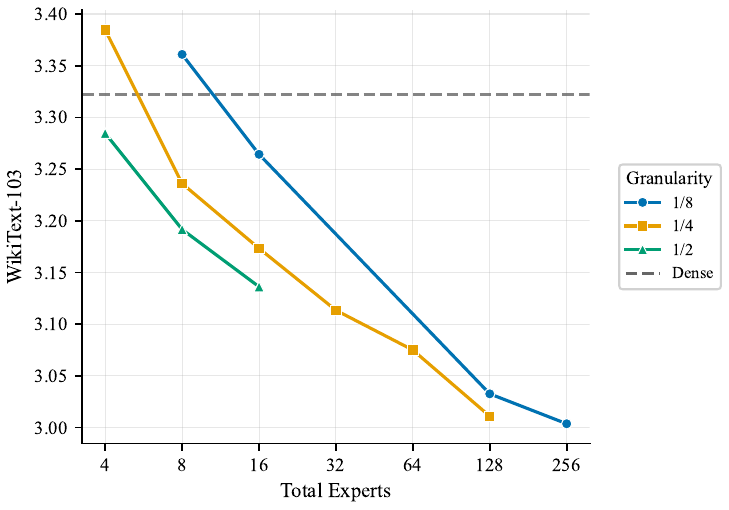}
        \end{subfigure}
        \begin{subfigure}[t]{0.33\textwidth}
            \centering
            \includegraphics[width=\linewidth]{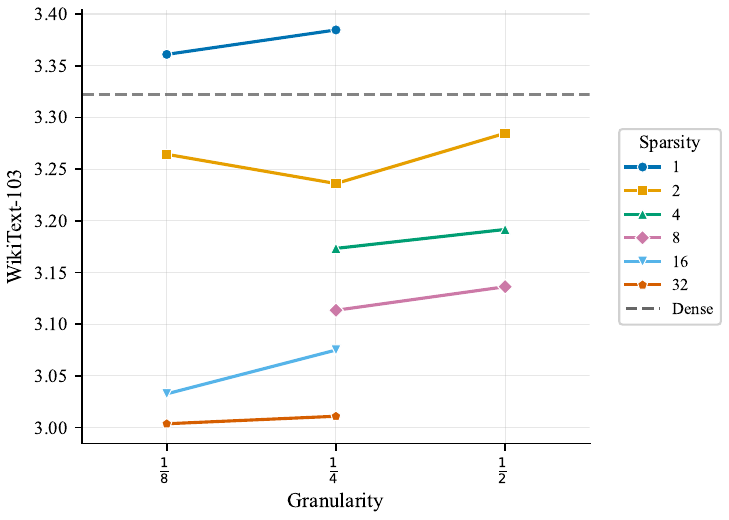}
        \end{subfigure}
        \caption{300M active, 300M - 6.6B total parameters}
    \end{subfigure}

    \caption{
    \textbf{Increasing inactive expert parameters via expert size (left) or total count (center) improves performance in MoEs (\S\ref{sec:expt_main}).} This effect is seen both when holding total number of experts fixed (left) and when holding expert granularity fixed (center). In general, increasing total parameters results in improved performance.  \textbf{Optimal tradeoff between expert count and granularity varies in MoEs (right). (\S\ref{sec:expt_main})}
    At each activation sparsity $s$ (equivalently, at each total parameter count), the optimal (total expert count, expert granularity) configuration varies. As $s$ increases, optimal expert granularity remains nearly fixed, suggesting that sparsity should be scaled up primarily by increasing total expert count $n$, while maintaining a near constant, slowly increasing expert granularity $g$. 
    }
    \label{fig:wikitext_103_experts}
\end{figure*}

\begin{figure*}[!ht]
    \centering
    
    \begin{subfigure}[t]{0.46\textwidth}
        \centering
        \includegraphics[width=\linewidth]{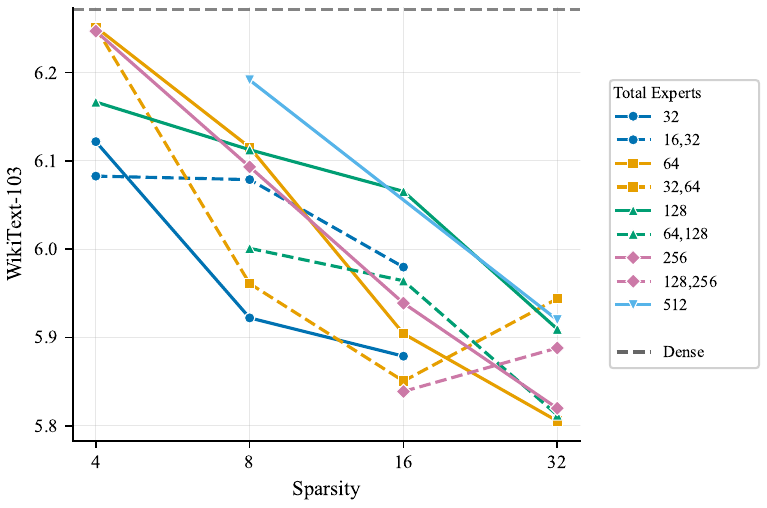}
        \caption{50M active, 50M - 930M total parameters}
    \end{subfigure}
    \vspace{1em}
    \begin{subfigure}[t]{0.46\textwidth}
        \centering
        \includegraphics[width=\linewidth]{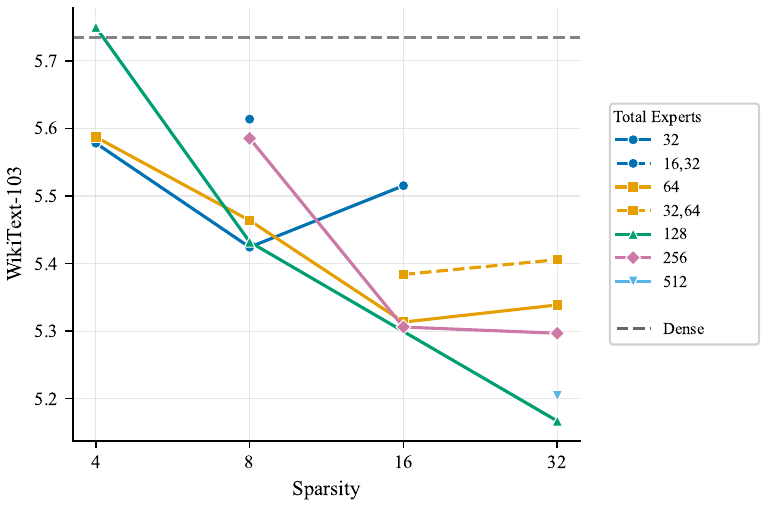}
        \caption{80M active, 80M - 765M total parameters}
    \end{subfigure}
    \caption{
    \textbf{Heterogeneity of expert size alone does not improve MoE performance (\S\ref{sec:expt_hetgen}).} To explore the potential benefits of their architectural flexibility, we compare heterogeneous MoEs (indicated by dotted lines) to active- and total-parameter-matched homogeneous MoEs. Heterogeneity alone does not result in performance gains, as, at each activation sparsity $s$, heterogeneous MoEs with $n_1, n_2 = a, b$ lie between or near the 2 closest homogeneous MoEs, with $n=a$ and with $n=b$.
    }
    \label{fig:wikitext_103_het}
\end{figure*}

\begin{figure*}[!ht]
    \centering
    
    \begin{subfigure}[t]{1.0\textwidth}
        \centering
        \includegraphics[width=\linewidth]{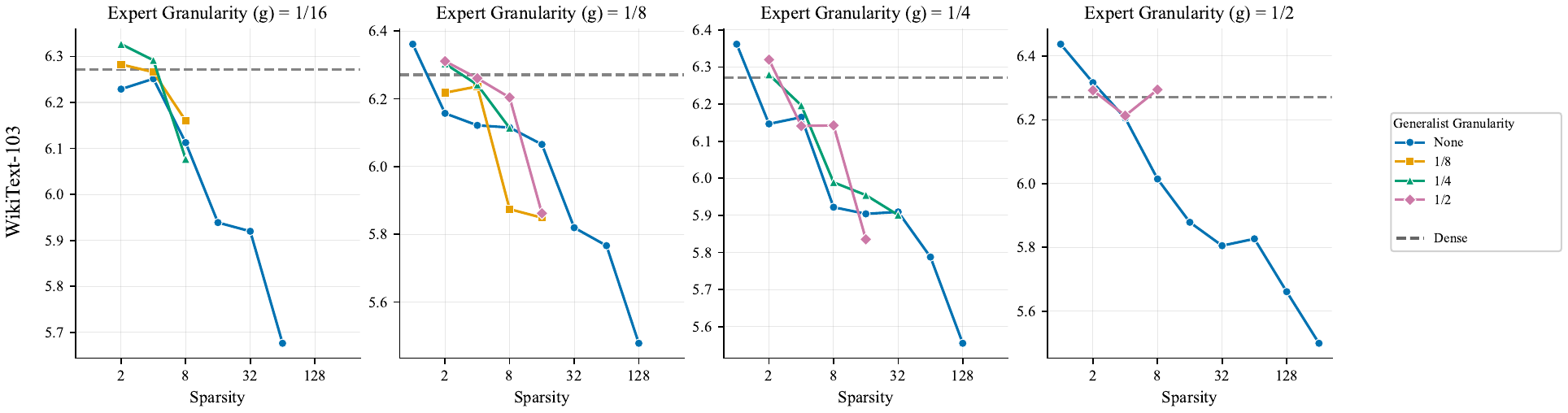}
        \caption{50M active, 50M - 930M total parameters}
    \end{subfigure}
    \par\bigskip\bigskip
    \begin{subfigure}[t]{1.0\textwidth}
        \centering
        \includegraphics[width=\linewidth]{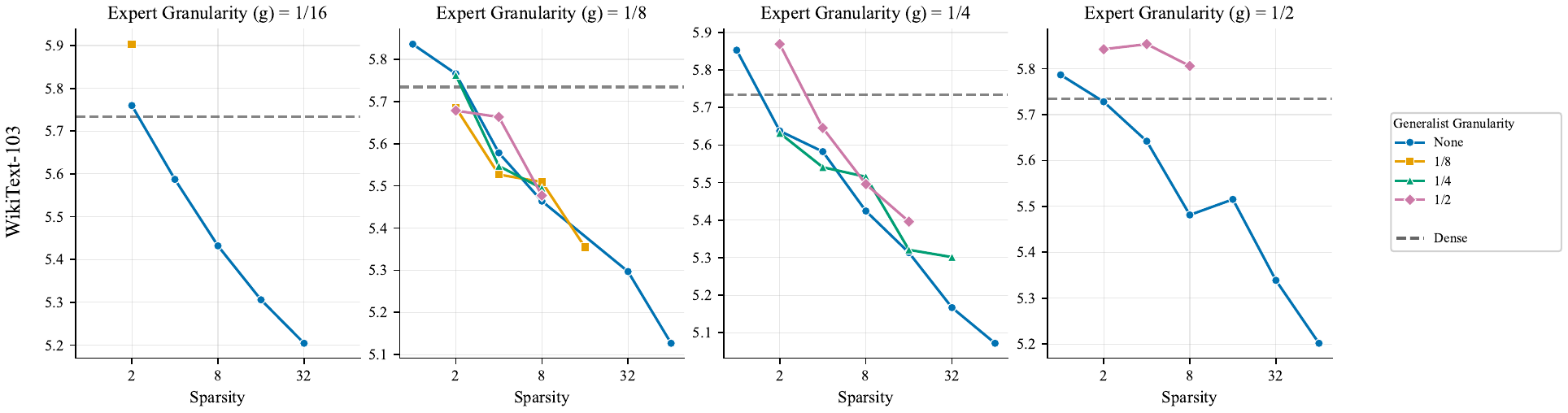}
        \caption{80M active, 80M - 765M total parameters}
    \end{subfigure}
    \par\bigskip\bigskip
    \begin{subfigure}[t]{1.0\textwidth}
        \centering
        \includegraphics[width=\linewidth]{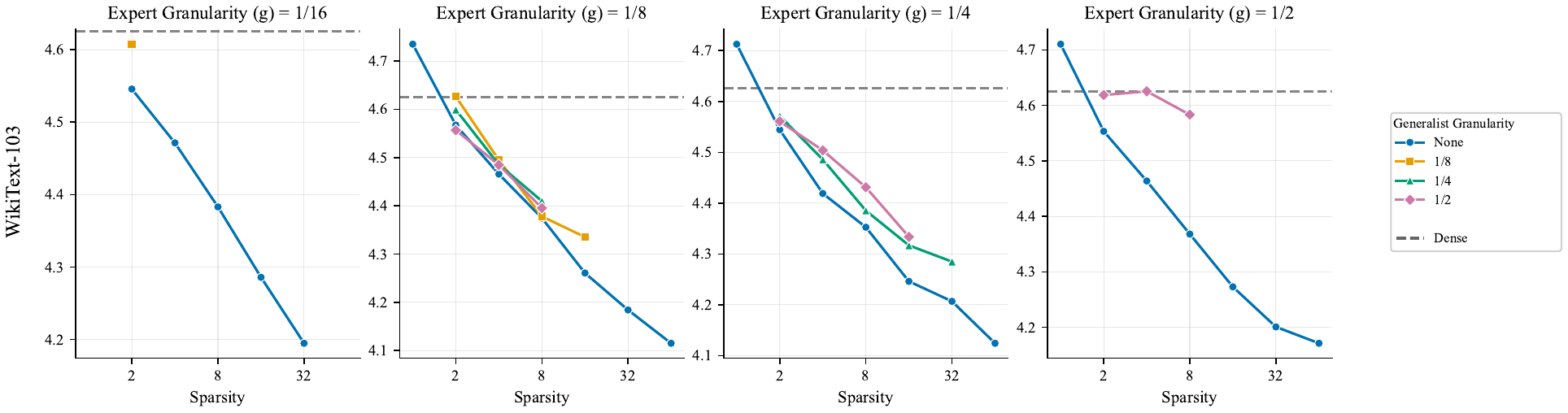}
        \caption{110M active, 110M - 1.4B total parameters}
    \end{subfigure}
    \caption{
    \textbf{The inclusion of a generalist consistently degrades performance in homogeneous MoEs (\S\ref{sec:expt_hetgen}).}
    We train MoE LMs which consist of some routed experts with granularity $g$, as well as a generalist with granularity $g_{gen}\in \{\frac{1}{2}, \frac{1}{4}, \frac{1}{8}\} $. We compare to settings with no generalist, only routed experts with granularity $g$. In all settings and configurations, the addition of any granularity generalist results in comparable or degraded performance. 
    }
    \label{fig:wikitext_103_gen}
\end{figure*}

\begin{figure*}[ht]
    \centering
    \begin{subfigure}[t]{1.0\textwidth}
        \centering
        \includegraphics[width=\linewidth]{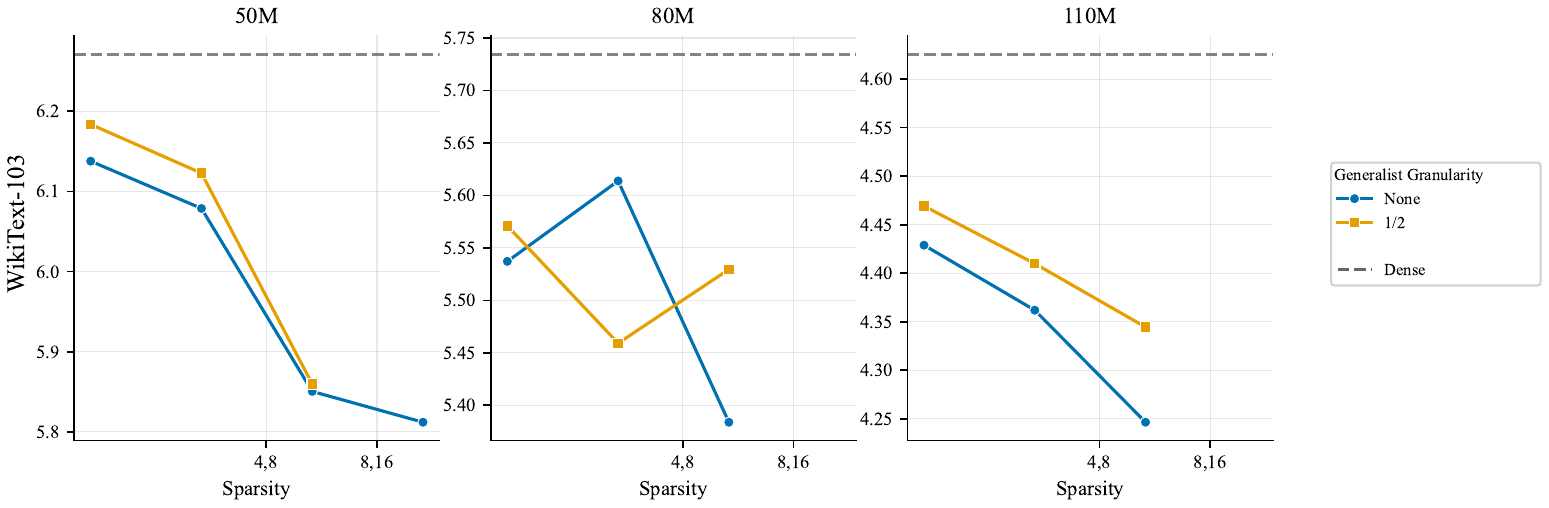}
    \end{subfigure}
    \caption{
    \textbf{The inclusion of a generalist consistently degrades performance in heterogeneous MoEs (\S\ref{sec:expt_hetgen}).}
    We train heterogeneous MoE LMs which consist of  routed experts with granularity $g_1, g_2$, as well as a generalist with granularity $g_{gen} = \frac{1}{2}$. We compare to settings with no generalist. In all settings and configurations, the addition of a generalist results in comparable or degraded performance. 
    }
    \label{fig:wikitext_103_hetgen}
\end{figure*}

\begin{figure*}[ht]
    \centering
    \begin{subfigure}[t]{\textwidth}
        \centering
        \begin{subfigure}[t]{0.45\textwidth}
            \includegraphics[width=\linewidth]{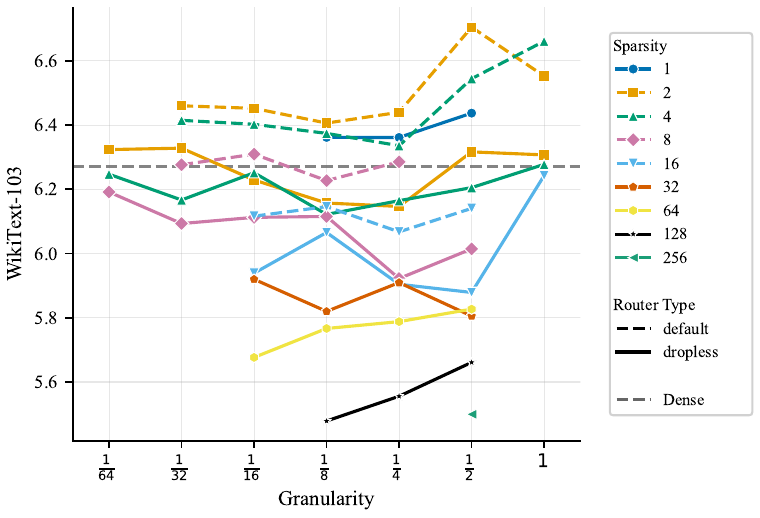}
            \caption{50M active, 50M - 930M total parameters}
        \end{subfigure}
    \hspace{1em}
        \begin{subfigure}[t]{0.45\textwidth}
            \centering
            \includegraphics[width=\linewidth]{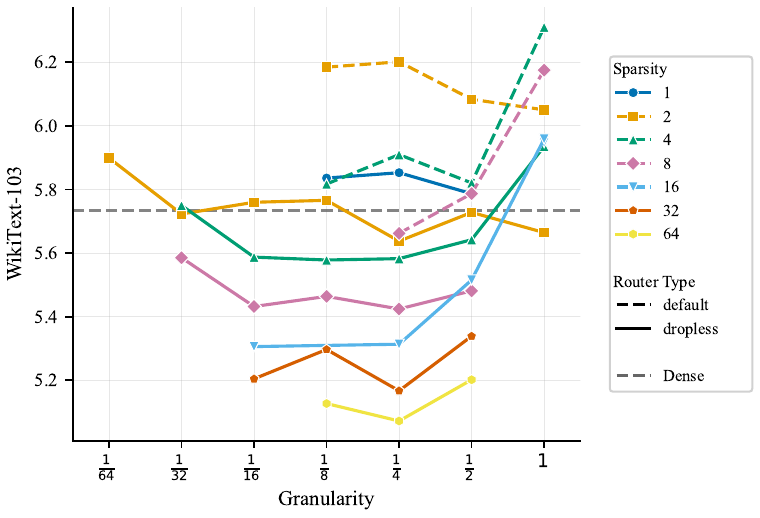}
            \caption{80M active, 80M - 765M total parameters}
        \end{subfigure}
    \end{subfigure}

    \par\bigskip\bigskip
    \begin{subfigure}[t]{0.45\textwidth}
        \centering
        \includegraphics[width=\linewidth]{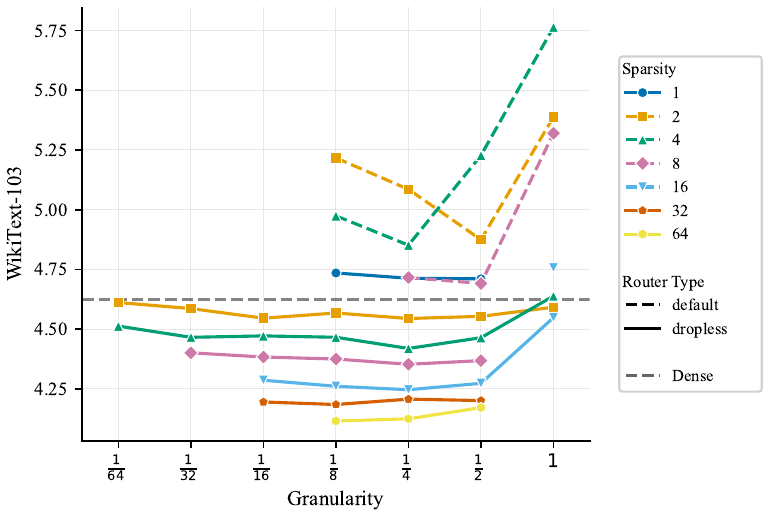}
        \caption{110M active, 110M - 1.4B total parameters}
    \end{subfigure}
    \caption{ 
    \textbf{Dropless routing outperforms default routing (\S\ref{sec:expt_router}).}
    We compare dropless routing to the default setting, which allow tokens to be dropped. Across all scales, we find that dropless routing outperforms or performs comparably to default routing. 
    }
    \label{fig:wikitext_103_dropless}
\end{figure*}

\begin{figure*}[ht]
    \centering
    \begin{subfigure}[t]{0.45\textwidth}
        \centering
        \includegraphics[width=\linewidth]{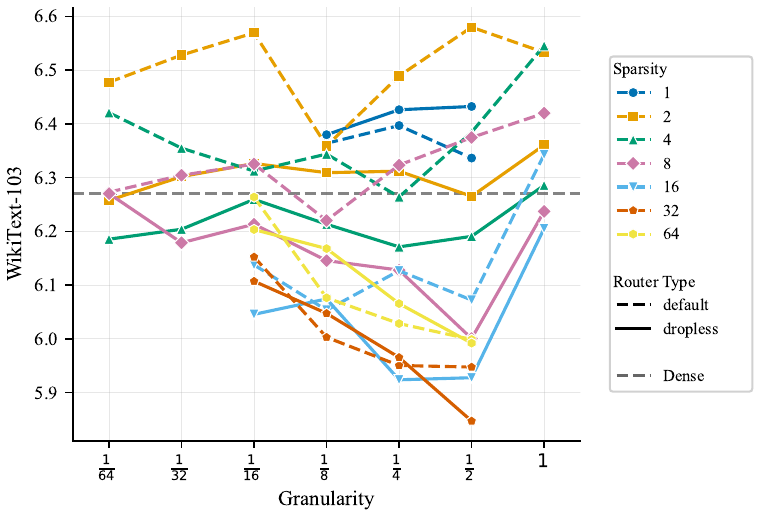}
        \caption{50M active, 50M - 930M total parameters}
    \end{subfigure}
    \hspace{1em}
    \begin{subfigure}[t]{0.45\textwidth}
        \centering
        \includegraphics[width=\linewidth]{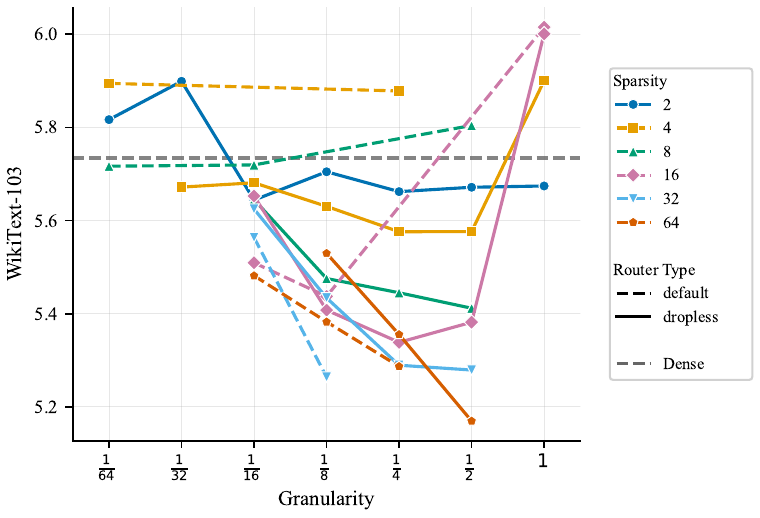}
        \caption{80M active, 80M - 765M total parameters}
    \end{subfigure}
    \caption{
    \textbf{Dropless routing, with bias $\gamma=\num{1e-3}$ (\S\ref{sec:expt_router}).} 
    As in Figure~\ref{fig:lm_avg_dropless}, we compare dropless routing to the default setting, which allow tokens to be dropped. Across all scales, we find that dropless routing outperforms or performs comparably to default routing. We see here with additional higher sparsity default routing runs that as sparsity increases, default routing performance approaches that of dropless routing.
    }
    \label{fig:wikitext_103_dropless_with_lf}
\end{figure*}

\begin{figure*}[ht]
    \centering
    \begin{subfigure}[]{\textwidth}
        \centering
        \includegraphics[width=0.46\linewidth]{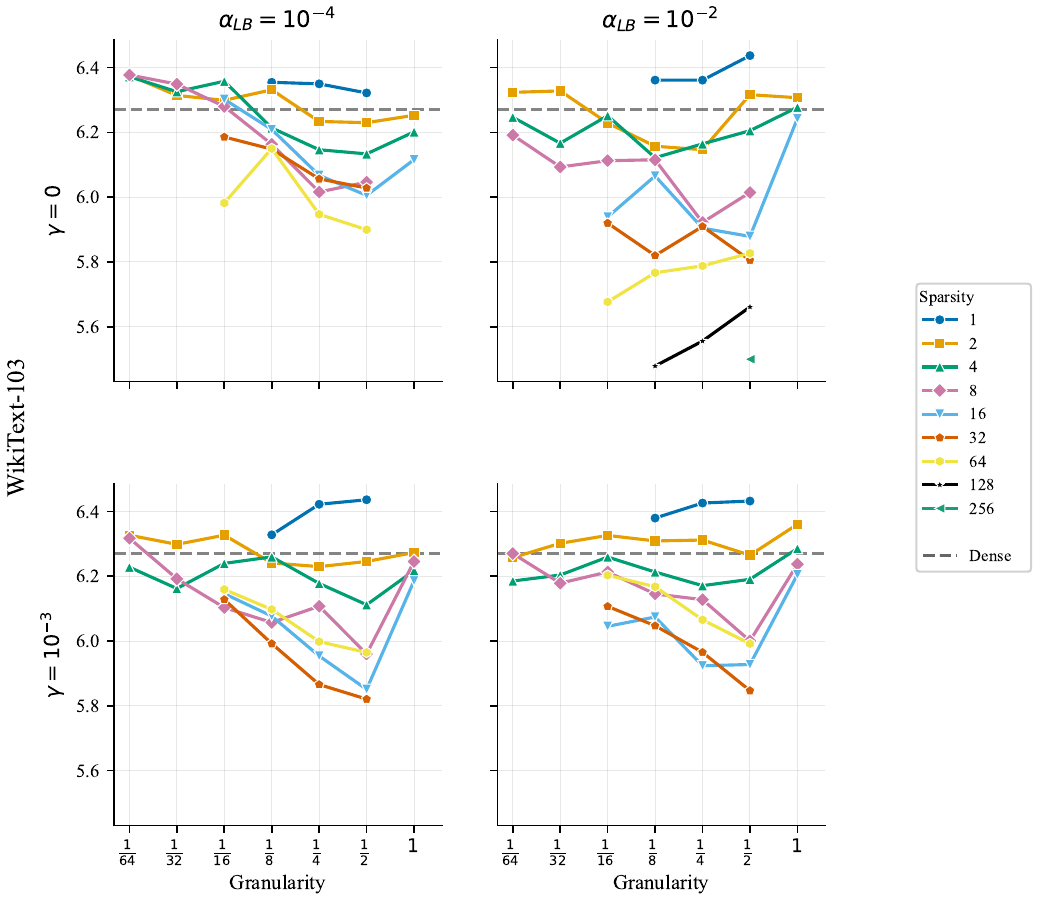}
        \hspace{1em}
        \includegraphics[width=0.46\linewidth]{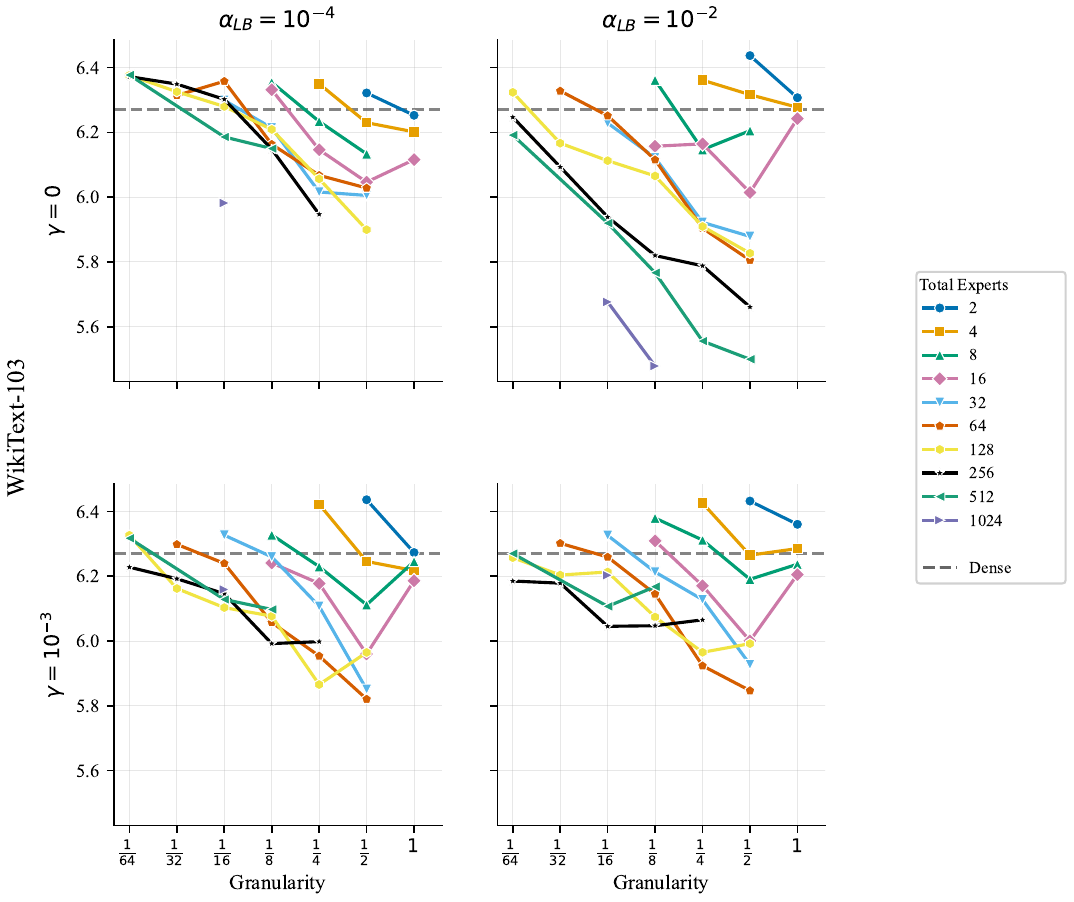}
        \caption{50M active, 50M - 930M total parameters}
    \end{subfigure}
    \par\bigskip\bigskip
    \begin{subfigure}[]{\textwidth}
        \centering
        \includegraphics[width=0.46\linewidth]{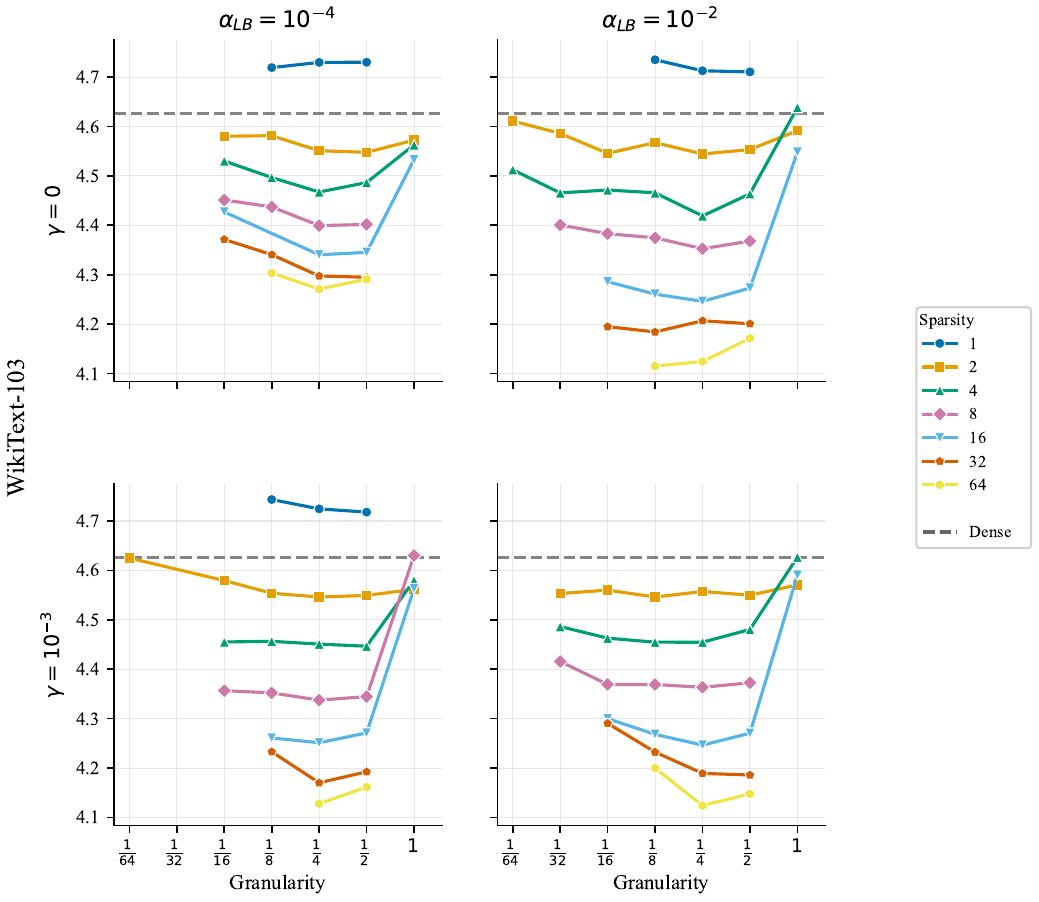}
        \hspace{1em}
        \includegraphics[width=0.46\linewidth]{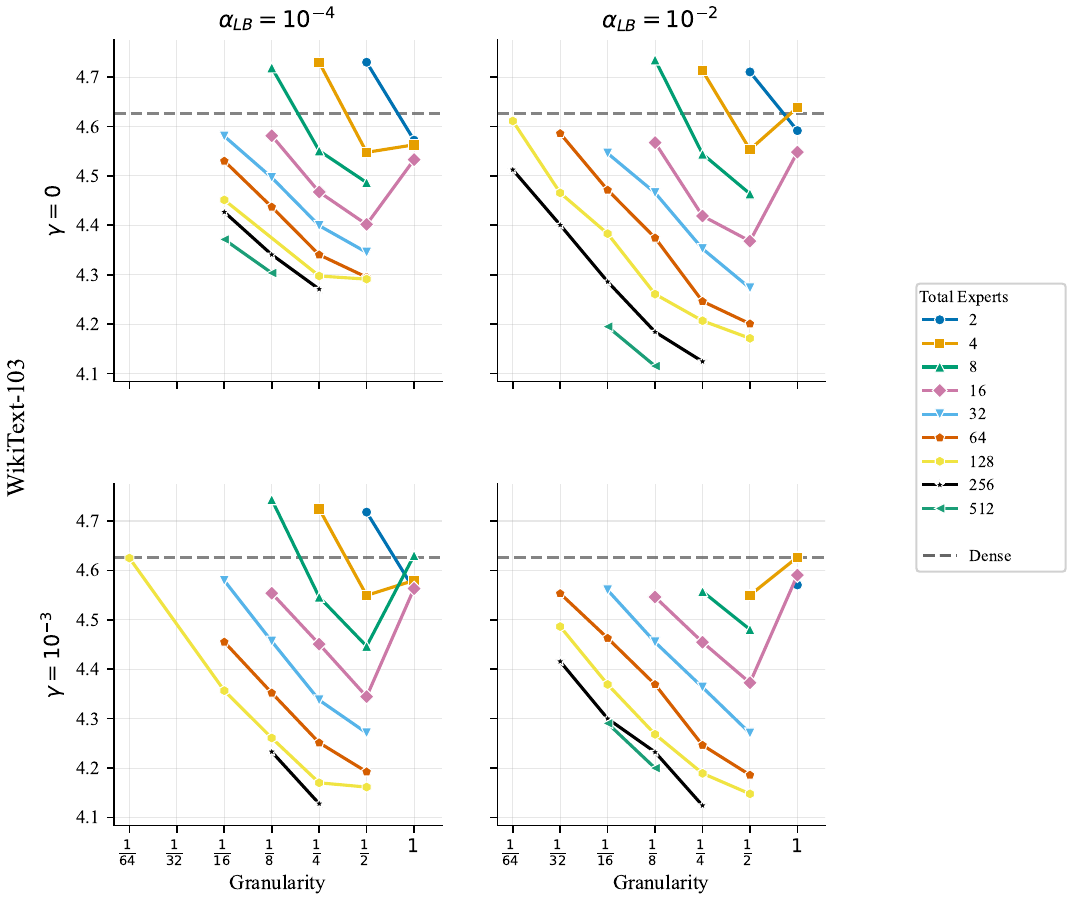}
        \caption{80M active, 80M - 765M total parameters}
    \end{subfigure}
    \par\bigskip\bigskip
    \begin{subfigure}[t]{\textwidth}
        \centering
        \includegraphics[width=0.46\linewidth]{figures/lm/wikitext_103-validation/ce_loss/lb_sweep_hgn_gxs_110M.pdf}
        \hspace{1em}
        \includegraphics[width=0.46\linewidth]{figures/lm/wikitext_103-validation/ce_loss/lb_sweep_hgn_gxn_110M.pdf}
        \caption{110M active, 110M - 1.4B total parameters}
    \end{subfigure}

    \end{figure*} 

\clearpage  

\begin{figure*}[ht]
    \addtocounter{figure}{-1}
    \centering
    \begin{subfigure}[t]{\textwidth}
        \centering
        \includegraphics[width=0.46\linewidth]{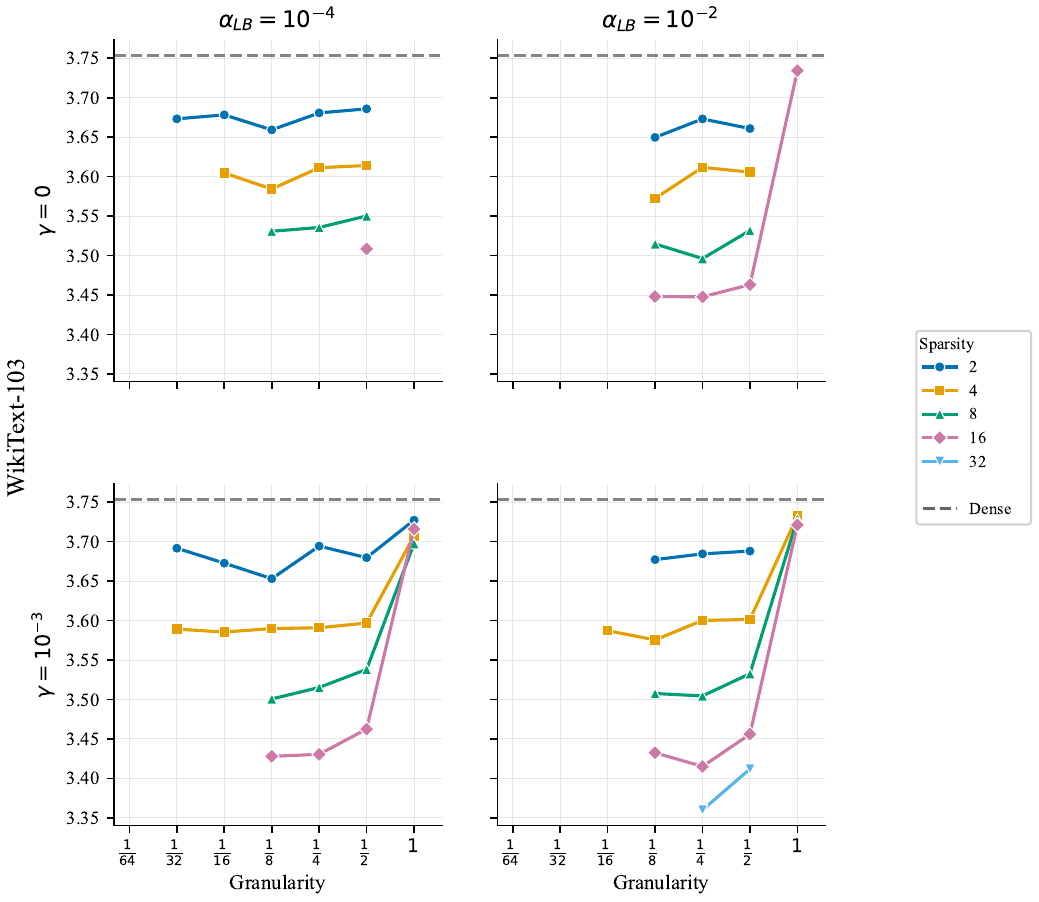}
        \hspace{1em}
        \includegraphics[width=0.46\linewidth]{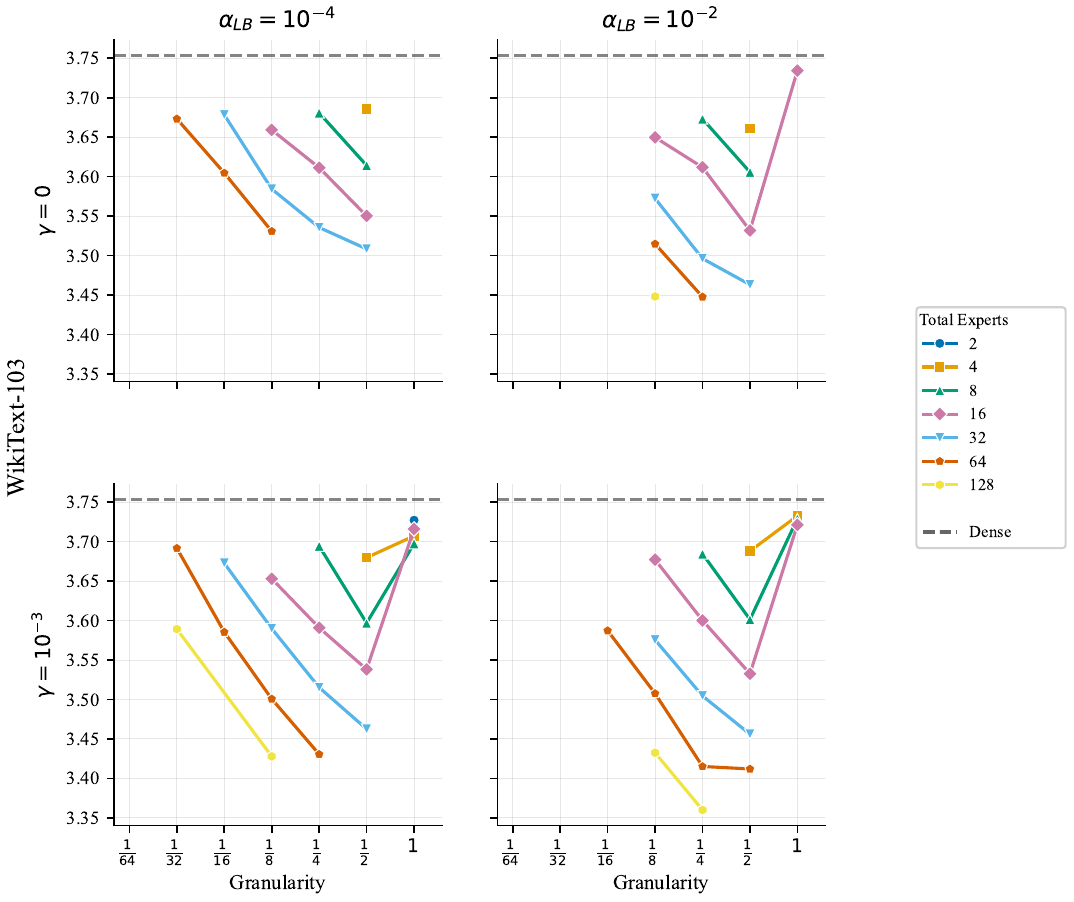}
        \caption{200M active, 200M - 3.3B total parameters}
    \end{subfigure}
    \par\bigskip\bigskip
    \begin{subfigure}[t]{\textwidth}
        \centering
        \includegraphics[width=0.3\linewidth]{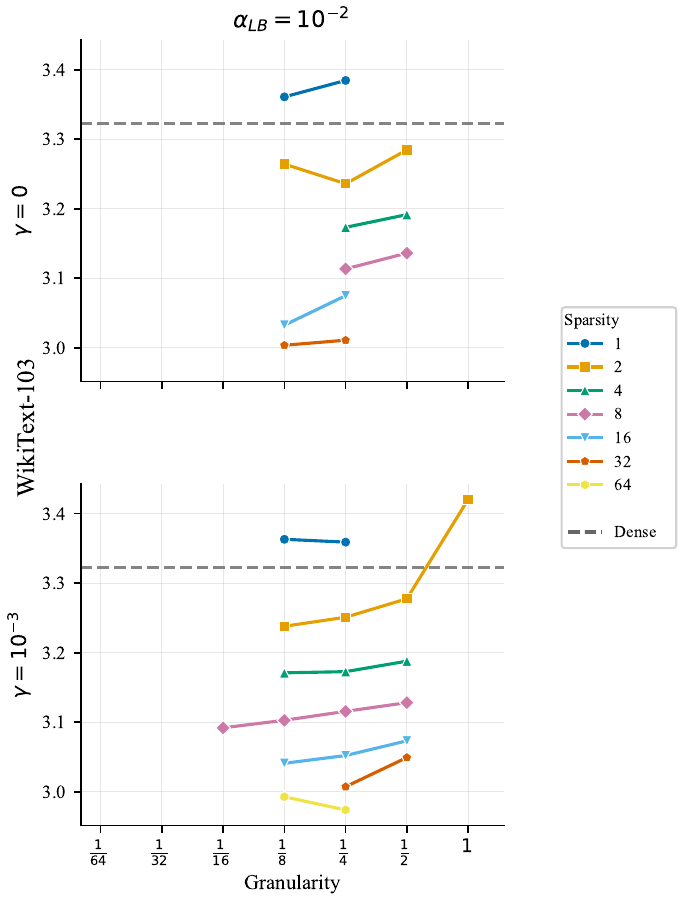}
        \hspace{1em}
        \includegraphics[width=0.3\linewidth]{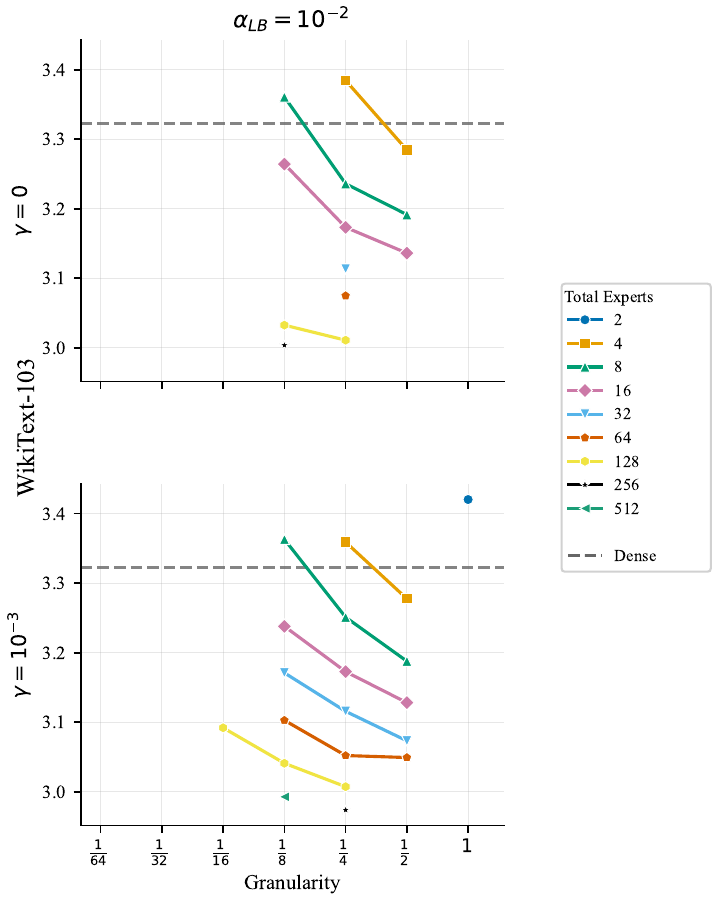}
        \caption{300M active, 300M - 6.6B total parameters}
    \end{subfigure}

    \caption{
    \textbf{Load balancing mechanisms must be tuned correctly (\S\ref{sec:expt_router}).}
    We consider load balancing loss weight $\alpha_{LB} \in \{\num{1e-2}, \num{1e-4}\}$ and loss-free load balancing with bias $\gamma\in\{0, \num{1e-3}\}$ ($\gamma=0$ indicates no loss-free mechanism). Results show that poorly chosen hyperparameters, such as high bias $\gamma = 1e-3$ with total experts $n\geq 512$, may impair performance. However, all settings other than $(\alpha_{LB}=\num{1e-2}, \gamma=\num{1e-3})$ perform comparably for $n \leq 512$, suggesting that a wide range of load balancing settings achieve near-optimal performance. 
    }
    \label{fig:wikitext_103_lb}
\end{figure*}

%% file: fig_tex/downstream/boolq.tex
\begin{figure*}[!ht]
    \centering
        \begin{subfigure}[t]{\textwidth}
        \begin{subfigure}[t]{0.33\textwidth}
            \centering
            \caption*{\scriptsize Fixed total experts (n)}
            \includegraphics[width=\linewidth]{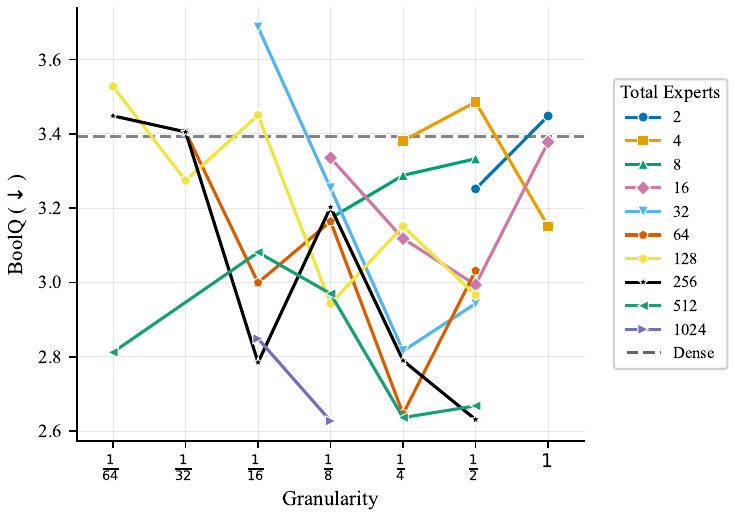}
        \end{subfigure}
        \begin{subfigure}[t]{0.33\textwidth}
            \centering
            \caption*{\scriptsize Fixed granularity (g)}
            \includegraphics[width=\linewidth]{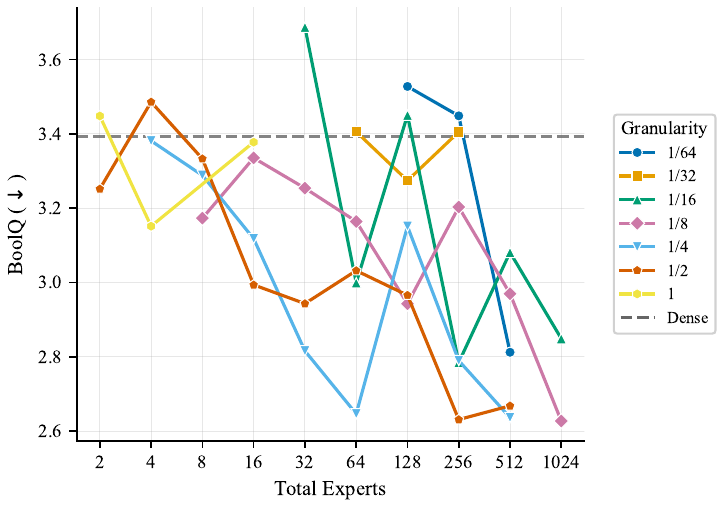}
        \end{subfigure}
        \begin{subfigure}[t]{0.33\textwidth}
            \centering
            \caption*{\scriptsize Fixed activation sparsity (s)}
            \includegraphics[width=\linewidth]{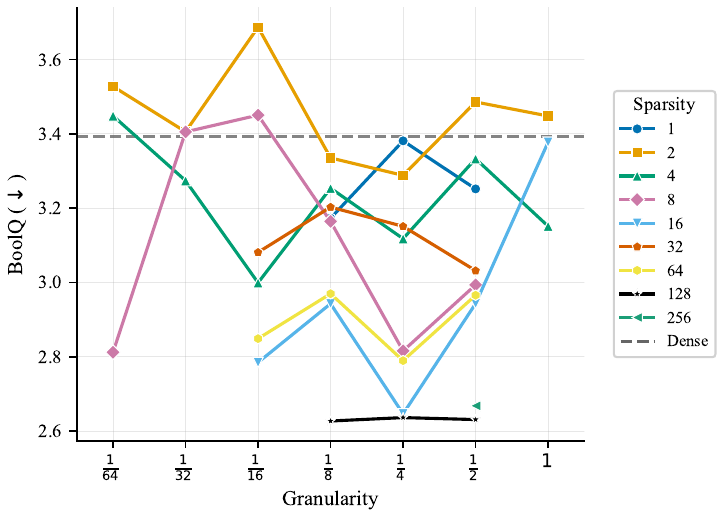}
        \end{subfigure}
        \caption{50M active, 50M - 930M total parameters}
    \end{subfigure}
\par\bigskip\bigskip
    \begin{subfigure}[t]{\textwidth}
        \begin{subfigure}[t]{0.33\textwidth}
            \centering
            \includegraphics[width=\linewidth]{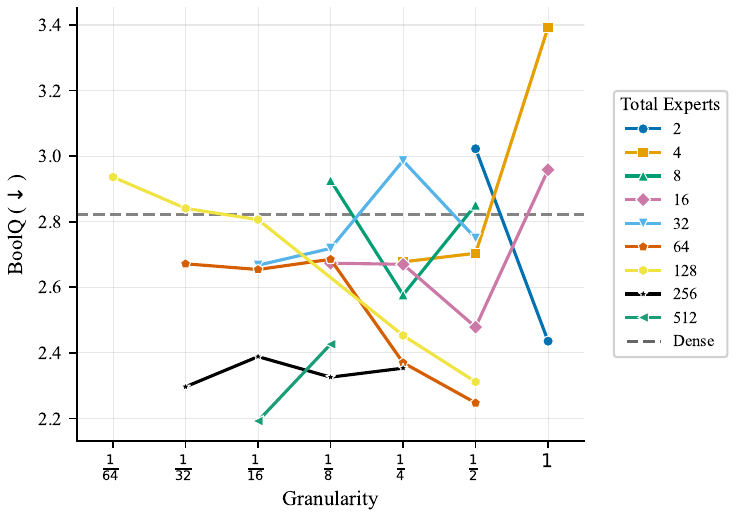}
        \end{subfigure}
        \begin{subfigure}[t]{0.33\textwidth}
            \centering
            \includegraphics[width=\linewidth]{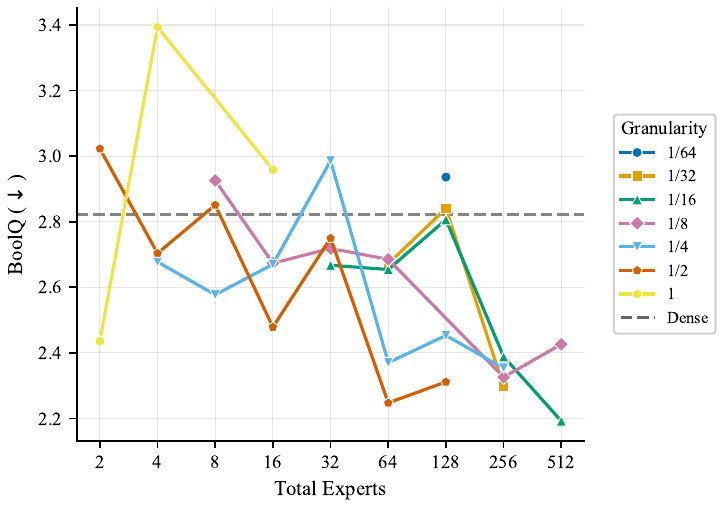}
        \end{subfigure}
        \begin{subfigure}[t]{0.33\textwidth}
            \centering
            \includegraphics[width=\linewidth]{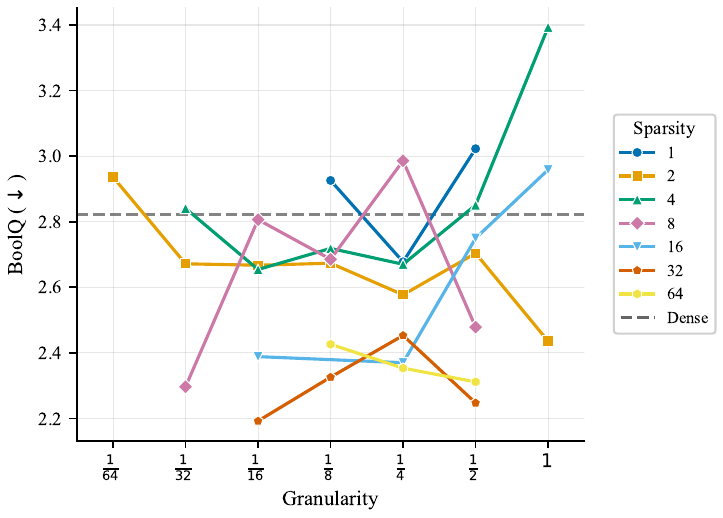}
        \end{subfigure}
        \caption{80M active, 80M - 765M total parameters}
    \end{subfigure}
    \par\bigskip\bigskip
        \begin{subfigure}[t]{\textwidth}
        \begin{subfigure}[t]{0.33\textwidth}
            \centering
            \includegraphics[width=\linewidth]{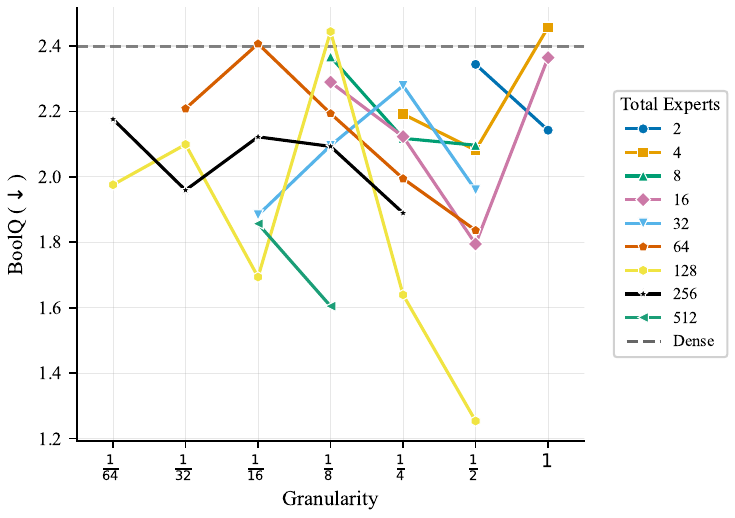}
        \end{subfigure}
        \begin{subfigure}[t]{0.33\textwidth}
            \centering
            \includegraphics[width=\linewidth]{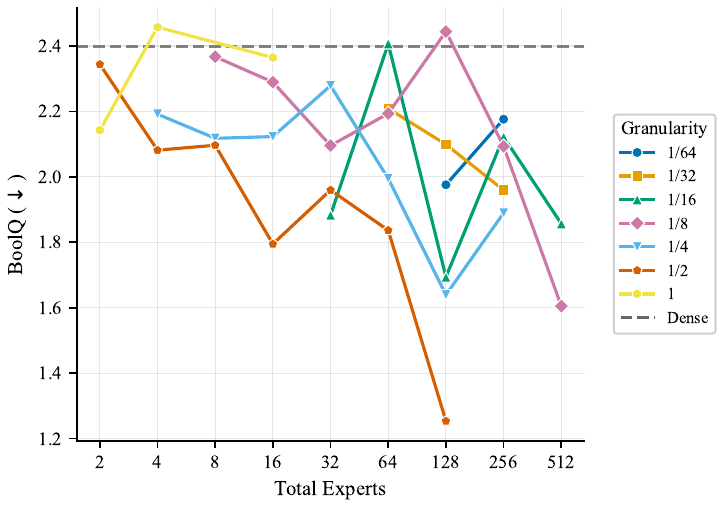}
        \end{subfigure}
        \begin{subfigure}[t]{0.33\textwidth}
            \centering
            \includegraphics[width=\linewidth]{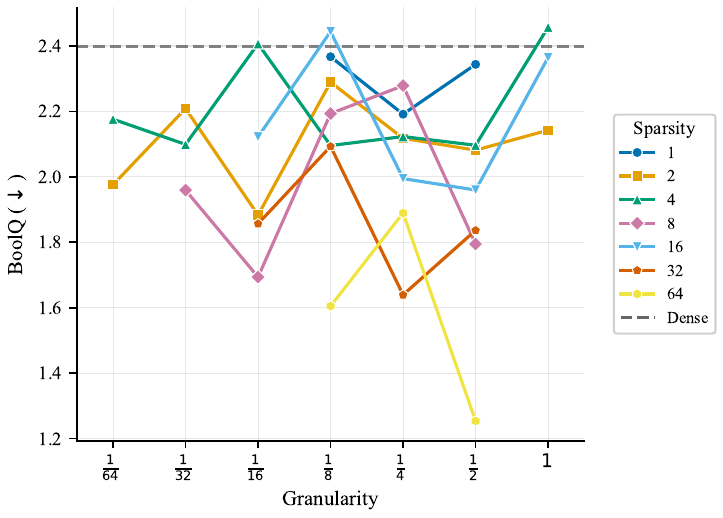}
        \end{subfigure}
        \caption{110M active, 110M - 1.4B total parameters}
    \end{subfigure}
    \end{figure*}

\clearpage  

\begin{figure*}[!ht]
        \addtocounter{figure}{-1}
    \begin{subfigure}[t]{\textwidth}
        \addtocounter{subfigure}{3}
        \begin{subfigure}[t]{0.33\textwidth}
            \centering
            \caption*{\scriptsize Fixed total experts (n)}
            \includegraphics[width=\linewidth]{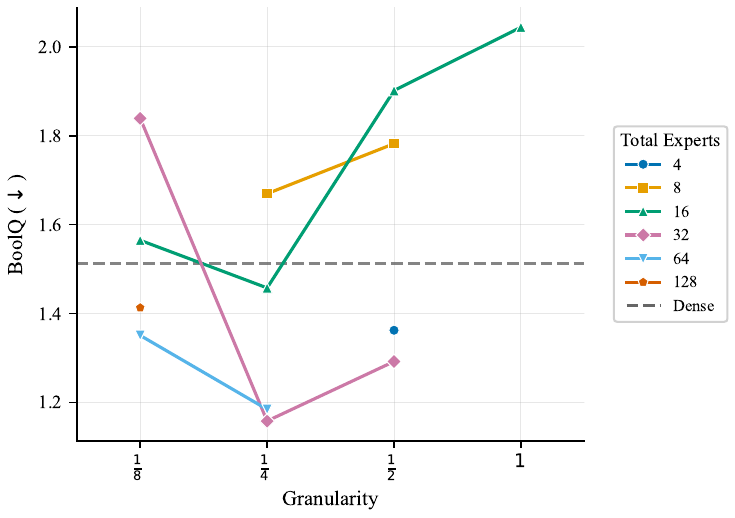}
        \end{subfigure}
        \begin{subfigure}[t]{0.33\textwidth}
            \centering
            \caption*{\scriptsize Fixed granularity (g)}
            \includegraphics[width=\linewidth]{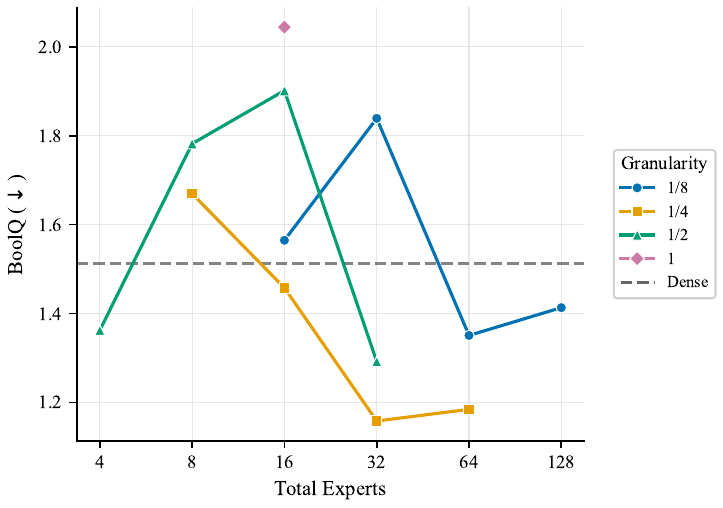}
        \end{subfigure}
        \begin{subfigure}[t]{0.33\textwidth}
            \centering
            \caption*{\scriptsize Fixed activation sparsity (s)}
            \includegraphics[width=\linewidth]{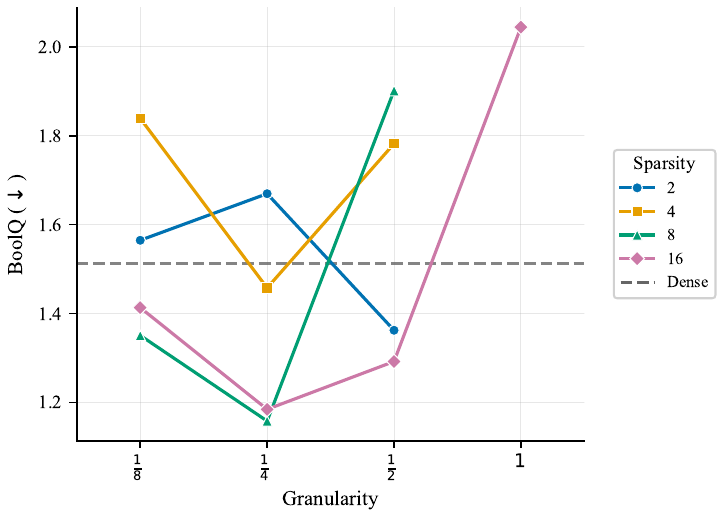}
        \end{subfigure}
        \caption{200M active, 200M - 3.3B total parameters}
    \end{subfigure}
    \par\bigskip\bigskip
        \begin{subfigure}[t]{\textwidth}
        \begin{subfigure}[t]{0.33\textwidth}
            \centering
            \includegraphics[width=\linewidth]{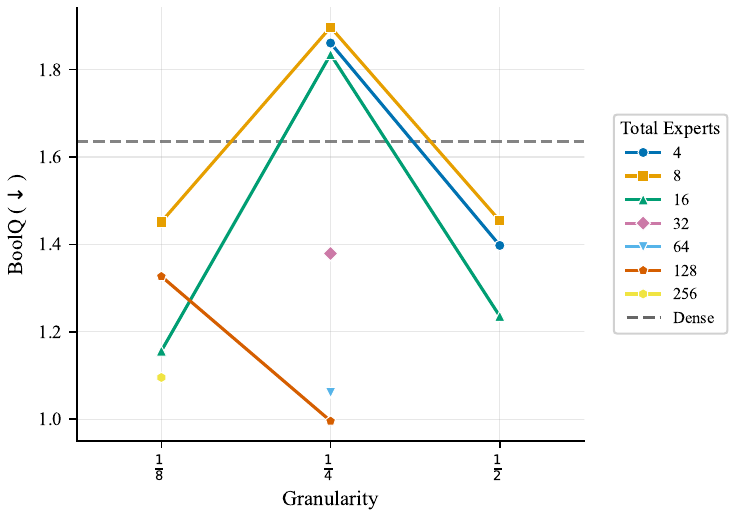}
        \end{subfigure}
        \begin{subfigure}[t]{0.33\textwidth}
            \centering
            \includegraphics[width=\linewidth]{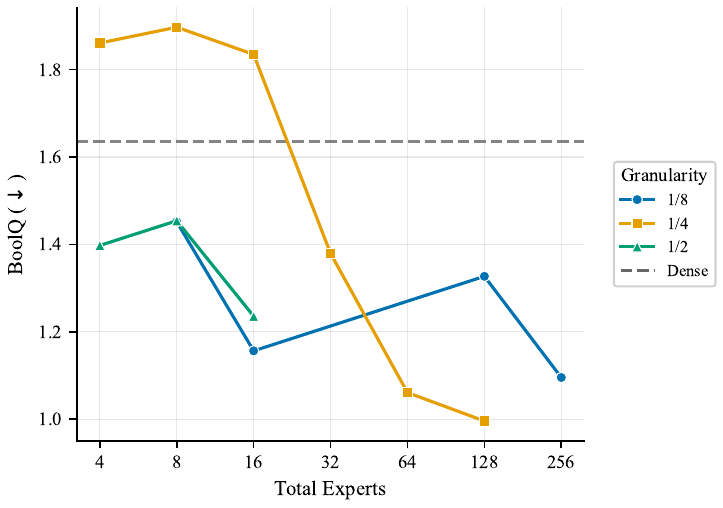}
        \end{subfigure}
        \begin{subfigure}[t]{0.33\textwidth}
            \centering
            \includegraphics[width=\linewidth]{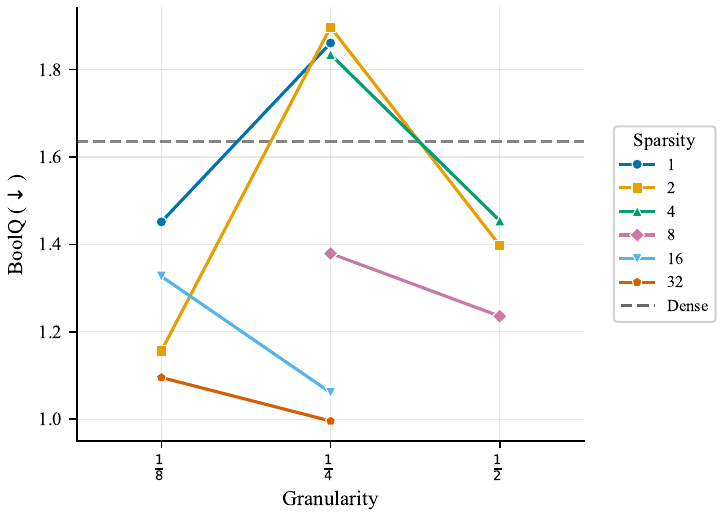}
        \end{subfigure}
        \caption{300M active, 300M - 6.6B total parameters}
    \end{subfigure}

    \caption{
    \textbf{Increasing inactive expert parameters via expert size (left) or total count (center) improves performance in MoEs (\S\ref{sec:expt_main}).} This effect is seen both when holding total number of experts fixed (left) and when holding expert granularity fixed (center). In general, increasing total parameters results in improved performance.  \textbf{Optimal tradeoff between expert count and granularity varies in MoEs (right). (\S\ref{sec:expt_main})}
    At each activation sparsity $s$ (equivalently, at each total parameter count), the optimal (total expert count, expert granularity) configuration varies. As $s$ increases, optimal expert granularity remains nearly fixed, suggesting that sparsity should be scaled up primarily by increasing total expert count $n$, while maintaining a near constant, slowly increasing expert granularity $g$. 
    }
    \label{fig:boolq_experts}
\end{figure*}

\begin{figure*}[!ht]
    \centering
    
    \begin{subfigure}[t]{0.46\textwidth}
        \centering
        \includegraphics[width=\linewidth]{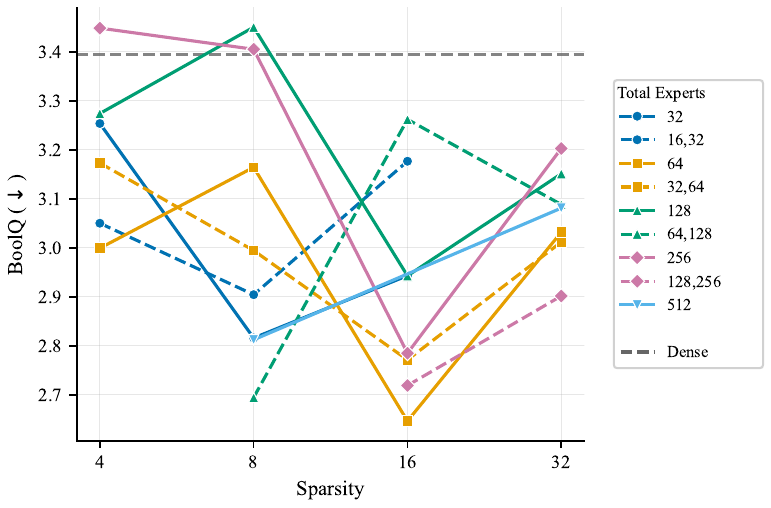}
        \caption{50M active, 50M - 930M total parameters}
    \end{subfigure}
    \vspace{1em}
    \begin{subfigure}[t]{0.46\textwidth}
        \centering
        \includegraphics[width=\linewidth]{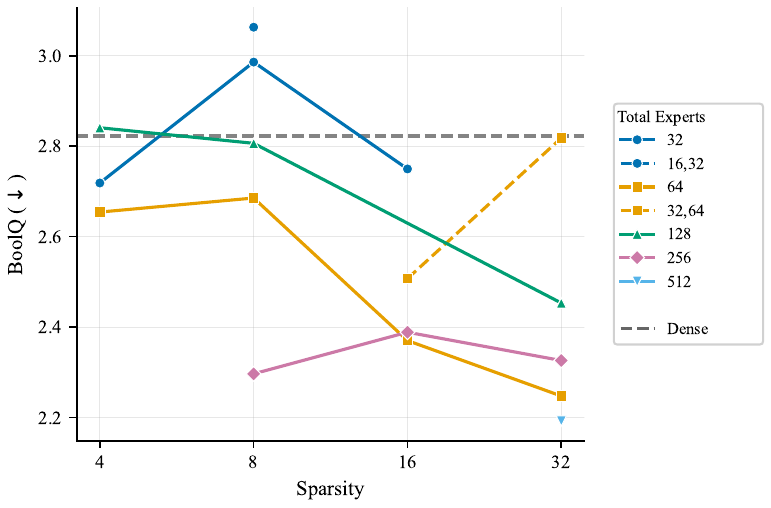}
        \caption{80M active, 80M - 765M total parameters}
    \end{subfigure}
    \caption{
    \textbf{Heterogeneity of expert size alone does not improve MoE performance (\S\ref{sec:expt_hetgen}).} To explore the potential benefits of their architectural flexibility, we compare heterogeneous MoEs (indicated by dotted lines) to active- and total-parameter-matched homogeneous MoEs. Heterogeneity alone does not result in performance gains, as, at each activation sparsity $s$, heterogeneous MoEs with $n_1, n_2 = a, b$ lie between or near the 2 closest homogeneous MoEs, with $n=a$ and with $n=b$.
    }
    \label{fig:boolq_het}
\end{figure*}

\begin{figure*}[!ht]
    \centering
    
    \begin{subfigure}[t]{1.0\textwidth}
        \centering
        \includegraphics[width=\linewidth]{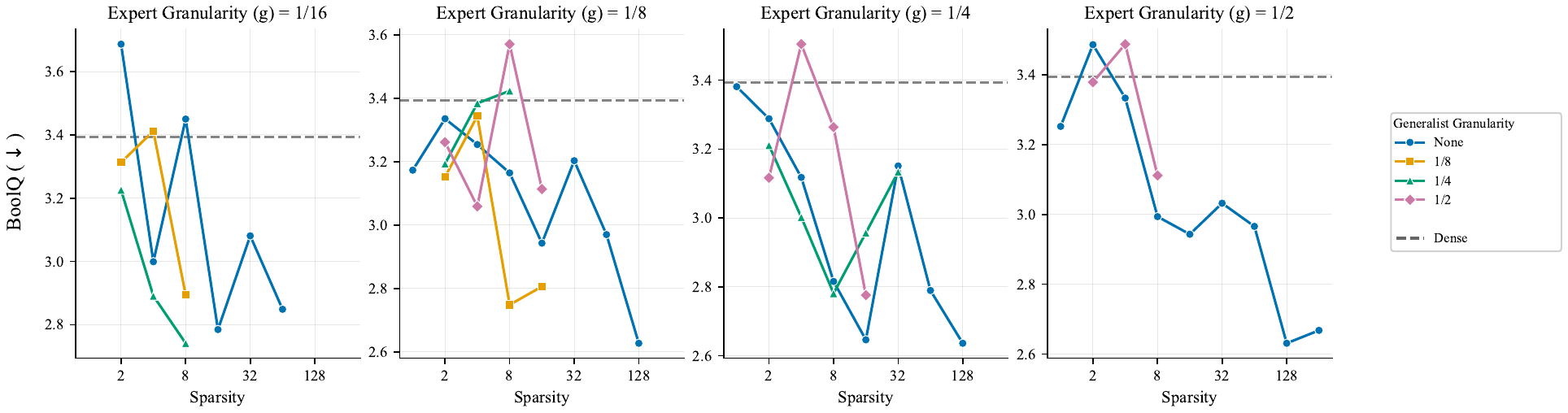}
        \caption{50M active, 50M - 930M total parameters}
    \end{subfigure}
    \par\bigskip\bigskip
    \begin{subfigure}[t]{1.0\textwidth}
        \centering
        \includegraphics[width=\linewidth]{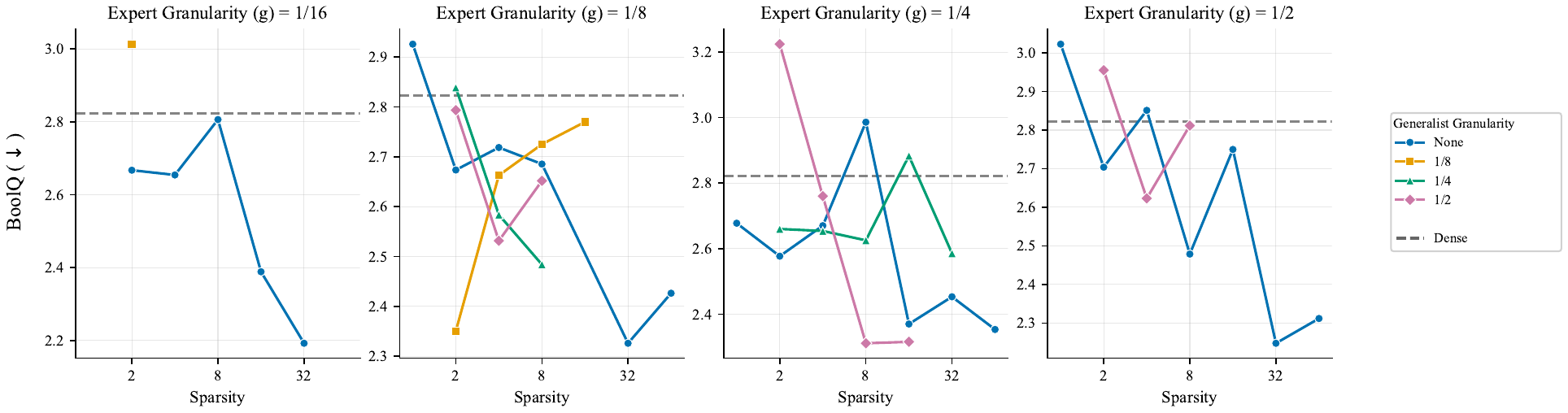}
        \caption{80M active, 80M - 765M total parameters}
    \end{subfigure}
    \par\bigskip\bigskip
    \begin{subfigure}[t]{1.0\textwidth}
        \centering
        \includegraphics[width=\linewidth]{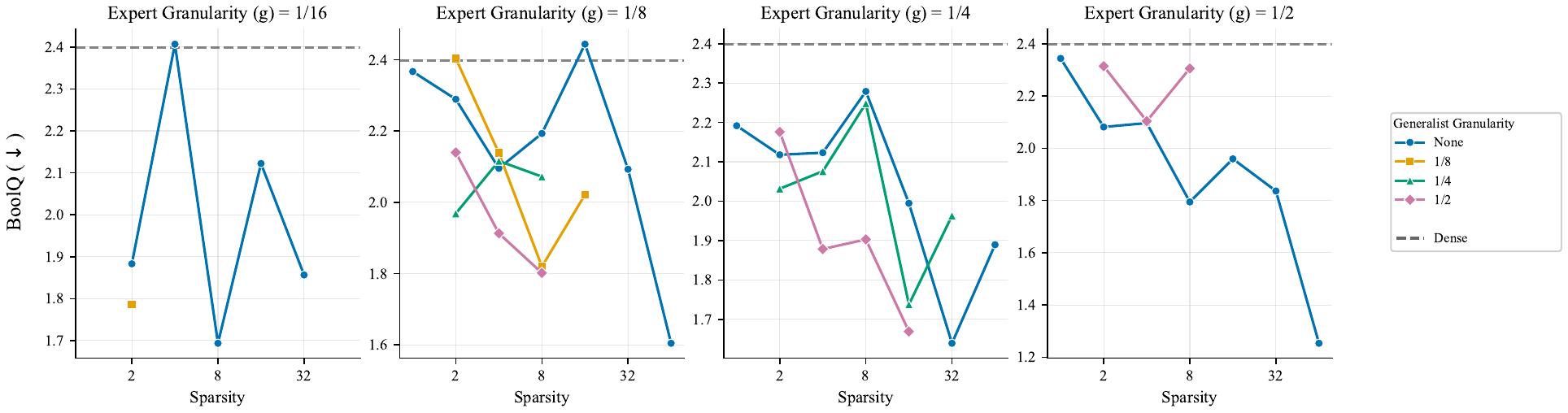}
        \caption{110M active, 110M - 1.4B total parameters}
    \end{subfigure}
    \caption{
    \textbf{The inclusion of a generalist consistently degrades performance in homogeneous MoEs (\S\ref{sec:expt_hetgen}).}
    We train MoE LMs which consist of some routed experts with granularity $g$, as well as a generalist with granularity $g_{gen}\in \{\frac{1}{2}, \frac{1}{4}, \frac{1}{8}\} $. We compare to settings with no generalist, only routed experts with granularity $g$. In all settings and configurations, the addition of any granularity generalist results in comparable or degraded performance. 
    }
    \label{fig:boolq_gen}
\end{figure*}

\begin{figure*}[ht]
    \centering
    \begin{subfigure}[t]{1.0\textwidth}
        \centering
        \includegraphics[width=\linewidth]{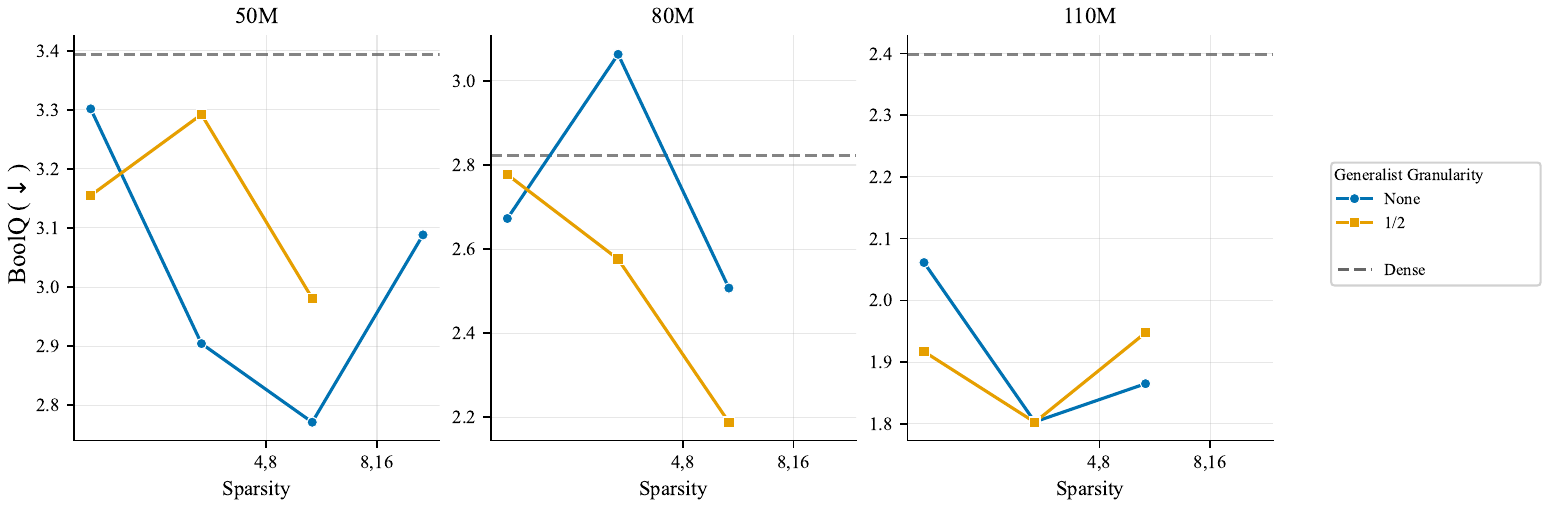}
    \end{subfigure}
    \caption{
    \textbf{The inclusion of a generalist consistently degrades performance in heterogeneous MoEs (\S\ref{sec:expt_hetgen}).}
    We train heterogeneous MoE LMs which consist of  routed experts with granularity $g_1, g_2$, as well as a generalist with granularity $g_{gen} = \frac{1}{2}$. We compare to settings with no generalist. In all settings and configurations, the addition of a generalist results in comparable or degraded performance. 
    }
    \label{fig:boolq_hetgen}
\end{figure*}

\begin{figure*}[ht]
    \centering
    \begin{subfigure}[t]{\textwidth}
        \centering
        \begin{subfigure}[t]{0.45\textwidth}
            \includegraphics[width=\linewidth]{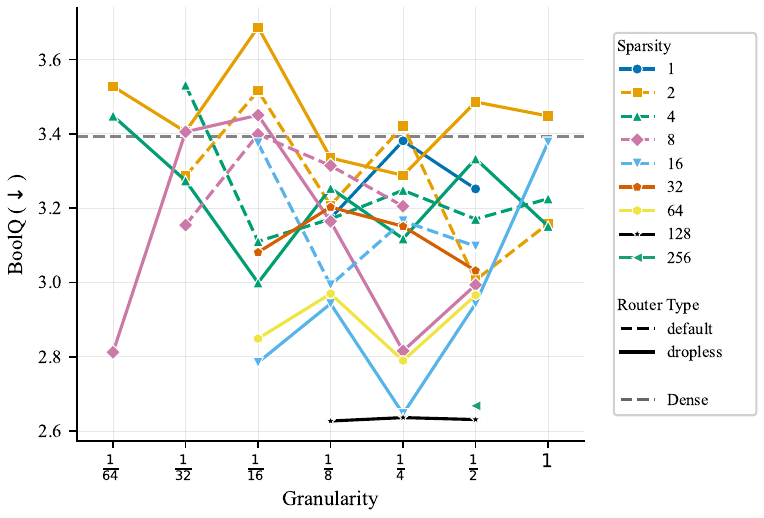}
            \caption{50M active, 50M - 930M total parameters}
        \end{subfigure}
    \hspace{1em}
        \begin{subfigure}[t]{0.45\textwidth}
            \centering
            \includegraphics[width=\linewidth]{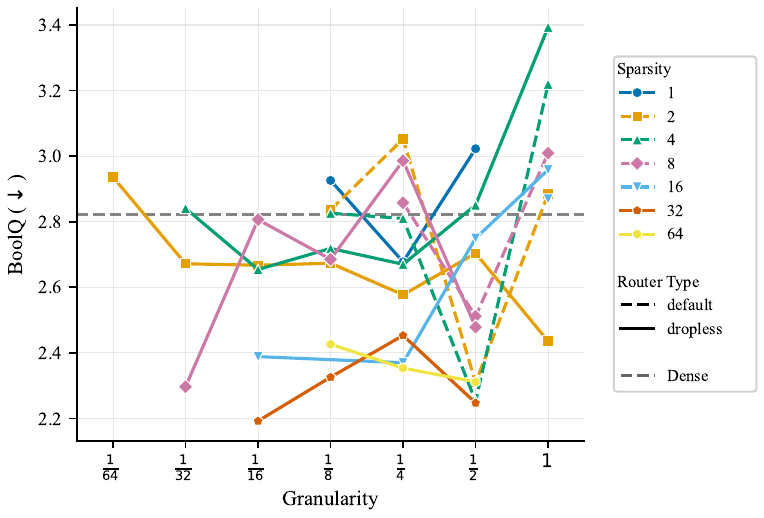}
            \caption{80M active, 80M - 765M total parameters}
        \end{subfigure}
    \end{subfigure}

    \par\bigskip\bigskip
    \begin{subfigure}[t]{0.45\textwidth}
        \centering
        \includegraphics[width=\linewidth]{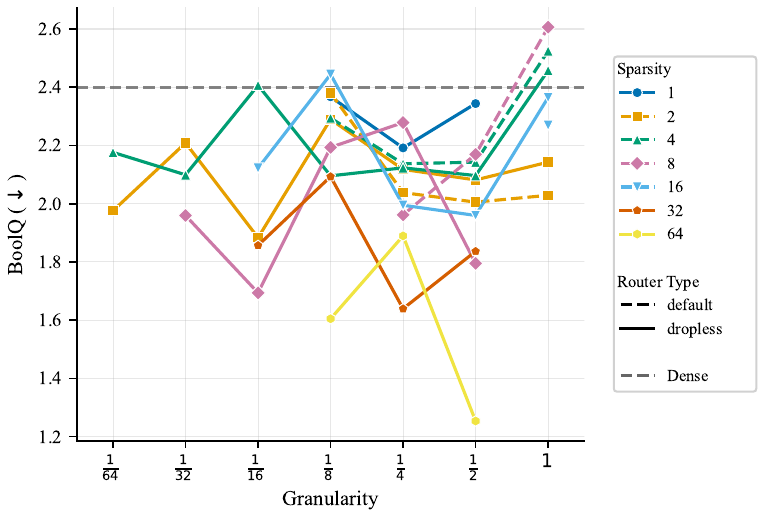}
        \caption{110M active, 110M - 1.4B total parameters}
    \end{subfigure}
    \caption{ 
    \textbf{Dropless routing outperforms default routing (\S\ref{sec:expt_router}).}
    We compare dropless routing to the default setting, which allow tokens to be dropped. Across all scales, we find that dropless routing outperforms or performs comparably to default routing. 
    }
    \label{fig:boolq_dropless}
\end{figure*}

\begin{figure*}[ht]
    \centering
    \begin{subfigure}[t]{0.45\textwidth}
        \centering
        \includegraphics[width=\linewidth]{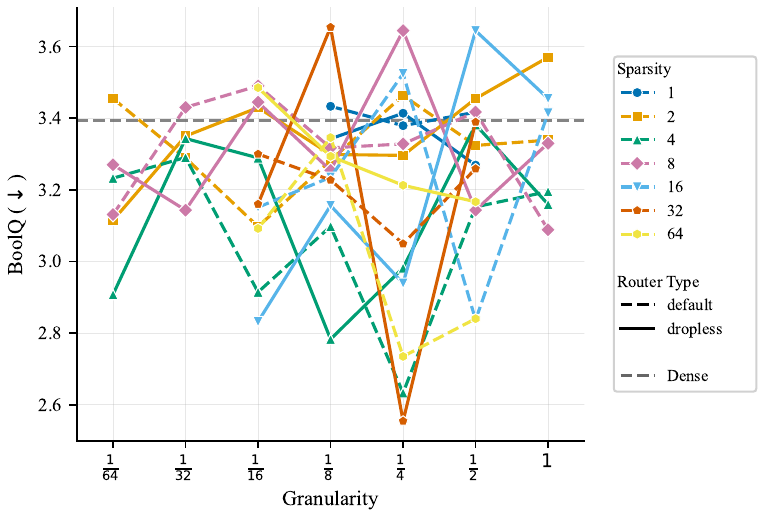}
        \caption{50M active, 50M - 930M total parameters}
    \end{subfigure}
    \hspace{1em}
    \begin{subfigure}[t]{0.45\textwidth}
        \centering
        \includegraphics[width=\linewidth]{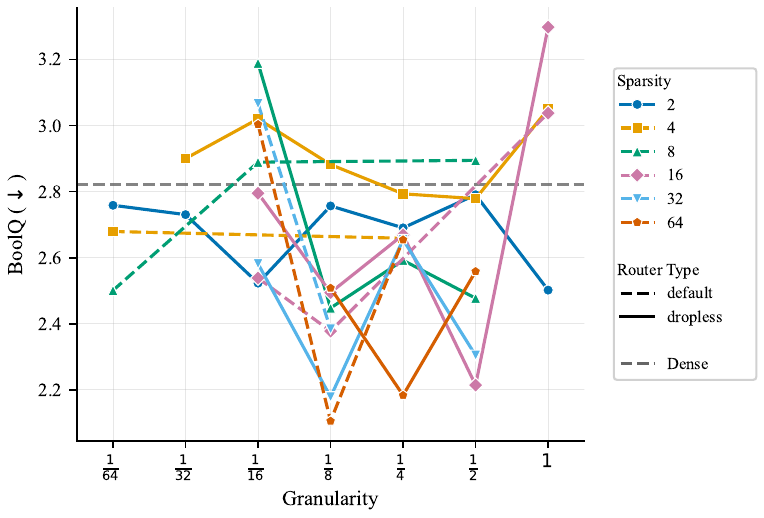}
        \caption{80M active, 80M - 765M total parameters}
    \end{subfigure}
    \caption{
    \textbf{Dropless routing, with bias $\gamma=\num{1e-3}$ (\S\ref{sec:expt_router}).} 
    As in Figure~\ref{fig:lm_avg_dropless}, we compare dropless routing to the default setting, which allow tokens to be dropped. Across all scales, we find that dropless routing outperforms or performs comparably to default routing. We see here with additional higher sparsity default routing runs that as sparsity increases, default routing performance approaches that of dropless routing.
    }
    \label{fig:boolq_dropless_with_lf}
\end{figure*}

\begin{figure*}[ht]
    \centering
    \begin{subfigure}[]{\textwidth}
        \centering
        \includegraphics[width=0.46\linewidth]{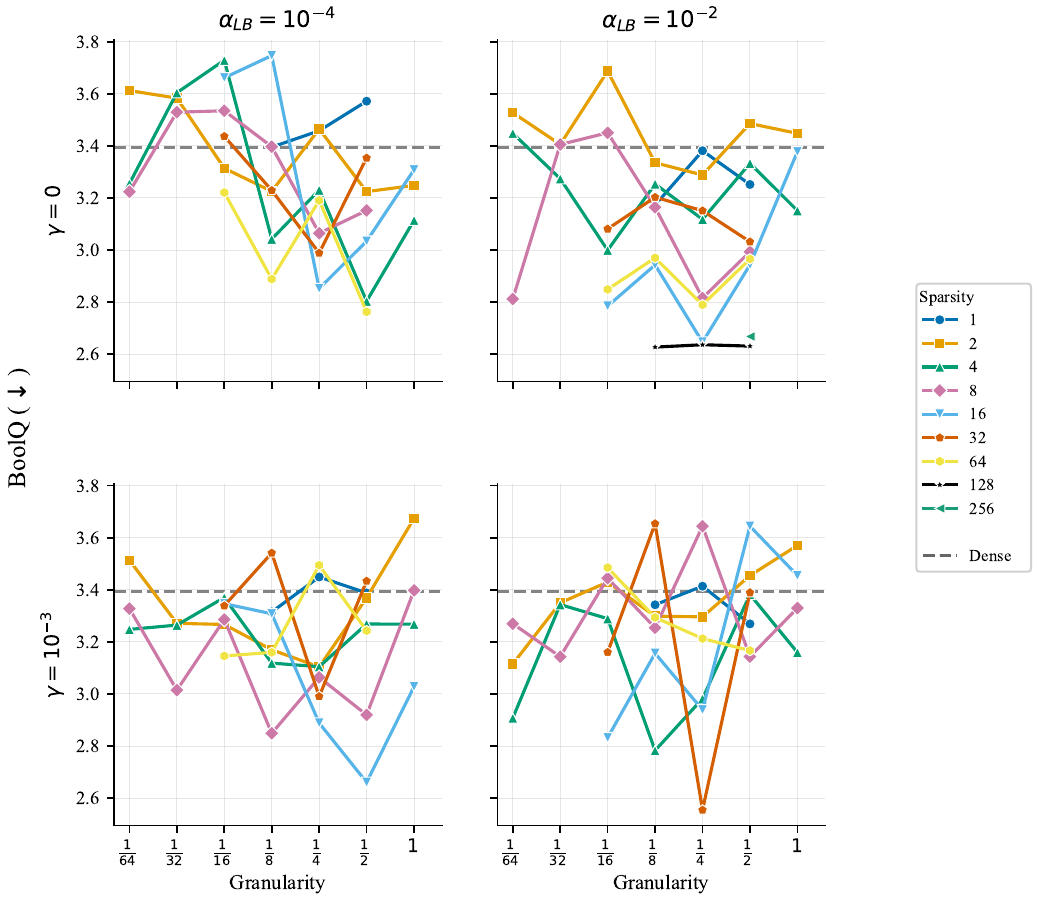}
        \hspace{1em}
        \includegraphics[width=0.46\linewidth]{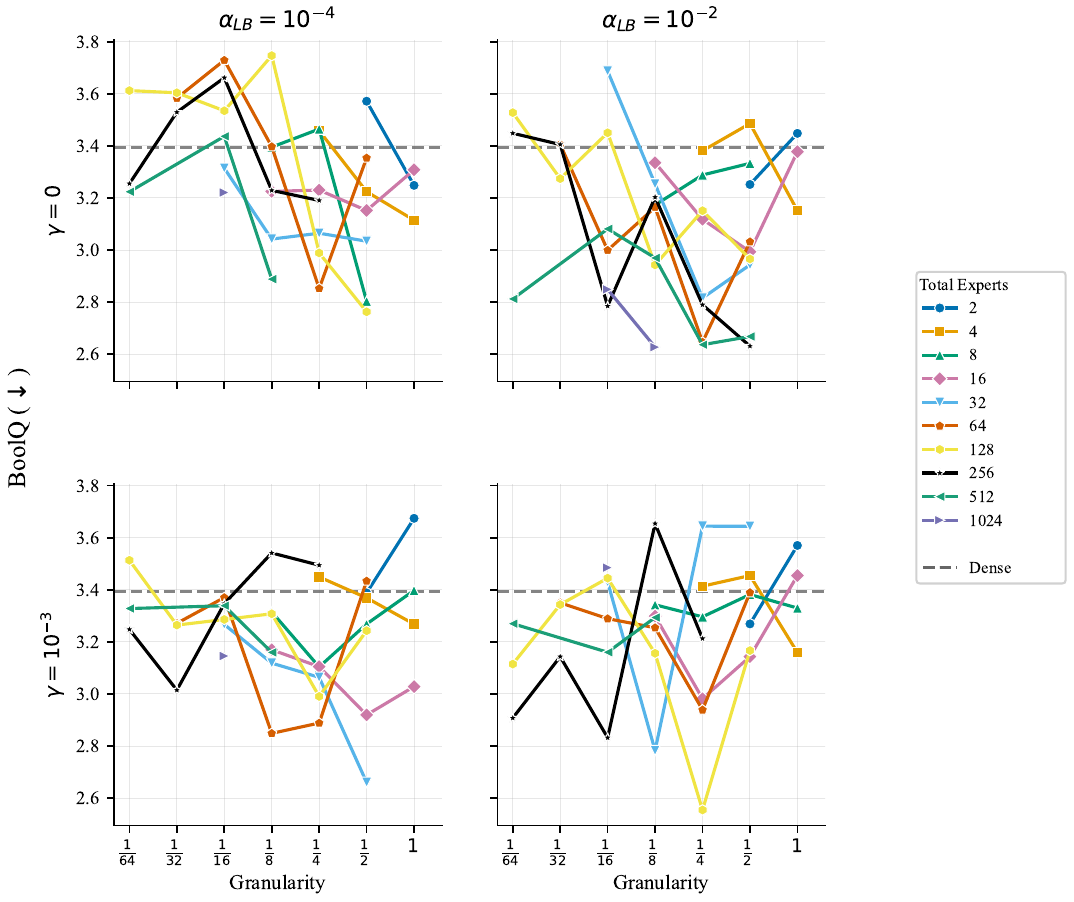}
        \caption{50M active, 50M - 930M total parameters}
    \end{subfigure}
    \par\bigskip\bigskip
    \begin{subfigure}[]{\textwidth}
        \centering
        \includegraphics[width=0.46\linewidth]{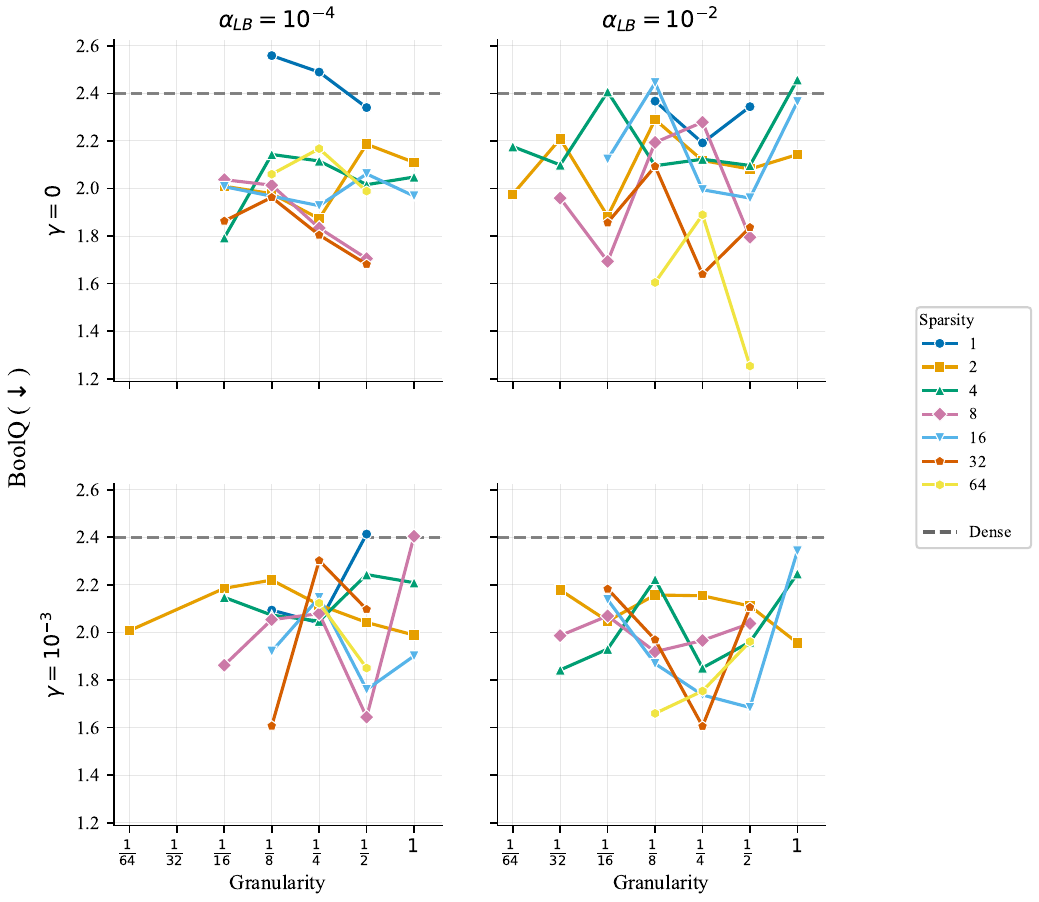}
        \hspace{1em}
        \includegraphics[width=0.46\linewidth]{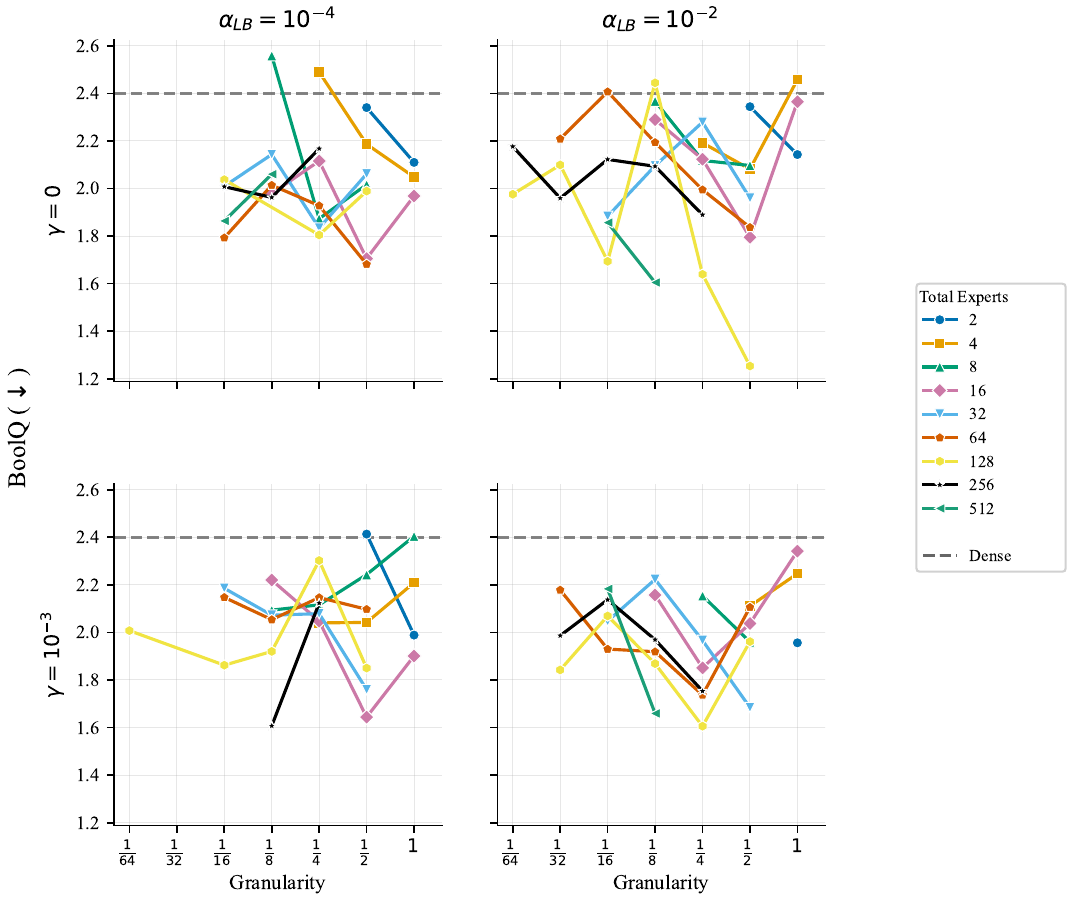}
        \caption{80M active, 80M - 765M total parameters}
    \end{subfigure}
    \par\bigskip\bigskip
    \begin{subfigure}[t]{\textwidth}
        \centering
        \includegraphics[width=0.46\linewidth]{figures/downstream/boolq/ce_loss/lb_sweep_hgn_gxs_110M.pdf}
        \hspace{1em}
        \includegraphics[width=0.46\linewidth]{figures/downstream/boolq/ce_loss/lb_sweep_hgn_gxn_110M.pdf}
        \caption{110M active, 110M - 1.4B total parameters}
    \end{subfigure}

    \end{figure*} 

\clearpage  

\begin{figure*}[ht]
    \addtocounter{figure}{-1}
    \centering
    \begin{subfigure}[t]{\textwidth}
        \centering
        \includegraphics[width=0.46\linewidth]{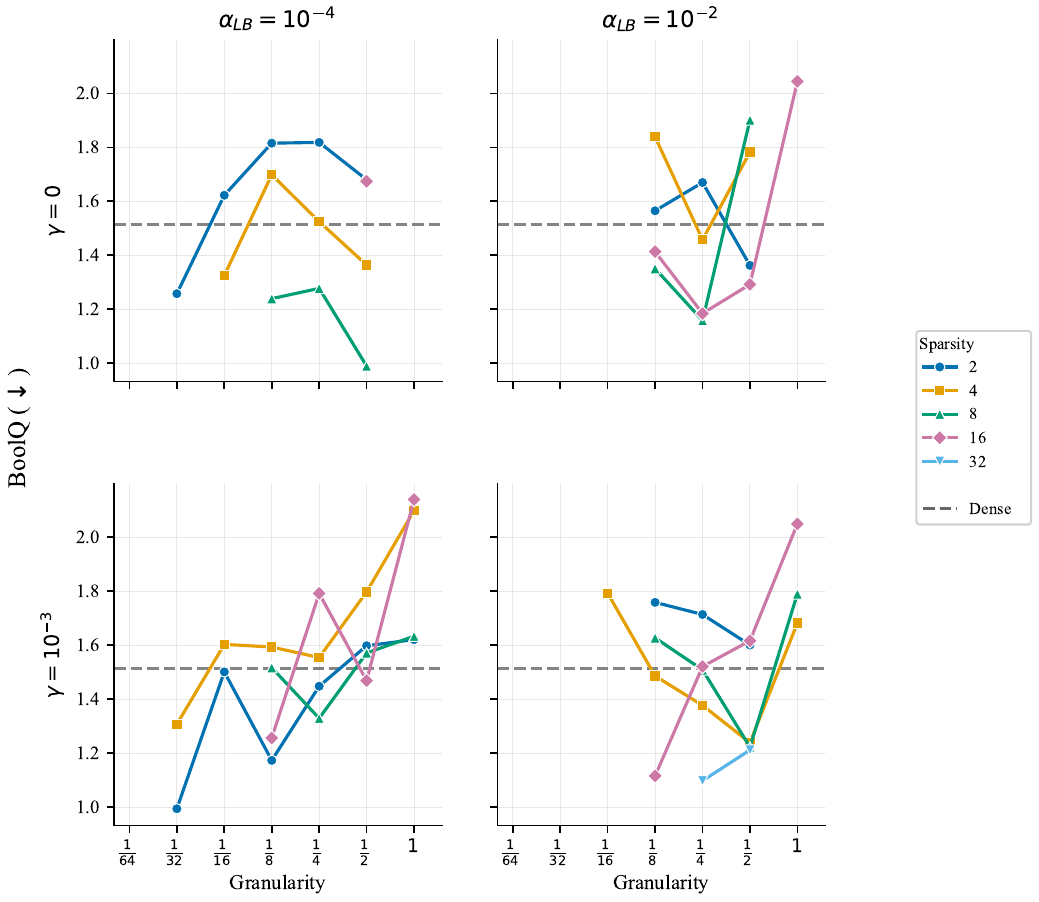}
        \hspace{1em}
        \includegraphics[width=0.46\linewidth]{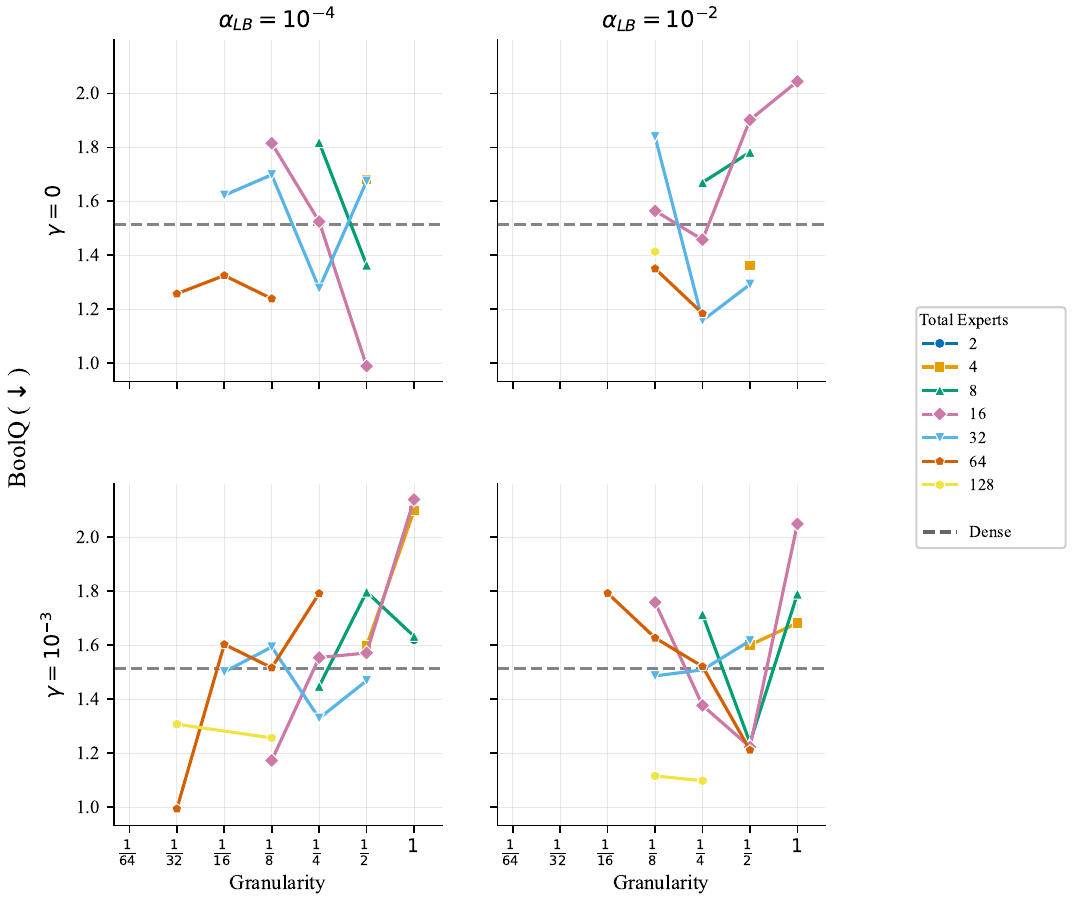}
        \caption{200M active, 200M - 3.3B total parameters}
    \end{subfigure}
    \par\bigskip\bigskip
    \begin{subfigure}[t]{\textwidth}
        \centering
        \includegraphics[width=0.3\linewidth]{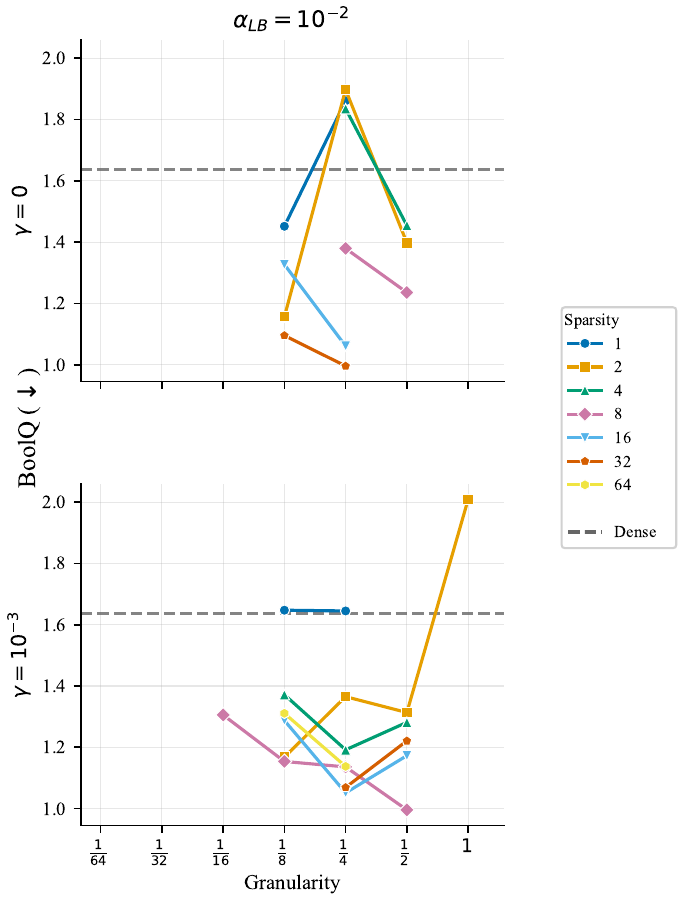}
        \hspace{1em}
        \includegraphics[width=0.3\linewidth]{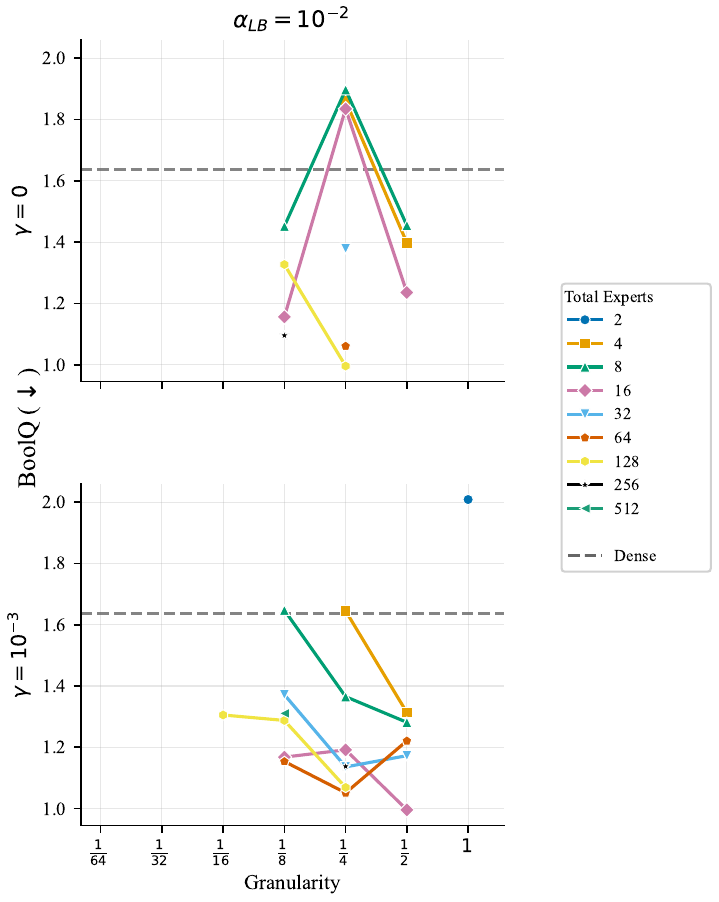}
        \caption{300M active, 300M - 6.6B total parameters}
    \end{subfigure}

    \caption{
    \textbf{Load balancing mechanisms must be tuned correctly (\S\ref{sec:expt_router}).}
    We consider load balancing loss weight $\alpha_{LB} \in \{\num{1e-2}, \num{1e-4}\}$ and loss-free load balancing with bias $\gamma\in\{0, \num{1e-3}\}$ ($\gamma=0$ indicates no loss-free mechanism). Results show that poorly chosen hyperparameters, such as high bias $\gamma = 1e-3$ with total experts $n\geq 512$, may impair performance. However, all settings other than $(\alpha_{LB}=\num{1e-2}, \gamma=\num{1e-3})$ perform comparably for $n \leq 512$, suggesting that a wide range of load balancing settings achieve near-optimal performance. 
    }
    \label{fig:boolq_lb}
\end{figure*}

%% file: fig_tex/downstream/hellaswag.tex
\begin{figure*}[!ht]
    \centering
        \begin{subfigure}[t]{\textwidth}
        \begin{subfigure}[t]{0.33\textwidth}
            \centering
            \caption*{\scriptsize Fixed total experts (n)}
            \includegraphics[width=\linewidth]{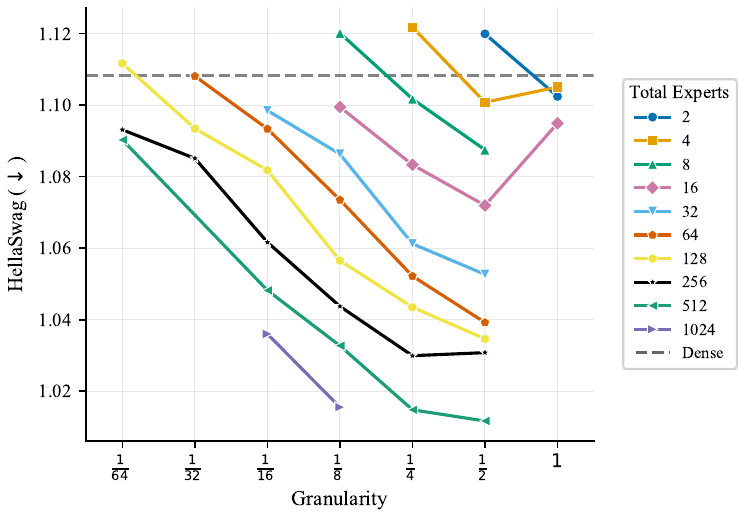}
        \end{subfigure}
        \begin{subfigure}[t]{0.33\textwidth}
            \centering
            \caption*{\scriptsize Fixed granularity (g)}
            \includegraphics[width=\linewidth]{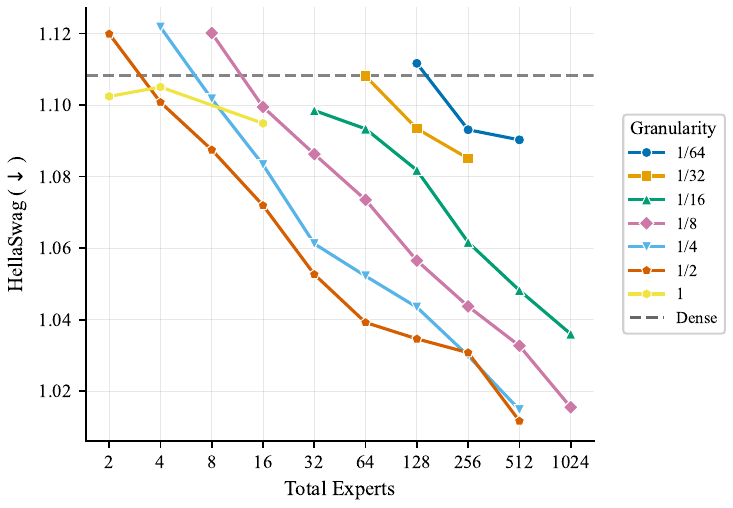}
        \end{subfigure}
        \begin{subfigure}[t]{0.33\textwidth}
            \centering
            \caption*{\scriptsize Fixed activation sparsity (s)}
            \includegraphics[width=\linewidth]{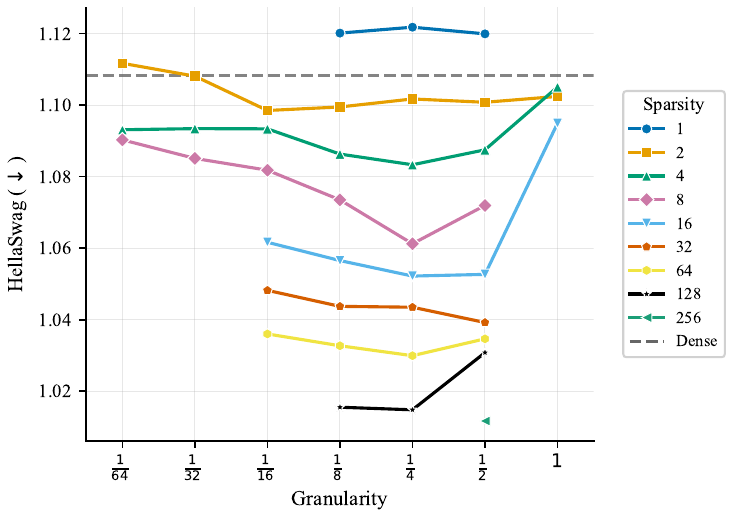}
        \end{subfigure}
        \caption{50M active, 50M - 930M total parameters}
    \end{subfigure}
\par\bigskip\bigskip
    \begin{subfigure}[t]{\textwidth}
        \begin{subfigure}[t]{0.33\textwidth}
            \centering
            \includegraphics[width=\linewidth]{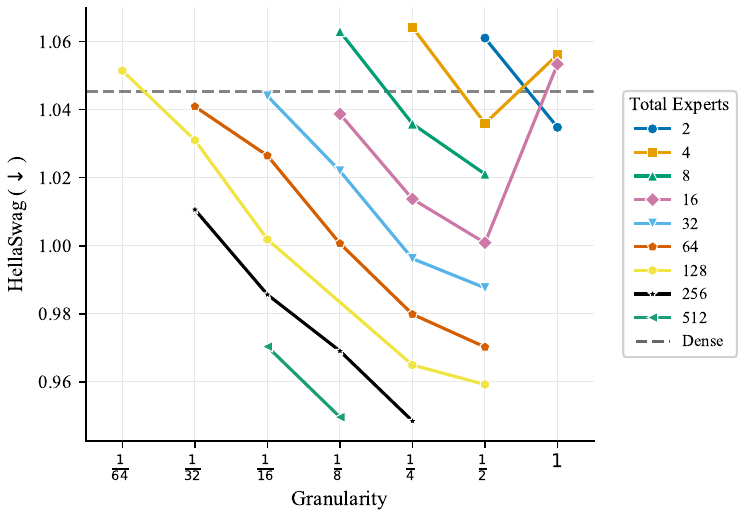}
        \end{subfigure}
        \begin{subfigure}[t]{0.33\textwidth}
            \centering
            \includegraphics[width=\linewidth]{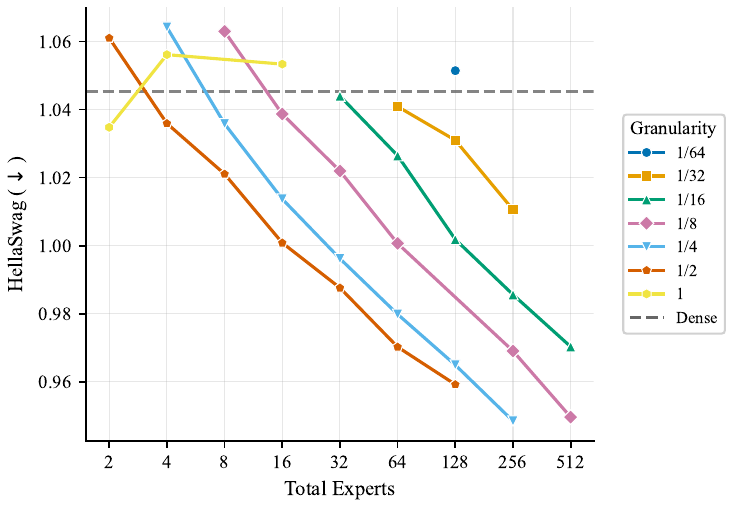}
        \end{subfigure}
        \begin{subfigure}[t]{0.33\textwidth}
            \centering
            \includegraphics[width=\linewidth]{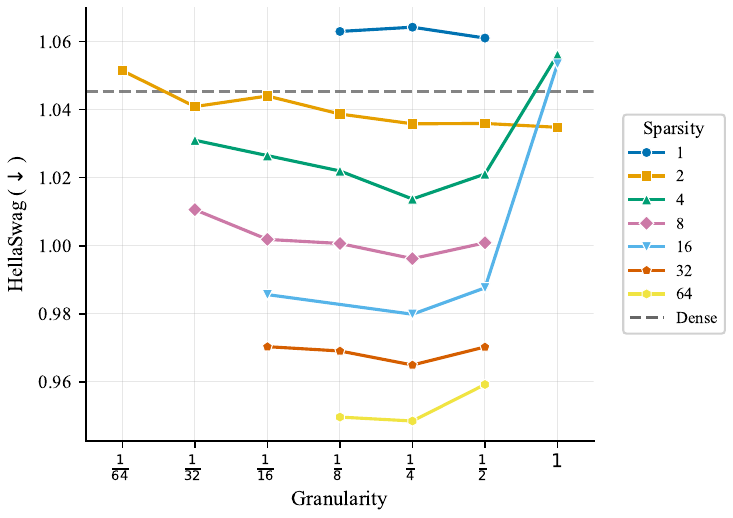}
        \end{subfigure}
        \caption{80M active, 80M - 765M total parameters}
    \end{subfigure}
    \par\bigskip\bigskip
        \begin{subfigure}[t]{\textwidth}
        \begin{subfigure}[t]{0.33\textwidth}
            \centering
            \includegraphics[width=\linewidth]{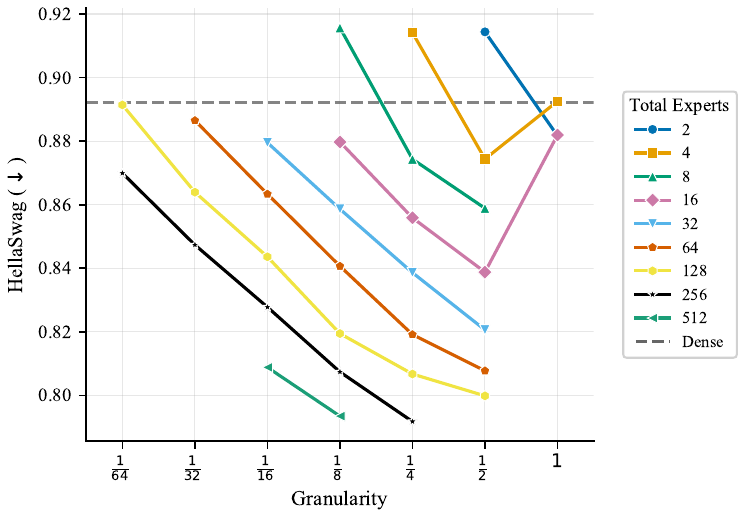}
        \end{subfigure}
        \begin{subfigure}[t]{0.33\textwidth}
            \centering
            \includegraphics[width=\linewidth]{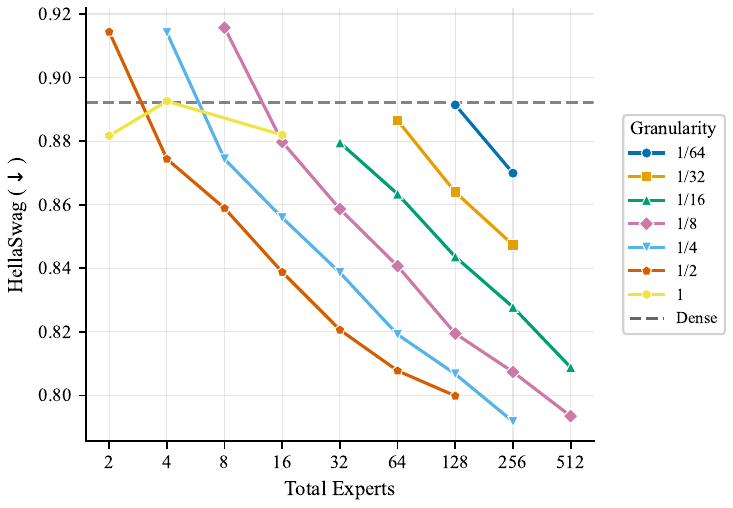}
        \end{subfigure}
        \begin{subfigure}[t]{0.33\textwidth}
            \centering
            \includegraphics[width=\linewidth]{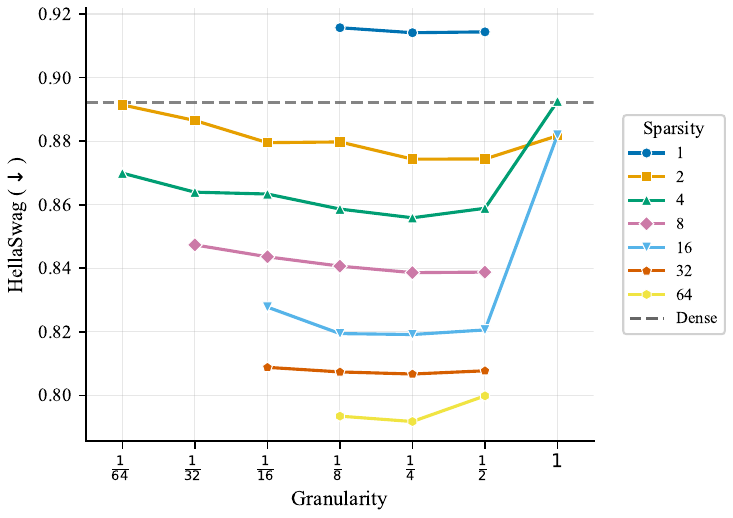}
        \end{subfigure}
        \caption{110M active, 110M - 1.4B total parameters}
    \end{subfigure}
    \end{figure*}

\clearpage  

\begin{figure*}[!ht]
        \addtocounter{figure}{-1}
    \begin{subfigure}[t]{\textwidth}
        \addtocounter{subfigure}{3}
        \begin{subfigure}[t]{0.33\textwidth}
            \centering
            \caption*{\scriptsize Fixed total experts (n)}
            \includegraphics[width=\linewidth]{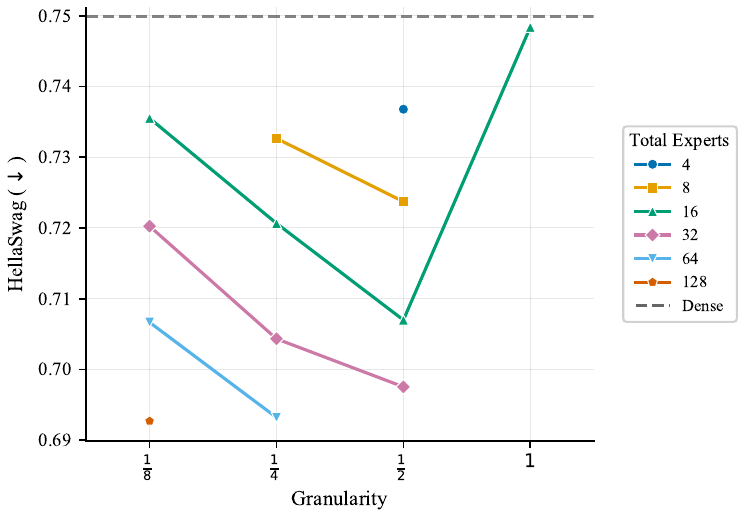}
        \end{subfigure}
        \begin{subfigure}[t]{0.33\textwidth}
            \centering
            \caption*{\scriptsize Fixed granularity (g)}
            \includegraphics[width=\linewidth]{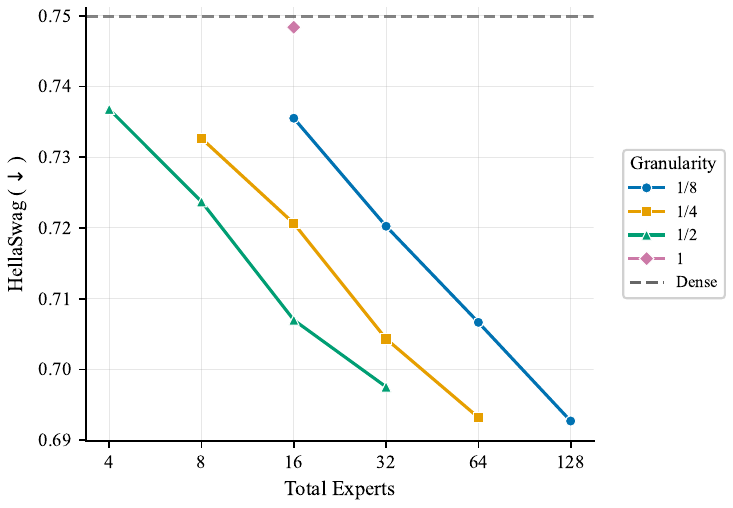}
        \end{subfigure}
        \begin{subfigure}[t]{0.33\textwidth}
            \centering
            \caption*{\scriptsize Fixed activation sparsity (s)}
            \includegraphics[width=\linewidth]{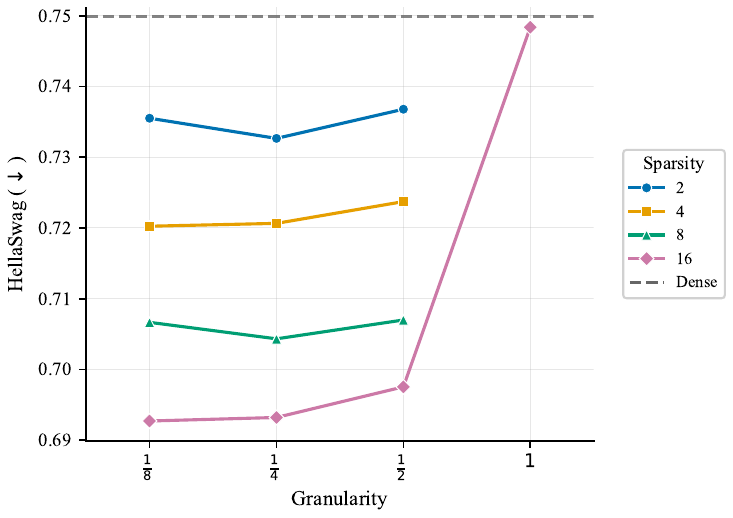}
        \end{subfigure}
        \caption{200M active, 200M - 3.3B total parameters}
    \end{subfigure}
    \par\bigskip\bigskip
        \begin{subfigure}[t]{\textwidth}
        \begin{subfigure}[t]{0.33\textwidth}
            \centering
            \includegraphics[width=\linewidth]{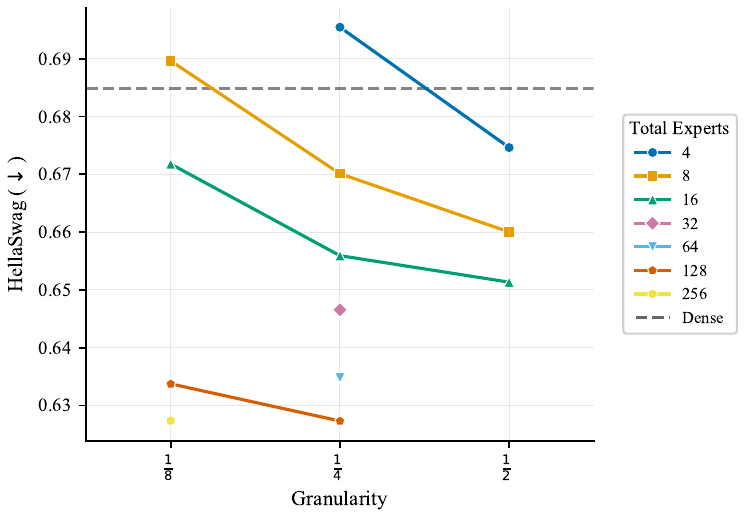}
        \end{subfigure}
        \begin{subfigure}[t]{0.33\textwidth}
            \centering
            \includegraphics[width=\linewidth]{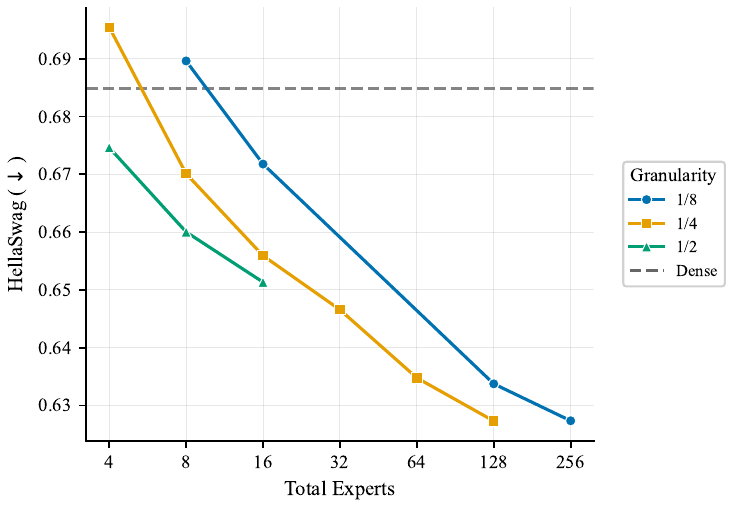}
        \end{subfigure}
        \begin{subfigure}[t]{0.33\textwidth}
            \centering
            \includegraphics[width=\linewidth]{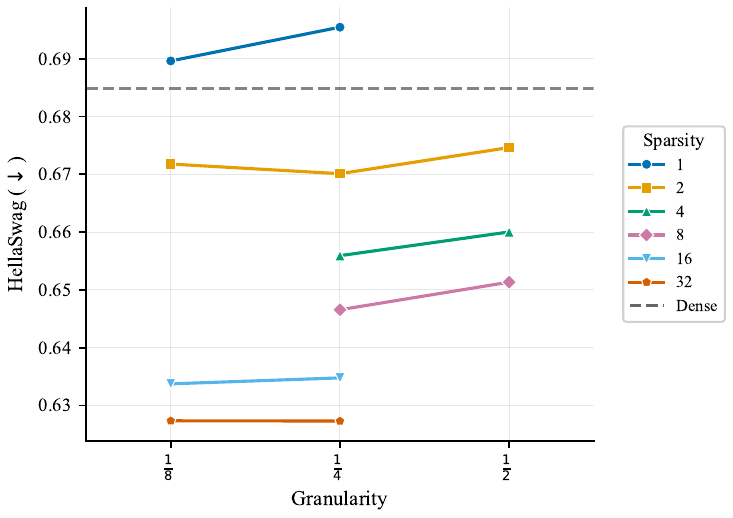}
        \end{subfigure}
        \caption{300M active, 300M - 6.6B total parameters}
    \end{subfigure}

    \caption{
    \textbf{Increasing inactive expert parameters via expert size (left) or total count (center) improves performance in MoEs (\S\ref{sec:expt_main}).} This effect is seen both when holding total number of experts fixed (left) and when holding expert granularity fixed (center). In general, increasing total parameters results in improved performance.  \textbf{Optimal tradeoff between expert count and granularity varies in MoEs (right). (\S\ref{sec:expt_main})}
    At each activation sparsity $s$ (equivalently, at each total parameter count), the optimal (total expert count, expert granularity) configuration varies. As $s$ increases, optimal expert granularity remains nearly fixed, suggesting that sparsity should be scaled up primarily by increasing total expert count $n$, while maintaining a near constant, slowly increasing expert granularity $g$. 
    }
    \label{fig:hellaswag_experts}
\end{figure*}

\begin{figure*}[!ht]
    \centering
    
    \begin{subfigure}[t]{0.46\textwidth}
        \centering
        \includegraphics[width=\linewidth]{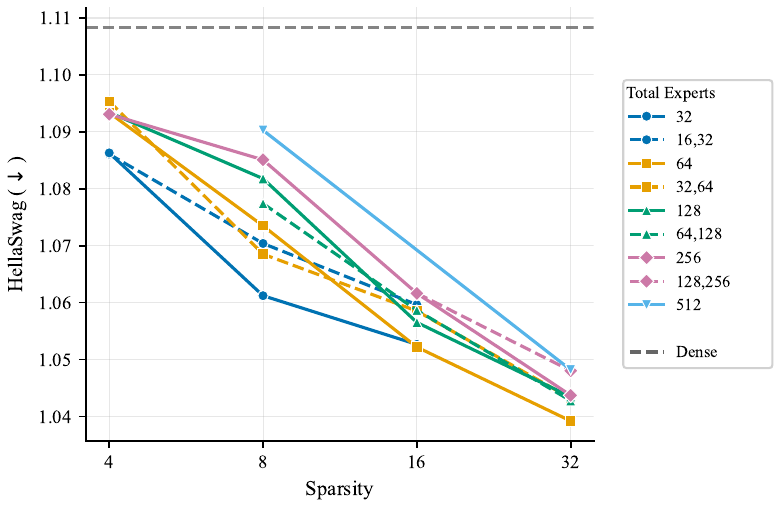}
        \caption{50M active, 50M - 930M total parameters}
    \end{subfigure}
    \vspace{1em}
    \begin{subfigure}[t]{0.46\textwidth}
        \centering
        \includegraphics[width=\linewidth]{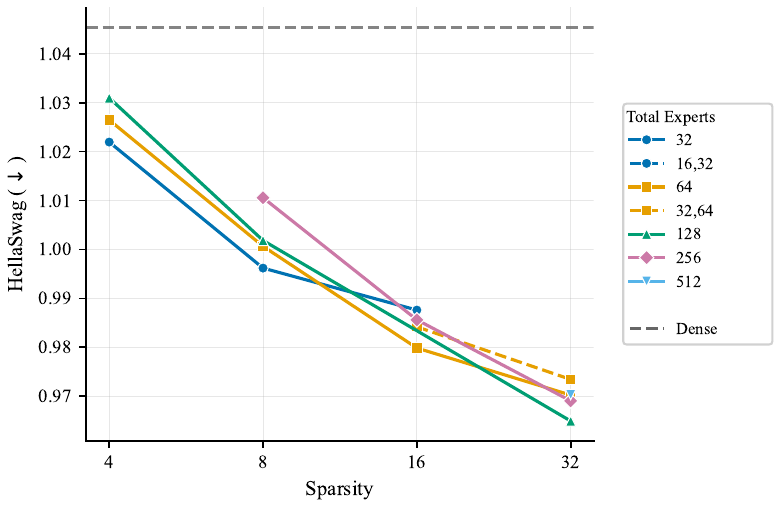}
        \caption{80M active, 80M - 765M total parameters}
    \end{subfigure}
    \caption{
    \textbf{Heterogeneity of expert size alone does not improve MoE performance (\S\ref{sec:expt_hetgen}).} To explore the potential benefits of their architectural flexibility, we compare heterogeneous MoEs (indicated by dotted lines) to active- and total-parameter-matched homogeneous MoEs. Heterogeneity alone does not result in performance gains, as, at each activation sparsity $s$, heterogeneous MoEs with $n_1, n_2 = a, b$ lie between or near the 2 closest homogeneous MoEs, with $n=a$ and with $n=b$.
    }
    \label{fig:hellaswag_het}
\end{figure*}

\begin{figure*}[!ht]
    \centering
    
    \begin{subfigure}[t]{1.0\textwidth}
        \centering
        \includegraphics[width=\linewidth]{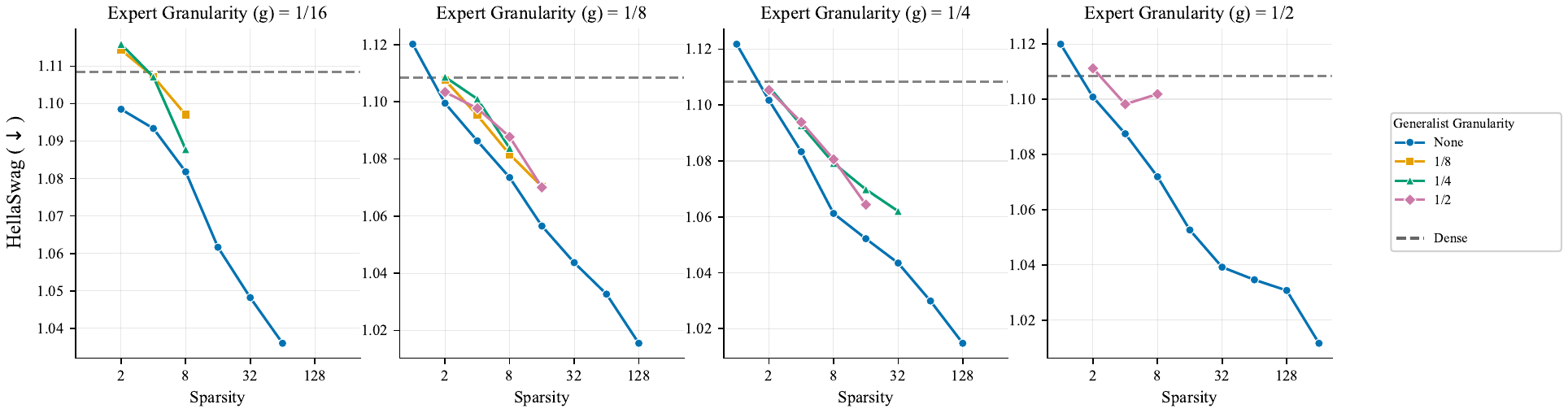}
        \caption{50M active, 50M - 930M total parameters}
    \end{subfigure}
    \par\bigskip\bigskip
    \begin{subfigure}[t]{1.0\textwidth}
        \centering
        \includegraphics[width=\linewidth]{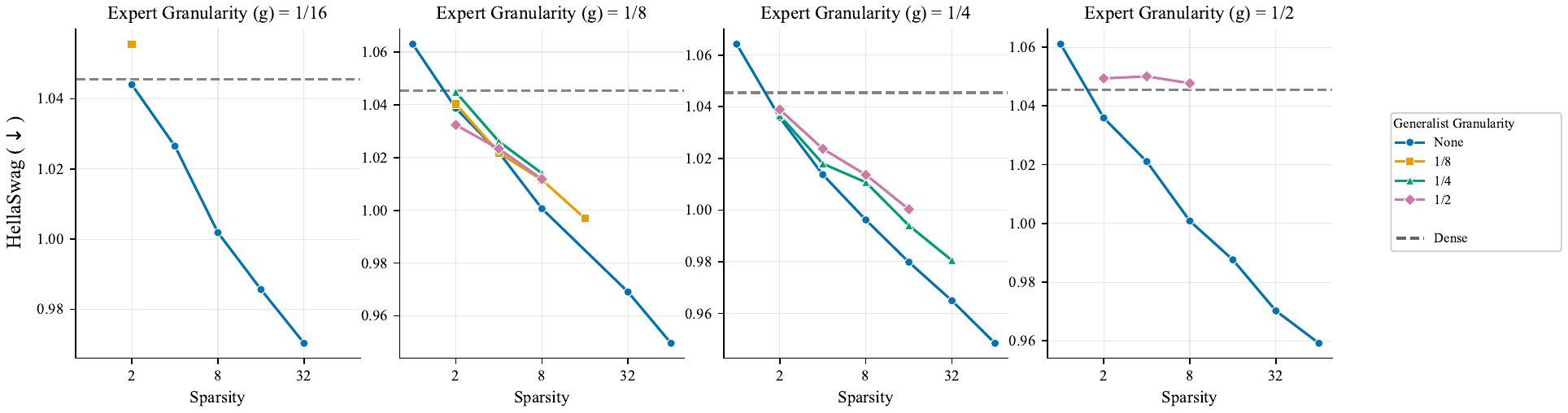}
        \caption{80M active, 80M - 765M total parameters}
    \end{subfigure}
    \par\bigskip\bigskip
    \begin{subfigure}[t]{1.0\textwidth}
        \centering
        \includegraphics[width=\linewidth]{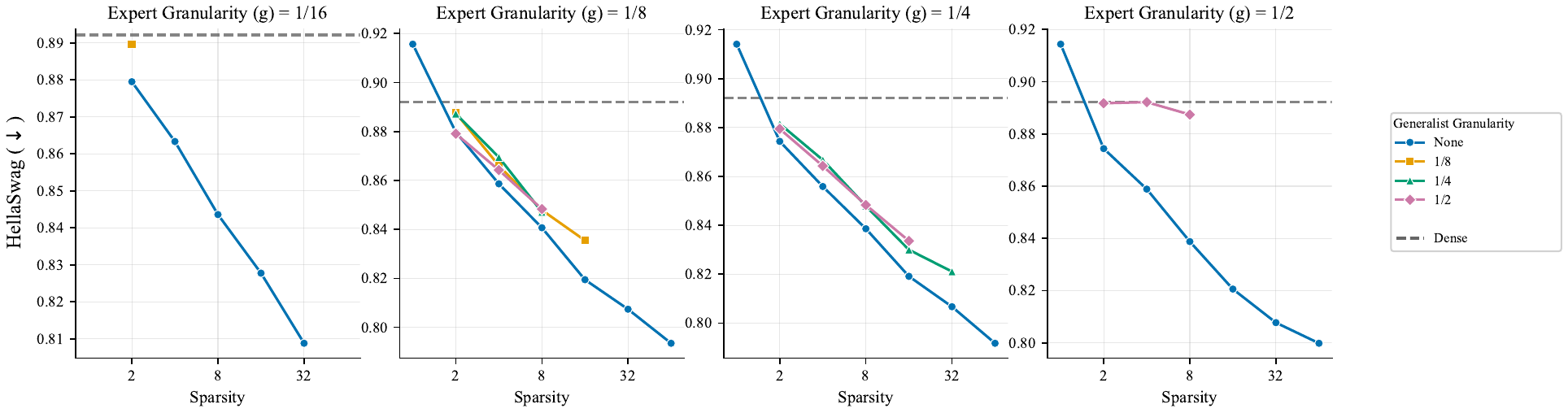}
        \caption{110M active, 110M - 1.4B total parameters}
    \end{subfigure}
    \caption{
    \textbf{The inclusion of a generalist consistently degrades performance in homogeneous MoEs (\S\ref{sec:expt_hetgen}).}
    We train MoE LMs which consist of some routed experts with granularity $g$, as well as a generalist with granularity $g_{gen}\in \{\frac{1}{2}, \frac{1}{4}, \frac{1}{8}\} $. We compare to settings with no generalist, only routed experts with granularity $g$. In all settings and configurations, the addition of any granularity generalist results in comparable or degraded performance. 
    }
    \label{fig:hellaswag_gen}
\end{figure*}

\begin{figure*}[ht]
    \centering
    \begin{subfigure}[t]{1.0\textwidth}
        \centering
        \includegraphics[width=\linewidth]{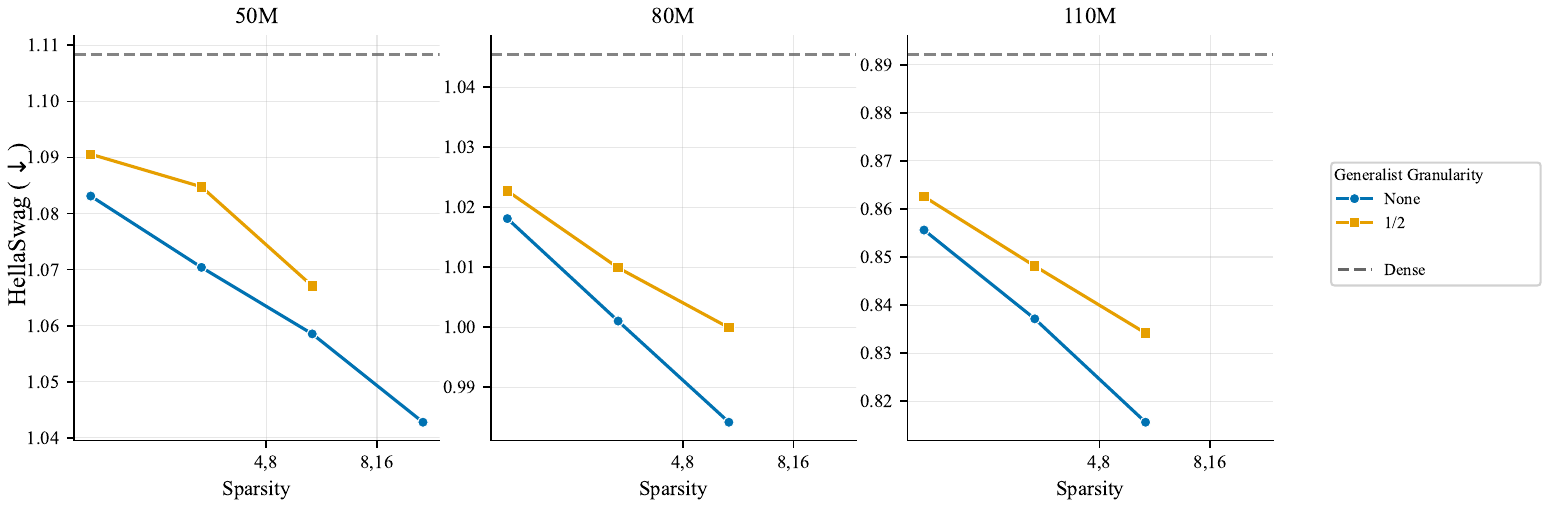}
    \end{subfigure}
    \caption{
    \textbf{The inclusion of a generalist consistently degrades performance in heterogeneous MoEs (\S\ref{sec:expt_hetgen}).}
    We train heterogeneous MoE LMs which consist of  routed experts with granularity $g_1, g_2$, as well as a generalist with granularity $g_{gen} = \frac{1}{2}$. We compare to settings with no generalist. In all settings and configurations, the addition of a generalist results in comparable or degraded performance. 
    }
    \label{fig:hellaswag_hetgen}
\end{figure*}

\begin{figure*}[ht]
    \centering
    \begin{subfigure}[t]{\textwidth}
        \centering
        \begin{subfigure}[t]{0.45\textwidth}
            \includegraphics[width=\linewidth]{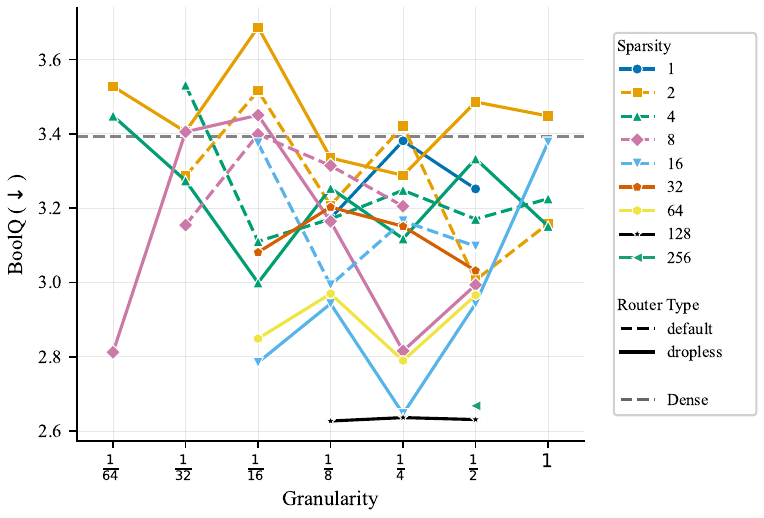}
            \caption{50M active, 50M - 930M total parameters}
        \end{subfigure}
    \hspace{1em}
        \begin{subfigure}[t]{0.45\textwidth}
            \centering
            \includegraphics[width=\linewidth]{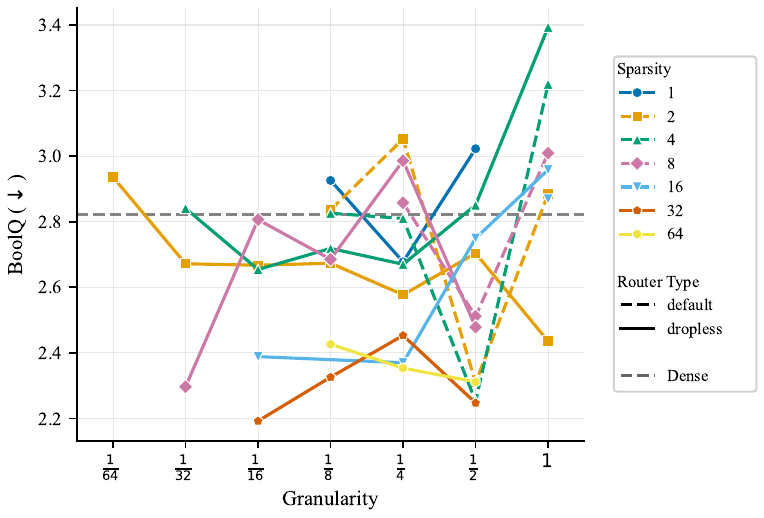}
            \caption{80M active, 80M - 765M total parameters}
        \end{subfigure}
    \end{subfigure}

    \par\bigskip\bigskip
    \begin{subfigure}[t]{0.45\textwidth}
        \centering
        \includegraphics[width=\linewidth]{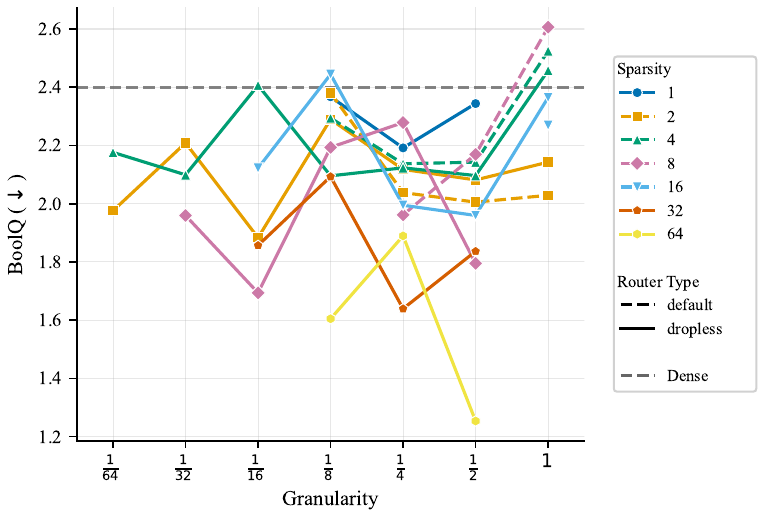}
        \caption{110M active, 110M - 1.4B total parameters}
    \end{subfigure}
    \caption{ 
    \textbf{Dropless routing outperforms default routing (\S\ref{sec:expt_router}).}
    We compare dropless routing to the default setting, which allow tokens to be dropped. Across all scales, we find that dropless routing outperforms or performs comparably to default routing. 
    }
    \label{fig:hellaswag_dropless}
\end{figure*}

\begin{figure*}[ht]
    \centering
    \begin{subfigure}[t]{0.45\textwidth}
        \centering
        \includegraphics[width=\linewidth]{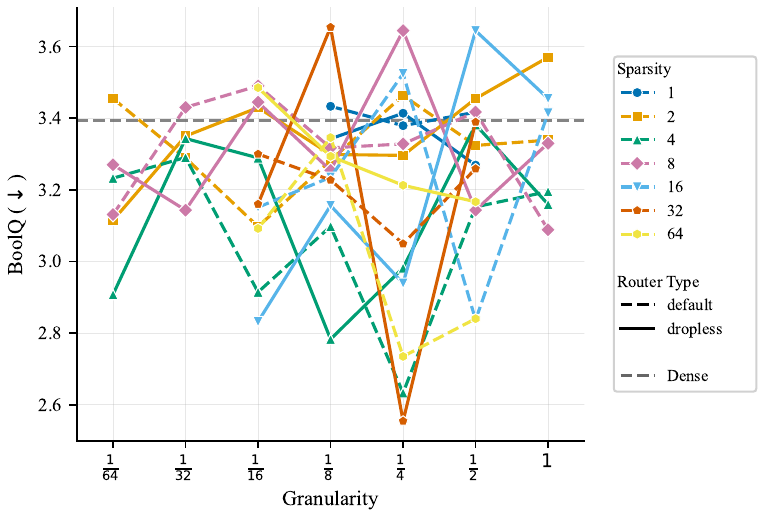}
        \caption{50M active, 50M - 930M total parameters}
    \end{subfigure}
    \hspace{1em}
    \begin{subfigure}[t]{0.45\textwidth}
        \centering
        \includegraphics[width=\linewidth]{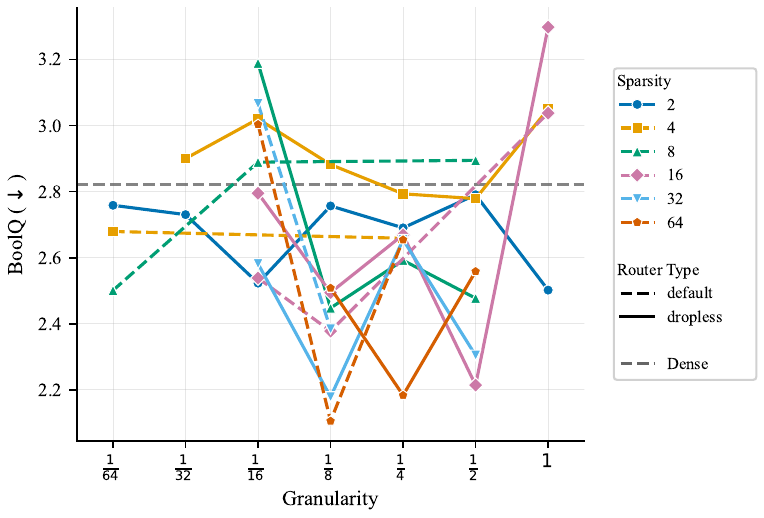}
        \caption{80M active, 80M - 765M total parameters}
    \end{subfigure}
    \caption{
    \textbf{Dropless routing, with bias $\gamma=\num{1e-3}$ (\S\ref{sec:expt_router}).} 
    As in Figure~\ref{fig:lm_avg_dropless}, we compare dropless routing to the default setting, which allow tokens to be dropped. Across all scales, we find that dropless routing outperforms or performs comparably to default routing. We see here with additional higher sparsity default routing runs that as sparsity increases, default routing performance approaches that of dropless routing.
    }
    \label{fig:hellaswag_dropless_with_lf}
\end{figure*}

\begin{figure*}[ht]
    \centering
    \begin{subfigure}[]{\textwidth}
        \centering
        \includegraphics[width=0.46\linewidth]{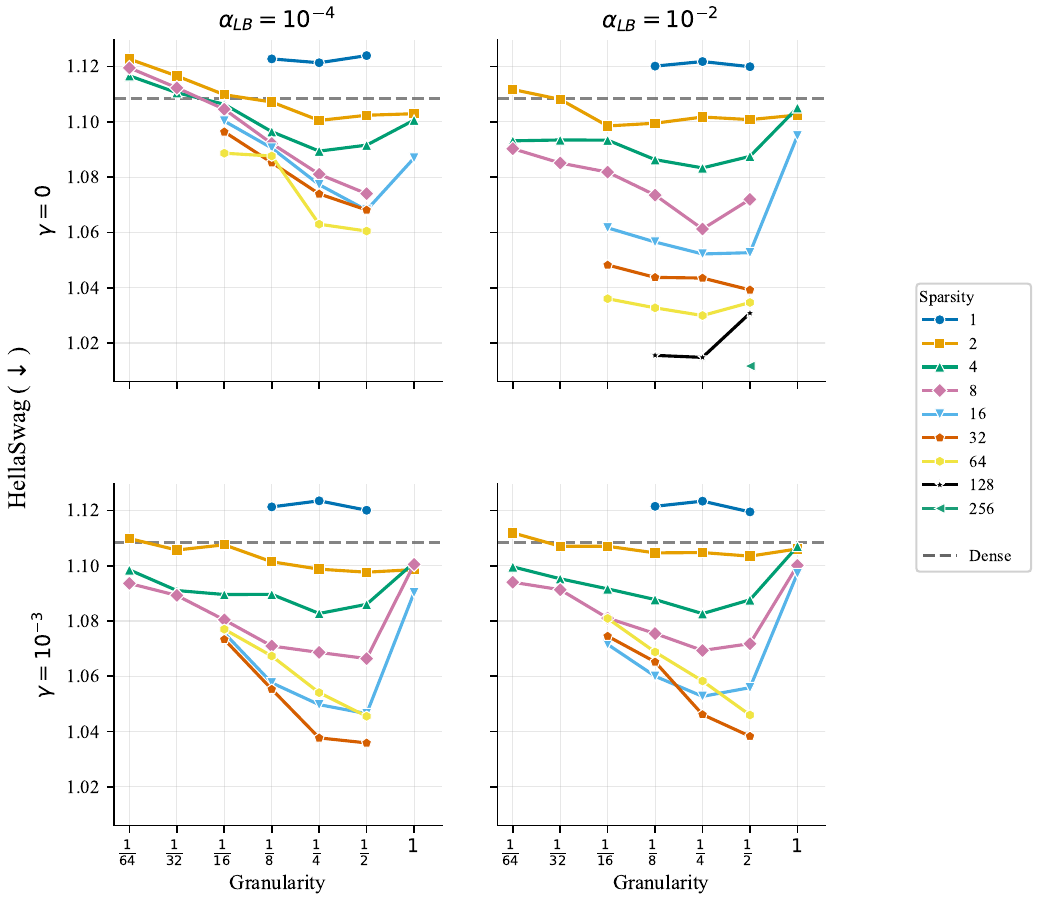}
        \hspace{1em}
        \includegraphics[width=0.46\linewidth]{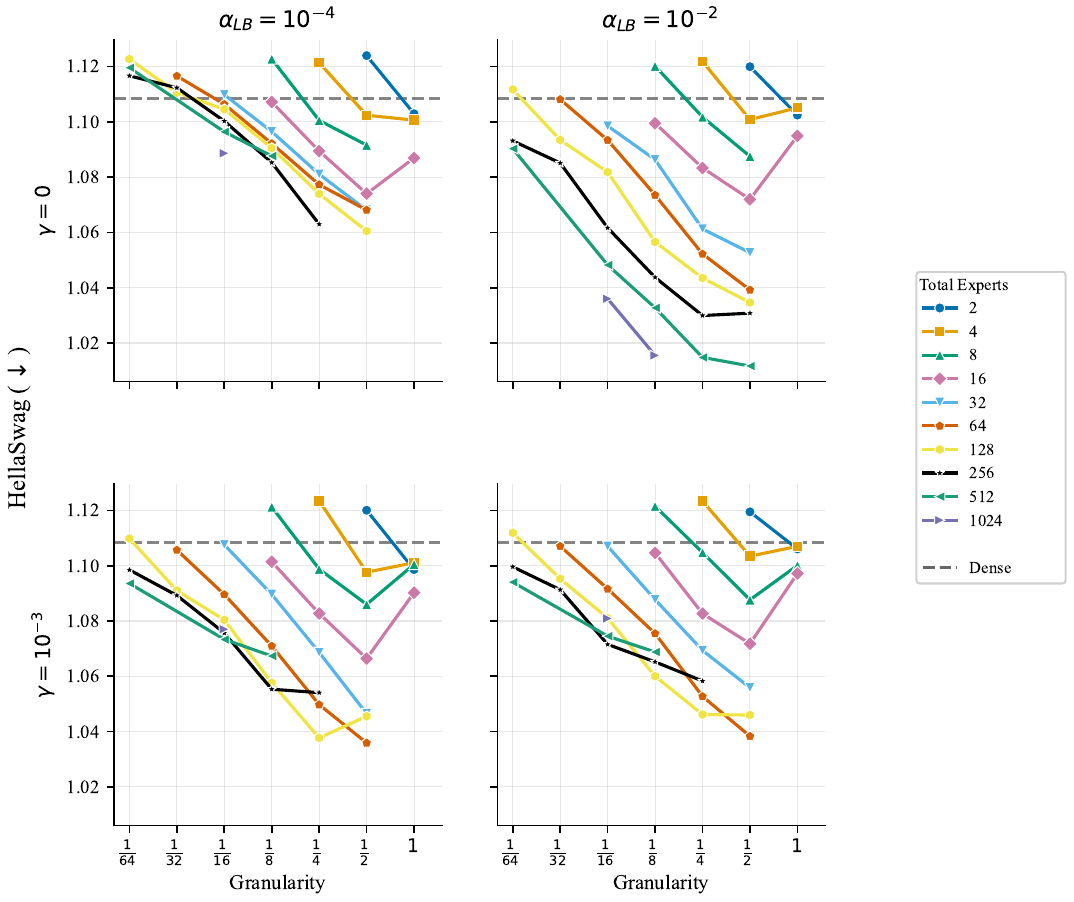}
        \caption{50M active, 50M - 930M total parameters}
    \end{subfigure}
    \par\bigskip\bigskip
    \begin{subfigure}[]{\textwidth}
        \centering
        \includegraphics[width=0.46\linewidth]{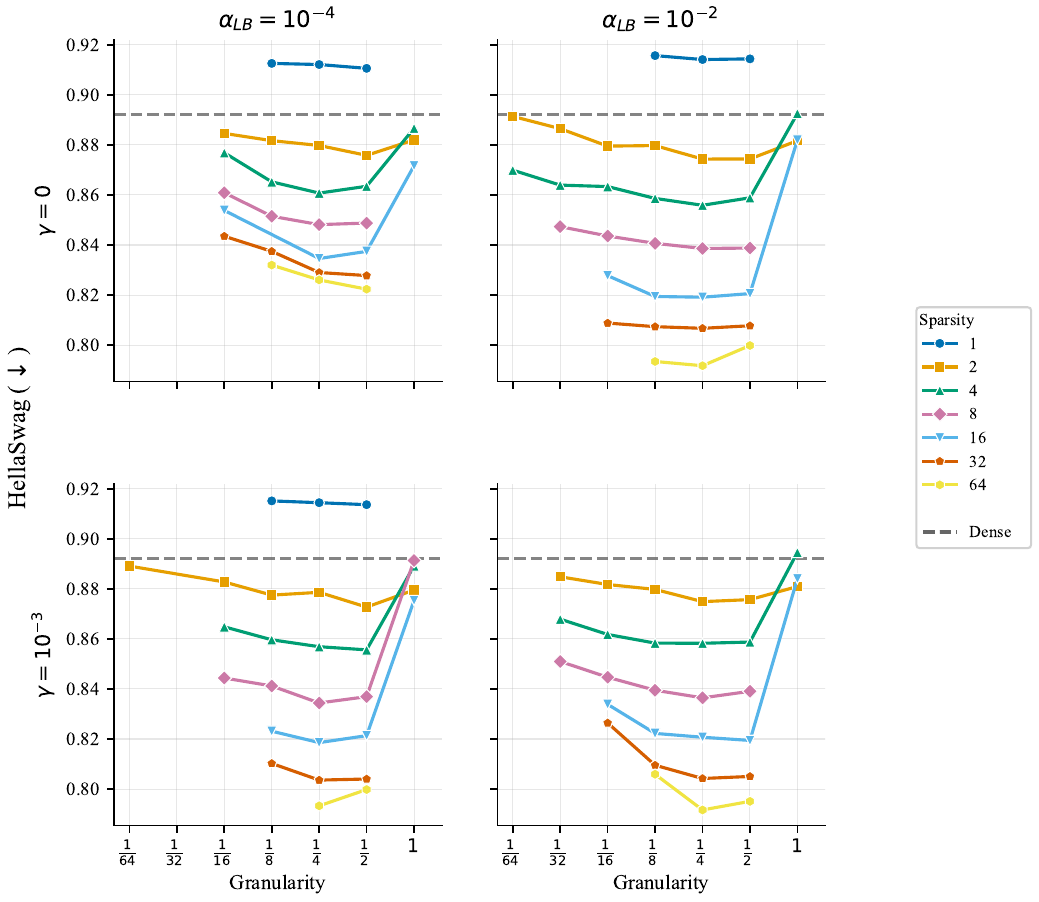}
        \hspace{1em}
        \includegraphics[width=0.46\linewidth]{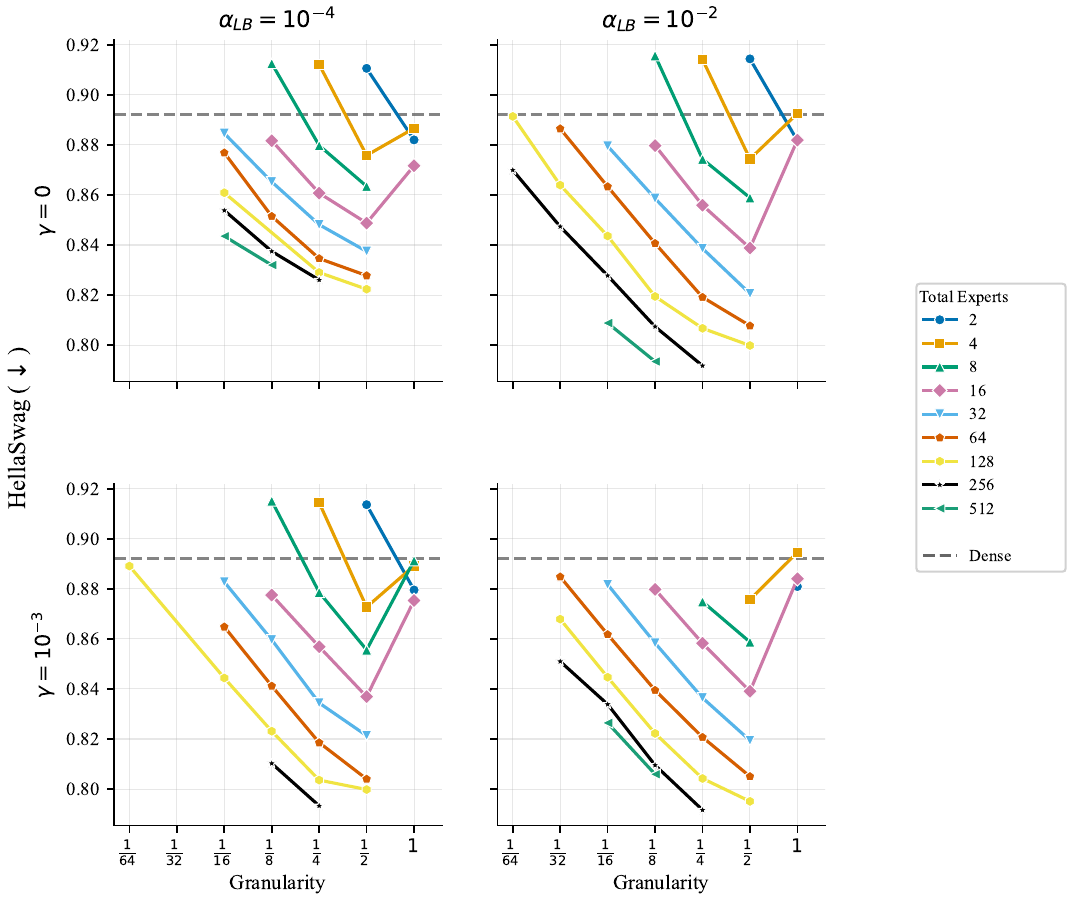}
        \caption{80M active, 80M - 765M total parameters}
    \end{subfigure}
    \par\bigskip\bigskip
    \begin{subfigure}[t]{\textwidth}
        \centering
        \includegraphics[width=0.46\linewidth]{figures/downstream/hellaswag/ce_loss/lb_sweep_hgn_gxs_110M.pdf}
        \hspace{1em}
        \includegraphics[width=0.46\linewidth]{figures/downstream/hellaswag/ce_loss/lb_sweep_hgn_gxn_110M.pdf}
        \caption{110M active, 110M - 1.4B total parameters}
    \end{subfigure}

    \end{figure*} 

\clearpage  

\begin{figure*}[ht]
    \addtocounter{figure}{-1}
    \centering
    \begin{subfigure}[t]{\textwidth}
        \centering
        \includegraphics[width=0.46\linewidth]{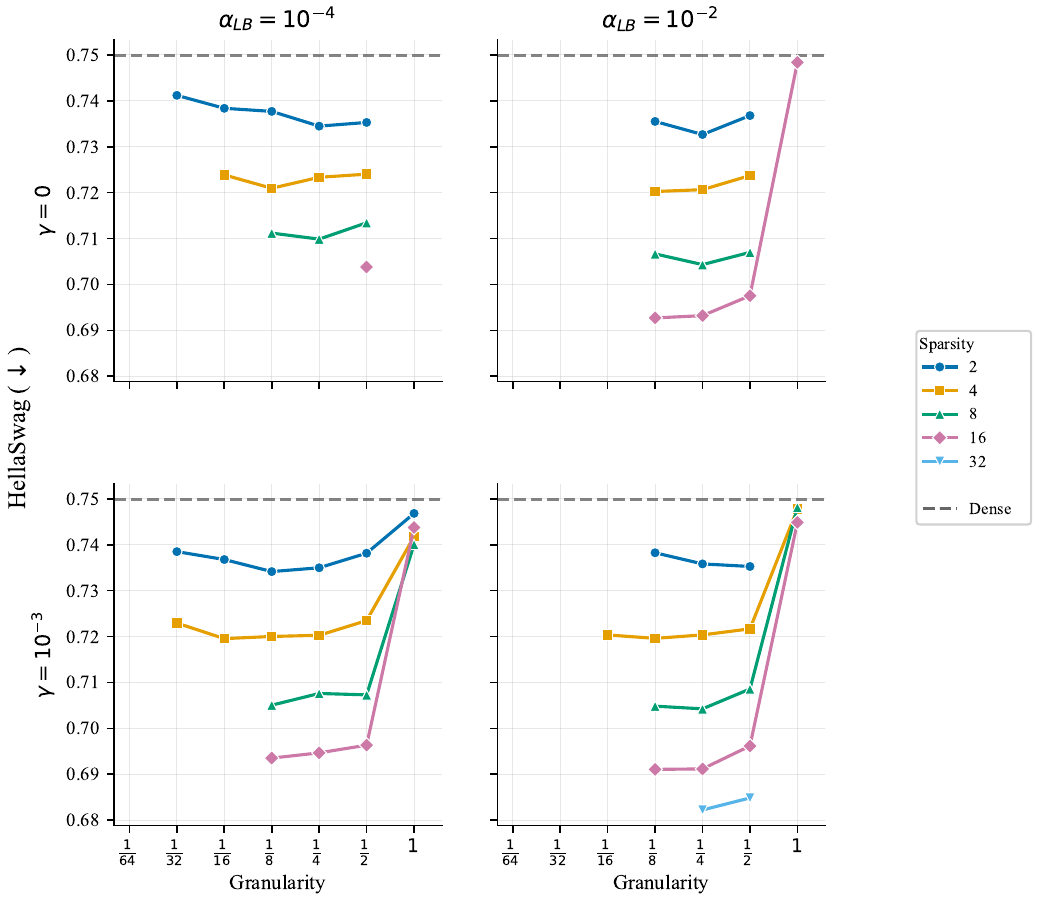}
        \hspace{1em}
        \includegraphics[width=0.46\linewidth]{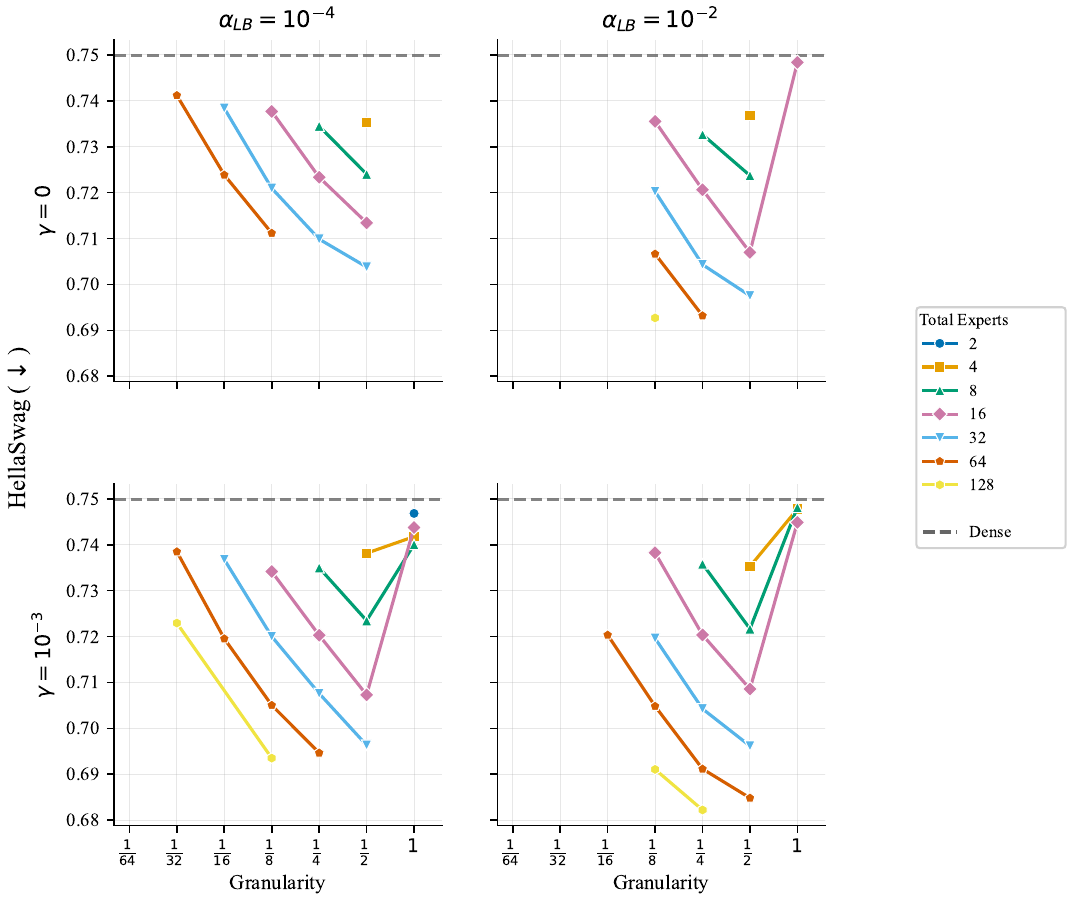}
        \caption{200M active, 200M - 3.3B total parameters}
    \end{subfigure}
    \par\bigskip\bigskip
    \begin{subfigure}[t]{\textwidth}
        \centering
        \includegraphics[width=0.3\linewidth]{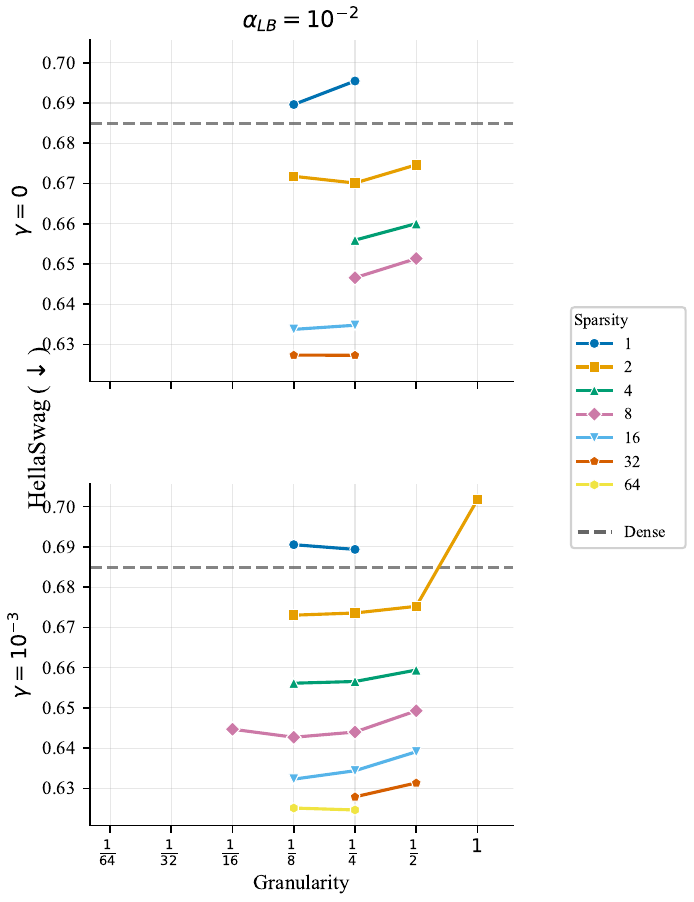}
        \hspace{1em}
        \includegraphics[width=0.3\linewidth]{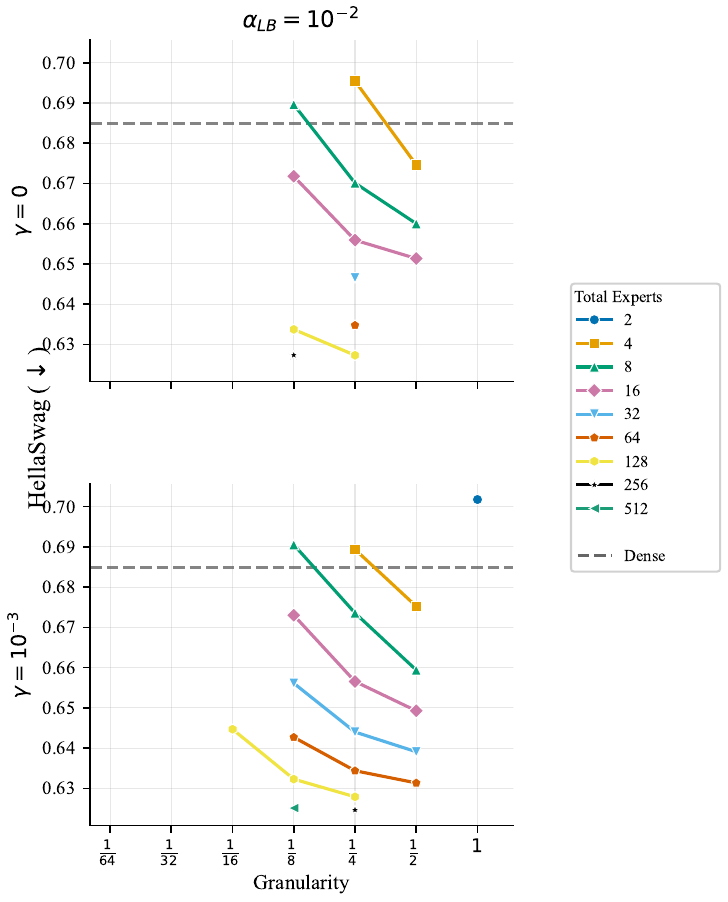}
        \caption{300M active, 300M - 6.6B total parameters}
    \end{subfigure}

    \caption{
    \textbf{Load balancing mechanisms must be tuned correctly (\S\ref{sec:expt_router}).}
    We consider load balancing loss weight $\alpha_{LB} \in \{\num{1e-2}, \num{1e-4}\}$ and loss-free load balancing with bias $\gamma\in\{0, \num{1e-3}\}$ ($\gamma=0$ indicates no loss-free mechanism). Results show that poorly chosen hyperparameters, such as high bias $\gamma = 1e-3$ with total experts $n\geq 512$, may impair performance. However, all settings other than $(\alpha_{LB}=\num{1e-2}, \gamma=\num{1e-3})$ perform comparably for $n \leq 512$, suggesting that a wide range of load balancing settings achieve near-optimal performance. 
    }
    \label{fig:hellaswag_lb}
\end{figure*}

%% file: fig_tex/downstream/mmlu_humanities.tex
\begin{figure*}[!ht]
    \centering
        \begin{subfigure}[t]{\textwidth}
        \begin{subfigure}[t]{0.33\textwidth}
            \centering
            \caption*{\scriptsize Fixed total experts (n)}
            \includegraphics[width=\linewidth]{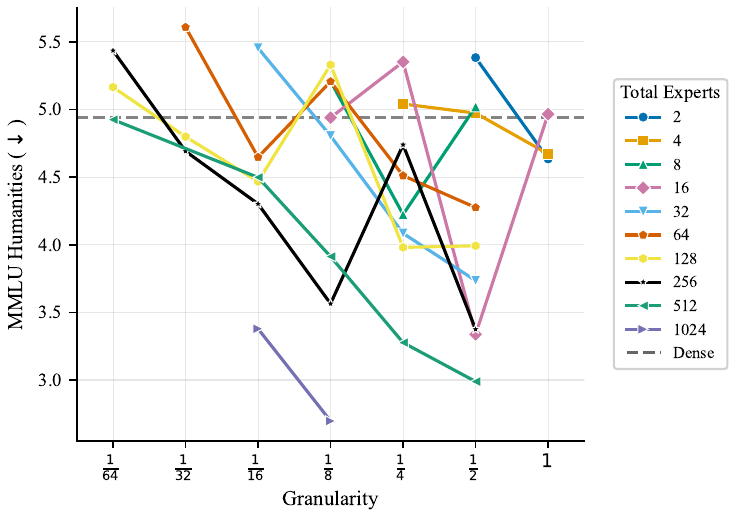}
        \end{subfigure}
        \begin{subfigure}[t]{0.33\textwidth}
            \centering
            \caption*{\scriptsize Fixed granularity (g)}
            \includegraphics[width=\linewidth]{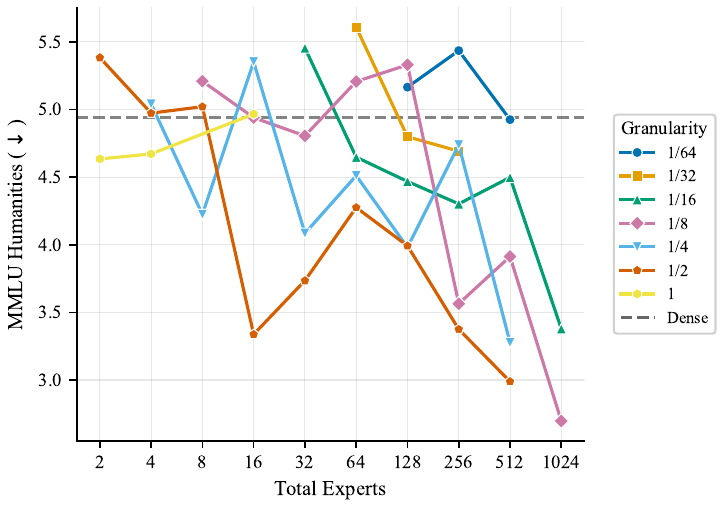}
        \end{subfigure}
        \begin{subfigure}[t]{0.33\textwidth}
            \centering
            \caption*{\scriptsize Fixed activation sparsity (s)}
            \includegraphics[width=\linewidth]{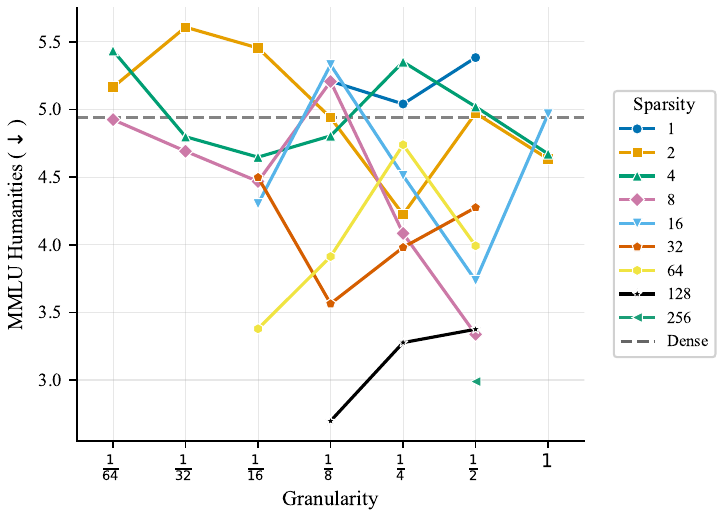}
        \end{subfigure}
        \caption{50M active, 50M - 930M total parameters}
    \end{subfigure}
\par\bigskip\bigskip
    \begin{subfigure}[t]{\textwidth}
        \begin{subfigure}[t]{0.33\textwidth}
            \centering
            \includegraphics[width=\linewidth]{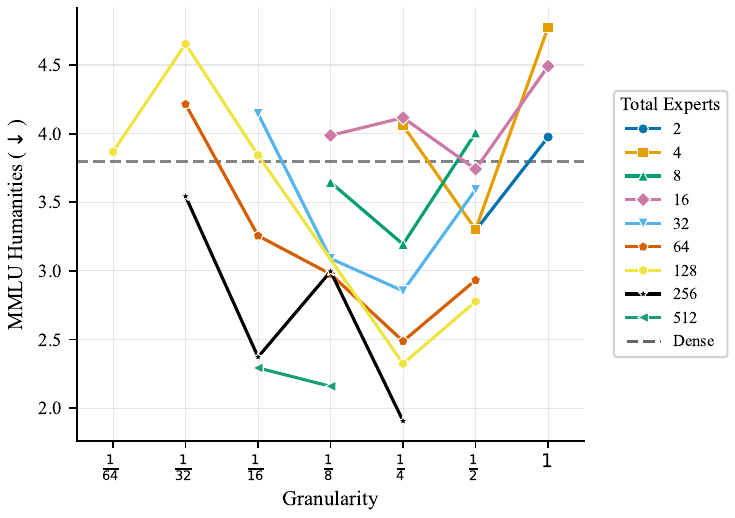}
        \end{subfigure}
        \begin{subfigure}[t]{0.33\textwidth}
            \centering
            \includegraphics[width=\linewidth]{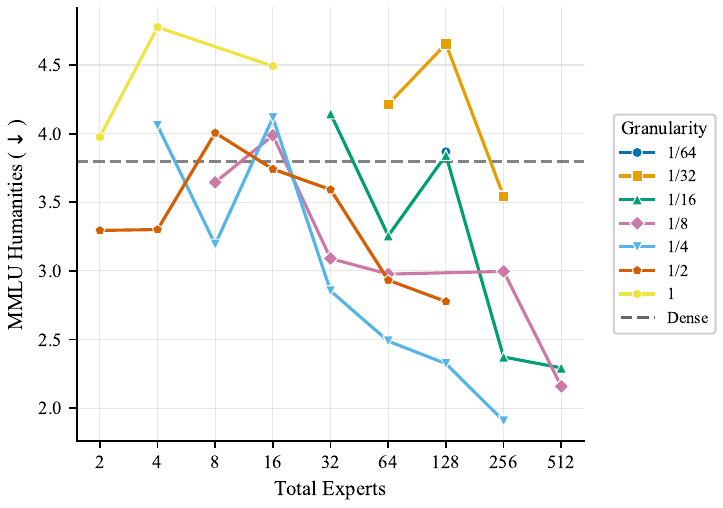}
        \end{subfigure}
        \begin{subfigure}[t]{0.33\textwidth}
            \centering
            \includegraphics[width=\linewidth]{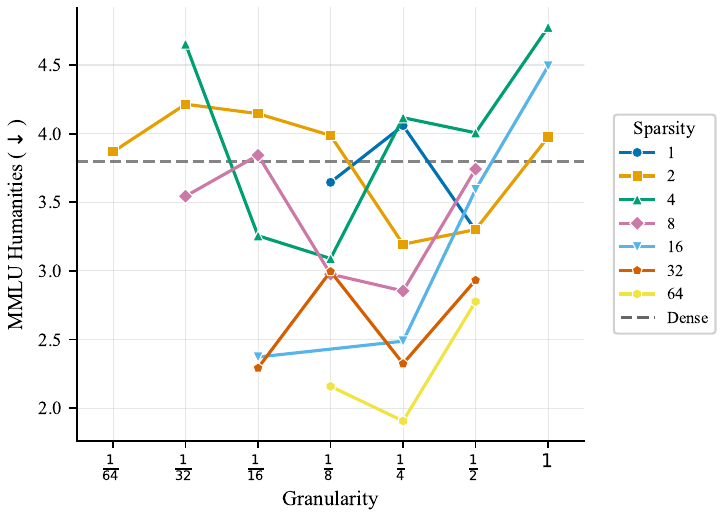}
        \end{subfigure}
        \caption{80M active, 80M - 765M total parameters}
    \end{subfigure}
    \par\bigskip\bigskip
        \begin{subfigure}[t]{\textwidth}
        \begin{subfigure}[t]{0.33\textwidth}
            \centering
            \includegraphics[width=\linewidth]{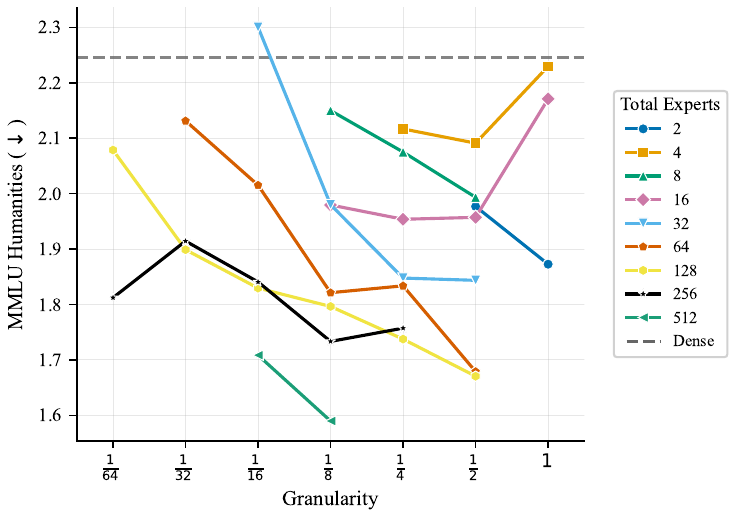}
        \end{subfigure}
        \begin{subfigure}[t]{0.33\textwidth}
            \centering
            \includegraphics[width=\linewidth]{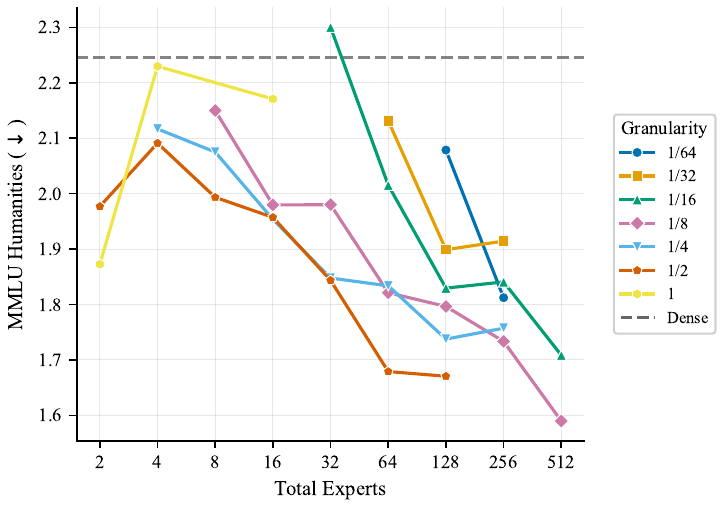}
        \end{subfigure}
        \begin{subfigure}[t]{0.33\textwidth}
            \centering
            \includegraphics[width=\linewidth]{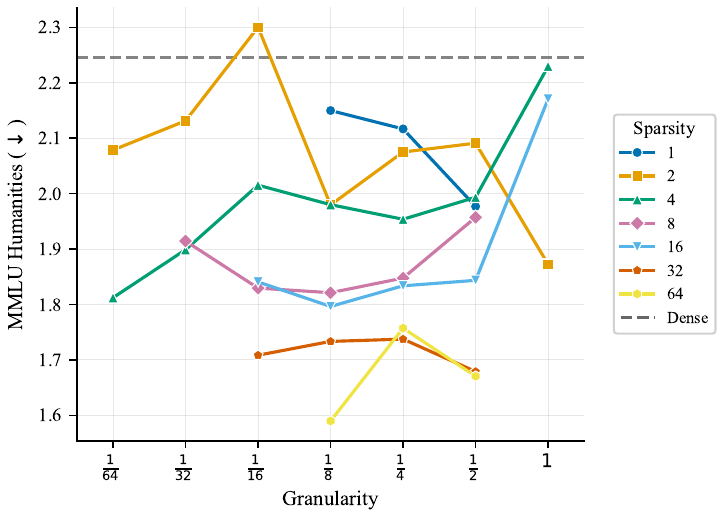}
        \end{subfigure}
        \caption{110M active, 110M - 1.4B total parameters}
    \end{subfigure}
    \end{figure*}

\clearpage  

\begin{figure*}[!ht]
        \addtocounter{figure}{-1}
    \begin{subfigure}[t]{\textwidth}
        \addtocounter{subfigure}{3}
        \begin{subfigure}[t]{0.33\textwidth}
            \centering
            \caption*{\scriptsize Fixed total experts (n)}
            \includegraphics[width=\linewidth]{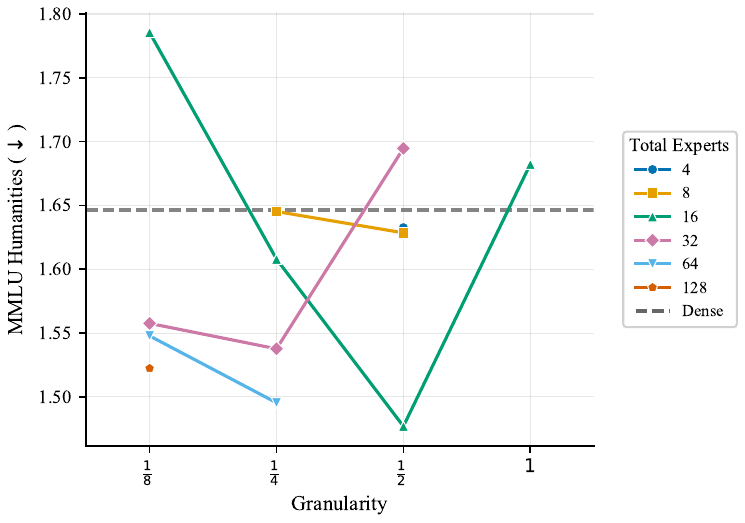}
        \end{subfigure}
        \begin{subfigure}[t]{0.33\textwidth}
            \centering
            \caption*{\scriptsize Fixed granularity (g)}
            \includegraphics[width=\linewidth]{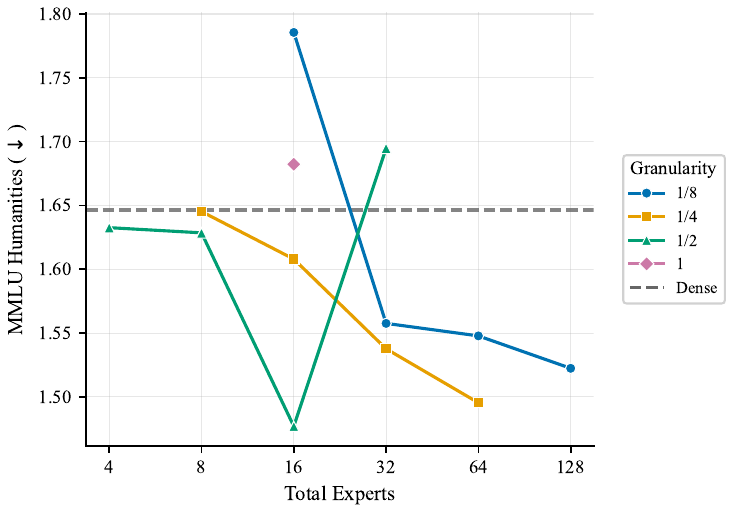}
        \end{subfigure}
        \begin{subfigure}[t]{0.33\textwidth}
            \centering
            \caption*{\scriptsize Fixed activation sparsity (s)}
            \includegraphics[width=\linewidth]{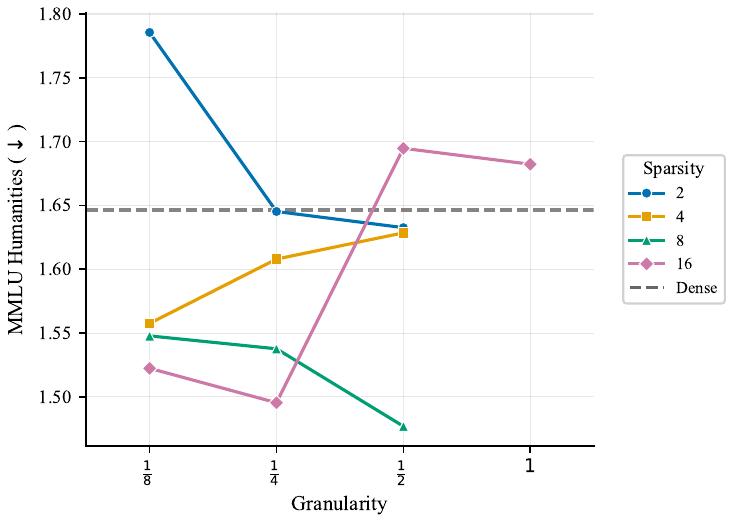}
        \end{subfigure}
        \caption{200M active, 200M - 3.3B total parameters}
    \end{subfigure}
    \par\bigskip\bigskip
        \begin{subfigure}[t]{\textwidth}
        \begin{subfigure}[t]{0.33\textwidth}
            \centering
            \includegraphics[width=\linewidth]{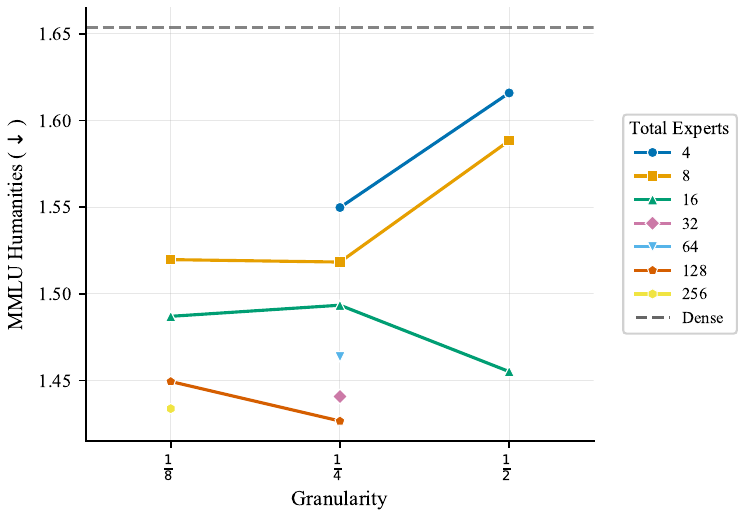}
        \end{subfigure}
        \begin{subfigure}[t]{0.33\textwidth}
            \centering
            \includegraphics[width=\linewidth]{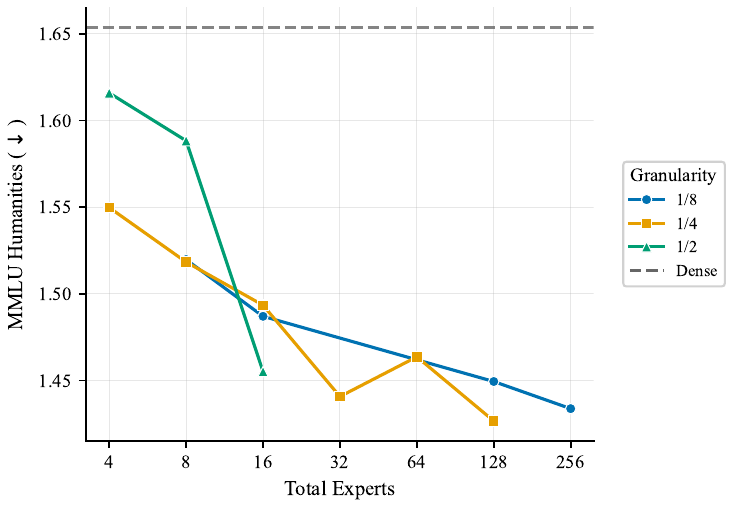}
        \end{subfigure}
        \begin{subfigure}[t]{0.33\textwidth}
            \centering
            \includegraphics[width=\linewidth]{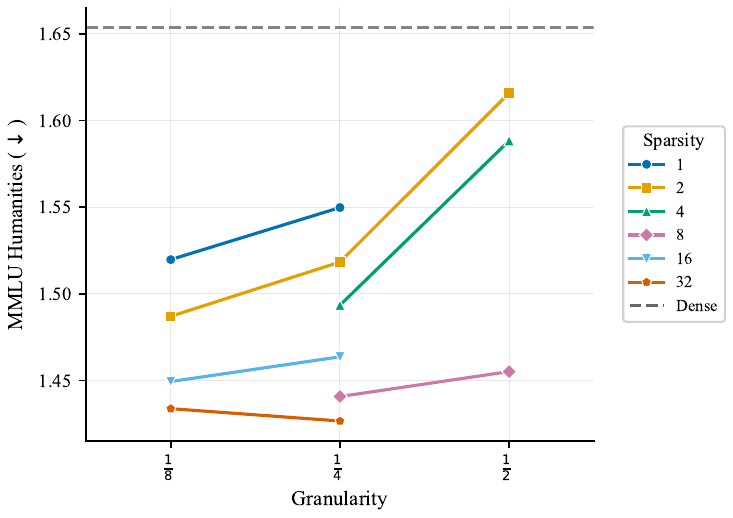}
        \end{subfigure}
        \caption{300M active, 300M - 6.6B total parameters}
    \end{subfigure}

    \caption{
    \textbf{Increasing inactive expert parameters via expert size (left) or total count (center) improves performance in MoEs (\S\ref{sec:expt_main}).} This effect is seen both when holding total number of experts fixed (left) and when holding expert granularity fixed (center). In general, increasing total parameters results in improved performance.  \textbf{Optimal tradeoff between expert count and granularity varies in MoEs (right). (\S\ref{sec:expt_main})}
    At each activation sparsity $s$ (equivalently, at each total parameter count), the optimal (total expert count, expert granularity) configuration varies. As $s$ increases, optimal expert granularity remains nearly fixed, suggesting that sparsity should be scaled up primarily by increasing total expert count $n$, while maintaining a near constant, slowly increasing expert granularity $g$. 
    }
    \label{fig:mmlu_humanities_experts}
\end{figure*}

\begin{figure*}[!ht]
    \centering
    
    \begin{subfigure}[t]{0.46\textwidth}
        \centering
        \includegraphics[width=\linewidth]{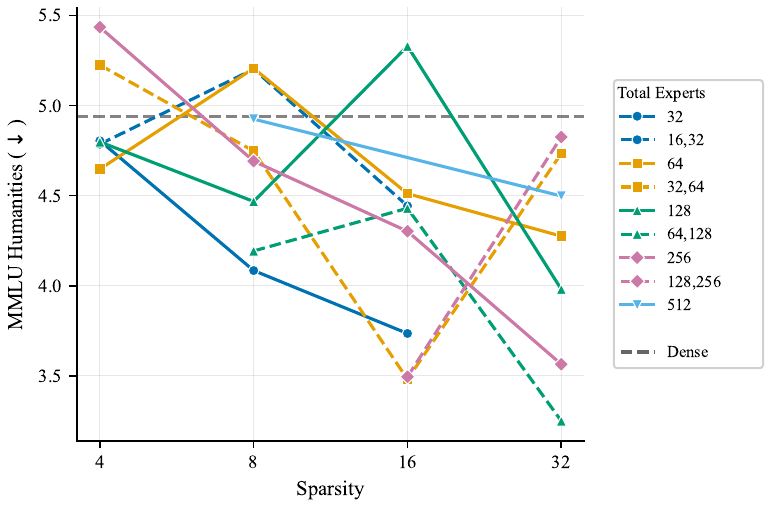}
        \caption{50M active, 50M - 930M total parameters}
    \end{subfigure}
    \vspace{1em}
    \begin{subfigure}[t]{0.46\textwidth}
        \centering
        \includegraphics[width=\linewidth]{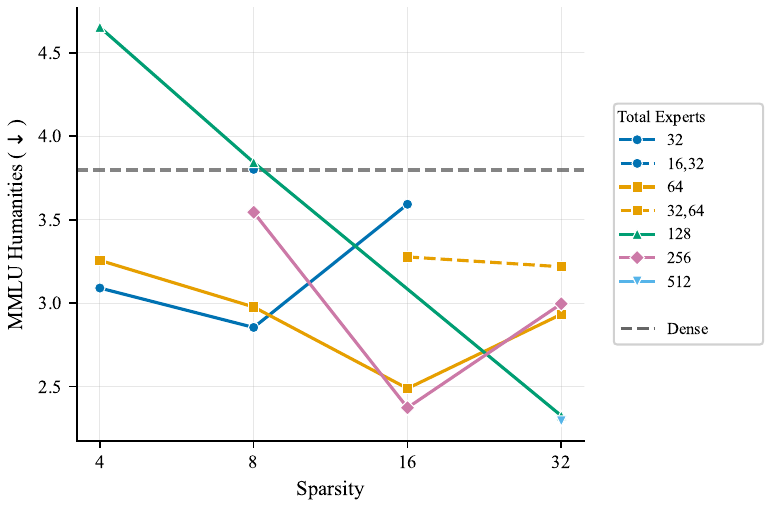}
        \caption{80M active, 80M - 765M total parameters}
    \end{subfigure}
    \caption{
    \textbf{Heterogeneity of expert size alone does not improve MoE performance (\S\ref{sec:expt_hetgen}).} To explore the potential benefits of their architectural flexibility, we compare heterogeneous MoEs (indicated by dotted lines) to active- and total-parameter-matched homogeneous MoEs. Heterogeneity alone does not result in performance gains, as, at each activation sparsity $s$, heterogeneous MoEs with $n_1, n_2 = a, b$ lie between or near the 2 closest homogeneous MoEs, with $n=a$ and with $n=b$.
    }
    \label{fig:mmlu_humanities_het}
\end{figure*}

\begin{figure*}[!ht]
    \centering
    
    \begin{subfigure}[t]{1.0\textwidth}
        \centering
        \includegraphics[width=\linewidth]{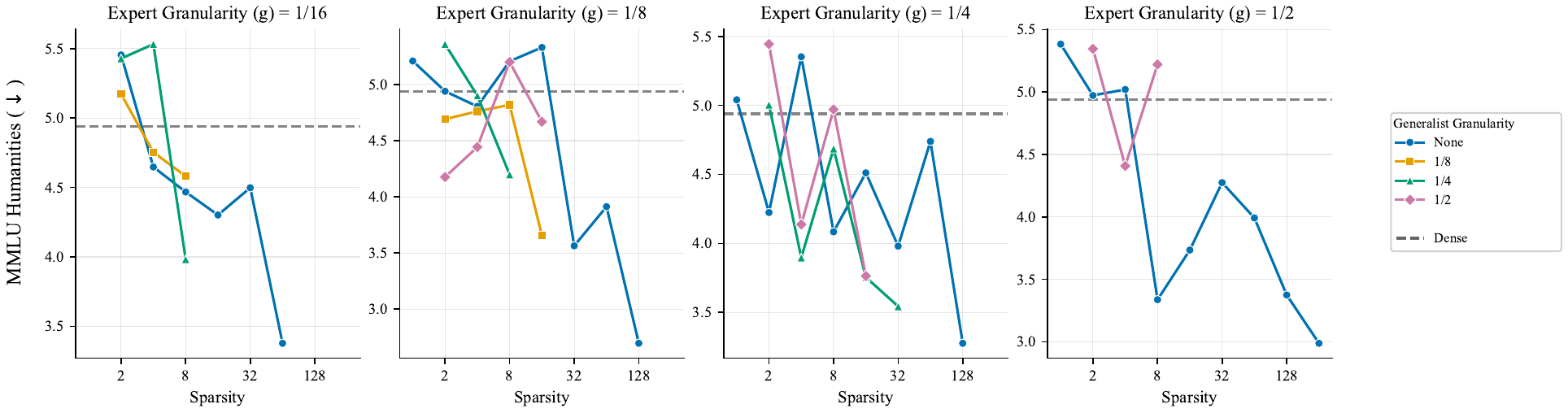}
        \caption{50M active, 50M - 930M total parameters}
    \end{subfigure}
    \par\bigskip\bigskip
    \begin{subfigure}[t]{1.0\textwidth}
        \centering
        \includegraphics[width=\linewidth]{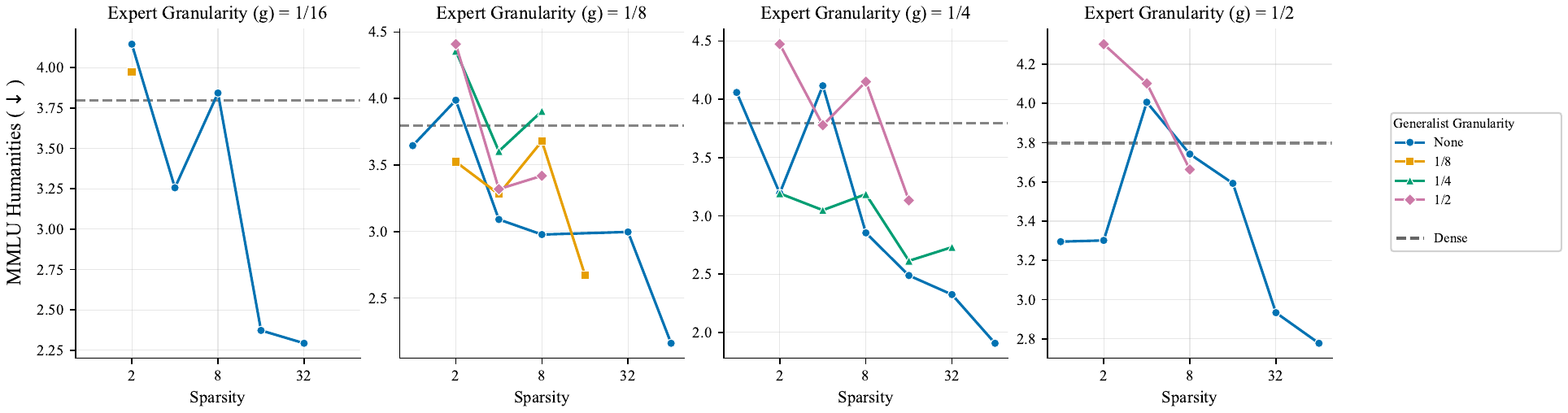}
        \caption{80M active, 80M - 765M total parameters}
    \end{subfigure}
    \par\bigskip\bigskip
    \begin{subfigure}[t]{1.0\textwidth}
        \centering
        \includegraphics[width=\linewidth]{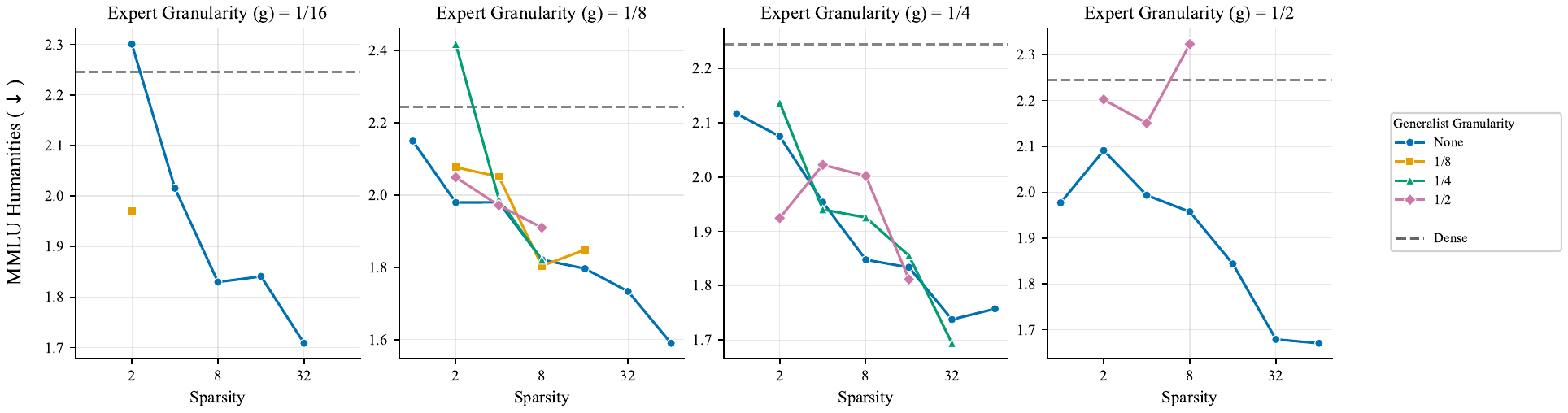}
        \caption{110M active, 110M - 1.4B total parameters}
    \end{subfigure}
    \caption{
    \textbf{The inclusion of a generalist consistently degrades performance in homogeneous MoEs (\S\ref{sec:expt_hetgen}).}
    We train MoE LMs which consist of some routed experts with granularity $g$, as well as a generalist with granularity $g_{gen}\in \{\frac{1}{2}, \frac{1}{4}, \frac{1}{8}\} $. We compare to settings with no generalist, only routed experts with granularity $g$. In all settings and configurations, the addition of any granularity generalist results in comparable or degraded performance. 
    }
    \label{fig:mmlu_humanities_gen}
\end{figure*}

\begin{figure*}[ht]
    \centering
    \begin{subfigure}[t]{1.0\textwidth}
        \centering
        \includegraphics[width=\linewidth]{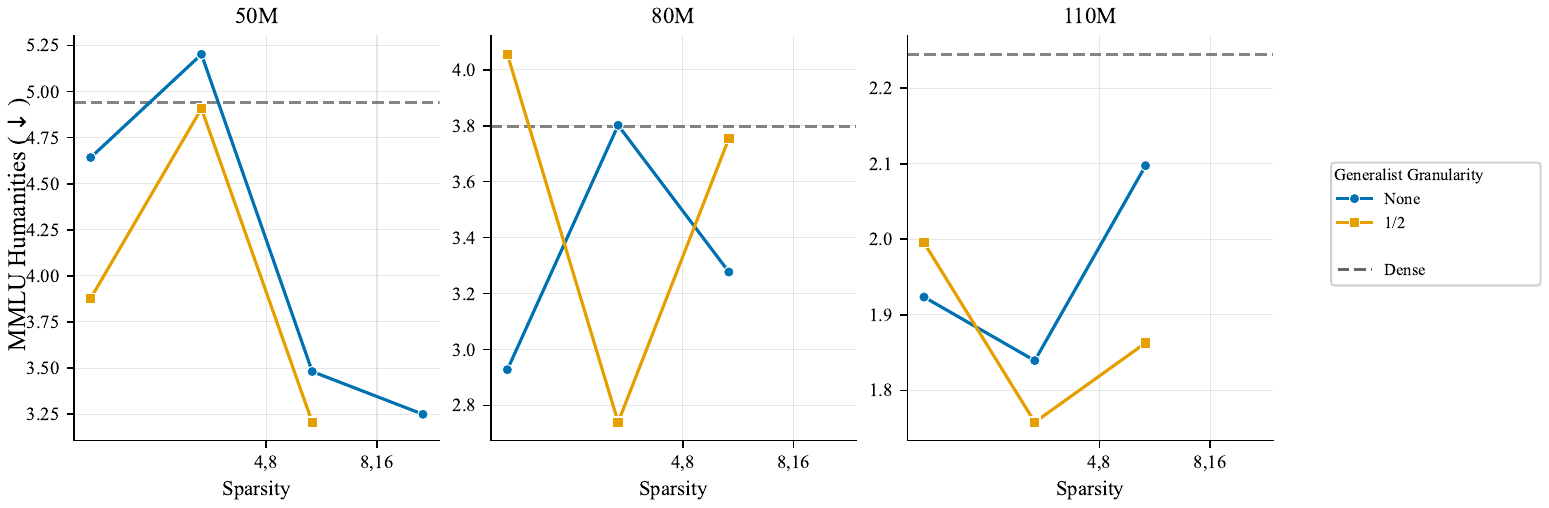}
    \end{subfigure}
    \caption{
    \textbf{The inclusion of a generalist consistently degrades performance in heterogeneous MoEs (\S\ref{sec:expt_hetgen}).}
    We train heterogeneous MoE LMs which consist of  routed experts with granularity $g_1, g_2$, as well as a generalist with granularity $g_{gen} = \frac{1}{2}$. We compare to settings with no generalist. In all settings and configurations, the addition of a generalist results in comparable or degraded performance. 
    }
    \label{fig:mmlu_humanities_hetgen}
\end{figure*}

\begin{figure*}[ht]
    \centering
    \begin{subfigure}[t]{\textwidth}
        \centering
        \begin{subfigure}[t]{0.45\textwidth}
            \includegraphics[width=\linewidth]{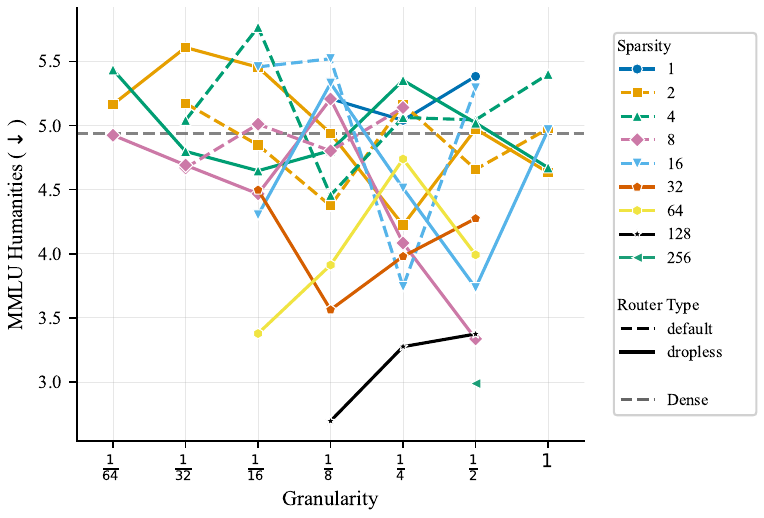}
            \caption{50M active, 50M - 930M total parameters}
        \end{subfigure}
    \hspace{1em}
        \begin{subfigure}[t]{0.45\textwidth}
            \centering
            \includegraphics[width=\linewidth]{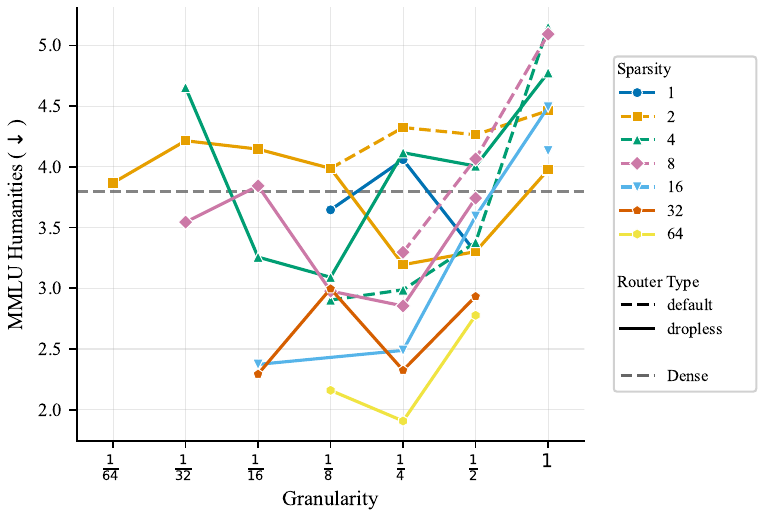}
            \caption{80M active, 80M - 765M total parameters}
        \end{subfigure}
    \end{subfigure}

    \par\bigskip\bigskip
    \begin{subfigure}[t]{0.45\textwidth}
        \centering
        \includegraphics[width=\linewidth]{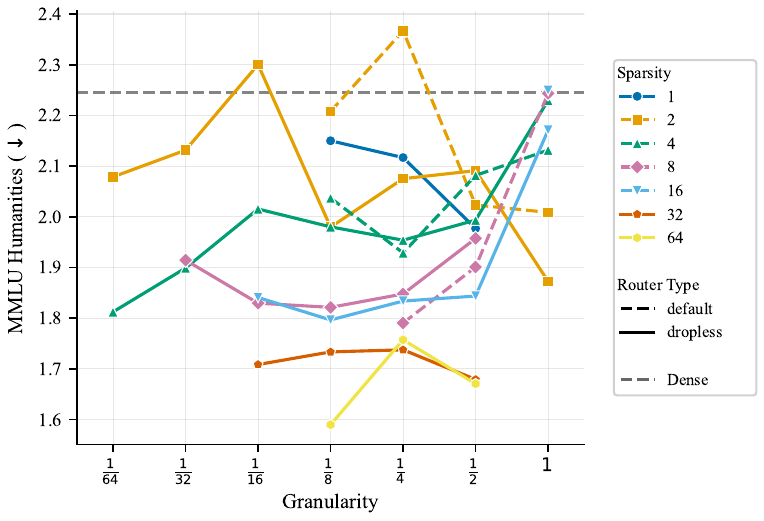}
        \caption{110M active, 110M - 1.4B total parameters}
    \end{subfigure}
    \caption{ 
    \textbf{Dropless routing outperforms default routing (\S\ref{sec:expt_router}).}
    We compare dropless routing to the default setting, which allow tokens to be dropped. Across all scales, we find that dropless routing outperforms or performs comparably to default routing. 
    }
    \label{fig:mmlu_humanities_dropless}
\end{figure*}

\begin{figure*}[ht]
    \centering
    \begin{subfigure}[t]{0.45\textwidth}
        \centering
        \includegraphics[width=\linewidth]{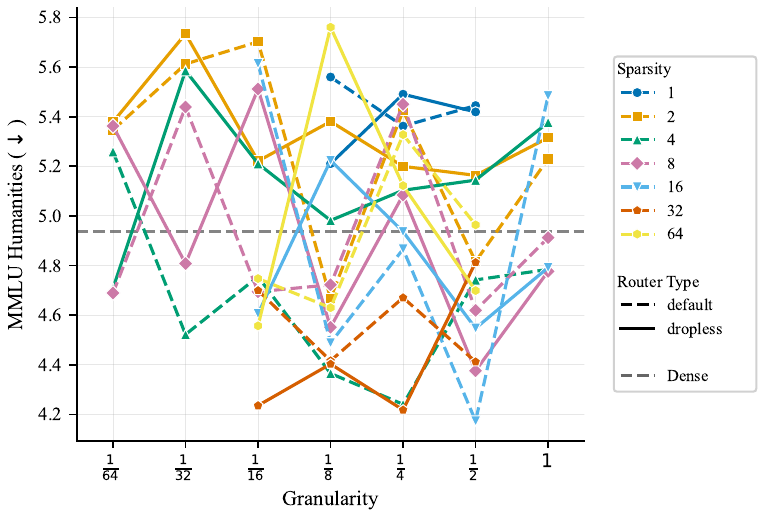}
        \caption{50M active, 50M - 930M total parameters}
    \end{subfigure}
    \hspace{1em}
    \begin{subfigure}[t]{0.45\textwidth}
        \centering
        \includegraphics[width=\linewidth]{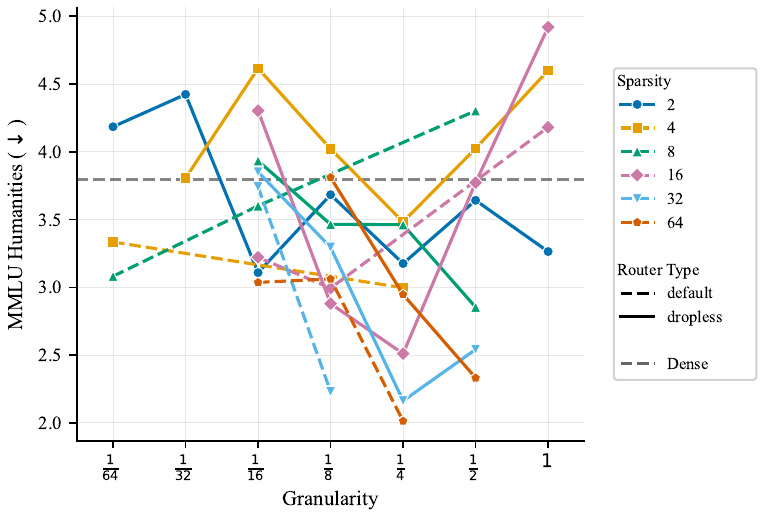}
        \caption{80M active, 80M - 765M total parameters}
    \end{subfigure}
    \caption{
    \textbf{Dropless routing, with bias $\gamma=\num{1e-3}$ (\S\ref{sec:expt_router}).} 
    As in Figure~\ref{fig:lm_avg_dropless}, we compare dropless routing to the default setting, which allow tokens to be dropped. Across all scales, we find that dropless routing outperforms or performs comparably to default routing. We see here with additional higher sparsity default routing runs that as sparsity increases, default routing performance approaches that of dropless routing.
    }
    \label{fig:mmlu_humanities_dropless_with_lf}
\end{figure*}

\begin{figure*}[ht]
    \centering
    \begin{subfigure}[]{\textwidth}
        \centering
        \includegraphics[width=0.46\linewidth]{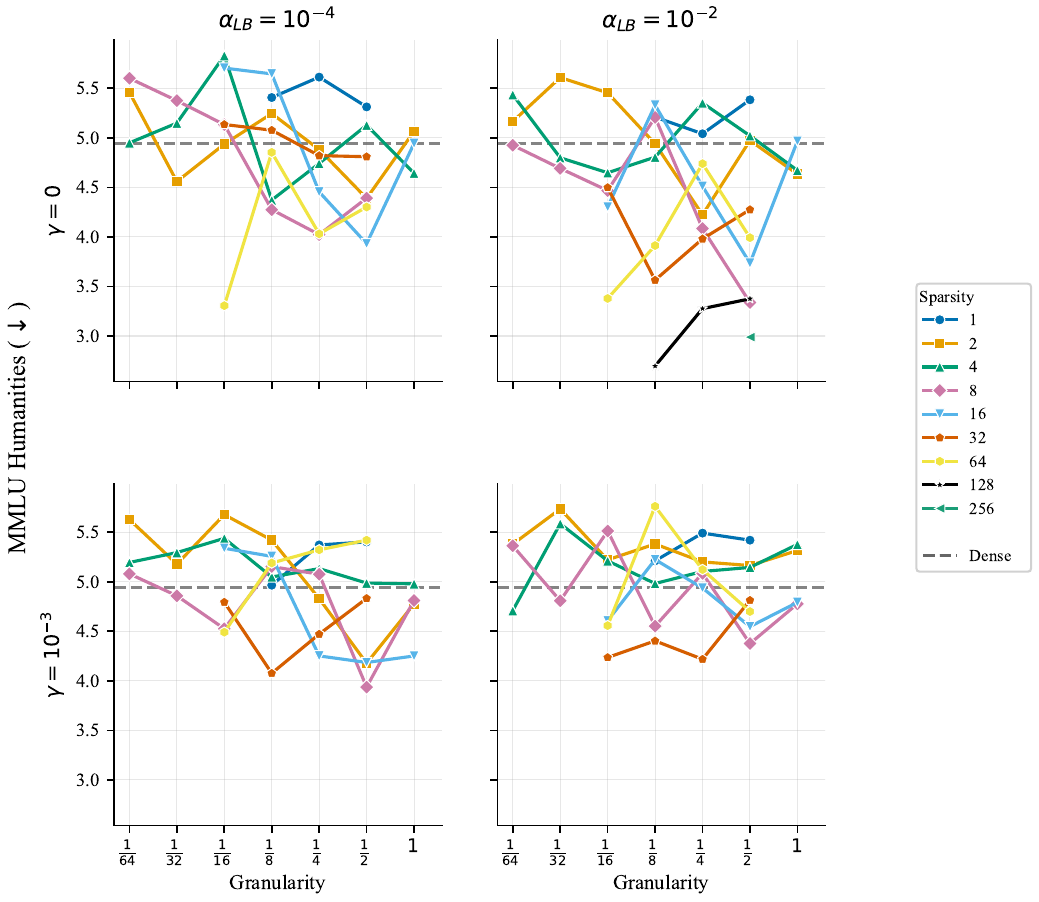}
        \hspace{1em}
        \includegraphics[width=0.46\linewidth]{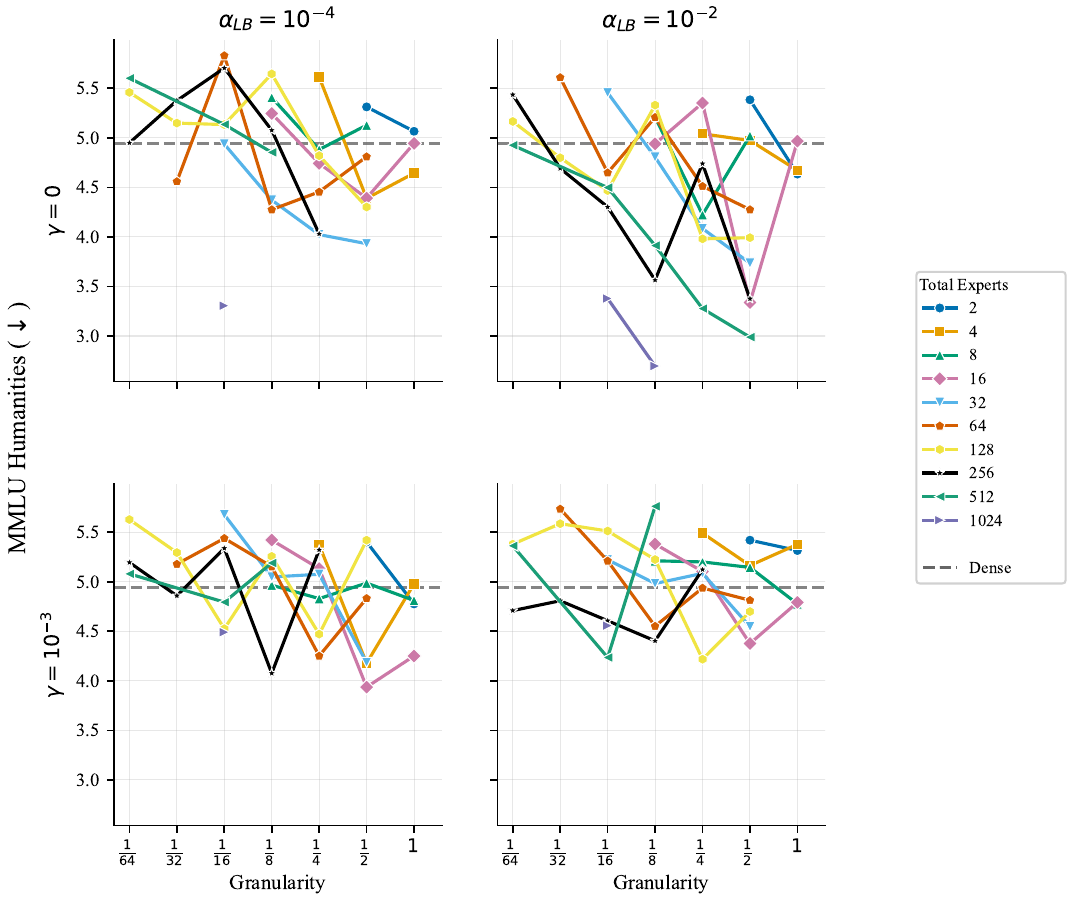}
        \caption{50M active, 50M - 930M total parameters}
    \end{subfigure}
    \par\bigskip\bigskip
    \begin{subfigure}[]{\textwidth}
        \centering
        \includegraphics[width=0.46\linewidth]{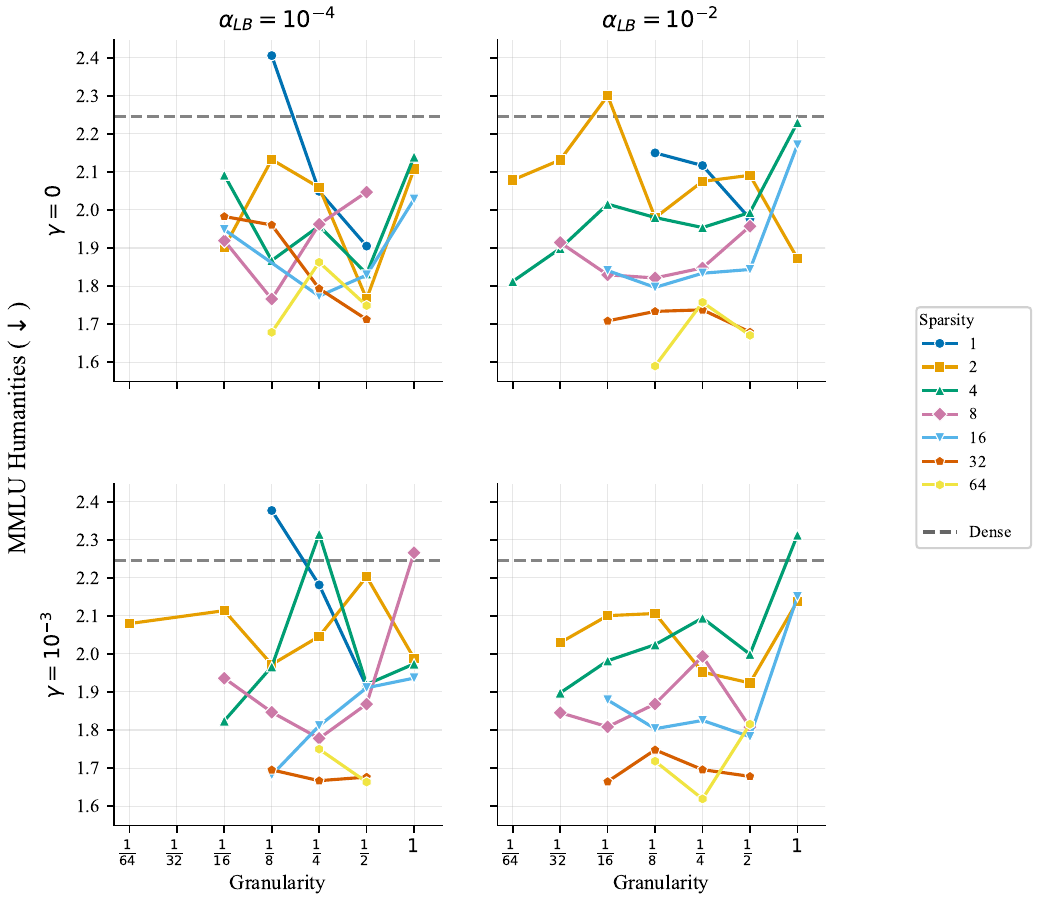}
        \hspace{1em}
        \includegraphics[width=0.46\linewidth]{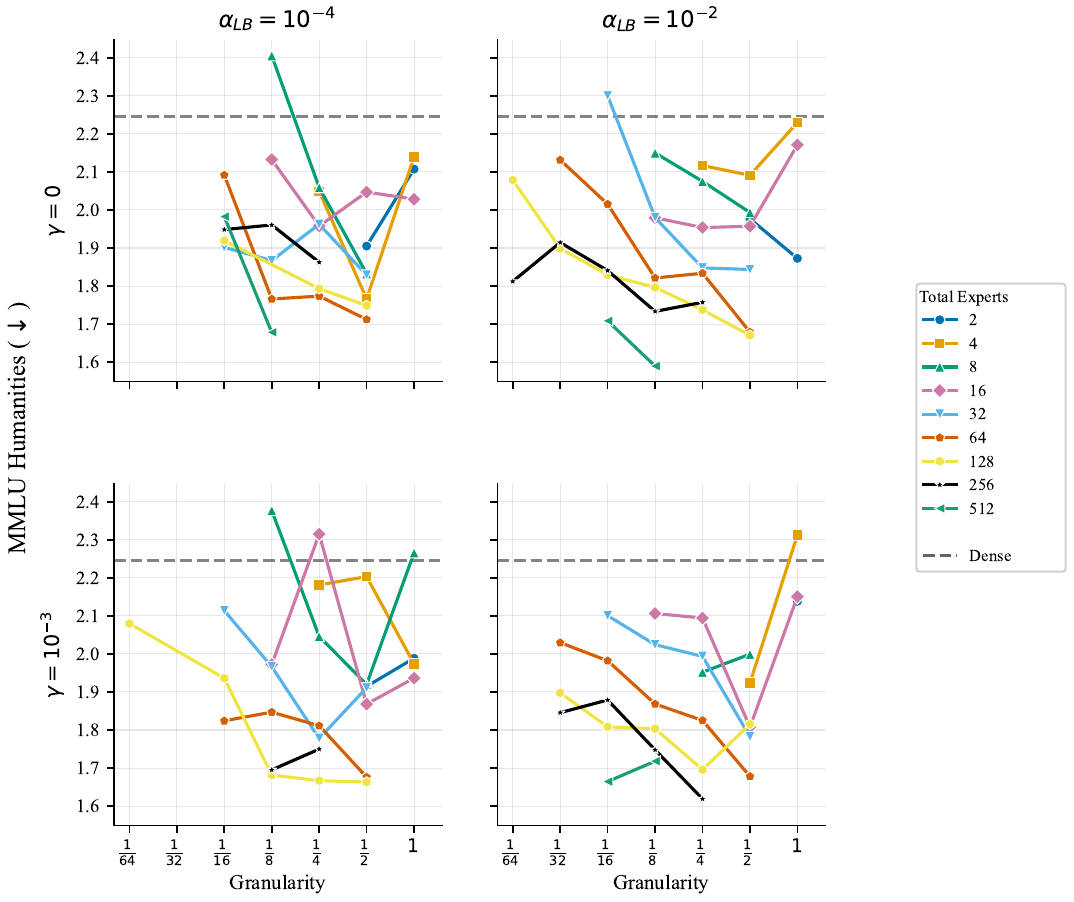}
        \caption{80M active, 80M - 765M total parameters}
    \end{subfigure}
    \par\bigskip\bigskip
    \begin{subfigure}[t]{\textwidth}
        \centering
        \includegraphics[width=0.46\linewidth]{figures/downstream/mmlu_humanities/ce_loss/lb_sweep_hgn_gxs_110M.pdf}
        \hspace{1em}
        \includegraphics[width=0.46\linewidth]{figures/downstream/mmlu_humanities/ce_loss/lb_sweep_hgn_gxn_110M.pdf}
        \caption{110M active, 110M - 1.4B total parameters}
    \end{subfigure}

    \end{figure*} 

\clearpage  

\begin{figure*}[ht]
    \addtocounter{figure}{-1}
    \centering
    \begin{subfigure}[t]{\textwidth}
        \centering
        \includegraphics[width=0.46\linewidth]{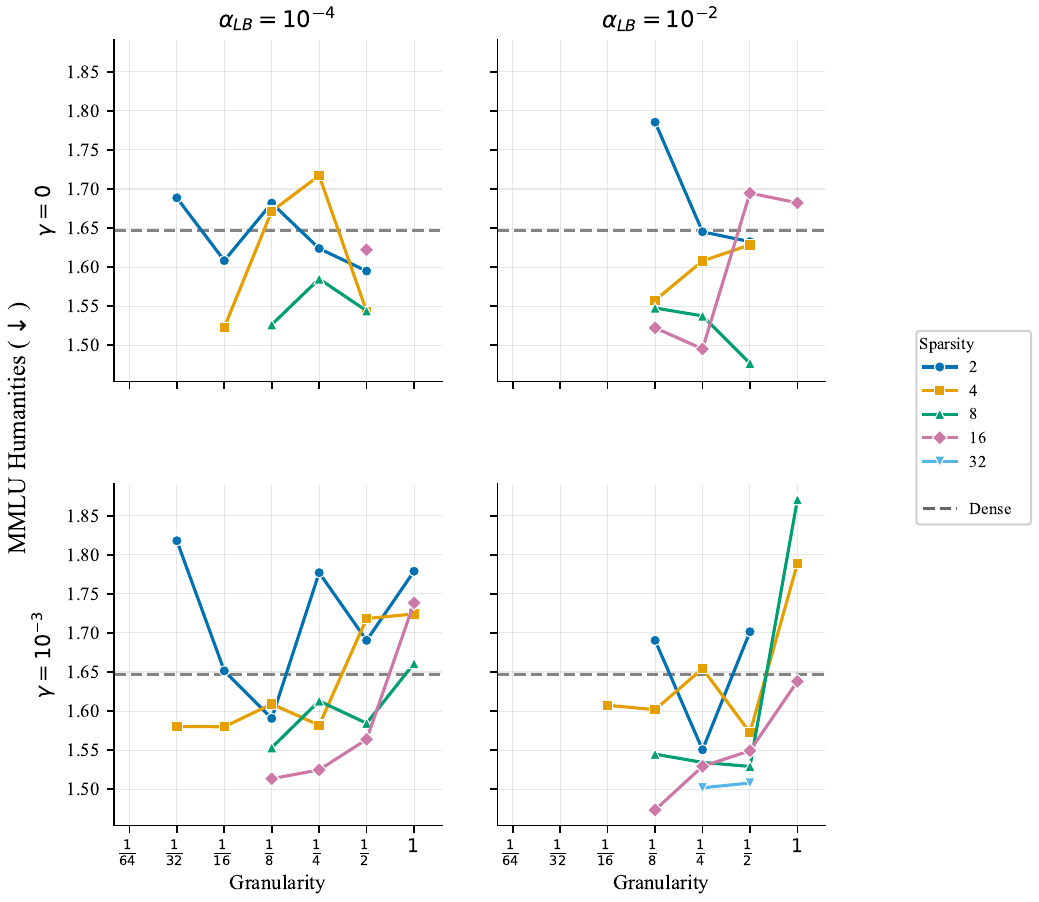}
        \hspace{1em}
        \includegraphics[width=0.46\linewidth]{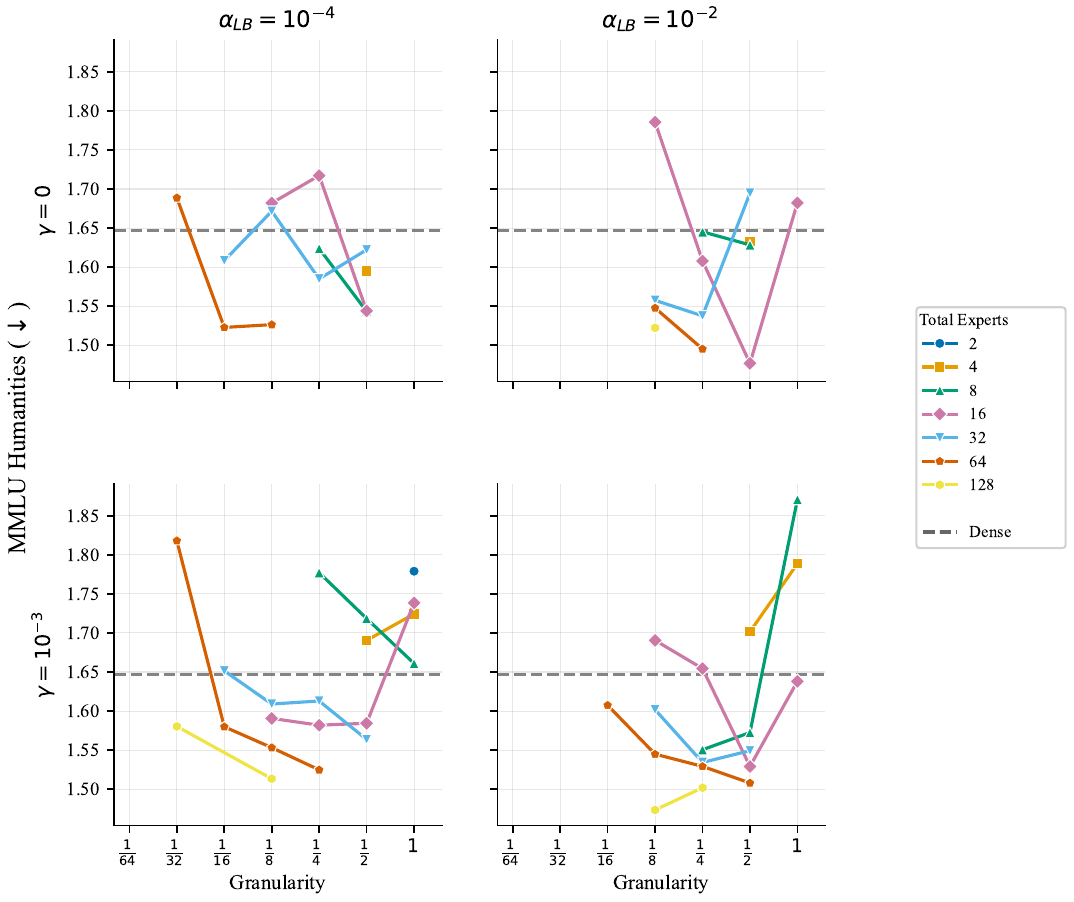}
        \caption{200M active, 200M - 3.3B total parameters}
    \end{subfigure}
    \par\bigskip\bigskip
    \begin{subfigure}[t]{\textwidth}
        \centering
        \includegraphics[width=0.3\linewidth]{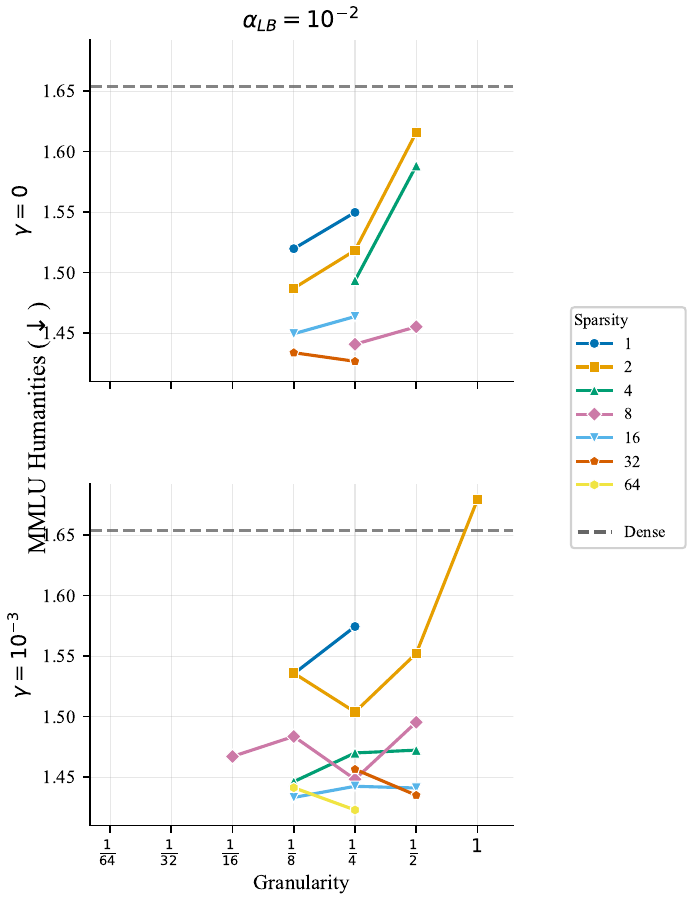}
        \hspace{1em}
        \includegraphics[width=0.3\linewidth]{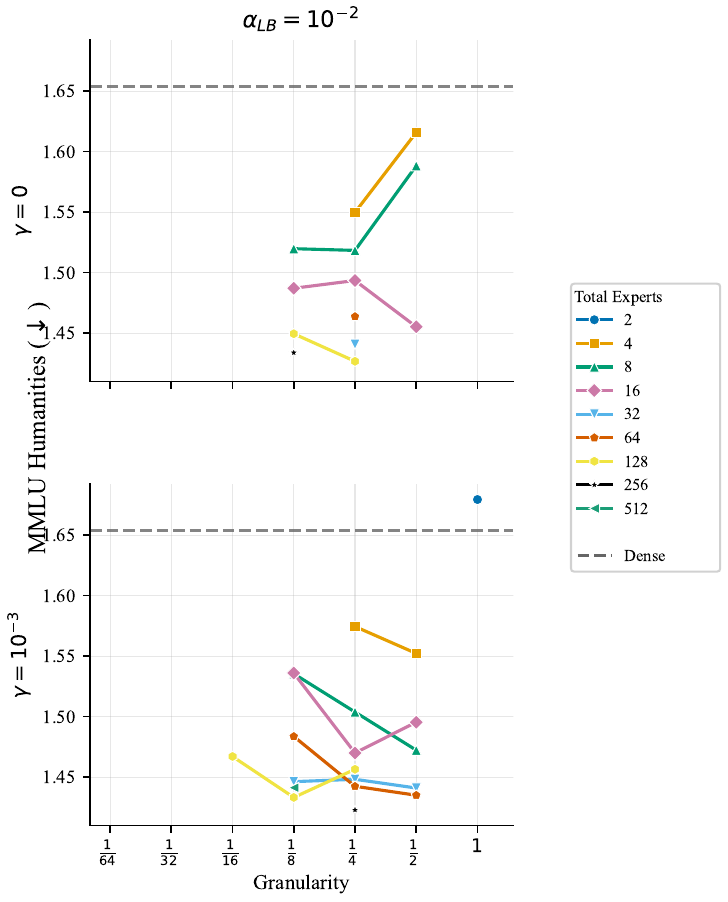}
        \caption{300M active, 300M - 6.6B total parameters}
    \end{subfigure}

    \caption{
    \textbf{Load balancing mechanisms must be tuned correctly (\S\ref{sec:expt_router}).}
    We consider load balancing loss weight $\alpha_{LB} \in \{\num{1e-2}, \num{1e-4}\}$ and loss-free load balancing with bias $\gamma\in\{0, \num{1e-3}\}$ ($\gamma=0$ indicates no loss-free mechanism). Results show that poorly chosen hyperparameters, such as high bias $\gamma = 1e-3$ with total experts $n\geq 512$, may impair performance. However, all settings other than $(\alpha_{LB}=\num{1e-2}, \gamma=\num{1e-3})$ perform comparably for $n \leq 512$, suggesting that a wide range of load balancing settings achieve near-optimal performance. 
    }
    \label{fig:mmlu_humanities_lb}
\end{figure*}

%% file: fig_tex/downstream/mmlu_other.tex
\begin{figure*}[!ht]
    \centering
        \begin{subfigure}[t]{\textwidth}
        \begin{subfigure}[t]{0.33\textwidth}
            \centering
            \caption*{\scriptsize Fixed total experts (n)}
            \includegraphics[width=\linewidth]{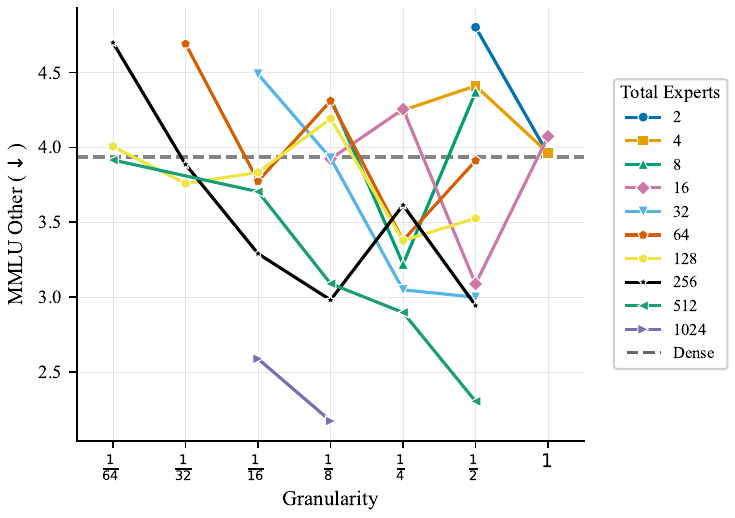}
        \end{subfigure}
        \begin{subfigure}[t]{0.33\textwidth}
            \centering
            \caption*{\scriptsize Fixed granularity (g)}
            \includegraphics[width=\linewidth]{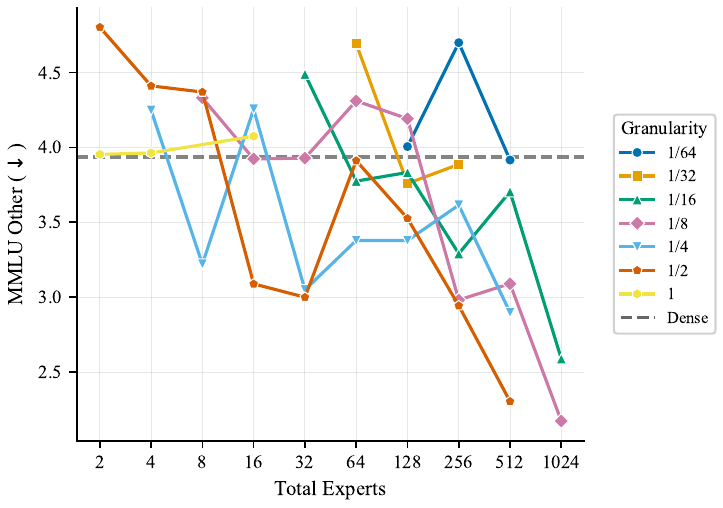}
        \end{subfigure}
        \begin{subfigure}[t]{0.33\textwidth}
            \centering
            \caption*{\scriptsize Fixed activation sparsity (s)}
            \includegraphics[width=\linewidth]{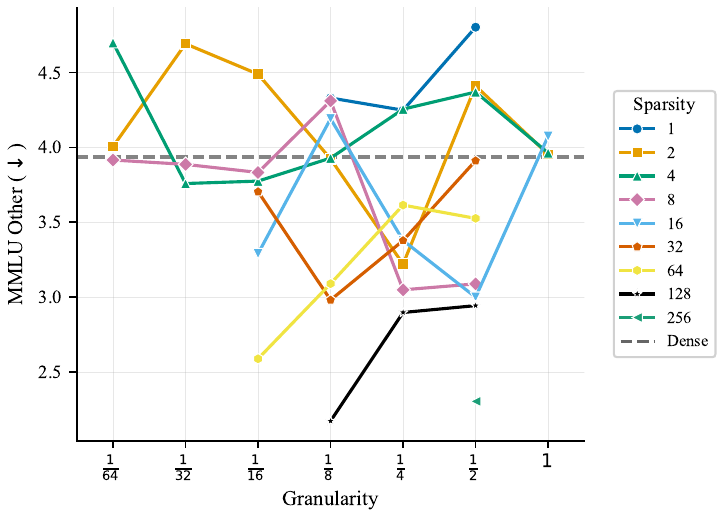}
        \end{subfigure}
        \caption{50M active, 50M - 930M total parameters}
    \end{subfigure}
\par\bigskip\bigskip
    \begin{subfigure}[t]{\textwidth}
        \begin{subfigure}[t]{0.33\textwidth}
            \centering
            \includegraphics[width=\linewidth]{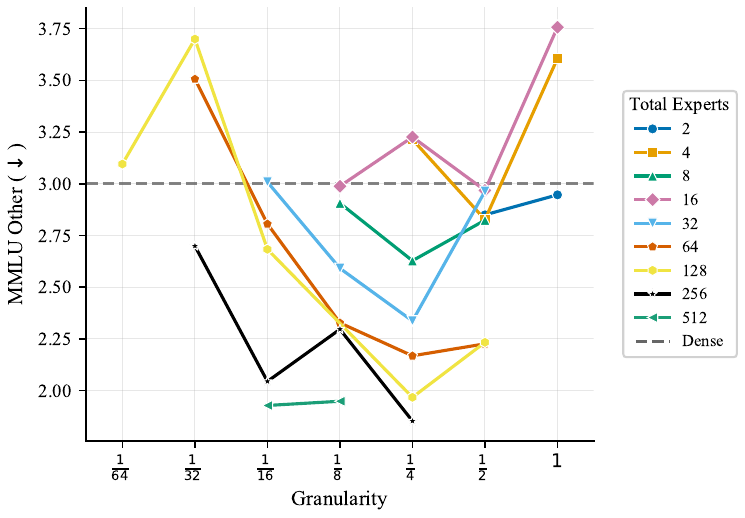}
        \end{subfigure}
        \begin{subfigure}[t]{0.33\textwidth}
            \centering
            \includegraphics[width=\linewidth]{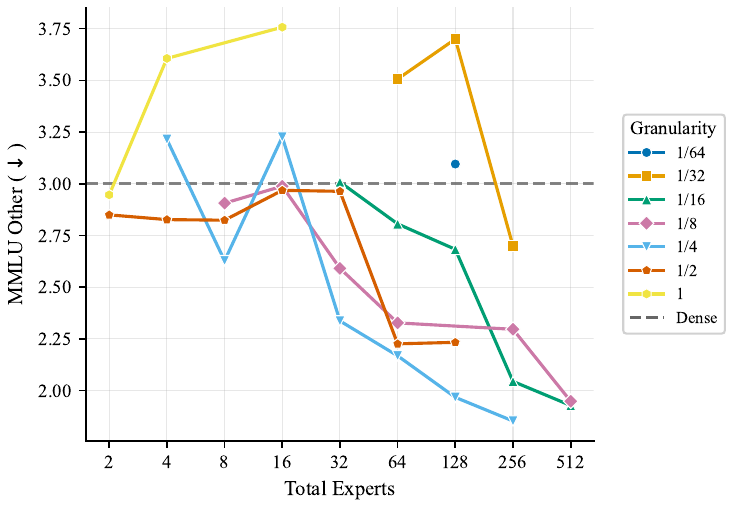}
        \end{subfigure}
        \begin{subfigure}[t]{0.33\textwidth}
            \centering
            \includegraphics[width=\linewidth]{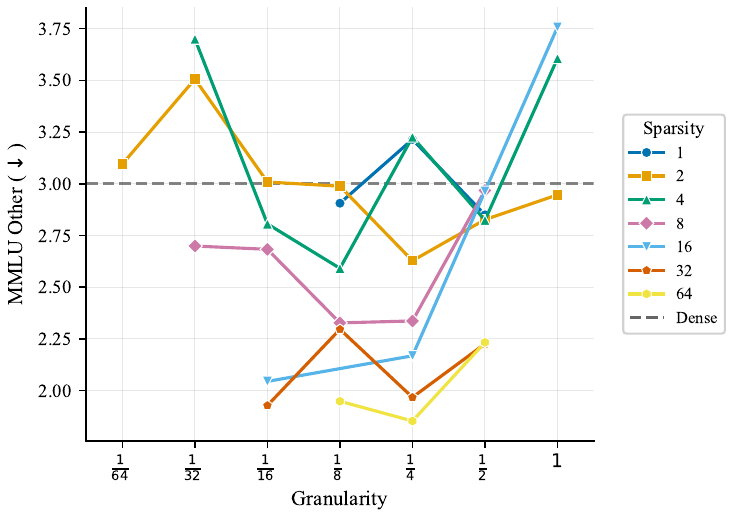}
        \end{subfigure}
        \caption{80M active, 80M - 765M total parameters}
    \end{subfigure}
    \par\bigskip\bigskip
        \begin{subfigure}[t]{\textwidth}
        \begin{subfigure}[t]{0.33\textwidth}
            \centering
            \includegraphics[width=\linewidth]{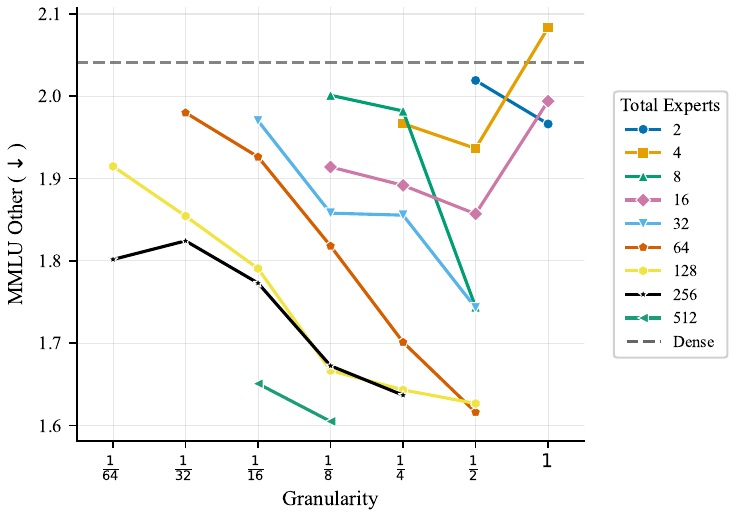}
        \end{subfigure}
        \begin{subfigure}[t]{0.33\textwidth}
            \centering
            \includegraphics[width=\linewidth]{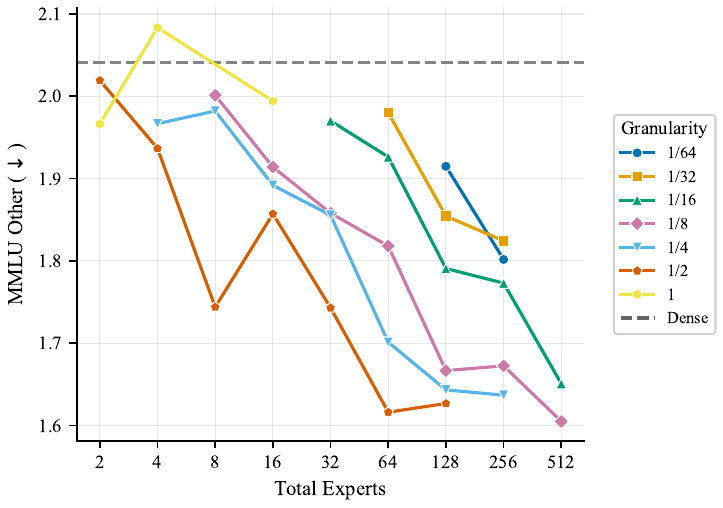}
        \end{subfigure}
        \begin{subfigure}[t]{0.33\textwidth}
            \centering
            \includegraphics[width=\linewidth]{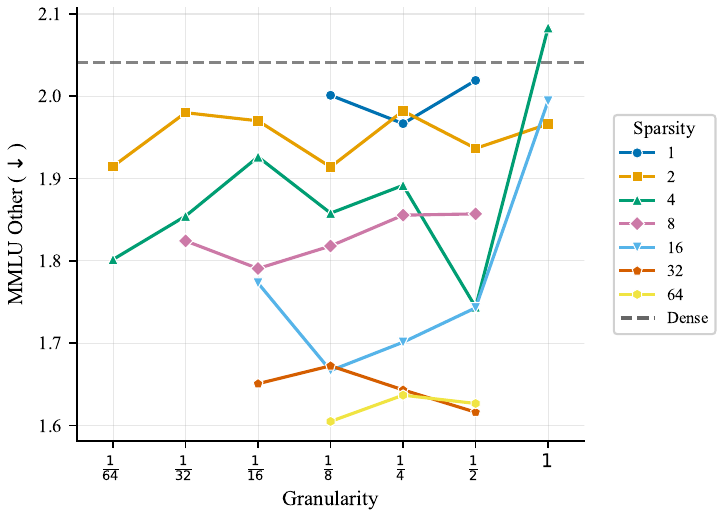}
        \end{subfigure}
        \caption{110M active, 110M - 1.4B total parameters}
    \end{subfigure}
    \end{figure*}

\clearpage  

\begin{figure*}[!ht]
        \addtocounter{figure}{-1}
    \begin{subfigure}[t]{\textwidth}
        \addtocounter{subfigure}{3}
        \begin{subfigure}[t]{0.33\textwidth}
            \centering
            \caption*{\scriptsize Fixed total experts (n)}
            \includegraphics[width=\linewidth]{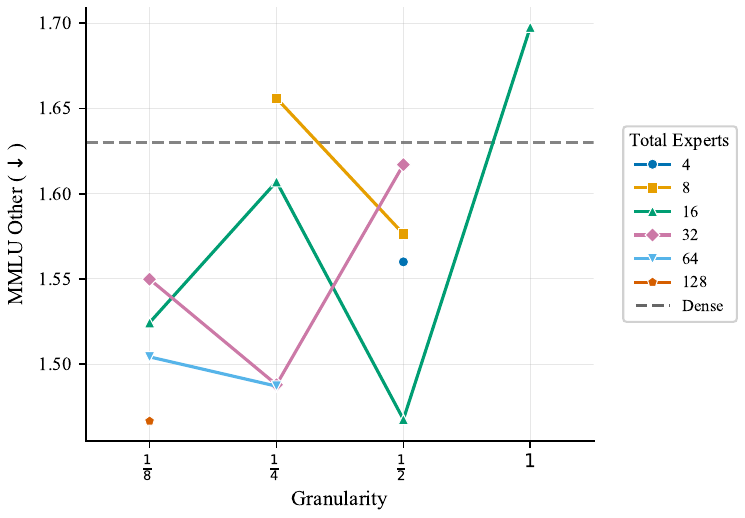}
        \end{subfigure}
        \begin{subfigure}[t]{0.33\textwidth}
            \centering
            \caption*{\scriptsize Fixed granularity (g)}
            \includegraphics[width=\linewidth]{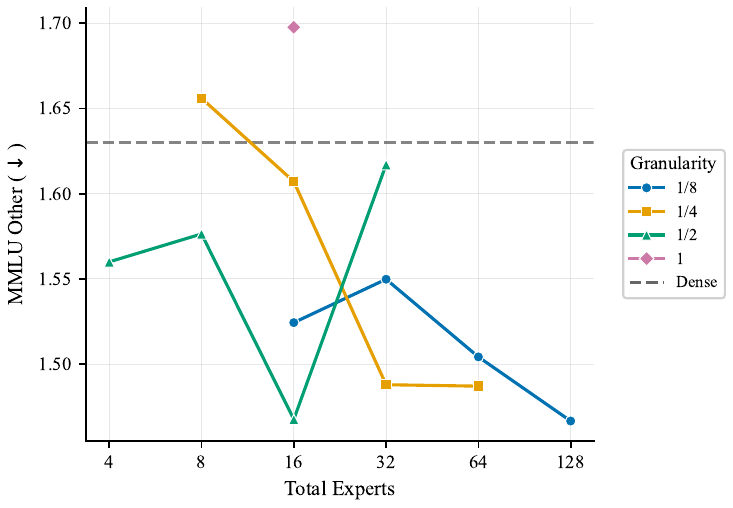}
        \end{subfigure}
        \begin{subfigure}[t]{0.33\textwidth}
            \centering
            \caption*{\scriptsize Fixed activation sparsity (s)}
            \includegraphics[width=\linewidth]{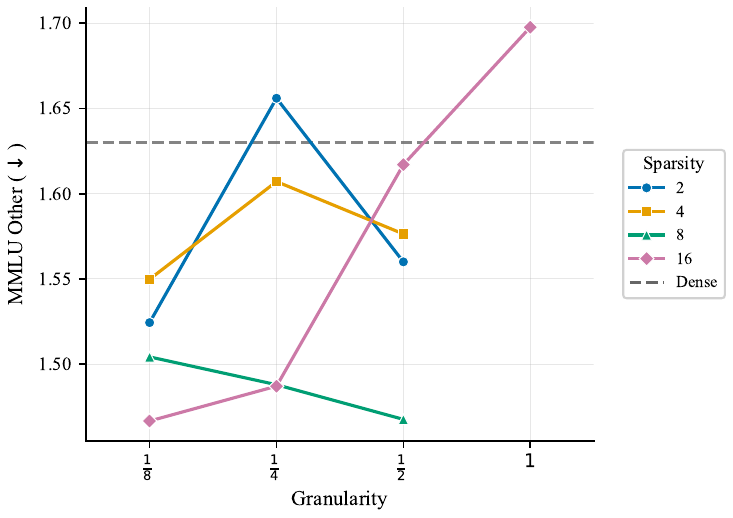}
        \end{subfigure}
        \caption{200M active, 200M - 3.3B total parameters}
    \end{subfigure}
    \par\bigskip\bigskip
        \begin{subfigure}[t]{\textwidth}
        \begin{subfigure}[t]{0.33\textwidth}
            \centering
            \includegraphics[width=\linewidth]{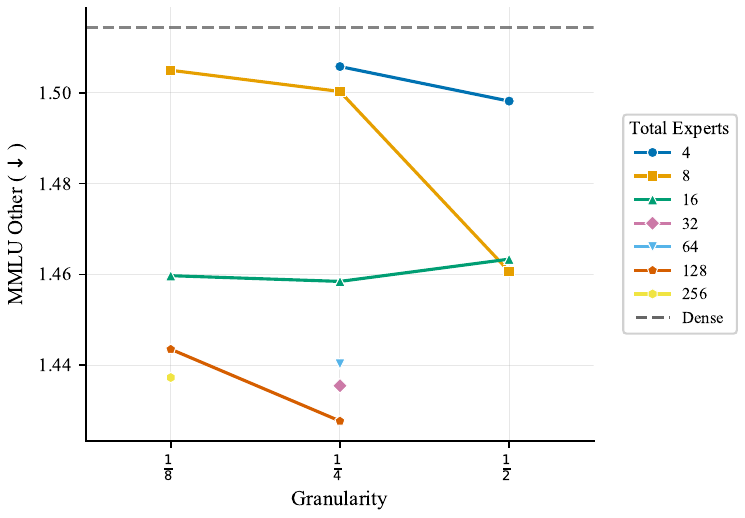}
        \end{subfigure}
        \begin{subfigure}[t]{0.33\textwidth}
            \centering
            \includegraphics[width=\linewidth]{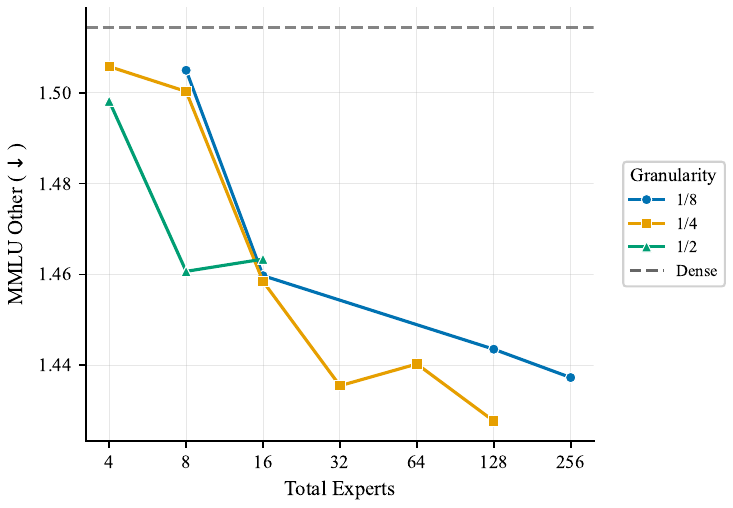}
        \end{subfigure}
        \begin{subfigure}[t]{0.33\textwidth}
            \centering
            \includegraphics[width=\linewidth]{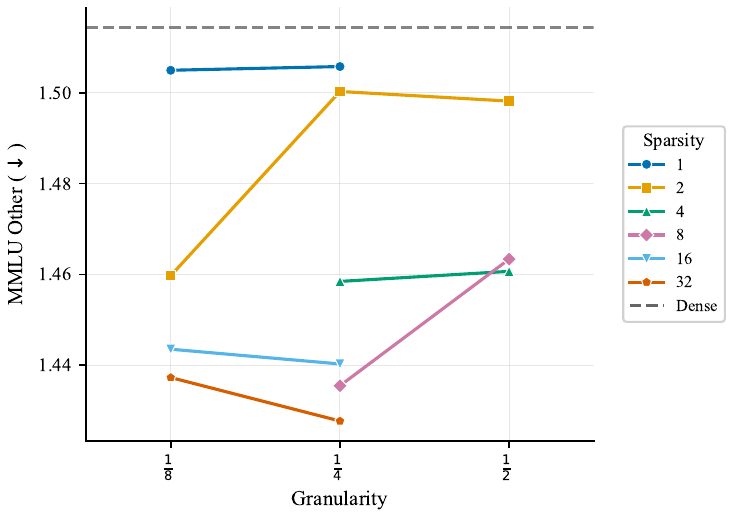}
        \end{subfigure}
        \caption{300M active, 300M - 6.6B total parameters}
    \end{subfigure}

    \caption{
    \textbf{Increasing inactive expert parameters via expert size (left) or total count (center) improves performance in MoEs (\S\ref{sec:expt_main}).} This effect is seen both when holding total number of experts fixed (left) and when holding expert granularity fixed (center). In general, increasing total parameters results in improved performance.  \textbf{Optimal tradeoff between expert count and granularity varies in MoEs (right). (\S\ref{sec:expt_main})}
    At each activation sparsity $s$ (equivalently, at each total parameter count), the optimal (total expert count, expert granularity) configuration varies. As $s$ increases, optimal expert granularity remains nearly fixed, suggesting that sparsity should be scaled up primarily by increasing total expert count $n$, while maintaining a near constant, slowly increasing expert granularity $g$. 
    }
    \label{fig:mmlu_other_experts}
\end{figure*}

\begin{figure*}[!ht]
    \centering
    
    \begin{subfigure}[t]{0.46\textwidth}
        \centering
        \includegraphics[width=\linewidth]{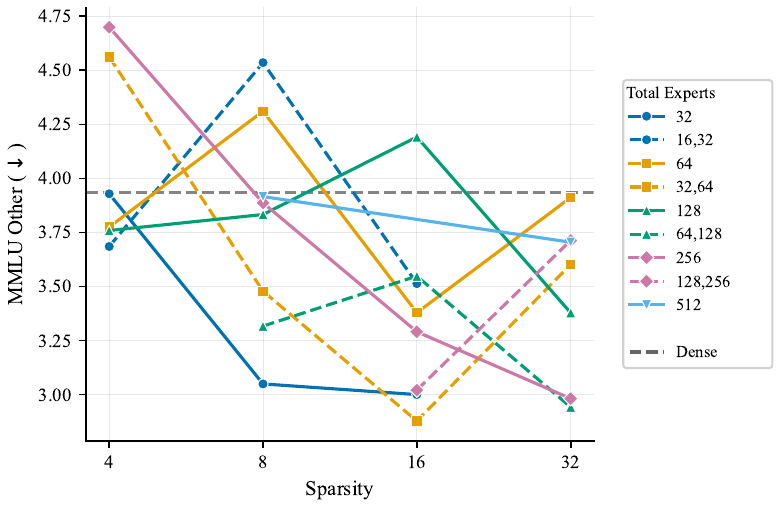}
        \caption{50M active, 50M - 930M total parameters}
    \end{subfigure}
    \vspace{1em}
    \begin{subfigure}[t]{0.46\textwidth}
        \centering
        \includegraphics[width=\linewidth]{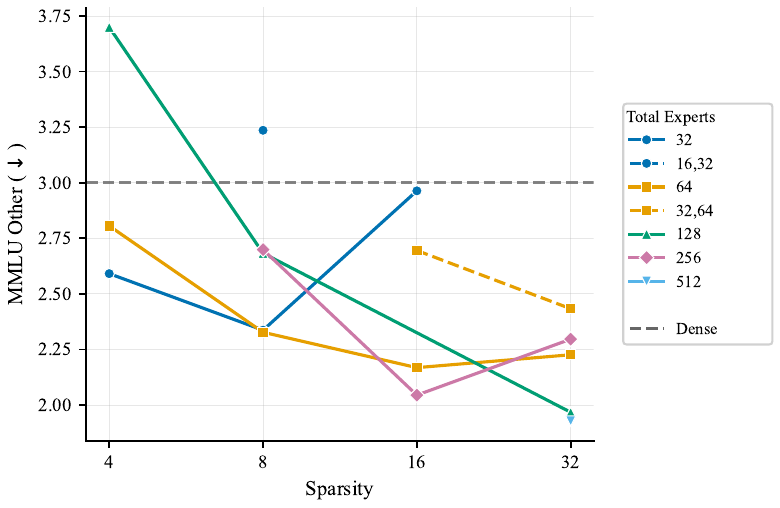}
        \caption{80M active, 80M - 765M total parameters}
    \end{subfigure}
    \caption{
    \textbf{Heterogeneity of expert size alone does not improve MoE performance (\S\ref{sec:expt_hetgen}).} To explore the potential benefits of their architectural flexibility, we compare heterogeneous MoEs (indicated by dotted lines) to active- and total-parameter-matched homogeneous MoEs. Heterogeneity alone does not result in performance gains, as, at each activation sparsity $s$, heterogeneous MoEs with $n_1, n_2 = a, b$ lie between or near the 2 closest homogeneous MoEs, with $n=a$ and with $n=b$.
    }
    \label{fig:mmlu_other_het}
\end{figure*}

\begin{figure*}[!ht]
    \centering
    
    \begin{subfigure}[t]{1.0\textwidth}
        \centering
        \includegraphics[width=\linewidth]{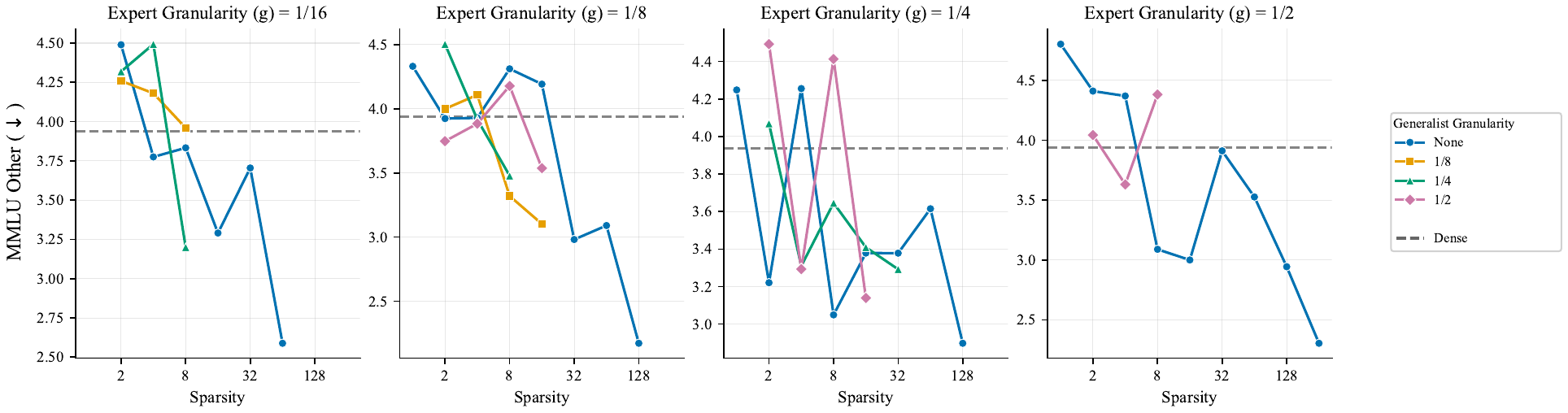}
        \caption{50M active, 50M - 930M total parameters}
    \end{subfigure}
    \par\bigskip\bigskip
    \begin{subfigure}[t]{1.0\textwidth}
        \centering
        \includegraphics[width=\linewidth]{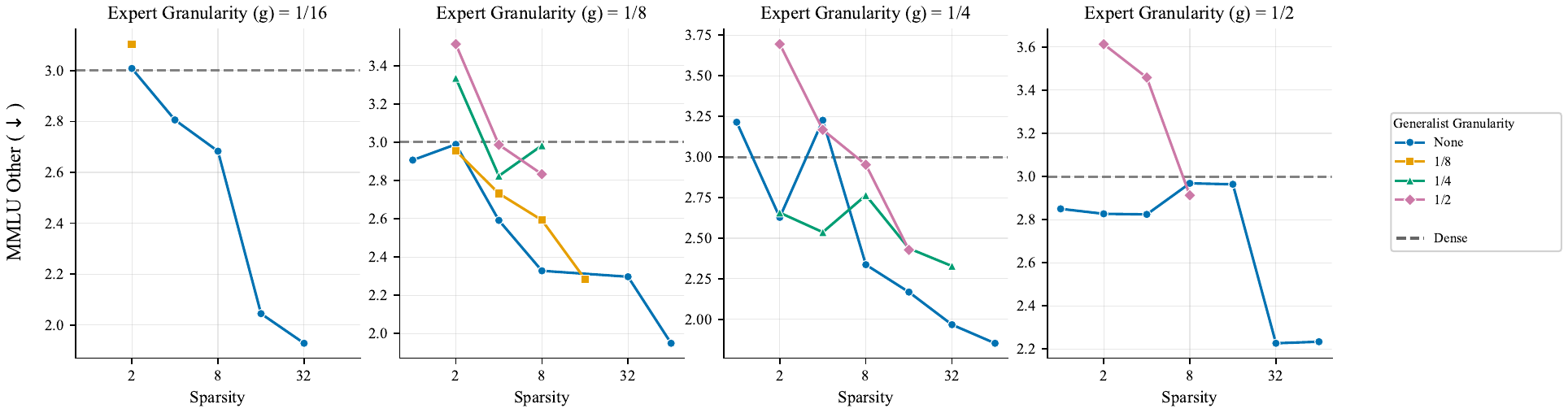}
        \caption{80M active, 80M - 765M total parameters}
    \end{subfigure}
    \par\bigskip\bigskip
    \begin{subfigure}[t]{1.0\textwidth}
        \centering
        \includegraphics[width=\linewidth]{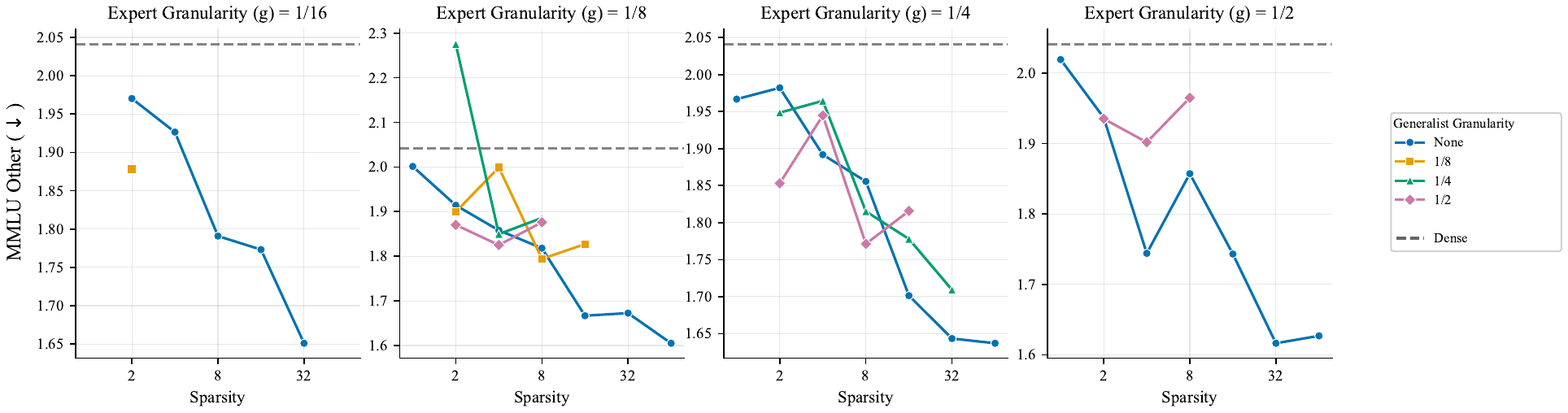}
        \caption{110M active, 110M - 1.4B total parameters}
    \end{subfigure}
    \caption{
    \textbf{The inclusion of a generalist consistently degrades performance in homogeneous MoEs (\S\ref{sec:expt_hetgen}).}
    We train MoE LMs which consist of some routed experts with granularity $g$, as well as a generalist with granularity $g_{gen}\in \{\frac{1}{2}, \frac{1}{4}, \frac{1}{8}\} $. We compare to settings with no generalist, only routed experts with granularity $g$. In all settings and configurations, the addition of any granularity generalist results in comparable or degraded performance. 
    }
    \label{fig:mmlu_other_gen}
\end{figure*}

\begin{figure*}[ht]
    \centering
    \begin{subfigure}[t]{1.0\textwidth}
        \centering
        \includegraphics[width=\linewidth]{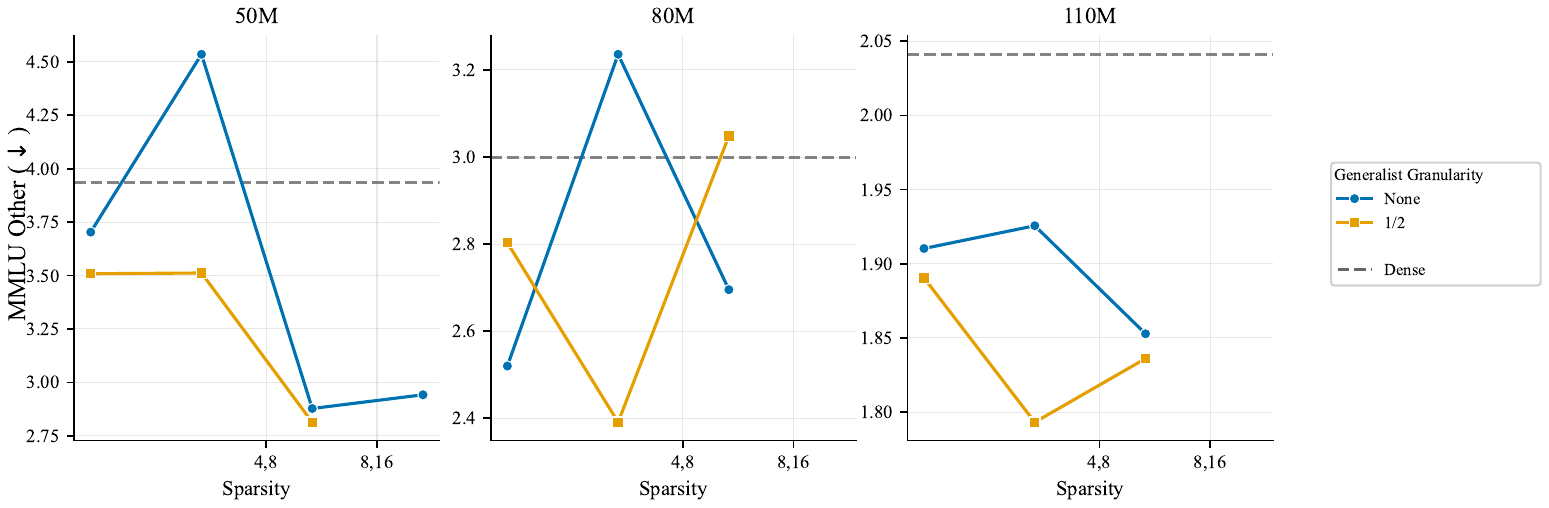}
    \end{subfigure}
    \caption{
    \textbf{The inclusion of a generalist consistently degrades performance in heterogeneous MoEs (\S\ref{sec:expt_hetgen}).}
    We train heterogeneous MoE LMs which consist of  routed experts with granularity $g_1, g_2$, as well as a generalist with granularity $g_{gen} = \frac{1}{2}$. We compare to settings with no generalist. In all settings and configurations, the addition of a generalist results in comparable or degraded performance. 
    }
    \label{fig:mmlu_other_hetgen}
\end{figure*}

\begin{figure*}[ht]
    \centering
    \begin{subfigure}[t]{\textwidth}
        \centering
        \begin{subfigure}[t]{0.45\textwidth}
            \includegraphics[width=\linewidth]{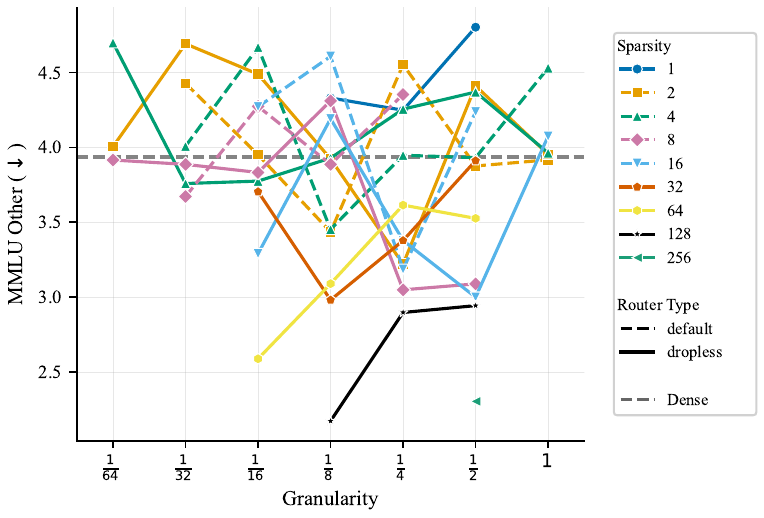}
            \caption{50M active, 50M - 930M total parameters}
        \end{subfigure}
    \hspace{1em}
        \begin{subfigure}[t]{0.45\textwidth}
            \centering
            \includegraphics[width=\linewidth]{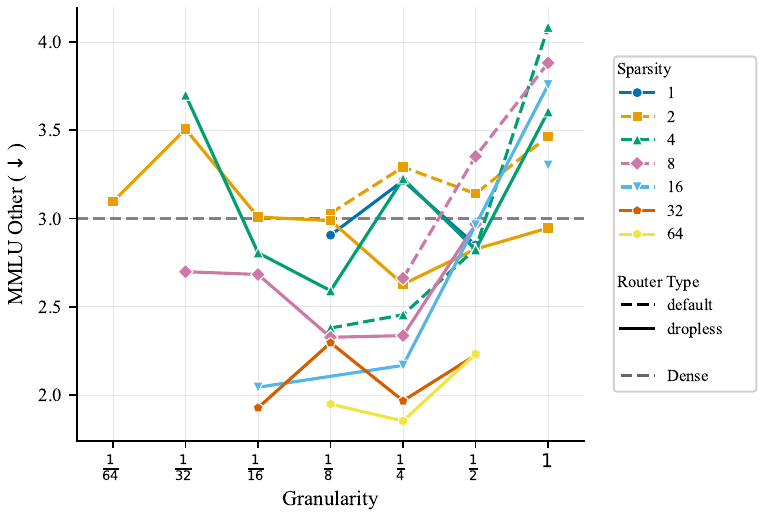}
            \caption{80M active, 80M - 765M total parameters}
        \end{subfigure}
    \end{subfigure}

    \par\bigskip\bigskip
    \begin{subfigure}[t]{0.45\textwidth}
        \centering
        \includegraphics[width=\linewidth]{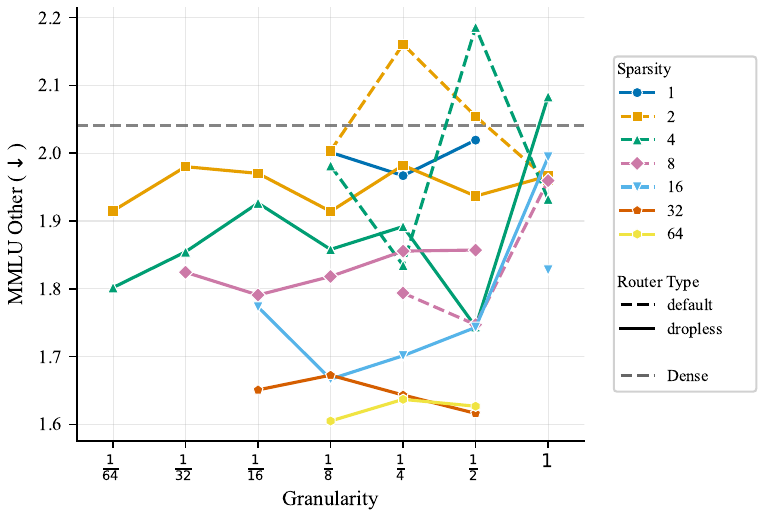}
        \caption{110M active, 110M - 1.4B total parameters}
    \end{subfigure}
    \caption{ 
    \textbf{Dropless routing outperforms default routing (\S\ref{sec:expt_router}).}
    We compare dropless routing to the default setting, which allow tokens to be dropped. Across all scales, we find that dropless routing outperforms or performs comparably to default routing. 
    }
    \label{fig:mmlu_other_dropless}
\end{figure*}

\begin{figure*}[ht]
    \centering
    \begin{subfigure}[t]{0.45\textwidth}
        \centering
        \includegraphics[width=\linewidth]{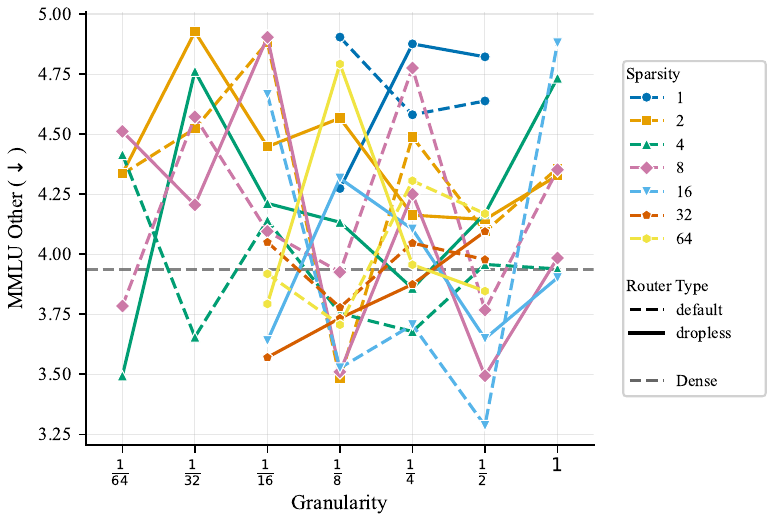}
        \caption{50M active, 50M - 930M total parameters}
    \end{subfigure}
    \hspace{1em}
    \begin{subfigure}[t]{0.45\textwidth}
        \centering
        \includegraphics[width=\linewidth]{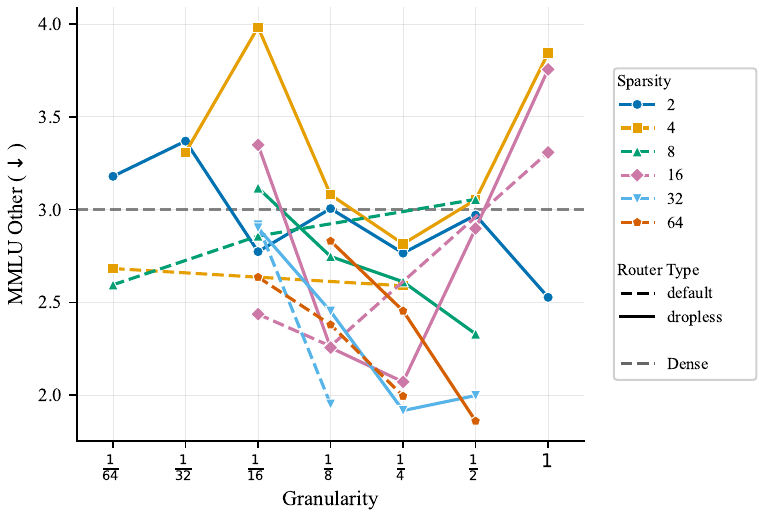}
        \caption{80M active, 80M - 765M total parameters}
    \end{subfigure}
    \caption{
    \textbf{Dropless routing, with bias $\gamma=\num{1e-3}$ (\S\ref{sec:expt_router}).} 
    As in Figure~\ref{fig:lm_avg_dropless}, we compare dropless routing to the default setting, which allow tokens to be dropped. Across all scales, we find that dropless routing outperforms or performs comparably to default routing. We see here with additional higher sparsity default routing runs that as sparsity increases, default routing performance approaches that of dropless routing.
    }
    \label{fig:mmlu_other_dropless_with_lf}
\end{figure*}

\begin{figure*}[ht]
    \centering
    \begin{subfigure}[]{\textwidth}
        \centering
        \includegraphics[width=0.46\linewidth]{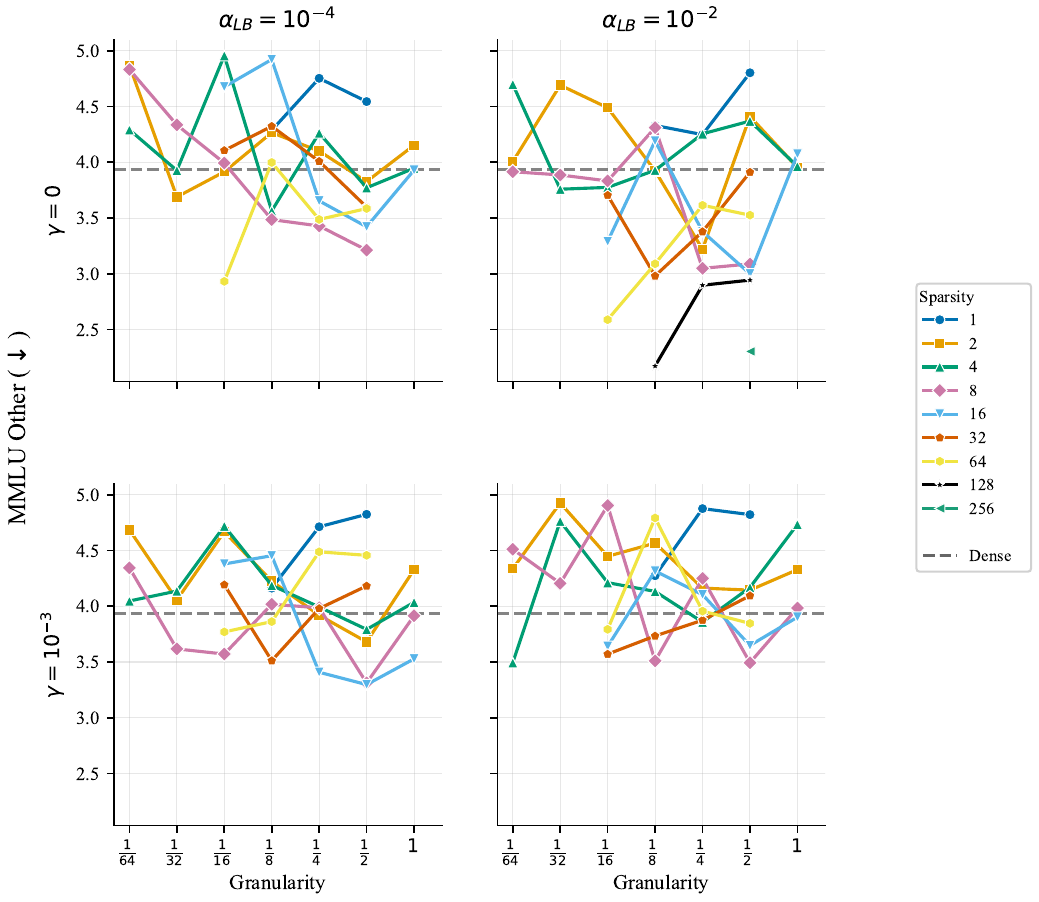}
        \hspace{1em}
        \includegraphics[width=0.46\linewidth]{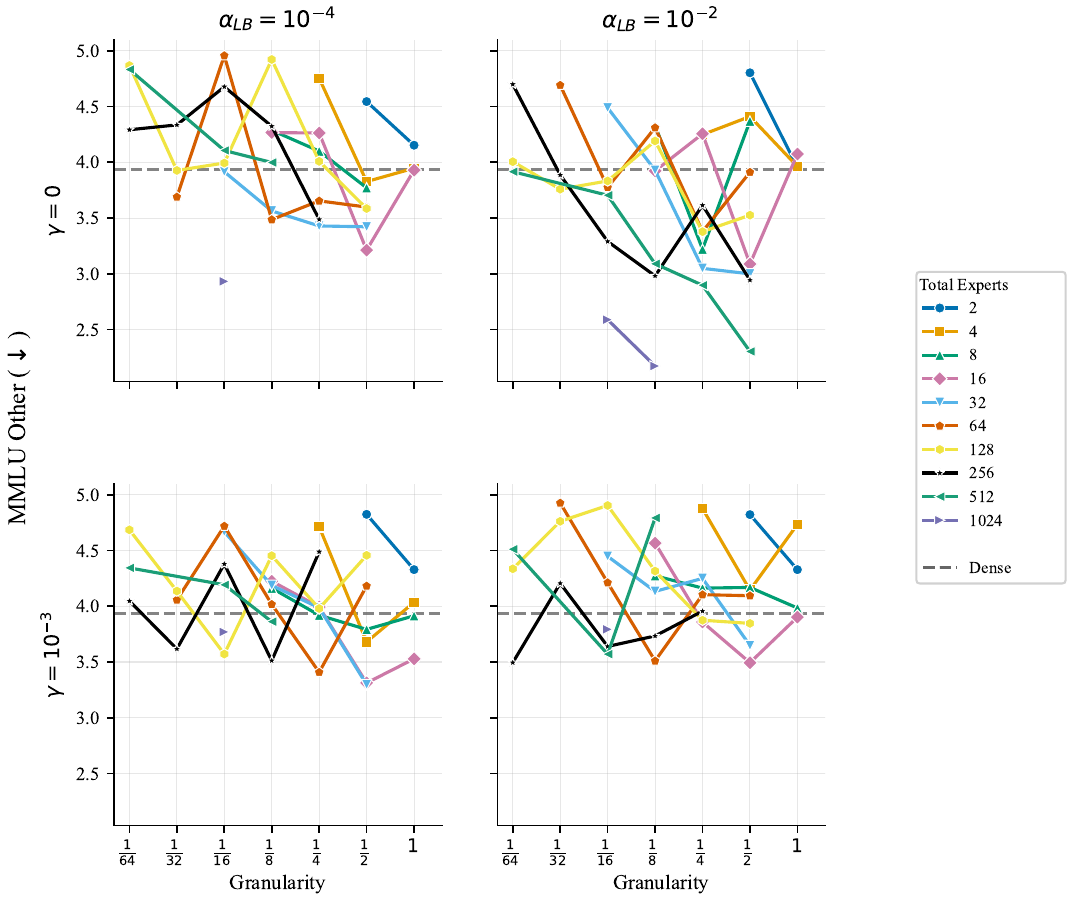}
        \caption{50M active, 50M - 930M total parameters}
    \end{subfigure}
    \par\bigskip\bigskip
    \begin{subfigure}[]{\textwidth}
        \centering
        \includegraphics[width=0.46\linewidth]{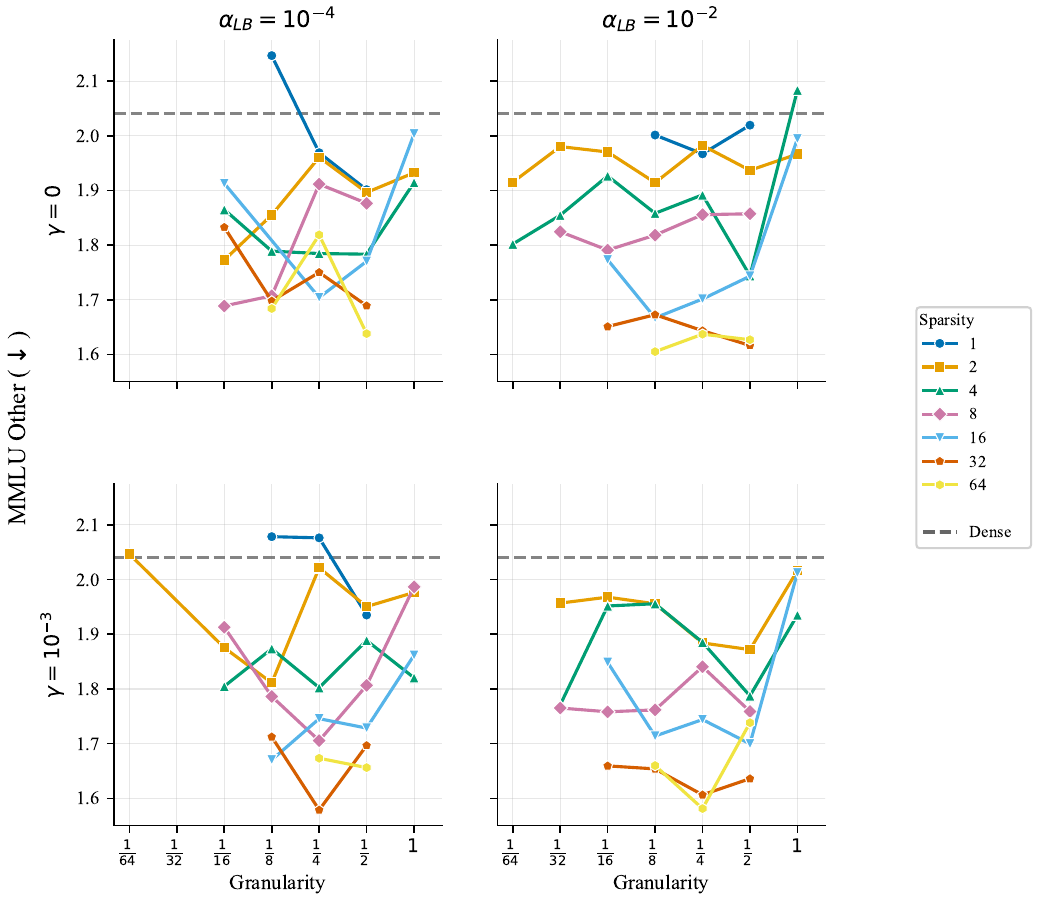}
        \hspace{1em}
        \includegraphics[width=0.46\linewidth]{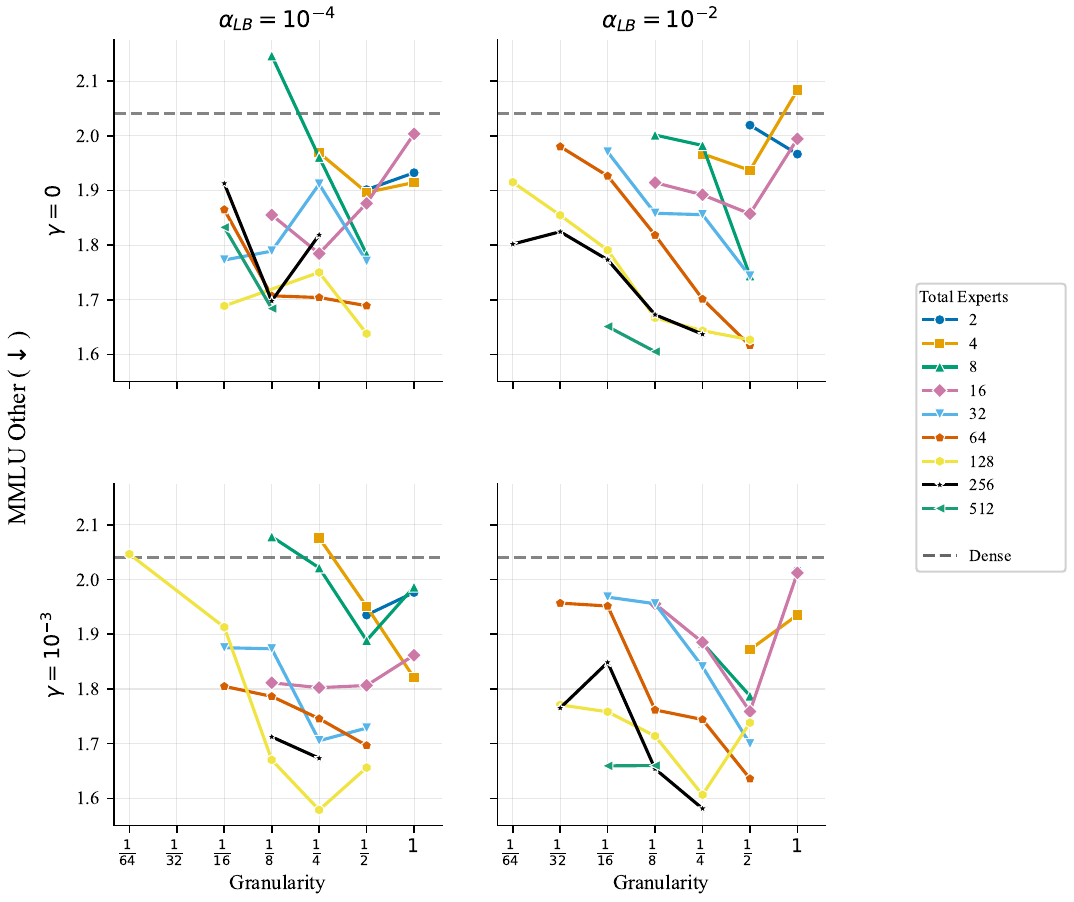}
        \caption{80M active, 80M - 765M total parameters}
    \end{subfigure}
    \par\bigskip\bigskip
    \begin{subfigure}[t]{\textwidth}
        \centering
        \includegraphics[width=0.46\linewidth]{figures/downstream/mmlu_other/ce_loss/lb_sweep_hgn_gxs_110M.pdf}
        \hspace{1em}
        \includegraphics[width=0.46\linewidth]{figures/downstream/mmlu_other/ce_loss/lb_sweep_hgn_gxn_110M.pdf}
        \caption{110M active, 110M - 1.4B total parameters}
    \end{subfigure}

    \end{figure*} 

\clearpage  

\begin{figure*}[ht]
    \addtocounter{figure}{-1}
    \centering
    \begin{subfigure}[t]{\textwidth}
        \centering
        \includegraphics[width=0.46\linewidth]{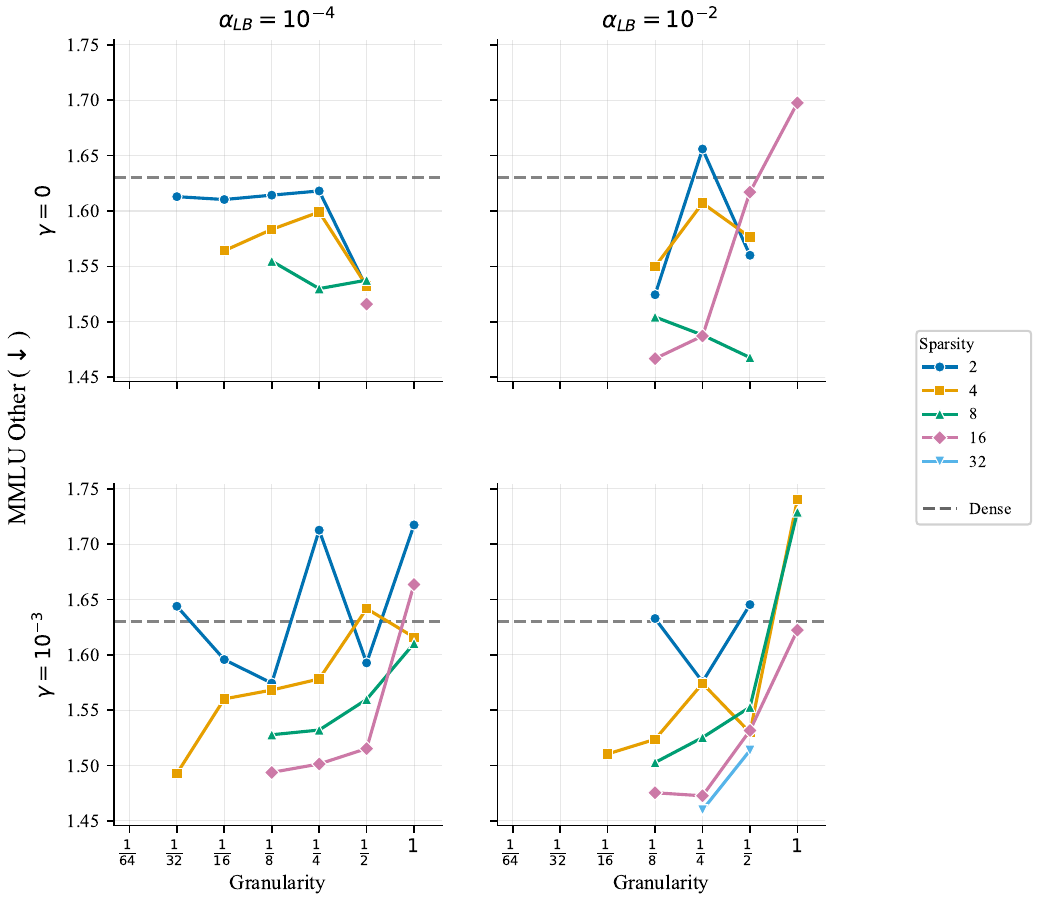}
        \hspace{1em}
        \includegraphics[width=0.46\linewidth]{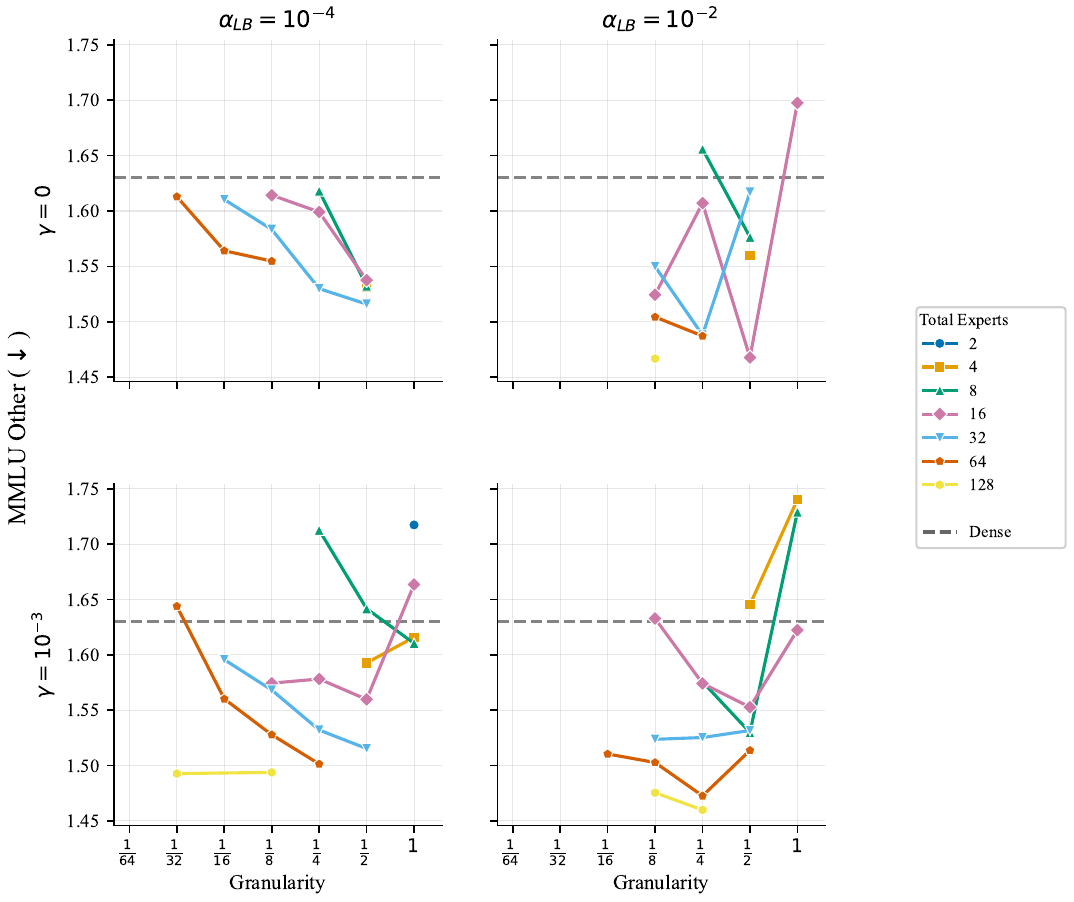}
        \caption{200M active, 200M - 3.3B total parameters}
    \end{subfigure}
    \par\bigskip\bigskip
    \begin{subfigure}[t]{\textwidth}
        \centering
        \includegraphics[width=0.3\linewidth]{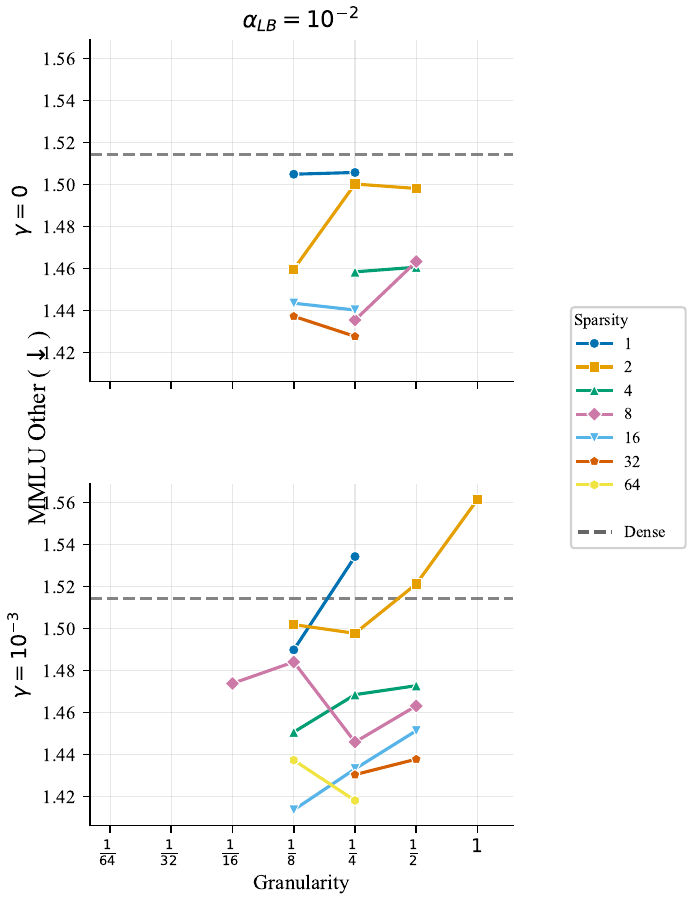}
        \hspace{1em}
        \includegraphics[width=0.3\linewidth]{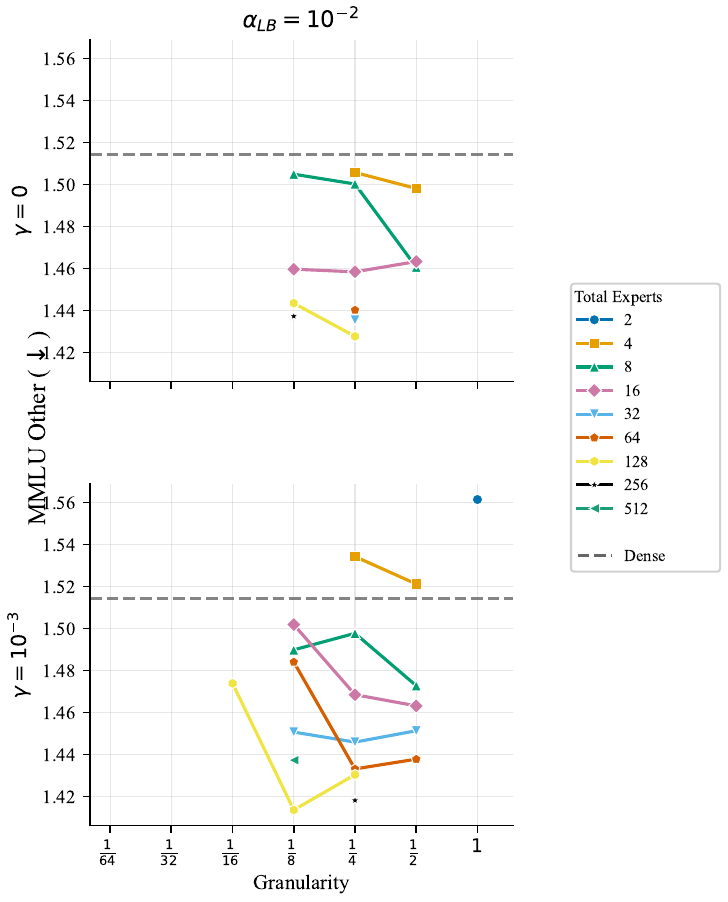}
        \caption{300M active, 300M - 6.6B total parameters}
    \end{subfigure}

    \caption{
    \textbf{Load balancing mechanisms must be tuned correctly (\S\ref{sec:expt_router}).}
    We consider load balancing loss weight $\alpha_{LB} \in \{\num{1e-2}, \num{1e-4}\}$ and loss-free load balancing with bias $\gamma\in\{0, \num{1e-3}\}$ ($\gamma=0$ indicates no loss-free mechanism). Results show that poorly chosen hyperparameters, such as high bias $\gamma = 1e-3$ with total experts $n\geq 512$, may impair performance. However, all settings other than $(\alpha_{LB}=\num{1e-2}, \gamma=\num{1e-3})$ perform comparably for $n \leq 512$, suggesting that a wide range of load balancing settings achieve near-optimal performance. 
    }
    \label{fig:mmlu_other_lb}
\end{figure*}

%% file: fig_tex/downstream/mmlu_social_sciences.tex
\begin{figure*}[!ht]
    \centering
        \begin{subfigure}[t]{\textwidth}
        \begin{subfigure}[t]{0.33\textwidth}
            \centering
            \caption*{\scriptsize Fixed total experts (n)}
            \includegraphics[width=\linewidth]{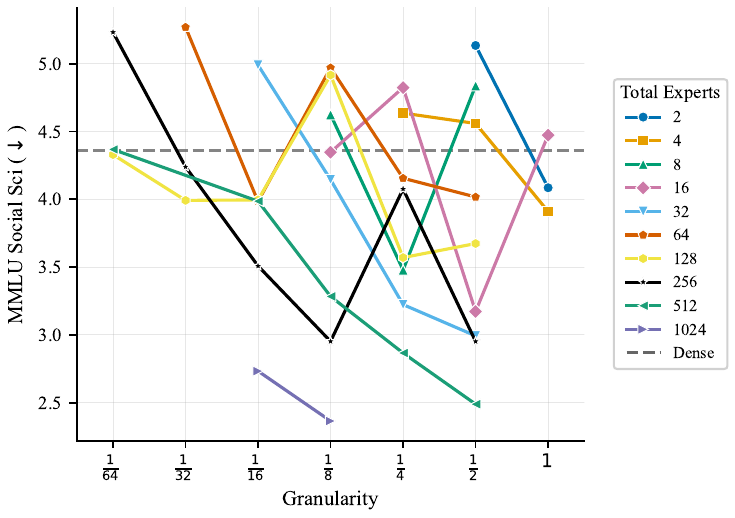}
        \end{subfigure}
        \begin{subfigure}[t]{0.33\textwidth}
            \centering
            \caption*{\scriptsize Fixed granularity (g)}
            \includegraphics[width=\linewidth]{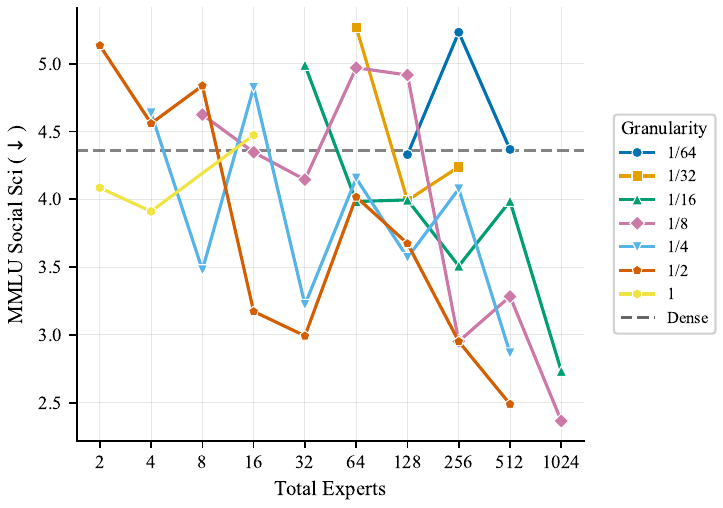}
        \end{subfigure}
        \begin{subfigure}[t]{0.33\textwidth}
            \centering
            \caption*{\scriptsize Fixed activation sparsity (s)}
            \includegraphics[width=\linewidth]{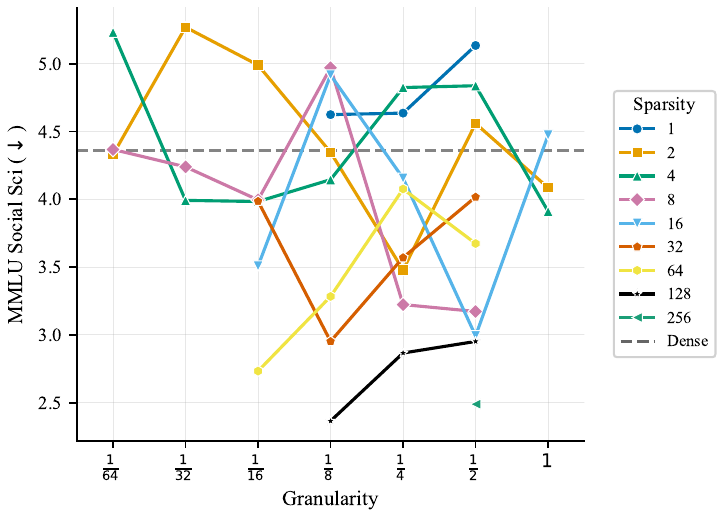}
        \end{subfigure}
        \caption{50M active, 50M - 930M total parameters}
    \end{subfigure}
\par\bigskip\bigskip
    \begin{subfigure}[t]{\textwidth}
        \begin{subfigure}[t]{0.33\textwidth}
            \centering
            \includegraphics[width=\linewidth]{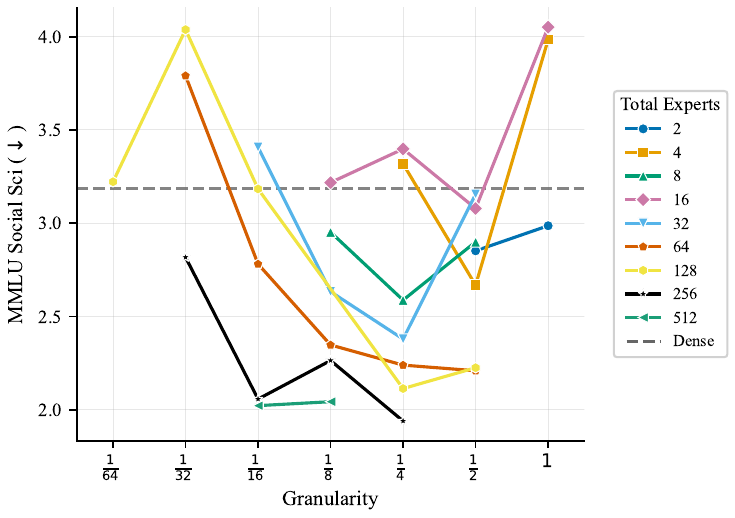}
        \end{subfigure}
        \begin{subfigure}[t]{0.33\textwidth}
            \centering
            \includegraphics[width=\linewidth]{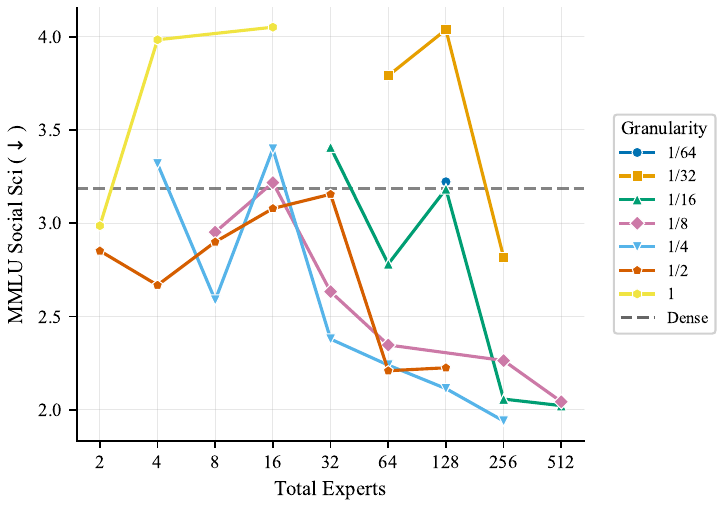}
        \end{subfigure}
        \begin{subfigure}[t]{0.33\textwidth}
            \centering
            \includegraphics[width=\linewidth]{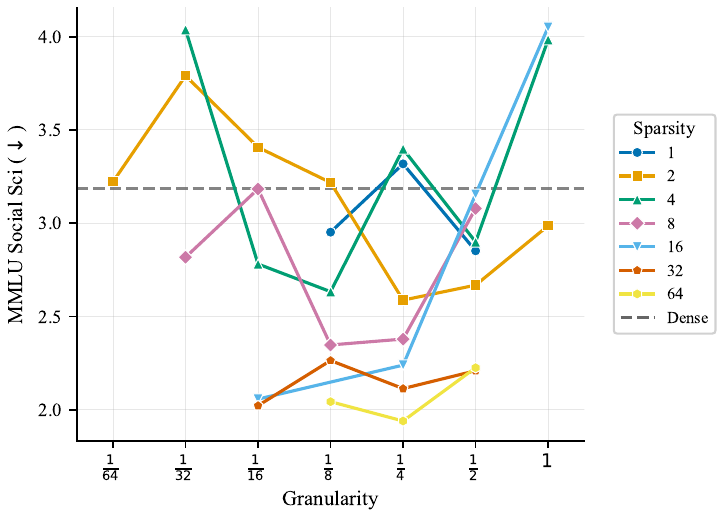}
        \end{subfigure}
        \caption{80M active, 80M - 765M total parameters}
    \end{subfigure}
    \par\bigskip\bigskip
        \begin{subfigure}[t]{\textwidth}
        \begin{subfigure}[t]{0.33\textwidth}
            \centering
            \includegraphics[width=\linewidth]{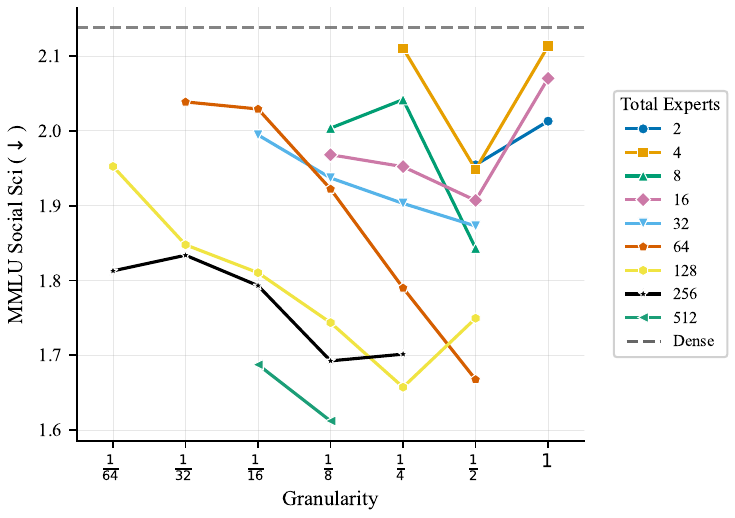}
        \end{subfigure}
        \begin{subfigure}[t]{0.33\textwidth}
            \centering
            \includegraphics[width=\linewidth]{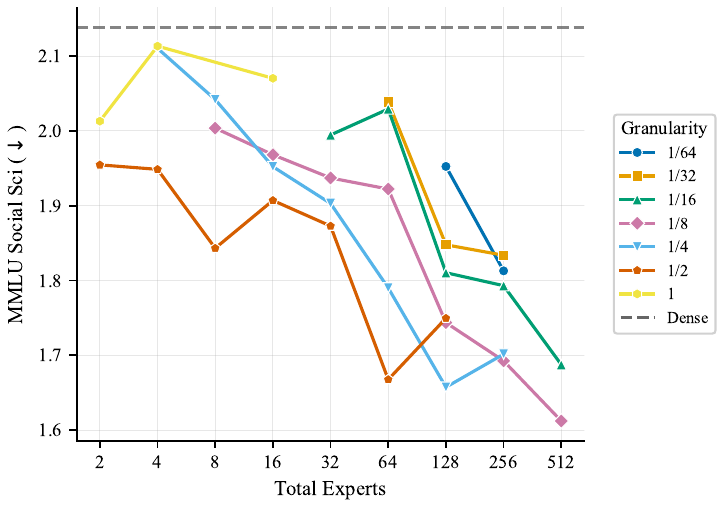}
        \end{subfigure}
        \begin{subfigure}[t]{0.33\textwidth}
            \centering
            \includegraphics[width=\linewidth]{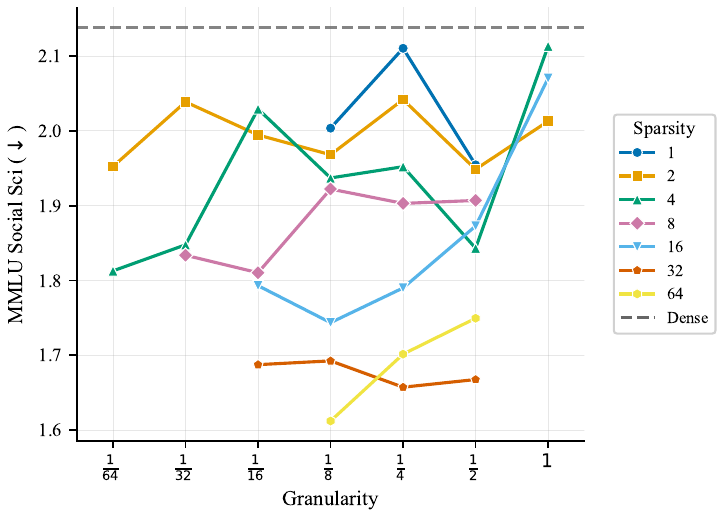}
        \end{subfigure}
        \caption{110M active, 110M - 1.4B total parameters}
    \end{subfigure}
    \end{figure*}

\clearpage  

\begin{figure*}[!ht]
        \addtocounter{figure}{-1}
    \begin{subfigure}[t]{\textwidth}
        \addtocounter{subfigure}{3}
        \begin{subfigure}[t]{0.33\textwidth}
            \centering
            \caption*{\scriptsize Fixed total experts (n)}
            \includegraphics[width=\linewidth]{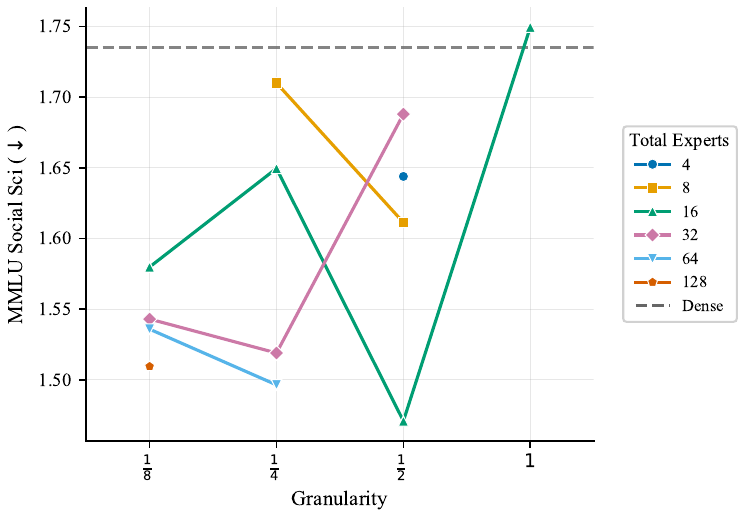}
        \end{subfigure}
        \begin{subfigure}[t]{0.33\textwidth}
            \centering
            \caption*{\scriptsize Fixed granularity (g)}
            \includegraphics[width=\linewidth]{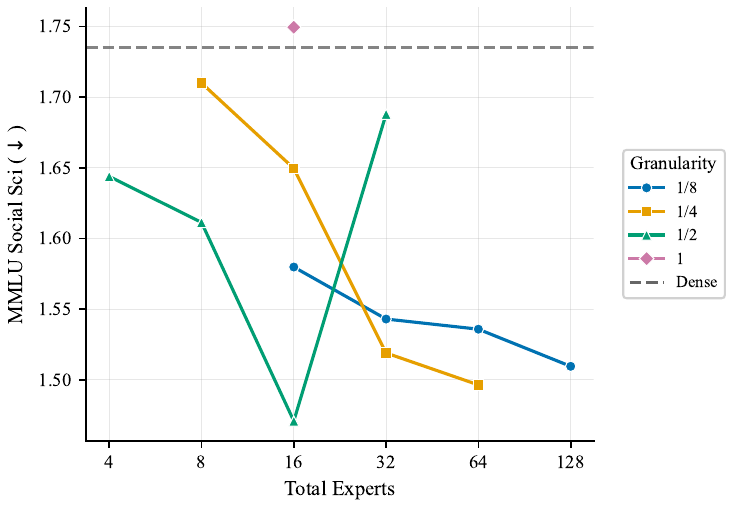}
        \end{subfigure}
        \begin{subfigure}[t]{0.33\textwidth}
            \centering
            \caption*{\scriptsize Fixed activation sparsity (s)}
            \includegraphics[width=\linewidth]{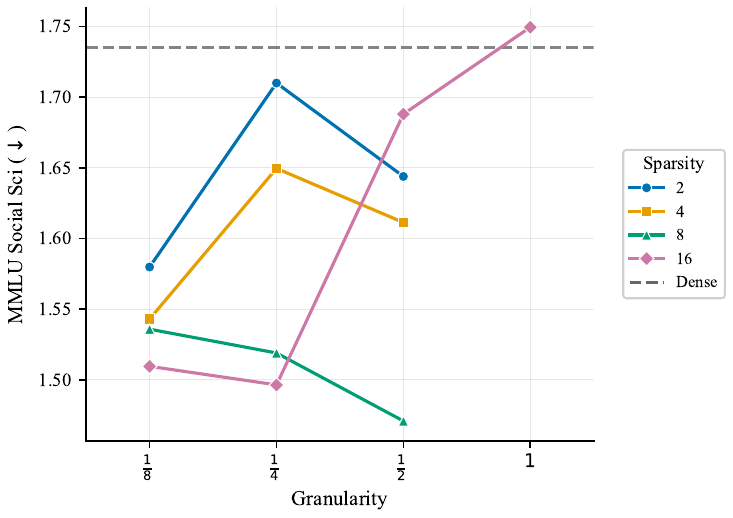}
        \end{subfigure}
        \caption{200M active, 200M - 3.3B total parameters}
    \end{subfigure}
    \par\bigskip\bigskip
        \begin{subfigure}[t]{\textwidth}
        \begin{subfigure}[t]{0.33\textwidth}
            \centering
            \includegraphics[width=\linewidth]{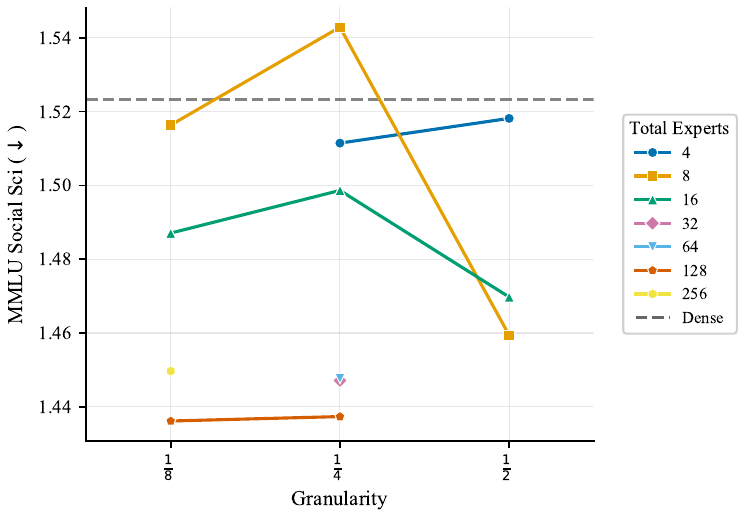}
        \end{subfigure}
        \begin{subfigure}[t]{0.33\textwidth}
            \centering
            \includegraphics[width=\linewidth]{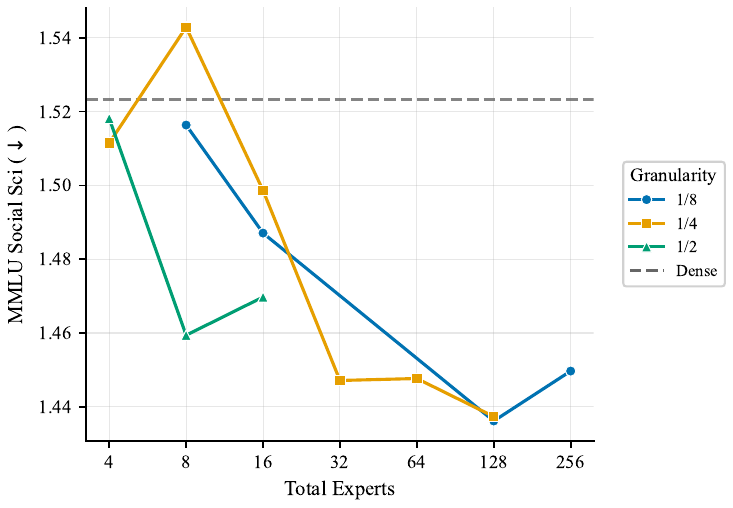}
        \end{subfigure}
        \begin{subfigure}[t]{0.33\textwidth}
            \centering
            \includegraphics[width=\linewidth]{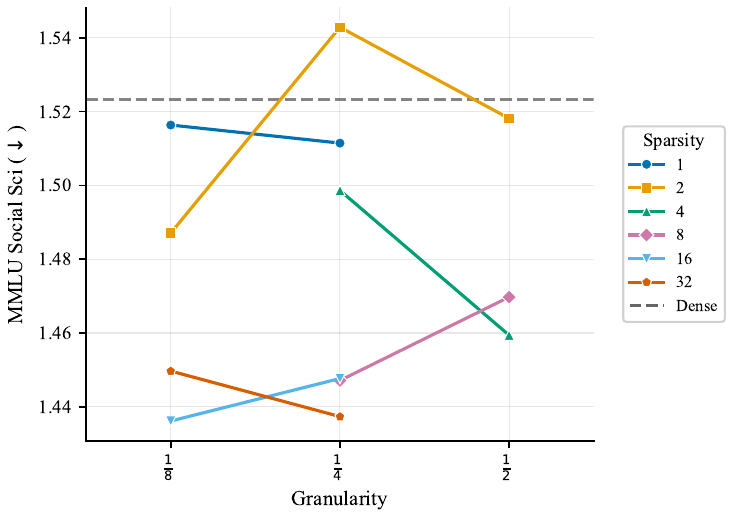}
        \end{subfigure}
        \caption{300M active, 300M - 6.6B total parameters}
    \end{subfigure}

    \caption{
    \textbf{Increasing inactive expert parameters via expert size (left) or total count (center) improves performance in MoEs (\S\ref{sec:expt_main}).} This effect is seen both when holding total number of experts fixed (left) and when holding expert granularity fixed (center). In general, increasing total parameters results in improved performance.  \textbf{Optimal tradeoff between expert count and granularity varies in MoEs (right). (\S\ref{sec:expt_main})}
    At each activation sparsity $s$ (equivalently, at each total parameter count), the optimal (total expert count, expert granularity) configuration varies. As $s$ increases, optimal expert granularity remains nearly fixed, suggesting that sparsity should be scaled up primarily by increasing total expert count $n$, while maintaining a near constant, slowly increasing expert granularity $g$. 
    }
    \label{fig:mmlu_social_sciences_experts}
\end{figure*}

\begin{figure*}[!ht]
    \centering
    
    \begin{subfigure}[t]{0.46\textwidth}
        \centering
        \includegraphics[width=\linewidth]{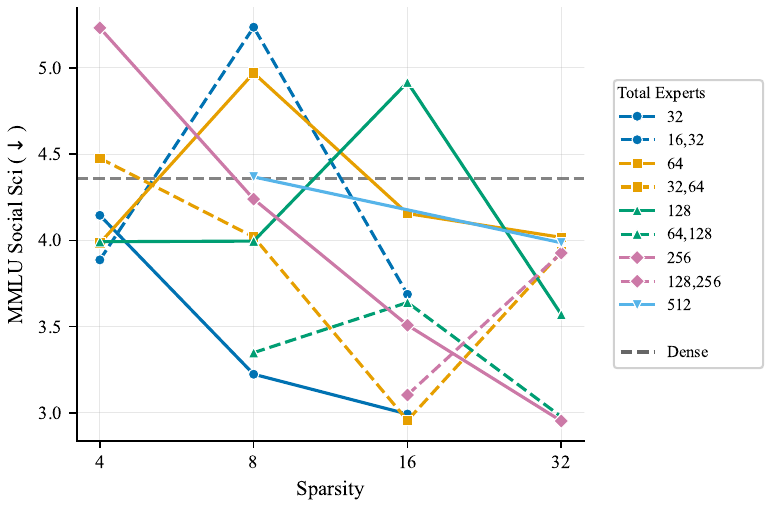}
        \caption{50M active, 50M - 930M total parameters}
    \end{subfigure}
    \vspace{1em}
    \begin{subfigure}[t]{0.46\textwidth}
        \centering
        \includegraphics[width=\linewidth]{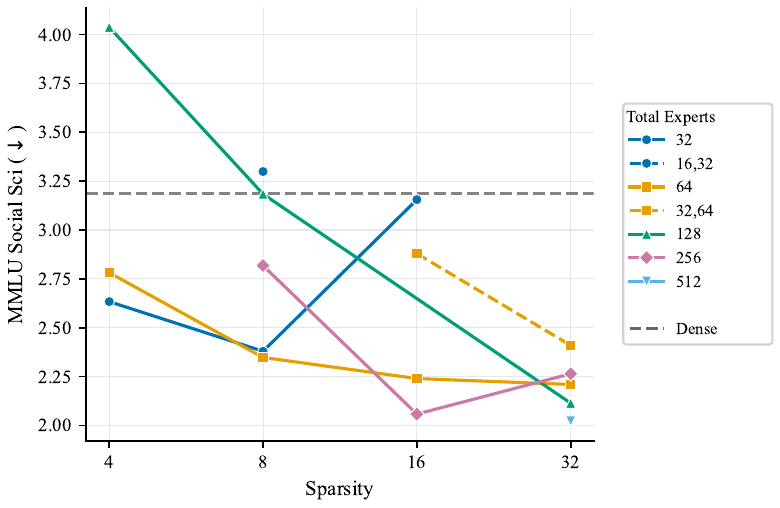}
        \caption{80M active, 80M - 765M total parameters}
    \end{subfigure}
    \caption{
    \textbf{Heterogeneity of expert size alone does not improve MoE performance (\S\ref{sec:expt_hetgen}).} To explore the potential benefits of their architectural flexibility, we compare heterogeneous MoEs (indicated by dotted lines) to active- and total-parameter-matched homogeneous MoEs. Heterogeneity alone does not result in performance gains, as, at each activation sparsity $s$, heterogeneous MoEs with $n_1, n_2 = a, b$ lie between or near the 2 closest homogeneous MoEs, with $n=a$ and with $n=b$.
    }
    \label{fig:mmlu_social_sciences_het}
\end{figure*}

\begin{figure*}[!ht]
    \centering
    
    \begin{subfigure}[t]{1.0\textwidth}
        \centering
        \includegraphics[width=\linewidth]{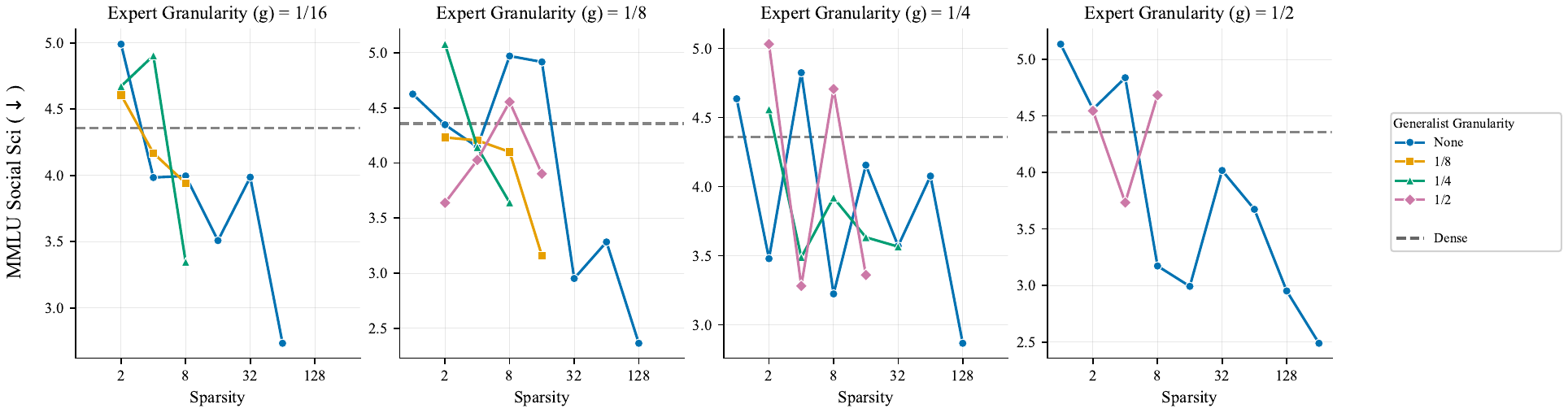}
        \caption{50M active, 50M - 930M total parameters}
    \end{subfigure}
    \par\bigskip\bigskip
    \begin{subfigure}[t]{1.0\textwidth}
        \centering
        \includegraphics[width=\linewidth]{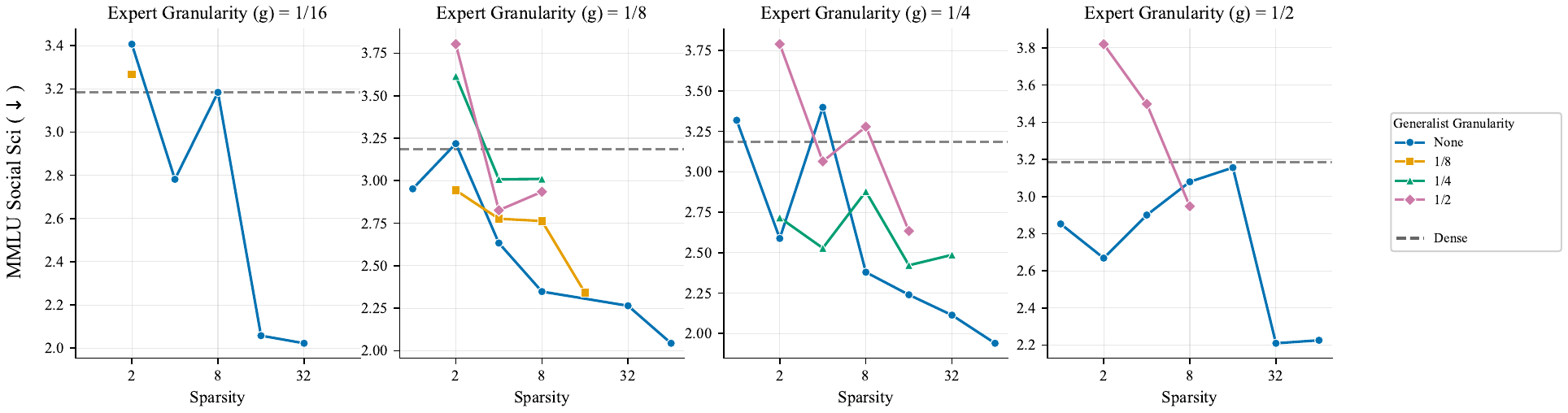}
        \caption{80M active, 80M - 765M total parameters}
    \end{subfigure}
    \par\bigskip\bigskip
    \begin{subfigure}[t]{1.0\textwidth}
        \centering
        \includegraphics[width=\linewidth]{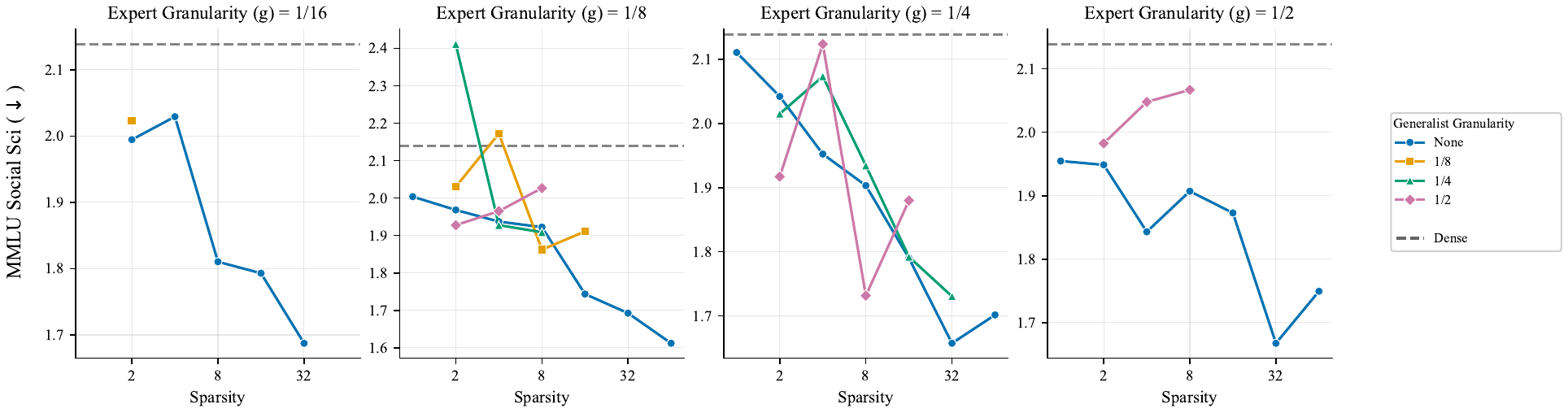}
        \caption{110M active, 110M - 1.4B total parameters}
    \end{subfigure}
    \caption{
    \textbf{The inclusion of a generalist consistently degrades performance in homogeneous MoEs (\S\ref{sec:expt_hetgen}).}
    We train MoE LMs which consist of some routed experts with granularity $g$, as well as a generalist with granularity $g_{gen}\in \{\frac{1}{2}, \frac{1}{4}, \frac{1}{8}\} $. We compare to settings with no generalist, only routed experts with granularity $g$. In all settings and configurations, the addition of any granularity generalist results in comparable or degraded performance. 
    }
    \label{fig:mmlu_social_sciences_gen}
\end{figure*}

\begin{figure*}[ht]
    \centering
    \begin{subfigure}[t]{1.0\textwidth}
        \centering
        \includegraphics[width=\linewidth]{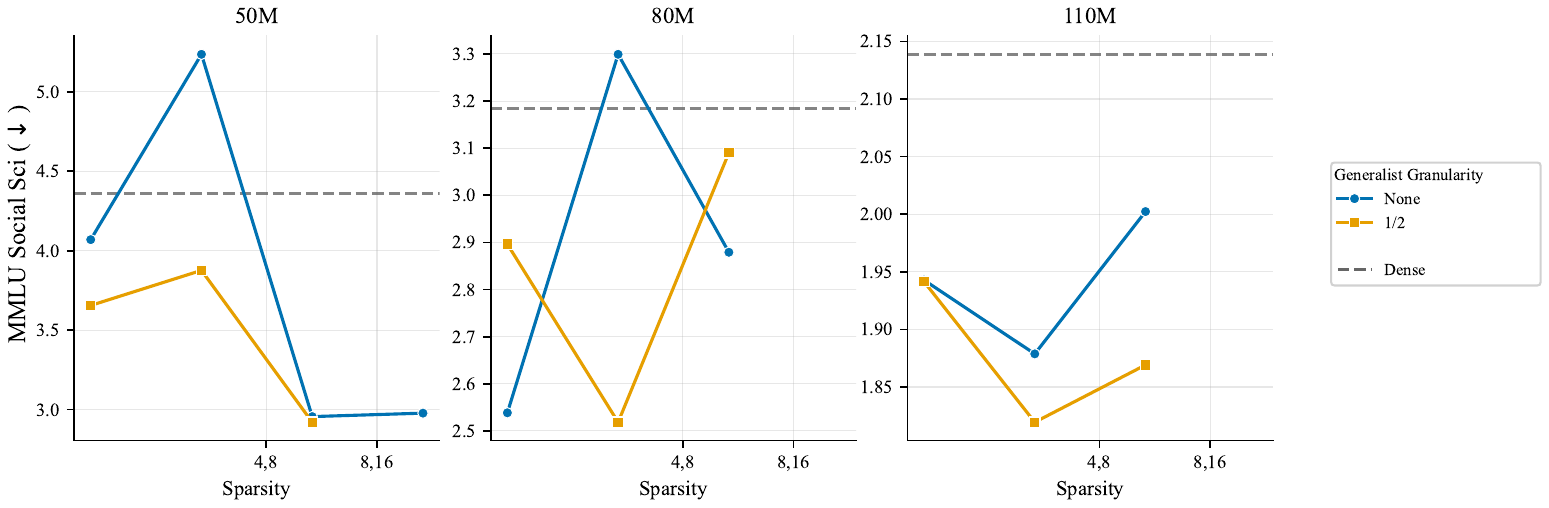}
    \end{subfigure}
    \caption{
    \textbf{The inclusion of a generalist consistently degrades performance in heterogeneous MoEs (\S\ref{sec:expt_hetgen}).}
    We train heterogeneous MoE LMs which consist of  routed experts with granularity $g_1, g_2$, as well as a generalist with granularity $g_{gen} = \frac{1}{2}$. We compare to settings with no generalist. In all settings and configurations, the addition of a generalist results in comparable or degraded performance. 
    }
    \label{fig:mmlu_social_sciences_hetgen}
\end{figure*}

\begin{figure*}[ht]
    \centering
    \begin{subfigure}[t]{\textwidth}
        \centering
        \begin{subfigure}[t]{0.45\textwidth}
            \includegraphics[width=\linewidth]{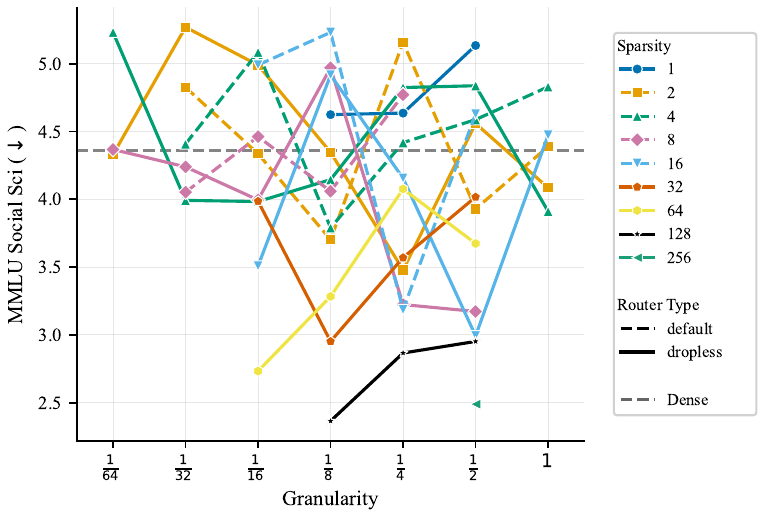}
            \caption{50M active, 50M - 930M total parameters}
        \end{subfigure}
    \hspace{1em}
        \begin{subfigure}[t]{0.45\textwidth}
            \centering
            \includegraphics[width=\linewidth]{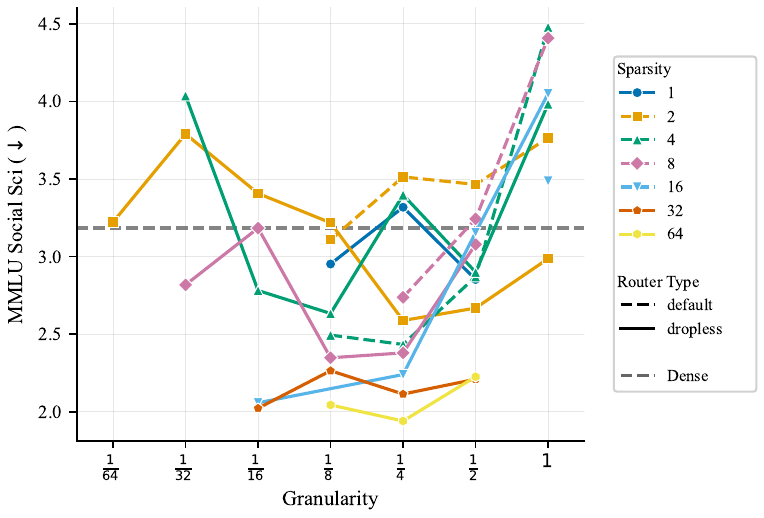}
            \caption{80M active, 80M - 765M total parameters}
        \end{subfigure}
    \end{subfigure}

    \par\bigskip\bigskip
    \begin{subfigure}[t]{0.45\textwidth}
        \centering
        \includegraphics[width=\linewidth]{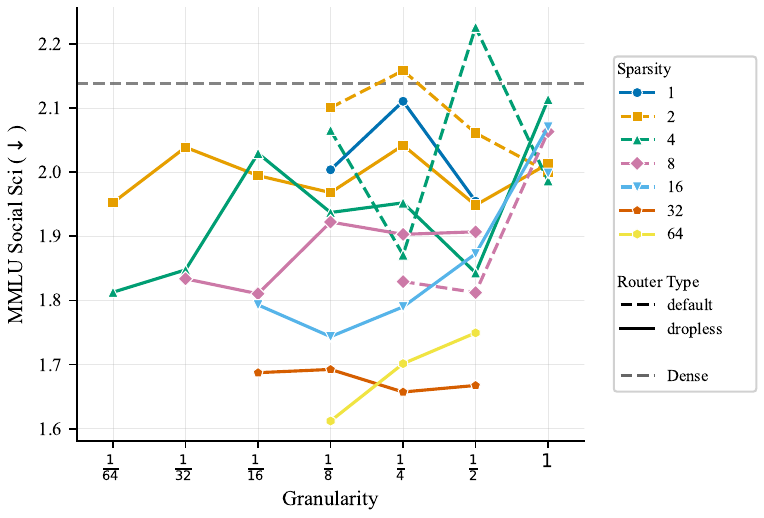}
        \caption{110M active, 110M - 1.4B total parameters}
    \end{subfigure}
    \caption{ 
    \textbf{Dropless routing outperforms default routing (\S\ref{sec:expt_router}).}
    We compare dropless routing to the default setting, which allow tokens to be dropped. Across all scales, we find that dropless routing outperforms or performs comparably to default routing. 
    }
    \label{fig:mmlu_social_sciences_dropless}
\end{figure*}

\begin{figure*}[ht]
    \centering
    \begin{subfigure}[t]{0.45\textwidth}
        \centering
        \includegraphics[width=\linewidth]{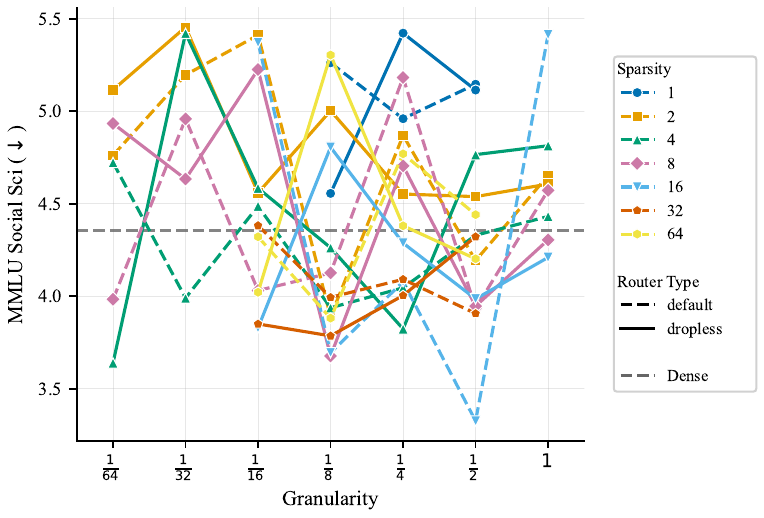}
        \caption{50M active, 50M - 930M total parameters}
    \end{subfigure}
    \hspace{1em}
    \begin{subfigure}[t]{0.45\textwidth}
        \centering
        \includegraphics[width=\linewidth]{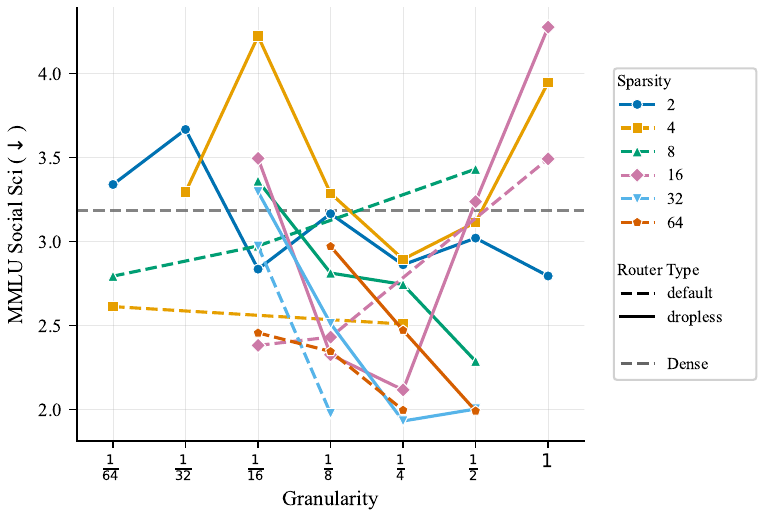}
        \caption{80M active, 80M - 765M total parameters}
    \end{subfigure}
    \caption{
    \textbf{Dropless routing, with bias $\gamma=\num{1e-3}$ (\S\ref{sec:expt_router}).} 
    As in Figure~\ref{fig:lm_avg_dropless}, we compare dropless routing to the default setting, which allow tokens to be dropped. Across all scales, we find that dropless routing outperforms or performs comparably to default routing. We see here with additional higher sparsity default routing runs that as sparsity increases, default routing performance approaches that of dropless routing.
    }
    \label{fig:mmlu_social_sciences_dropless_with_lf}
\end{figure*}

\begin{figure*}[ht]
    \centering
    \begin{subfigure}[]{\textwidth}
        \centering
        \includegraphics[width=0.46\linewidth]{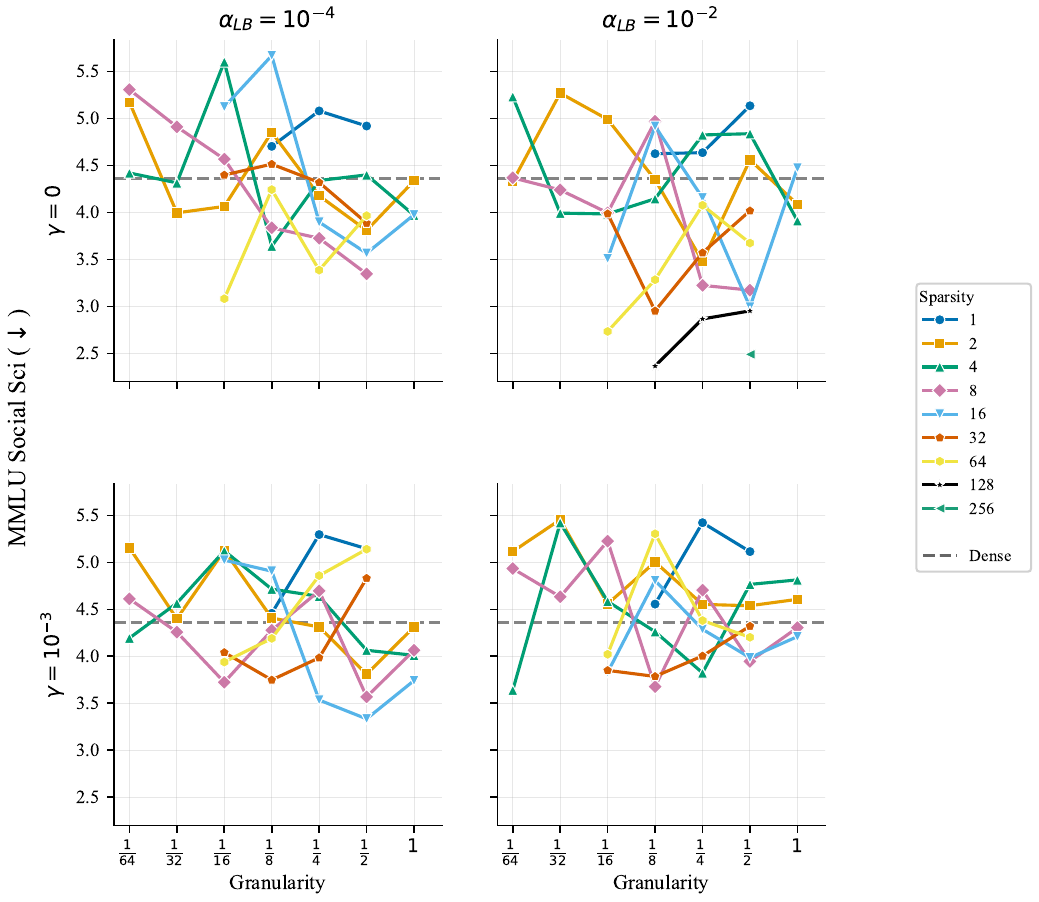}
        \hspace{1em}
        \includegraphics[width=0.46\linewidth]{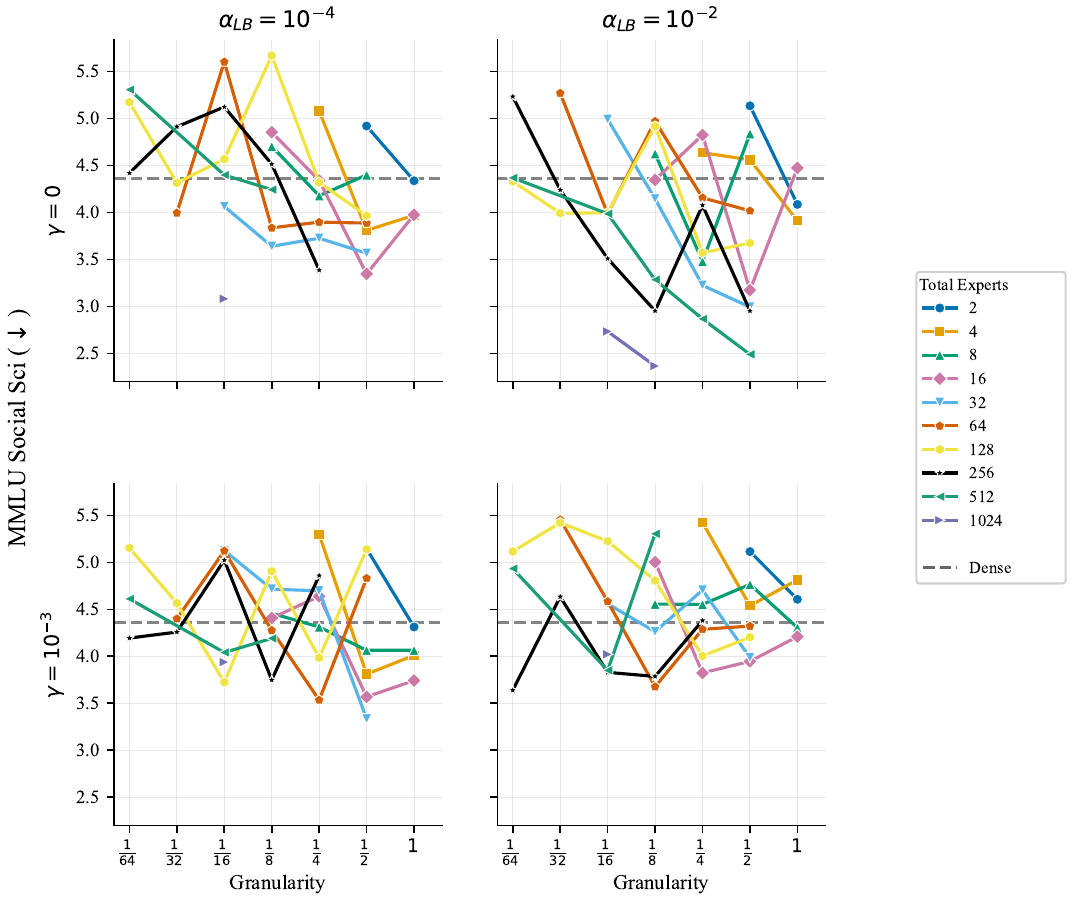}
        \caption{50M active, 50M - 930M total parameters}
    \end{subfigure}
    \par\bigskip\bigskip
    \begin{subfigure}[]{\textwidth}
        \centering
        \includegraphics[width=0.46\linewidth]{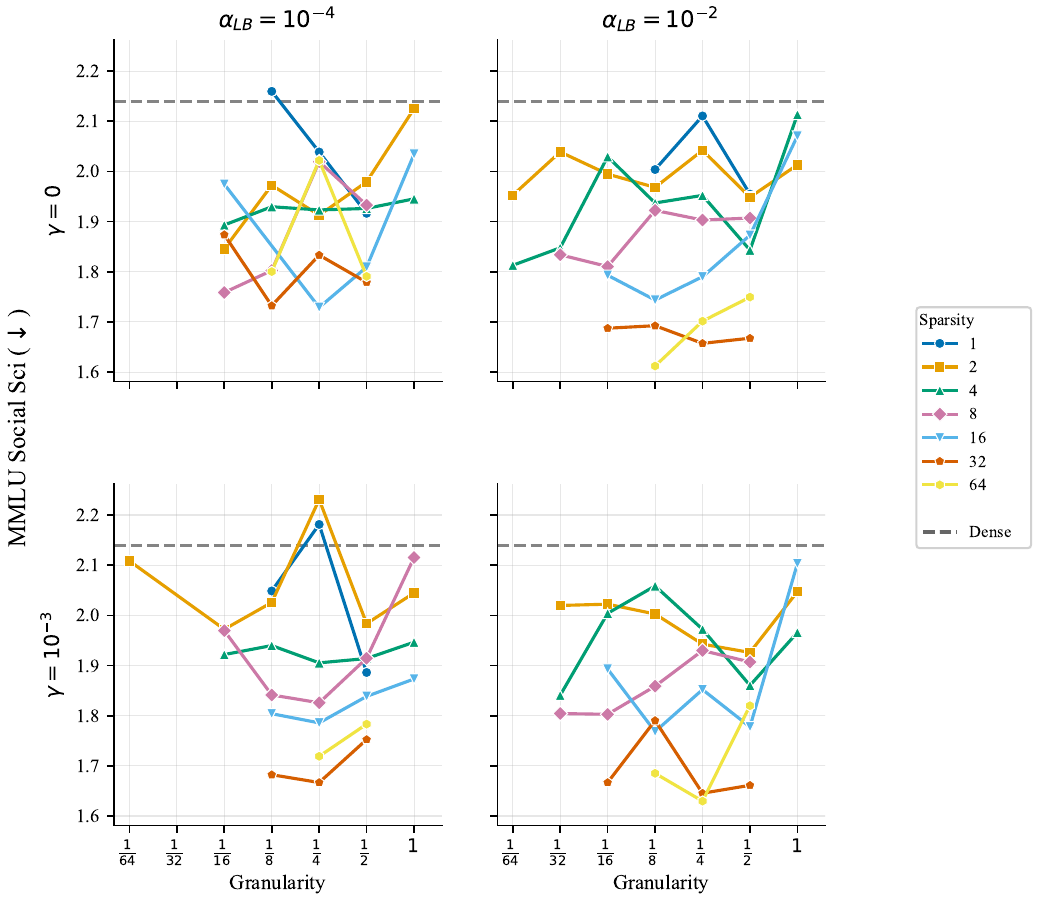}
        \hspace{1em}
        \includegraphics[width=0.46\linewidth]{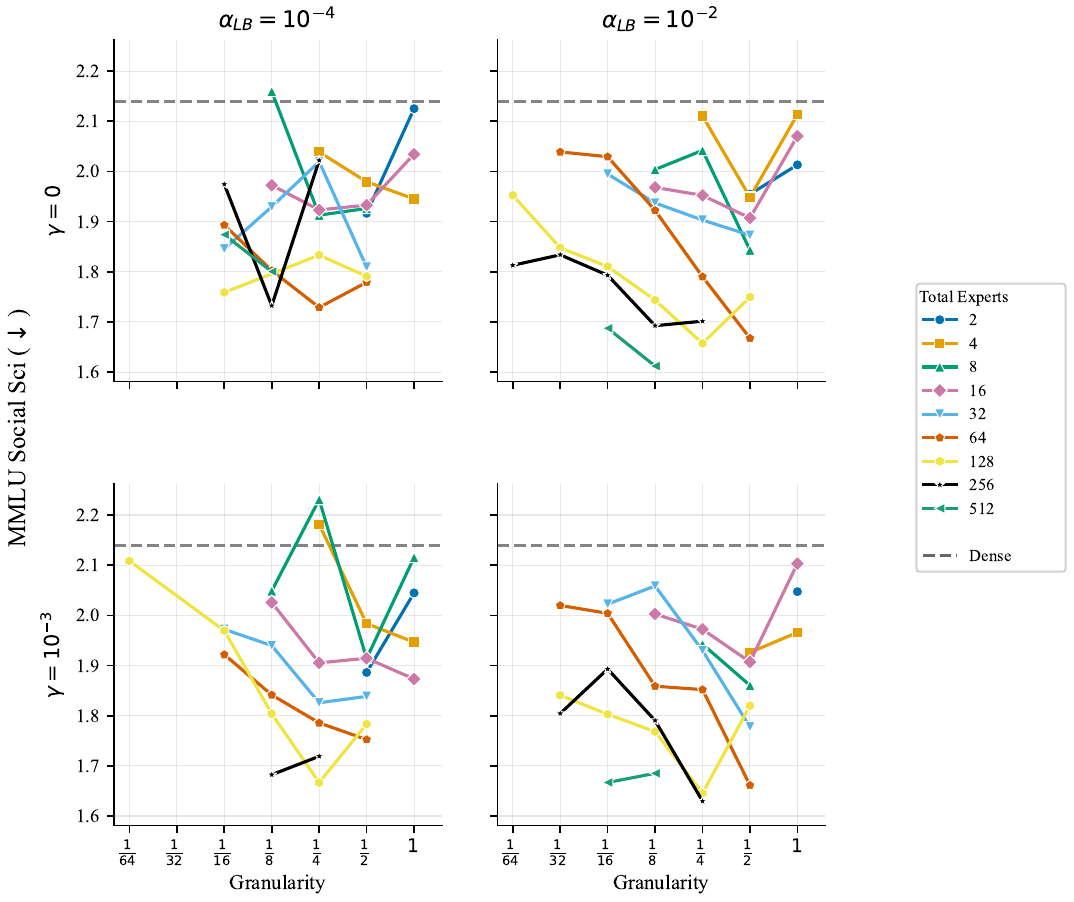}
        \caption{80M active, 80M - 765M total parameters}
    \end{subfigure}
    \par\bigskip\bigskip
    \begin{subfigure}[t]{\textwidth}
        \centering
        \includegraphics[width=0.46\linewidth]{figures/downstream/mmlu_social_sciences/ce_loss/lb_sweep_hgn_gxs_110M.pdf}
        \hspace{1em}
        \includegraphics[width=0.46\linewidth]{figures/downstream/mmlu_social_sciences/ce_loss/lb_sweep_hgn_gxn_110M.pdf}
        \caption{110M active, 110M - 1.4B total parameters}
    \end{subfigure}

    \end{figure*} 

\clearpage  

\begin{figure*}[ht]
    \addtocounter{figure}{-1}
    \centering
    \begin{subfigure}[t]{\textwidth}
        \centering
        \includegraphics[width=0.46\linewidth]{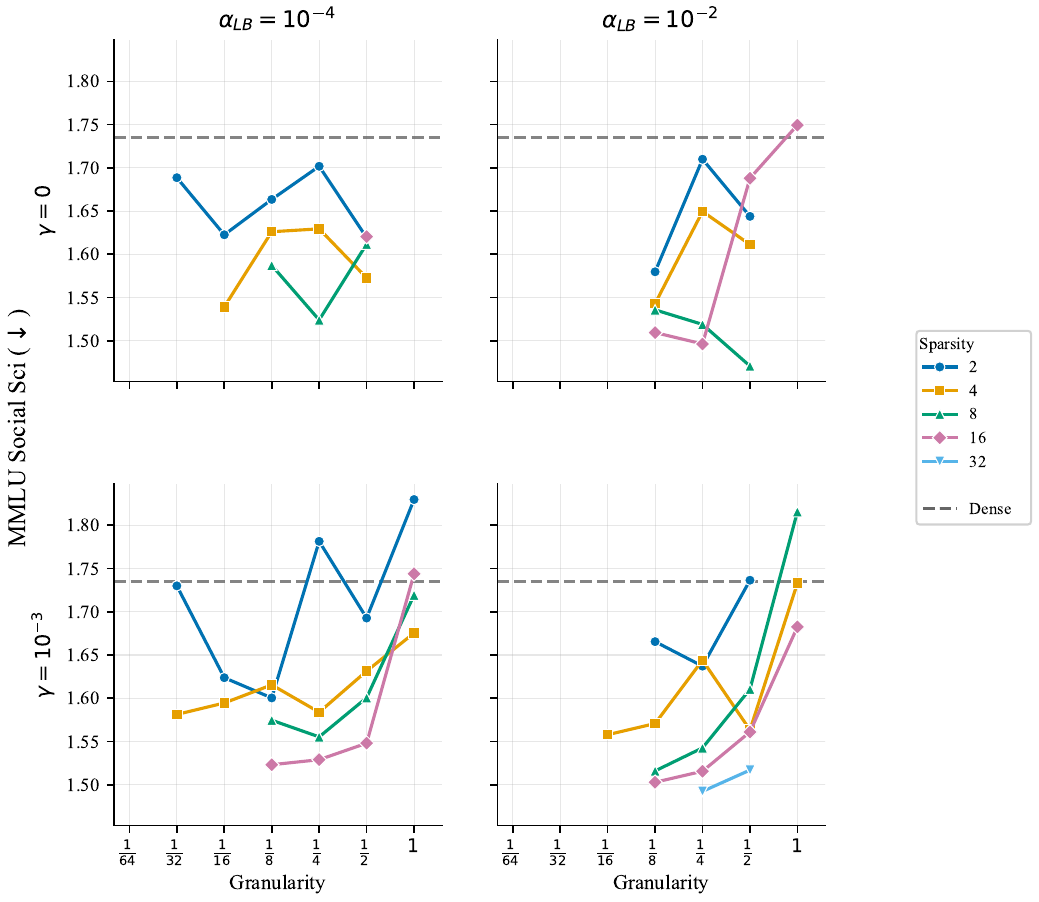}
        \hspace{1em}
        \includegraphics[width=0.46\linewidth]{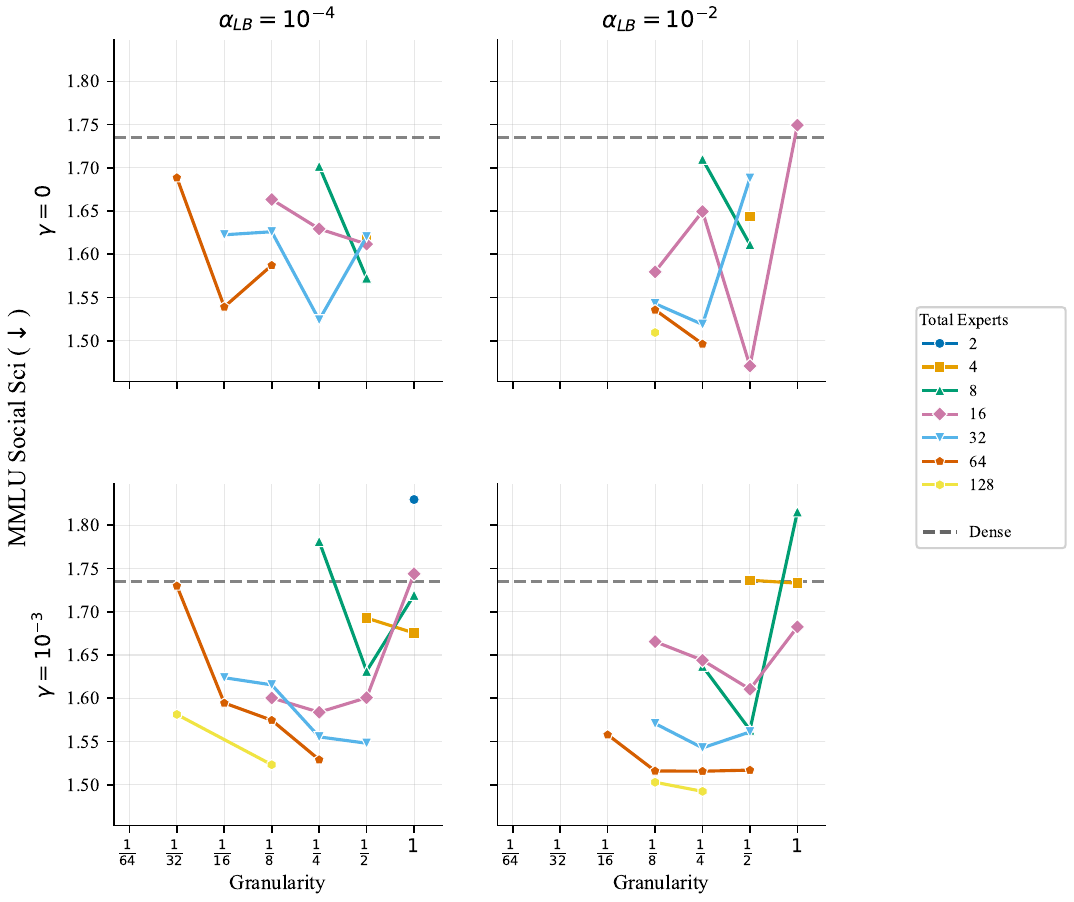}
        \caption{200M active, 200M - 3.3B total parameters}
    \end{subfigure}
    \par\bigskip\bigskip
    \begin{subfigure}[t]{\textwidth}
        \centering
        \includegraphics[width=0.3\linewidth]{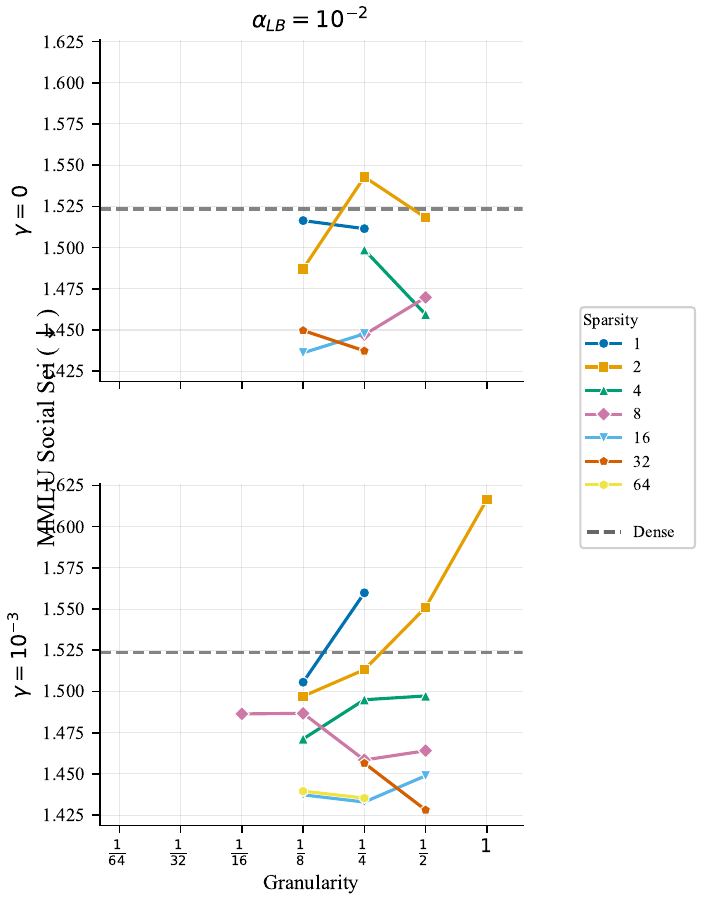}
        \hspace{1em}
        \includegraphics[width=0.3\linewidth]{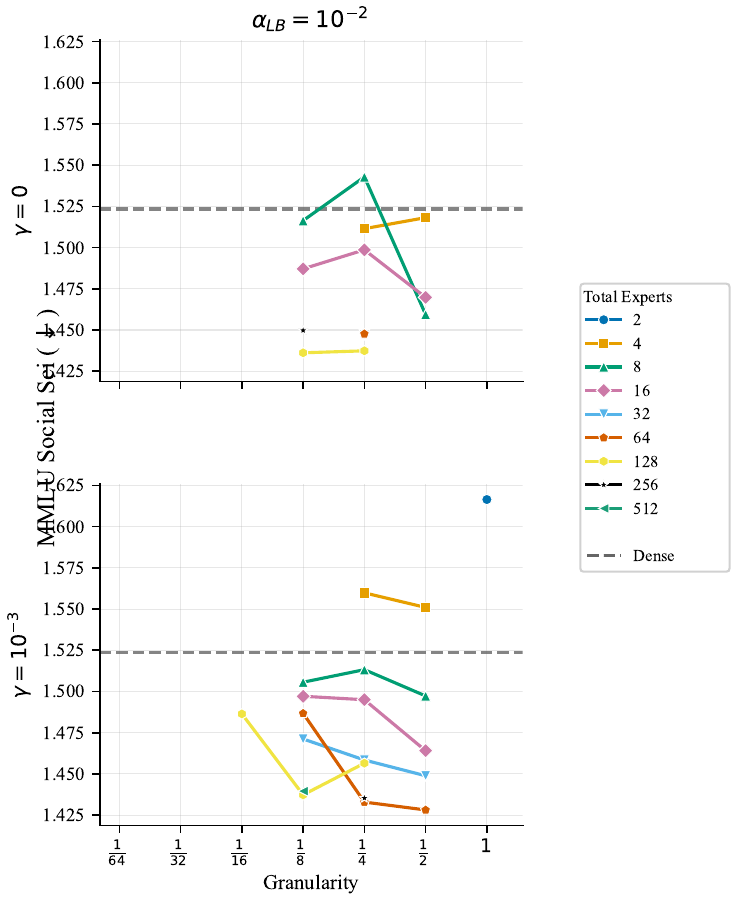}
        \caption{300M active, 300M - 6.6B total parameters}
    \end{subfigure}

    \caption{
    \textbf{Load balancing mechanisms must be tuned correctly (\S\ref{sec:expt_router}).}
    We consider load balancing loss weight $\alpha_{LB} \in \{\num{1e-2}, \num{1e-4}\}$ and loss-free load balancing with bias $\gamma\in\{0, \num{1e-3}\}$ ($\gamma=0$ indicates no loss-free mechanism). Results show that poorly chosen hyperparameters, such as high bias $\gamma = 1e-3$ with total experts $n\geq 512$, may impair performance. However, all settings other than $(\alpha_{LB}=\num{1e-2}, \gamma=\num{1e-3})$ perform comparably for $n \leq 512$, suggesting that a wide range of load balancing settings achieve near-optimal performance. 
    }
    \label{fig:mmlu_social_sciences_lb}
\end{figure*}

%% file: fig_tex/downstream/mmlu_stem.tex
\begin{figure*}[!ht]
    \centering
        \begin{subfigure}[t]{\textwidth}
        \begin{subfigure}[t]{0.33\textwidth}
            \centering
            \caption*{\scriptsize Fixed total experts (n)}
            \includegraphics[width=\linewidth]{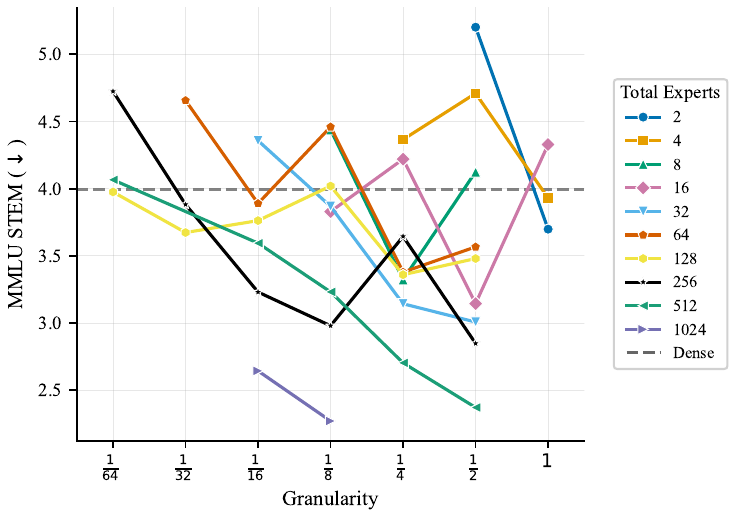}
        \end{subfigure}
        \begin{subfigure}[t]{0.33\textwidth}
            \centering
            \caption*{\scriptsize Fixed granularity (g)}
            \includegraphics[width=\linewidth]{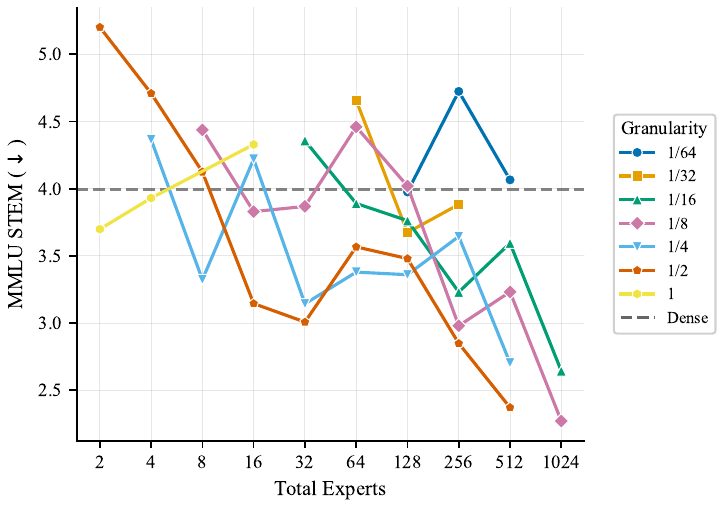}
        \end{subfigure}
        \begin{subfigure}[t]{0.33\textwidth}
            \centering
            \caption*{\scriptsize Fixed activation sparsity (s)}
            \includegraphics[width=\linewidth]{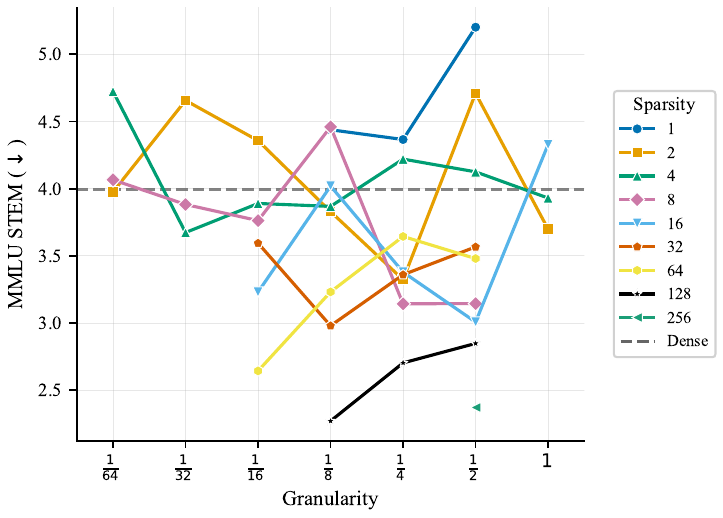}
        \end{subfigure}
        \caption{50M active, 50M - 930M total parameters}
    \end{subfigure}
\par\bigskip\bigskip
    \begin{subfigure}[t]{\textwidth}
        \begin{subfigure}[t]{0.33\textwidth}
            \centering
            \includegraphics[width=\linewidth]{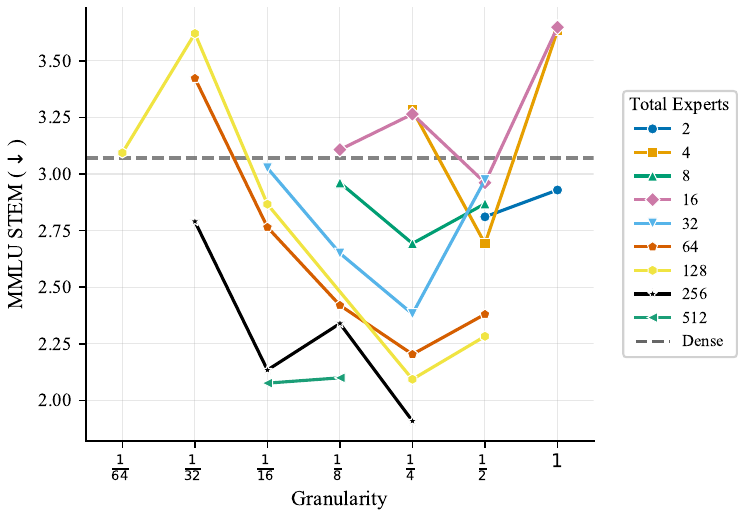}
        \end{subfigure}
        \begin{subfigure}[t]{0.33\textwidth}
            \centering
            \includegraphics[width=\linewidth]{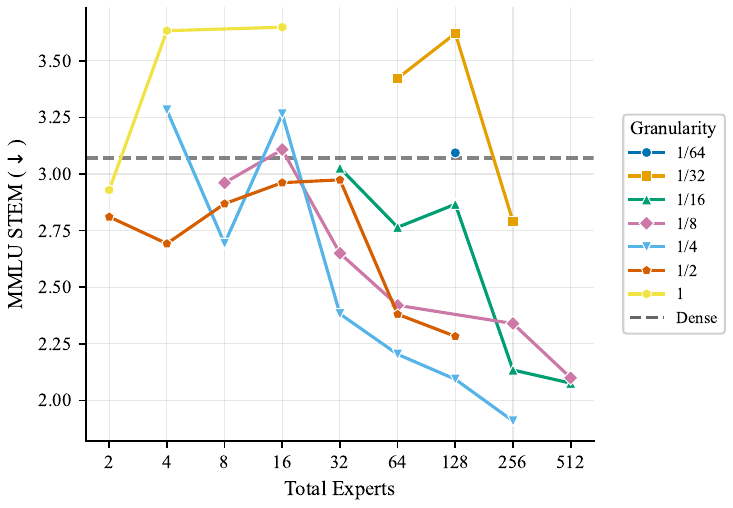}
        \end{subfigure}
        \begin{subfigure}[t]{0.33\textwidth}
            \centering
            \includegraphics[width=\linewidth]{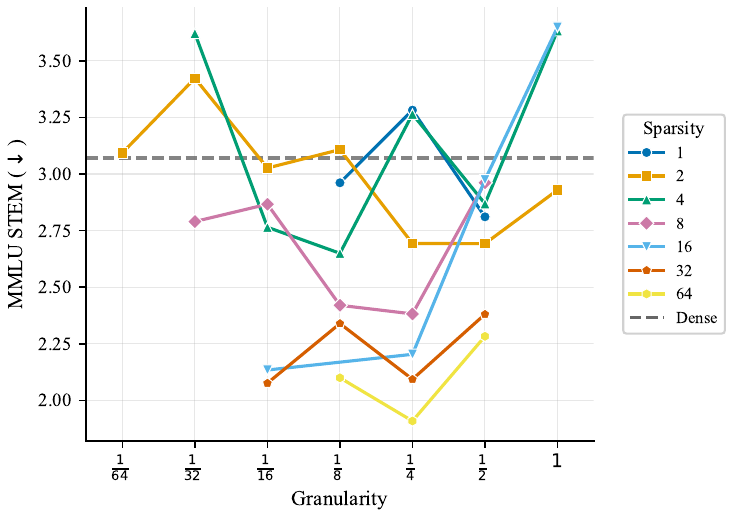}
        \end{subfigure}
        \caption{80M active, 80M - 765M total parameters}
    \end{subfigure}
    \par\bigskip\bigskip
        \begin{subfigure}[t]{\textwidth}
        \begin{subfigure}[t]{0.33\textwidth}
            \centering
            \includegraphics[width=\linewidth]{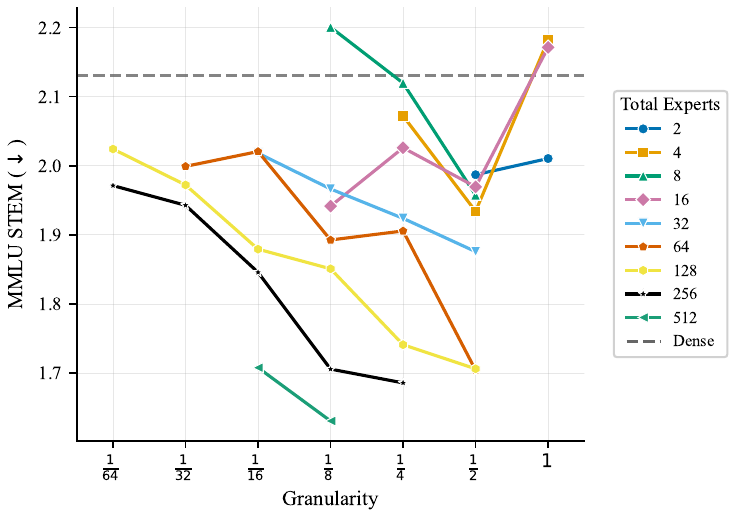}
        \end{subfigure}
        \begin{subfigure}[t]{0.33\textwidth}
            \centering
            \includegraphics[width=\linewidth]{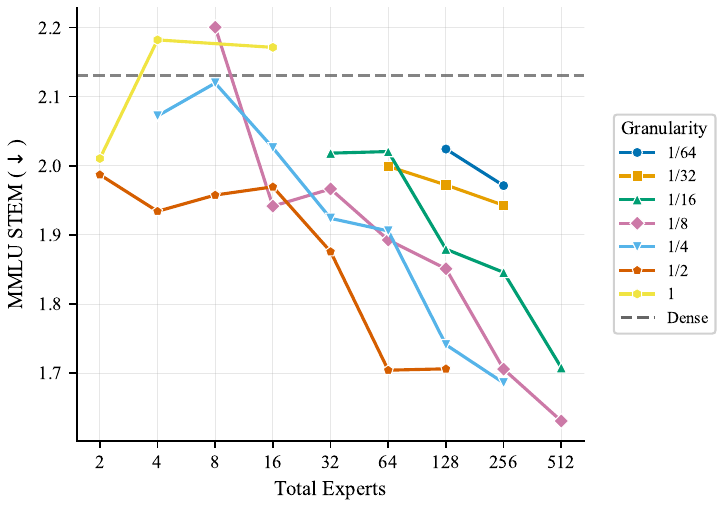}
        \end{subfigure}
        \begin{subfigure}[t]{0.33\textwidth}
            \centering
            \includegraphics[width=\linewidth]{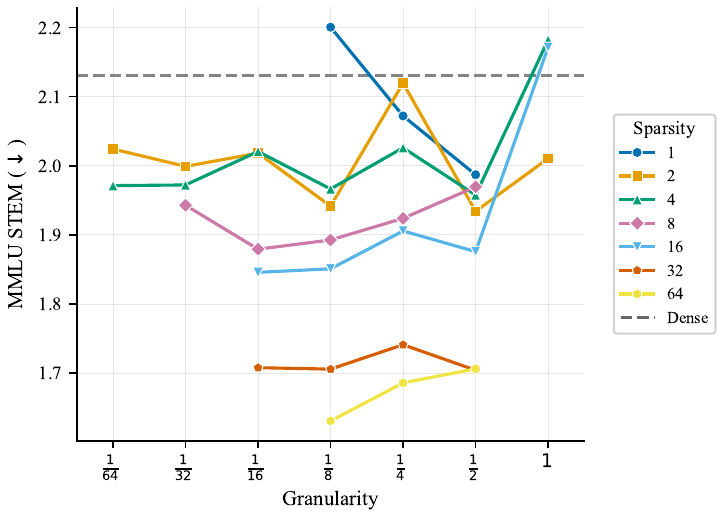}
        \end{subfigure}
        \caption{110M active, 110M - 1.4B total parameters}
    \end{subfigure}
    \end{figure*}

\clearpage  

\begin{figure*}[!ht]
        \addtocounter{figure}{-1}
    \begin{subfigure}[t]{\textwidth}
        \addtocounter{subfigure}{3}
        \begin{subfigure}[t]{0.33\textwidth}
            \centering
            \caption*{\scriptsize Fixed total experts (n)}
            \includegraphics[width=\linewidth]{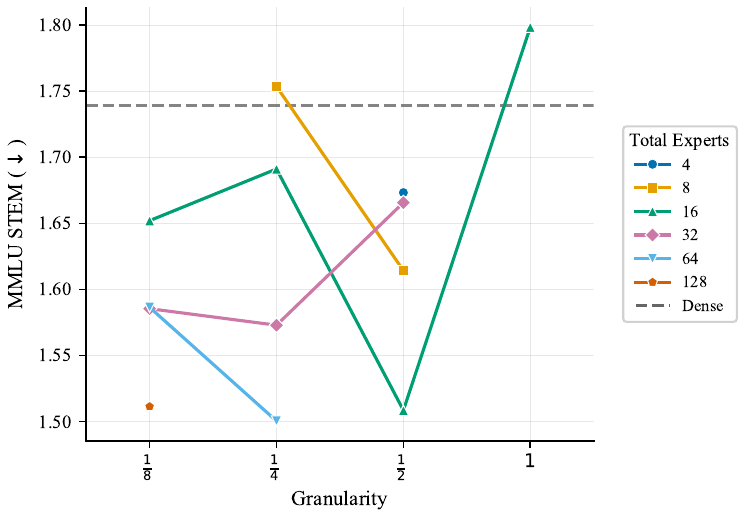}
        \end{subfigure}
        \begin{subfigure}[t]{0.33\textwidth}
            \centering
            \caption*{\scriptsize Fixed granularity (g)}
            \includegraphics[width=\linewidth]{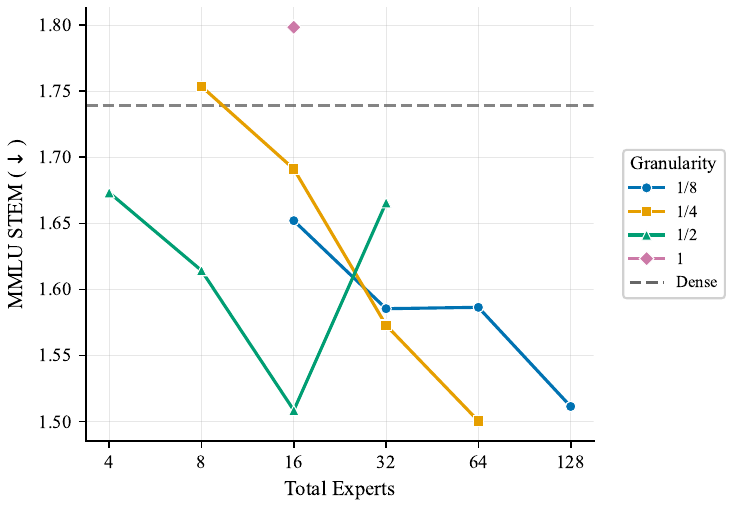}
        \end{subfigure}
        \begin{subfigure}[t]{0.33\textwidth}
            \centering
            \caption*{\scriptsize Fixed activation sparsity (s)}
            \includegraphics[width=\linewidth]{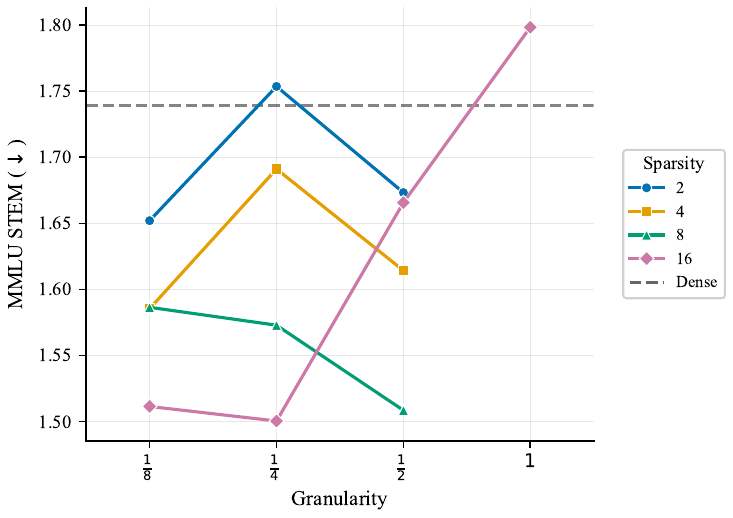}
        \end{subfigure}
        \caption{200M active, 200M - 3.3B total parameters}
    \end{subfigure}
    \par\bigskip\bigskip
        \begin{subfigure}[t]{\textwidth}
        \begin{subfigure}[t]{0.33\textwidth}
            \centering
            \includegraphics[width=\linewidth]{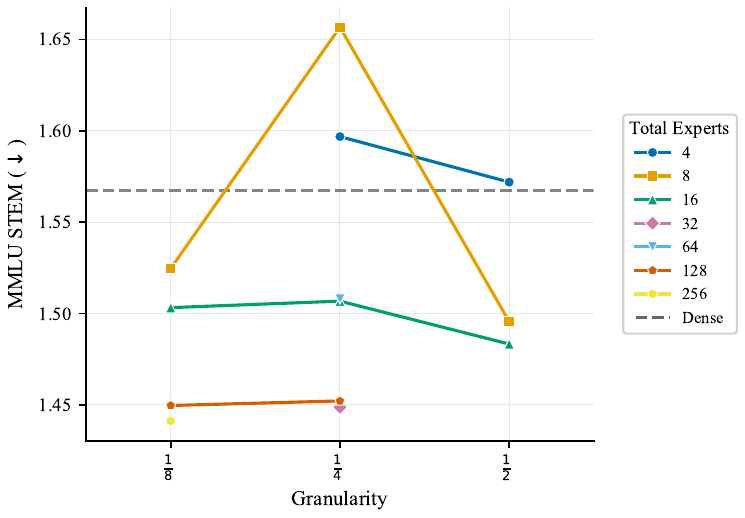}
        \end{subfigure}
        \begin{subfigure}[t]{0.33\textwidth}
            \centering
            \includegraphics[width=\linewidth]{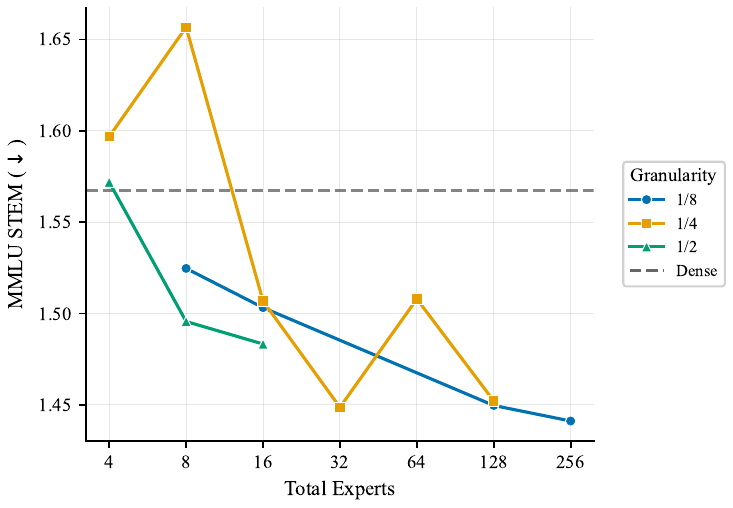}
        \end{subfigure}
        \begin{subfigure}[t]{0.33\textwidth}
            \centering
            \includegraphics[width=\linewidth]{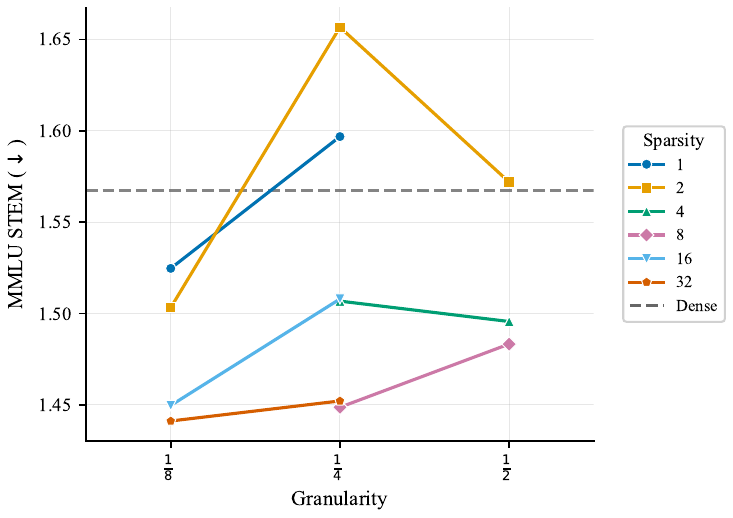}
        \end{subfigure}
        \caption{300M active, 300M - 6.6B total parameters}
    \end{subfigure}

    \caption{
    \textbf{Increasing inactive expert parameters via expert size (left) or total count (center) improves performance in MoEs (\S\ref{sec:expt_main}).} This effect is seen both when holding total number of experts fixed (left) and when holding expert granularity fixed (center). In general, increasing total parameters results in improved performance.  \textbf{Optimal tradeoff between expert count and granularity varies in MoEs (right). (\S\ref{sec:expt_main})}
    At each activation sparsity $s$ (equivalently, at each total parameter count), the optimal (total expert count, expert granularity) configuration varies. As $s$ increases, optimal expert granularity remains nearly fixed, suggesting that sparsity should be scaled up primarily by increasing total expert count $n$, while maintaining a near constant, slowly increasing expert granularity $g$. 
    }
    \label{fig:mmlu_stem_experts}
\end{figure*}

\begin{figure*}[!ht]
    \centering
    
    \begin{subfigure}[t]{0.46\textwidth}
        \centering
        \includegraphics[width=\linewidth]{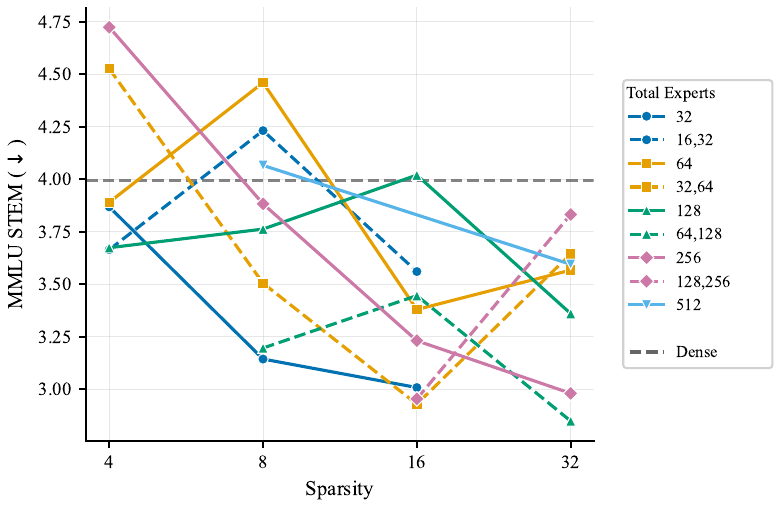}
        \caption{50M active, 50M - 930M total parameters}
    \end{subfigure}
    \vspace{1em}
    \begin{subfigure}[t]{0.46\textwidth}
        \centering
        \includegraphics[width=\linewidth]{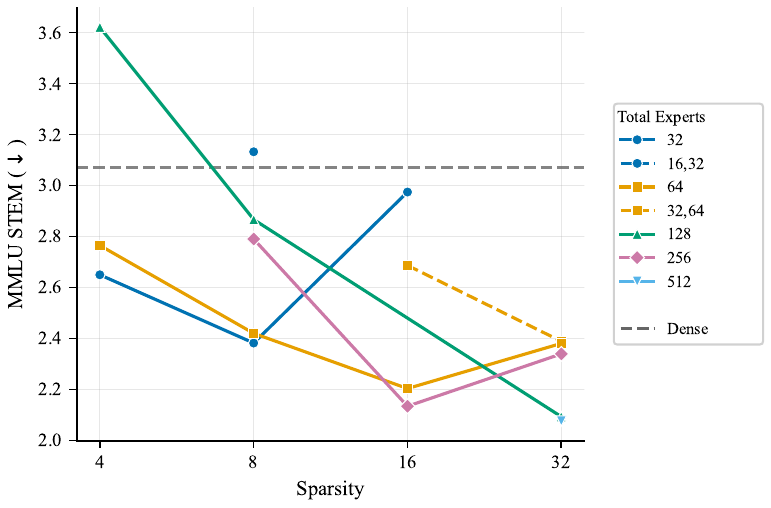}
        \caption{80M active, 80M - 765M total parameters}
    \end{subfigure}
    \caption{
    \textbf{Heterogeneity of expert size alone does not improve MoE performance (\S\ref{sec:expt_hetgen}).} To explore the potential benefits of their architectural flexibility, we compare heterogeneous MoEs (indicated by dotted lines) to active- and total-parameter-matched homogeneous MoEs. Heterogeneity alone does not result in performance gains, as, at each activation sparsity $s$, heterogeneous MoEs with $n_1, n_2 = a, b$ lie between or near the 2 closest homogeneous MoEs, with $n=a$ and with $n=b$.
    }
    \label{fig:mmlu_stem_het}
\end{figure*}

\begin{figure*}[!ht]
    \centering
    
    \begin{subfigure}[t]{1.0\textwidth}
        \centering
        \includegraphics[width=\linewidth]{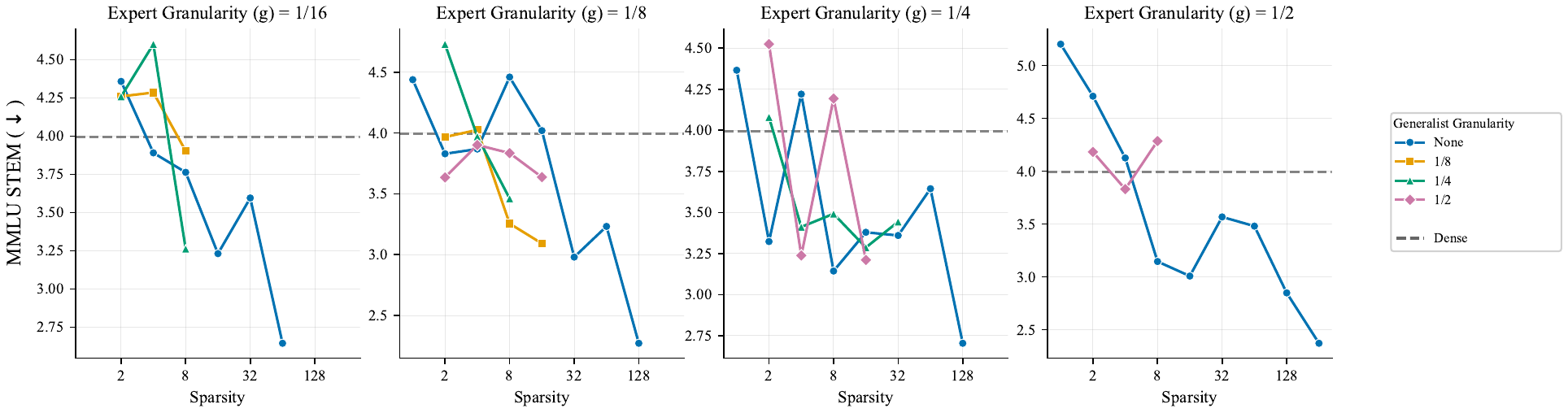}
        \caption{50M active, 50M - 930M total parameters}
    \end{subfigure}
    \par\bigskip\bigskip
    \begin{subfigure}[t]{1.0\textwidth}
        \centering
        \includegraphics[width=\linewidth]{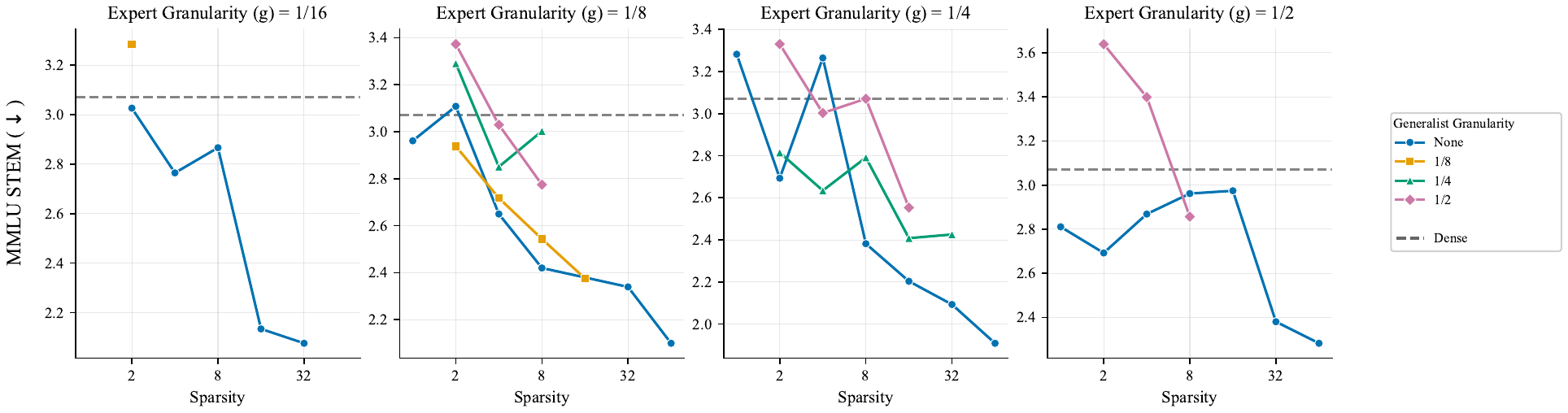}
        \caption{80M active, 80M - 765M total parameters}
    \end{subfigure}
    \par\bigskip\bigskip
    \begin{subfigure}[t]{1.0\textwidth}
        \centering
        \includegraphics[width=\linewidth]{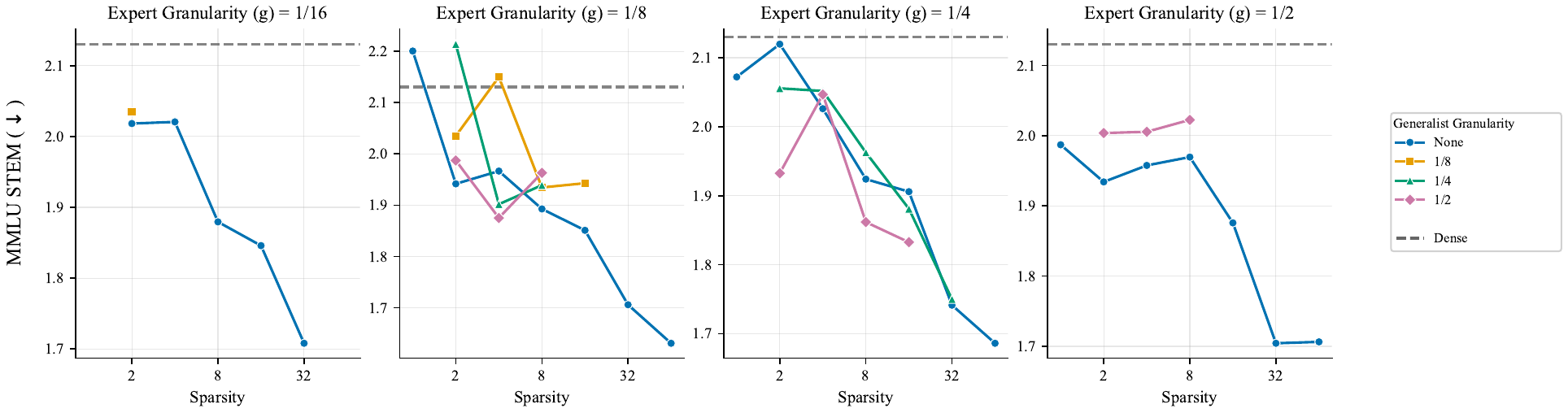}
        \caption{110M active, 110M - 1.4B total parameters}
    \end{subfigure}
    \caption{
    \textbf{The inclusion of a generalist consistently degrades performance in homogeneous MoEs (\S\ref{sec:expt_hetgen}).}
    We train MoE LMs which consist of some routed experts with granularity $g$, as well as a generalist with granularity $g_{gen}\in \{\frac{1}{2}, \frac{1}{4}, \frac{1}{8}\} $. We compare to settings with no generalist, only routed experts with granularity $g$. In all settings and configurations, the addition of any granularity generalist results in comparable or degraded performance. 
    }
    \label{fig:mmlu_stem_gen}
\end{figure*}

\begin{figure*}[ht]
    \centering
    \begin{subfigure}[t]{1.0\textwidth}
        \centering
        \includegraphics[width=\linewidth]{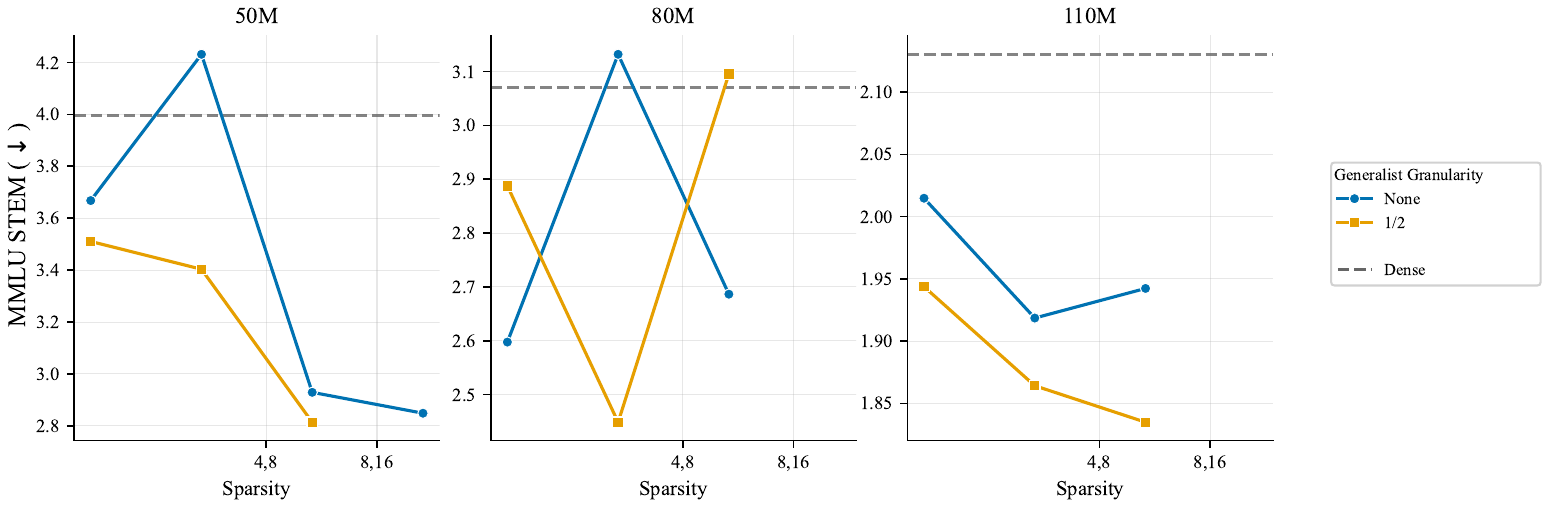}
    \end{subfigure}
    \caption{
    \textbf{The inclusion of a generalist consistently degrades performance in heterogeneous MoEs (\S\ref{sec:expt_hetgen}).}
    We train heterogeneous MoE LMs which consist of  routed experts with granularity $g_1, g_2$, as well as a generalist with granularity $g_{gen} = \frac{1}{2}$. We compare to settings with no generalist. In all settings and configurations, the addition of a generalist results in comparable or degraded performance. 
    }
    \label{fig:mmlu_stem_hetgen}
\end{figure*}

\begin{figure*}[ht]
    \centering
    \begin{subfigure}[t]{\textwidth}
        \centering
        \begin{subfigure}[t]{0.45\textwidth}
            \includegraphics[width=\linewidth]{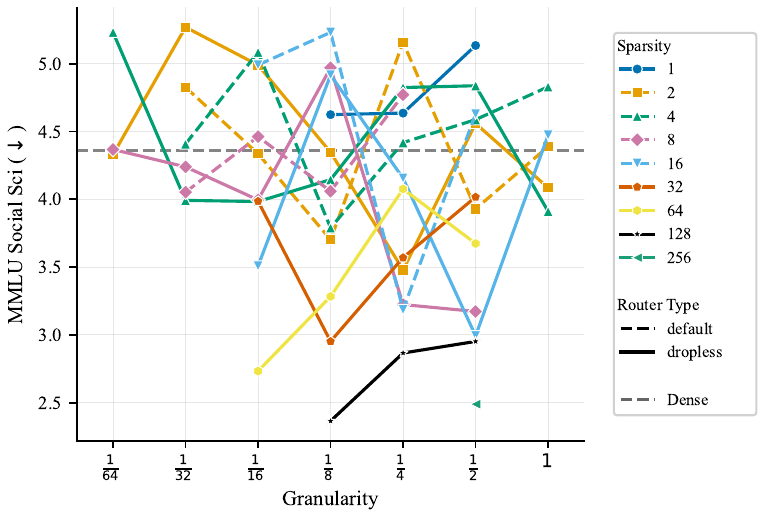}
            \caption{50M active, 50M - 930M total parameters}
        \end{subfigure}
    \hspace{1em}
        \begin{subfigure}[t]{0.45\textwidth}
            \centering
            \includegraphics[width=\linewidth]{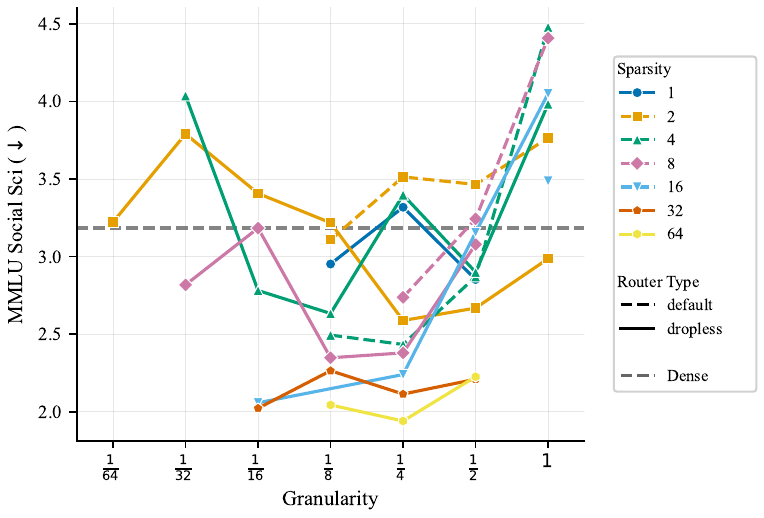}
            \caption{80M active, 80M - 765M total parameters}
        \end{subfigure}
    \end{subfigure}

    \par\bigskip\bigskip
    \begin{subfigure}[t]{0.45\textwidth}
        \centering
        \includegraphics[width=\linewidth]{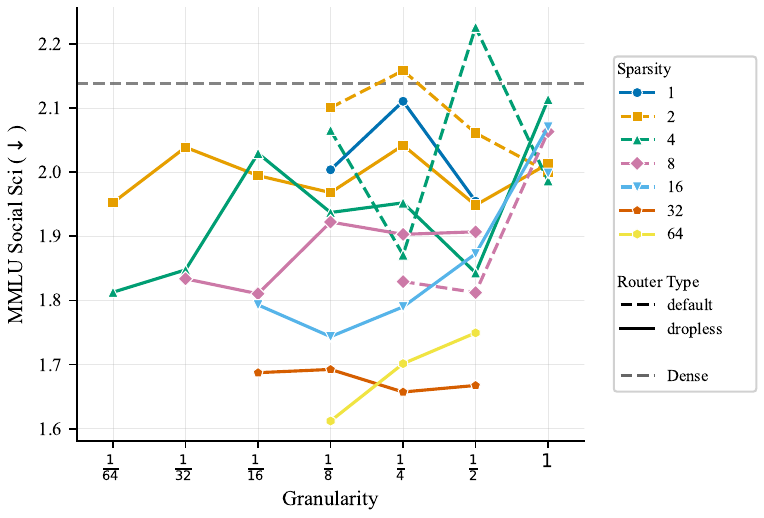}
        \caption{110M active, 110M - 1.4B total parameters}
    \end{subfigure}
    \caption{ 
    \textbf{Dropless routing outperforms default routing (\S\ref{sec:expt_router}).}
    We compare dropless routing to the default setting, which allow tokens to be dropped. Across all scales, we find that dropless routing outperforms or performs comparably to default routing. 
    }
    \label{fig:mmlu_stem_dropless}
\end{figure*}

\begin{figure*}[ht]
    \centering
    \begin{subfigure}[t]{0.45\textwidth}
        \centering
        \includegraphics[width=\linewidth]{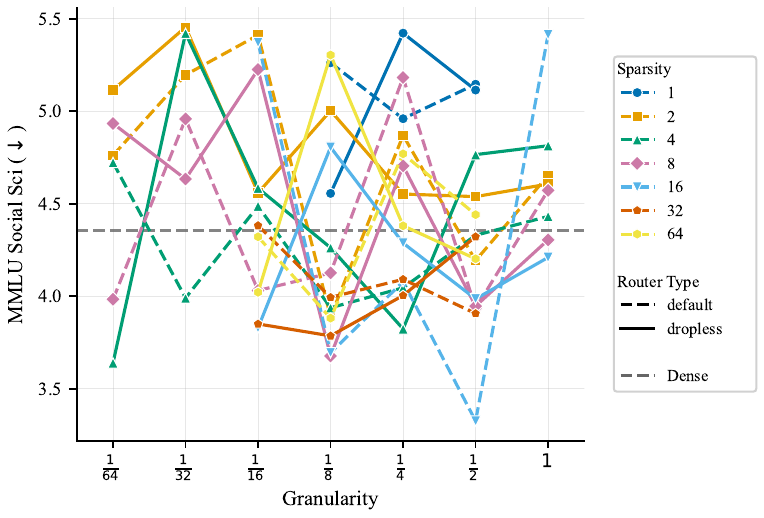}
        \caption{50M active, 50M - 930M total parameters}
    \end{subfigure}
    \hspace{1em}
    \begin{subfigure}[t]{0.45\textwidth}
        \centering
        \includegraphics[width=\linewidth]{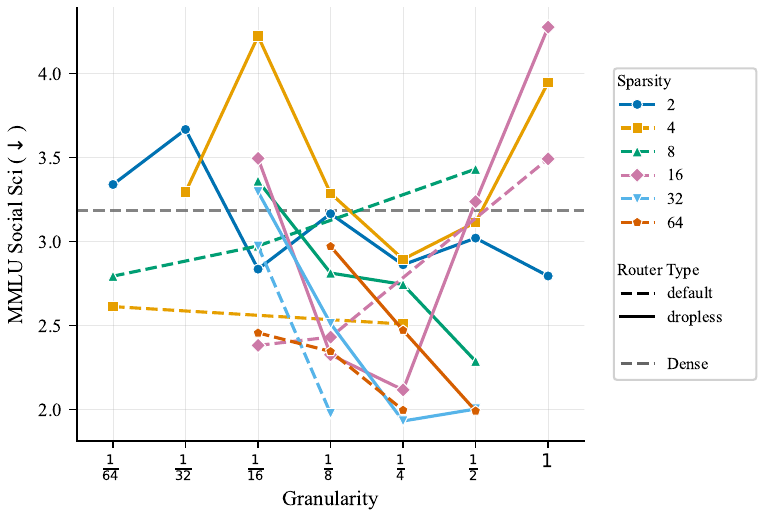}
        \caption{80M active, 80M - 765M total parameters}
    \end{subfigure}
    \caption{
    \textbf{Dropless routing, with bias $\gamma=\num{1e-3}$ (\S\ref{sec:expt_router}).} 
    As in Figure~\ref{fig:lm_avg_dropless}, we compare dropless routing to the default setting, which allow tokens to be dropped. Across all scales, we find that dropless routing outperforms or performs comparably to default routing. We see here with additional higher sparsity default routing runs that as sparsity increases, default routing performance approaches that of dropless routing.
    }
    \label{fig:mmlu_stem_dropless_with_lf}
\end{figure*}

\begin{figure*}[ht]
    \centering
    \begin{subfigure}[]{\textwidth}
        \centering
        \includegraphics[width=0.46\linewidth]{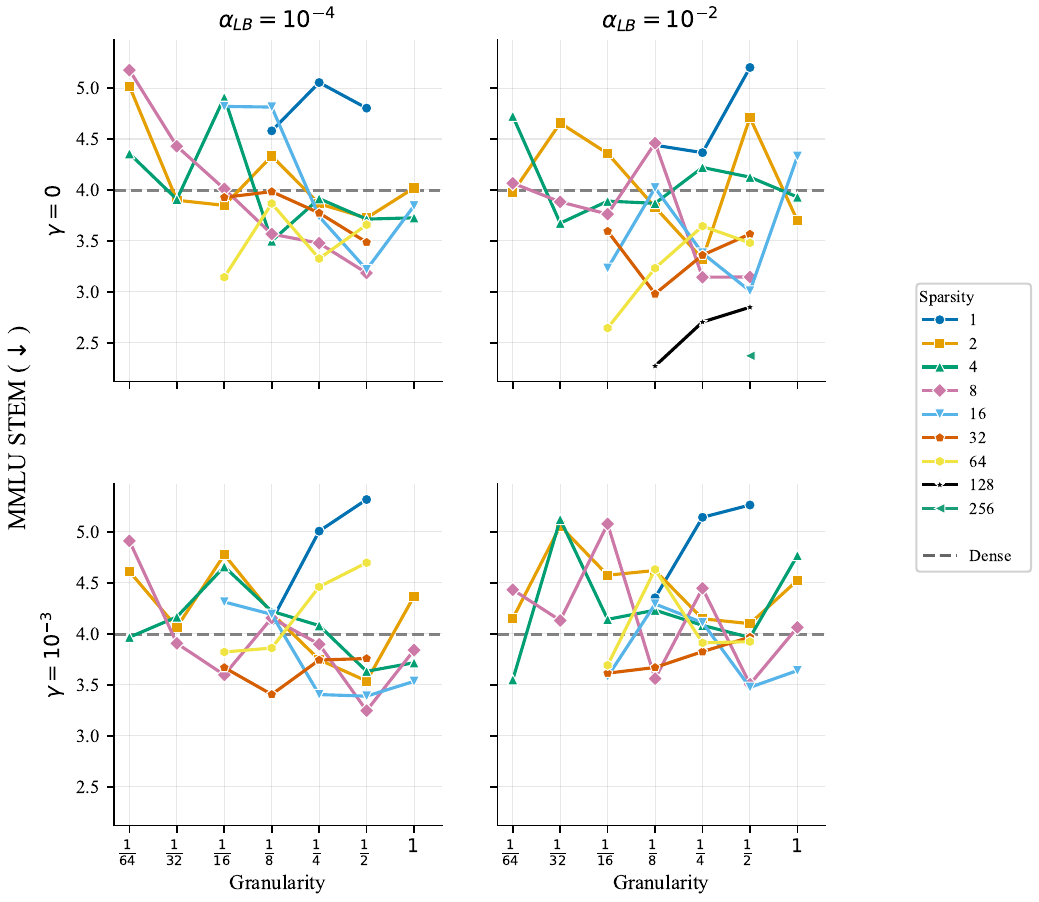}
        \hspace{1em}
        \includegraphics[width=0.46\linewidth]{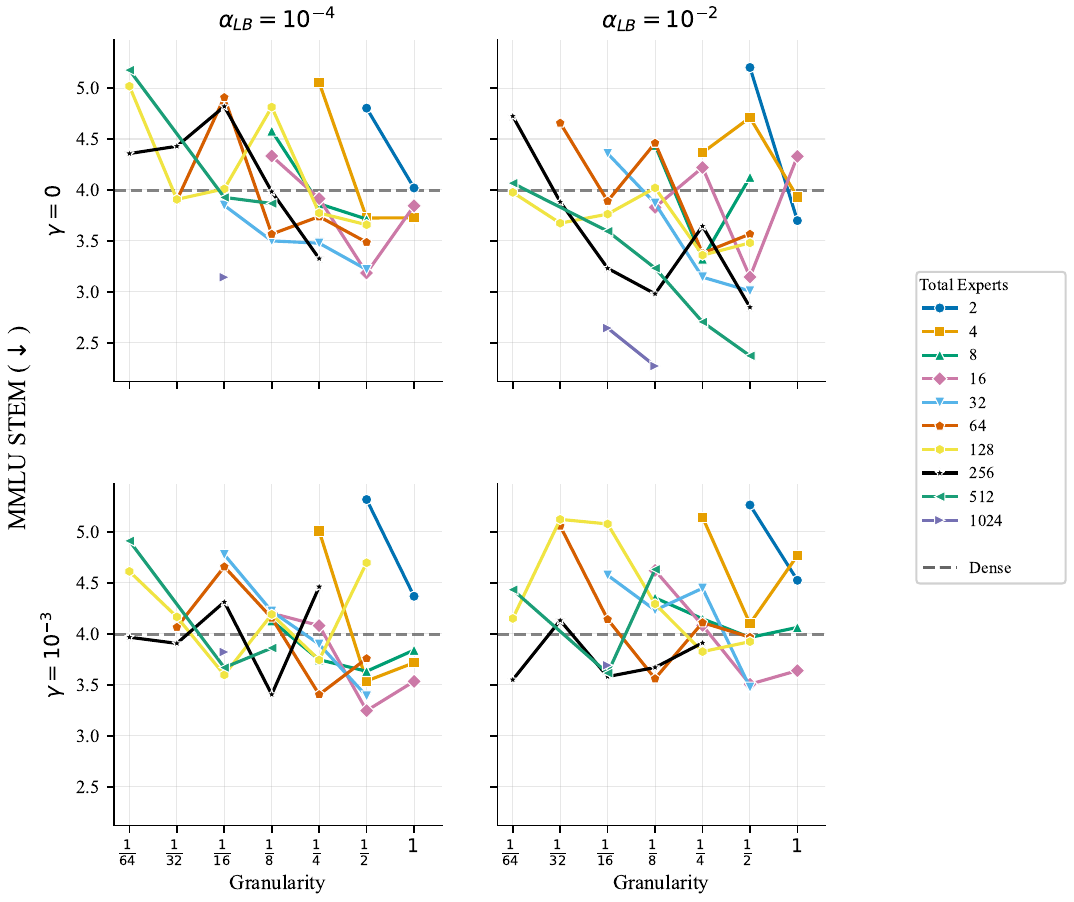}
        \caption{50M active, 50M - 930M total parameters}
    \end{subfigure}
    \par\bigskip\bigskip
    \begin{subfigure}[]{\textwidth}
        \centering
        \includegraphics[width=0.46\linewidth]{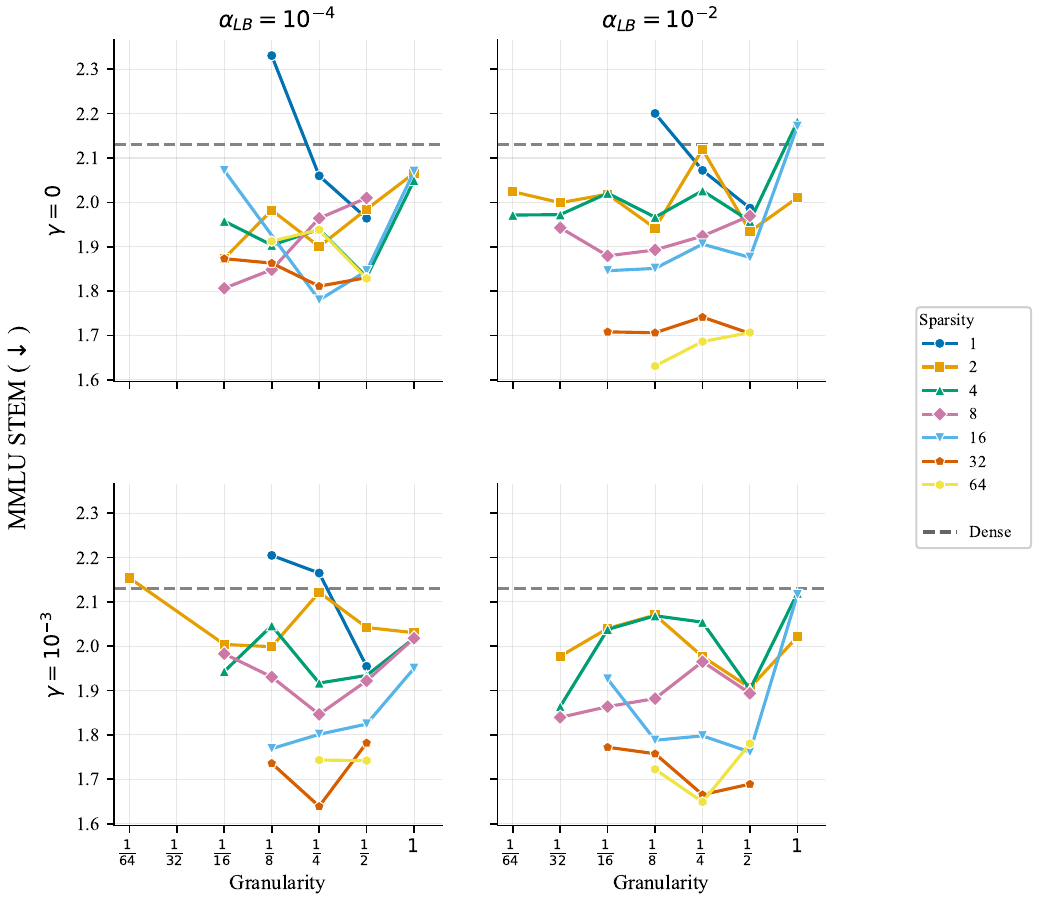}
        \hspace{1em}
        \includegraphics[width=0.46\linewidth]{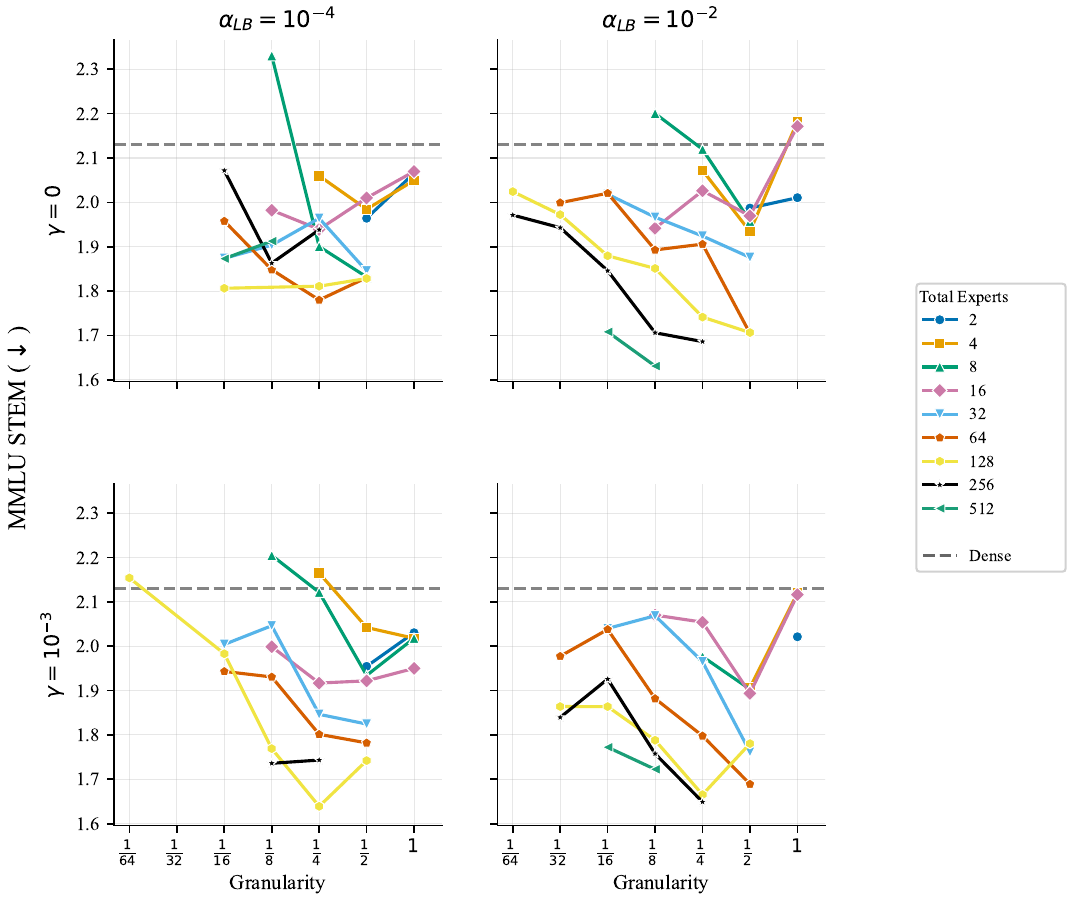}
        \caption{80M active, 80M - 765M total parameters}
    \end{subfigure}
    \par\bigskip\bigskip
    \begin{subfigure}[t]{\textwidth}
        \centering
        \includegraphics[width=0.46\linewidth]{figures/downstream/mmlu_stem/ce_loss/lb_sweep_hgn_gxs_110M.pdf}
        \hspace{1em}
        \includegraphics[width=0.46\linewidth]{figures/downstream/mmlu_stem/ce_loss/lb_sweep_hgn_gxn_110M.pdf}
        \caption{110M active, 110M - 1.4B total parameters}
    \end{subfigure}

    \end{figure*} 

\clearpage  

\begin{figure*}[ht]
    \addtocounter{figure}{-1}
    \centering
    \begin{subfigure}[t]{\textwidth}
        \centering
        \includegraphics[width=0.46\linewidth]{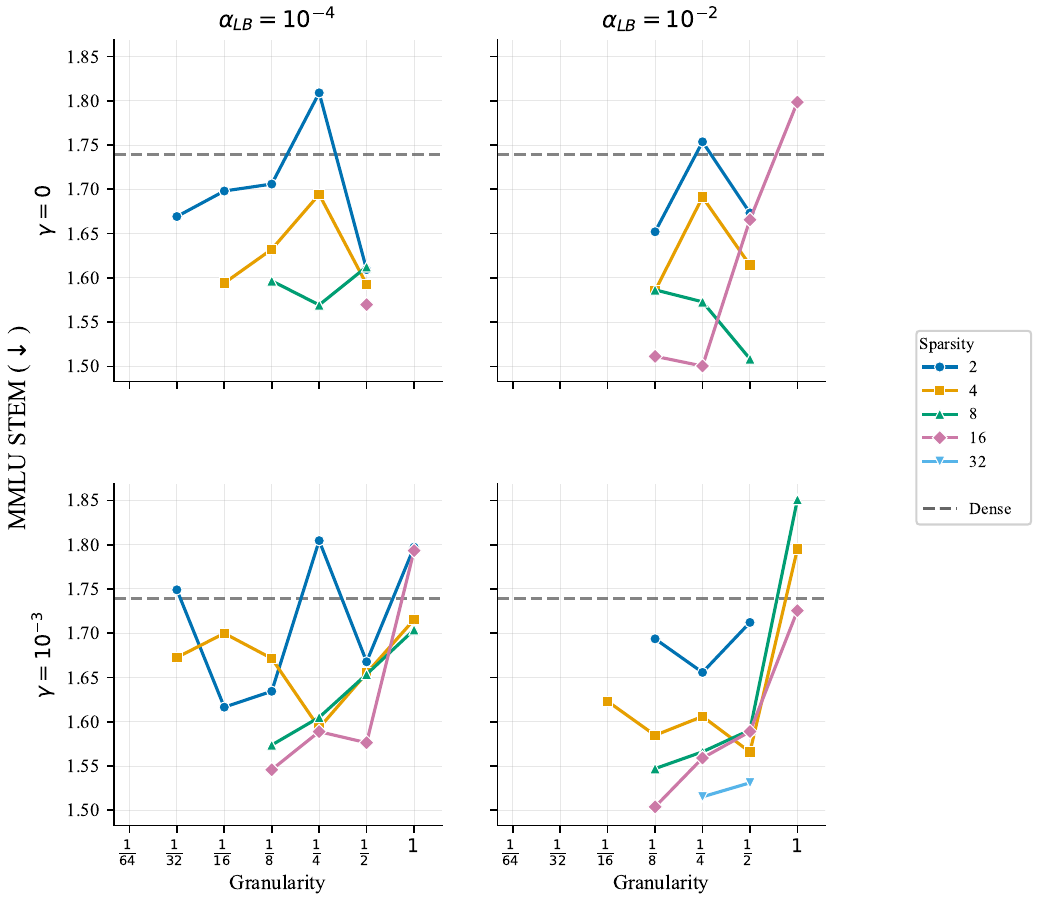}
        \hspace{1em}
        \includegraphics[width=0.46\linewidth]{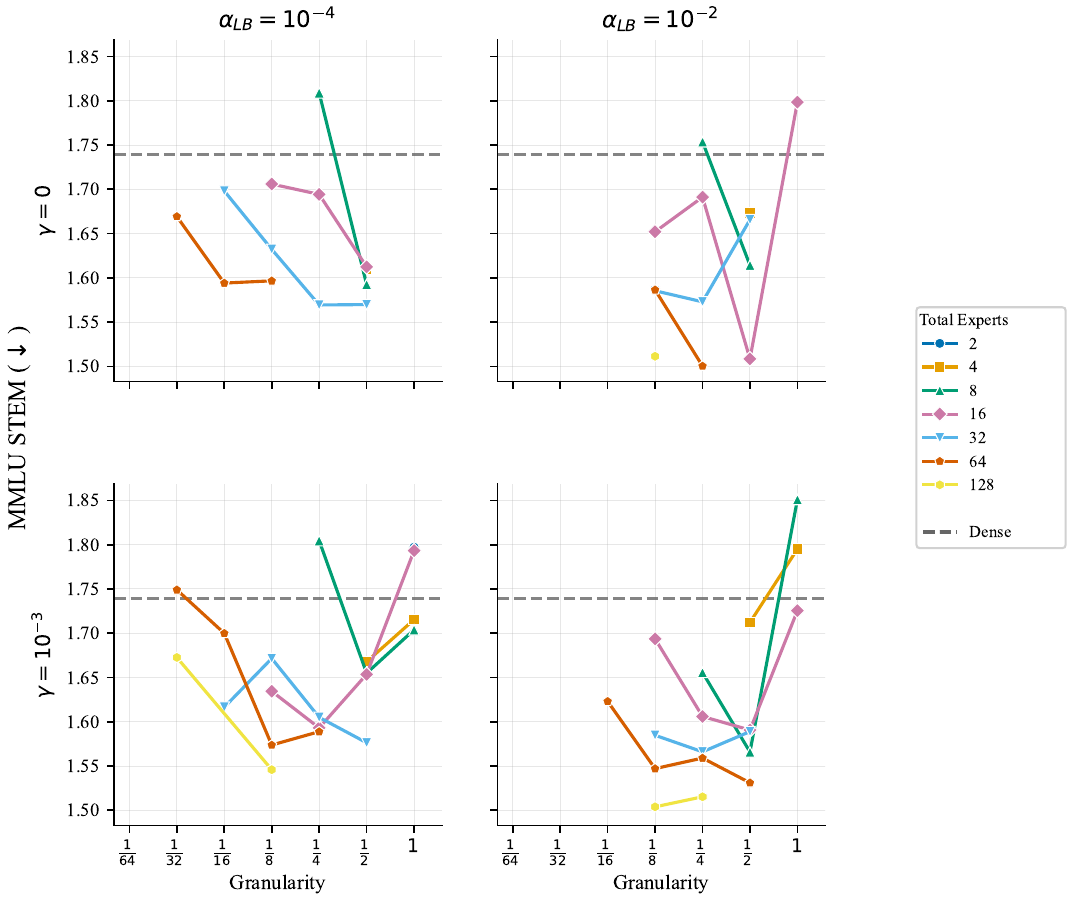}
        \caption{200M active, 200M - 3.3B total parameters}
    \end{subfigure}
    \par\bigskip\bigskip
    \begin{subfigure}[t]{\textwidth}
        \centering
        \includegraphics[width=0.3\linewidth]{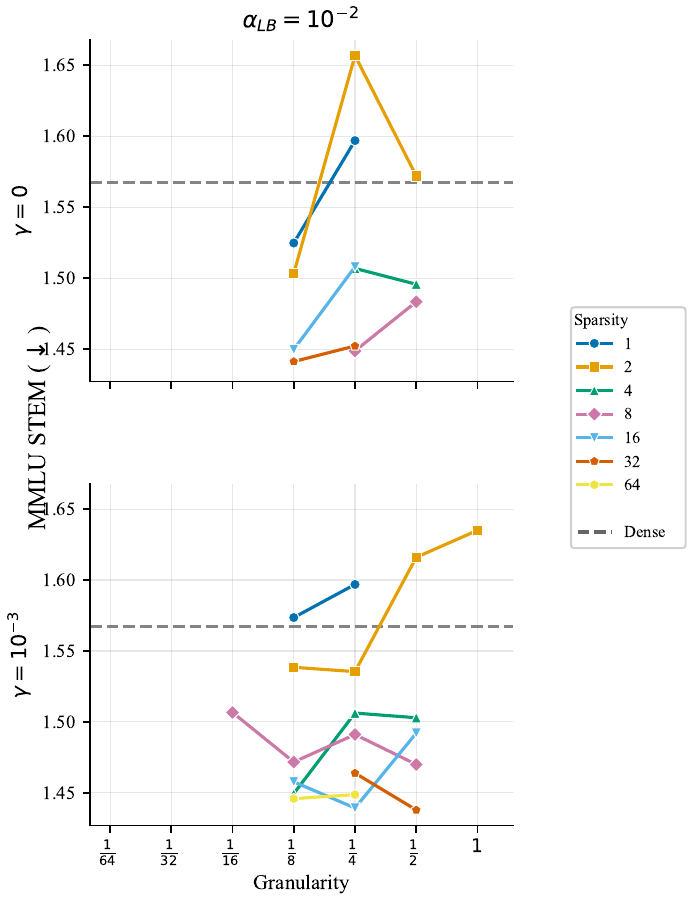}
        \hspace{1em}
        \includegraphics[width=0.3\linewidth]{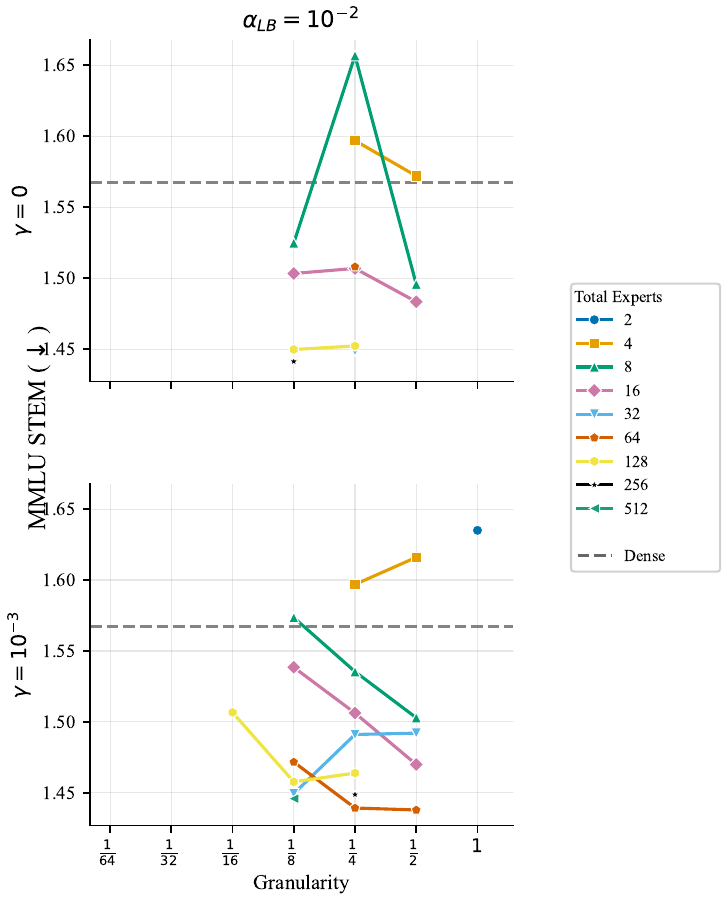}
        \caption{300M active, 300M - 6.6B total parameters}
    \end{subfigure}

    \caption{
    \textbf{Load balancing mechanisms must be tuned correctly (\S\ref{sec:expt_router}).}
    We consider load balancing loss weight $\alpha_{LB} \in \{\num{1e-2}, \num{1e-4}\}$ and loss-free load balancing with bias $\gamma\in\{0, \num{1e-3}\}$ ($\gamma=0$ indicates no loss-free mechanism). Results show that poorly chosen hyperparameters, such as high bias $\gamma = 1e-3$ with total experts $n\geq 512$, may impair performance. However, all settings other than $(\alpha_{LB}=\num{1e-2}, \gamma=\num{1e-3})$ perform comparably for $n \leq 512$, suggesting that a wide range of load balancing settings achieve near-optimal performance. 
    }
    \label{fig:mmlu_stem_lb}
\end{figure*}